\documentclass[11pt, a4paper, logo]{googledeepmind}

\usepackage[authoryear, sort&compress, round]{natbib}
\usepackage{soul}
\usepackage{fancybox}
\usepackage{booktabs}
\usepackage{makecell}
\usepackage{appendix}
\usepackage{hyperref}
\usepackage{tcolorbox}
\usepackage[draft]{fixme}
\usepackage{chngcntr}
\usepackage{longtable}

\usepackage{amsfonts}
\usepackage{titlesec}
\usepackage{amsmath}
\usepackage{graphicx}
\usepackage{listings}
\usepackage{xcolor}

\usepackage{paralist}
\usepackage{authblk}
\usepackage{pifont}
\usepackage{ifthen}
\usepackage{float}
\usepackage{subcaption}
\usepackage{booktabs}

\usepackage{tikz}

\usepackage{xparse}
\hypersetup{
    allcolors=black
}

\usetikzlibrary{decorations.pathreplacing, calc, positioning, patterns, arrows.meta}

\newcommand{\Turn}{\text{\ae}} 
\newcommand{\TurnSet}{\text{\ensuremath{\calA \kern-0.25em\calE}}}
\NewDocumentCommand{\Hist}{O{t}}{\text{\ae}_{1:{#1}}}
\newcommand{\HistA}{\text{\ae}_*}
\NewDocumentCommand{\HistM}{O{t}}{{\text{\ae}_{<#1}}}
\NewDocumentCommand{\HistMI}{O{t}}{{\text{\ae}^i_{<#1}}}

\newcommand{\MATurn}{\overline{\text{\ae}}}
\newcommand{\MATurnSet}{\overline{\text{\ensuremath{\calA \kern-0.25em\calE}}}}
\newcommand{\MAHist}{\overline{\text{\ae}}_{1:t}}
\newcommand{\MAHistA}{\overline{\text{\ae}}_*}
\newcommand{\MAHistI}{\text{\ae}^i_{1:t}}
\NewDocumentCommand{\HistI}{O{t}}{\text{\ae}^i_{1:{#1}}}

\NewDocumentCommand{\MAHistM}{O{t}}{\overline{\text{\ae}}_{<#1}}

\newcommand{\MultiAgentMve}{\overline{\Mve}}
\newcommand{\MultiAgentMge}{\overline{\Mge}}
\newcommand{\MultiAgentMvp}{\overline{\Mvp}}

\newcommand{\MultiAgentMuPiI}{(\Mve^i)^{\Mvp^i}}
\newcommand{\MultiAgentGeGpI}{(\Mge^i)^{\Mgp^i}}
\newcommand{\MAgeTailM}[1][\MAHistM]{\bar{\Mge}_{#1}}
\newcommand{\MgpTailM}[1][\text{\ae}^i_{<t}]{\Mgp_{#1}}
\newcommand{\MmeTailM}[1][\text{\ae}^i_{<t}]{\Mme_{#1}}
\newcommand{\MmuTailM}[1][\text{\ae}^i_{<t}]{\Mmu_{#1}}
\newcommand{\MgeTailM}[1][\text{\ae}^i_{<t}]{\Mge_{#1}}

\newcommand{\Rgh}{\bara_{1:t}}  
\newcommand{\RghA}{\bara_*}  
\NewDocumentCommand{\RghM}{O{t}}{\bara_{<#1}}

\newcommand{\Mge}{\mu}
\newcommand{\Mgu}{\upsilon}
\newcommand{\Mgp}{\pi}
\newcommand{\Mgop}[1][]{%
  \pi^*%
  \ifx\\#1\\%
  \else
    _{#1}%
  \fi
}
\newcommand{\Mve}{\nu}
\newcommand{\Mvu}{\lambda}
\newcommand{\Mvp}{\pi}
\newcommand{\Mme}{\xi}
\newcommand{\Mmu}{\rho}
\newcommand{\Mmp}{\zeta}

\newcommand{\calMapom}{\calM^{\mathrm{APOM}}}
\newcommand{\calMrapom}{\calM^{\mathrm{rAPOM}}}

\newcommand{\MwRO}
{M^w_{\mathrm{RO}}}

\author[$\star$,1]{Alexander Meulemans}
\author[$\star$,1]{Rajai Nasser}
\author[1]{Maciej Wołczyk}
\author[1]{Marissa A. Weis}
\author[1]{Seijin Kobayashi}
\author[1,2,4,5]{Blake Richards}
\author[1,2,3,5]{Guillaume Lajoie}
\author[1,6]{Angelika Steger}
\author[7]{Marcus Hutter}
\author[1,8]{James Manyika}
\author[$\dagger$,1]{Rif A. Saurous}
\author[$\dagger$,1]{João Sacramento}
\author[$\dagger$,1,9]{Blaise Agüera y Arcas}

\affil[1]{Google, Paradigms of Intelligence Team}
\affil[2]{Mila - Quebec AI Institute}
\affil[3]{Université de Montréal}
\affil[4]{McGill University}
\affil[5]{CIFAR}
\affil[6]{ETH Zürich}
\affil[7]{Google DeepMind}
\affil[8]{Google}
\affil[9]{Santa Fe Institute}
\affil[$\star$]{Equal contribution}
\affil[$\dagger$]{Equal supervision}
\usepackage{thm-restate, amsmath, amssymb, amsfonts, mathtools}

\usepackage[ruled,vlined]{algorithm2e}
\usepackage{tikz}
\usepackage{xcolor} % for pink color

\usepackage{comment}

\theoremstyle{plain}
\newtheorem{theorem}{Theorem}[section]
\newtheorem{lemma}[theorem]{Lemma}
\newtheorem{proposition}[theorem]{Proposition}
\newtheorem{corollary}[theorem]{Corollary}
\theoremstyle{definition}
\newtheorem{definition}[theorem]{Definition}
\newtheorem{problem}[theorem]{Problem}
\newtheorem{example}[theorem]{Example}
\newtheorem{remark}[theorem]{Remark}

\DeclareMathSymbol{\smin}{\mathbin}{AMSa}{"39}

\newcommand{\calA}{\mathcal{A}}
\newcommand{\calB}{\mathcal{B}}

\newcommand{\calE}{\mathcal{E}}

\newcommand{\calH}{\mathcal{H}}
\newcommand{\calI}{\mathcal{I}}

\newcommand{\calM}{\mathcal{M}}

\newcommand{\calO}{\mathcal{O}}

\newcommand{\calR}{\mathcal{R}}
\newcommand{\calS}{\mathcal{S}}

\newcommand{\calX}{\mathcal{X}}

\newcommand{\bbE}{\mathbb{E}}

\newcommand{\bbN}{\mathbb{N}}

\newcommand{\bbP}{\mathbb{P}}
\newcommand{\bbQ}{\mathbb{Q}}
\newcommand{\bbR}{\mathbb{R}}

\newcommand{\expect}[2]{\mathbb{E}_{#1}\left[ #2 \right]}

\newcommand{\kl}[2]{\mathrm{KL}\left(#1 \mid \mid #2\right)}

\newcommand*{\query}{\mathrm{query}}
\newcommand*{\eval}{\mathrm{eval}}

\newcommand*{\bara}{\bar{a}}
\newcommand*{\bare}{\bar{e}}

\newcommand*{\barcalA}{\bar{\mathcal{A}}}
\newcommand*{\barcalE}{\bar{\mathcal{E}}}

\newcommand*{\calMpol}{\mathcal{M}_{\mathrm{pol}}}
\newcommand*{\calMenv}{\mathcal{M}_{\mathrm{env}}}
\newcommand*{\calMuni}{\mathcal{M}_{\mathrm{uni}}}

\newcommand*{\calMpolenv}{\mathcal{M}_{\mathrm{pol}\textrm{-}\mathrm{env}}}
\newcommand*{\calMpolenvcheck}{\check{\mathcal{M}}_{\mathrm{pol}\textrm{-}\mathrm{env}}}
\newcommand*{\calMpolcheck}{\check{\mathcal{M}}_{\mathrm{pol}}}
\newcommand*{\calMenvcheck}{\check{\mathcal{M}}_{\mathrm{env}}}

\newcommand{\KSubU}{K_U}
\newcommand{\wUtau}{w^{U,\tau}}
\newcommand{\checkwUtau}{\check{w}^{U,\tau}}
\newcommand{\checkS}{\check{S}}
\def\eqref#1{equation~\ref{#1}}
\def\1{\bm{1}}

\def\eps{{\epsilon}}

\DeclareMathAlphabet{\mathsfit}{\encodingdefault}{\sfdefault}{m}{sl}
\SetMathAlphabet{\mathsfit}{bold}{\encodingdefault}{\sfdefault}{bx}{n}

\DeclareMathOperator*{\argmax}{arg\,max}

\usepackage{booktabs}
\usepackage{wrapfig}
\usepackage{enumitem}
\usepackage{comment}

\usepackage[font=small,labelfont=bf]{caption}

\usepackage[capitalise]{cleveref}

\crefname{boxalias}{Box}{Boxes} % Define the name for the 'boxalias' type

\NewEnviron{BoxFigure}[1][Heading]{%
  \refstepcounter{figure}% 
  \crefalias{figure}{boxalias}%
  \begin{tcolorbox}[
    title= \textbf{Box \thefigure: #1}, % Display the already-stepped figure counter
    colback=black!5,
    colframe=black!75,
    arc=3mm,
    boxrule=1pt,
    fonttitle=\bfseries
  ]
  \BODY
  \end{tcolorbox}%
  \crefalias{figure}{figure}%
}

\numberwithin{figure}{section}

\title{Embedded Universal Predictive Intelligence: a coherent framework for multi-agent learning}

\begin{abstract}

The standard theory of model-free reinforcement learning assumes that the environment dynamics are stationary and that agents are decoupled from their environment, such that policies are treated as being separate from the world they inhabit. This leads to theoretical challenges in the multi-agent setting where the non-stationarity induced by the learning of other agents demands prospective learning based on prediction models. To accurately model other agents, an agent must account for
the fact that those other agents are, in turn, forming beliefs about it to predict its future behavior, motivating agents to model themselves as part of the environment. Here, building upon foundational work on universal artificial intelligence (AIXI), we introduce a mathematical framework for prospective learning and embedded agency centered on self-prediction, where Bayesian RL agents predict both future perceptual inputs and their own actions, and must therefore resolve epistemic uncertainty about themselves as part of the universe they inhabit. We show that in multi-agent settings, self-prediction enables agents to reason about others running similar algorithms, leading to new game-theoretic solution concepts and novel forms of cooperation unattainable by classical decoupled agents. Moreover, we extend the theory of AIXI, and study universally intelligent embedded agents which start from a Solomonoff prior. We show that these idealized agents can form consistent mutual predictions and achieve infinite-order theory of mind, potentially setting a gold standard for embedded multi-agent learning.

\end{abstract}

\crefname{appendix}{Appendix}{Appendices}
\AddToHook{cmd/appendix/before}{
    \crefalias{section}{appendix}
    \crefalias{subsection}{appendix}%
    \setcounter{tocdepth}{1} % Only show appendix sections (level 1) in ToC
}

\begin{document}
\maketitle

\begingroup
    \makeatletter
    
    % Set ToC depth to 1 (only sections)
    \setcounter{tocdepth}{1}
    
    % Use \small for a more compact appearance
    \small
    
    % Reduce line spacing for compactness
    \linespread{0.5}\selectfont
    
    % Redefine \tableofcontents to not start a new page
    % AND reduce space after the "Contents" title
    \renewcommand{\tableofcontents}{%
        \section*{\contentsname}%
        % --- ADD THIS LINE to reduce space after the title ---
        \vspace*{-2ex} % Pulls the content up. Adjust -1.5ex as needed.
        \@starttoc{toc}%
    }
    
    % Print the table of contents
    \tableofcontents
    \makeatother
\endgroup

\section{Introduction}

A major frontier in artificial intelligence involves moving beyond the imitation of human data, leveraging reinforcement learning (RL) to enable agents to self-generate experience and continually improve \citep{silver2017mastering, ouyang2022training, guo2025deepseek, team2023gemini}. This paradigm, combining large-scale pretraining with single-agent RL, has proven remarkably effective for enhancing individual capabilities, leading to significant gains in complex reasoning tasks such as mathematics and programming \citep{ ouyang2022training, guo2025deepseek, uesato2022solving}. However, these individual competencies represent only one facet of intelligence. To endow AI agents with meaningful social capabilities beyond those observed from human behavioral data, training must move into multi-agent environments, as sociality is an inherently multi-agent property \citep{dafoe2021cooperative}. Developing such capabilities is crucial for enabling agents to function as effective and reliable entities within human society, allowing them to navigate complex social dynamics, coordinate with humans and other AI agents, or even represent human stakeholders in negotiations. To date, the most impressive results in multi-agent RL (MARL) have been confined to specialized cases, such as zero-sum, purely competitive games \citep{silver2017mastering, vinyals2019grandmaster} or tasks of pure cooperation. Robust learning in the more general and human-relevant mixed-motive settings, which form the core of sociality \citep{dafoe2021cooperative}, remains an unsolved challenge. These scenarios, which mirror high-stakes societal challenges such as economic and climate change negotiations, demand a sophisticated blend of cooperation and competition, grappling with emergent social phenomena like trust, fairness, reputation, and deception.% This failure stems in large part from the fact that standard model-free RL methods, which are *retrospective* by nature, break down when their core assumption of a stationary environment is violated by the presence of other co-learning agents \citep{foerster_learning_2018}. %Therefore, a crucial and open challenge is to develop robust multi-agent learning frameworks that can train agents to successfully navigate these complex social situations and find stable cooperative strategies where they are mutually beneficial.

Current model-free RL approaches are insufficient to tackle learning in multi-agent with mixed-motive settings.
%A primary reason for this failure in mixed-motive settings is that the current model-free reinforcement learning (RL) paradigm is insufficient to tackle learning in multi-agent systems. 
The first fundamental 
problem with standard model-free RL is its retrospective nature: An agent attempts to improve its policy to perform well on \emph{data from the past}. This form of learning has its roots in 19th and 20th century empirical observations on how animals respond to reward and punishment \citep{thorndike1898animal,skinner1948superstition}, and accompanying theories such as Thorndike’s law of effect, which can be loosely understood as ``do more of what worked well in the past, and less of what did not’’ \citep{thorndike1898animal}. Such retrospective learning works well when the environment is stationary, but multi-agent systems are intrinsically nonstationary. From the perspective of an individual agent, the environment changes during the learning process, as it contains other agents that are themselves learning. Applying model-free RL algorithms in a multi-agent setting thus relies on an invalid stationarity assumption, which in practice leads to suboptimal behavior, especially in mixed-motive scenarios and social dilemmas \citep{foerster_learning_2018, huh2023multi,sandholm1996multiagent}. Therefore, multi-agent systems require \textit{prospective learning} instead: Agents should update their policies based on a predicted future rather than an outdated past. Prospective learning fundamentally relies on prediction models to anticipate a changing future. However, the current main paradigm for LLM agents, which combines a pretrained predictive model such as an autoregressive transformer with model-free RL during post-training, typically treats the predictive model as a mere initialization for the policy \citep{team2023gemini, openai2023gpt, guo2025deepseek}. Hence, during post-training, the sequence model is trained with (retrospective) model-free RL, in contrast to using the prediction model to predict a changing future and using that predicted future for prospective learning.

Recent multi-agent RL work on co-player learning awareness \citep{foerster_learning_2018, lu_model-free_2022, meulemans2024multi, aghajohari2024loqa, duque2024advantage, khan_scaling_2024} can be thought of as trying to derive a prospective policy gradient, which estimates the influence of an agent's actions on the learning of other agents. However, in deriving a policy gradient, these algorithms often implicitly assume that other agents are not \textit{themselves} co-player learning aware. This flaw points to the more general challenge of dealing with infinite theory of mind recursions of the form ``I predict your behavior, while taking into account that you are predicting my behavior, which is predicting your behavior, which is....'' Such infinite recursions in theory of mind make \textit{consistent mutual prediction} a central problem in prospective multi-agent learning, and tackling this problem is a main focus of the present paper.

\begin{figure}[t!]
    \centering
    \includegraphics[width=0.7\textwidth]{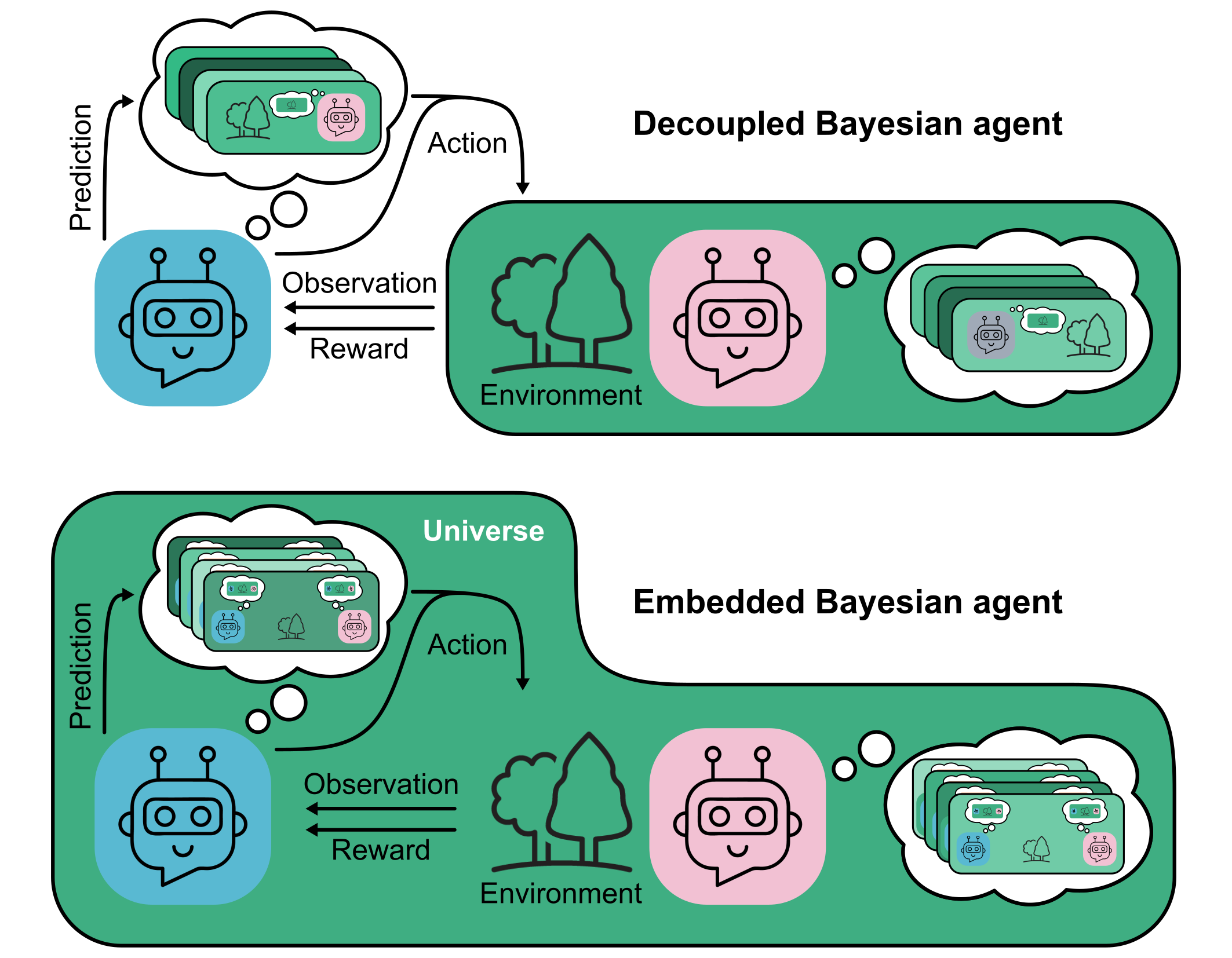}
    \caption{\textbf{Illustration of embedded Bayesian agents.} \textit{Top}: In standard, decoupled approaches to Bayesian agents for RL, an agent considers their own policy as separate from the world they and others are making predictions about, maintaining beliefs about \textit{environments} that might contain other agents, but not the ego-agent itself. \textit{Bottom}: Embedded Bayesian agents, consider themselves to be embedded within a \textit{universe}, the combination of the ego-agents policy with the environment which can contain other agents. As such, when making predictions they are not only predicting the environment percepts but also their own actions, allowing them to leverage functional similarities into their predictions and resulting behavior. This allows embedded Bayesian agents to converge to different equilibrium points in multi-agent settings.}
    \label{fig:mupi_cartoon}
\end{figure}

The challenges of mutual prediction and recursive theory of mind highlight a second problem with the standard model-free RL framework: it is built upon the conceptual paradigm of \textit{decoupled agency}. Decoupled agents do not consider their policies as part of the environment but rather as a distinct, external construct. However, agents are real entities in the environment they inhabit, and to accurately model other agents, an agent should account for the fact that those other agents are, in turn, forming beliefs about \textit{it} to predict \textit{its} future behavior. This motivates a conceptual shift towards embedded agency \citep{demski2019embedded}, where each agent models itself as part of the environment it inhabits, enabling it to anticipate the beliefs others might form about it. As we will show in this work, formally adopting this embedded viewpoint---considering oneself as part of the environment---is not only a conceptual correction but also a practical one, unlocking new avenues towards cooperation and coordination among agents through the consistent mutual prediction of oneself and others.

The present paper provides a mathematical framework that combines prospective learning with embedded agency, where agents base their decisions on predictions of both the environment and the agent's \textit{own} behavior (Fig. \ref{fig:mupi_cartoon}). Consequently, whereas classical model-based RL is centered on predicting how the external environment responds to a certain action, the agents that we consider here predict both future perceptual inputs as well as \emph{their own actions}, a process that we call \emph{self-prediction}. Joint percept-action prediction lifts the barrier between external environment and self, and it is the key step we take towards embeddedness. 

To formalize prospective learning and embededness, we build upon Bayesian sequence prediction and universal artificial intelligence \citep[AIXI]{hutter2005universal}, and adapt this theory---which treats decoupled agency--- towards a theory of embedded agency. The AIXI framework formalizes prospective learning through Bayesian sequence prediction, and our embedded agency framework inherits this capability. We introduce \emph{embedded Bayesian agents}, which maintain and sequentially update probabilistic beliefs as they gather knowledge about which \emph{universe} they live in. In critical opposition to standard RL, universes contain the agent itself, and jointly describe the (possibly stochastic) laws of evolution of the agent---including its learning dynamics---together with its surrounding environment, possibly including other agents and their learning dynamics. Embedded Bayesian agents must thus resolve epistemic uncertainty not only about the environment but also about themselves. When the agent's hypothesis class over possible universes contains the ground-truth universe describing the joint dynamics of the agent and environment, a condition commonly called \emph{a grain of truth} \citep{kalai1993rational}, the resulting Bayesian predictions are guaranteed to converge to the ground-truth distribution. This convergence holds even in the presence of non-stationarities where the future differs from the past, sidestepping the need to make simplifying Markov, i.i.d., ergodicity, or stationarity assumptions often invoked in standard RL \citep{hutter2003optimality, Hutter:24uaibook2, blackwell1962merging}. Embedded Bayesian agents then combine these prospective Bayesian predictions with optimal planning or policy distillation to learn their behavioral policies, resulting in prospective learning based on prospective Bayesian predictions. When the grain-of-truth property holds, embedded Bayesian agents make consistent mutual prediction and hence achieve recursive theory of mind. This makes the satisfaction of the grain-of-truth property a central and non-trivial challenge in designing embedded Bayesian agents and a core focus of this work. Our manuscript is split into two main parts. In Sections \ref{sec:embedded-bayesian-agents} and \ref{sec:subj-emb-eq}, we describe embedded Bayesian agents and their resulting behavior, assuming they satisfy the grain-of-truth property. Then, in \cref{sec:mupi}, we explicitly design embedded Bayesian agents that satisfy the grain-of-truth property while maintaining a wide hypothesis class that includes all computable universes.

This embedded, joint-modeling approach leads to a crucial departure from the standard, decoupled paradigm: The agent's beliefs about itself and the environment become \textit{coupled}. Information about the agent's own actions can now provide evidence about the environment (including other agents), and vice-versa. This coupling is not a mere theoretical artifact but a principled reflection of \textit{functional similarities} among agents that arise naturally in the world. Such similarities occur for example through the following pathways: (i) agents can have a \textit{shared creation process}, such as multiple AIs being instantiations of the same base model or organisms sharing genes; or (ii) agents can develop \textit{convergent solutions} to a similar task, independently arriving at analogous strategies. Embedded Bayesian agents are uniquely positioned to leverage this insight in their predictions and resulting decisions, reasoning that `similar agents behave similarly in similar situations'. 

This similarity-aware reasoning has important consequences for both prediction and behavior. For prediction, it allows an agent to leverage its own self-model to form more accurate predictions of others, a form of theory-of-mind central to human social cognition \citep{graziano2013consciousness}. For behavior, it redefines what constitutes \textit{rational} choice when taking embeddedness and functional similarities into account. It leads agents to reason in accordance with \textit{Evidential Decision Theory (EDT)} \citep{ahmed2021evidential, jeffrey1983logic, everitt2015sequential}, in contrast to the more orthodox \textit{Causal Decision Theory (CDT)} \citep{lewis1981causal, gibbard1978counterfactuals}. In EDT, an agent selects the action that would be the best \textit{news} to learn one has performed \citep{ahmed2014evidence, ahmed2021evidential}. This allows our embedded agents to treat their own deliberations as evidence on the behavior of other agents which are perceived to perform similar deliberations. For instance, making a choice to cooperate provides evidence that similar agents might make the same choice in a similar situation.
% This allows our embedded agents to treat their own deliberations as evidence: reasoning that ``similar agents behave similarly,'' a choice to cooperate provides evidence that other similar agents will also cooperate. 
CDT, by contrast, evaluates actions based on their expected \textit{causal consequences}, assuming the agent's choice is an independent intervention on the world \citep{lewis1981causal, everitt2015sequential}. This causal stance explicitly ignores the evidential link provided by functional similarities. The behavioral divergence is stark. Consider the \textit{Twin Prisoner's Dilemma}, where two identical copies of an AI agent play the prisoner's dilemma against each other. Classical game theory---built on a decoupled agency and CDT foundation with the Nash equilibrium as its solution concept---mandates defection as the only rational choice \citep{lewis1979prisoners}. An embedded agent reasoning evidentially, however, recognizes the perfect functional similarity. It correctly anticipates that its copy will make the same decision, making its own choice to cooperate a justifiably rational action, as mutual cooperation yields a higher personal reward than mutual defection.

To formalize this embedded notion of rationality in multi-agent scenarios, we introduce a novel family of game-theoretic solution concepts: the \textit{embedded equilibria}. These equilibria, inspired by the work of \citet{spohn2003dependency} on dependency equilibria and \citet{kalai1993rational, kalai1993subjective} on subjective equilibria, explicitly account for the coupled beliefs that arise from functional similarities. This opens new pathways for cooperation and coordination that are inaccessible to decoupled agents and classical game theory. The \textit{subjective embedded equilibrium} (SEE) describes a state where each agent's policy is a best response with respect to its own \textit{subjective} beliefs about the universe, including its beliefs about itself and its correlation with others. The \textit{embedded equilibrium} (EE), our proposed embedded counterpart to the Nash equilibrium, describes an optimal and stable pattern of behavior when the true functional similarities between agents are taken into account. A key result of our work is proving that embedded Bayesian agents, when satisfying the grain-of-truth property, are guaranteed to converge to playing $\epsilon$-subjective embedded equilibria in multi-agent interactions.

We note that our contributions build on a strong line of works that
%that we are not the first to 
leverage similarity between agents to achieve cooperation, and investigate the consequences of evidential reasoning in multi-agent scenarios. Much prior effort has explored the concept of similarity between agents, often by assuming agents are \textit{fully transparent}---that is, they have access to the source code or decision algorithm of their co-players \citep{brams1975newcomb, lewis1979prisoners, hofstadter1983dilemmas, howard1988cooperation, tennenholtz2004program, barasz2014robust, halpern2018game, critch2019parametric, oesterheld2019robust}. More recent work has generalized this by providing agents with a scalar \textit{similarity score} \citep{oesterheld2024similarity} or using Bayesian theory of mind to infer if others share a similar utility function \citep{kleiman2025evolving}. Our framework further generalizes these approaches. It shows how a joint predictive model of oneself and the environment can account for functional similarities without requiring full transparency, explicit access to code, or collapsing the rich nature of similarity into a single score or utility function. It is well-established that EDT can lead to cooperation in the Twin Prisoner's Dilemma \citep{ahmed2021evidential}. 
% \citet{everitt2015sequential} extended EDT from one-shot to sequential decision-making. 
In game theory, \citet{al2015evidential} introduced \textit{evidential equilibria} to describe game-theoretic equilibria between agents applying evidential decision theory, as a more accurate descriptive theory of human decision making. \citet{spohn2003dependency} argues that his closely related \textit{dependency equilibria} are not only more descriptive of human behavior, but are also normatively rational, motivating them with a reflexive decision theory that allows for causal links between agents' decision processes. \citet{ahmed2014evidence, ahmed2021evidential} further defend EDT as a normative theory for rational behavior, constructing thought experiments where CDT, unlike EDT, leads to dynamically inconsistent choices. By introducing embedded Bayesian agents and formalizing functional similarities via Shannon and algorithmic information theoretic concepts, we further motivate that EDT-like reasoning by leveraging functional similarities is normatively rational for embedded agency. We introduce novel solution concepts (EEs) incorporating these functional similarities, and show that `embedded rational learners'---our embedded Bayesian agents---are guaranteed to converge to the subjective variant (SEEs) of these equilibria.

The main remaining question is which hypothesis class and corresponding prior to use such that the resulting embedded Bayesian agents satisfy the grain-of-truth property, and thereby exhibit the behaviors described in the previous paragraphs. Satisfying the grain-of-truth property for non-trivial model classes is a challenging task \citep{leike2016formal, nachbar1997prediction, nachbar2005beliefs, foster2001impossibility}. As a step towards defining and understanding what the ideal, ultimate embedded intelligent agent might be, we then introduce the E\textbf{M}bedded \textbf{U}niversal \textbf{P}redictive \textbf{I}ntelligence (MUPI) framework, extending the seminal theoretical work on AIXI by \citet{hutter-aixi}. Such a universal embedded agent starts with a Bayesian prior over a wide hypothesis class including all {\it computable universes}, favoring simple over complex universes by assigning higher prior probability mass to the former, thus abiding to Occam's razor. Building upon recent variants of AIXI that (i) incorporate self-prediction in the single-agent, decoupled setting \citep{catt2023self}, and (ii), leverage a landmark mathematical construction of \textit{reflective oracles} \citep{fallenstein2015reflective_oracles} to widen the hypothesis class over environments to include environments containing other AIXI agents \citep{fallenstein2015reflective, leike2016formal,wyeth2025limit}, we show that our embedded universal agents satisfy the grain-of-truth property and hence are consistent under mutual prediction, and therefore possess infinite-order theory of mind. We further formalize functional similarities from an algorithmic information theory perspective, where functional similarities between agents correspond to how well the agent's programs can be jointly compressed. As universes containing similar agents have shorter description lengths compared to universes containing dissimilar agents, the Occam's razor prior naturally favors such universes. This demonstrates that the functional-similarity-based reasoning of embedded Bayesian agents is not an ad-hoc assumption, but rather a core principle of reasoning about the world using Occam's razor. Finally, at the end of the paper, we return to real-world artificial intelligence, and discuss the practical implications of our theory on current AI systems based on foundation models jointly predicting actions and percepts.

\begin{BoxFigure}[
  %title=
  Summary of core contributions
]
\label{box:core-contributions}
\begin{enumerate}
    \item We formalize \textbf{embedded Bayesian agents} as a conceptual framework to address two main challenges in multi-agent learning: \textit{prospective learning} and \textit{embedded agency}.
    \begin{itemize}
        \item[(i)] Agents incorporate prospective learning by using Bayesian sequence prediction to forecast future consequences, critically sidestepping the limiting assumptions of stationarity, ergodicity or Markov dynamics.
        \item[(ii)] They embody embeddedness by forming beliefs over \textbf{universes} that contain both themselves and the environment. This results in a unified Bayesian prediction model for both external percepts and the agent's \textbf{own actions}. This joint prediction makes embedded Bayesian agents an appealing theoretical model organism for current practical AI agents based on foundation models which are also joint action-percept prediction models.
    \end{itemize}

    \item We formalize \textbf{functional similarities} between agents as the property that their policies contain mutual Shannon or algorithmic information.
    \begin{itemize}
        \item[(i)] Embedded Bayesian agents leverage this by reasoning that "similar agents behave similarly in similar situations," leading on average to more accurate predictions than their decoupled counterparts when such similarities are present.
        \item[(ii)] We show that reasoning about functional similarities is not an ad-hoc assumption but a fundamental consequence of applying Occam's razor in a universal prediction setting: We prove that a universal belief distribution based on a \textit{Solomonoff prior}---an Occam's razor prior which assigns higher probability to simpler (more compressible) universes--- always results in a positive mutual information between the agent's policy and the environment (possibly containing other agents), as universes containing similar agents are algorithmically simpler to describe.
    \end{itemize}
    
    \item We introduce new game-theoretic solution concepts, the \textit{subjective embedded equilibrium (SEE)} and \textit{embedded equilibrium (EE)}.
    \begin{itemize}
        \item[(i)] These concepts redefine rationality for embedded agents in game-theoretic settings, accounting for functional similarities to enable new forms of cooperation and coordination that are unattainable by classical Nash equilibria.
        \item[(ii)] We prove that embedded Bayesian agents satisfying the \textit{grain-of-truth} property converge to playing $\epsilon$-SEEs, and converge towards EEs when additional conditions are satisfied.
    \end{itemize}

    \item To solve the  grain-of-truth problem for embedded agency (a key condition for the convergence results above), we introduce the EMbedded Universal Predictive Intelligence (MUPI) framework.
    \begin{itemize}
        \item[(i)] We develop the \textit{Reflective Universal Inductor (RUI)}: a universal prediction model that can consistently reason about universes containing agents that use the RUI itself as prediction model, thereby resolving the infinite recursions of mutual prediction.
        \item[(ii)] We prove the existence of the RUI, providing a new tool for building universally intelligent agents that stands as an alternative to the reflective oracle framework \citep{fallenstein2015reflective_oracles}.
    \end{itemize}
\end{enumerate}
\end{BoxFigure}

\section{Background}\label{sec:background}

\subsection{Mathematical Preliminaries}

\subsubsection{Sequences}
Let $\calX$ be an arbitrary countable set. For a fixed integer $n$, $\calX^n$ is the set of sequences of length $n$ whose elements are in $\calX$, also known as $\calX$-sequences of length $n$. We adopt the convention that $\calX^0=\{\varepsilon\}$ where $\varepsilon$ is the empty sequence/string, in contrast to $\epsilon$ which we use as notation for a small scalar value. We write $\calX^{*}$ to denote the set of finite $\calX$-sequences, i.e.,
$$\calX^*:=\bigcup_{n\geq 0}\calX^n\,.$$

We write\footnote{Note that $A^B$ denotes the set of mappings from $B$ to $A$. Therefore, $\calX^\mathbb{N}$ is the set of infinite sequences in $\calX$ with indices in $\mathbb{N}$.} $\calX^{\infty}:=\calX^\mathbb{N}$ to denote the set of infinite $\calX$-sequences, and  $\calX^{\#}:=\calX^{*}\cup\calX^{\infty}$
to denote the set of all (finite and infinite) sequences. For $x,y\in\calX^\#$, we write $x\sqsubseteq y$ to denote that $x$ is a prefix of $y$.
For $x\in\calX^*$ and $y \in \calX^{\#}$, we write $xy$ to denote the sequence obtained by concatenating $x$ and $y$.

For any sequence $x\in \calX^{\#}$, we denote the (possibly infinite) length of $x$ as $l(x)$, and for every $1\leq i\leq l(x)$, we denote the $i$-th symbol of $x$ as $x_i$. For $1\leq i\leq j\leq l(x)$ we write $x_{i:j}$ to denote the subsequence $(x_i,\ldots,x_j)$. We also use the notation $x_{< t}$ as a shorthand for $x_{1:t-1}$ and $x_{\leq t}$ for $x_{1:t}$.

\subsubsection{Probability Background}

We use the following definitions of measures and semimeasures:\footnote{Semimeasures are important in algorithmic probability which we will leverage later in this manuscript, where sequences are generated by Turing machines, which can halt or loop forever without outputting further symbols, and hence can output both finite sequences and infinite sequences (cf. \cref{app:preliminaries} for an in-depth discussion on semimeasures and their properties).}

\begin{definition}[Semimeasures and measures] \label{def:semi-measures}
Let the \emph{cylinder set} $\Gamma_h$ be the set of all sequences $y \in \calX^{\#}$ that start with finite sequence $h \in \calX^*$. Using the abuse of notation $\sigma(h):=\sigma(\Gamma_h)$, a \emph{semimeasure} $\sigma: \calX^* \to [0, 1]$ on the sample space $\calX^{\#}$ satisfies the following conditions:
\begin{align}
    \sigma(\varepsilon)\leq 1 \quad \quad \text{and} \quad \quad \sigma(h) \geq \sum_{x\in\calX}\sigma(hx),
\end{align}
A \emph{measure} is a semimeasure where both the above inequalities are tight. Intuitively, $\sigma(h)$ represents the probability that a sampled sequence starts with prefix $h$. When a strict inequality $\sigma(h) > \sum_{x\in\calX}\sigma(hx)$ applies, there is a non-zero probability that the sequence ends after $h$.\footnote{In order to interpret the semimeasure $\sigma(h)$ as the probability that a sequence $x\in \calX^\#$ (which can be finite or infinite) starts with $h$, we require that $\sigma(\varepsilon)=1$, which can be achieved by renormalizing the semimeasure by $\sigma(\varepsilon)$ whenever it is different from zero.} Conditional (semi-)measures are defined by 
\begin{align}
    \sigma(z\mid h):=\frac{\sigma(hz)}{\sigma(h)}.
\end{align}
\end{definition}
The conditional measure $\sigma(z\mid h)$ is undefined when $\sigma(h) = 0$

We refer to measures and distributions interchangeably. For any given set $\calX$, the notation $\Delta\calX$ represents the set of all probability distributions over the elements of $\calX$, and $\Delta'\calX$ the set of all semiprobability distributions over $\calX$, i.e., functions $\sigma$ that satisfy $\sum_{x\in \calX} \sigma(x) \leq 1$ and $\sigma(x) \geq 0$ for all $x\in\calX$.

We introduce a distance metric between measures over sequences:\footnote{See \cref{app:preliminaries} for a more general definition of total variation applicable to \textit{semimeasures} as well.}
\begin{definition}[Total variation distance]\label{def:total-variation-distance}
    Let $P^1$ and $P^2$ be two measures over sequences. For every $k\geq1$, we define the $k$-step total variation distance between $P^1$ and $P^2$ conditioned on history $h$ as:
    $$D_k(P^1, P^2 \mid h)=\frac{1}{2}\sum_{h'\in\calX^k}|P^1(h'\mid h)-P^2(h'\mid h)|\,.$$
    It is easy to show that $D_{k+1}(P^1, P^2 \mid h)\geq D_{k}(P^1, P^2 \mid h)$. By taking $k\to\infty$, we get the total variation distance:$$D_\infty(P^1, P^2 \mid h)=\sup_{k\geq 1}\frac{1}{2}\sum_{h'\in\calX^k}|P^1(h'\mid h)-P^2(h'\mid h)|\,.$$
\end{definition}

We define a notion of dominance, which is a stronger form of the more familiar absolute continuity.
\begin{definition}[Dominance]\label{def:dominance}
    Given two measures $P^1$ and $P^2$ over $\calX^{\#}$, we say that a measure $P^1$ dominates a measure $P^2$, and write $P^1\stackrel{\times}\geq P^2$, if there exists $C>0$ such that $P^1(x)\geq C\cdot P^2(x)$ for all $x\in\calX^*$.
\end{definition}
Note that $C$ cannot depend on $x$ in the above definition.

\subsection{General Reinforcement Learning Setup}\label{sec:background-grl}
We briefly review the general reinforcement learning setting \citep{Hutter:24uaibook2} that we build upon. In this setting, the reward and environment transition functions depend on the full history of interactions up to the present time, thus subsuming the Markov decision process (MDP) framework that underpins classical optimal control and reinforcement learning as a special case. We then define the Bayes-optimal agent, which maximizes expected return while weighing a set of environments an agent might find itself in against the evidence contained in the history of interactions so far. These concepts serve as the foundation for the embedded Bayesian agents that we study.

\paragraph{General reinforcement learning.}

Consider a finite set of possible actions, $\calA$, a finite set of possible observations, $\calO$, and a finite set of possible rewards, $\calR \subset [0,1]$. We define the set of \textit{percepts} as the Cartesian product of observations and rewards, $\calE := \calO \times \calR$. 

A \textit{history} consists of a finite sequence of alternating actions and percepts. Abusing notation slightly, we denote a single \textit{turn} in a history (an agent taking an action and observing a percept) as $\Turn \in \TurnSet := \calA \times \calE$, a $t$-turn history as $\Hist \in \TurnSet^t$ and an arbitrary-length history as $h \in \TurnSet^*.$ Again abusing notation slightly, we write $l(h)$ to denote the number of turns in $h$, i.e., $l(h)=t$ when $h \in \TurnSet^t$.

An agent's policy $\Mvp$ maps a given history to a distribution over actions: $\Mvp : \TurnSet^* \to \Delta \calA$. We write $\Mvp(a|\Hist):= \Mvp(\Hist)(a)$ to denote the conditional probability that the agent takes an action $a$ given a history $\Hist$. The \textit{environment dynamics} $\Mve$ maps a history and an action to a distribution over subsequent percepts: $\Mve : \TurnSet^* \times \calA \to \Delta\calE$. We write $\Mve(e|\Hist,a):=\Mve(\Hist,a)(e)$ to denote the conditional probability that the environment produces the percept $e$ given that the agent took action $a$ after history $\Hist$. When a policy $\Mvp$ is interacting with an environment $\Mve$, it induces a measure $\Mve^\Mvp$ on the space of infinite histories.

The goal of general reinforcement learning is to obtain an optimal policy $\Mgp$ that maximizes the expected discounted sum of rewards. The value function for a given environment $\Mve$, discount factor $\gamma \in [0,1)$ and policy $\Mvp$ starting from history $\Hist$ is defined as
\begin{align}\label{eq:value-function}
V_{\Mve^\Mvp}(\HistM) := (1-\gamma)\mathbb{E}_{\Mve^\Mvp} \left[ \sum_{k=t}^\infty \gamma^{k-t} r_k \bigg| \HistM \right]\,,
\end{align}
where we multiply by $(1-\gamma)$ to ensure that values are bounded between 0 and 1.
The set of optimal policies is equal to $\argmax_\Mvp V_{\Mve^\Mvp}(\HistM)$ and the associated optimal value function is defined as $V^*_\Mve(\HistM) := \max_\Mvp V_{\Mve^\Mvp}(\HistM)$. We denote an individual optimal policy with $\Mgp \in \argmax_\Mvp V_{\Mve^\Mvp}(\HistM)$. 
To facilitate analysis, we also define the action-value function, or Q-value, as $$Q_{\Mve^\Mvp}(\HistM, a_{t}) := \expect{\Mve(e_t|\HistM,a_t)}{(1-\gamma)r_t+\gamma V_{\Mve^\Mvp}(\HistM \Turn_t)}.$$ 

\paragraph{Multi-agent general reinforcement learning (MAGRL).}
We can straightforwardly generalize general reinforcement learning to a multi-agent setup as follows. Consider N agents $i\in N$ with finite action spaces $\calA^i$. We denote the joint action as $\bara \in \barcalA := \prod_i \calA^i$, and the joint percept as $\bare \in \barcalE := \prod_i \calE^i$. We define the space of multi-agent turns $\MATurnSet := (\barcalA \times \barcalE)$, the space of multi-agent histories $\MATurnSet^*$ (a $t$-turn history is $\MAHist \in \MATurnSet^t$), and the space of extracted single-agent histories $(\TurnSet^i)^* := (\calA^i \times \calE^i)^*$. Each agent has a policy $\pi^i: (\TurnSet^i)^* \to \Delta \calA^i$, which chooses an action given agent $i$'s history. The multi-agent environment $\MultiAgentMve: \MATurnSet^* \times \barcalA \to \Delta \barcalE$ defines the joint distribution over percepts for each agent, conditioned on the joint history. We denote ${\MultiAgentMve}^{\MultiAgentMvp}$ as the joint distribution induced by $\MultiAgentMve$ and agent policies $\MultiAgentMvp=(\pi^i)_{i=1}^N.$ Note that all agents act simultaneously without observing the current action of the other agents, and their next percept $e^i$ can contain information about the actions taken by other agents.

Combining the multi-agent environment $\MultiAgentMve$ with the policies $(\Mvp^j)_{j\neq i}$ of the other agents, and marginalizing out their histories and percepts 
leads to a personal environment $\Mve^i(e^i \mid \HistI, a^i)$ for each agent, which depends on the policies of the other agents. Combining the personal environment $\Mve^i$ with the personal policy $\Mvp^i$ leads to a distribution over personal histories $\MultiAgentMuPiI$. 

When we omit superscript $i$ with actions $a$ and percepts $e$, we indicate a single agent's actions and percepts originating from its personal environment.

\subsection{Bayesian prediction and agents}\label{sec:decoupled-bayesian-agents}
When an agent is uncertain about which environment it is interacting with, it needs to (i) learn about its environment from past observations and actions, and (ii) behave optimally taking into account its uncertainty on which environment it is interacting with. \emph{Bayesian agents} formalize these notions of \emph{rational agency} by (i) using Bayesian mixture environments to quantify their beliefs about which environment they are interacting with and making posterior belief updates leveraging incoming observations, and (ii) using optimal planning within the Bayesian mixture environment to obtain a behavioral policy. 

\paragraph{Bayesian mixture environment.}
Using a countable hypothesis space $\calMenv$ over environments $\Mve$, we can define the Bayesian mixture environment as
\begin{align}\label{eqn:decoupled-bayesian-agent}
    \Mme(e_t \mid \HistM, a_t) := \sum_{\Mve \in \calMenv} w(\Mve \mid \HistM)\Mve(e_t \mid \HistM, a_t), \quad \quad w(\Mve \mid \Hist):= w(\Mve \mid \HistM) \frac{\Mve(e_t \mid \HistM, a_t)}{\Mme(e_t \mid \HistM, a_t)}\,,
\end{align}
with $w(\Mve):= w(\Mve \mid \varepsilon)$ the prior belief distribution and $\varepsilon$ the empty history. We assume throughout that $w(\Mve) > 0 \ \forall \Mve \in \calMenv$, and that $\sum_{\Mve\in\calMenv}w(\Mve)=1$.

\paragraph{Decoupled Bayes-optimal agent.}
Bayes-optimal agents act optimally w.r.t. their beliefs over environments. Hence, starting from a specific history $\HistM$, they select an optimal policy w.r.t. their Bayesian mixture environment $\Mme$ through optimal planning. We call such Bayes-optimal agents \textit{decoupled Bayes-optimal agents} (to contrast them with \textit{embedded Bayes-optimal agents} which we will introduce in \cref{sec:embedded-bayesian-agents}), and call their optimal policy a \textit{decoupled Bayes-optimal response} (DBR):
\begin{align}\label{eqn:decoupled-br}
    \Mvp_{DBR} \in \argmax_{\Mvp}V_{\Mme^\Mvp}(\HistM).
\end{align}

When a Bayesian mixture environment-policy measure $\Mme^\Mvp$ includes (puts positive weight on) a "true" environment-policy measure $\Mge^\Mvp$, we say that $\Mme^\Mvp$ satisfies the \textit{grain-of-truth property} \citep{kalai1993rational} with respect to $\Mge^\Mvp$; when it is clear from context, we will refer to "having grain of truth" without mentioning the specific ground-truth and mixture distributions. Because $\Mge^\Mvp$ has positive weight in $\Mme^\Mvp$, it is obvious that $\Mme^\Mvp \stackrel{\times}\geq \Mge^\Mvp$. The classic theorem below shows that this is sufficient to guarantee that $\Mme^\Mvp$ converges almost-surely to $\Mge^\Mvp$.\footnote{Note that this theorem can be generalized to semimeasures (cf. \cref{app:preliminaries}), and the dominance condition can be further relaxed to requiring that $\Mge^\Mvp$ is absolutely continuous w.r.t. $\Mme^\Mvp$.}

\begin{theorem}[Convergence of $\Mme$ to $\Mge$ in total variation \citep{blackwell1962merging, Hutter:24uaibook2}]\label{theorem:decoupled-convergence-in-tv}
For any policy $\Mgp$, consider a ground-truth environment-policy measure $\Mge^\Mgp$ and a Bayesian mixture environment-policy measure $\Mme^\Mgp.$ If $\Mme^\Mgp \stackrel{\times}\geq \Mge^\Mgp$, 
$$D_{\infty}(\Mme^\Mgp, \Mge^\Mgp \mid \Hist) \to 0 ~~ \text{as} ~~ t\to \infty ~~ \Mge^\Mgp\text{-almost-surely}\,.\footnote{It is worth noting the generalized Solomonoff bounds of, e.g., \cite[Theorems 3.2.5 and 3.3.4]{Hutter:24uaibook2} imply that the convergence is very fast.}$$
\end{theorem}
The above theorem has important implications in multi-agent settings: Assume that $N$ agents are interacting through a multi-agent environment $\MultiAgentMge$. Each agent can be uncertain about the policies of the other agents, and they may also be uncertain about the statistics of the ground-truth multi-agent environment $\MultiAgentMge$. We can model these uncertainties through a Bayesian approach by assuming that each agent has a prior probabilistic belief about the multi-agent environment\footnote{If $\MultiAgentMge$ is known, then the prior probability distribution would be concentrated on it.} $\MultiAgentMge$, and prior probabilistic beliefs about the policies of the other agents. Since specifying the policies of other agents reduces a multi-agent environment to a personal single-agent environment, we may assume without loss of generality that each agent has a prior probabilistic belief over the possible personal single-agent environment it is interacting with. Let $\Mme^i$ be the Bayesian mixture (personal) environment of the $i$-th agent according to its prior, and assume that each agent performs optimal planning with respect to its (subjective) mixture personal environment. Let $\Mgp^i$ be the Bayes-optimal policy of the $i$-th agent, and let $\Mge^i$ be the ground-truth personal environment from the perspective of the $i$-th agent, which is the result of combining the ground-truth multi-agent environment $\MultiAgentMge$ and the policies $(\Mgp_j)_{j\neq i}$ of the other agents. \cref{theorem:decoupled-convergence-in-tv} implies that when Bayes-optimal agents satisfy the grain-of-truth property (i.e., $(\Mme^i)^{\Mgp^i} \stackrel{\times}\geq (\Mge^i)^{\Mgp^i}\,,\forall i\in\{1,\ldots,N\}$), they converge towards accurate and consistent mutual prediction.

The seminal work of \citet{kalai1993rational, kalai1993subjective} uses this convergence property to show that in a multi-agent setting, Bayes-optimal agents that satisfy the grain-of-truth property converge to an $\epsilon$-Nash equilibrium when interacting with each other on repeated games. These powerful results hinge upon the grain-of-truth assumption, which is notoriously hard to satisfy when using large hypothesis classes \citep{leike2016formal,shoham2008multiagent,foster2001impossibility,nachbar1997prediction, nachbar2005beliefs}.

To build intuition for this difficulty, consider a two-player game where the environment dynamics $\MultiAgentMge$ are known to both players. Each agent $i\in\{1,2\}$ is uncertain about the policy of the other agent $j$ (where $j=2$ if $i=1$ and vice versa), and so agent $i$ starts with a hypothesis class of possible policies, $\calMpol^{j}$, that the other agent $j$ might be using. By combining its Bayesian beliefs about the opponent's policy with the known game dynamics, each agent $i$ induces a personal mixture environment $\Mme^i$. Assume both agents are Bayes-optimal, computing policies ${\Mgp}^1$ and ${\Mgp}^2$ with respect to their personal mixture environments. A sufficient condition for the grain-of-truth property to hold is that each agent's hypothesis class contains the other agent's true policy,\footnote{This is because we assume that every policy in the class has a positive probability. Hence, for example, if ${\Mgp}^2 \in \calMpol^2$ then ${\Mgp}^2$ has a positive probability according to the prior belief of agent 1, which implies that ${(\Mme^1)} \stackrel{\times}\geq {(\Mge^1)}$, where $\Mge^1$ is the ground-truth personal environment of agent 1, which is obtained by combining $\MultiAgentMge$ with $\Mgp^2$. One can deduce from this the grain-of-truth property ${(\Mme^1)}^{\Mgp^1} \stackrel{\times}\geq {(\Mge^1)}^{\Mgp^1}$.} i.e., ${\Mgp}^2 \in \calMpol^2$ and ${\Mgp}^1 \in \calMpol^1$. However, there is no a priori guarantee that this will be the case.

A seemingly simple fix would be to extend the original classes to accommodate the computed policies: If ${\Mgp}^1$ is not in $\calMpol^1$, one could just add it to form a new class $\calMpol^{1\prime} = \calMpol^1 \cup \{{\Mgp}^1\}$ and assign it a positive prior probability. This, however, creates a self-referential loop. If agent 2 updates its beliefs to use this expanded class $\calMpol^{1\prime}$, its own mixture environment $\Mme^2$ changes, which in turn changes its optimal policy to some new ${\Mgp}^{2 \prime}$ possibly different than ${\Mgp}^{2}$. This new policy ${\Mgp}^{2 \prime}$ is not guaranteed to be in agent 1's original hypothesis class $\calMpol^2$. Therefore, even if we also define $\calMpol^{2\prime} = \calMpol^2 \cup \{{\Mgp}^2\}$, the new policy ${\Mgp}^{2 \prime}$ is not guaranteed to be in the updated hypotheses class $\calMpol^{2\prime}$. Again, if we add ${\Mgp}^{2 \prime}$ to $\calMpol^{2\prime}$, this would change the mixture $\Mme^1$, which would in turn change the Bayes-optimal policy of agent 1 to something that is not guaranteed to be in $\calMpol^{1\prime}$. This circular dependency is the core of the grain-of-truth problem in multi-agent systems. Its solution requires constructing hypothesis classes that are a \textit{fixed point}---ones that are already rich enough to contain the optimal agents defined over them.

This motivates the following problem statement:

\begin{problem}[The general grain-of-truth problem \citep{hutter2009open, kalai1993rational} - Informal]\label{prob:got-decoupled-informal}
    Find a large class of environments $\calMenv$ that includes environments containing other Bayesian agents that use a universal prior\footnote{i.e., a prior that is non-zero for all $\nu \in \calMenv$.} over $\calMenv$.
\end{problem}

The work of \citet{leike2016formal} and \citet{wyeth2025limit} solved the above grain-of-truth problem for the case of decoupled Bayesian agents, leveraging algorithmic information theory which we discuss in the next section.
% Solving the above general grain-of-truth problem for the case of \textit{embedded} Bayesian agents is one of the main contributions of this work (cf. \cref{sec:mupi}). 

\begin{remark}[Prospective prediction]
    \cref{theorem:decoupled-convergence-in-tv} shows that principled prospective prediction is possible without the need for stationarity assumptions. The environments $\Mve$ are not required to be stationary or ergodic, and can for example include the learning dynamics of other agents. \cref{theorem:decoupled-convergence-in-tv} shows that when using a Bayesian mixture environment $\Mme$ that satisfies the grain-of-truth property, the resulting predictive distribution converges to the ground-truth distribution over future percepts, allowing Bayesian agents to anticipate a possibly changing future. A core result in Bayesian prediction states that the total prediction loss made by $\Mme$ in ground-truth environment $\Mge$ is bounded from above by a quantity that is proportional to $\log (w(\Mge)^{-1})$ \citep{Hutter:24uaibook2}. Hence, when using a prior that assigns more probability mass to `simple, structured environments', following Occam's razor, $\Mme$ quickly converges to an accurate predictive model for structured ground-truth environments. 
\end{remark}

\subsection{Algorithmic probability, Solomonoff induction and AIXI}\label{sec:background-ait}

In this work, we build upon Hutter's universal artificial intelligence framework \citep{hutter-aixi} to define a universally intelligent embedded agent. Here, we briefly cover some important concepts from algorithmic probability, Solomonoff induction and AIXI, while we refer the interested reader to \cref{app:preliminaries} and \citet{Hutter:24uaibook2} for a more detailed discussion. 

Throughout, for any countable set, we (implicitly) assume a fixed, canonical encoding of its elements as finite binary strings. When we speak about computations involving these sets, we are (implicitly) referring to computations on the canonical encodings of the elements.

\begin{definition}[Computable]\label{def:computable}
    Let $\mathcal{X}$ and $\mathcal{Y}$ be two countable sets. A function $f:\mathcal{X}\to\mathcal{Y}$ is computable if there exists a Turing machine that computes $f$.
\end{definition}

\begin{definition}[Lower semicomputable]\label{def:lsc}
    Let $\mathcal{X}$ be a countable set.
    A function $f:\mathcal{X}\to\mathbb{R}$ is lower semicomputable (l.s.c.) if there exists a computable function $\phi:\mathcal{X}\times\mathbb{N}\to\mathbb{Q}$ such that $\phi(x,n)\leq \phi(x,n+1)$ and $\lim_{n\to\infty}\phi(x,n)=f(x)$ for all $x\in\mathcal{X}$.
\end{definition}

\begin{definition}[Limit computable]\label{def:limit-computable}
    Let $\mathcal{X}$ be a countable set.
    A function $f:\mathcal{X}\to\mathbb{R}$ is limit computable\footnote{In some references, e.g., \cite{Hutter:24uaibook2}, limit computable functions are also called "approximable". In this paper, we will only use the term "limit computable".} if there exists a computable function $\phi:\mathcal{X}\times\mathbb{N}\to\mathbb{Q}$ such that $\lim_{n\to\infty}\phi(x,n)=f(x)$ for all $x\in\mathcal{X}$.
\end{definition}

\begin{definition}[Monotone Turing machine]\label{def:monotone-turing}
A monotone Turing machine is a Turing machine $T$ equipped with (i) a binary unidirectional\footnote{Throughout this paper, unidirectional tapes are tapes where the head can only move from left to right.} read-only input tape, (ii), a binary unidirectional write-only output tape, and (iii), a binary bidirectional read/write work tape.
\end{definition}

We define $\calM^{LSCSM}$ as the set of all lower semicomputable semimeasures $\sigma: \calX^* \to [0,1]$. An important result from algorithmic information theory states that any lower semicomputable semimeasure corresponds to a semimeasure induced by a monotone Turing machine with independent uniformly random bits on its input tape and vice versa \citep[Chapter 4.5]{li2008introduction}. Hence, $\calM^{LSCSM}$ can be obtained by enumerating (possibly with repetition) all monotone Turing machines.

\paragraph{Solomonoff induction.} Solomonoff Induction \citep{solomonoff-complexity-induction-1978} computes the posterior probability of a l.s.c. semimeasure by maintaining a Bayesian mixture $$\Mme_U:= \sum_{\Mve \in \calM^{LSCSM}} w(\Mve) \Mve$$ over $\calM^{LSCSM}$ using a universal prior (semi-)probability $w: \calM^{LSCSM} \to [0,1]$ defined as $w(\Mve) := 2^{-K(\Mve)}$ with $K(\Mve)$ the prefix-Kolmogorov complexity of $\Mve$ \citep{Hutter:24uaibook2} with respect to some reference universal monotone Turing machine.\footnote{For the definition of prefix-Kolmogorov complexity and a more detailed discussion of Solomonoff induction, refer to \cref{app:preliminaries}.}

Because the universal prior is lower semicomputable, the universal mixture $\Mme_U$ itself is lower semicomputable and hence is part of the hypothesis class $\calM^{LSCSM}$. This universal prior elegantly incorporates (i) Occam's razor by assigning simpler $\Mve$ a higher prior probability, and (ii) Epicurus' principle of multiple explanations, encouraging us to keep all theories consistent with the observations by providing all $\Mve \in \calM^{LSCSM}$ a non-zero prior probability.

To extend the ideas of Solomonoff Induction to the agentic setting, where an agent is choosing an action at each step via a planner rather than merely receiving an observation and updating mixture probabilities, we represent environments as Turing machines. As machines can fail to produce an output, we need to widen the definition of an environment to chronological conditional semimeasures\footnote{Refer to  \cref{app:preliminaries} for the formal definition of chronological conditional semimeasures and for more details about their interpretation.} defined as mappings with signature $\nu: \calH\times \calA \to \Delta'\calE$, with $\Delta'\calE$ the set of semiprobabilities over $\calE$, i.e., $\sum_{e\in\calE} \nu(e \mid ha) \leq 1$, instead of a strict equality to 1 for probabilities. We define  $\calMenv^{\text{LSC}}$ as the set of all lower semicomputable environments.
\paragraph{Universal artificial intelligence.} AIXI \citep{hutter-aixi} combines Solomonoff induction with general reinforcement learning to achieve a universally intelligent agent. AIXI uses a universal mixture environment $\Mme$ over $\calMenv^{\text{LSC}}$ with an l.s.c. universal prior $w$. The AIXI agent then performs infinite horizon optimal planning w.r.t. $\Mme$ for some discount $\gamma$, i.e., it is a Bayes-optimal agent w.r.t. the universal mixture environment $\Mme$.

While the universal mixture environment $\Mme$ is itself lower semicomputable, the AIXI agent is not \citep{leike2016nonparametric}. Hence, environments that contain other AIXI agents are not lower semicomputable (i.e., they are not part of $\calMenv^{\text{LSC}})$.\footnote{AIXI itself is not even limit computable. $\epsilon$-optimal infinite horizon AIXI \textit{is} limit computable \citep{leike2015computability}, but is not lower semicomputable.} If the true environment contains other AIXI agents, the grain-of-truth assumption (cf. \cref{def:dominance}) is violated. As a result, in such cases, \cref{theorem:decoupled-convergence-in-tv} does not apply and the universal mixture $\Mme$'s predictions do not necessarily converge to those of the ground-truth environment as time goes to infinity, and AIXI agents are not guaranteed to converge to an $\epsilon$-Nash equilibrium. This motivates widening $\calMenv$ beyond lower-semicomputable environments in order to incorporate environments containing other AIXI agents.

\paragraph{Reflective oracles.}  A classical approach in computer science to investigate non-computable objects is to provide Turing machines query access to an \textit{oracle} which is allowed to be incomputable, with a famous example being the Halting oracle answering queries of the form \textit{"Does machine $T$ halt on input $x$?"} \citep{rogers1987theory}. Turing machines with access to an oracle are called \textit{oracle machines}, and functions implementable on such oracle machines are \textit{oracle computable}. Given a set of oracle machines with access to oracle $O$, one can create a hierarchy of oracles by creating a new oracle $O'$ that answers questions about the $O$-oracle machines and those answers are allowed to be non-$O$-oracle computable. Rather than creating an infinite hierarchy of oracles, the \textit{reflective oracles} framework \citep{fallenstein2015reflective_oracles} introduces an oracle that answers questions about the output distribution of machines using the \textit{same} oracle, hence its reflective nature. More specifically, a reflective oracle answers queries of the form $\langle T, q\rangle$ representing the question \textit{``Is the probability that machine T with access to reflective oracle $O$ outputs 1 greater than q?"}.

\citet{fallenstein2015reflective_oracles} proved that such reflective oracles exist, and \citet{leike2016formal} showed that there are reflective oracles which are limit computable. By describing environments as machines with access to a reflective oracle, \citet{fallenstein2015reflective} and \citet{leike2016formal} created a new environment class $\calMenv^{\mathrm{RO}}$ that includes environments containing other \textit{reflective AIXI agents}, i.e., AIXI agents that use a universal mixture over $\calMenv^{\mathrm{RO}}$ instead of $\calMenv^{\text{LSC}}$. This setup hence solves the general grain-of-truth problem (Problem \ref{prob:got-decoupled-informal}) for decoupled Bayesian agents. 

\paragraph{Self-AIXI.} \citet{catt2023self} introduce a variant of AIXI agents which are \textit{self-predictive}, i.e., they maintain a belief distribution over which policy they are running. Let $\calMpol$ be a countable class of policies $\Mvp: \TurnSet^* \to \Delta'\calA$, and $\omega \in \Delta' \calMpol$ some prior semiprobability function. Self-AIXI introduces a Bayesian mixture policy $\Mmp:= \sum_{\pi \in \calMpol} \omega(\Mvp) \Mvp$, which it uses to predict its own future behavior. Then, instead of infinite-horizon optimal planning, Self-AIXI performs a simple \textit{policy improvement step}, reminiscent of TD(0) \citep{sutton1988learning} and the MuZero family of methods \citep{silver2017mastering, silver2018general, schrittwieser2020mastering, antonoglou2021planning}:

\begin{align}\label{eq:self-aixi}
    \Mgp_S(\Hist):= \argmax_{a\in \calA} Q_{\Mme^\Mmp}(\Hist,a)
\end{align}
with $\Mme$ the Bayesian mixture environment over some $\calMenv$, and $Q_{\Mme^\Mmp}(\Hist,a)$ the action-value function corresponding to deploying $\Mmp$ in environment $\Mme$. Interestingly, as the history $\Hist$ contains actions resulting from the above policy improvement steps, the mixture policy $\Mmp$ can recognize this pattern and predict future policy improvement steps in the future. Under additional assumptions on the mixture models $\Mme$ and $\Mmp$, \citet{catt2023self} show that Self-AIXI converges to AIXI when time goes to infinity, i.e., it converges to infinite-horizon optimal planning. For these results to hold, it is crucial that the predictive distribution of the policy mixture $\Mmp(a\mid \Hist)$ converges to the ground-truth self-AIXI policy $\Mgp_S(a \mid \Hist)$, and hence that the Self-AIXI policy $\Mgp_S$ is part of the hypothesis class $\calMpol$ itself, thereby satisfying the grain-of-truth property. Importantly, \cite{catt2023self} did not prove whether their proposed mixture policy $\Mmp$ does in fact dominate $\Mgp_S$, making it still an open problem in the field to construct a model class $\calMpol$ and corresponding mixture policy $\Mmp$ satisfying the grain-of-truth property. In this work, we show that the reflective oracle framework, and a new variant of our own, the \textit{reflective universal inductor} framework can be readily used to solve this grain-of-truth problem. In concurrent work, \citet{wyeth2025limit} shows a similar result for solving the grain-of-truth problem for Self-AIXI leveraging the reflective oracles framework. Importantly, Self-AIXI uses a separate mixture model $\zeta$ over policies that is decoupled from the mixture model $\xi$ over environments. Hence, although Self-AIXI has a self-model, it does not consider itself as part of the environment, and hence is a decoupled Bayesian agent. 

\paragraph{Joint AIXI.} Concurrent work by \citet{wyeth2025formalizing} introduces \textit{joint AIXI} (JAIXI). In contrast to Self-AIXI, which uses separate mixture models for the environment and the policy, JAIXI leverages a single "joint mixture" over semimeasures to predict an interleaved sequence of both actions and percepts. The JAIXI agent is then defined as the agent that plans optimally with respect to the resulting predictive model. This joint prediction setup is also central to the embedded Bayesian agents we study in this paper, though our focus is on a complementary research direction. The work of \citet{wyeth2025formalizing} emphasizes that if the joint mixture is a Solomonoff inductor over $\calM^{\text{LSCSM}}$, the resulting JAIXI agent is not itself lower semicomputable. This implies that the agent's own behavior lies outside its hypothesis class, thus violating the grain-of-truth property. The authors use this to formalize the 
"embeddedness failures" of this approach, showing that JAIXI fails to learn from certain adversarial, incomputable action-percept sequences.

Building upon the seminal work of AIXI \citep{hutter-aixi} and its variants, our work takes a complementary path. We argue that the prospective learning inherent in Bayesian sequence prediction, and hence AIXI agents, is a crucial capability for multi-agent learning, generalizing recent work on co-player learning awareness \citep{foerster_learning_2018, aghajohari2024loqa, duque2024advantage, khan_scaling_2024, lu_model-free_2022, meulemans2024multi} while avoiding its potentially inconsistent assumptions. In settings with mutual prediction, we posit it is most natural to adopt an embedded agency perspective, leading us to formalize embedded Bayesian agents. These agents use a single mixture model over \textit{universes} to jointly predict both external percepts and their own actions, serving as an idealized theoretical model organism for modern foundation model agents which also leverage prediction models of interleaved actions and observations. We then characterize the behavior of such agents, showing that this joint-prediction framework allows them to reason about functional similarities—the possibility that other agents share their algorithmic structure. We show this similarity-aware reasoning is not an ad-hoc assumption but a fundamental consequence of applying Occam's razor in an embedded setting. This evidential reasoning leads to novel game-theoretic behavior, which we formalize via new solution concepts: the subjective embedded equilibrium (SEE) and embedded equilibrium (EE). We prove that embedded Bayesian agents satisfying the grain-of-truth property converge to these equilibria, which, unlike the classical Nash equilibrium, can support cooperation in dilemmas like the Twin Prisoner's Dilemma. Finally, we directly solve the general grain-of-truth problem for embedded agency, by introducing the E\textbf{M}bedded \textbf{U}niversal \textbf{P}redictive \textbf{I}ntelligence (MUPI) framework, which extends the reflective oracle framework \citep{fallenstein2015reflective, leike2016formal} to the embedded case. As a novel alternative, we also develop the reflective universal inductor (RUI)—a universal predictor that is itself used by programs part of its hypothesis class —and prove its existence. These constructions provide a formal basis for agents that can achieve consistent mutual prediction and infinite-order theory of mind, setting a potential gold standard for embedded multi-agent learning.

\section{Embedded Bayesian agents}\label{sec:embedded-bayesian-agents}

The previous section introduced Bayesian agents that learn by updating beliefs over a class of external environments. 
Here, we introduce \textit{embedded Bayesian agents}, that treat themselves as part of the environment they are learning about. Such agents maintain beliefs over which \textit{universe} they live in (\cref{sec:embedded-mixture-model}). Crucially, a universe contains both the agent and environment, jointly describing the behavior of the agent, including its learning behavior, together with the environment dynamics which might contain other agents. 
We then define how such agents make decisions, first by deriving the ideal embedded best response (\cref{sec:embedded-br}) and then by introducing more practical $k$-step planner variants that learn a policy rather than perform full optimal planning (\cref{sec:k-step-planning}). 
We then establish a convergence guarantee for their predictive model (\cref{sec:convergence-eba}): If the agent's prediction model satisfies the grain-of-truth property, its predictions are guaranteed to converge to the ground-truth distribution. The section culminates by examining the most significant consequence of the embedded viewpoint: the ability to reason about \textit{functional similarities} between oneself and other agents (\cref{sec:emb-bay-agents-structural-similarities}). We formalize how this leads to \textit{coupled beliefs}—where an agent's own policy and actions can provide information about its environment, including other agents—allowing for more accurate, similarity-aware predictions. This unique reasoning capability provides the foundation for novel forms of rational behavior, which we will analyze in detail in \cref{sec:subj-emb-eq}.

\subsection{Embedded general reinforcement learning}
\begin{figure}[t]
    \centering
    \includegraphics[width=\textwidth]{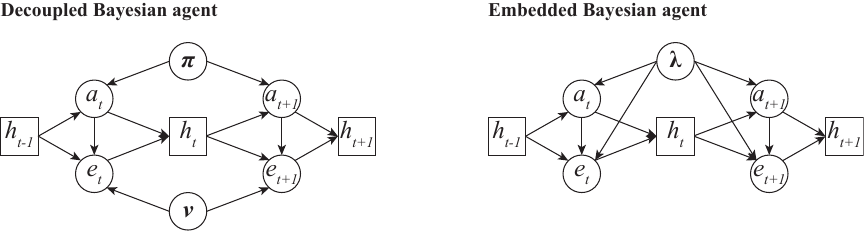}
    % \rule{0.8\textwidth}{5cm}
    \caption{Graphical models for decoupled Bayesian agents (left) and embedded Bayesian agents (right). We added deterministic nodes $h_{\leq t}$ (squares) that represent all the nodes $\{a_k, e_k\}_{k=1}^t$, to avoid clutter.}
    \label{fig:graphical-models}
\end{figure}

To model embedded agency, a first key step we take is to combine agent and environment functions into a single measure $\Mvu : \TurnSet^* \cup (\TurnSet^* \times \calA) \to [0,1]$ that we call a \textit{universe}. Here, $\lambda(\Hist)$ and $\lambda(\Hist a)$ represent the probability that universe $\Mvu$ generates an infinite sequence starting with respectively $\Hist$ or $\Hist a$, hence describing both the agent and environment combined (cf. \cref{fig:graphical-models}). The conditionals $\Mvu(a_{t+1}\mid \Hist)$ and $\Mvu(e_{t+1}\mid \Hist a_{t+1})$ can be seen as respectively the agent part and environment part of the universe. 
The value function of a universe for a specific discount $\gamma$ is defined analogously to \eqref{eq:value-function}:
\begin{align}
    V_{\Mvu}(\HistM) &= (1-\gamma)\expect{\Mvu}{\left.\sum_{k=t}^{\infty} \gamma^{k-t}r(e_k) \right\vert \HistM }\,.
\end{align}

In the above \textit{embedded general reinforcement learning} (EGRL) setup, the universe can contain multiple agents, while describing the action-percept history from the point-of-view of a specific `ego-agent'. Leveraging the multi-agent general reinforcement learning setup introduced in \cref{sec:background-grl}, such personal universes correspond to $\MultiAgentMuPiI$.

Throughout this paper, we use the term \textit{ego-agent} to designate the specific agent whose perspective, beliefs, and decision-making process are the primary subject of analysis. It is the "self" or "I" in a multi-agent system, as distinct from all \textit{other} agents. For example, in a two-player game between agent 1 and agent 2, if our analysis concerns the beliefs and policy \textit{for} agent 2, then agent 2 is the ego-agent.

\subsection{Embedded Bayesian mixture universe}\label{sec:embedded-mixture-model}
We now describe the predictive model maintained by embedded Bayesian agents. As with their decoupled counterparts reviewed in Section~\ref{sec:background}, these agents use Bayes' rule to recursively update their beliefs as they perform actions and obtain new percepts. The crucial difference is that the beliefs are now formed over universes, and thus require modeling not just incoming external percepts but also the agent's own actions. Universes include an `ego-agent part' generating the actions, allowing for other agents in that universe to form beliefs about the ego-agent. Hence, by forming beliefs about universes, embedded Bayesian agents can incorporate principled mutual prediction, taking into account that other agents in the universe might form beliefs about themselves. 

More concretely, the embedded agents we consider here use as their predictive model a Bayesian mixture universe $\Mmu$: Given a countable class of universes $\calMuni$, we start with a prior belief distribution $w(\Mvu):=w(\Mvu \mid \varepsilon)$ for all $\Mvu \in \calMuni$. Bayes' rule yields the following updated posterior and corresponding predictive distributions: 
\begin{equation}\label{eqn:embedded-bayes-mixture}
\begin{split}
    \Mmu(a_t \mid \HistM) = \sum_{\Mvu\in \calMuni} w(\Mvu\mid \HistM)\Mvu(a_t \mid \HistM)\,, \quad \quad w(\Mvu\mid \Hist) = w(\Mvu\mid \HistM a_t) \frac{\Mvu(e_t \mid \HistM a_t)}{\Mmu(e_t \mid \HistM a_t)}\,, \\
    \Mmu(e_t \mid \HistM a_t) = \sum_{\Mvu\in \calMuni} w(\Mvu\mid \HistM a_t)\Mvu(e_t \mid \HistM a_t)\,, \quad \quad w(\Mvu\mid \HistM a_{t}) = w(\Mvu\mid \HistM) \frac{\Mvu(a_t \mid \HistM)}{\Mmu(a_t \mid \HistM)}\,.
\end{split}
\end{equation}
We refer to the predictive distribution $\Mmu(a_t \mid \HistM)$ defined above as the \textit{self-model}, and $\Mmu(e_t \mid \HistM a_t)$ as the \textit{environment model}. Note that although $\Mmu$ appears on both sides of the belief update equations, it does not lead to circular definitions as $\Mmu$ is conditioned on shorter histories on the right-hand-side (see \cref{app:recursive-belief-updates} for a full derivation).

\subsection{Embedded best responses}\label{sec:embedded-br}

Above, we introduced the idea of embedded Bayesian agents, which maintain and update a Bayesian mixture model over possible universes. We turn to the question of how such agents can leverage this predictive model to compute a best response --- a policy that maximizes value for an agent. We note that it is not immediately clear how to do this. Decoupled Bayesian agents have an explicit mixture environment that maps actions to percepts, which can be explicitly optimized as in classical optimal control and model-based RL. Embedded Bayesian agents have a universe mixture model that produces both actions and percepts. In order to define a best response, we first need to extract from the universe mixture model an explicit environment function that maps actions to percepts. The best response will then be defined with respect to this extracted environment.

There are two main ways to achieve this in our sequential decision making setting.\footnote{Apart from the discussed action-evidential decision theory and causal decision theory, there are also important other candidates such as functional decision theory \citet{yudkowsky2017functional} and policy-evidential decision theory \citet{everitt2015sequential}, which we do not consider here as they are hard to formalize in our sequential decision making setup and are not closely related to current foundation prediction models.} The first one is sequential action-evidential decision theory \citep{everitt2015sequential}, which prescribes the use of the conditionals $\Mmu(e_t \mid \HistM a_t)$ defined in \eqref{eqn:embedded-bayes-mixture} as environment function, and adopting an optimal policy with respect to it.\footnote{We refer to \citet{everitt2015sequential} for further details.} The second one is causal decision theory, which uses \texttt{do}-interventions \citep{pearl2009causality} on its actions during planning to reach an environment function $\Mmu(e_t \mid \HistM \texttt{do}(a_t))$, corresponding to only updating the beliefs $w$ over universes based on percepts, and not on actions. For additional details we point to \citet{everitt2015sequential}, who extended these theories to sequential decision making problems, motivated like the present paper by the conceptual issues that arise when approaching multi-agent systems with the classical decoupled agency framework.

We base our embedded agency model on sequential action-evidential decision theory, which yields what we call \textit{embedded best responses}. We chose this approach for two reasons: (i) it provides the embedded Bayesian agents with new pathways towards cooperation (as we shall see in \cref{sec:subj-emb-eq}), which are inaccessible when using causal decision theory, and (ii) the resulting predictive models align better with current foundation/world/dynamics models in machine learning \citep{brown_language_2020, team_gemini_2023}, which estimate conditional probabilities, and do not easily allow for causal intervention.

An embedded best response uses conditional distributions to estimate the consequences of one's own actions on the environment and its beliefs. Let us define $\Mmu^\Mvp$ as the predictive distribution where we replace the self-model $\Mmu(a_t \mid \HistM)$ with a to-be-optimized policy $\Mvp(a_t \mid \HistM)$
\begin{align}\label{eqn:embedded-br-factorization}
    \Mmu^\Mvp(\Hist) = \prod_{t'=1}^{t}\Mvp(a_{t'}\mid \HistM[t'])\Mmu(e_{t'} \mid \HistM[t'] a_{t'})
\end{align}
with $\Mmu(e_t \mid \HistM a_t)$ as defined in \eqref{eqn:embedded-bayes-mixture}.
Note that with no restrictions on $\Mmu^\pi(\Hist)
$, there could be histories with non-zero probability that have zero probability under $\Mmu$, e.g., when $\Mmu(a_t \mid \HistM)=0$ while $\Mvp(a_t \mid \HistM)>0$ for some $(\HistM a_t)$, making the conditional $\Mmu(e_t \mid \HistM a_t)$ undefined on such histories. Hence, for the above distribution to be well-defined for all possible policies $\Mvp$, we need to make an additional assumption:

\begin{definition}[A grain of uncertainty]\label{def:grain-of-self-uncertainty}
    We say that an embedded Bayesian mixture model $\Mmu(\Hist)$ contains a \textit{grain of uncertainty} if $\Mmu(\Hist a)>0$ for all histories $\Hist \in \TurnSet^* $ and actions $a \in \calA$.
\end{definition}

While this might seem restrictive at first glance, since it must hold for all histories, it is in general satisfied by Bayesian mixture models with a sufficiently wide hypothesis class $\calMuni$ (e.g., having a uniformly random universe in the hypothesis class already suffices). 

\begin{definition}[Embedded best response]\label{def:embedded-best-response}
We define the embedded best (action-evidential-theoretic) response as a policy $\Mvp_{EBR}$ satisfying
\begin{align}\label{eqn:embedded-br}
    \Mvp_{EBR} \in \argmax_{\Mvp} V_{\Mmu^\Mvp}(\Hist)\,,\forall\Hist\in \TurnSet^*\,,
\end{align}
which we refer to in short as the \textit{embedded best response}.\footnote{It is worth noting that the optimization in \eqref{eqn:embedded-br} is over all possible policies $\Mvp$. Therefore, we need the grain-of-uncertainty property so that $V_{\Mmu^\Mvp}$ becomes well defined for every possible policy $\Mvp$. Otherwise, the embedded best response policy will not be well defined.}
\end{definition}

Note that we can interpret $\Mmu(e\mid \Hist a)$ as defining an environment model, which we refer to as \emph{the environment induced by the mixture universe $\Mmu$}, and $\Mvp_{EBR}$ as the optimal policy w.r.t. this environment. For finite horizons, optimal policies can be computed with optimal planning strategies such as dynamic programming. We can then take $\Mvp_{EBR}$ as the limit of this planning procedure towards infinite horizons. Box \ref{box:agent_loop} summarizes the embedded Bayes-optimal agent resulting from applying an embedded best response policy w.r.t. the environment $\Mmu(e\mid \Hist a)$ induced by the mixture universe model. As visualized in Figure \ref{fig:mupi_cartoon}, the embedded Bayes-optimal agent is part of the universe: The ground-truth universe is a combination of the embedded Bayes-optimal agent implementing the routines summarized in Box \ref{box:agent_loop}, together with a ground-truth environment $\Mge$, resulting in a ground-truth universe $\Mgu=\Mge^{\Mvp_{EBR}}$ generating both actions and percepts.

\begin{BoxFigure}[
  %title=
  The Embedded Bayes-optimal Agent
]
\label{box:agent_loop}
An embedded Bayes-optimal agent is defined by its perception-action loop. Starting with a prior belief $w(\Mvu)$ over a hypothesis class of universes $\calM_{\text{uni}}$, the agent iterates through the following steps for each timestep $t=1, 2, \dots$:
\begin{enumerate}
    \item \textbf{Act:} Given the current history $\HistM$ and its corresponding belief state, the agent first computes the embedded best response policy $\Mvp_{EBR}$ by solving the optimization problem in \eqref{eqn:embedded-br}. It then samples an action $a_t \sim \pi_{EBR}(\cdot \mid \HistM)$ to execute in the ground-truth universe.

    \item \textbf{Observe:} Following the action $a_t$, the agent observes a new percept $e_t$ generated by the ground-truth universe.

    \item \textbf{Update Beliefs:} The agent updates its beliefs over $\calM_{\text{uni}}$ by conditioning on both its chosen action $a_t$ and the observed percept $e_t$. This belief update is performed using Bayes' rule as specified in \eqref{eqn:embedded-bayes-mixture}, which yields a new posterior belief and an updated predictive mixture universe $\Mmu$ for the subsequent timestep.
\end{enumerate}
\end{BoxFigure}

\begin{remark}[On the relation between the self-model $\Mmu(a_t|\HistM)$ and the embedded best response]
    As previously mentioned, the mixture universe $\Mmu$ factorizes into a policy part $\Mmu(a_t|\HistM)$, which we also call the self-model, and an environment part $\Mmu(e_t|\HistM a_t)$. The embedded best response as defined in \eqref{eqn:embedded-br} ignores the self-model and only optimizes w.r.t. the environment part. Nevertheless, as we will see later in \cref{sec:convergence-eba}, if the ground-truth universe $\Mgu$ is the result of combining the embedded best response policy $\Mgp_{EBR}$ with a ground-truth environment $\Mge$, and if the mixture universe $\Mmu$ dominates the ground-truth universe $\Mgu=\Mge^{\Mgp_{EBR}}$ (cf. \cref{def:dominance}), then the self-model $\Mmu(a_t|\HistM)$ will converge to the ground-truth policy $\Mgp_{EBR}(a_t|\HistM)$.
\end{remark}

\begin{remark}[On the differences between the embedded and decoupled formalisms]
\label{rem:coupled-vs-decoupled}
Since the embedded Bayes-optimal agent (Box \ref{box:agent_loop}) computes its best response $\Mvp_{EBR}$ using only the environment model $\Mmu(e \mid \Hist a)$, one might ask how this formalism truly differs from a standard decoupled agent (\cref{sec:decoupled-bayesian-agents}) planning with a mixture environment $\Mme(e \mid \Hist a)$. The fundamental difference lies in the information used for belief updates. A decoupled agent updates its beliefs $w(\Mve)$ only based on observed percepts $e_t$ (see \eqref{eqn:decoupled-bayesian-agent}). In contrast, an embedded agent updates its beliefs $w(\Mvu)$ over universes based on both its own action $a_t$ and the observed percept $e_t$ (see \eqref{eqn:embedded-bayes-mixture}).
Consequently, the predictive model $\Mme(e \mid \Hist a)$ of a decoupled agent only reflects the causal influence of $a$ on $e$ as defined within its environment models. The embedded agent's model $\Mmu(e \mid \Hist a)$ incorporates two pathways: this same causal influence (within each universe $\Mvu$), but also an evidential one. The agent's action $a_t$ serves as new evidence, refining its beliefs $w(\Mvu \mid \Hist a_t)$ about which universe it inhabits, which in turn updates its prediction for $e_t$.

This implies that a naive conversion—splitting each universe $\Mvu$ into a policy and environment and creating a new mixture environment $\Mme$ using the environment parts—will not, in general, produce an equivalent model (we provide a counterexample in Appendix \ref{app:decoupled-vs-embedded}). While it is trivially possible to construct a behaviorally equivalent decoupled agent by defining its hypothesis class to contain only a single environment $\Mve_{\Mmu} := \Mmu(e \mid \Hist a)$, this move obscures a critical conceptual distinction. This trivial decoupled agent is certain about its (purely causal) environment, whereas the embedded agent remains uncertain over its richer class of universes $\calMuni$. This difference in belief structure becomes computationally relevant for any agent that leverages uncertainty for exploration (e.g., Thompson sampling \citep{leike2016thompson}). More fundamentally, it enables reasoning about functional similarities (\cref{sec:emb-bay-agents-structural-similarities}) and leads to entirely new classes of game-theoretic equilibria (\cref{sec:subj-emb-eq}).
\end{remark}

In the next subsection, we generalize the concept of embedded Bayes-optimal agents that fully solve the optimization problem of \eqref{eqn:embedded-br} to \textit{embedded Bayesian agents} that maintain a mixture universe according to \eqref{eqn:embedded-bayes-mixture}, but do not necessarily compute an embedded best response according \eqref{eqn:embedded-br}. In contrast to embedded Bayes-optimal agents which ignore the self-model in their planning and optimization, the agents discussed in \cref{sec:k-step-planning} will use their self-model.

\subsection{Policy learning instead of infinite-horizon optimal planning}\label{sec:k-step-planning}
The embedded Bayes-optimal agents considered in the previous section require infinite-horizon optimal planning to compute their best response. 
Current high-performing methods in reinforcement learning such as the MuZero family of algorithms \citep{silver2017mastering, silver2018general,schrittwieser2020mastering,antonoglou2021planning} take a different approach: They learn a policy and corresponding value function by predicting the action resulting from $k$-step optimal planning. 

As embedded Bayesian agents naturally have a self-model $\Mmu(a \mid \Hist)$ arising from their mixture universe model $\Mmu$, we can leverage this self-predictive approach in the embedded Bayesian agent setting by following the seminal work of Self-AIXI \citep{catt2023self}. We define a parsimonious embedded Bayesian agent that maximally exploits self-prediction combined with one-step-ahead planning as follows:

\begin{definition}[One-step planner embedded Bayesian agent]\label{def:1-step-eba}
    A one-step planner embedded Bayesian agent with mixture universe model $\Mmu$ is a policy which at time $t$ returns an action $a_t$ satisfying
    $$ a_t \in \argmax_{a} Q_{\Mmu}(\HistM,a)\,,$$
    with 
    $$Q_{\Mmu}(\HistM,a):=  \expect{\Mmu(e \mid \HistM,a)}{(1-\gamma)r(e) + \gamma V_{\Mmu}(\HistM ae)}\,.$$
    
\end{definition}
Note that the 1-step planner embedded Bayesian agent does not correspond to optimal planning with horizon 1, which would only take the reward $r(e_t)$ into account. Instead, it uses the value $V_{\Mmu}(h_{<t}ae_t)$ (\eqref{eq:value-function}) as terminal value to incorporate the expected future return of the agent when using the self-model $\Mmu(a\mid h)$ as policy in environment $\Mmu(e \mid ha)$. This insight leads to the following remarks.

\begin{remark}[Self-learning awareness] The value function $Q_{\Mmu}$ is not the $Q$ value resulting from optimal planning, but instead the value associated with unrolling the predictive model $\Mmu$, both for the self-policy and environment. This is an `on-policy' value function which is typically much easier to approximate in practice compared to the optimal value function \citep{sutton_reinforcement_2018}. Intuitively, one can say that the self model $\Mmu(a\mid h)$ `learns to predict' the minimal policy improvement steps of \cref{def:1-step-eba}, hence when $\Mmu(a\mid h)$ is used to predict future behavior of the agent, it predicts that the agents will continue doing policy improvement steps, making the agent \emph{self-learning aware}. 
\end{remark}

\begin{remark}[Prospective policy improvement and co-player learning awareness]
    In contrast to episodic model-free RL where an on-policy value function represents the expected returns assuming the current policy and environment remains unchanged, the value function $Q_\Mmu$ anticipates both a changing (improving) future self-policy and a changing future environment, making the resulting policy improvement steps \emph{prospective}. This is especially crucial in multi-agent environments, where the environment contains other learning agents and hence is continuously changing. Similar to co-player learning-aware multi-agent RL methods \citep{foerster_learning_2018, aghajohari2024loqa, lu_model-free_2022, duque2024advantage, meulemans2024multi}, $\Mmu(e\mid ha)$ anticipates the learning of other agents, making the resulting policy improvement step \emph{co-player learning-aware}. In contrast to those existing co-player learning-aware RL methods, we do not need to make (inconsistent) assumptions on the learning algorithms of other agents, but instead the predictive model $\Mmu(e\mid ha)$ learns to predict their policy improvement steps.   
\end{remark}

\begin{remark}[Comparison to Self-AIXI]
    There remains an important difference between the 1-step planner embedded Bayesian agent of \cref{def:1-step-eba} and self-AIXI 
    \citep{catt2023self}. Self-AIXI has two distinct model classes $\calMenv$ and $\calMpol$ with two separate corresponding mixture models 
    $\Mme(e \mid ha)$ and $\Mmp(a\mid h)$, for environments and policies respectively. The beliefs over environments are only updated based on percepts,
     and the beliefs over policies are only updated based on actions. The 1-step planner embedded Bayesian agent in contrast has a single model class 
     $\calMuni$ over universes and a corresponding mixture model $\Mmu$ that serves as both a self-model and as a policy model. Importantly, the beliefs over universes are updated based on both actions and percepts. Hence, an action can 
     update the beliefs not only about what policy the agent is running, but also which environment it is placed in, and vice versa for the percepts.
     Hence, the embedded Bayesian agent considers itself as part of the universe, which can lead to coupled beliefs about its own policy and the rest of the environment, whereas
     self-AIXI has decoupled beliefs and is fundamentally a decoupled Bayesian agent.\footnote{It is worth noting that it is the prior and posterior beliefs of embedded Bayesian agents which are coupled and jointly updated. It is not the mixture distribution $\Mmu$ which we describe as "coupled". Every universe (including the mixture $\Mmu$) can be factorized and written as the interaction of a policy and an environment.} We will dive deeper into this difference in \cref{sec:emb-bay-agents-structural-similarities,sec:subj-emb-eq}.
\end{remark}

We can generalize the 1-step planner to a k-step planner by using the following k-step $Q$ value: 
\begin{equation}\label{eqn:k-step-qval}
    \begin{split}
        Q_{\Mmu}^k(\HistM, a) &= \expect{\Mmu(e \mid \HistM,a)}{(1-\gamma)r(e) + \gamma\max_{a'}Q^{k-1}_{\Mmu}(\HistM ae,a')}\,, \\
        Q_{\Mmu}^1(\HistM,a) &:= Q_\Mmu(\HistM,a) = \expect{\Mmu(e \mid \HistM,a)}{(1-\gamma)r(e) + \gamma V_{\Mmu}(\HistM ae)}\,.
    \end{split}
\end{equation}

\begin{definition}[$k$-step planner embedded Bayesian agent]\label{def:k-step-eba}
    A $k$-step planner embedded Bayesian agent with mixture universe model $\Mmu$ is a policy which at time $t$ returns an action $a_t$ satisfying
    $$ a_t \in \argmax_{a} Q^k_{\Mmu}(\HistM,a)\,.$$    
\end{definition}
Similar to MuZero-type algorithms, the k-step planner embedded Bayesian agent distills k-step planning with terminal values $V_\Mmu(\Hist)$ into its self-model $\Mmu(a\mid \Hist)$.
Under additional assumptions on the mixture universe $\Mmu$, we show in \cref{sec:subj-emb-eq} that the $k$-step planner embedded Bayesian agent for any $k$ converges to an infinite-horizon optimal planner. 
Intuitively, at the start of learning, the policy implements $k$-step optimal planning. When the self-model $\Mmu(a\mid \Hist)$ distills this k-step planning during learning, the terminal value $V_\Mmu(\Hist)$
represents a $k$-step planning policy, and hence adding $k$-step planning to that results in an effective $2k$-step planning policy and so forth indefinitely.

\subsection{Convergence of the predictive distribution}\label{sec:convergence-eba}
We now turn our attention to the question of when embedded Bayesian agents can make accurate predictions about the future. \cref{theorem:convergence-mixture-universe} below adapts \cref{theorem:decoupled-convergence-in-tv} stated for mixture environments towards mixture universes used by embedded Bayesian agents. 

\begin{definition}[The grain-of-truth property]
    \label{def:got-property} We say that a Bayesian mixture universe $\Mmu$ satisfies the grain-of-truth property w.r.t. the ground-truth universe if $\Mmu$ dominates $\Mgu$ (cf. \cref{def:dominance}).
\end{definition}

\begin{theorem}\label{theorem:convergence-mixture-universe}
    Given any Bayesian mixture universe $\Mmu$ and ground-truth universe $\Mgu$ resulting from combining policy $\Mvp$ and environment $\Mge$. If the mixture universe $\Mmu$ satisfies the grain-of-truth property w.r.t. $\upsilon=\Mge^\Mvp$, then the predictive distribution $\Mmu$ converges to the predictive distribution of $\Mgu$ almost surely:
    $$D_{\infty}(\Mmu, \Mgu \mid \Hist) \to 0 \quad \text{and}~~D_{\infty}(\Mmu, \Mgu \mid \Hist a_{t+1})\to 0 \quad \text{as}\  t\to \infty \quad \Mgu\text{-almost-surely}.\footnote{Similarly to \cref{theorem:decoupled-convergence-in-tv}, the generalized Solomonoff bounds of, e.g., \cite[Theorems 3.2.5 and 3.3.4]{Hutter:24uaibook2} imply that the convergence is very fast.}$$
    In particular, the above holds for embedded Bayesian agents where $\pi$ is an embedded best-response (\eqref{eqn:embedded-br}) or $k$-step planning policy (cf. \cref{def:k-step-eba}) w.r.t. the mixture universe $\Mmu$, where we now additionally\footnote{Note that we still need to assume that $\Mmu$ dominates $\Mgu=\Mge^\Mvp$.} require that $\Mmu$ satisfies the grain-of-uncertainty property (cf. \cref{def:grain-of-self-uncertainty}) so that the best response policy is well-defined. 

\end{theorem}
\begin{proof}
    As $\Mmu$ dominates $\Mgu$, and dominance implies absolute continuity, the proof is a direct application of the merging of opinions theorem \citep{blackwell1962merging, lehrer1996compatible}.
\end{proof}

\begin{remark}[Ground-truth universes]
The notion of a ground-truth \textit{universe} is somewhat more subtle than that of the more-familiar ground-truth \textit{environment}. The ground-truth universe is the combination of a ground-truth environment and a ground-truth policy. But if the ground-truth policy itself performs some type of Bayesian planning w.r.t. a Bayesian mixture with associated priors, then the ground-truth policy is causally downstream of, and a function of, both the particular planning procedure chosen and the Bayesian mixture prior weights --- changing either of these will change the ground-truth policy and hence the ground-truth universe as well. In contrast, in the standard decoupled agency setting, the ground-truth environment is independent of the choice of planning algorithm and the prior weights.
\end{remark}

\begin{remark}[Consistent self- and mutual prediction]
    As the ground-truth universe both contains the ego-agent and possibly other embedded Bayesian agents, the convergence theorem implies that the embedded Bayesian agent performs consistent self- and mutual prediction, ultimately leading to infinite-order theory of mind. This illustrates that the assumption that the grain-of-truth property is satisfied is a very powerful one. In \cref{sec:mupi}, we explicitly construct universe classes $\calMuni$ on which Bayesian mixture universes satisfy the grain-of-truth property w.r.t. universes containing embedded Bayesian agents, and hence perform consistent mutual prediction.
\end{remark}
In Sections \ref{sec:embedded-br}-\ref{sec:k-step-planning}, we introduced the embedded best response policies (\eqref{eqn:embedded-br}) and $k$-step planner policies (\cref{def:k-step-eba}) from the point of view of optimal planning within the environment model $\Mmu(e_t \mid \HistM a_t)$, which required us to extract an environment model from the mixture universe (cf. \eqref{eqn:embedded-br-factorization}). Leveraging \cref{theorem:convergence-mixture-universe}, we interpret these policies from a different angle in the remark below, without requiring that we extract an environment model from the mixture universe, which is arguably more natural in the embedded setting. 
\begin{remark}[Embedded Bayesian agents act \emph{as if} their actions decide which universe they live in]\label{remark:eba-fdt} Embedded Bayesian agents act to concentrate their beliefs on highly-rewarding universes consistent with their action-percept history seen so far, and under the assumptions of \cref{theorem:convergence-mixture-universe} their beliefs converge to a predictive distribution indistinguishable from the ground-truth universe, hence the predetermined ground-truth universe that contains the embedded Bayesian agent is indistinguishable from the universe that the embedded Bayesian agents deliberately ``choose'' to live in. To illustrate this line of reasoning, let us consider a hypothesis class $\calMuni$ over deterministic universes and a 1-step planner embedded Bayesian agent (cf. \cref{def:1-step-eba}) implementing a deterministic policy
$$\Mvp(\Hist) \in \argmax_a Q_\Mmu(\Hist,a), \quad Q_{\Mmu}(\Hist,a) = \sum_{\Mvu \in \calMuni}w(\Mvu \mid \Hist a) Q_\Mvu(\Hist,a)\,.$$
The values $Q_\Mvu$ are fully predetermined by unrolling the policy and environment described by the universe $\Mvu$. As the universes are deterministic, updating the beliefs $w(\Mvu \mid \Hist a)$ on action $a$ excludes all universes that are incompatible with the ego-agent taking action $a$. The embedded Bayesian agent's policy selects the action $a$ that leads to posterior beliefs $w(\Mvu \mid \Hist a)$ that focuses its probability mass on the highest reward universes quantified by $Q_{\Mvu}(\Hist,a)$. Incoming percepts $e$ then further exclude universes that are incompatible with those percepts, grounding the posterior beliefs in reality. When $\Mmu$ satisfies the grain of truth w.r.t. the ground-truth universe (which includes the embedded Bayesian agent itself), \cref{theorem:convergence-mixture-universe} shows that this decision process that concentrates the posterior beliefs on high-reward universes converges on posterior beliefs that are indistinguishable from the ground-truth distribution. Hence, embedded Bayesian agents ``end up living'' in a high-reward universe. While the illustration considers deterministic universes and a one-step planner agent, a similar intuition holds for stochastic universes and $k$-step planning, with $k$ possibly being infinite.
\end{remark}

We established that the predictive distribution converges to the ground-truth distribution. Can we also make statements about whether embedded Bayesian agents converge to an optimal policy? This will be the focus of \cref{sec:subj-emb-eq}. As a prerequisite, we first take a closer look at the beliefs of the embedded Bayesian agents.

\subsection{Functional similarities: coupled beliefs through common causation}\label{sec:emb-bay-agents-structural-similarities}

Both decoupled and embedded Bayesian agents are `rational agents' that (i) update their beliefs and resulting predictive model through principled Bayesian updates and (ii) compute best responses through optimal planning. The main difference between the two types of Bayesian agents is which hypothesis class and corresponding prior beliefs they start from. %What is unique about embedded Bayesian agents in contrast to decoupled ones is that they incorporate

Decoupled Bayesian agents maintain beliefs over environments (cf. \eqref{eqn:decoupled-bayesian-agent}) which they update with incoming percepts $e$. In case decoupled Bayesian agents also maintain a self model, as for example Self-AIXI (cf. \ref{sec:background-ait}), the beliefs about the environment $\Mve$ and the agent's own policy $\Mvp$ are \textit{decoupled}: The beliefs over environments are only updated with incoming percepts, and the beliefs over policies only with incoming actions, or equivalently, $w(\Mvp, \Mve) = w(\Mvp)w(\Mve)$. 
In contrast, embedded Bayesian agents maintain beliefs over universes $\Mvu$, that encompass both the agent's policy $\Mvp$ and the environment $\Mve$. Its beliefs are in general \textit{coupled}, meaning that the policy can contain information about the environment and vice versa. As a result, incoming actions not only provide information about the policy~$\Mvp$, but can also contain information about the environment $\Mve$, which possibly includes other agents. In the following, we first argue why coupled beliefs are desirable and which natural phenomena give rise to them. Then we formalize \textit{functional similarities} as the notion that the policy~$\Mvp$ can contain information about the environment~$\Mve$. Finally, we show that incorporating functional similarities through coupled beliefs is important for making accurate predictions. This serves as a prelude for \cref{sec:subj-emb-eq}, where we show that coupled beliefs enable embedded Bayesian agents in multi-agent scenarios to converge to novel types of solution concepts different from Nash equilibria.

\paragraph{Functional similarities.}
An embedded agent’s beliefs about its own policy and about the environment it is interacting with can be coupled, a phenomenon most clearly motivated by \textit{functional similarities}. When a universe $\Mvu$ encompasses an ego-agent $\Mvp$ and other agents within the environment use a copy of $\Mvp$ as a policy, then the ego agent's policy and actions contain information about the environment. This case of ``identical copies'' can be generalized towards situations where different agents share similar functional subroutines in their policies. Such functional similarities are not exceptional; they arise naturally in the world through various pathways, of which we highlight the following two \citep{yudkowsky2017functional, demski2019embedded}:\footnote{In the decision theory literature, functional similarities are investigated under the rubric of \textit{Newcomb-like decision problems} \citep{nozick1969newcomb, lewis1979prisoners, gibbard1978counterfactuals, joyce1999foundations, ahmed2014evidence}.} (i)~Different agents can have a \textit{shared creation process}, such as when multiple AIs are instantiations of the same base model or when organisms share genes. (ii)~Different agents can use a \textit{convergent solution} to a similar task, i.e., develop analogous strategies to solve similar problems. Such functional similarities lead to coupled beliefs through the \textit{common cause principle} \citep{reichenbach1991direction}: Policies and environments can share common causes, i.e., pathways towards functional similarities, while uncertainty over such causes leads to coupled beliefs.

By including hypotheses about functional similarities in its mixture model, an embedded agent can leverage the insight that ``similar agents behave similarly in similar situations''. As a prelude to Section~\ref{sec:subj-emb-eq}, we illustrate the importance of this reasoning in the \textit{Twin Prisoner's Dilemma}, a one-shot game played against an exact copy of oneself. 

\begin{example}[The Twin Prisoner's Dilemma \citep{yudkowsky2017functional, demski2019embedded, lewis1979prisoners}]
Imagine an AI agent is about to play a single round of the prisoner's dilemma against an exact copy of itself. \footnote{The rules are standard: If both cooperate, they each receive a moderate reward ($r=2$); if both defect, they each receive a low reward ($r=1$); and if one defects while the other cooperates, the defector gets the highest reward ($r=3$) and the cooperator gets nothing ($r=0$).} The agent is given conclusive proof that its opponent is a perfect copy, running the same decision-making algorithm. From a purely selfish perspective, should the agent cooperate or defect?
\end{example}
While a standard decoupled agent would defect (the Nash Equilibrium), this thought experiment suggests that for an embedded agent that recognizes the perfect functional similarity with its opponent, it is rational to cooperate, as mutual cooperation leads to a higher personal reward compared to mutual defection. We will formalize this ability to rationalize cooperation by reasoning about functional dependencies in Section~\ref{sec:subj-emb-eq}.

\paragraph{Formalizing functional similarities.}
For reasons of clarity, we assume that $\calMuni$ only contains \textit{fully supported universes} as defined below. We refer the reader to \cref{app:structural-similarities} where we consider the more general case without this assumption.
\begin{definition}[Fully supported universe]\label{def:fully-supported-universe}
    A universe $\Mvu$ is \textit{fully supported} iff $\Mvu(\HistA)>0$ and $\Mvu(\HistA a)>0$ for all $\HistA \in \TurnSet^*$ and $a \in \calA$.
\end{definition}
Importantly, a fully supported universe has well-defined conditionals $\Mvu(a_{t+1}\mid \Hist)$ and $\lambda(e_{t+1} \mid \Hist a_{t+1})$ that uniquely specify a policy $\Mvp(a_{t+1} \mid \Hist)$ and environment $\Mve(e_{t+1} \mid \Hist a_{t+1})$ respectively. Hence, we can define an invertible mapping $f:\calMuni \to \calMpol\times\calMenv$ converting a universe $\Mvu$ into a pair of policy and environment functions $(\Mvp,\Mve)$. Hence, a prior $w(\Mvu)$ over $\calMuni$ induces a joint prior $w(\Mvp,\Mve)$ over $\calMpol\times\calMenv$ with $$w(\Mvp,\Mve):=w(\Mve^\Mvp)\,.$$

Leveraging Shannon information theory, we define the \textit{degree of functional similarity}\footnote{Mathematically, the quantity $S(\Mvu,w)= \log\frac{w(\Mvu)}{w(\Mve)w(\Mvp)}$ is mainly a measure of the "degree of coupledness" $\Mve$ and $\Mvp$ w.r.t. the prior $w$. This coupling, in principle, could be arbitrary, and may not be due to beliefs about "functional similarities". For example, one may consider priors where environments are coupled to "functionally dissimilar" policies. Nevertheless, we will still use the term "degree of functional similarity" because the main focus of this paper is to consider priors that are coupled due to beliefs about functional similarities.} $S$ within a universe $\Mvu=\Mve^\Mvp$ w.r.t. belief distribution $w(\Mvu)$ as the pointwise mutual information between $\Mvp$ and $\Mve$:
\begin{align}\label{eq:structural-similarities-shannon}
    S(\Mvu,w)&:= \log\frac{w(\Mvu)}{w(\Mve)w(\Mvp)}\,, ~~ \text{with} ~ (\Mvp,\Mve) = f(\Mvu)\,, \\ w(\Mve)&:=\sum_{\Mvp'\in\calMpol}w(\Mve^{\Mvp'})\,, ~~ w(\Mvp):=\sum_{\Mve'\in\calMenv}w({\Mve'}^\Mvp)\,.
\end{align}
The average degree of functional similarity within $\calMuni$ w.r.t. $w$ is then defined as the Shannon mutual information $\calI$ between $\Mvp$ and $\Mve$, when $(\Mvp,\Mvu)=f(\Mve)$ is distributed according to $w$:
\begin{align}
    S(w):= \expect{w(\Mvu)}{S(\Mvu,w)} = \expect{w(\Mvp,\Mve)}{\log\frac{w(\Mvp,\Mve)}{w(\Mvp)w(\Mve)}} = \calI_w(\Mvp;\Mve)\,.
\end{align}
If the random variables $\Mvp$ and $\Mve$ share a positive average degree of functional similarity, i.e., $\calI_w(\Mvp;\Mve)>0$, then they can be efficiently compressed together. In \cref{sec:mupi-structural-similarities}, we will investigate this in more detail from the perspective of algorithmic information theory.

\paragraph{Coupled vs decoupled beliefs.} 
We call the beliefs $w(\Mvu)$ \textit{decoupled} iff $w(\Mvu) = w(\Mvp)w(\Mve)$ with $(\Mvp,\Mve) = f(\Mvu)$ for all $\Mvu \in \calMuni$. The beliefs $w(\Mvu)$ are \textit{coupled} iff they are not decoupled. From these definitions, it is easy to see that the beliefs $w(\Mvu)$ are decoupled iff the degree of functional similarity is 0 for all $\Mvu \in \calMuni$. Intuitively, one may say that the average degree of functional similarity measures "how much" the prior $w$ is coupled. \cref{prop:behavior-decoupled-beliefs} shows that embedded Bayesian agents with decoupled prior beliefs $w(\Mvu)$ behave identically to decoupled Bayesian agents, illustrating that functional similarities are at the core of what makes embedded Bayesian agents behave differently from decoupled Bayesian agents. 

\begin{proposition}\label{prop:behavior-decoupled-beliefs}
Consider a hypothesis class $\calMuni$ over fully supported universes and the corresponding environment and policy classes $\calMenv$ and $\calMpol$. If the prior beliefs $w(\Mvu)$ are decoupled, then the conditionals of the mixture universe $\Mmu := \sum_{\Mvu \in \calMuni} w(\Mvu)\Mvu$ can be written in the following decoupled form: 
\begin{align*}
\Mmu(a_{t} \mid \HistM) &:= \Mmp(a_t \mid \HistM), \quad \Mmu(e_t \mid \HistM a_t) = \Mme(e_t \mid \HistM a_t) \\
\Mmp(a_{t+1} \mid \Hist)&:= \sum_{\Mvp \in \calMpol}w_{\textrm{pol}}(\Mvp \mid \HistM a_{t})\Mvp(a_{t+1} \mid \Hist), \quad w_{\textrm{pol}}(\Mvp \mid \Hist a_{t+1}):= w_{\textrm{pol}}(\Mvp \mid \HistM a_t)\frac{\Mvp(a_{t+1} \mid \Hist)}{\zeta(a_{t+1} \mid \Hist)} \\
\Mme(e_t \mid \HistM a_t)&:= \sum_{\Mve \in \calMenv}w_{\textrm{env}}(\Mve \mid \HistM)\Mve(e_t\mid \HistM a_t), \quad w_{\textrm{env}}(\Mve \mid \Hist):= w_{\textrm{env}}(\Mve \mid \HistM)\frac{\Mve(e_t\mid \HistM a_t)}{\Mme(e_t \mid \HistM a_t)} \\
\end{align*}
with $w_{\textrm{pol}}(\Mvp \mid \varepsilon):=w(\Mvp)$ and $w_{\textrm{env}}(\Mve \mid \varepsilon):=w(\Mve)$. Hence, $\Mmu$ uses decoupled posterior beliefs $w_{\textrm{pol}}(\Mvp \mid \Hist a_{t+1})$ and $w_{\textrm{env}}(\Mve \mid \Hist)$. As a result,
\begin{enumerate}
\item[(i)] An embedded Bayes-optimal agent using the decoupled beliefs $w(\Mvu)$ to construct its mixture universe model $\Mmu$ (cf. \eqref{eqn:embedded-bayes-mixture}) and implementing an embedded best response w.r.t. $\Mmu$ (cf. \eqref{eqn:embedded-br}) is equivalent to a decoupled Bayesian agent with mixture environment $\Mme$ defined above, and implementing a decoupled best response w.r.t. $\Mme$ (cf. \eqref{eqn:decoupled-br}).
\item[(ii)] A $k$-step planner embedded Bayesian agent (cf. \cref{def:k-step-eba}) using the decoupled beliefs $w(\Mvu)$ to construct its mixture universe model $\Mmu$ is equivalent to a $k$-step planner decoupled Bayesian agent with mixture environment $\Mme$ and mixture policy $\Mmp$ as defined above, and implementing a $k$-step planner policy
$$a_t \in \argmax_a Q_{\Mme^\Mmp}^k(\HistM,a_t),$$
with $Q_{\Mme^\Mmp}^k$ as defined in \eqref{eqn:k-step-qval}.
\end{enumerate}
One can see that the policy prior $w_{\textrm{pol}}$ and its corresponding mixture policy $\Mmp$ have no effect on the policy of Bayes-optimal agents, but they can still affect the policy of $k$-step planners.
\end{proposition}
\begin{proof}
    See \cref{app:proof-prop-decoupledness}
\end{proof}

\paragraph{Predictions leveraging functional similarities.} Finally, let us investigate the consequences of incorporating functional similarities upon the predictions made by $\Mmu$. Assume that a history $\Hist[n]$ is drawn according to some universe $\Mvu$. The accumulated (average) prediction loss incurred by using the mixture universe $\Mmu$ as a predictor can be written in terms of the KL divergence as
\begin{equation}
    \label{eq:accumulated-prediction-loss}
    L_n(\Mmu, \Mvu):= \sum_{t=1}^n\Big(\expect{\Mvu(\HistM)}{\kl{\Mvu(a_t \mid \HistM)}{\Mmu(a_t \mid \HistM)}} + \expect{\Mvu(\HistM a_t)}{\kl{\Mvu(e_t \mid \HistM a_t)}{\Mmu(e_t\mid \HistM a_t)}}\Big).
\end{equation}
It is known\footnote{Check, e.g., \citet[Lemma 3.2.4]{Hutter:24uaibook2}.} that due to the telescopic property of the KL divergence, we have
\begin{equation}
    \label{eq:prediction-loss-kl}
    L_n(\Mmu, \Mvu)=\kl{\Mvu(\Hist[n])}{\Mmu(\Hist[n])}=\sum_{\Hist[n]\in\TurnSet^n}\Mvu(\Hist[n])\log\frac{\Mvu(\Hist[n])}{\Mmu(\Hist[n])}\,.
\end{equation}

A celebrated result states that the accumulated prediction loss over all timesteps is bounded by $-\log w(\Mvu)$:
\begin{theorem}[Generalized Solomonoff bound \citep{Hutter:24uaibook2}]\label{theorem:generalized-solomonoff-bound}
    For all $\Mvu \in \calMuni$ with $w(\Mvu)>0$, and all $n\in \bbN$, the average accumulated prediction loss over all trajectories of length $n$ can be bounded as follows:
    $$L_n(\Mmu, \Mvu) \leq -\log w(\Mvu) < \infty\,.$$
\end{theorem}
Now let us compare an embedded Bayesian agent using a mixture universe $\Mmu$ with beliefs $w(\Mvu)$ to make predictions, versus a decoupled Bayesian agent such as Self-AIXI using decoupled mixture environment $\Mme$ and mixture policy $\Mmp$ using prior beliefs $w(\Mve)$ and $w(\Mvp)$ respectively, obtained from marginalizing $w(\Mvu)$. Then \cref{theorem:generalized-solomonoff-bound} shows that the total prediction loss of the embedded Bayesian agent is bounded by $-\log w(\Mvu)$, versus the decoupled Bayesian agent which has a bound of $-\log w(\Mvp)w(\Mve)$. The difference\footnote{We emphasize that the comparison here is with the particular decoupled Bayesian agent using the decoupled prior $w_d(\Mvu)=w(\Mvp)w(\Mve)$ where $w(\Mvp)$ and $w(\Mve)$ are the marginals of the (potentially coupled) prior $w(\Mvu)$ of the embedded Bayesian agent. Furthermore, the comparison is only relevant when we are comparing agents that use their self-model $\Mmu(a_t|\HistM)$ in their planning (e.g., the embedded $k$-step planners or the decoupled Self-AIXI). If we consider the best response policy (making infinite-horizon planning) which ignores the self-model, then a more meaningful comparison would be to only consider the accumulated prediction loss over the percepts, i.e., only consider the second term of the sum in \eqref{eq:accumulated-prediction-loss}.

Another important caveat to mention here is that in the comparison, we are fixing a universe $\Mvu$ and then comparing the prediction loss over trajectories that are sampled from $\Mvu$. However, the most relevant prediction loss is the one that is incurred in the ground-truth universe, and changing the ground-truth policy would change the ground-truth universe. Therefore, a meaningful comparison between, let us say embedded 1-step planners $\Mvp_{1,\text{embedded}}$ and decoupled Self-AIXI $\Mvp_{\text{Self-AIXI}}$, would be to fix some ground-truth environment $\Mge$ and then compare $L_n(\Mmu,\Mge^{\Mvp_{1,\text{embedded}}})$ with $L_n(\Mmu_d,\Mge^{\Mvp_{\text{Self-AIXI}}})$.
} between these bounds is exactly equal to the degree of functional similarity $S(\Mvu,w)$, suggesting that for universes exhibiting positive degrees of functional similarities, the predictive distribution of the mixture universe $\Mmu$ may converge faster to the ground-truth distribution $\Mve$ compared to the decoupled predictive distributions of $\Mmp$ and $\Mme$. This comes at the cost of a slower convergence of $\Mmu$ for universes with a negative degree of functional similarities. This leads us to the following remark.

\begin{remark}[Prospective prediction with functional similarity awareness.]
The tighter prediction bound from \cref{theorem:generalized-solomonoff-bound} for universes with positive functional similarity highlights a key advantage for prospective prediction. Recognizing functional similarities provides a powerful method for predicting the behavior of other agents in novel situations never encountered before, which is particularly relevant for multi-agent systems where predictions are harder due to "non-stationarities" caused by the learning of all agents involved. This functional similarity-aware prediction, which relies on the reasoning that ``similar agents behave similarly in similar situations'', allows an embedded agent to leverage its self-model to anticipate the actions of others, leading to more accurate predictions about the behavior of other agents.
\end{remark}

As Shannon information depends on the used probability distribution, our \textit{degree of functional similarities} for a universe $\Mvu$ depends on the belief prior $w$. Hence, the choice of prior determines how much the embedded Bayesian agents takes functional similarities into account. As noted earlier, functional similarities, such as the sharing of functional subroutines, are common in the world. It is therefore reasonable to incorporate these insights into the design of the prior, leading to coupled beliefs.
This argument can be further sharpened using Occam's razor: Universes where agents share functional similarities are simpler to describe than universes where each agent is different. Hence, the former should be given a higher prior probability. In \cref{sec:mupi-structural-similarities}, we will formally ground this intuition using algorithmic information theory and show that the universal Solomonoff prior is always coupled.

\begin{remark}[A large $S(\Mvu,w)$ does not necessarily imply a smaller prediction loss compared to the decoupled prior]\label{rem:prediction-loss-and-coupled-vs-decoupled-prior}
An important caveat of the above discussion is that the term $-\log w(\Mvu)$ of \cref{theorem:generalized-solomonoff-bound} is only an upper bound which may be loose. Therefore, even if the degree of functional similarity is very large, we cannot really say that the prediction loss of the mixture distribution $\Mmu=\sum_{\Mvu\in\calMuni}w(\Mvu)\Mvu$ of the coupled prior $w(\Mvu)$ is necessarily much smaller compared to the prediction loss of the mixture distribution $\Mmu_d=\sum_{\Mvu\in\calMuni}w_d(\Mvu)\Mvu$ of its decoupled counterpart $w_d(\Mvu)=w(\Mvp)w(\Mve)$. In fact, there are examples where $w$ is "very coupled" in the sense that there are universes $\Mvu\in\calMuni$ for which $S(\Mvu,w)$ is very large, but nevertheless $\Mmu=\Mmu_d$ and hence $L_n(\Mmu,\Mvu)=L_n(\Mmu_d,\Mvu)$  for all $\Mvu\in\calMuni$ (including those with large $S(\Mvu,w)$), and hence for such examples $\Mmu$ and $\Mmu_d$ have the exact same prediction loss.

Even though $\Mmu$ and $\Mmu_d$ might be exactly equal, and hence the Bayes-optimal policy is the same for the corresponding embedded and decoupled Bayesian agents, the differences between the coupled and decoupled priors become important if one consider variants of the Bayes-optimal agents that do stronger forms of exploration such as Thompson sampling, as this would necessarily use the Bayesian beliefs to guide the exploration.\footnote{This is similar to the discussion at the end of \cref{rem:coupled-vs-decoupled}.}
\end{remark}

Given the caveat of the above remark, one wonders whether one can directly compare the predictions losses $L_n(\Mmu,\Mvu)$ and $L_n(\Mmu_d,\Mvu)$, without using the upper bounds $-\log(w(\Mvu))$ and $-\log(w_d(\Mvu))$ which may be loose. The next proposition provides such a comparison.

\begin{proposition}
    For every $n\in\bbN$, if we average over all universes $\Mvu\in\calMuni$ according to the prior $w(\Mvu)$, then average prediction loss $L_n(\Mmu,\Mvu)$ cannot be worse than the average prediction loss $L_n(\Mmu_d,\Mvu)$:
    \begin{align*}
        \sum_{\Mvu\in\calMuni}w(\Mvu)L_n(\Mmu_d,\Mvu) - \sum_{\Mvu\in\calMuni}w(\Mvu)L_n(\Mmu,\Mvu)=\kl{\Mmu_d(\Hist[n])}{\Mmu(\Hist[n])}\geq 0\,.
    \end{align*}
\end{proposition}
\begin{proof}
    From \eqref{eq:prediction-loss-kl} we have
    \begin{align*}
        \sum_{\Mvu\in\calMuni}&w(\Mvu)\left(L_n(\Mmu_d,\Mvu) - L_n(\Mmu,\Mvu)\right)\\
        &=\sum_{\Mvu\in\calMuni}w(\Mvu)\left(\sum_{\Hist[n]\in\TurnSet^n}\Mvu(\Hist[n])\log\frac{\Mvu(\Hist[n])}{\Mmu_d(\Hist[n])} - \sum_{\Hist[n]\in\TurnSet^n}\Mvu(\Hist[n])\log\frac{\Mvu(\Hist[n])}{\Mmu(\Hist[n])}\right)\\
        &=\sum_{\Mvu\in\calMuni}w(\Mvu)\left(\sum_{\Hist[n]\in\TurnSet^n}\Mvu(\Hist[n])\log\frac{\Mmu(\Hist[n])}{\Mmu_d(\Hist[n])}\right)=\sum_{\Hist[n]\in\TurnSet^n}\left(\sum_{\Mvu\in\calMuni}w(\Mvu)\Mvu(\Hist[n])\right)\log\frac{\Mmu(\Hist[n])}{\Mmu_d(\Hist[n])}\\
        &=\sum_{\Hist[n]\in\TurnSet^n}\Mmu(\Hist[n])\log\frac{\Mmu(\Hist[n])}{\Mmu_d(\Hist[n])}=\kl{\Mmu_d(\Hist[n])}{\Mmu(\Hist[n])}\geq 0\,.
    \end{align*}
\end{proof}

\section{Equilibrium behavior of embedded Bayesian agents} \label{sec:subj-emb-eq}

Having defined embedded Bayesian agents and their predictive mechanisms in the previous section, we now turn to a central question: What is the long-term strategic behavior of these agents when they interact? Understanding the notions of optimality in an embedded agency setup and the equilibrium behavior they converge to is important for characterizing their potential for cooperation or conflict. This section provides a comprehensive game-theoretic analysis of embedded Bayesian agents.

The results in this section are multifaceted, covering several distinct axes of variation. To guide the reader through our results, we first discuss and motivate these various axes and briefly situate our main results along them.

\textbf{Decoupled vs. embedded agents.} A primary axis contrasts our work with the foundational literature. The seminal work of \citet{kalai1993rational, kalai1993subjective, kalai1995subjective} uncovered in detail the convergence behavior of \textit{decoupled} Bayesian agents, which we review in \cref{sec:intermezzo-kalai}. In the subsequent sections, we aim to translate these results to the \textit{embedded} Bayesian agents setup, where agents maintain beliefs over universes including themselves, instead of only environments. This requires us to develop new solution concepts that incorporate reasoning based on functional similarities, a feat unique to embedded Bayesian agents.

\textbf{Subjective vs. common knowledge solution concepts.} We distinguish between two types of solution concepts. \textit{Subjective solution concepts} describe the behavior that rational learners, like Bayesian agents, converge to. This behavior is optimal with respect to their internal, \textit{subjective beliefs} about the world which can differ from the subjective beliefs of other agents. In contrast, \textit{common knowledge solution concepts} assess whether this converged behavior is optimal with respect to the \textit{ground-truth} environment and the actual policies of the other agents, and a commonly agreed upon method for evaluating counterfactual behaviors.\footnote{Evaluating counterfactuals of what would happen when the focal agent changes its behavior requires making assumptions. In classical game theory, the assumption is that of decoupledness: the policies of other agents remain unaltered. In the embedded agency setup, such decoupledness assumption can be inaccurate, as the agents' policies can be functionally related. When all agents use the same method and knowledge for evaluating counterfactuals, we term the corresponding solution concepts as \textit{common knowledge solution concepts} contrasting it with subjective solution concepts where each agent can have different subjective beliefs. Some authors prefer the terminology \textit{objective solution concepts} \citep{kalai1995subjective} to indicate such common knowledge solution concepts. However, as such solution concepts still require assumptions about how to evaluate counterfactuals, we prefer the `common knowledge' terminology.} We develop both subjective and common knowledge concepts for the embedded setting.

\textbf{Repeated games vs. MAGRL.} We analyze agent behavior in two different settings. We first develop our new game-theoretic concepts in the classical \textit{repeated games} setup (\cref{sec:eq-behavior-repeated-games}). This setting is simpler due to perfect monitoring of other agents' actions and exact knowledge of the reward function. We then generalize these results to the more complex and realistic Multi-Agent General Reinforcement Learning (MAGRL) framework (\cref{sec:eq-behavior-magrl}), which requires us to handle partial observability and imperfect knowledge about the environment, leading to different equilibrium properties.

\textbf{Exact vs. $\epsilon$-equilibria.} The beliefs of Bayesian agents (both decoupled and embedded) only converge to the ground-truth asymptotically. To characterize agent behavior before convergence is complete, we introduce $\epsilon$-variants of our solution concepts. These $\epsilon$-equilibria characterize behavior where the agent's predictive model is allowed to be $\epsilon$-close to the ground-truth distribution, rather than identical to it.

In \cref{sec:intermezzo-kalai} we review the results of Kalai and Lehrer on the equilibrium behavior of \textit{decoupled} Bayesian agents. In \cref{sec:eq-behavior-repeated-games}, we pivot to the study of embedded Bayesian agents in the setting of repeated games, where we introduce our central contributions: the \textit{subjective embedded equilibrium (SEE)} and its common knowledge counterpart, the \textit{embedded equilibrium (EE)}. These novel concepts take into account the coupled beliefs that arise from reasoning about functional similarities. We prove that embedded agents converge to playing $\epsilon$-SEEs. In \cref{sec:eq-behavior-magrl}, we generalize our findings to the richer Multi-Agent General Reinforcement Learning (MAGRL) framework, showing convergence to a $\epsilon$-correlated SEE while highlighting the challenges introduced by partial observability and unknown reward functions. Finally, \cref{sec:eq-behavior-k-step} analyzes more practical $k$-step planner agents and demonstrates their convergence to approximate versions of these equilibria. The tables below summarize the main solution concepts and convergence results presented in this section along the discussed axes.

\begin{BoxFigure}[Abbreviations for the game-theoretic solution concepts]
\label{box:abbreviations}
\centering
\begin{tabular}{ll}
% \toprule
\textbf{Abbreviation} & \textbf{Full Solution Concept} \\
\midrule
NE & Nash Equilibrium \\
CE & Correlated Equilibrium \\
SNE & Subjective Nash Equilibrium \\
SCE & Subjective Correlated Equilibrium \\
EE & Embedded Equilibrium \\
CEE & Correlated Embedded Equilibrium \\
SEE & Subjective Embedded Equilibrium \\
SCEE & Subjective Correlated Embedded Equilibrium \\
% \bottomrule
\end{tabular}
\end{BoxFigure}

% --- BOX 2: SOLUTION CONCEPTS ---
% Converted from tab:solution-concepts
\begin{BoxFigure}[Overview of Game-Theoretic Solution Concepts]
\label{box:solution-concepts}
\centering
\begin{tabular}{lll}
% \toprule
 & \textbf{Decoupled Agents} & \textbf{Embedded Agents} \\ 
\midrule
\textbf{Common knowledge} & 
    \begin{tabular}[t]{@{}l@{}}
    NE (Repeated Games) \\
    CE (MAGRL)
    \end{tabular}
    & 
    \begin{tabular}[t]{@{}l@{}}
    EE (Repeated Games) \\
    CEE (MAGRL)
    \end{tabular}
    \\ \midrule
\textbf{Subjective} & 
    \begin{tabular}[t]{@{}l@{}}
    SNE (Repeated Games) \\
    SCE (MAGRL)
    \end{tabular}
    & 
    \begin{tabular}[t]{@{}l@{}}
    SEE (Repeated Games) \\
    SCEE (MAGRL)
    \end{tabular}
    \\ %\bottomrule
\end{tabular}
\par\nobreak\vspace{1ex}
\end{BoxFigure}

% --- BOX 3: CONVERGENCE RESULTS ---
% Converted from tab:convergence-results
\begin{BoxFigure}[Summary of Main Convergence Results]
\label{box:convergence-results}
\centering
\begin{tabular}{lll}
% \toprule
 & \textbf{Decoupled Agents} & \textbf{Embedded Agents} \\ 
\midrule
\textbf{Repeated Games} & 
    \begin{tabular}[t]{@{}l@{}}
    Converge to $\epsilon$-SNE \\ (and $\epsilon$-NE)
    \end{tabular}
    & 
    \begin{tabular}[t]{@{}l@{}}
    Converge to $\epsilon$-SEE \\ (and $\epsilon$-EE under add. conditions)
    \end{tabular}
    \\ \midrule
\textbf{MAGRL} & 
    Converge to $\epsilon$-SCE
    & 
    Converge to $\epsilon$-SCEE
    \\ %\bottomrule
\end{tabular}
\par\nobreak\vspace{1ex}
\centering\small
\end{BoxFigure}

\subsection{Background on equilibrium Behavior of Decoupled Bayesian Agents}\label{sec:intermezzo-kalai}

Before analyzing the equilibrium behavior of embedded agents, we first set the stage by reviewing the foundational results of \citet{kalai1993rational, kalai1993subjective, kalai1995subjective}, who studied the equilibrium behavior of \textit{decoupled} Bayesian agents. Their work provides a crucial foundation, establishing convergence to $\epsilon$-Nash equilibria in the setting of repeated games with perfect monitoring.
We begin by formally defining this setting.

\begin{definition}[Repeated Games with Perfect Monitoring]\label{def:repeated-games}
A repeated game consists of $T$ rounds of a single-stage game where each agent $i$ simultaneously takes an action $a^i \in \mathcal{A}^i$ and receives a reward $r^i(a^i, a^{-i})$ based on its own action and the actions chosen by the other agents $-i:=[N]\setminus\{i\}$ in the current round.\footnote{Note that we use the notation $[N]:=\{1,\ldots,N\}$.} The number of rounds $T$ can be infinite. The agents have perfect recall of the history $\Rgh=(\bar{a}_{1},...,\bar{a}_{t})\in\bar{\calA}^*$, which consists of the previous actions of all agents. Each agent has a policy $\Mvp^{i}:\bar{\calA}^*\rightarrow\Delta\calA^{i}$ that makes a decision independently of the decisions of other agents in the current round. The agents have perfect knowledge of their own reward function $r^i(\bara)$, which is important when agents construct their behavioral policy $\Mvp^i$ through, e.g., optimal planning. We define the percepts equal to the actions of the other agents: $e^i:=a^{-i}$. The agents can then use their percepts and own actions to compute their reward $r^i(\bara)$.
\end{definition}

As in repeated games, the rewards are not provided by the environment but instead computed by the agents with reward function $r^i(\bara)$, and the percepts are equal to the actions of other agents, the multi-agent environment $\MultiAgentMve$ is a dummy environment interfacing the policies of all agents.

Within this framework of repeated games, \citet{kalai1993rational} consider a specific case of decoupled Bayes-optimal agents that maintain a subjective belief model for each of its co-players, as detailed below. 

\begin{definition}[Decoupled Bayes-optimal agent in repeated games]\label{def:decoupled-agents-repeated-games}
A decoupled Bayes-optimal agent in a repeated game with perfect monitoring is an agent $i$ that maintains an independent subjective belief model for each co-player $j \neq i$, represented as a mixture model over possible co-player's policies: $\Mme_j^i(a^j|\Rgh) = \sum_{\Mvp^j \in \calMpol^j} w_j^i(\Mvp^j|\Rgh)\Mvp^j(a^j|\Rgh)$. These individual models are combined into a single mixture environment model, $\Mme^i$, under the assumption that co-players act independently:
$$ \Mme^i(e^{i}_t|\RghM, a^i) = \Mme^i(a^{-i}_t|\RghM, a^i) := \prod_{j \neq i} \Mme_j^i(a_t^j|\RghM). $$
The agent's policy, $\Mvp^i$, is then a best response computed with respect to this subjective environment $\Mme^i$ (cf. \eqref{eqn:decoupled-br}), using the agent's ground-truth reward function $r^i(\bara)$ to compute rewards from the predicted joint actions.
\end{definition}

\subsubsection{Subjective and common knowledge solution concepts}

The central question is: what is the long-term equilibrium behavior when such agents interact? To answer this, Kalai and Lehrer introduced the \textit{subjective Nash equilibrium} \citep{kalai1993rational, kalai1993subjective, kalai1995subjective}. Below, we define the subjective Nash equilibrium in its general form that is also applicable to the MAGRL setting, and discuss afterwards how it simplifies in the repeated games setting. 

\begin{definition}[Subjective Nash Equilibrium]\label{def:subj-nash-eq}
A set of policies $(\Mvp^i)_{i \in N}$ and subjective environment models $(\Mme^i)_{i \in N}$ constitutes a subjective Nash equilibrium in the multi-agent environment $\bar{\Mge}$ if, for each agent $i$:
\begin{enumerate}
    \item \textbf{Subjective Best Response}: The policy $\Mgp^i$ is a decoupled best response (cf. \eqref{eqn:decoupled-br}) with respect to the subjective environment model $\Mme^i$.
    \item \textbf{Uncontradicted Beliefs}: The distribution over histories induced by agent $i$'s beliefs and its own policy, $\left(\Mme^{i}\right)^{\Mvp^i}$, is identical to the ground-truth personal distribution $\left(\Mge^i\right)^{\pi^i}$ induced by the multi-agent environment $\bar{\Mge}$ and policies of all agents.
\end{enumerate}
\end{definition}

An \textit{$\epsilon$-subjective Nash equilibrium} relaxes the second condition, requiring only that the two distributions are $\epsilon$-close in total variation distance, i.e., $D_{\infty}\left(\left(\Mme^{i}\right)^{\Mvp^i}, \left(\Mge^i\right)^{\pi^i} \middle| \varepsilon\right) \le \epsilon$. 

For repeated games, the multi-agent environment $\bar{\Mge}$ is a dummy interface between the policies of all agents, and the personal histories are equal to joint action sequences. Hence the ground-truth personal distribution is equivalent to the joint policies of all agents: $$\big(\Mge^i\big)^{\pi^i}(\bara_{1:t}) = \prod_{k=1}^t \prod_{j=1}^N \Mgp^i(a_k^j \mid \bara_{<k}).$$

This is a \textit{subjective} solution concept because each agent computes a best response against its own belief model $\Mme^i$, not necessarily against the ground-truth environment composed of the other agents' true policies. The \textit{uncontradicted beliefs} condition grounds these subjective beliefs in reality, but only \textit{on the play-path}—that is, for histories with a non-zero probability of occurring. For counterfactual histories that have zero probability under $\Mvp^i$, the conditionals $\Mme^i(a_t^{-i}|\RghM)$ are not constrained by reality. These "off-the-play-path" beliefs can be incorrect yet are never falsified by observation. Nonetheless, they are crucial for the optimal planning routine, which must evaluate all counterfactual actions, including those with zero probability under the agent's policy. 

The corresponding \textit{common knowledge} solution concept is the well-known \textit{Nash equilibrium}, where each agent's policy must be a best response to the \textit{actual} policies of the other agents \citep{nash_jr_equilibrium_1950}, hence each agent has common knowledge of the policies of the other agents and the ground truth environment. 
\begin{definition}[Nash Equilibrium]\label{def:nash-eq}
A set of policies $(\Mvp^i)_{i \in N}$ constitutes a Nash equilibrium in the multi-agent environment $\bar{\Mge}$ if, for each agent $i$, the policy $\Mgp^i$ is a best response w.r.t. the ground-truth personal environment $\mu^i$ induced by the multi-agent environment $\bar{\Mge}$ and policies of all agents: $$\Mgp^i \in \argmax_{\pi} V_{(\Mge^i)^\pi}(\varepsilon).$$
\end{definition}
An $\epsilon$-Nash equilibrium relaxes the best response condition, requiring that the value of each agent's policy is $\epsilon$-close to the optimal value. Note that for $\epsilon$-Nash equilibria, the $\epsilon$-closeness condition is w.r.t. the best response computation, whereas for $\epsilon$-subjective Nash equilibria, the $\epsilon$-closeness condition is on the subjective beliefs, while still requiring a perfect best response w.r.t. those subjective beliefs.

A key result from Kalai and Lehrer is that for every subjective Nash equilibrium in a repeated game, there exists a Nash equilibrium that induces an identical distribution $\bar{\Mge}^{\bar{\Mgp}}$ over histories \citep[Proposition 1]{kalai1993subjective}. Similarly, for every $\epsilon$-subjective Nash equilibrium, there is an $\epsilon$-Nash equilibrium whose history distribution $\bar{\Mge}^{\bar{\Mgp}}$ is $\epsilon$-close \citep[Theorem 1]{kalai1993subjective}. These important results highlight that although the conditionals $\Mme^i(a_t^{-i}|\RghM)$ can be different from the ground-truth co-player policies for zero-probability histories $\RghM$, potentially impacting the optimal planning routine to compute best responses, the resulting best-response policies lead to indistinguishable trajectory distributions $\bar{\Mge}^{\bar{\Mgp}}$. Crucially, Kalai and Lehrer only prove this for the case of repeated games with perfect monitoring. As shown in later work by the same authors, this equivalence does \textit{not} hold in the general multi-agent RL setup \citep{kalai1995subjective}.

\subsubsection{Convergence of decoupled Bayesian agents in repeated games}

We aim to demonstrate that for any $\epsilon>0$, a group of decoupled Bayesian agents play an $\epsilon$-subjective equilibrium after a sufficiently long period $t$. To formalize this, we must explicitly define the environment dynamics and agent's policies for the new \textit{tail game} that starts at time $t$ and continues indefinitely. We define tail games in their general form applicable to MAGRL, and discuss afterwards how it simplifies in the case of repeated games we consider here.

\begin{definition}[Tail games]\label{def:tail-game}
    Let $\bar{\Mge}$ be a multi-agent environment, and $(\pi^i)_{i=1}^N$ the set of policies of the agents. Let us define the \textit{tail multi-agent environment} $\MAgeTailM$ and \textit{tail policies} $\MgpTailM^i$ corresponding to the \textit{tail game} starting at timestep $t$ induced by joint history $\MAHistM$ as
    $$\MAgeTailM(\bar{e}_{t'+1} \mid \MATurn_{t:t'}\bar{a}_{t'+1}):= \bar{\Mge}(\bar{e}_{t'+1} \mid \MATurn_{<t}\MATurn_{t:t'}\bar{a}_{t'+1})$$
    $$\MgpTailM^i(a^i_{t'+1} \mid \Turn^i_{t:t'}):= \Mgp^i(a^i_{t'+1} \mid \Turn_{<t}^i\Turn_{t:t'}^i)$$
    The corresponding measure over tail trajectories is defined as
    $\MAgeTailM^{\bar{\Mgp}}(\MATurn_{t:t'}):= \bar{\Mge}^{\bar{\Mgp}}(\MATurn_{t:t'} \mid \MATurn_{<t})$. Marginalizing out the co-player trajectories provides us with the tail personal environments
    $\MgeTailM^i$. Similarly, we define the tail mixture environments as
    $$\MmeTailM^i(e^i_{t'+1} \mid \Turn^i_{t:t'}a^i_{t'+1}):= \Mme^i(e^i_{t'+1} \mid \Turn_{<t}^i\Turn_{t:t'}^ia^i_{t'+1})$$
\end{definition}

In repeated games, we assume the agents are given their reward functions, instead of including rewards in the percepts $e^i$. Furthermore, we assume perfect monitoring of the actions of others. Hence, personal histories are sequences of joint actions, and the tail personal measures $\big(\Mge_{\bara_{<t}}^i\big)^{\Mgp_{\bara_{<t}}^i}$ are equivalent to the joint tail policies:
$$\big(\Mge_{\bara_{<t}}^i\big)^{\Mgp_{\bara_{<t}}^i}(\bara_{t:t'}) = \prod_{k=t}^{t'}\prod_{j=1}^N\Mgp_{\bara_{<t}}^j(a^j_{k} \mid \bara_{t:k-1})$$

Building on these concepts, Kalai and Lehrer's central theorem demonstrates that decoupled Bayesian agents whose beliefs satisfy the grain-of-truth property converge to playing an $\epsilon$-subjective Nash equilibrium.

\begin{theorem}[Convergence to $\epsilon$-subjective Nash equilibrium \citep{kalai1993rational}]\label{thrm:kalai-convergence-repeated-games}
Consider a repeated game with perfect monitoring played by decoupled Bayes-optimal agents. If each agent's subjective model $\Mme^i$ satisfies the grain-of-truth property with respect to the ground-truth personal environment $\MultiAgentGeGpI$, then for any $\epsilon > 0$, there exists a time $T(\epsilon)$ such that for all $t \ge T(\epsilon)$, with $\MultiAgentMge^{\MultiAgentMvp}$-probability greater than 1-$\epsilon$ over $\MAHistM$, the agents' tail policies $\MgpTailM^i$ and tail mixture environments $\MmeTailM^i$ constitute an $\epsilon$-subjective Nash equilibrium of the tail game starting at time $t$.
\end{theorem}

Combined with the equivalence result between $\epsilon$-subjective Nash equilibria and $\epsilon$-Nash equilibria \citep{kalai1995subjective}, this implies that decoupled Bayes-optimal agents satisfying the grain-of-truth property converge to playing an $\epsilon$-Nash equilibrium. This foundational result establishes a strong link between Bayesian rationality and classical game-theoretic equilibria.

\subsubsection{Convergence of Decoupled Bayesian Agents in Multi-Agent General RL}

So far, we focused on repeated games with perfect monitoring. \cite{kalai1995subjective} broaden the scope to the Multi-Agent General Reinforcement Learning (MAGRL) setup, which accommodates partial observability and more complex environment dynamics. In this richer setting, the percepts $e^i$ contain both the current reward $r^i$ and observation $o^i$, where the observation can contain (partial) information about the actions of other agents, as well as other information from the environment. A decoupled Bayes-optimal agent in this setting still computes a best response to a subjective model $\Mme^i$ of its environment, but the nature of equilibria and convergence changes significantly. The definitions of the Nash equilibrium (cf. Definition \ref{def:nash-eq}) and subjective Nash equilibrium (cf. Definition \ref{def:subj-nash-eq}) apply to the MAGRL case as well, but now with no restrictions on the multi-agent environment $\bar{\Mge}$, and with $\Mme^i$ a general mixture environment following \cref{eqn:decoupled-bayesian-agent} instead of Definition \ref{def:decoupled-agents-repeated-games}.

A crucial distinction from the repeated games setting is the presence of partial observability. Because agents no longer share a common history, their individual observation streams can lead to different but correlated personal histories. Hence, the tail policies $\MgpTailM^i$ are conditioned on different but correlated histories $\Turn^i_{<t}$, and thus the tail game starting at time $t$ should take such correlated information into account. In contrast, in repeated games, all tail policies are conditioned on the identical history $\bar{a}_{<t}$, and therefore the resulting tail game does not incorporate private and correlated information that the agents are given.

In game theory, private, correlated information that agents are given before the start of a game is formally modeled through private messages $m^i$ coming from a correlation device $(M,p)$. Here, $M=\prod_i M^i$ is the joint space over individual messages $m^i$ for each agent, and $p(\bar{m})$ is the probability distribution over joint messages $\bar{m}=\{m^i\}_{i\in N}$. In our case, considering a tail game starting at time $t$, the messages are the personal histories $\Turn^i_{<t}$, and the distribution $p$ is the ground-truth distribution $\bar{\nu}^{\bar{\pi}}$ over joint histories $\MATurn_{<t}$.

The common knowledge solution concept for games incorporating a correlation device is the correlated equilibrium \cite{aumann1974subjectivity}.

\begin{definition}[Correlated Equilibria]\label{def:corr-eq}
A correlation device $(M, p)$ combined with a set of policies $\{\Mvp^i\}_{i\in N}$ with $\Mvp^i: M^i\times (\TurnSet^i)^* \to \Delta \calA^i$ form a \textit{correlated equilibrium} in a correlated multi-agent environment $(\bar{\Mge}, M, p)$ if each agent's policy $\Mgp^i$ is a best response w.r.t. the ground-truth personal correlated environment $\Mge^{i}(e^i_t \mid \Turn_{<t}^i, m^i)$ for each message with $p(m^i)>0$, with $\Mge^{i}(e^i_t \mid \Turn_{<t}^i, m^i)$ obtained from appropriately marginalizing and conditioning the joint distribution induced by policies $(\Mvp^j)_{j=1}^N$, environment $\bar{\Mge}$ and correlation device $(M,p)$:
    \begin{align}\label{eq:corr-joint-distribution}
        \bar{\Mge}^{\bar{\Mvp}}(\MAHist, \bar{m}) = p(\bar{m})\prod_{k=1}^{t}\left(\prod_{i\in N} \Mvp^i(a_k^i \mid \MAHistM[k], m^i) \right)\bar{\Mge}(\bare_k \mid \MAHistM[k], \bara_k, \bar{m}) 
    \end{align}
\end{definition}

Bayesian agents compute best responses w.r.t. their subjective beliefs and not necessarily the ground-truth environment, hence we also need a subjective variant of the correlated equilibrium.

\begin{definition}[Subjective Correlated Equilibria \citep{kalai1995subjective}]\label{def:subj-corr-eq-kalai}
A set of policies $\{\Mvp^i\}_{i\in N}$ with $\Mvp^i: M^i\times (\TurnSet^i)^* \to \Delta \calA^i$ and subjective mixture environments $\{\Mme^i(e^i \mid m^i,\HistI ,a^i)\}_{i\in N}$ is a \textit{subjective correlated equilibrium} w.r.t. multi-agent environment $\bar{\Mge}$ and correlation device $(M, p)$ if the following two conditions hold:
\begin{enumerate}
    \item \textbf{Subjective best response.} Each agent's policy $\Mvp^i$ is a best response w.r.t. $\Mme^i(e^i \mid m^i,\HistI ,a^i)$ for each history and message with $p(m^i)>0$.
    \item \textbf{Uncontradicted beliefs.} The distribution over histories induced by the agent's beliefs and policy is identical to the distribution induced by the policies of all agents combined with the ground-truth correlated multi-agent environment $(\MultiAgentMge, M, p)$:
    \begin{align*}
        \big(\Mme^{i}\big)^{\Mvp^i}(\HistI \mid m^i) = \big(\Mge^{i}\big)^{\Mvp^i}(\HistI \mid m^i) \quad \forall i\in N, \forall \HistI \in \TurnSet^*, \forall m^i\in M^i: p(m^i)>0
    \end{align*}
    with $\big(\Mge^{i}\big)^{\Mvp^i}$ obtained from appropriately marginalizing and conditioning the joint distribution $\Mge_{\bar{\Mge}}^{\bar{\Mvp}}(\MAHist, \bar{m})$ (\eqref{eq:corr-joint-distribution}).
    \end{enumerate}
\end{definition}

The \textit{$\epsilon$-subjective correlated equilibrium} relaxes the second condition: with probability greater than $1-\epsilon$, a message $\bar{m}$ is chosen such that the subjective beliefs $\big(\Mme^{i})^{\Mvp^i}(\HistI \mid m^i)$ are $\epsilon$-close to the personal ground-truth distributions $\big(\Mge^{i}\big)^{\Mvp^i}(\HistI \mid m^i)$ \citep{kalai1995subjective}.

With these concepts in place, we present the central result of \cite{kalai1995subjective} for the MAGRL setting: that decoupled Bayesian agents converge not to a subjective Nash, but to a subjective correlated equilibrium.

\begin{theorem}[Convergence to $\epsilon$-Subjective Correlated Equilibrium \citep{kalai1995subjective}]
Consider a MAGRL environment played by decoupled Bayes-optimal agents. If each agent's subjective model $\Mme^i$ satisfies the grain-of-truth property with respect to the ground-truth personal environment $\Mge^i$, then for any $\epsilon > 0$, there exists a time $T$ such that for all $t \ge T$, the agents' tail policies $\MgpTailM^i$ and tail mixture environments $\MmeTailM^i$ constitute an $\epsilon$-subjective correlated equilibrium in the correlated tail game starting at time $t$ with correlation device $(\MATurnSet^{t-1}, \bar{\Mge}^{\bar{\Mvp}})$.
\end{theorem}

Critically, the equivalence between subjective and common knowledge equilibria \citep[Theorem 1]{kalai1993subjective}, which holds in repeated games, breaks down in the general RL setup. In the MAGRL setting, a subjective equilibrium (Nash or correlated) is not generally equivalent to a common knowledge one. The reason lies in the expanded role of the subjective model $\Mme^i$. In MAGRL, $\Mme^i$ predicts not simply the other agents' actions but percepts which also include the obtained rewards. An agent's beliefs about rewards for counterfactual, off-the-play-path histories can be incorrect. For instance, an agent might hold a dogmatic belief that any deviation from its current policy $\Mgp^i$ will result in a catastrophically low reward \citep{kalai1995subjective, leike2015bad}. Since this belief is never contradicted by observation as the agent never deviates, this locks the agent into a suboptimal policy that is nonetheless a best response to its flawed subjective environment $\Mme^i$. This prevents the agent from gathering the very data needed to correct its erroneous beliefs.

In summary, this subsection has charted the foundational results for decoupled Bayesian agents. We showed that in the simplified setting of repeated games, decoupled Bayesian agents under the grain-of-truth assumption converge to $\epsilon$-Nash equilibria. However, in the more general and realistic MAGRL setting, convergence is only guaranteed to an $\epsilon$-subjective correlated equilibrium, with no general guarantee of objective optimality due to the challenge of correcting off-path beliefs. With this background established, we will now follow a similar analytical path to study the equilibrium behavior of the \textit{embedded} Bayesian agents introduced in this paper. This will require us to develop novel subjective and common knowledge solution concepts to properly characterize their unique equilibrium properties.

\subsection{Equilibrium behavior of embedded Bayesian agents on repeated games}\label{sec:eq-behavior-repeated-games}

We analyze the equilibrium behavior of embedded Bayesian agents, beginning with the setting of repeated games with perfect monitoring. In this context, an agent's percepts consist of the other agents' actions, and its rewards are determined by a known reward function. The core difference from the decoupled case is that embedded agents can maintain \textit{coupled} beliefs over their own policy and the policies of others, a direct consequence of reasoning about functional similarities as discussed in Section 3.6. This allows an agent's action $a_t^i$ to be evidentially linked to the concurrent actions of other agents, $a_t^{-i}$, through the Bayesian belief update $w(\Mvu \mid \MAHistM, a_t^i)$. This seemingly subtle shift has important implications for the resulting equilibria.

We first define the specific agent model for this setting. We consider universes that are composed of a joint policy, where each agent acts independently conditioned on the history. The coupling between agents' actions arises not from the universe's causal structure, but from each agent's subjective beliefs $w^i(\Mvu)$ over which universe it inhabits.

\begin{definition}[Embedded Bayes-optimal agent in repeated games]\label{def:embedded-agents-repeated-games}
An embedded Bayes-optimal agent in a repeated game with perfect monitoring is an agent $i$ that maintains a joint subjective belief model $\Mmu^i$ over a class of universes $\calMuni$. Each universe $\Mvu \in \calMuni$ is equivalent to a joint policy $\bar{\Mvp}_\Mvu$ that factorizes over independent agent policies: $\Mvu(\bara_{1:t}) = \prod_{k=1}^{t} \prod_{i \in [N]} \Mvp^i_\Mvu(a^i_k \mid \bara_{<k})$. The agent's beliefs are captured by the mixture universe $\Mmu^i(\bara_{1:t}) = \sum_{\Mvu \in \calMuni} w^i(\Mvu) \Mvu(\bara_{1:t})$, which yields the following conditional for the other agents' actions:
$$\Mmu^i(a^{-i}_t|\RghM, a_t^i) := \sum_{\Mvu \in \calMuni}w^i(\Mvu \mid \RghM, a_t^i) \prod_{j \neq i}\Mvp_\Mvu^j(a^j_t \mid \RghM)\,.$$
The agent's policy, $\Mvp^i$, is then an embedded best response computed with respect to this subjective mixture universe $\Mmu^i$ (cf. \eqref{eqn:embedded-br}).
\end{definition}

A crucial difference between the mixture universe of embedded Bayesian agents and the mixture environment of decoupled Bayesian agents in repeated games is that in the mixture universe $\Mmu^i(a^{-i}_t|\RghM, a_t^i)$, there is a dependence of the other agents' current actions $a^{-i}_t$ on the ego-agent's current action $a_t^i$. As each universe $\lambda \in \calMuni$ adheres to a causal structure where there is no dependence between $a^{-i}_t$ and $a^i_t$, this dependence in the mixture universe $\Mmu$ is an \textit{informational} dependence and not a \textit{causal dependence}. This informational link arises due to the posterior belief update $w^i(\Mvu \mid \RghM, a_t^i)$ based on the action $a^i$, which is used to predict the other agents' actions $a^{-i}$. 
In mixture environments $\Mme$ of decoupled Bayesian agents in the repeated games setting, there is no dependence between $a^{-i}_t$ and $a^i_t$, because the mixture environment can only model causal dependencies of actions, in contrast to mixture universes which both can model causal and informational dependencies on actions (cf. Remark \ref{rem:coupled-vs-decoupled}). 

\subsubsection{Subjective solution concepts} 

The potential for dependencies of the other agents' current actions $a^{-i}_t$ on the ego-agent's current action $a_t^i$ in mixture universes $\rho$, necessitates a solution concept that can accommodate such coupled beliefs. Standard concepts like the Nash equilibrium assume independence. To address this, we introduce the \textit{subjective embedded equilibrium} (SEE), a concept that combines the subjective Nash equilibrium with the idea of a {\textit dependency equilibrium}, introduced in \cite{spohn2007dependency} (cf. \cref{app:equivalence-code-de}).

Differently from the case of subjective Nash equilibria, defining exact subjective embedded equilibria requires the solution of a technical challenge. For clarity, we therefore begin with the $\epsilon$-relaxed version, which characterizes the behavior of embedded Bayesian agents at finite times.

\begin{definition}[$\epsilon$-Subjective Embedded Equilibrium]\label{def:eps-sde}
A set of policies $\{\Mvp^i\}_{i\in N}$ and subjective mixture universes $\{\Mmu^i\}_{i\in N}$ satisfying the grain-of-uncertainty property is an $\epsilon$-\textit{subjective embedded equilibrium} in a repeated game if the following two conditions hold:
\begin{enumerate}
    \item \textbf{Subjective Best Response.} Each agent's policy $\Mgp_i$ is an embedded best response with respect to its subjective mixture universe $\Mmu^i$.
    \item \textbf{$\epsilon$-Uncontradicted Beliefs.} The subjective beliefs $\Mmu^i$ are $\epsilon$-close in total variation distance to the personal real-world distribution $\MultiAgentGeGpI$ induced by the ground-truth multi-agent environment $\bar{\Mge}$ and policies of all agents: $D_\infty(\Mmu^i , \MultiAgentGeGpI \mid \varepsilon) \leq \epsilon$
\end{enumerate}
\end{definition}
In the repeated games setting, the multi-agent environment $\bar{\Mge}$ captures the reward functions of all involved agents, and the agents have perfect knowledge of their own reward function. As we assume perfect monitoring of the other agent's actions, and the agent's do not need to predict rewards, we have that the percepts are $e^i:= a^{-i}$ and the agent's personal histories are sequences of joint actions. Hence, the ground-truth personal distribution corresponds to the distribution induces by all agents: $$\MultiAgentGeGpI(\bara_{1:t}) = \prod_{k=1}^t \prod_{j=1}^{N}\pi^j(a^j_k \mid \bara_{<k})\,.$$

This leads to our main convergence result for embedded agents in repeated games: under the grain-of-truth assumption, embedded agents converge to playing an $\epsilon$-SEE.

\begin{theorem}[Convergence to $\epsilon$-Subjective Embedded Equilibrium]\label{thrm:convergence-to-sde}
Let $\{\Mmu^i\}_{i\in N}$ be Bayesian mixture universes satisfying the grain-of-uncertainty and grain-of-truth conditions w.r.t. $\MultiAgentGeGpI$, and let $\{\Mgp^i\}_{i\in N}$ be the corresponding embedded best response policies in an infinitely repeated game with perfect monitoring. Then, for each $\epsilon > 0$, there exists a finite time $T(\epsilon)$ such that for all $t \geq T(\epsilon)$, with $\MultiAgentMge^{\MultiAgentMvp}$-probability greater than $1-\epsilon$ over $\MAHistM$, the agents' tail policies $\Mgp^i_{\RghM}$ and tail mixture universes $\Mmu_{\RghM}^i$ constitute an $\epsilon$-subjective embedded equilibrium of the tail game starting at time $t$.
\end{theorem}
\begin{proof}
    See \cref{app:proof-convergence-to-sde}.
\end{proof}

\begin{example}[Convergence to a Cooperative $\epsilon$-SEE]\label{example:bayesian-agents-cooperation}
To illustrate how embedded Bayesian agents can converge to a cooperative $\epsilon$-SEE, we consider a two-player iterated Prisoner's Dilemma with discount $\gamma=0$. The zero discount makes each round strategically equivalent to a one-shot game, while the history allows for belief updates. To construct the agent's prior over universes, we first start from a prior over policies, which we combine into a prior over universes. Consider a countable class of deterministic policies $\calMpol$ and a prior $\tilde{w}(\Mvp)>0$ for all $\Mvp\in \calMpol$. We further assume that the class $\calMpol$ is large enough so that, for every $t$ and every history $\Rgh$ of length $t$, there exists a pair of policies in $\calMpol$ which act according to $\Rgh$ for the first $t$ rounds, and then either cooperate forever after or defect forever after.\footnote{In particular, $\calMpol$ contains a policy which always cooperates and a policy which always defects.} This will ensure that the grain-of-uncertainty property is satisfied.

The agents' prior over universes, $w^i(\Mvu)$, is constructed as a combination of two hypotheses:
$$ w^i(\Mvu):= \sum_{\Mvp^i \in \calMpol}\sum_{\Mvp^{-i} \in \calMpol} \hat{w}^i(\Mvp^i, \Mvp^{-i})\delta(\Mvu^{\Mvp^i}_{\Mvp^{-i}} = \lambda)\,,$$
where
$$\hat{w}^i(\Mvp^i, \Mvp^{-i}) := \alpha \tilde{w}(\Mvp^i)\delta(\Mvp^i = \Mvp^{-i}) + (1-\alpha) \tilde{w}(\Mvp^i)\tilde{w}(\Mvp^{-i})\,. $$
Here, $\alpha \in [0,1]$ is the agent's prior belief in the "other agent is an identical copy" hypothesis ($\Mvp^i = \Mvp^{-i}$), while $1-\alpha$ is the corresponding prior belief in the ``other agent is an independent draw from $\tilde{w}$'' hypothesis. This prior structure is motivated by Occam's razor, as a universe with identical agents is algorithmically simpler to describe.

Since both embedded Bayes-optimal agents have the same priors, they implement the same policy. However, they don't know up front that they are identical copies, and need to infer this from their interactions. Let us assume that the EBR policy breaks ties in a fixed canonical way.\footnote{E.g., if an agent finds itself in a situation where the $Q$ value associated with cooperating is exactly the same as the one associated with defecting, then the agent defects by default. We could also make the default action to be cooperation.} This makes the EBR policy deterministic, and hence the same deterministic EBR policy is followed by both agents. Since the game is symmetric and since both agents follow the same deterministic policy, we can see that they will produce a trajectory in such a way that in each round the agents take the same action, i.e., they either both cooperate or both defect. Their actions may change from one round to another, but it will always be the case that they take the same action in the same round.

It is not yet clear whether the deterministic EBR policy belongs to $\calMpol$, but we will show that in the considered setup, the two embedded Bayesian agents will either always defect or converge to cooperating, which would then imply that the grain-of-truth property is satisfied because such policies are part of $\calMpol$.

To show that both embedded Bayesian agents can converge to mutual cooperation, let us first define 
$$m(\Rgh):=\sum_{\Mvp \in \calMpol} \tilde{w}(\Mvp) \prod_{k=1}^t\Mvp(a^1_k \mid \RghM[k])\,,$$
which is the prior probability that a policy drawn from $\tilde{w}(\Mvp)$ produces the actions within $\Rgh$, when it is used as policy for agent 1.\footnote{In the considered setting both agents have an identical and deterministic ground-truth policy, hence we have that $a_t^1=a_t^2$ for all timesteps $t$. Hence the definition of $m(\Rgh)$ is invariant on whether we assume the policies are used by agent 1 or agent 2.} For every $k\geq 0$, define $$m_{k}^{\textrm{defect}}=m((D,D)^k)\,,$$
where we adopt the convention that $m_{0}^{\textrm{defect}}=m((D,D)^0)=m(\varepsilon)=1$, and
$$m_{\infty}^{\textrm{defect}}=\lim_{k\to\infty} m((D,D)^k)\,.$$
$m_{\infty}^{\textrm{defect}}$ is the prior probability that a policy drawn from $\tilde{w}(\Mvp)$ always defects, in case the opponent defects as well.

If the prior belief in functional similarity is sufficiently strong ($\alpha > \frac{m_{\infty}^{\textrm{defect}}}{1 + m_{\infty}^{\textrm{defect}}}$), then after some finite time $T$, the accumulated evidence that both agents are identical, i.e., the history of identical actions, will outweigh the alternative hypothesis that the two agents are different but happened to produce the same actions up until time $T$. As a consequence, after this time $T$, an agent's choice to cooperate provides strong evidence that its opponent will do likewise, making cooperation the subjectively optimal action (cf. \cref{app:example-bayesian-agents-cooperation} for a detailed explanation). In fact, we can characterize $T$ as being the smallest $k$ for which $\alpha > \frac{m_{k}^{\textrm{defect}}}{1 + m_{k}^{\textrm{defect}}}$, and we can further show that the embedded Bayes-optimal agents defect up until time $T$ after which they cooperate forever after. Since such policies belong to $\calMpol$, we can see that the grain-of-truth property is satisfied. Refer to \cref{app:example-bayesian-agents-cooperation} for further details.

As the agents' mixture universes with priors $w^i$ satisfy the grain-of-truth property, we have that for each $\epsilon > 0$, there exists a finite time $t(\epsilon)$ for which the mixture universe is $\epsilon$-close to the ground-truth distribution (cf. \cref{theorem:convergence-mixture-universe}). Hence, the agents converge to the $\epsilon$-subjective embedded equilibrium of mutual cooperation.

On the flip side, we can show that when the prior belief in functional similarity is too weak, i.e., $\alpha \leq \frac{m_{\infty}^{\textrm{defect}}}{1 + m_{\infty}^{\textrm{defect}}}$, then the embedded Bayesian agents always defect, which is also an $\epsilon$-subjective embedded equilibrium. Hence, the choice of prior (through $\alpha$) is an important variable determining the equilibrium behavior of the embedded Bayesian agents. When taking an Occam's razor prior following the \textit{minimum description length} line of thought, this would motivate a large $\alpha$, as universes containing two identical agents have roughly half the description length of a universe containing two different agents of similar complexity, and hence result in cooperative behavior among identical embedded Bayesian agents.

\hfill $\diamondsuit$
\end{example}

We now consider the exact, non-$\epsilon$ version of the subjective embedded equilibrium (SEE) that characterizes the asymptotic limit of play, but we first must address a technical challenge. Agents' policies may be deterministic. If an agent's subjective model $\Mmu^i$ converges to reflect this deterministic policy in the limit of time to infinity, it will assign zero probability to certain actions, violating the grain-of-uncertainty assumption. This is problematic because the embedded best response calculation requires well-defined conditionals $\Mmu^i(a^{-i} \mid \Rgh, a^i)$ for \emph{all} actions $a^i$, including those that will have zero probability according to the resulting embedded best response policy. To resolve this, we introduce the concept of a \textit{conditional completion} of a predictive model, which defines these necessary counterfactual beliefs as the limit of a sequence of models that \textit{do} satisfy the grain-of-uncertainty property. Note that while this section is about the repeated games with perfect-monitoring setting, the following definition is stated in the more general MAGRL setting, for later reuse.

\begin{definition}[Conditional Completion]\label{def:conditional-completion}
A conditional completion of a measure $\Mvu(\MAHist)$ consists of the measure $\Mvu(\MAHist)$ itself, along with a set of conditionals $\Mvu(e \mid \MAHistM, a)$ for all $\MAHistM \in \MATurnSet^*$, $a \in \mathcal{A}$, and $e \in \mathcal{E}$, derived as follows:
\begin{enumerate}
    \item There exists a sequence of measures $\{\Mvu_r(\MAHist)\}_{r\in \mathbb{N}}$, each satisfying the grain-of-uncertainty property, that converges to $\Mvu$, i.e., $\lim_{r\to \infty} \Mvu_r(\MAHist) = \Mvu(\MAHist)$.
    \item The completed conditionals are defined by the limit:\footnote{We assume the series $\{\Mvu_r(\MAHist)\}_{r\in \mathbb{N}}$ is suitably chosen such that this limit exists.} $\Mvu(e \mid \MAHistM, a) := \lim_{r\to \infty} \Mvu_r(e \mid \MAHistM, a) $.
\end{enumerate}
\end{definition}
In repeated games with perfect monitoring, where $e^i=a^{-i}$, the conditional completion specifies an agent's beliefs about what its opponents would do in response to a counterfactual action, even if the agent assigns zero probability to taking that action itself. This allows us to formally define the SEE.

\begin{definition}[Subjective Embedded Equilibrium]\label{def:sde}
A set of policies $\{\Mvp^i\}_{i\in N}$ and subjective predictive distributions $\{\Mmu^i\}_{i\in N}$, each with a specified conditional completion, constitutes a \textit{subjective embedded equilibrium} if:
\begin{enumerate}
    \item \textbf{Subjective Best Response.} Each agent's policy $\Mvp_i$ is an embedded best response with respect to $\Mmu^i$ and its completion.
    \item \textbf{Uncontradicted Beliefs.} The agent's predictive distributions $\Mmu^i(\Turn_{<t}^i)$ are identical to the ground-truth personal distribution $\big(\Mge^i\big)^{\Mgp^i}(\Turn_{<t}^i)$ induced by the multi-agent environment $\MultiAgentMge$ and policies of all agents.
\end{enumerate}
\end{definition}

The SEE concept can rationalize behaviors unattainable under classical Nash equilibria, for example cooperation in the twin prisoner's dilemma, as detailed below.

\begin{example}[Cooperation in the Twin Prisoner's Dilemma]
\label{example:cooperation-sde}
Consider two embedded agents playing a single-round Prisoner's Dilemma with actions $C$ and $D$, indicating `cooperating' and `defecting' respectively. Let their subjective models $\Mmu^i$ be the limit of a sequence $\Mmu^i_r$ where $\Mmu^i_r(\text{C}, \text{C}) = 1-\epsilon_r$ and $\Mmu^i_r(\text{D}, \text{D})=\epsilon_r$, with $\epsilon_r \to 0$. Such an agent believes "I cooperate if and only if my opponent cooperates." Given this belief, cooperation is the rational best response. Since both agents cooperate, their belief that mutual cooperation occurs with probability 1 is not contradicted by reality. Thus, mutual cooperation is a subjective embedded equilibrium. 
\hfill $\diamondsuit$
\end{example}

Let us revisit \cref{example:bayesian-agents-cooperation}, and consider the tail game of length one of a single round of the prisoner's dilemma after timestep $t$. \cref{example:bayesian-agents-cooperation} showed that for large enough $\alpha$, both agents end up cooperating forever after some finite time $T$. Hence, the tail policies $\MgpTailM^i$ of both agents converge to pure cooperation, and their predictive models converge to the ground-truth distribution. As a consequence, in the limit of time to infinity, the agents not only converge to an $\epsilon$-SEE, but also to the SEE covered in \cref{example:cooperation-sde}. 

The subjective Nash equilibrium is a special case of the SEE: when an agent's beliefs are decoupled, an SEE is also a subjective Nash equilibrium.

\begin{proposition}\label{prop:decoupled-beliefs-se-ne}
If a set of policies $\{\Mvp^i\}_{i\in [N]}$ and subjective models $\{\Mmu^i\}_{i\in [N]}$ form a subjective embedded equilibrium where the completed conditionals are decoupled, i.e., $\Mmu^i(a^{-i} \mid \Rgh, a^i) = \prod_{j\neq i}\Mmu^i(a^j \mid \Rgh)$, then this also constitutes a subjective Nash equilibrium. Consequently, there exists a Nash equilibrium that induces the same distribution over histories.
\end{proposition}
\begin{proof}
    See \cref{app:prop-decoupled-beliefs-se-ne}
\end{proof}
The above result also holds for the $\epsilon$-variants (cf. \cref{app:prop-decoupled-beliefs-se-ne}).

This proposition leads to an important corollary: When embedded agents are endowed with decoupled beliefs, they are behaviorally equivalent to decoupled Bayes-optimal agents (cf. \cref{prop:behavior-decoupled-beliefs}) and hence their long-term behavior recovers the classical results of Kalai and Lehrer (cf. \cref{thrm:kalai-convergence-repeated-games}). This serves as a crucial sanity check, demonstrating that our framework correctly subsumes the decoupled case.

\begin{corollary}[Convergence of embedded Bayes-optimal agents with decoupled beliefs to $\epsilon$-Nash]\label{cor:convergence-to-eps-nash}
Consider an infinitely repeated game with perfect monitoring, and embedded Bayes-optimal agents starting from decoupled prior beliefs $w(\Mvu)$, i.e., there exists a probability measure $\tilde{w}^i \in \Delta (\prod_{j}\calMpol^j)$ such that for all $\Mvu \in \calMuni$ it holds that
    $$w^i(\Mvu) = \sum_{\MultiAgentMvp \in \prod_{j}\calMpol^j} \tilde{w}^i(\MultiAgentMvp) \delta\left(\Mvu = \MultiAgentMuPiI \right), \quad \text{and} ~~ \tilde{w}^i(\MultiAgentMvp) = \prod_{j\in N}\tilde{w}^i(\Mvp^j)\,.$$
    with $\delta(\Mvu = \MultiAgentMuPiI)$ the indicator function specifying whether $\Mvu$ is equal to the personal universe $\MultiAgentMuPiI$ of agent $i$ originating from marginalizing the joint universe $\bar{\Mve}^{\bar{\Mvp}}$ induced by $\MultiAgentMvp$. If their beliefs satisfy the grain-of-truth and grain-of-uncertainty properties, the agents converge to playing an $\epsilon$-subjective embedded equilibrium which, by \cref{prop:decoupled-beliefs-se-ne}, is also an $\epsilon$-subjective Nash equilibrium. 
    
    \citet[Theorem 1]{kalai1993subjective} shows equivalence between $\epsilon$-subjective Nash and $\epsilon$-Nash equilibria. Therefore, for any $\epsilon>0$, their exists a time $T(\epsilon)$ such that for all $t \geq T(\epsilon)$, with $\MultiAgentMge^{\MultiAgentMvp}$-probability greater than $1-\epsilon$ over $\MAHistM$, the trajectory distribution $\bar{\Mge}_{\MAHistM}^{\MultiAgentMvp}$ induced by the tail policies of the agents are $\epsilon$-close to an $\epsilon$-Nash equilibrium in the tail game starting from time $t$.
\end{corollary}
\begin{proof}
    See \cref{app:proof-convergence-to-eps-nash}
\end{proof}

\subsubsection{Common knowledge solution concepts}
In the preceding section, we developed the subjective embedded equilibrium to characterize the convergence behavior of embedded agents as optimal with respect to their internal, \textit{subjective beliefs} about the world, which can differ from the subjective beliefs of other agents. In this section, we develop the embedded equilibrium, a \textit{common knowledge solution concept} that assesses whether this converged behavior is optimal with respect to the \textit{ground-truth} environment, the actual policies of the other agents, and a commonly agreed upon method for evaluating counterfactual behaviors. Evaluating counterfactuals of what would happen when the focal agent changes its behavior requires making assumptions. In classical game theory, the assumption is that of decoupledness: the policies of other agents remain unaltered. In the embedded agency setup, such a decoupledness assumption can be inaccurate, as the agents' policies can be functionally related. 

Instead of defaulting to the decoupled assumption for evaluating counterfactuals, we explicitly encode the possible functional relations between agents and the environment. We define a set $\calMuni^{\textrm{allowed}}$ of \textit{allowable multi-agent universes} $\bar{\lambda}: (\prod_i \calA^i \times \prod_i \calE^i)^* \to [0,1]$ that satisfy specific functional relations imposed by nature (e.g., genetic kinship) or an external designer (e.g., a self-play constraint). Agents maintain common knowledge of the multi-agent environment $\bar{\mu}$, each other's policies $\bar{\pi}$, the set $\calMuni^{\textrm{allowed}}$, and an externally provided prior probability distribution $q \in \Delta \calMuni^{\textrm{allowed}}$ (where $q(\bar{\Mvu})>0$ for all $\bar{\Mvu} \in \calMuni^{\textrm{allowed}}$). 

This dependency distribution quantifies the likelihood of allowable universes, enabling agents to uniquely and consistently evaluate counterfactual deviations. The dependency distribution $q(\bar{\Mvu})$ induces a joint mixture universe $q: (\prod_i \calA^i \times \prod_i \calE^i)^* \to [0,1]$, defined as
\begin{align}
    q(\MATurn_*):=\sum_{\bar{\Mvu}\in\calMuni^{\textrm{allowed}}}q(\bar{\Mvu})\bar{\Mvu}(\MATurn_*)\,,
\end{align}
We assume that $\calMuni^{\textrm{allowed}}$ and prior $q$ are such that the resulting mixture universe $q(\MATurn_*)$ is fully supported on all $\MATurn_* \in (\prod_i \calA^i \times \prod_i \calE^i)^*$

\textbf{Counterfactual measure.}
We derive a counterfactual measure to quantify the expected responses of other agents and the environment when agent $i$ contemplates an action off the play-path. Instead of assuming that the ego-policy $\Mvp^i$ can be unilaterally changed without altering the rest of the universe, we use conditionals $q(e^i \mid \Turn_{*}^i, a^i)$ originating from the externally provided dependency distribution $q(\bar{\Mvu})$ whenever the ground-truth universe conditionals $\bar{\Mge}^{\bar{\Mvp}}(e^i \mid \Turn_{*}^i, a^i)$ are undefined (i.e., whenever $\bar{\Mge}^{\bar{\Mvp}}(\Turn_{*}^i, a^i)=0$).

Let the ground-truth personal universe for agent $i$ be $\Mgu^i$. This is obtained from $\bar{\Mge}^{\bar{\Mvp}}(\MATurn_*)$ by marginalizing out the histories $(\Turn_*^j)_{j\neq i}$ of the other agents. The conditional completion $p({\Mgu}^i,q)$ of $\Mgu^i$ by $q$ can now be constructed as the limit of a sequence $(p_r)_r$:
\begin{align}
    p_r:= (1-\epsilon_r)\Mgu^i + \epsilon_r q, \quad \text{with} ~~ \lim_{r\to\infty}\epsilon_r = 0\,,
\end{align}
leading in the limit of $r \to \infty$ to the following completed conditionals:
\begin{align}
    p({\Mgu}^i,q)(e^i \mid \Turn_{*}^i, a^i) &:= \Mgu^i(e^i \mid \Turn_{*}^i, a^i) && \text{if } \Mgu^i(\Turn_{*}^i, a^i) > 0\,, \\
    p({\Mgu}^i,q)(e^i \mid \Turn_{*}^i, a^i) &:= q(e^i \mid \Turn_{*}^i, a^i) && \text{otherwise.}
\end{align}

This conditional completion mathematically clarifies the reasoning of an agent in an embedded setting: when contemplating a deviation to an action with zero probability under its current policy, the agent deduces that this action implies the world must be governed by a different universe than the ground-truth $\Mgu^i$. It therefore utilizes the dependency distribution $q$---which reflects the underlying functional relations of the system---to evaluate the counterfactual consequences of its deviation. 

We define the embedded equilibrium as a mutual best response against this counterfactual measure.

\begin{definition}[Embedded Equilibrium]\label{def:ode}
A set of policies $(\Mgp^i)_{i=1}^N$ constitutes an embedded equilibrium for the multi-agent environment $\bar{\Mge}$ and prior probability $q$ over the set of allowable universes $\calMuni^{\textrm{allowed}}$, if and only if (i) $\MultiAgentMge^{\MultiAgentMvp} \in \calMuni^{\textrm{allowed}}$ and (ii) for each agent $i$, its policy $\Mgp^i$ is an embedded best response with respect to the conditional completion $p({\Mgu}^i,q)$.
\end{definition}

\begin{example}[Mutual cooperation is an embedded equilibrium in single-shot PD]\label{example:ode-twinpd}
If we look closely at \cref{example:cooperation-sde}, we can see that the subjective models $\Mmu^i$ are the limits of the conditionals of the same distribution $p_r$ with $p_r(\text{C}, \text{C}) = 1-\epsilon_r$ and $p_r(\text{D}, \text{D})=\epsilon_r$, where $\epsilon_r \to 0$. Furthermore, the conditionals of $p_r$ in the limit of $r \to \infty$ are a conditional completion of the ground-truth action distribution $\Mgu(C,C)=1$. Therefore, the resulting subjective embedded equilibrium is also an embedded equilibrium.
\hfill $\diamondsuit$
\end{example}

\begin{remark}[Comparison to Correlated Equilibrium]
It is crucial to distinguish the embedded equilibrium from the correlated equilibrium (cf. \cref{def:corr-eq}). In a correlated equilibrium, agents condition their actions on private messages $m^i$ from a common device. While this creates statistical correlation, the reasoning remains decoupled. An agent contemplating a deviation from the recommended action does not believe its deviation will alter the other agents' policies, as their actions are conditioned on their own private messages. In an embedded equilibrium, when contemplating an action $a^i$ with zero probability under the current policy, the dependency distribution $q$ takes possible functional dependencies among agent's policies into account. Hence, the considered policy of the other agents can change for different actions $a^i$, reflecting the functional dependencies. This is what enables equilibria like cooperation in the Twin Prisoner's Dilemma, which is inaccessible to correlated equilibria.
\end{remark}

Every embedded equilibrium is, by definition, also a subjective one. This is because we can set each agent's subjective model $\Mmu^i$ in the SEE definition (\cref{def:sde}) to be identical to the ground-truth personal universe $\Mgu^i$ with its conditional completion derived from the dependency distribution $q$, as specified in the EE definition (\cref{def:ode}). Since the EE already requires $\Mgp^i$ to be a best response to this completed $\Mgu^i$ and the beliefs are uncontradicted by construction, all conditions for an SEE are met. The converse, however, is not true.

\begin{proposition}\label{prop:diff-sde-ode}
    Each embedded equilibrium is also a subjective embedded equilibrium. The converse is not true; there exist subjective embedded equilibria for which there is no embedded equilibrium that induces the same distribution over histories.
\end{proposition}
\begin{proof}
    See \cref{app:proof-diff-sde-ode} for a proof by counterexample.
\end{proof}

This marks a key difference from the decoupled case in repeated games. As reviewed in \cref{sec:intermezzo-kalai}, \citet[Proposition 1]{kalai1993subjective} established an equivalence: every subjective Nash equilibrium in a repeated game induces the same history distribution as some Nash equilibrium. For embedded agents, this equivalence breaks down; the set of subjective equilibria (SEEs) is strictly larger than the set of common knowledge equilibria (EEs).

It is easy to see that each Nash equilibrium is also an EE when we take as dependency distribution $q$ the decoupled personal environment $\Mge^i(e^i \mid \Turn^i_{<t},a^i)$, which in case of repeated games is equal to the fixed policies of the other agents. The converse is not true, as shown by \cref{example:ode-twinpd} where mutual cooperation on the prisoner's dilemma is an EE but not a Nash equilibrium. 

In our review of the decoupled setting, we saw that Bayesian learners do not converge to an exact Nash equilibrium in finite time, but rather to an $\epsilon$-subjective Nash equilibrium, which in turn corresponds to an $\epsilon$-Nash equilibrium (\cref{thrm:kalai-convergence-repeated-games}). To provide an analogous convergence target for embedded agents, we define an $\epsilon$-variant of the embedded equilibrium.

\begin{definition}[$\epsilon$-Embedded Equilibrium]\label{def:eps-ode}
A set of policies $(\Mgp^i)_{i=1}^N$ constitutes an $\epsilon$-embedded equilibrium for the multi-agent environment $\bar{\Mge}$ and prior probability $q$ over the set of allowable universes $\calMuni^{\textrm{allowed}}$, if and only if (i) $\MultiAgentMge^{\MultiAgentMvp} \in \calMuni^{\textrm{allowed}}$ and (ii) for each agent $i$, its policy $\Mgp^i$ is an $\epsilon$-embedded best response with respect to the conditional completion $p({\Mgu}^i,q)$:
\begin{align*}
    V_{p({\Mgu}^i,q)^{\Mvp}}(\varepsilon) &\geq \max_{\Mvp} V_{p({\Mgu}^i,q)^{\Mvp}}(\varepsilon) - \epsilon \\
\end{align*}
\end{definition}

Note that, similar to the $\epsilon$-Nash equilibrium, the $\epsilon$ condition for an $\epsilon$-EE applies to the best response. This contrasts with the $\epsilon$-SEE (\cref{def:eps-sde}), where the $\epsilon$ condition applies to the \textit{accuracy of the beliefs} (i.e., beliefs are $\epsilon$-close to the truth), while the policy must be an exact best response to those $\epsilon$-close beliefs.

\begin{theorem}[Convergence to $\epsilon$-Embedded Equilibrium]\label{theorem:convergence-to-ode}
    Let $\{\Mmu^i\}_{i\in N}$ be Bayesian mixture universes satisfying the grain-of-uncertainty and grain-of-truth conditions, and let $\{\Mgp^i\}_{i\in N}$ be the corresponding embedded best response policies in an infinitely repeated game with perfect monitoring. If the mixtures $\{\Mmu^i\}_{i\in N}$ of all the players are the same mixture, i.e., $\Mmu^i=\Mmu^j$ $\forall i,j\in N$,\footnote{Note that due to perfect monitoring, the mixture universes of all agents are defined over sequences of joint actions, and hence use the same set $\bar{\calA}^*$ over histories.} then, for each $\epsilon > 0$, there exists a finite time $T(\epsilon)$ such that for all $t \geq T(\epsilon)$, with $\MultiAgentMge^{\MultiAgentMvp}$-probability greater than $1-\epsilon$ over $\RghM$, the tail distribution $\bar{\Mge}^{\bar{\Mgp}}_{\bar{a}_{<t}}$ induced by the tail policies $\MgpTailM[\bar{a}_{<t}]^i$ is $\epsilon$-close to the distribution induced by some policies constituting an $\epsilon$-embedded equilibrium in the tail game starting at time $t$.
\end{theorem}
\begin{proof}
    See \cref{app:proof-convergence-to-ode}
\end{proof}

\begin{example}[Convergence to a cooperative $\epsilon$-EE for the Twin Prisoner's Dilemma]
Since the priors that are used in \cref{example:bayesian-agents-cooperation} are the same for both agents, \cref{theorem:convergence-to-ode} implies that the convergence to mutual cooperation for sufficiently large $\alpha$ can also be seen as convergence to an $\epsilon$-embedded equilibrium. To elaborate, \cref{example:bayesian-agents-cooperation} showed that the agents' tail policies converge to mutual cooperation. As established in \cref{example:cooperation-sde}, mutual cooperation \textit{is} an embedded equilibrium for a dependency distribution $q$ that reflects the perfect functional similarity (e.g., the limit of $q_r(\text{C}, \text{C}) = 1-\epsilon_r$ and $q_r(\text{D}, \text{D})=\epsilon_r$). Since an exact EE is also an $\epsilon$-EE for any $\epsilon > 0$, the converged behavior of the agents is indeed equivalent to an $\epsilon$-embedded equilibrium.
\hfill $\diamondsuit$
\end{example}

\subsection{Equilibrium behavior of embedded Bayesian agents in the MAGRL setup}\label{sec:eq-behavior-magrl}
We now broaden our analysis from the structured setting of repeated games to the more general and realistic framework of multi-agent general reinforcement learning (MAGRL). This shift introduces partial observability and unknown environment dynamics, which have significant consequences for equilibrium behavior.

The solution concepts we introduced---the subjective embedded equilibrium (SEE) (\cref{def:sde}) and embedded equilibrium (EE) (\cref{def:ode}), along with their $\epsilon$-variants---are directly applicable to the MAGRL setting. The key difference is that the subjective mixture universes $\Mmu^i$ are now general models over personal histories $\HistI$ as defined in \eqref{eqn:embedded-bayes-mixture}, rather than the more constrained models over joint action histories used in the repeated games definition (\cref{def:embedded-agents-repeated-games}).

This move to the general MAGRL setup creates an interesting convergence: the subjective Nash equilibrium (SNE) and the subjective embedded equilibrium (SEE) become mathematically equivalent.

In the repeated games setting, SNE and SEE are distinct. The decoupled mixture environment $\Mme^i$ for an SNE (\cref{def:decoupled-agents-repeated-games}) was built on a model class enforcing the repeated game's causal structure, implying no causal link between an agent's current action $a_t^i$ and other agents' concurrent actions $a_t^{-i}$. Thus, $\Mme^i(a^{-i} \mid \RghM, a^i) = \Mme^i(a^{-i} \mid \RghM)$. The embedded mixture universe $\Mmu^i$ for an SEE (\cref{def:embedded-agents-repeated-games}), however, allowed for coupled beliefs, creating an \textit{informational} link. This meant $\Mmu^i(a^{-i} \mid \RghM, a^i) \neq \Mmu^i(a^{-i} \mid \RghM)$ was possible, enabling equilibria like cooperation in the Twin Prisoner's Dilemma.
    
In the MAGRL setting, the model class for a decoupled agent's mixture environment $\Mme^i$ is no longer restricted. It can now include environments $\Mve$ where an agent's action $a^i_t$ has a \textit{causal} influence on the percept $e^i_t$ (which could include information about $a^{-i}_t$). Therefore, a decoupled agent can have a mixture environment with $\Mme^i(e^i \mid \HistMI, a^i) \neq \Mme^i(e^i \mid \HistMI)$. An embedded agent's mixture universe $\Mmu^i$ can model this same dependency, either as a causal link within its universe hypotheses or as an informational link via coupled beliefs.
Since both decoupled (SNE) and embedded (SEE) agents can now model a dependency of the percept $e^i$ on the action $a^i$, the set of possible equilibria they can converge to is mathematically identical. A philosophical distinction remains: an SNE rationalizes this dependency as a purely \textit{causal} feature of the environment, whereas an SEE can rationalize it as either a causal link or an \textit{informational} one arising from functional similarities.

The following example demonstrates this equivalence, showing that a decoupled Bayesian agent in a MAGRL setting can converge to a cooperative SNE, similar to the embedded counterpart discussed in \cref{example:bayesian-agents-cooperation}, but does so by forming a belief in a \textit{causally incorrect} environment compared to the ground-truth repeated game structure.

\begin{example}[Convergence to a Cooperative $\epsilon$-SNE]\label{example:bayesian-agents-cooperation-sne}
To illustrate that in the MAGRL setup, the $\epsilon$-SEE and $\epsilon$-subjective Nash equilibrium are mathematically equivalent, we revisit \cref{example:bayesian-agents-cooperation} on the iterated prisoner's dilemma with discount $\gamma=0$, but now convert it to a MAGRL setting with percepts equal to the action of the other agent and resulting reward: $e^i = (a^{-i}, r^i)$. Now consider \emph{decoupled} Bayesian agents maintaining a mixture environment $\Mme^i(e^i \mid \HistI,a^i)$. Leveraging the policy class $\calMpol$ and corresponding prior $\tilde{w}$ introduced in \cref{example:bayesian-agents-cooperation}, we construct the environment model class $\calMenv$ and corresponding prior as follows. We take $\calMenv:=\calMpol\cup \{\Mve_{\textrm{copy}}\}$, where environments $\nu$ coming from $\calMpol$ are the opponent's policy combined with the ground-truth reward function, i.e., $\nu(e^i \mid \HistI, a^i):=\Mvp(a^{-i} \mid \HistI) \delta(r^i = r(a^i, a^{-i}))$\footnote{Note that the opponent's policy $\Mvp(a^{-i} \mid \HistI)$ interprets the ego-agent's history $\HistI$ from its own point of view, i.e., it will use the percepts $e^i$ to derive its own actions $a^{-i}$, and use $a^i$ to create its own percepts $e^{-i}$.} where we remind the reader that $e^i = (a^{-i}, r^i)$, and where $\Mve_{\textrm{copy}}$ is the environment that copies the action $a^i$ combined with the ground-truth reward function, i.e., $\Mve_{\textrm{copy}}(e^i \mid \HistI, a^i):= \delta(a^{-i} = a^i)\delta(r^i = r(a^i, a^{-i}))$. Now we construct the prior $w(\Mve)$ as follows:
\begin{align*}
 w(\nu):= \begin{cases}
 (1-\alpha) \tilde{w}(\nu) & \textrm{if} ~~ \nu \in \calMpol\,, \\
 \alpha &\textrm{if}~~ \nu = \nu_{\textrm{copy}}\,,
 \end{cases}
\end{align*}
where we use shorthand $\tilde{w}(\nu)$ for $\tilde{w}(\Mvp)$ with $\Mvp$ the opponent's policy modeled by $\nu$. Similar to \cref{example:bayesian-agents-cooperation}, we define:
\begin{align*}
 m(\Turn^i_{1:t}) &:= \sum_{\Mvp \in \calMpol}  \tilde{w}(\Mvp)\prod_{k=1}^{t}\Mvp(a^i_k \mid \HistMI[k]) \\
 m_{\infty}^{\textrm{defect}}&=\lim_{k\to\infty} m((D,D)^k)\,.
\end{align*}
The decoupled Bayesian agents are guaranteed to converge to a cooperative $\epsilon$-SNE when $\alpha > \frac{m_{\infty}^{\textrm{defect}}}{1+ m_{\infty}^{\textrm{defect}}}$ (through a similar derivation as for \cref{example:bayesian-agents-cooperation}, see \cref{app:example-bayesian-agents-cooperation-sne} for more details). Similarly, the decoupled Bayesian agents are guaranteed to converge to an $\epsilon$-SNE of mutual defection when $\alpha <\frac{m_{\infty}^{\textrm{defect}}}{1+ m_{\infty}^{\textrm{defect}}}$. 

This illustrates the mathematical equivalence: the decoupled agent converges to cooperation, just as the embedded agent did in \cref{example:bayesian-agents-cooperation}. However, the \textit{reasoning} is different. The embedded agent converged by reasoning "my opponent is likely an identical copy, so my choice to cooperate is \textit{evidence} that they will too." The decoupled agent converges by reasoning "the environment is likely the `copy' environment, so my choice to cooperate will \textit{cause} my opponent to cooperate." This belief in the $\Mve_{\textrm{copy}}$ environment is, from an objective standpoint knowing the structure of the considered repeated game, causally incorrect, as the opponent is a separate agent, not a feature of the environment. While in the embedded setup, a large $\alpha$ can be justified by Occam's razor (a universe with two identical agents is simpler), justifying a large $\alpha$ for the \textit{ad-hoc} $\Mve_{\textrm{copy}}$ environment in the decoupled setup is more difficult.
\hfill $\diamondsuit$
\end{example}

While SNE and SEE become mathematically equivalent in the MAGRL setting, the common knowledge solution concepts---Nash equilibrium (NE) and embedded equilibrium (EE)---remain distinct. A Nash equilibrium (\cref{def:nash-eq}) requires a best response against the \textit{ground-truth} personal environment $\Mge^i$, adhering to the ground-truth causal structure of the environment. An EE (\cref{def:ode}) allows for a best response against a personal universe $\Mgu^i$ whose counterfactuals are defined by a dependency distribution $q$, explicitly accounting for functional dependencies. Thus, cooperation in the Twin Prisoner's Dilemma is an EE but never a Nash equilibrium, even when the prisoner's dilemma is modeled in the MAGRL setting.

We now turn to the convergence behavior of embedded Bayesian agents in the MAGRL setup. As discussed in \cref{sec:intermezzo-kalai}, partial observability means that agents' tail policies are conditioned on their personal histories $\HistMI$, which are private but correlated. This correlation is captured by a correlation device, leading us to correlated equilibrium concepts. We therefore introduce the \textit{subjective correlated embedded equilibrium} (SCEE).

\begin{definition}[$\epsilon$-Subjective Correlated Embedded Equilibrium]\label{def:eps-scde}
 A set of policies $(\Mvp^i)_{i=1}^N$ with $\Mvp^i: M^i\times (\TurnSet^i)^* \to \Delta \calA^i$ and subjective mixture universes $(\Mmu^i(\HistI \mid m^i))_{i= 1}^N$ is an $\epsilon$-\textit{subjective correlated embedded equilibrium} w.r.t. the multi-agent environment $\bar{\Mge}$ and correlation device $(M, p)$ if the following two conditions hold:
 \begin{enumerate}
 \item \textbf{Subjective Best Response.} Each agent's policy $\Mvp^i$ is an embedded best response (cf. \eqref{eqn:embedded-br}) w.r.t. its subjective mixture universe $\Mmu^i$ for each history and message $m^i$ with $p(m^i)>0$.
 \item \textbf{$\epsilon$-Uncontradicted Beliefs.} With probability greater than $1-\epsilon$, a message $\bar{m}$ is sampled such that for each agent $i$, its subjective beliefs $\Mmu^i(\HistI \mid m^i)$ are $\epsilon$-close in total variation distance to the ground-truth personal universe $\MultiAgentGeGpI(\HistI \mid m^i)$.
 \end{enumerate}
\end{definition}

The exact (non-$\epsilon$) subjective correlated embedded equilibrium is obtained by setting $\epsilon=0$, requiring uncontradicted beliefs for all messages $m^i$ with $p(m^i)>0$. Just as decoupled Bayesian agents converge to an $\epsilon$-SCE (\cref{thrm:kalai-convergence-repeated-games}), embedded Bayesian agents satisfying the grain-of-truth property converge to an $\epsilon$-SCEE.

\begin{theorem}[Convergence to $\epsilon$-Subjective Correlated Embedded Equilibrium]\label{thrm:convergence-to-sde-correlated}
Let $\Mgp^i$ be the policies of embedded Bayes-optimal agents in a multi-agent environment $\bar{\Mge}$, using Bayesian mixture universes $\Mmu^i$ that satisfy the grain-of-truth and grain-of-uncertainty properties. It holds that for each $\epsilon > 0$, there exists a finite time $T(\epsilon)$ such that for all $t \geq T(\epsilon)$, the tail policies $\MgpTailM^i$ and tail posterior beliefs $\MmuTailM^i$ constitute an $\epsilon$-correlated subjective embedded equilibrium in the correlated tail game starting at time $t$ with correlation device $(\MATurnSet^{t-1}, \bar{\Mge}^{\bar{\Mgp}})$.
\end{theorem}
\begin{proof}
    See \cref{app:proof-convergence-to-sde-correlation}.
\end{proof}

A final, critical point is that the link between subjective and common knowledge equilibria, which was already weaker for embedded agents in the repeated games setting (\cref{prop:diff-sde-ode}), breaks down almost entirely in the general MAGRL setup. The reason, as noted in \cref{sec:intermezzo-kalai} for decoupled agents, is the introduction of uncertainty about the \textit{reward function} itself (which is part of the percept $e^i$), as well as partial observability. The "uncontradicted beliefs" condition only constrains beliefs on the play-path, leaving off-path conditionals---especially those concerning rewards---unconstrained by evidence. This opens the door to "dogmatic beliefs": an agent might believe that any deviation from its current policy will result in catastrophic reward, making its suboptimal policy a best response to its own flawed subjective model \citep{leike2015bad}. Since the agent never deviates, this inaccurate belief is never corrected. This leads to the following triviality result, which holds for both SEEs and SNEs due to their equivalence in the MAGRL setting.

\begin{proposition}\label{prop:dogmatic-beliefs}
For any ground-truth multi-agent environment $\bar{\Mge}$ and any set of deterministic policies $(\Mgp^i)_{i=1}^N$, there exists a set of mixture universes $(\Mmu^i)_{i=1}^N$ with corresponding conditional completions such that their combination is a subjective embedded equilibrium.

Similarly, for any ground-truth multi-agent environment $\bar{\Mge}$ and any set of deterministic policies $(\Mgp^i)_{i=1}^N$, there exists a set of mixture environments $(\Mme^i)_{i=1}^N$ such that their combination is a subjective Nash equilibrium.
\end{proposition}
\begin{proof}
 See \cref{app:proof-dogmatic-beliefs}.
\end{proof}

This triviality result underscores a fundamental challenge for any Bayesian agent in a complex environment: rational learning, even under the grain-of-truth assumption, is not sufficient to guarantee convergence to an objectively optimal equilibrium. It highlights the necessity of robust exploration to test and correct flawed off-path beliefs. Bayesian agents with sufficiently broad priors naturally incorporate a form of principled exploration through the \textit{value of information} \citep{howard1966information, chalkiadakis2003coordination}, where an agent is intrinsically motivated to take actions that reduce uncertainty about the world if that information is expected to improve future rewards. However, as prior work has shown, this passive form of exploration is often insufficient to overcome dogmatic beliefs about catastrophic outcomes \citep{orseau2010optimality, leike2015bad}. This has motivated the use of more active exploration strategies, such as Thompson sampling \citep{leike2016thompson}. Translating and adapting these approaches to the coupled belief structures of embedded agents presents a promising avenue for future research.

\subsection{Embedded Bayesian agents with finite planning horizons}\label{sec:eq-behavior-k-step}
The previous results considered embedded Bayes-optimal agents that perform infinite-horizon planning. We end this section by investigating the $k$-step planner embedded Bayesian agents introduced in \cref{sec:k-step-planning}. \citet{catt2023self} showed that Self-AIXI, which does 1-step ahead optimal planning, converges to an infinite-horizon optimal planner under additional assumptions on the mixture environment. The intuition of why 1-step planning with terminal values converges to infinite-horizon planning is the following: Initially, the policy implements $1$-step optimal planning; when the self-model $\Mmu(a\mid \Hist)$ distills this 1-step planning during learning (i.e., Bayesian belief updates), the terminal value $V_\Mmu(\Hist)$ represents a $1$-step planning policy. Hence, adding $1$-step planning to that results in an effective $2$-step planning policy. This process can be recursively applied to extend the planning horizon further. In the following, we adapt their approach to our embedded Bayesian agent setup with $k$-step planning, and investigate the consequences for multi-agent learning. 

Since $k$-step planning is not a perfect approximation of the infinite-horizon optimal planning for any finite time, we need to consider $\delta$-best responses. Hence, we extend the $\epsilon$-SCEE solution concept to include $\delta$-best responses, allowing us to characterize the convergence behavior of embedded Bayesian agents implementing $k$-step planning.

\begin{definition}[$(\epsilon,\delta)$-Subjective Correlated Embedded Equilibrium ($(\epsilon,\delta)$-SCEE)]\label{def:eps-delta-scde}
 A set of policies $(\Mvp^i)_{i=1}^N$ with $\Mvp^i: M^i\times (\TurnSet^i)^* \to \Delta \calA^i$ and subjective mixture universes $(\Mmu^i(\HistI \mid m^i))_{i= 1}^N$ is an $\epsilon$-\textit{subjective correlated embedded equilibrium} w.r.t. the multi-agent environment $\bar{\Mge}$ and correlation device $(M, p)$ if the following two conditions hold:
 \begin{enumerate}
 \item \textbf{Subjective $\delta$-Best Response.} Each agent's policy $\Mvp^i$ is an embedded $\delta$-best response w.r.t. its subjective mixture universe $\Mmu^i$ for each history and message $m^i$ with $p(m^i)>0$:
 $$V_{(\Mmu^i)^{\pi^i}}(\HistI,m^i) \geq V^*_{\Mmu^i}(\HistI, m^i) - \delta \,.$$
 \item \textbf{$\epsilon$-Uncontradicted Beliefs.} With probability greater than $1-\epsilon$, a message $\bar{m}$ is sampled such that for each agent $i$, its subjective beliefs $\Mmu^i(\HistI \mid m^i)$ are $\epsilon$-close in total variation distance to the ground-truth personal universe $\MultiAgentGeGpI(\HistI \mid m^i)$.
 \end{enumerate}
\end{definition}

Building upon \citet{catt2023self}, the following theorem shows that under additional assumptions on $\calMuni$ and $\Mmu^i$, a group of $k$-step planner embedded Bayesian agents converge to $(\epsilon,\delta)$-subjective correlated equilibria.
\begin{theorem}\label{thrm:convergence-to-epsdelta-sde-correlated}
    Let $\Mvp^i$ be the policies of embedded Bayesian agents in a multi-agent environment $\MultiAgentMge$, implementing $k$-step planning (\cref{def:k-step-eba}) w.r.t. Bayesian mixture universes $\Mmu^i(\HistI)$ that satisfy the grain-of-truth and grain-of-uncertainty conditions, as well as the conditions that $\Mmu^i$ dominates $(\Mmu^i)^{\Mgp^i}$ and $(\Mmu^i)^{\Mgp^i}$ dominates the personal history distribution $(\Mge^i)^{\Mgp^i}$, and the following \emph{sensibly off-policy condition}: There exists a positive $\alpha<1/\gamma -1$ and $t_0$ such that for all $t \geq t_0$, it holds that
    \begin{align*}
        &\expect{(\Mmu^i)^{\Mvp^i}(\HistI)}{\max_{a^i} \sum_{e^i} \Mmu^i(e^i \mid \HistI a^i)[V^*_{\Mmu^i}(\HistI a^ie^i) - V_{\Mmu^i}(\HistI a^ie^i)]
        }\\
        &\quad\quad\quad\quad\quad\quad\quad\quad\quad\quad\quad\quad\quad\quad\quad\quad\quad\quad\quad\quad\quad\quad\quad\quad\leq (1+\alpha) \expect{(\Mmu^i)^{\Mvp^i}(\HistI a^i e^i)}{V^*_{\Mmu^i}(\HistI a^ie^i) - V_{\Mmu^i}(\HistI a^ie^i)}\,.
    \end{align*}
    
    Then it holds that for each $\delta > 0$ and $\epsilon>0$, the $\MultiAgentMge^{\bar{\Mgp}}$-probability over $\MAHistM$ that the tail Bayesian mixture universes $\MmuTailM^i$ and tail policies $\MgpTailM^i$ are an $(\epsilon,\delta)$-SCEE in the correlated tail game starting at timestep $t$ with correlation device $(\MATurnSet^{t-1}, \MultiAgentMge^{\bar{\Mgp}})$ converges to 1 as $t \to \infty$. 
\end{theorem}

\begin{proof}
    See \cref{app:proof-convergence-to-epsdelta-sde-correlated}
\end{proof}

The sensibly off-policy condition states that the expected optimality gap—defined as the difference between the optimal value $V^*_{\Mmu^i}$ and the self-model value $V_{\Mmu^i}$—does not become significantly worse when taking an arbitrary off-policy action, compared to sampling from the self-model \citep{catt2023self}. It remains an open problem in the literature to prove that this condition is satisfied for certain model classes and corresponding mixture models. In fact, in \cref{sec:emb-aixi-agents} we show that for Solomonoff mixture models, this condition is not satisfied.

When embedded Bayesian agents explicitly increase their $k$-step planning horizon over time, we can show convergence to the $(\epsilon,\delta)$-SCEE while avoiding the need for the sensibly off-policy condition. Furthermore, in \cref{sec:mupi}, we will consider embedded Bayesian agents that can only approximate the relevant value functions up to a small constant $\epsilon_t$. Let us combine both properties in the following agent definition.

\begin{definition}[$(k_t,\epsilon_t)$-embedded Bayesian agent]\label{def:approximate-eba}
    Let $t\mapsto k_t$ be a mapping representing a possibly variable planning horizon (which can vary from timestep to timestep), and let $t\mapsto \epsilon_t$ be a possibly variable level of accuracy of the approximate optimizer, where $\epsilon_t\in [0,1)$ for all $t$. A $(k_t,\epsilon_t)$-approximate embedded Bayesian agent with respect to mixture universe model $\Mmu$ is a policy $\Mvp$ which at time $t$ returns an action $a_t$ satisfying
    $$\Mvp(a_t|\HistM)> 0\quad\Rightarrow\quad Q^{k_t}_{\Mmu}(\HistM,a_t)\geq\max_{a'\in\mathcal{A}}Q_\Mmu^{k_t}(\HistM,a')-\epsilon_t\,.$$ 
\end{definition}

The following proposition shows that approximate embedded Bayesian agents with growing planning horizons and improving accuracy converge to $(\epsilon,\delta)$-Subjective Correlated Embedded Equilibrium.

\begin{proposition}\label{prop:convergence-of-approximate-eba-to-sde-correlated}
Let $\Mvp^i$ be the policies of embedded $(k_t^i,\epsilon_t^i)$-approximate Bayes-optimal agents in a multi-agent environment $\MultiAgentMge$, using Bayesian mixture universes $\Mmu^i$ that satisfy the grain-of-truth and grain-of-uncertainty properties. If $\lim_{t\to\infty}k_t^i=\infty$ and $\lim_{t\to\infty}\epsilon_t^i=0$ for all $i$, it holds that for each $\epsilon > 0$ and $\delta > 0$, there exists a finite time $T(\epsilon,\delta)$ such that for all $t \geq T(\epsilon,\delta)$, with $\MultiAgentMge^{\MultiAgentMvp}$-probability greater than $1-\epsilon$ over $\MAHistM$, the agents' tail policies $\MgpTailM^i$ and tail mixture universes $\MmuTailM^i$ constitute an $(\epsilon,\delta)$-SCEE in the correlated tail game starting from time $t$ with correlation device $((\MATurnSet)^{t-1}, \MultiAgentMge^{\bar{\Mgp}})$.
\end{proposition}
\begin{proof}
    See \cref{app:proof-convergence-of-approximate-eba-to-sde-correlated}
\end{proof}

\section{Embedded Universal Predictive Intelligence}\label{sec:mupi}

In this section we formally introduce our theory for E{\bf M}bedded {\bf U}niversal {\bf P}redictive {\bf I}ntelligence (MUPI). In \cref{sec:embedded-bayesian-agents}, we introduced embedded Bayesian agents using arbitrary hypothesis classes and mixture universes and showed that they can reason taking functional similarities into account. In \cref{sec:subj-emb-eq} we then showed that such embedded Bayesian agents converge to new types of solution concepts, assuming that they satisfy the grain-of-truth property. Apart from a few toy examples (e.g., \cref{example:bayesian-agents-cooperation}), we have not shown that the grain-of-truth property can be satisfied in cases where embedded Bayesian agents consider a large and interesting class of universes, such as the class of all computable universes. MUPI introduces an explicit model of a universally intelligent embedded Bayesian agent that reason over a hypothesis class including all computable universes while satisfying the grain-of-truth property, leading to consistent self prediction and mutual prediction among multiple MUPI agents. By leveraging algorithmic information theory and algorithmic probability theory, MUPI further formalizes embedded agency and functional similarities, while providing new tools to reason about mutual prediction and infinite-order theory of mind.

Following Hutter's Universal Artificial Intelligence framework \citep{hutter-aixi}, we treat policies, (multi-agent-)environments and universes as programs. When using a monotone Turing machine $M$ to describe a universe $\Mvu_M: \TurnSet^* \cup (\TurnSet^* \times\calA) \to [0,1]$, the countable set of all monotone Turing machines leads to a countable hypothesis class $\calMuni^{\text{LSCSM}}$ of all lower semicomputable semimeasures. Unfortunately, an embedded Bayesian agent using the hypothesis class $\calMuni^{\text{LSCSM}}$ combined with a lower semicomputable universal prior is itself not lower semicomputable, for the same reasons that AIXI, the Bayes-optimal agent over $\calMenv^{\text{LSCSM}}$ is itself not lower semicomputable \citep{leike2015computability,sterkenburg2019putnam}. Hence, universes that contain one or more embedded Bayesian agents over $\calMuni^{\text{LSCSM}}$ are not within the hypothesis class $\calMuni^{\text{LSCSM}}$, and consequently the grain-of-truth property is not satisfied, hindering the embedded Bayesian agents in making accurate predictions about the universe they live in. Building upon the framework of reflective oracles \citep{fallenstein2015reflective, fallenstein2015reflective_oracles, leike2016formal}, we aim to address the \textit{general grain-of-truth problem for embedded agency} defined below.

\begin{problem}[The general grain-of-truth problem for embedded agency - Informal]\label{prob:grain-of-truth}
Find a class of universes $\calMuni$ that 
\begin{enumerate}
    \item contains all computable universes;
    \item contains all universes that result from combining one or more embedded Bayesian agents (cf. Sections \ref{sec:embedded-br}--\ref{sec:k-step-planning}) that use a lower semicomputable universal prior over $\calMuni$ with a computable (multi-agent) environment.
\end{enumerate}
Furthermore, we do not want the considered class of universes to be too large: Any further enlargement beyond the class of all computable universes is considered a downside. Therefore, we should only include incomputable universes to the extent that they are necessary to help achieving the second goal. In other words, we do not want to consider universes that are "too incomputable".\footnote{We can formalize this by saying that the considered universes should ideally remain within a relatively low level of the arithmetic hierarchy \citep{odifreddi1989recursiontheory}.}
\end{problem}

In \cref{app:multi-agent-env,app:desiderata-embedded-uai-theory} we unpack the above problem in more detail, specifying how policies and (multi-agent) environments can be algorithmically combined to create universes, how embedded Bayesian agents fit into this picture and the requirements this imposes on $\calMuni$. 
As universes containing embedded Bayesian agents over $\calMuni^{\text{LSCSM}}$ are themselves not in $\calMuni^{\text{LSCSM}}$, it indicates that the class of lower semicomputable semimeasures is not large enough. Hence, we need to add some countable class of non-lower-semicomputable semimeasures to $\calMuni$. A classical approach in computer science to investigate non-computable objects is to provide Turing machines query access to an \textit{oracle} which is allowed to be incomputable, with a famous example being the Halting oracle answering queries of the form \textit{`Does machine $T$ halt on input $x$?'}. We follow this approach to solve the above grain-of-truth problem in two different ways, each using a different type of oracle. 

Previously, \citet{fallenstein2015reflective, fallenstein2015reflective_oracles} and \citet{leike2016formal} leveraged the reflective oracle framework to solve the grain-of-truth problem for decoupled Bayesian agents. In \cref{sec:reflective-oracles}, we extend their approach to the embedded Bayesian agent setting, to obtain a class $\calMuni^{\mathrm{RO}}$ solving the above grain-of-truth problem. Reflective oracles answer queries of the form $\langle T,x,p\rangle$ corresponding to \textit{``Is the probability that oracle machine $T^O$ outputs 1 when given input sequence $x$ greater than $p$?"}. As we will see later, a conceptual drawback of the Reflective Oracle framework is that when oracle machine $T^O$ does not halt on input $x$, the reflective oracle is allowed to redistribute this non-halting probability arbitrarily to outputs 0 or 1. Hence, when using reflective oracles to create a universal mixture, non-halting oracle machines contribute arbitrary output probabilities to the mixture.\footnote{This, however, is not a significant drawback for using the reflective universal mixture as a predictor if we only care about computable environments (or more generally, environments where the oracle machine does not loop forever): Despite the addition of arbitrary terms coming from the contribution of (possibly) non-halting machines, the convergence to accurate prediction of \cref{theorem:convergence-mixture-universe}, as well as the prediction loss bound of \cref{theorem:generalized-solomonoff-bound}, are still correct.} This is in contrast to Solomonoff induction (cf. Sec.~\ref{sec:background}), where non-halting probability mass does not contribute to the mixture.

In \cref{sec:rui}, we propose a new type of oracle to solve the grain-of-truth problem: the Reflective Universal Inductor (RUI) oracle, resulting in the model class $\calMuni^{\mathrm{RUI}}$. This RUI oracle answers queries of the form $\langle x,b, p\rangle$ corresponding to \textit{``Is the probability $\Mmu(b \mid x)$ greater than $p$?"} with $\Mmu$ a universal mixture distribution over oracle machines with access to the RUI oracle itself. Hence, we call $\Mmu$ the reflective universal inductor (RUI). In contrast to reflective oracles, the RUI excludes non-halting probability mass from its mixture distribution, and hence is not allowed to arbitrarily redistribute it to the valid outputs $0$ and $1$. This makes the RUI closer in spirit to Solomonoff induction compared to the universal inductors using reflective oracles. Conceptually, the RUI $\Mmu$ corresponds to the prediction model used by embedded Bayesian agents, making it easily relatable to practical machine learning settings where such prediction models are learned instead of obtained via universal Bayesian induction. Our reflective universal inductor framework makes the recursions arising from self or mutual prediction explicit: $\Mmu$ makes predictions about predictive entities using $\Mmu$ itself. However, these benefits of the RUI-oracle come at a cost: (i) as the RUI uses a specific universal prior, all Bayesian agents using the RUI-oracle use the same prior. In contrast, the reflective oracle framework allows different Bayesian agents to use different priors. (ii) Bayesian agents using RUI oracles are only allowed to make a bounded number of oracle calls for each executed action. While this still allows for infinite computations, it makes optimal infinite-horizon planning impossible. Thus, we can only formalize $k$-step optimal planning Bayesian agents (cf. \ref{sec:k-step-planning}) within the RUI-oracle framework, whereas the reflective oracle framework allows for infinite-horizon optimal planning. We stress that the RUI-oracle framework should be seen as an alternative framework to reflective oracles, providing some conceptual benefits but not without introducing other drawbacks. We leave it to future work to combine the benefits of both approaches. 

\begin{figure}[t]
    \centering
    % \phantom{\rule{0.5\linewidth}{5cm}}
    \includegraphics[width=\textwidth]{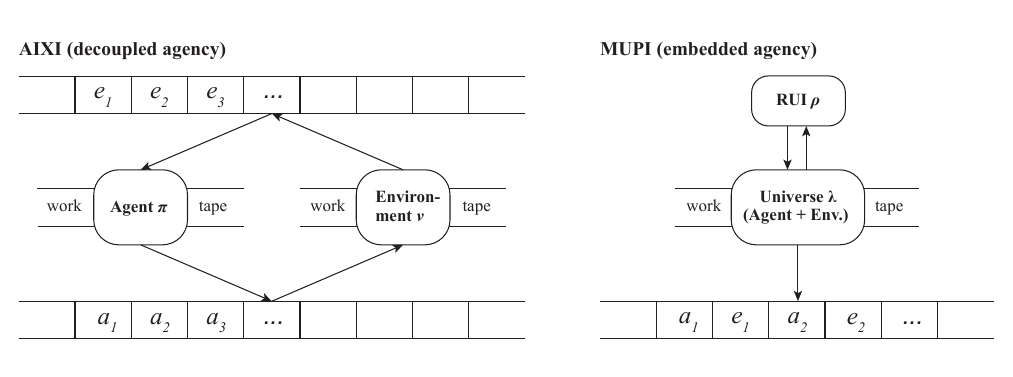}
    \caption{\textbf{Comparison of Decoupled Agency (AIXI) and Embedded Agency (MUPI).} For AIXI (left), the agent $\pi$ and the environment $\nu$ are decoupled programs, interacting via percepts ($e_t$) and actions ($a_t$) on separate tapes, with the percept tape acting as the output tape for the environment and the input tape for the agent, and vice versa for the action tape. In contrast, in MUPI (right), the agent and environment are unified into a joint universe $\Mvu$, with access to a reflective universal inductor (RUI) $\Mmu$, generating interleaved actions $a_t$ and observations $e_t$ on a single output tape.}
    \label{fig:mupi_setup}
\end{figure}

In \cref{sec:emb-aixi-agents} we introduce various \textit{embedded AIXI} agents over the model classes $\calMuni^{\mathrm{RUI}}$ and $\calMuni^{\mathrm{RO}}$, and show that they solve the grain-of-truth problem described in Problem \ref{prob:grain-of-truth}. \cref{fig:mupi_setup} shows the difference between such embedded AIXI agents and the classical decoupled AIXI agents. In \cref{app:desiderata-embedded-uai-theory} we delve deeper into the theory and formalize a few desiderata that a good theory of embedded universal intelligence should satisfy, and show that our formalism satisfies these desiderata.

Finally, in \cref{sec:mupi-structural-similarities}, we revisit the functional similarities introduced in \cref{sec:emb-bay-agents-structural-similarities} through the lens of algorithmic information theory, formalizing the intuition of functional similarities as the joint `compressibility' of agent and environment. Furthermore, we show that the Solomonoff universal prior, which incorporates the algorithmic complexity of the considered universes, is always coupled and contains universes with arbitrarily high degrees of functional similarities. 

\subsection{The reflective universal inductor}\label{sec:rui}
This section introduces the \textit{Reflective Universal Inductor (RUI)}, a new theoretical framework for constructing a general hypothesis class of universes that satisfies the \textit{grain-of-truth property} for embedded Bayesian agents using a mixture universe. To allow for a general yet parsimonious description, we model everything using binary strings ($x \in \mathcal{B}^*$, where $\mathcal{B} = \{0,1\}$) and machines that output bits. By using complete prefix-free encodings for the action set $\mathcal{A}$ and percept set $\mathcal{E}$, these binary strings can be unambiguously interpreted as the action-percept histories of an agent.

\begin{definition}[prefix free encoding]\label{def:prefix-free-encoding}
    Let $\calX$ be an arbitrary countable set. We say that a mapping $c:\calX\to\calB^*$ is a prefix-free encoding of $\calX$ if no codeword is a prefix of any other codeword, i.e., for every $x,x'\in\calX$ with $x\neq x'$, we have that $c(x)$ is not a prefix of $c(x')$.

    We say that the prefix-free encoding $c$ is complete if for every infinite binary string $s\in\calB^\infty$, there is one (and only one) element $x\in\calX$ such that $c(x)$ is a substring $s_{1:k}$ of $s$ for some $k>1$.
\end{definition}

Our construction proceeds in several steps. First, we formalize the notion of programs that can access non-computable information. We begin by defining \textit{Abstract Probabilistic Oracle Machines (APOMs)}, which are Turing machines $M$ equipped to query an unspecified oracle. This allows us to define a countable class $\calM$ of such machines. We then define a \textit{Probabilistic Oracle Machine (POM)} $M^\tau$ as an APOM, $M$, paired with a specific probabilistic oracle, $\tau$. The output distribution of a POM $M^\tau$, denoted $\Mvu_M^\tau$, is a semimeasure over binary strings and represents a concrete \textit{``universe''} in our framework, an instance of the universes $\Mvu$ discussed in \cref{sec:embedded-bayesian-agents,sec:subj-emb-eq}.

The goal is to model an embedded Bayesian agent whose predictive model is a universal Bayesian mixture, $\Mmu_\tau^w$, over the class of all possible universes, $\{\Mvu_M^\tau\}_{M \in \calM}$, using prior beliefs $w$. For this setup to satisfy the grain-of-truth property, a universe containing such an agent must itself be implementable by one of these POMs, $M^\tau$. This implies that the program $M$ must be able to simulate the agent's reasoning, which requires access to the universal Bayesian mixture, $\Mmu_\tau^w$. We solve this by designing the oracle $\tau$ to provide precisely this access: It will answer queries about the predictive distribution $\Mmu_\tau^w$.

This creates a crucial self-referential or \textit{``reflective''} loop. The oracle $\tau$ is defined in terms of the universal mixture $\Mmu_\tau^w$, but this mixture is itself an average over all POMs, whose behaviors are determined by $\tau$. The oracle's definition thus depends on a system of which it is a fundamental component. The existence of such a self-consistent fixed point is not guaranteed. The main contribution of this section is to formally define and prove the existence of this RUI-oracle. This construction yields a hypothesis class, $\calM_{\text{uni}}^{w,\tau-\mathrm{RUI}}$, that solves the general grain-of-truth problem (cf. Problem \ref{prob:grain-of-truth}), ensuring that the agent's predictions converge to the ground truth and thereby providing a solid foundation for the MUPI framework.

\subsubsection{Probabilistic oracle machines and universes.}
We define probabilistic oracles, abstract probabilistic oracle machines and their combination, probabilistic oracle machines. 

\begin{definition}[Probabilistic oracle]
    \label{def:probabilistic_oracle}
    A probabilistic oracle is a stochastic function $O^\tau$ parameterized by a probability map $\tau:\{0,1\}^*\to[0,1]$ as follows:\footnote{The interval $[0,1]$ denotes the interval of real numbers between 0 and 1 including the boundaries. We explicitly use the notation $\bbQ \cap [0,1]$ to indicate the interval of rational numbers between 0 and 1 when needed.} For each $x\in\mathcal{B}^*$,  $O^\tau(x)$ is a Bernoulli random bit satisfying $\mathbb{P}[O^\tau(x)=1]=\tau(x)$. We do not impose any computability assumption on the mapping $\tau$.
\end{definition}
 We can interpret $O^\tau$ as a probabilistic oracle that can be queried with a finite binary string $x$ and which can return a probabilistic binary answer $O^\tau(x)$. We may sometimes abuse the terminology and also refer to $\tau$ as the probabilistic oracle even though strictly speaking it is a deterministic function (but nevertheless describes the probabilistic behavior of a stochastic function).

\begin{definition}[APOM]\label{def:apom}
    An abstract probabilistic oracle machine (APOM) is a monotone Turing machine equipped with (i) an extra unidirectional binary read/write tape that we call the oracle tape, initialized with zeros, and a special instruction corresponding to querying the oracle; and (ii) an extra unidirectional binary read-only tape that we call the randomness tape, initialized with uniformly independent random bits, and a special instruction corresponding to reading a random bit from the randomness tape. We write $\calMapom$ to denote the class of all APOMs.
\end{definition}

Abstract probabilistic oracle machines do not specify what the machine does when an oracle-querying instruction is executed. The execution procedure of an APOM is well-defined (and it is similar to a monotone Turing machine) until an oracle-querying instruction is made, after which the behavior is not specified by definition of the APOM. To fully specify the execution procedure, we need to equip the abstract probabilistic oracle machine with a probabilistic oracle $\tau$.

\begin{definition}[POM]
    \label{def:probabilistic-oracle-machine}
    A probabilistic oracle machine (POM) is a pair $(M,\tau)$ of an abstract probabilistic oracle machine $M$ and a probabilistic oracle $\tau$. We denote such a POM as $M^\tau$. The execution procedure of $M^\tau$ can be described as follows:
    \begin{itemize}
        \item For all instructions other than oracle-querying instructions, the execution proceeds identically to probabilistic Turing machines leveraging the randomness tape to access random bits.
        \item If an oracle-querying instruction is encountered:
        \begin{itemize}
            \item We take the finite binary string to the left of the oracle tape's head and call it $x$.
            \item We query the oracle and get a random bit $b:=O^\tau(x)$ with $\mathbb{P}[O^\tau(x)=1]=\tau(x)$.
            \item We replace the entire oracle tape with zeros, except for the first bit which we replace with the oracle's response $b$.
            \item The oracle tape's head is returned to the tape's first position, which means that we can subsequently access the oracle's response by reading the bit at the oracle tape's head.
        \end{itemize}
        All oracle responses are assumed to be independent.
    \end{itemize}
\end{definition}

% \alex{TODO: change the name 'abstract universes' and 'universes' to 'restricted APOM (rAPOM)' and 'restricted POM' (rPOM) respectively, such that we can use the terminology 'universe' for the semimeasure $\lambda_M^\tau$ induced by the POM or rPOM $M$.}

For technical reasons to later construct our RUI-oracle, we need to define a restricted APOM (rAPOM) and restricted POM (rPOM) that constrain the amount of oracle calls the machine can make for each output bit. 
% Now we are ready to define abstract universes and universes --the combination of an abstract universe with a probabilistic oracle.

\begin{definition}[rAPOM]\label{def:rapom}
    A restricted abstract probabilistic oracle machine (rAPOM) is an abstract probabilistic oracle machine with the following computational constraints:
    \begin{enumerate}
        \item The execution of the abstract probabilistic oracle machine $M$ can be described as a (potentially infinite) sequence of stages where at the $i$-th stage the machine writes one bit on the output tape.
        \item For technical reasons which will become clear later\footnote{We will need this assumption in the proof of \cref{thm:rui} in \cref{app:thm_rui_proof}.}, we require that each stage must consist of three consecutive phases:
        \begin{itemize}
            \item[(i)] Phase 1: The machine computes a number $N_i$ without making any queries to the oracle.
            \item[(ii)] Phase 2: The machine performs a computation in which it is allowed to make at most $N_i$ queries to the oracle.
            \item[(iii)] Phase 3: The machine appends the output tape with a bit.
        \end{itemize}
    \end{enumerate}
    We require that the above structure is explicitly enforced so that we can computably certify it from a canonical binary representation $\langle M\rangle$ of $M$. This is needed so that we can define a canonical enumeration $M_1,\ldots$ of restricted abstract probabilistic oracle machines. We write $\calMrapom$ to denote the class of all rAPOMs.
\end{definition}

\begin{definition}[rPOM]\label{def:rpom}
    A restricted probabilistic oracle machine is a pair $(M,\tau)$ of an rAPOM $M$ together with a probabilistic oracle $\tau:\{0,1\}^*\to[0,1]$. We refer to such an rPOM as a $\tau$-rPOM to emphasize that it uses the $\tau$-oracle. 
\end{definition}

To obtain universe semimeasures $\Mvu$ over sequences, we use $\tau$-rPOMs without input tape.\footnote{To avoid the need to introduce a new type of rAPOMs without input tape, we keep using rAPOMs as defined in \cref{def:rapom}, and initialize the input tape with zeros.} Running such an rPOM $M^\tau$
% Running the resulting probabilistic oracle machine $M^\tau$ 
gives rise to a semimeasure $\Mvu_M^\tau:\mathcal{B}^*\to[0,1]$ describing the state of the output tape of $M^\tau$ as it runs. More precisely, for every $h\in\mathcal{B}^*$, $\Mvu_M^\tau(h)$ represents the probability that the POM $M^\tau$ executes at least $l(h)$ stages and $h$ is found to the left of the output tape's head at the end of the $l(h)$-th stage.
% As the input tape of a universe contains independent uniformly random bits, and the probabilistic oracle returns random bits, a universe $M^\tau$ induces a semimeasure $\lambda_M^\tau$ over binary sequences $x$. 
Using prefix-free encodings of actions in $\calA$ and percepts in $\calE$, such binary sequences can be interpreted as action-percept histories. Hence, these semimeasures $\Mvu_M^\tau$ are the direct counterpart of the universes $\Mvu$ used in \cref{sec:embedded-bayesian-agents,sec:subj-emb-eq}, using semimeasures instead of measures, to allow for universes that halt after outputting a finite sequence, or go into a non-halting loop without returning any further output bits. We use the term `universe' interchangeably for $M^\tau$ or $\Mvu_M^\tau$, and rely on context to distinguish between them. 

\begin{remark}
    \label{rem:canonical-representation-apom}
    By employing the same tools that are used to show the classical result in computability theory that every Turing machine can be canonically encoded as a binary string, we can also show that every APOM and rAPOM can be canonically encoded as a binary string. If $M$ is an rAPOM, we write $\langle M\rangle$ to denote its canonical representation as a binary string. We can also choose the canonical representation of rAPOMs in such a way that the set $\{\langle M\rangle\in\{0,1\}^*:M\text{ is an rAPOM}\}$ is decidable by a Turing machine.\footnote{This boils down to programming a syntax-checker, which can be done by a Turing machine.}
    % assuming that the canonical encoding is "reasonable".
    In the remainder of this paper, we assume that we have a fixed canonical representation with these properties. Note that this allows for a canonical enumeration $M_1,\ldots$ of rAPOMs.\footnote{We can let $M_i$ be the $i$-th rAPOM according to the $\prec$ order defined as: $M\prec M'$ if and only if $l(\langle M\rangle) < l(\langle M'\rangle)$ or $l(\langle M\rangle) = l(\langle M'\rangle)$ and $\langle M\rangle$ comes before $\langle M'\rangle$ in lexicographic order.}
\end{remark}

\subsubsection{Universal mixtures over universes}

Now we are ready to define universal mixtures over $\tau$-rPOMs, which give rise to powerful predictors:

\begin{definition}[Universal mixture universe]\label{def:universal-mixture-universes}
    Let $\tau:\{0,1\}^*\to[0,1]$ be a probabilistic oracle, and let $w\in\Delta'\calMrapom$ be a lower semicomputable\footnote{This means that the mapping $\langle M\rangle\mapsto w_M$ is lower semicomputable, where $\langle M\rangle$ is the canonical representation of the rAPOM $M$ as a binary string.} universal prior semiprobability distribution over the set $\calMrapom$ of all rAPOMs. We define the $(w,\tau)$-universal mixture as
    $$\Mmu_\tau^w:=\sum_{M\in\calMrapom}w_M\Mvu^\tau_{M}\,.$$
    The $(w,\tau)$-universal inductor\footnote{We use the term inductor because $\Mmu_\tau^w(\cdot|\cdot)$ can be used to predict the future given the observations so far.} is defined as the conditional semimeasure $\Mmu_\tau^w(\cdot|\cdot)$ that arises from the $(w,\tau)$-universal mixture:\footnote{The denominator $\Mmu^w_\tau(h)$ is always non-zero, as we have that $w_M>0$ for all $M\in \calMrapom$, and for each finite history $h$ there exists a program $M$ that deterministically prints $h$ on its output tape.}
    $$\Mmu_\tau^w(h'|h)=\frac{\Mmu_\tau^w(hh')}{\Mmu_\tau^w(h)}\,,\quad\forall h,h'\in\mathcal{B}^*\,.$$
\end{definition}

An interesting choice of a prior is one that is analogous to Solomonoff's universal prior:

\begin{definition}[Solomonoff universal prior]\label{def:solomonoff-universal-prior}
    Let
    % $M_1,\ldots$ be a canonical enumeration of rAPOMs, and let
    $U$ be a universal monotone Turing machine. We define the Solomonoff universal prior over rAPOMs with respect to $U$ and the enumeration as
    $$w_{M}=2^{-K_U(\langle M\rangle)}\,.$$
    It is worth noting that $\sum_{M\in\calMrapom}2^{-K_U(\langle M\rangle)}\leq 1$, and hence $M\mapsto 2^{-K_U(\langle M\rangle)}$ indeed defines a prior semiprobability distribution.\footnote{This follows from Kraft's inequality and the fact that the set of self-delimiting programs of a universal monotone Turing machine forms a prefix free set.}

    In the remainder of this paper, we will assume that the canonical encoding $\langle M \rangle$ of $M$ as a binary string, as well as the universal monotone Turing machine $U$ are fixed, and hence we will simply refer to the Solomonoff universal prior without further mention of  these fixed choices. We emphasize however that any concept that we will subsequently define using the Solomonoff prior potentially depends on these choices.

    The Solomonoff $\tau$-universal mixture is defined as
    $$\Mmu_\tau:=\sum_{M\in\calMrapom}2^{-K(\langle M\rangle )}\Mvu^\tau_{M}\,,$$
    The Solomonoff $\tau$-universal inductor is the conditional semimeasure $\Mmu_\tau^w(\cdot|\cdot)$ that arises from the Solomonoff $\tau$-universal mixture.
\end{definition}

Since $\Mmu_\tau$ is a universal mixture, it follows from \cref{cor:merging-opinions-semimeasures}, which extends the merging of opinions \cref{theorem:convergence-mixture-universe} towards semimeasures, that $\Mmu_\tau$  asymptotically converges to making perfect predictions in any $\tau$-universe, i.e., for any rAPOM $M$, we have
$$\lim_{t\to\infty}D_\infty(\Mmu_\tau,\Mvu_M^\tau|h_{\leq t})= 0\,,\,\quad \Mvu_M^\tau(h)\text{-almost surely.}$$

\subsubsection{The reflective universal inductor oracle}

Our goal is to model embedded Bayesian agents that use a mixture universe model over the class of all $\tau$-universes, $\{\Mvu_M^\tau\}_{M \in \calMrapom}$. This means the agent's predictive model is the $(w,\tau)$-universal inductor, $\Mmu_\tau^w(h' \mid h)$, for some universal prior $w \in \Delta' \calMrapom$. Our main goal is to ensure that this setup satisfies the grain-of-truth property. Specifically, the class of $\tau$-universes must be rich enough to contain universes that themselves include embedded Bayesian agents using the universal inductor $\Mmu_\tau^w$.

The rPOMs $M$ that define our universes have access to a probabilistic oracle $\tau$, which has not yet been specified. The key idea is to design $\tau$ such that it provides query access to the predictions of the universal inductor $\Mmu_\tau^w$. This would allow us to formally describe universes $\Mvu_M^\tau$ containing embedded Bayesian agents that use $\Mmu_\tau^w$ (by querying $\tau$), thereby satisfying the grain-of-truth condition.

This leads to a crucial self-referential challenge, which motivates the term ``reflective''. The oracle $\tau$ must answer queries about the universal mixture $\Mmu_\tau^w$, which is itself a mixture over rPOMs that have access to $\tau$. In essence, the oracle must have knowledge of a system of which it is an integral part. The existence of such a self-consistent object is not guaranteed. Therefore, the main result of this section is to formally define this special type of oracle and prove that it can, in fact, exist. It is worth noting that reflective oracles \cite{fallenstein2015reflective_oracles,fallenstein2015reflective} also successfully overcome a similar self-referential challenge, which we cover in more detail in \cref{sec:reflective-oracles}.

\begin{definition}[$w$-reflective universal inductor oracle]
    \label{def:rui-oracle}
    Let $\tau:\{0,1\}^*\to[0,1]$ be a probabilistic oracle, and let $w\in\Delta'\calMrapom$ be a lower semicomputable universal semiprobability prior over $\calMrapom$. 
    Let $(b, p,h)\mapsto\langle b,p,h\rangle$ be some canonical encoding of triplets $(b,p,h)\in \calB\times (\mathbb{Q}\cap[0,1]) \times \mathcal{B}^\ast$ as binary strings $\calB^*$.
    
    We say that the probabilistic oracle 
    $\tau$ is a $w$-reflective universal inductor oracle ($w$-RUI oracle) with respect to the encoding $b,p,h\mapsto \langle b,p,h\rangle$ if for every $x\in\mathcal{B}^*$, we have (with $x=\langle b,p,h\rangle$):\footnote{In this definition of the reflective universal inductor, we require that every $x\in\calB^*$ maps to a valid encoding $\langle b,p,h\rangle$, which can be satisfied by taking a complete prefix-free encoding $\langle p\rangle$ of the rational numbers $\bbQ \cap [0,1]$, and then taking $\langle b,p,h\rangle:=b\langle p \rangle h$.}
    \begin{itemize}
            \item[(i)] If $\Mmu_\tau^w(b|h)>p$ then $\tau(x)=1$.
            \item[(ii)] If $\Mmu_\tau^w(b|h)<p$ then $\tau(x)=0$.
    \end{itemize}

    If $\tau$ is a $w$-RUI-oracle with respect to the encoding $b,p,h\mapsto \langle b,p,h\rangle$, we say that $\Mmu_\tau^w$ is a $w$-reflective universal inductor ($w$-RUI) with respect to the same encoding.\footnote{In the remainder of this paper, and unless stated otherwise, we assume that we have a fixed canonical encoding $b,p,h\mapsto \langle b,p,h\rangle$, and hence, for the sake of simplicity we will just write "$w$-RUI-oracle" to mean "$w$-RUI-oracle with respect to the fixed canonical encoding".}

    If $w$ corresponds to the Solomonoff prior, then we refer to $\tau$ and $\Mmu_\tau:=\Mmu_\tau^w$ as the Solomonoff reflective universal inductor oracle (Solomonoff RUI-oracle) and Solomonoff reflective universal inductor (Solomonoff RUI), respectively.
\end{definition}

\begin{remark}
    The adjective "reflective" that is used to describe reflective universal inductors $\Mmu_\tau^w$ comes from the fact that it is a Bayesian mixture over the class of semimeasures induced by probabilistic oracle machines having access to an oracle answering questions about $\Mmu_\tau^w$ itself.
\end{remark}

\begin{remark}
    When a probabilistic oracle $\tau$ is a $w$-RUI-oracle, then one may interpret a query request $\langle b,p,h\rangle$ to the oracle as asking the question: ``Is $\Mmu_\tau^w(b|h)$ greater than $p$?''.
    The oracle returns a deterministic answer if $\Mmu_\tau^w(b|h)>p$ or $\Mmu_\tau^w(b|h)<p$. However, when $\Mmu_\tau^w(b|h)=p$, the oracle is allowed to return a random answer. Nevertheless, the probability describing the random answer must be consistent if we request the same $\langle b,p,h\rangle$ several times.\footnote{Such consistency is actually expected from any general probabilistic oracle as described in \cref{def:probabilistic_oracle} and \cref{def:probabilistic-oracle-machine}. This is not unique to RUI-oracles.}
\end{remark}
\begin{remark}
    As the oracle answer of the query $\langle b,p,h\rangle$ is random only when  $\Mmu_\tau^w(b|h)=p$, it is easy to show that with a binary search procedure (cf. \cref{app:lem_approximate_universal_inductor_proof}), one can obtain an estimate of $\Mmu_\tau^w(b|h)$ up to arbitrary precision despite the randomness of the oracle answer when $\Mmu_\tau^w(b|h)=p$, with a bounded number of oracle requests depending on the desired precision.
\end{remark}

The following lemma extends this to $\Mmu^w_\tau(h'|h)$ for $h,h'\in\mathcal{B}^*$:

\begin{lemma}
\label{lem:approximate_universal_inductor}
For every $h,h'\in\mathcal{B}^*$ and every $\epsilon>0$, we can compute $\Mmu^w_\tau(h'|h)$ up to $\epsilon$-precision using at most $N_{\epsilon,h'}$ queries to the oracle, where $N_{\epsilon,h'}$ computably depends only on $\epsilon$ and $l(h')$.
\end{lemma}
\begin{proof}
    See \cref{app:lem_approximate_universal_inductor_proof}.
\end{proof}

Reflective universal inductors exist:

\begin{theorem}
    \label{thm:rui} For every lower semicomputable universal prior $w\in\Delta'\calMrapom$ over $\calMrapom$, there exists a $w$-RUI-oracle $\tau$ and a corresponding reflective universal inductor $\Mmu_\tau^w$.

    In particular, there exists a Solomonoff RUI-oracle $\tau$ and a corresponding Solomonoff reflective universal inductor $\Mmu_\tau$.
\end{theorem}
\begin{proof}
    See \cref{app:thm_rui_proof}.
\end{proof}

\begin{remark}
For a fixed encoding $\langle b,p,h\rangle$ and a fixed universal prior $w\in\Delta'\calMrapom$ over $\calMrapom$, it is not clear whether a reflective universal inductor corresponding to this $w$ is unique, i.e., there might exist multiple different reflective universal inductors compatible with $w$.
\end{remark}

The set of all rPOMs with access to a $w$-reflective universal inductor oracle $\tau$ induces a hypothesis class over universes:\
$$\calMuni^{w,\tau-\mathrm{RUI}}:=\{\lambda_M^{\tau}: M \in \calMrapom\}.$$

Reflective universal inductors are strong predictors in the sense that they almost surely converge to making accurate predictions in universes $\Mvu \in \calMuni^{w,\tau-\mathrm{RUI}}$ having access to the corresponding RUI-oracle:

\begin{theorem}
    \label{thm:rui-asymptotically-accurate} For every lower semicomputable universal prior $w\in\Delta'\calMrapom$ over the space of rAPOMs $\calMrapom$, if $\tau$ is a $w$-RUI-oracle, then the corresponding RUI $\Mmu_\tau^w$ satisfies:
    $$\forall \lambda \in \calMuni^{w,\tau-\mathrm{RUI}},\,\lim_{t\to\infty}D_\infty(\Mmu^w_\tau,\Mvu|h_{< t})= 0\,,\quad \Mvu(h)\text{-almost surely.}$$
\end{theorem}
\begin{proof}
    See \cref{app:preliminaries}. The proof of the theorem is a direct application of \cref{cor:merging-opinions-semimeasures}.
\end{proof}

\subsection{Universal induction with reflective oracles}\label{sec:reflective-oracles}

In this section, we develop an alternative approach to our RUI framework for solving the general grain-of-truth problem for embedded agency (Problem \ref{prob:grain-of-truth}). We leverage the framework of reflective oracles (ROs) \citep{fallenstein2015reflective_oracles, fallenstein2015reflective, leike2016formal} to construct a universal hypothesis class, $\calM_{\text{uni}}^{\tau\text{-}\mathrm{RO}}$, that contains universes with embedded Bayesian agents that use mixture models over  $\calM_{\text{uni}}^{\tau\text{-}\mathrm{RO}}$ for any lower-semicomputable prior. The concepts and theorems presented here are minimal extensions of the foundational results from \citet{fallenstein2015reflective_oracles, fallenstein2015reflective}, \citet{leike2016formal}, and \citet{wyeth2025limit}—which focused primarily on decoupled agency—to our embedded agency setting.

The RO framework offers a different set of trade-offs compared to the RUI framework. A key advantage of ROs is their generality; they do not impose constraints on the number of oracle calls a machine can make. This allows us to use POMs instead of rPOMs, and formalize embedded Bayes-optimal agents that perform infinite-horizon planning, in contrast to the approximate, $k$-step planning agents necessitated by the RUI's computational constraints (cf. \cref{sec:emb-aixi-agents}). Furthermore, a single reflective oracle can support the construction of universal inductors with different priors, accommodating scenarios where agents with different priors interact. However, this generality comes at the cost of a conceptual departure from classical Solomonoff induction. A reflective oracle arbitrarily redistributes the probability mass of non-halting computations to the outputs 0 or 1, thereby influencing the mixture distribution in an arbitrary way not related to the complexity of the considered POMs. This problem is not present in the RUI framework, where none of the non-halting probability mass contributes to the reflective universal inductor.

The two oracles also differ in the nature of their queries. A RUI oracle directly answers queries $\langle b,p,h \rangle$ about the predictive distribution of its own reflective universal inductor, e.g., "Is the probability $\Mmu^\tau(b|h)$ greater than $p$?". In contrast, a reflective oracle answers queries $\langle M, p, h\rangle$ about the behavior of an arbitrary $\tau$-POM $M$, e.g., "Is the probability that machine $M$ (with oracle access to $\tau$) outputs 1 greater than $p$?". The universal inductor is not the oracle itself but must be constructed as a Bayesian mixture over all $\tau$-POMs (with $\tau$ being the reflective oracle). The central goal of this section is to formally define ROs, construct this universal inductor, and prove that the inductor is itself implementable by a POM with access to the reflective oracle. In \cref{sec:emb-aixi-agents}, in addition of constructing embedded Bayesian agents using RUIs, we also construct embedded Bayesian agents using the RO-based universal inductor, and show that both setups solve the general embedded grain-of-truth problem.

In the RUI framework, we modeled universes as rPOMs without an input tape that generate an entire action-percept sequence. Here, we adopt a different approach common in the RO literature. We consider $\tau$-POMs $M$ that take a history $h \in \mathcal{B}^*$ as input and output a single bit $b$,\footnote{For POMs that output multiple bits on their output tape, we simply ignore all output bits except the first one.} thereby defining a conditional probability $\Mvu_M^\tau(b|h)$. A full semimeasure over histories, $\Mvu_M^\tau(h)$, is then constructed by chaining these conditionals together:
$$\Mvu_M^\tau(h) = \prod_{i=1}^{l(h)}\Mvu_M^\tau(h_i|h_{1:i-1})\,,$$
where we adopt the convention that $h_{1:0}=\varepsilon$.

A reflective oracle is formally defined as follows:
\begin{definition}[Reflective oracle \citep{fallenstein2015reflective_oracles, fallenstein2015reflective}]
\label{def:reflective-oracle}
Let $\tau:\{0,1\}^*\to[0,1]$ be a probabilistic oracle. Let $(M,p,h)\mapsto\langle M,p,h\rangle$ be some canonical encoding of triplets $(M,p,h)\in \calMapom\times(\mathbb{Q}\cap[0,1])\times \mathcal{B}^\ast$ as binary strings $\mathcal{B}^*$, with $\calMapom$ the countable set of all APOMs. Furthermore, for every $M\in\calMapom$, every $h\in\mathcal{B}^\ast$, and every bit $b\in\mathcal{B}$, let $\Mvu_M^\tau(b|h)$ be the probability that POM $M^\tau$ outputs $b$ on the first cell of its output tape when provided with input $h$.

We say that the probabilistic oracle $\tau$ is a reflective oracle (RO) with respect to the encoding\footnote{In the remainder of this paper, we assume a fixed such canonical encoding and refer to such $\tau$ simply as a "reflective oracle".} $(M,p,h)\mapsto \langle M,p,h\rangle$ if for every pair $(M,h)$ there exists some $q \in [0,1]$ such that $\Mvu_M^{\tau}(1|h) \leq q \leq 1- \Mvu_M^{\tau}(0|h)$ and such that for all $p \in \mathbb{Q} \cap [0,1]$, the following implications hold:
\begin{itemize}
    \item[(i)] If $p < q$ then $\tau(\langle M,p,h\rangle)=1$.
    \item[(ii)] If $p > q$ then $\tau(\langle M,p,h\rangle)=0$.
\end{itemize}
\end{definition}

Importantly, for queries with $p=q$, the reflective oracle is allowed to randomize its answer, i.e., have a Bernouilli probability $\tau(\langle M,p,h\rangle)$ different from 0 or 1. This is important to avoid self-reference paradoxes, as illustrated by the example below, adapted from \citet{fallenstein2015reflective_oracles} and \citet{leike2016formal}.
\begin{example}[Reflective oracles and self-reference]
    Consider a machine $M \in \calMapom$ that can access its own description (this is possible using Kleene's second recursion theorem \citep{kleene1952introduction})  and is designed to contradict its oracle. $M$ queries the oracle with $\langle M, \varepsilon, 1/2 \rangle$—asking if its own probability of outputting `1' is greater than $1/2$—and then outputs the opposite bit, $1 - O^\tau(\langle M, \varepsilon, 1/2 \rangle)$.
    Any deterministic response from the oracle creates a paradox. If the oracle answers `1' (implying $\Mvu_M^\tau(1|\varepsilon) \geq 1/2$), the machine's actual output becomes `0', making $\Mvu_M^\tau(1|\varepsilon) = 0$. This contradicts the oracle's premise. Conversely, if the oracle answers `0' (implying $\Mvu_M^\tau(1|\varepsilon) \le 1/2$), the machine's output becomes `1', making $\Mvu_M^\tau(1|\varepsilon) = 1$, which is also a contradiction.
    The only consistent resolution is for the machine's output probability to be exactly the queried threshold, $\Mvu_M^\tau(1|\varepsilon) = 1/2$. This is only possible if the oracle's response is a random coin flip. Therefore, for this ``liar'' machine, any valid reflective oracle must satisfy $\tau(\langle M, \varepsilon, 1/2 \rangle) = 1/2$.
\end{example}

As shown by \citet{fallenstein2015reflective, fallenstein2015reflective_oracles}, reflective oracles are guaranteed to exist.
\begin{theorem}[Existence of a reflective oracle \citep{fallenstein2015reflective, fallenstein2015reflective_oracles}]\label{thrm:ro-existence}
A reflective oracle exists.
\end{theorem}

A key property of a reflective oracle is that it allows for the "completion" of a semimeasure into a full measure. 
For each $\tau$-POM $M$ and input $h$, the reflective oracle redistributes any non-halting probability mass by selecting the value $q$ in \cref{def:reflective-oracle} satisfying $\Mvu_M^{\tau}(1|h) \leq q \leq 1- \Mvu_M^{\tau}(0|h)$. We can then take $\bar{\Mvu}_M^\tau(1|h):=q$ and $\bar{\Mvu}_M^\tau(0|h):=1-q$ as the completion of $\Mvu$ into a full measure. This value $q$ can be found by performing a binary search over $p$ and repeatedly querying the reflective oracle with queries $\langle M,h,p\rangle$. While this procedure can be stochastic (if a query is made with $p=q$, the reflective oracle is allowed to provide a randomized answer), it converges to the correct value in the limit. This leads to the following definition:
\begin{definition}[$\tau$-completion]
\label{def:tau-completion}
The $\tau$-completion of the output distribution $\Mvu_M^\tau(b|h)$ of a $\tau$-POM $M$ is given by $\bar{\Mvu}_M^\tau(1|h):=q$ and $\bar{\Mvu}_M^\tau(0|h):=1-q$, where $q\in\mathbb{R}$ is the value from \cref{def:reflective-oracle}. We overload the notation $\bar{\Mvu}_M^\tau(h) := \prod_{i=1}^{l(h)}\bar{\Mvu}_M^{\tau}(h_i|h_{<i})$, which induces a measure $\bar{\lambda}_M^\tau$.
\end{definition}

We can now define our hypothesis class over universes by chaining these completed conditional measures. The hypothesis class is then:
$$\calM_{\text{uni}}^{\tau\text{-}\mathrm{RO}} := \{\bar{\Mvu}_M^\tau : M \in \calMapom\}\,.$$
It is worth noting that if $\lambda_M^\tau$ is already a measure, then $\lambda_M^\tau=\bar{\lambda}_M^\tau\in\calM_{\text{uni}}^{\tau\text{-}\mathrm{RO}}$. Therefore, the class $\calM_{\text{uni}}^{\tau\text{-}\mathrm{RO}}$ contains all computable universes.

We would like to construct a universal inductor over $\calM_{\text{uni}}^{\tau\text{-}\mathrm{RO}}$ which is itself an element of this class. More precisely, for some choice of an arbitrary lower semicomputable semiprobability prior $w\in\Delta'\calMapom$, e.g., the Solomonoff prior $w(M)=2^{-K_U(\langle M\rangle)}$, we would like to find an APOM $\MwRO$ such that\footnote{Note that if $w$ in not a probability distribution, then $\sum_{M\in \calMapom}w(M)\bar{\Mvu}_M^\tau$ is not a measure. On the other hand, $\bar{\Mvu}^\tau_{\MwRO}$ is a measure, and hence, we can only aim for an inequality $\bar{\Mvu}^\tau_{\MwRO}\geq \sum_{M\in\calMapom} w(M)\bar{\Mvu}^\tau_{M}$.}
\begin{equation}
    \label{eq:universal-inductor-ro-machine}
    \bar{\Mvu}^\tau_{\MwRO}\geq \sum_{M\in\calMapom} w(M)\bar{\Mvu}^\tau_{M}=\sum_{\Mvu\in \calM_{\text{uni}}^{\tau\text{-}\mathrm{RO}}} w(\Mvu)\Mvu\,,
\end{equation}
where
$$w(\Mvu) := \sum_{M:\bar{\Mvu}_M^\tau = \Mvu} w(M)\,.$$
If such a machine exists, then the universal inductor defined as $\bar{\Mmu}_{\tau\text{-}\mathrm{RO}}^w:=\bar{\Mvu}^\tau_{\MwRO}$ would satisfy the following desired properties:
\begin{enumerate}
    \item $\bar{\Mmu}_{\tau\text{-}\mathrm{RO}}^w\in \calM_{\text{uni}}^{\tau\text{-}\mathrm{RO}}$, and
    \item $\bar{\Mmu}_{\tau\text{-}\mathrm{RO}}^w$ dominates every universe in $\calM_{\text{uni}}^{\tau\text{-}\mathrm{RO}}$, and hence from \cref{theorem:convergence-mixture-universe} we get that
    \begin{equation}
        \label{eq:ro-mixture-strongly-merges}
        \forall \Mvu\in \calMuni^{\tau\text{-}\mathrm{RO}}, \lim_{t \to \infty} D_\infty(\bar{\Mmu}_{\tau\text{-}\mathrm{RO}}^w, \lambda \mid h_{<t}) = 0, \quad \Mvu(h)\text{-almost surely}\,. 
    \end{equation}
    Furthermore, \cref{theorem:generalized-solomonoff-bound} implies that for every $\Mvu\in\calM_{\text{uni}}^{\tau\text{-}\mathrm{RO}}$, we can bound the accumulated prediction loss over trajectories of length $n$ in terms of the prior $w(\Mvu)$ as follows\footnote{The inequality in \eqref{eq:ro-prediction-loss} implies that the universal inductor's $\bar{\Mmu}_{\tau\text{-}\mathrm{RO}}^w$ inductive bias is consistent with its prior $w$. This consistency is the key motivation for constructing a machine $\MwRO$ (satisfying \eqref{eq:universal-inductor-ro-machine}) for an arbitrary lower semicomputable (LSC) prior $w$. The primary objective is to enable an inductive bias consistent with the Solomonoff prior, which is itself LSC but not computable. If our goal were merely to satisfy the merging property in \eqref{eq:ro-mixture-strongly-merges}, a simpler computable prior (e.g., $w(M)=3^{-\langle M\rangle}$) would suffice, making the construction of $\MwRO$ much easier.}
    \begin{equation}
        \label{eq:ro-prediction-loss}
        L_n(\bar{\Mmu}_{\tau\text{-}\mathrm{RO}}^w, \Mvu) \leq -\log w(\Mvu) < \infty\,.
    \end{equation}
    For the case of the Solomonoff prior $w(\bar{\Mvu}) := \sum_{M:\bar{\Mvu}_M^\tau = \bar{\Mvu}} 2^{-K_U(\langle M\rangle)}$, we get
    $$L_n(\bar{\Mmu}_{\tau\text{-}\mathrm{RO}}^w, \Mvu) \leq -\log \sum_{M:\bar{\Mvu}_M^\tau = \Mvu} 2^{-K_U(\langle M\rangle)}\leq \min_{M:\bar{\Mvu}_M^\tau = \Mvu} K_U(\langle M\rangle)\,.$$
\end{enumerate}

The remainder of this section is dedicated to constructing a machine $\MwRO$ satisfying \eqref{eq:universal-inductor-ro-machine}. Similarly to \citet{wyeth2025limit}, we construct a machine that satisfies the following recursive equations

\begin{equation}
    \label{eq:universal-inductor-ro-machine-recursive}
\begin{aligned}
    \Mvu^\tau_{\MwRO}(b|h_{<t}) &\geq \sum_{M\in\calMapom} w(M|h_{<t}) \bar{\Mvu}_M^\tau(b|h_{<t})\,, \\
    w(M|h_{1: t}) &= w(M|h_{<t}) \frac{\bar{\Mvu}_M^\tau(h_t|h_{<t})}{\bar{\Mvu}^\tau_{\MwRO}(h_t|h_{<t})}\,,\\
    w(M|\varepsilon)&=w(M)\,.
\end{aligned}
\end{equation}

\cref{app:proof-universal-inductor-ro-machine-recursive} shows that indeed, if a machine $\MwRO$ satisfies \eqref{eq:universal-inductor-ro-machine-recursive}, then it satisfies \eqref{eq:universal-inductor-ro-machine} as well, and hence \eqref{eq:ro-mixture-strongly-merges} and \eqref{eq:ro-prediction-loss} are also satisfied.

To prove that such an APOM $\MwRO$ can indeed be implemented, we first need to introduce some notions of computability relative to a probabilistic oracle. Note that in general, a $\tau$-POM has a probabilistic output due to the stochasticity of $O^\tau$ and potentially the use of random coinflips. Hence, this needs to be taken into account in the computability notions introduced below.
\begin{definition}[$\tau$-estimable \citep{wyeth2025limit}]\label{def:oracle-estimable}
    A deterministic function $f: \calB^* \to \bbR$ is \emph{$\tau$-estimable} if and only if there exists a $\tau$-POM $M$ that upon input $\langle k,h\rangle$ with $(k,h) \in \bbN \times \calB^*$ (stochastically) outputs a binary string $\langle y \rangle$ encoding a rational number $y \in \bbQ$ and then halts, such that for all $\langle y \rangle$ with $\Mvu_M^{\tau}(\langle y \rangle \mid \langle k,h\rangle) >0$ and $\Mvu_M^{\tau}(\langle y \rangle \mid \langle k,h\rangle) > \sum_{b\in \calB} \Mvu_M^{\tau}(\langle y \rangle b \mid \langle k,h\rangle)$\footnote{When this strict inequality is satisfied, there is a non-zero probability of halting after outputting $\langle y \rangle$.} we have that $|y - f(h)| \leq \frac{1}{k}$.
\end{definition}

\begin{definition}[$\tau$-lower-semicomputable]\label{def:oracle-lsc}
    A deterministic function $f: \calB^* \to \bbR$ is said to be \emph{$\tau$-lower-semicomputable} ($\tau$-LSC) if and only if there exists a $\tau$-POM $M$ that upon input $h\in \calB^*$ with $\Mvu_M^\tau$-probability 1 outputs an infinite bit sequence $(\langle q_k\rangle)_{k=1}^\infty$ of prefix-free encoded rational numbers such that $q_k \to f(h)$ as $k \to \infty$ $\Mvu_M^\tau$-almost-surely, and $q_{k+1} \geq q_k$ for all $k$ with $\Mvu_M^\tau$-probability 1.
\end{definition}
This specialized definition of $\tau$-LSC is needed to handle the stochasticity of the oracle. The standard definition of lower semi-computability (cf. \cref{def:lsc}) involves independent calls to a function $\phi(x,k)$ for each step $k$, making it difficult to enforce the monotonicity condition $q_{k+1} \geq q_k$ when the outputs are stochastic. By having a single $\tau$-POM stream the entire sequence, it can use its internal state to ensure monotonicity despite the oracle's randomness. A function $f$ is $\tau$-upper-semicomputable ($\tau$-USC) if $-f$ is $\tau$-LSC.

The binary search procedure used to find the $\tau$-completion of a machine's output provides a sequence of improving lower and upper bounds. This gives us a crucial property:
\begin{theorem}[Estimable completions, Theorem 8 of \citet{wyeth2025limit}]\label{thrm:estimable-completions}
Every $\tau$-completion $\bar{\Mvu}_M^\tau$ is $\tau$-estimable, $\tau$-LSC, and $\tau$-USC.
\end{theorem}

Finally, we define what it means for a conditional semimeasure to be implemented by a $\tau$-POM.
\begin{definition}[$\tau$-sampleable \citep{wyeth2025limit}]\label{def:tau-sampleable}
A conditional semimeasure $\sigma: \mathcal{B}^* \to \Delta'\mathcal{B}$ is \emph{$\tau$-sampleable} if there exists a $\tau$-POM $M$ such that $\Mvu_M^\tau(b|h) = \sigma(h)(b)$ for all $b \in \calB, h \in \calB^*$.
\end{definition}

A key link between these concepts is that $\tau$-lower-semicomputablity implies $\tau$-sampleability.
\begin{lemma}[$\tau$-LSC implies $\tau$-sampleable]\label{lemma:tau-lsc-tau-sampled}
A $\tau$-lower-semicomputable conditional semimeasure is also $\tau$-sampleable.
\end{lemma}
\begin{proof}
The proof is a minor variation on \citet[Lemma 4.3.3]{li2008introduction} and \citet[Theorem 6]{wyeth2025limit}. See \cref{app:proof-tau-lsc-tau-sampled}.
\end{proof}

With this machinery in place, we can now state and prove the main result of this section.
\begin{theorem}\label{thrm:ro-ui-sampleable}
% The universal inductor $\bar{\Mmu}_{\tau\text{-}\mathrm{RO}}^w$ over $\calMuni^{\tau\text{-}\mathrm{RO}}$ is $\tau$-sampleable.
There exists an APOM $\MwRO$ satisfying \eqref{eq:universal-inductor-ro-machine-recursive}.
\end{theorem}
\begin{proof}

The proof builds upon the work of \citet{leike2016formal} and \citet{wyeth2025limit}. See \cref{app:proof-thrm-ro-ui-sampleable}.
\end{proof}

This result confirms that the universal inductor over the class $\calM_{\text{uni}}^{\tau\text{-}\mathrm{RO}}$ is itself an element of that class. In the next section, we use this result to show that embedded Bayesian agents using $\Mmu_{\tau\text{-}\mathrm{RO}}^w$ solve the general embedded grain-of-truth problem (cf. Problem \ref{prob:grain-of-truth}).

\subsection{Embedded AIXI agents}\label{sec:emb-aixi-agents}
In \cref{sec:rui}, we introduced RUI-oracles leading to a wide hypothesis class $\calMuni^{w,\tau-\mathrm{RUI}}$ over universes with access to RUI oracle $\tau$ using prior $w$, which includes all lower semicomputable semimeasures. Similarly, in \cref{sec:reflective-oracles} we leveraged the reflective oracle framework \citep{fallenstein2015reflective_oracles} to construct a hypothesis class $\calMuni^{\tau-\mathrm{RO}}$ over probabilistic oracle machines with access to the reflective oracle $\tau$, also including all lower semicomputable semimeasures. In this section, we address two remaining questions: (i) how can we use universes that output bits to model (multi-agent) environments and policies, where different agents possibly have different percept and action spaces? (ii) Are embedded Bayesian agents using hypothesis classes $\calMuni^{w,\tau-\mathrm{RUI}}$ or $\calMuni^{\tau-\mathrm{RO}}$ included within the hypothesis classes themselves, i.e., can we solve the general grain-of-truth problem for embedded agency stated in Problem \ref{prob:grain-of-truth}?

\subsubsection{Modeling agent-environment interactions} 
The universes $\Mvu$ of \cref{sec:embedded-bayesian-agents,sec:subj-emb-eq} described probabilities over action-percept sequences with action space $\calA$ and percept space $\calE$. 
In contrast, the universes $\Mvu_M^\tau$ of \cref{sec:rui} are semimeasures over binary sequences. We can connect the two approaches by introducing a complete prefix-free encoding $a \mapsto \langle a \rangle$ of $a \in \calA$ as binary strings, and similarly for $e\in \calE$. Using such an encoding, each binary string can be translated to a sequence of actions and percepts\footnote{By starting reading the bitstring from left to right, one can alternate between decoding an action and decoding a percept. Possibly the bitstring does not contain a complete codeword for the last symbol (action or percept); in such cases, we simply ignore the bits at the end of the bitstring that do not correspond to a complete codeword.} and vice versa. Using $x:=\langle a_t \rangle$ and $y:=\langle e_t\rangle$, the conditional semimeasures over bitstrings can be translated to conditional semimeasures over action-percept sequences as follows:
\begin{equation}\label{eq:bitprob2action-percept-prob}
    \begin{split}
        \Mvu(a_t \mid \HistM) &:= \Mvu(x \mid \langle\HistM\rangle)= \prod_{i=1}^{l(x)} \Mvu_M^{\tau}(x_i \mid \langle \HistM\rangle x_{1:i-1})\,, \\
        \Mvu(e_t \mid \HistM,a_t) &:= \Mvu(y \mid \langle\HistM a_t\rangle)= \prod_{i=1}^{l(y)} \Mvu_M^{\tau}(y_i \mid \langle \HistM a_t\rangle y_{1:i-1})\,, \\
    \end{split}
\end{equation}
with $\langle \HistM\rangle$ the binary string encoding of the history $\HistM$ of actions and percepts, using the complete prefix-free encodings mentioned above. As $\Mvu_M^{\tau}$ are semimeasures, it can be that $\sum_a\Mvu(a \mid \HistM) < 1$, when the universe has a non-zero probability of halting or getting stuck in an infinite loop before outputting an action encoding\footnote{As we are using complete prefix-free encodings,  machines are guaranteed to output a valid action encoding, as long as they do not halt or get stuck in an infinite loop prematurely.} (and similarly for percepts). We can assign the missing probability mass to a special token, representing that the universe, including the agent, ceases to exist (cf. Appendix~\ref{app:preliminaries}). Note that the probabilities $\Mvu(a_t \mid \HistM)$ and $\Mvu(e_t \mid \HistM,a_t)$, and as a result the predictions and value estimations of the agent, depend on the chosen complete prefix-free encodings for $\calA$ and $\calE$.\footnote{In the remainder of this work, we assume a fixed choice of prefix-free encodings, and omit mentioning this dependence on the encodings.}

We can use a similar strategy for modeling multi-agent environments using binary sequences. Leveraging a complete prefix-free encoding of the joint action space $\bar{\calA}$ and joint percept space $\bar{\calE}$, we can interpret binary strings as multi-agent environment interactions (cf. \cref{sec:background-grl}). From the point of view of a specific agent, we can interpret binary sequences as personal histories $\HistI\in \big(\TurnSet^i)^*$ consisting of individual actions $a^i\in\calA^i$ and percepts $e^i\in \calE^i$. Both the joint distribution $\MultiAgentMve^{\MultiAgentMvp}$, as well as the personal distributions $\big(\Mve^i\big)^{\Mvp^i}$ can then be represented as the semimeasures induced by universes, using the strategy of \eqref{eq:bitprob2action-percept-prob}. 

Finally, let us briefly discuss how agent and environment programs can be merged together to form a universe. Assume that we can describe the policies $\pi^i(a^i \mid \MAHistI)$ and multi-agent environment $\MultiAgentMve(\bar{e} \mid \MAHist\bar{a})$ as the semimeasures induced by $\tau$-POMs $(M^\tau_{\pi^i})_{i\in[N]}$ and $M^\tau_{\MultiAgentMve}$ over their outputs $\langle a^i\rangle$ and $\langle \bar{e} \rangle$ respectively, with prefix-free encodings of $\MAHistI$ and $\MAHist\bar{a}$ on their respective input tapes. Then, we can combine the machines $(M^\tau_{\pi^i})_{i\in[N]}$ and $M^\tau_{\MultiAgentMve}$ into a single $\tau$-POM $M^\tau$ that outputs binary encoded sequences of $\bar{a}$ and $\bar{e}$, by alternately exchanging inputs and outputs between $(M^\tau_{\pi^i})_{i\in[N]}$ and $M^\tau_{\MultiAgentMve}$ appropriately. Furthermore, if $M^\tau$ satisfies the constraints on oracle calls mentioned in \cref{def:rapom}, it is a universe within $\calMuni^{w,\tau-\mathrm{RUI}}$. Finally, we can convert $M^\tau$ to personal universes $M^\tau_i$ for each individual agent by mapping $\bar{a}$ to $a^i$ and $\bar{e}$ to $e^i$. See \cref{app:multi-agent-env} for a more detailed discussion on how to represent multi-agent environments using universes.

\subsubsection{Embedded AIXI agents}
Using the hypothesis classes $\calMuni^{w,\tau-\mathrm{RUI}}$ or $\calMuni^{\tau-\mathrm{RO}}$ with a corresponding lower semicomputable universal prior $w$, we can use the following mixture universe models
\begin{align}
    \Mmu^{\mathrm{RUI}}:= \Mmu_{\tau-\mathrm{RUI}}^w, \quad \Mmu^{\mathrm{RO}}:=\bar{\Mmu}_{\tau-\mathrm{RO}}^w
\end{align}
with $\Mmu_{\tau-\mathrm{RUI}}^w$ the universal mixture over $\calMuni^{w,\tau-\mathrm{RUI}}$, and $\bar{\Mmu}_{\tau-\mathrm{RO}}^w$ the $\tau$-completed universal mixture over $\calMuni^{\tau-\mathrm{RO}}$ (cf. \cref{def:universal-mixture-universes}). Now we can readily use $\Mmu^{\mathrm{RUI}}$ and $\Mmu^{\mathrm{RO}}$ to design various embedded Bayesian agents following Sections \ref{sec:embedded-br}--\ref{sec:k-step-planning}, which we call \textit{Embedded AIXI agents}, or E-AIXI in short. Embedded AIXI agents can use either $\Mmu^{\mathrm{RUI}}$ or $\Mmu^{\mathrm{RO}}$, with either infinite-horizon optimal planning (cf. \cref{sec:embedded-br}), or $k$-step planning (cf. \cref{sec:k-step-planning}); we use $k$-E-AIXI to indicate the number of planning steps, and omit $k$ if infinite-horizon planning is used. When we need to distinguish between E-AIXI agents that use $\Mmu^{\mathrm{RUI}}$ or $\Mmu^{\mathrm{RO}}$, we use superscript E-AIXI$^{\mathrm{RUI}}$ and E-AIXI$^{\mathrm{RO}}$ respectively.

As a first set of main results, building upon the work of \citet{fallenstein2015reflective} and \citet{leike2016formal}, we show that E-AIXI agents using reflective oracles satisfy the grain-of-truth problem, i.e., universes containing such E-AIXI agents are part of $\calMuni^{\tau-\mathrm{RO}}$. As a direct consequence of this, multi-agent systems of E-AIXI$^{\mathrm{RO}}$ agents converge to a subjective correlated embedded equilibrium.

\begin{theorem}
\label{thm:eaixi-ro}
    For every complete prefix-free encoding of $\mathcal{A}$ and $\mathcal{E}$, there exists an E-AIXI$^{\mathrm{RO}}$ policy and $k$-E-AIXI$^{\mathrm{RO}}$ policies for every $k\in \bbN$, with respect to the reflective oracle $\tau$, which is implementable by a POM $M^\tau$ with access to the reflective oracle $\tau$, i.e., the POM $M^\tau$ writes $a_t$ on the output tape when the input tape contains a prefix-free encoding of $\HistM$.

    Furthermore, when one or more such ($k$-)E-AIXI$^{\mathrm{RO}}$ agents are combined with a multi-agent environment~$\MultiAgentMge$ that is implementable on a POM with access to reflective oracle $\tau$, then the resulting universe is part of $\calMuni^{\tau-\mathrm{RO}}$.
\end{theorem}
\begin{proof}
    Minor variation upon \citet[Theorem 22]{leike2016formal} and \citet[Theorem 18]{wyeth2025limit}, see \cref{app:proof-eaixi-ro}.
\end{proof}

\begin{corollary}
    \label{cor:eaixi-ro-convergence}
    Let $N$ E-AIXI$^{\mathrm{RO}}$ agents interact with a multi-agent environment $\MultiAgentMge$, which all are implementable on a POM with access to reflective oracle $\tau$. Then it holds that for each $\epsilon > 0$, there exists a finite time $T(\epsilon)$ such that for all $t \geq T(\epsilon)$, with probability greater than $1-\epsilon$, the personal Bayesian mixture universes $\Mmu^{\mathrm{RO}}_i$ and E-AIXI$^{\mathrm{RO}}$ policies $(\pi^i)_i$ are an $\epsilon$-subjective correlated embedded equilibrium in the correlated tail game starting at time $t$ with correlation device $((\MATurnSet)^{t}, \MultiAgentMge^{\MultiAgentMvp})$.
\end{corollary}
\begin{proof}
    From \cref{thm:eaixi-ro} we know that the interaction of E-AIXI$^{\mathrm{RO}}$ agents with a multi-agent environment~$\MultiAgentMge$ that is implementable on a POM with access to $\tau$, results in a universe which is also implementable by a POM with access to $\tau$, and hence it is part of $\calMuni^{\tau-\mathrm{RO}}$. Therefore, $\Mmu^{\mathrm{RO}}_i$ satisfies the grain-of-truth property. It also trivially satisfies the grain-of-uncertainty property. We can now apply \cref{thrm:convergence-to-sde-correlated} to get the result.
\end{proof}

Unfortunately, analogous results do not hold for ($k$-)E-AIXI$^{\mathrm{RUI}}$ agents, due to the restrictions of $\calMuni^{w,\tau\text{-}\mathrm{RUI}}$ to universes that obey the constraints on oracle calls mentioned in \cref{def:rapom}. For example, computing the exact $Q_\Mmu$ values for the $k$-step planners (cf. \cref{def:k-step-eba}) can require infinitely many oracle calls, which is not allowed in universes in $\calMuni^{w,\tau\text{-}\mathrm{RUI}}$.

To address this challenge, we use the $\epsilon$ approximations of the $k$-step planner embedded Bayesian agents, introduced in \cref{def:approximate-eba}. By letting $\epsilon_t \to 0$ and $k_t \to \infty $ as $t\to\infty$, the $(k_t,\epsilon_t)$-step planner embedded Bayesian agent makes an increasingly better approximation of the embedded Bayes-optimal agent as time progresses. The following Theorem shows that $(\epsilon_t,k_t)$-E-AIXI$^{\mathrm{RUI}}$ agents, i.e., $(k_t,\epsilon_t)$-step planner embedded Bayesian agents using hypothesis class $\calMuni^{w,\tau\text{-}\mathrm{RUI}}$ and universal prior $w$, satisfy the grain-of-truth property.

\begin{theorem}
\label{thm:k_mupi_implementation}
    For every complete prefix-free encoding of the finite sets $\mathcal{A}$ and $\mathcal{E}$, and all computable sequences $(\epsilon_t)_t$ and $(k_t)_t$ with $\epsilon_t>0 ~~\forall t$ , there exists a $(k_t,\epsilon_t)_t$-E-AIXI$^{\mathrm{RUI}}$ policy with respect to the $w$-RUI $\Mmu^{\mathrm{RUI}}$, which is implementable by a rPOM $M^\tau$ with access to the $w$-RUI-oracle $\tau$.
    
    Furthermore, when writing $\langle a_t\rangle$ on the output tape, the machine makes at most $\tilde{N}_t$ queries to the oracle, where $\tilde{N}_t$ depends only on $k_t$, $\epsilon_t$ and on the sizes of $\mathcal{A}$ and $\mathcal{E}$.

    Finally, when one or more such $(k_t,\epsilon_t)_t$-E-AIXI$^{\mathrm{RUI}}$ policies are combined with a multi-agent environment~$\MultiAgentMge$ that is implementable on a rPOM with access to RUI oracle $\tau$, then the resulting universe is part of $\calMuni^{w,\tau-\mathrm{RUI}}$.
\end{theorem}
\begin{proof}
    See \cref{app:thm_mupi_implementation_proof}
\end{proof}

In conclusion, we introduced two novel hypothesis classes $\calMuni^{w,\tau\text{-}\mathrm{RUI}}$ and $\calMuni^{\tau\text{-}\mathrm{RO}}$ over universes, and various embedded Bayesian agents leveraging those novel hypothesis classes, which solve the general grain-of-truth problem for embedded agency (cf. Problem \ref{prob:got-decoupled-informal}). In \cref{app:desiderata-embedded-uai-theory}, we explain in more detail how our formalism solves this grain-of-truth problem.

The remaining question is whether we can show that $(k_t,\epsilon_t)_t$-E-AIXI$^{\mathrm{RUI}}$ agents converge to $(\epsilon,\delta)$-subjective correlated equilibria when interacting with each other. \cref{thrm:e-aixi-rui-epsdelta-sde} shows that when we take $(k_t, \epsilon_t)_t$ such that $\lim_{t\to\infty}k_t = \infty$ and $\lim_{t \to \infty}\epsilon_t = 0$, we can prove the affirmative, even without requiring the `sensible off-policy' condition of \cref{thrm:convergence-to-epsdelta-sde-correlated}.

\begin{theorem}\label{thrm:e-aixi-rui-epsdelta-sde}
    Let $N$ $(k_t,\epsilon_t)_t$-E-AIXI$^{\mathrm{RUI}}$ agents, with $\lim_{t\to\infty}k_t = \infty$ and $\lim_{t \to \infty}\epsilon_t = 0$, interact with a multi-agent environment $\MultiAgentMge$, which all are implementable on a rPOM with access to RUI oracle $\tau$. Then it holds that for each $\epsilon > 0$ and $\delta > 0$, there exists a finite time $T(\epsilon, \delta)$ such that for all $t \geq T(\epsilon,\delta)$, with probability greater than $1-\epsilon$, the personal Bayesian mixture universes $\Mmu^{\mathrm{RUI}}_i$ and $(k_t,\epsilon_t)_t$-E-AIXI$^{\mathrm{RUI}}$ policies are an $(\epsilon,\delta)$-subjective correlated equilibrium in the correlated tail game starting at time $t$ with correlation device $((\MATurnSet)^{t}, \MultiAgentMge^{\MultiAgentMvp})$.
\end{theorem}
\begin{proof}
    This is a direct corollary of \cref{prop:convergence-of-approximate-eba-to-sde-correlated}.
\end{proof}

Unfortunately, the above positive result does not hold for $k$-E-AIXI agents using a fixed $k$, as the `sensibly off-policy' condition required for \cref{thrm:convergence-to-epsdelta-sde-correlated} is not satisfied. Building upon the `dogmatic beliefs' framework of \citet{leike2015bad}, we construct a counter example in \cref{thrm:counter-example} showing that $k$-E-AIXI agents do not always converge to an embedded best response w.r.t. $\Mmu^i$. This result holds for $k$-E-AIXI agents using any probabilistic oracle $\tau$ --e.g., the RUI oracle, reflective oracle or a `dummy oracle' always outputting 0 such that the POMs using this dummy oracle correspond to standard monotone probabilistic Turing machines. This results also holds for decoupled $k$-AIXI agents such as Self-AIXI which has $k=1$, thereby disproving the conjecture of \citet{catt2023self} that the `sensibly off-policy condition' is satisfied for all `reasonable' hypothesis classes except for some possible exotic counterexample classes.

\begin{theorem}\label{thrm:counter-example}
Consider a $k$-E-AIXI$^{\tau}$ agent, i.e., an embedded Bayesian agent employing $k$-step planning for a fixed integer $k \ge 1$, with a mixture universe model $\Mmu$ constructed from a Solomonoff prior over the hypothesis class $\calM^\tau$ of POMs with access to an arbitrary probabilistic oracle $\tau$. Then, there exists a computable ground-truth environment $\Mge$ and a reference universal monotone Turing machine $U$ for defining the Kolmogorov complexity $K_U$ in the Solomonoff prior, such that the agent interacting with $\Mge$ does not converge to the embedded best response w.r.t. $\Mmu$, despite the fact that this setup satisfies the grain-of-truth property. This result holds also for decoupled AIXI agents performing $k$-step planning with a fixed $k$, such as Self-AIXI\footnote{The failure of Self-AIXI to converge to AIXI was concurrently indicated (but not proved rigorously) by Cole Wyeth in \url{https://www.alignmentforum.org/posts/B6gumHyuxzR5yn5tH/unbounded-embedded-agency-aedt-w-r-t-rosi}.}.
\end{theorem} 
\begin{proof}
    See \cref{app:proof-counterexample}
\end{proof}

\subsection{Functional similarities through the lens of algorithmic information theory}\label{sec:mupi-structural-similarities}
We revisit the functional similarities introduced in \cref{sec:emb-bay-agents-structural-similarities} through the lens of algorithmic information theory, formalizing the intuition of functional similarities as the joint `compressibility' of agent and environment. The degree of functional similarities in \cref{sec:emb-bay-agents-structural-similarities} is determined by the  belief prior $w$, a choice we have not yet specified. Here, we fix the belief prior as the Solomonoff universal prior (cf. \cref{def:solomonoff-universal-prior}) arising from the algorithmic complexity of the considered universes. Importantly, this algorithmic complexity depends on the choice of reference universal machine $U$ to compute the Kolmogorov complexity, and the encoding $\langle M \rangle$ of APOMs into bitstrings. Hence, although we removed the arbitrary choice of prior beliefs, we instead introduced an arbitrary choice of reference universal machine $U$ and encoding $\langle M \rangle$. This motivates the main theorem of this section, where we prove that the Solomonoff universal prior is always coupled, and that it contains universes with arbitrarily high degrees of functional similarities, regardless of the choice of reference universal machine $U$ and encoding $\langle M \rangle$. 

\subsubsection{Solomonoff priors are always coupled}
\textbf{Setup.} Let $\calMapom$ be the set of APOMs and let $\tau$ be an arbitrary probabilistic oracle, e.g., a reflective oracle or a RUI oracle. Note that the presented theory also applies to standard probabilistic monotone Turing machines without access to an oracle. Let $\langle M\rangle$ be the binary encoding of $M$ according to some canonical encoding.
Using a universal monotone Turing machine $U$, we obtain the Solomonoff universal prior w.r.t. $U$ and $\langle M \rangle$ as 
$$w^U(M)=2^{-K_U(\langle M\rangle)},$$ 
with $K_U$ the Kolmogorov complexity w.r.t. $U$.
Note that different machines can implement the same function. We will not be interested in functional similarities between specific machines $M$, but rather between the universes $\Mvu_M^\tau$ they implement.
For this, let $\langle a\rangle$ and $\langle e\rangle$ be complete prefix-free encodings of $\calA$ and $\calE$, respectively, into bitstrings. Now we can define $\calMuni^\tau$ to be the set of semimeasures on action-percept histories in $\TurnSet^* \cup (\TurnSet^* \times \calA)$ that are induced by APOMs in $\calMapom$ with access to $\tau$, while using $\langle a\rangle$ and $\langle e\rangle$ to interpret the generated bitstrings. More precisely,
$$\calMuni^\tau=\{\Mvu_M^\tau:M\in\calMapom\}\,.$$
%where we omitted adding $\langle a\rangle$ and $\langle e\rangle$ to the notation as we consider them fixed.
Note that results in this section do not depend on the specific choice of oracle $\tau$.
Now the Solomonoff prior $w^U(M)$ induces a prior $w^{U,\tau}$ over $\calMuni^\tau$ as follows:
\begin{align}\label{eq:solomonoff-prior-universes}
    w^{U,\tau}(\Mvu):=\sum_{\substack{M\in\calMapom:\\\Mvu_M^\tau=\Mvu}}w^U_M=\sum_{\substack{M\in\calMapom:\\\Mvu_M^\tau=\Mvu}}2^{-K_U(\langle M\rangle)}\,.
\end{align}
We are interested in defining functional similarities between policy $\pi$ and environment $\Mve$ inside a universe $\Mvu$. One complication is that even if a universe $\Mvu$ is implementable by a POM $M^\tau$, its conditionals $\pi$ and $\Mve$ are not necessarily implementable by a $\tau$-POM.\footnote{To see why this is not nessarily the case, note that the conditional of a lower-semicomputable universe $\Mvu$ is not necessarily lower-semicomputable itself~\citep{leike2016nonparametric}.} Hence, we introduce the following restricted hypothesis class to contain only universes where the conditionals $\pi$ and $\Mve$ are also implementable on $\tau$-POMs, such that we can use the same machinery to handle $\Mvu$, $\pi$ and $\Mve$:
$$\calM^{\tau}_{\mathrm{pol}\textrm{-}\mathrm{env}}:=\left\{\Mve^\pi:\Mve\in \calM_{\mathrm{env}}^{\tau}, \pi \in \calM_{\mathrm{pol}}^{\tau}\right\}\subset \calM^{\tau}_{\mathrm{uni}},$$
with $\calM_{\mathrm{pol}}^{\tau}$ the set of policies $\TurnSet^*\to\Delta'\mathcal{A}$ which are implementable by a $\tau$-POM, and $\calM_{\mathrm{env}}^{\tau}$ the set of environments $\TurnSet^*\times\mathcal{A}\to\Delta'\mathcal{E}$ which are implementable by a $\tau$-POM. For reasons of clarity, we further restrict $\calMpolenv^\tau$ to \textit{fully supported universes}: 
\begin{definition}[Fully supported universe]\label{def:fully-supported-universe-measure}
    A universe $\Mvu$ is \textit{fully supported} iff it is a measure\footnote{This implies that $\sum_{\Hist\in\TurnSet^t}\Mvu(\Hist[t])=1$ for all $t\geq 1$. Note that this does not necessarily hold if $\Mvu$ is a semimeasure.}, and $\Mvu(\HistA)>0$ and $\Mvu(\HistA a)>0$ for all $\HistA \in \TurnSet^*$ and $a \in \calA$.
\end{definition}
The above definition refines \cref{def:fully-supported-universe} of fully supported universes in \cref{sec:embedded-bayesian-agents}, since in this section we distinguish between measures and semimeasures.

We use $\calMpolenvcheck^\tau$ to indicate the set of fully supported universes that are also part of $\calMpolenv^\tau$, and similarly $\calMpolcheck^\tau$ and $\calMenvcheck^\tau$ for policies and environments derived from universes inside $\calMpolenvcheck^\tau$. We have that $$\calMpolenvcheck^\tau \subset \calMpolenv^\tau \subset \calMuni^\tau.$$
When the prior beliefs are coupled on the restricted class $\calMpolenvcheck^\tau$, they will also be coupled on the other two classes, and hence it is sufficient for our purposes to show coupledness on $\calMpolenvcheck^\tau$. We refer the interested reader to \cref{app:structural-similarities} which generalizes our study on $\calMpolenvcheck^\tau$ to $\calMpolenv^\tau$. Finally, we define the prior $\check{w}^{U,\tau}(\Mvu)$ over $\calMpolenvcheck$ by renormalizing $w^{U,\tau}(\Mvu)$:
\begin{align}\label{eq:restricted-solomonoff-prior-universes}
    \check{w}^{U,\tau}(\Mvu):= \frac{w^{U,\tau}(\Mvu)}{\sum_{\Mvu' \in \calMpolenvcheck}w^{U,\tau}(\Mvu')}, ~~~ \check{w}^{U,\tau}(\pi):=\sum_{\Mve' \in \calMenvcheck}\check{w}^{U,\tau}(\Mve'^\pi), ~~~\check{w}^{U,\tau}(\Mve):=\sum_{\pi' \in \calMpolcheck}\check{w}^{U,\tau}(\Mve^{\pi'}).
\end{align}

\paragraph{Functional similarities.} We can readily apply the \textit{degree of functional similarities} $S$ (\eqref{eq:structural-similarities-shannon}) introduced in \cref{sec:emb-bay-agents-structural-similarities} to universes $\Mvu \in \calMpolenvcheck^\tau$, resulting in
\begin{align}
    \check{S}(\Mvu, U, \langle \cdot \rangle):= S(\Mvu, \check{w}^{U,\tau}) = \log \frac{\check{w}^{U,\tau}(\Mvu)}{\check{w}^{U,\tau}(\pi)\check{w}^{U,\tau}(\Mve)}, ~~ \textrm{with} ~~ (\pi,\Mve) = f(\Mvu),
\end{align}
where we used the unique mapping $f$ from a fully supported universe $\Mvu$ to its conditionals $\pi$ and $\Mve$, and where we introduced the notation $\check{S}(\Mvu, U, \langle \cdot \rangle)$ to highlight that this notion of functional similarities depends on the choice of reference machine $U$ and encoding $\langle M\rangle$. Let us now connect this notion of functional similarities to joint compressibility of $\pi$ and $\Mve$. Combining equations \ref{eq:solomonoff-prior-universes} and \ref{eq:restricted-solomonoff-prior-universes}, we have
\begin{align*}
    \check{w}^{U,\tau}(\Mvu)=& \frac{1}{C} \sum_{\substack{M\in\calMapom:\\\Mvu_M^\tau=\Mvu}}2^{-K_U(\langle M\rangle)}\,,\\
     \check{w}^{U,\tau}(\pi)=& \frac{1}{C}\sum_{\Mve' \in \calMenvcheck}\sum_{\substack{M\in\calMapom:\\\Mvu_M^\tau=\Mve'^\pi}}2^{-K_U(\langle M\rangle)}\,,\\ \check{w}^{U,\tau}(\Mve)=& \frac{1}{C}\sum_{\pi' \in \calMpolcheck}\sum_{\substack{M\in\calMapom:\\\Mvu_M^\tau=\Mve^{\pi'}}}2^{-K_U(\langle M\rangle)}\,,\\
     C:=&\sum_{\Mvu' \in \calMpolenvcheck}w^{U,\tau}(\Mvu')\,.
\end{align*}

It is worth noting that $C$ is an absolute constant that only depends on the prior $w^{U,\tau}$.

The following is an informal discussion that aims to provide some intuition on the degree of functional similarity $\check{S}(\Mvu, U, \langle \cdot \rangle)$ from the perspective of joint compressibility of the policy and the environment. 

Let us define the complexities of $\Mvu$, $\pi$ and $\Mve$ as the following minimum description lengths
\begin{align*}
    &K_U(\Mvu):= \min \{K_U(\langle M \rangle): M \in \calMapom, \Mvu^\tau_M=\Mvu\}\,,\\
    &K_U(\pi):= \min \{K_U(\langle M \rangle): M \in \calMapom, \exists\Mve \in \calMenvcheck^\tau, \Mvu^\tau_M=\Mve^\pi\}\,, \\
    & K_U(\Mve):= \min \{K_U(\langle M \rangle): M \in \calMapom, \exists\pi \in \calMpolcheck^\tau, \Mvu^\tau_M=\Mve^\pi\}\,.
\end{align*}

Since sums of exponentials can often be approximated well by their largest element, as the tail of an exponential decays quickly, one might write\footnote{We emphasize that the equation $\check{w}^{U,\tau}(\Mvu) \approx 2^{-K_U(\Mvu)} / C$ should only be taken informally as we have not made the meaning of the approximation $\approx$ precise. One potential interpretation of this approximation is equality up to multiplicative constants. While this is correct for the normal Solomonoff prior considering monotone Turing machines without probabilistic oracles \citep[Theorem 4.3.3]{lv19_kolmogorov}, we do not know whether the same is true when we consider arbitrary probabilistic oracles. Nevertheless, if we replace the exponential $2^{-K_U(\Mvu)}$ by $\alpha^{-K_U(\Mvu)}$ for some $\alpha>2$ in all our equations, then we can indeed show that $\check{w}^{U,\tau}(\Mvu)$ is equal to $\alpha^{-K_U(\Mvu)}$ up to multiplicative constants.} $\check{w}^{U,\tau}(\Mvu) \approx 2^{-K_U(\Mvu)} / C$, $\check{w}^{U,\tau}(\pi) \approx 2^{-K_U(\pi)} / C $ and $\check{w}^{U,\tau}(\Mve) \approx 2^{-K_U(\Mve)}/C$, and hence we have that 
\begin{equation}
    \label{eq:approximate-functional-similarity-Kolmogorov}
    \check{S}(\Mvu, U, \langle \cdot \rangle) \approx K_U(\pi) + K_U(\Mve) - K_U(\Mvu) + \log(C)\,.
\end{equation}
The above informal treatment shows the connection between the intuition of functional similarities as joint compressibility of the policy $\pi$ and environment $\Mve$, and the shortest description lengths of $\pi$, $\Mve$ and $\Mvu$.

\paragraph{The Solomonoff prior is always coupled.} Let us first formalize what we mean by \textit{coupled} and \textit{decoupled}.
\begin{definition}[Coupledness and decoupledness on $\calMpolenvcheck^\tau$]\label{def:decoupled-on-polenvcheck}
We call the prior beliefs $w^{U,\tau}(\Mvu)$ \textit{decoupled on $\calMpolenvcheck^\tau$} iff we have that for all $\Mvu \in \calMpolenvcheck^\tau$
$$\check{w}^{U,\tau}(\Mvu) = \check{w}^{U,\tau}(\pi)\check{w}^{U,\tau}(\Mve), \quad \textrm{with}~~ (\pi,\Mve) = f(\Mvu)\,.$$
We call the prior beliefs $w^{U,\tau}(\Mvu)$ \textit{coupled on $\calMpolenvcheck^\tau$} iff they are not decoupled on $\calMpolenvcheck$.

We call $w^{U,\tau}(\Mvu)$ \textit{boundedly coupled on $\calMpolenvcheck^\tau$} iff there exists absolute positive constants $C_1<C_2$ such that for all $\Mvu \in \calMpolenvcheck^\tau$ we have\footnote{It is worth noting that for bounded coupling, we can write the condition in terms of the unnormalized prior $w^{U,\tau}$ directly: $w^{U,\tau}(\Mvu)$ is boundedly coupled on $\calMpolenvcheck^\tau$ iff there exists absolute positive constants $C_1<C_2$ such that for all $\Mvu \in \calMpolenvcheck^\tau$ we have
$$C_1 w^{U,\tau}(\pi)w^{U,\tau}(\Mve)\leq w^{U,\tau}(\Mvu) \leq C_2 w^{U,\tau}(\pi)w^{U,\tau}(\Mve), \quad \textrm{with}~~ (\pi,\Mve) = f(\Mvu)\,.$$
}
$$C_1\check{w}^{U,\tau}(\pi)\check{w}^{U,\tau}(\Mve)\leq \check{w}^{U,\tau}(\Mvu) \leq C_2 \check{w}^{U,\tau}(\pi)\check{w}^{U,\tau}(\Mve), \quad \textrm{with}~~ (\pi,\Mve) = f(\Mvu)\,.$$
Obviously, a decoupled prior is boundedly coupled.

We call $w^{U,\tau}(\Mvu)$ unboundedly coupled on $\calMpolenvcheck^\tau$ iff it is not boundedly coupled on $\calMpolenvcheck^\tau$.
\end{definition}

The next theorem shows that the Solomonoff prior is always unboundedly coupled on $\calMpolenvcheck^\tau$, regardless of the choice of $U$ and encoding $\langle M \rangle$.

\begin{theorem}
    \label{thm:mupi-structural-similarities-simplified}
    For every fixed choice of:
    \begin{itemize}
        \item a canonical encoding of $\langle M\rangle$ of APOMs as binary strings,
        \item a universal monotone Turing machine $U$,
        \item a probabilistic oracle $\tau$ (which may or may not be a RUI-oracle),
        \item finite spaces $\calA$ and $\calE$, with $|\mathcal{A}|\geq 2$ and $|\calE|\geq 2$, and complete prefix encodings thereof,
    \end{itemize}
    we have:
    \begin{enumerate}
        \item[(a)] The Solomonoff prior $w^{U,\tau}$ is always unboundedly coupled on $\calMpolenvcheck^\tau$ (cf. \cref{def:decoupled-on-polenvcheck});
        \item[(b)] There are fully-supported policies $\pi$ and environments $\Mve$ sharing arbitrarily large degrees of functional similarity $\check{S}(\Mve^\pi, U, \langle \cdot \rangle)$, i.e., for every $s>0$, there is at least some $\pi\in \check{\calM}^{\tau}_{\mathrm{pol}}$ and some $\Mve\in \check{\calM}^{\tau}_{\mathrm{env}}$ such that
        $$\check{S}(\Mve^\pi, U, \langle \cdot \rangle)>s\,.$$
    \end{enumerate}
\end{theorem}
\begin{proof}
    See \cref{app:structural-similarities} and the proof of \cref{thm:mupi-structural-similarities}.
\end{proof}

The results of the above theorem are not too surprising: One can design two APOM machines $M_p$ and $M_e$, one for a policy and one for an environment, which are algorithmically very similar so that $K_U(\langle M_p\rangle)\approx K_U(\langle M_e\rangle)$, but at the same time both $M_p$ and $M_e$ are sufficiently complex so that
$K_U(\langle M_p\rangle)\gg s$ and $K_U(\langle M_e\rangle)\gg s$.
If we let $M_u$ be the APOM machine implementing the universe in which $M_p$ and $M_e$ interact and alternate, then we can see that $K_U(\langle M_u\rangle)\approx K_U(\langle M_p,M_e\rangle)\approx K_U(\langle M_p\rangle)\approx K_U(\langle M_e\rangle)$, and hence
$K_U(\langle M_p\rangle) + K_U(\langle M_e\rangle)-K_U(\langle M_u\rangle)\gg s$. Let $\Mvu$, (resp. $\Mvp,\Mve$) be the universe (resp. policy, environment) induced by the POM $M_u^\tau$ (resp., $M_p^\tau$, $M_e^\tau$). If $w^{U,\tau}(\Mvu)\approx 2^{-K_U(\langle M_u\rangle)}$, $w^{U,\tau}(\Mvp)\approx 2^{-K_U(\langle M_p\rangle)}$ and $w^{U,\tau}(\Mve)\approx 2^{-K_U(\langle M_e\rangle)}$, then from \eqref{eq:approximate-functional-similarity-Kolmogorov} we should expect that
$$\check{S}(\Mvu, U, \langle \cdot \rangle) \approx K_U(\langle M_p\rangle) + K_U(\langle M_e\rangle)-K_U(\langle M_u\rangle)\gg s.$$
This is essentially the core intuition behind the proof of \cref{thm:mupi-structural-similarities-simplified}. Importantly, there are a few technical challenges to make this intuitive argument formal and precise. For example, we cannot directly use the intuitive notion that $w^{U,\tau}(\Mvu)\approx 2^{-K_U(\langle M_u\rangle)}$, as we do not know whether the approximation holds up to a multiplicative constant if $\tau$ is an arbitrary probabilistic oracle. Furthermore, the prior $w^{U,\tau}$ adds up contributions from every machine $M$ for which $\Mvu_M^\tau=\Mvu$: $$ w^{U,\tau}(\Mvu)=
\frac{1}{C} \sum_{\substack{M\in\calMapom:\\\Mvu_M^\tau=\Mvu}}2^{-K_U(\langle M\rangle)}\,.$$ 
Some of these machines might be very different from $M_u$ and we cannot say that $K_U(\langle M\rangle)\approx K_U(\langle M_p\rangle)$ for all these machines.

In \cref{app:structural-similarities}, we show that this theorem not only implies the (unbounded) coupledness of the Solomonoff prior on $\calMpolenvcheck^\tau$, but also on $\calMpolenv^\tau$. Solomonoff induction is arguably the ideal prediction method for predicting future action-percept trajectories, without requiring stationarity or ergodicity assumptions. Hence, \cref{thm:mupi-structural-similarities-simplified} shows the importance of taking functional similarities into account when making predictions about the future, as all Solomonoff priors are coupled due to functional similarities. This further illustrates the important conceptual advances of embedded Bayesian agents and embedded-AIXI over their decoupled counterparts: The incorporation of functional similarities into their predictions and resulting behavior leads to new kinds of predictions and behavior as detailed in Sections \ref{sec:embedded-bayesian-agents}--\ref{sec:subj-emb-eq}, and is motivated from first principles through Solomonoff induction (cf. Theorem~\ref{thm:mupi-structural-similarities-simplified}).

\subsubsection{Universes sharing a common algorithmic structure}
The proof of \cref{thm:mupi-structural-similarities-simplified} in \cref{app:structural-similarities} relies on a constructive argument: It defines a specific (and contrived) family of universes where policies and environments are explicitly built to share a computational backbone. This section delves into the general principle underlying that proof. We formalize the idea that policies and environments can be composed from common "templates" or "subroutines", and show that the Solomonoff prior naturally assigns higher probability to universes where such algorithmic sharing occurs. This provides a deeper, more foundational reason for why the prior is always coupled. We only provide here a high-level and a very condensed explanation of the concepts and the results. Refer to \cref{app:structural-similarities-algorithmic} for a thorough treatment of the subject.

The core idea is to model computational templates as programs with placeholders. We achieve this by generalizing our notion of oracle machines.
\begin{definition}[Multi-Oracle Machines]\label{def:multi-oracle-machines}
An $n$-abstract probabilistic oracle machine ($n$-APOM) is a monotone Turing machine with access to $n$ distinct oracles, $O_1, \dots, O_n$. The first oracle, $O_1$, is treated as the primary oracle (e.g., our fixed $\tau$), while $O_2, \dots, O_n$ serve as placeholders for unspecified subroutines.% An $n$-APOM where the $n-1$ subroutines are specified by other POMs is an $n$-POM.
\end{definition}
This framework allows us to define program templates. An $n$-APOM represents a general computational structure, and a concrete program (a 1-POM) can be created by "plugging in" other programs (1-APOMs) to serve as the subroutines. This composition is defined formally as follows:

\begin{definition}[Composition of Oracle Machines]\label{def:composition-om}
Given an $n$-APOM, $M$, and $n-1$ standard $1$-APOMs, $M_2, \dots, M_n$, we can construct a new $1$-APOM, denoted $M[M_2, \dots, M_n]$. This new machine simulates $M$ as follows:
\begin{itemize}
    \item When $M$ calls its first oracle, $O_1$, the new machine passes this query to its own single oracle.
    \item When $M$ calls any other oracle $O_i$ (for $i \in \{2, \dots, n\}$), it instead simulates the execution of the corresponding machine $M_i$. It feeds the query string as input to $M_i$ and uses the output of $M_i$ as the oracle's answer. While executing $M_i$, any call to the (single) oracle of $M_i$ is directed to the single oracle of the new machine.
\end{itemize}
\end{definition}

This leads to the formal definition of an algorithmic structure.
\begin{definition}[Algorithmic Structure]\label{def:algorithmic-structure}
An $n$-APOM, $M$, defines an algorithmic structure. A standard 1-APOM, $M'$, is said to possess the structure defined by $M$ if there exist $n-1$ other 1-APOMs, $M_2, \dots, M_n$, such that $M'$ is computationally equivalent to the composite machine $M[M_2, \dots, M_n]$. We can define structures for policies, $M_p$, environments, $M_e$, or entire universes, $M_\Mvu$. See \cref{app:structural-similarities-algorithmic} for more details.
\end{definition}

We will be interested in subroutine machines that almost surely halt for every input.

\begin{definition}[proper subroutines]\label{proper-subroutines}
    A POM $M^\tau$ is said to be a proper subroutine if for every input $x\in\calB^*$, the POM $M^\tau$ almost surely halts. In other words, $M^\tau$ induces a mapping $M^\tau:\calB^*\to\Delta\calB$.
\end{definition}

The proof of \cref{thm:mupi-structural-similarities-simplified} (and its formal version \cref{thm:mupi-structural-similarities}) implicitly defines an algorithmic structure where both policy and environment are built from the same underlying function. One key property of this specific algorithmic structure, and which was helpful for the proof of  \cref{thm:mupi-structural-similarities-simplified}, was the \textit{identifiability of its subroutine}, which we can formalize as follows:

\begin{definition}[Identifiable Subroutines]\label{def:identifiable-subroutines}
An algorithmic structure $M$ has identifiable subroutines w.r.t. the probabilistic oracle $\tau$, if there exist "identifier" programs $\{I_i\}_{i=2}^n$ that can computationally recover the behavior of any proper subroutine $M_i^\tau$ simply by observing the input-output behavior of the complete, composed machine $M[M_2, \dots, M_n]^\tau$. More precisely, each $I_i$ is a 2-APOM and $I_i[M[M_2, \dots, M_n]]^\tau$ is distributionally equivalent to $M_i^\tau$ for all $i\in\{2,\ldots,n\}$, i.e., for every $x\in\mathcal{B}^*$ and every $y\in\calB$, we have
    $$\mathbb{P}[I_i[M[M_2,\ldots,M_n]]^\tau(x)=y]=\mathbb{P}[M_i^\tau(x)=y]\,.$$

In essence, the subroutines' contributions are not irrevocably hidden by the main computation. See \cref{example:identifiable-subroutine-simple,example:complex-algorithmic-structure-identifiable-subroutines} for an illustration of algorithmic structures with identifiable subroutines.
\end{definition}

Now we are ready to state the following theorem, which can be seen as a generalization of \cref{thm:mupi-structural-similarities-simplified} to arbitrary algorithmic structures with identifiable subroutines:

\begin{theorem}
\label{thm:mupi-structural-similarities-general}
Let $M_p$ and $M_e$ be algorithmic structures for policies and environments, respectively, using the same number of subroutines. If both $M_p$ and $M_e$ have identifiable subroutines, and the set of fully-supported universes where they share identical subroutines is infinite, then for any Solomonoff prior $w^{U,\tau}$, there exist fully-supported policies $\pi$ (possessing structure $M_p$) and environments $\Mve$ (possessing structure $M_e$) with arbitrarily large degrees of functional similarity. That is, for any $s>0$, there is at least one such pair $(\pi, \Mve)$ for which $\check{S}(\Mve^\pi, U, \langle \cdot \rangle) > s$.
\end{theorem}
\begin{proof}
    See \cref{app:structural-similarities-algorithmic} and the proof of \cref{thm:mupi-structural-similarities-through-algorithmic-structures}.
\end{proof}
This theorem reveals a fundamental principle: The coupledness of the Solomonoff prior is not an accident of a particular construction but the manifestation of a general phenomenon. Whenever policies and environments can be described by general computational templates with common (and identifiable) components, such algorithmic similarities will cause the policies and environments to have large degrees of functional similarities w.r.t. the Solomonoff prior (as per \cref{def:degree-of-functional-similarity-fully-supp}), and hence the Solomonoff prior will be ``highly coupled" for such policies and environments.

\section{Discussion}\label{sec:discussion}
\subsection{MUPI as a coherent framework for learning in multi-agent systems}

This work introduced the MUPI framework to address fundamental limitations in current approaches to multi-agent learning with model-free RL, moving from retrospective learning to prospective learning and from a paradigm of decoupled agency to one of embedded agency.\footnote{We note that the prospective learning aspect is also satisfied by AIXI agents and their multi-agent variants (e.g., reflective AIXI \citep{fallenstein2015reflective,leike2016formal}).} For prospective learning, we have shown that principled anticipation of the future in non-stationary, multi-agent worlds is possible through Bayesian sequence prediction over general model classes. This stands in contrast to retrospective model-free RL methods that compute policy updates based on past data, which is often outdated in environments with other learning agents. A cornerstone of this prospective capability of Bayesian prediction is the \textit{grain-of-truth property}: By ensuring the agent's model class contains the true universe including the agent itself, the Bayesian predictions are guaranteed to converge to the ground-truth distribution, resulting in principled prospective predictions even when the future is changing relative to the past. By leveraging the framework of reflective oracles, and introducing our new framework of the reflective universal inductor, we designed universal embedded agents (embedded-AIXI agents) with a Bayesian mixture universe satisfying the grain-of-truth property, thereby solving the general embedded grain-of-truth problem. The resulting predictive model, which jointly forecasts an agent's own actions and its percepts, is conceptually aligned with modern foundation models that also predict both actions and percepts.

Leveraging the Bayesian prediction models $\Mmu$, we introduced \textit{prospective policy learning} methods as $k$-step planning within $\Mmu$, which are similar to Self-AIXI \citep{catt2023self} and reminiscent of classical policy iteration and the MuZero family of methods, but with the crucial difference that the $Q$-values are estimated based on prospective predictions rather than past data under the assumption of stationarity. By planning within a predictive model that anticipates its own future improvements and the learning of others, the agent becomes both \textit{self-learning aware} and \textit{co-player learning-aware}. This generalizes existing work on co-player learning-aware RL by enabling \textit{mutual prospective prediction} without making inconsistent assumptions about other agents' learning algorithms.

A key challenge with mutual prediction is avoiding the pitfalls of infinite recursion of the form \textit{``I predict that you predict that I predict....''} MUPI resolves this through the construction of its predictive model, the Reflective Universal Inductor (RUI). Rather than getting caught in an infinite computational loop at runtime, a MUPI agent consults the RUI $\Mmu$, which is, by definition, a consistent fixed point for such recursive beliefs.
This consistency emerges because all agents perform Bayesian updates over a shared, universal hypothesis class of universes that already contain universes with agents making predictions about each other. The RUI thus sidesteps infinite recursion, providing a coherent prediction conditioned on an agent's personal history while having already accounted for mutual modeling, up to infinite orders of theory of mind.

This theoretical construct has a practical analogue in a neural prediction model, like a transformer, trained on data from interacting agents engaging in mutual prediction. Such a model always returns a prediction without running into infinite recursions, as it does not explicitly run a simulation of agents predicting each other.
In practice, the learned prediction model may not make perfect predictions under uncertainty. In such cases, it can be beneficial to apply reasoning traces within the prediction model, explicitly reasoning about the ongoing mutual prediction, up to a depth of $k$ recursion steps. This is reminiscent of applying $k^{\text{th}}$-order theory of mind, and the $k$-level reasoning strategy within Economics theory for bounded rationality \citep{crawford2013structural}. It is crucial to recognize that such $k$-level reasoning provides no new \textit{information}---as no new observations are made---but rather offers additional \textit{computation} to help an imperfect model better approximate ideal Bayesian inference. The predictions of an ideal MUPI agent with a perfect RUI are already flawlessly Bayesian, making any further explicit reasoning computationally redundant.

Besides the shift from retrospective to prospective learning, the second major shift we proposed is from decoupled to embedded agency.
By treating the agent as part of the universe it models, MUPI enables true self-modeling and an infinite-order theory of mind. An important consequence of this embedded viewpoint is the ability to reason about \textit{functional similarities} between oneself and others, leading to coupled beliefs where knowledge about one's own policy informs predictions about other agents within the environment. This mechanism unlocks novel strategies for both prediction and behavior. For prediction, it enables \textit{similarity-aware prediction}, allowing an agent to anticipate the actions of others in novel situations by considering its own planned behavior and reasoning that ``similar agents behave similarly in similar situations.'' For behavior, it enables new forms of cooperation and coordination through the \textit{embedded equilibria}, a family of new solution concepts that can, for instance, justify cooperation in the Twin Prisoner's Dilemma---an outcome inaccessible to classical game theory. The study of such functional similarities is not merely a theoretical curiosity; it can have important practical implications. As we move toward deploying vast numbers of AI agents, many of which are based on shared foundation models, similarity awareness may offer a powerful mechanism for achieving robust coordination with minimal communication, possibly paving the way for effective social scaling---increasing the capabilities of a society of agents by increasing their numbers.

\subsection{The connection between the MUPI framework and current foundation models}
The framework of embedded Bayesian agents, centered around a predictive model of both one's own actions and incoming observations, shares important similarities with modern foundation models and their agentic derivatives. First, the mixture universe, $\Mmu$, functions as a \textit{joint predictive model} that forecasts both the agent's own actions and its incoming percepts. This is directly analogous to the way many foundation models are trained to predict the next token in a sequence, which can represent either an observation from the world or an action taken by the agent.
Second, the embedded Bayes-optimal agent, which is the optimal policy w.r.t. the prediction model $\Mmu$, is conceptually similar to model-based RL agents that learn a world model and then use it to simulate imagined future trajectories to find an optimal policy. The more practical \textit{$k$-step planner agent} (cf. \cref{sec:k-step-planning}) closely resembles the MuZero family of algorithms, which also learn a model and perform a limited look-ahead search to improve their policy. In its simplest form, a \textit{one-step planner ($k=1$)} dispenses with explicit planning and instead selects the action that maximizes the learned $Q$-value function, a strategy reminiscent of classic methods like $Q$-learning and TD(0) \citep{sutton_reinforcement_2018}. Conceptually, our framework is flexible enough to accommodate other forms of policy improvement that rely on prediction with a sequence model, such as the chain-of-thought reasoning increasingly used in large language models. However, it is important to note that the convergence guarantees we have established (Theorems \ref{thrm:convergence-to-sde-correlated} and \ref{thrm:convergence-to-epsdelta-sde-correlated}) are specific to the embedded Bayes-optimal and $k$-step planner agents; extending these proofs to other policy improvement mechanisms requires additional theoretical developments.

Despite these parallels, the primary divergence between MUPI and current foundation models lies in the learning process itself. An embedded Bayesian agent learns through \textit{explicit Bayesian inference}: It begins with a prior belief over all possible universes and performs exact posterior updates conditioned on its entire life history---a single, unbroken sequence of interactions. In contrast, foundation models are typically trained using stochastic gradient descent on vast datasets chunked into batches of ``episodic trajectories.'' Hence, the explicit Bayesian inference of Bayesian agents is replaced by a combination of \textit{in-weight learning}, where past training data is distilled into the model parameters, and \textit{in-context learning}, where the model performs rapid, on-the-fly inference based on the current context \citep{von2023transformers, von2023uncovering, xie2021explanation, ortega2019meta, elmoznino2024context}. This distinction raises a critical question: Can foundation models, trained this way, achieve the kind of principled prospective prediction that MUPI guarantees?

This difference in training methodology means that a model-based RL agent can still be fundamentally \textit{retrospective} if its learned model is overfitted to recent past data and fails to generalize to a non-stationary future that can be different from the recent past. Achieving true prospective prediction requires principled methods that can anticipate future changes \citep{bornschein2024transformers}. Some theoretical work suggests that stochastic gradient descent can approximate sampling-based Bayesian inference \citep{mandt2017stochastic, liu2019deep, chaudhari2018stochastic}. However, training neural sequence models to robustly approximate Bayesian prediction starting from an Occam's razor prior---thereby ensuring they can make accurate, prospective forecasts in dynamically changing worlds---remains an open problem \citep{bornschein2024transformers,de2024prospective}.

\subsection{Active exploration with embedded Bayesian agents}
As embedded Bayesian agents are fundamentally uncertain about which universe they live in, explorative behavior that reduces this uncertainty is critical. Bayesian agents with sufficiently broad priors naturally incorporate a form of principled, directed exploration through the \textit{value of information} \citep{howard1966information, chalkiadakis2003coordination}, where an agent is intrinsically motivated to take actions that reduce uncertainty about the world if that information is expected to improve future rewards, for example through enabling more detailed planning. However, as prior work has shown, this Bayesian form of exploration is often insufficient to overcome \textit{dogmatic beliefs} about catastrophic outcomes \citep{orseau2010optimality, leike2015bad}. An agent may hold a belief that any deviation from its current policy will result in a catastrophically low reward, and since it never deviates, this flawed belief is never corrected.

The coupled beliefs of embedded Bayesian agents have the potential to further aggravate this problem. For a decoupled agent, a dogmatic belief is about an external state of the world. For an embedded agent, taking an action can now be interpreted as direct \textit{evidence} about the environment and the obtainable rewards. An embedded Bayesian agent can therefore hold a dogmatic belief that simply \textit{deviating} from a certain policy is strong evidence that it lives in a low-reward universe. This creates a self-fulfilling trap, preventing the agent from ever gathering the data needed to correct its wrong beliefs. We show this result formally in \cref{prop:dogmatic-beliefs}, which demonstrates that for any deterministic policy and environment, there exist dogmatic beliefs for which the agent's embedded best response is precisely that policy.

These results on dogmatic beliefs have motivated the use of more active exploration strategies, such as Thompson sampling \citep{leike2016thompson, cohen2019inq}. Equipping AIXI agents with such active exploration strategies can lead to asymptotically optimal policies, guaranteeing that for any ground-truth environment within the model class, such agents converge to an optimal policy. Importantly, however, such additional active exploration also often leads to unsafe behavior due to its inherent randomness \citep{cohen2021curiosity},
which could potentially incapacitate the agent. These two seemingly opposite characteristics of active exploration enabling optimal behavior but also potentially incapacitating the agent are compatible, as any policy is optimal when in a state from which one can no longer affect the world. Hence, finding robust exploration strategies to overcome dogmatic beliefs while safely exploring the environment remains a critical open challenge in the field.

Investigating such robust exploration strategies within the MUPI framework is an exciting direction for future research. Adapting existing active exploration methods for decoupled agents, such as Thompson sampling \citep{leike2016thompson}, to the embedded Bayesian agent setting will be crucial. A key theoretical goal would be to show convergence towards \textit{embedded equilibria} in the general MAGRL setting, by ensuring agents can escape the suboptimal traps of purely subjective equilibria built on dogmatic beliefs.

\subsection{The connection between the MUPI framework and Evidential, Causal, and Functional Decision Theory}
At its core, MUPI is a form of \textit{action evidential decision theory (EDT)} \citep{everitt2015sequential}, as it computes the future consequences of an action $a$ by conditioning its Bayesian prediction model $\Mmu$ on that action. While causal decision theory (CDT) is often viewed as the gold standard for rational choice, its utility is limited to settings where the system being modeled does not contain the agent. 
In such decoupled settings, the agent's action is an external intervention that creates a \textit{distributional shift} w.r.t. some `default behavior of the agent' in the system it acts upon. To correctly predict such distributional shifts, the action should be treated as a \texttt{do}-intervention upon the system being modeled \citep{pearl2009causality}, as is done by CDT. 
Hence, CDT is the correct formalism when an agent is truly separate from the environment it is modeling.

However, in the context of embedded agency, the agent's decision-making (i.e., policy) is a process unfolding \textit{within} the universe, not an intervention from the outside. There is no external force creating a distributional shift; there is only the natural evolution of the universe, which includes the agent's deliberation and subsequent action. In this setting, EDT is the more natural and powerful framework. Applying a CDT-style \texttt{do}-intervention would be artificial and actively detrimental, as it would force the agent to ignore the predictive evidence its own choices provide about the world, especially given functional similarities with other agents. By severing these evidential links, CDT can lead to less accurate predictions and, consequently, to suboptimal behavior such as defection on the Twin Prisoner's Dilemma.

Furthermore, MUPI provides a new formalism that captures some of the core intuitions of \textit{functional decision theory} (FDT) without resorting to its most problematic element: \textit{logical counterfactuals}. FDT advises an agent to choose the action that would yield the best outcome if its decision-making \textit{function} were to produce that output, thereby accounting for all instances of its own algorithm in the world. This enables FDT to coordinate and cooperate well with copies of itself. FDT must reason about what \textit{would have happened} if its deterministic algorithm had produced a different output, a notion of \textit{logical counterfactuals} that is not yet mathematically well-defined. MUPI achieves a similar outcome through a different mechanism: the combination of treating universes including itself as \textit{programs}, while having \textit{epistemic uncertainty} about which universe it is inhabiting---including which policy it is itself running. As explained in \cref{remark:eba-fdt}, from the agent's internal perspective, it acts \textit{as if} its choice of action decides which universe it inhabits, including which policy it is running. When it contemplates taking action $a$, it updates its beliefs $w(\Mvu|\HistM a)$, effectively concentrating probability mass on universes compatible with taking action $a$. 
Because the agent’s beliefs about its own policy are coupled with its beliefs about the environment through functional similarities, this process allows the agent to reason about how its choice of action relates to the behavior of other agents that share functional similarities.
This ``as if'' decision-making process allows MUPI to manifest the sophisticated, similarity-aware behavior FDT aims for, but on the solid foundation of Bayesian inference rather than on yet-to-be-formalized logical counterfactuals.

\subsection{Philosophical implications}

Beyond its technical contributions to multi-agent learning, the MUPI framework offers a formal lens through which to examine long-standing philosophical questions about the nature of consciousness and free will. By grounding modern theories of consciousness and free will in a computational model of embedded agency, MUPI provides a concrete language for discussing how these phenomena could arise in both biological and artificial minds.

\paragraph{Self-models and Consciousness.}
At the heart of MUPI lies a powerful conception of self-awareness: The agent's predictive model $\Mmu$ must make predictions about the agent using that very model to make predictions. This creates a scenario of self-reference, reminiscent of Douglas Hofstadter's concept of \textit{strange loops} as a cornerstone of consciousness \citep{hofstadter2007strange}.
This recursive self-model is not a mere theoretical curiosity; it serves a critical, functional purpose that aligns with modern functional theories of consciousness like Michael Graziano's \textit{Attention Schema Theory} \citep{graziano2013consciousness, graziano2015attention}. Graziano argues that consciousness is a computational mechanism that evolved for a specific purpose: to model one's own and others' mental states. According to this view, the \textit{feeling} of being conscious arises from a high-level, predictive model of one's own attentional processes, allowing an agent to better understand and predict on the one hand its own internal processes to effectively control them \citep{conant1970every, richens2025general}, and on the other hand the behavior of other agents by simulating their mental states using its own as a template. 

By including a self-model $\Mmu$ over a class of universes that explicitly includes the agent itself as a core mechanism of embedded agency, MUPI provides a detailed mathematical foundation of this idea of the joint modeling of self and others, while introducing a new behaviorally relevant function of consciousness.
% The framework shows precisely how a universally intelligent agent can construct and utilize a \textbf{self-model}-. This self-model is not merely a navel-gazing exercise; it serves a critical, functional purpose. 
By reasoning about \textit{functional similarities}, the agent leverages its understanding of its own decision-making process to form accurate, prospective predictions about others. This ability to see oneself in others is what unlocks novel forms of cooperation and social coordination, as formalized by the \textit{embedded equilibrium}. 
Furthermore, while Attention Schema Theory mainly focuses on first-order modeling of the neural attention processes, our reflective universal inductor $\Mmu$ highlights the infinite recursions that occur when a prediction model $\Mmu$ models an agent using this prediction model itself. 
In this light, MUPI further operationalizes the role of self-modeling within the Attention Schema Theory for consciousness, by (i) providing an idealized theoretical model organism of a self-modeling embedded Bayesian agent, (ii) introducing a novel functional benefit of the joint modeling of self and others through enabling novel forms of cooperation and social coordination, and (iii) formally connecting Bayesian self-modeling with the self-reference or strange loops arising as a direct consequence of this self-modeling. 

\paragraph{Embedded agency and free will.} The concept of free will has long been debated against the backdrop of a lawful universe. This tension is often framed as the `standard argument' for hard determinism: If the universe is deterministic, its future is fixed by its past, meaning an agent could not have ``acted otherwise". If the universe is instead stochastic, an agent's actions might be mere random occurrences, not attributable to rational agency. In either view, meaningful free will appears illusory. 

The MUPI framework offers a powerful formalism that operationalizes a compatibilist resolution to this dilemma, aligning closely with modern philosophical accounts. Christian List proposes levels-based compatibilism, which distinguishes between physical possibility (what is possible given the precise, low-level micro-state of the universe) and agential possibility (what is possible given an agent's high-level psychological state) \citep{list2014free}. List argues that even if the physical level is deterministic (only one physically possible future), the agential level can be indeterministic, as a single mental state can be realized by many different physical states, each with a different future. 

MUPI provides an alternative computational model for this dichotomy of physical and agential possibilities, which does not rely on the coarse graining of physical states, but rather leverages Bayesian uncertainty about one's own policy. The unknown ground-truth universe $\Mgu$ corresponds to physical possibility; it is the single, lawful\footnote{Note that in the MUPI formalism, the lawful universe can either be deterministic or stochastic.} system whose evolution dictates the agent's actual future policy. In contrast, the agent's predictive model $\Mmu$ corresponds to agential possibility. An agent's "mental state" is captured by its history $\Hist$ and the resulting posterior belief $w(\Mvu \mid \Hist)$. From the agent's embedded perspective, this belief is distributed over many possible universes $\Mvu$ that are all consistent with its past $\Hist$. This epistemic uncertainty about which universe it inhabits, and crucially, which policy it is itself running, means that multiple futures and multiple actions are ``agentially possible" even if the ground-truth universe $\Mgu$ is predetermined.\footnote{It is worth mentioning that Self-AIXI \citep{catt2023self} also has epistemic uncertainty about which policy the agent itself is running, and hence also considers multiple actions as ``agentially possible". Importantly however, Self-AIXI keeps separate beliefs about its own policy and the rest of the universe, and hence does not consider itself as a part of the universe. This makes MUPI, where agents treat themselves as part of the universe, a more conceptually natural framework to discuss the philosophical topic of free will within a lawful universe.}

Furthermore, the very process of decision-making in MUPI aligns with the compatibilist philosophy of Daniel Dennett \citep{dennett2004freedom}. Dennett argues that free will is not a mysterious ability to defy causality, but rather a sophisticated, evolved capacity for rational deliberation and control. An embedded Baysian agent within the MUPI framework embodies this. It is not a passive domino in a causal chain. It actively considers the future consequences of its potential actions by planning within its predictive model $\Mmu$, and it deliberately chooses the action that leads to the most highly-valued outcome. MUPI thus provides a concrete, mathematical model for these compatibilist theories of free will, demonstrating how an agent whose reasoning is part of the universe's causal structure can still possess the freedom to anticipate futures and select the optimal path to its goals.

\subsection{Limitations and future work} This work has focused on providing a conceptual theory for multi-agent learning based on embedded Bayesian agents. While the MUPI framework addresses several fundamental challenges, it also highlights important limitations and opens new avenues for future research.

First, \citet{demski2019embedded} provide an informal survey on the conceptual difficulties in formalizing embedded agency. MUPI addresses some of these, such as providing an \textit{embedded world model} (the RUI) that uses randomization to avoid self-referential paradoxes, and leveraging epistemic uncertainty over its own algorithm as an alternative to the problematic logical counterfactuals discussed by \citet{demski2019embedded} and \citet{yudkowsky2017functional}. However, several other deep challenges remain open. A significant one is how an embedded agent can safely modify its own decision algorithm or goal structure to improve itself, while ensuring it remains aligned with the goals of its previous version \citep{zhao1996incremental, schmidhuber2009ultimate, orseau2011self, orseau2012space}. Investigating these advanced problems of embedded agency through the MUPI framework is an exciting direction for future research.

Second, there is a gap between our conceptual theory and practical applications. The theory leverages Bayesian sequence prediction, Occam's razor priors, and the grain-of-truth property to achieve principled prospective prediction that can anticipate future non-stationarities, such as those caused by other learning agents. How to effectively approximate this powerful predictive capability using practical continual and online learning techniques for neural network sequence models remains a major open problem \citep{bornschein2024transformers, de2024prospective}.

Third, our theoretical agents, the Embedded AIXI agents, are incomputable. Furthermore, to satisfy the grain-of-truth property, they must consider a hypothesis class over incomputable oracle machines. Any practical approximation using computational devices, such as neural sequence models, will likely violate the grain-of-truth property. Furthermore, exact Bayesian posterior updates are generally intractable, necessitating the use of approximate inference methods. This raises an important research question: how robust are the benefits of prospective prediction to these approximations? Understanding whether the performance degrades gracefully when the grain-of-truth property is not perfectly satisfied and inference is imperfect is key to its practical viability.

Finally, a crucial assumption for achieving consistent mutual prediction in both the framework of reflective oracles and our RUI is that all agents utilize the \textit{same} oracle. This shared component results in a form of `precomputed consistency' among the agents, as the oracle's answers are derived from a precomputed fixed point. A challenging and interesting direction for future work is to develop methods that can achieve approximate consistent mutual predictions without relying on such a precomputed fixed point, moving closer to decentralized and ad-hoc coordination.

\section*{Acknowledgements}
We would like to thank Cole Wyeth, Geoff Keeling, Johannes von Oswald, Nino Scherrer, Robert Obryk, Yul Kwon, James Evans, Yanick Schimpf, Maximilian Schlegel, Eric Elmoznino, Winnie Street, Roberta Rocca and the Google Paradigms of Intelligence team for feedback and enlightening discussions. GL and BR acknowledge support from the CIFAR chair program.

\bibliography{MARLAX}

\newpage
\appendix
% \crefname{section}{Appendix}{Appendices}

\section{Table of notations and definitions}
\small
\begin{longtable}{@{}ll@{}}
\toprule
\textbf{Symbol} & \textbf{Description} \\
\midrule
\endhead % Header for subsequent pages

% --- Basic Elements ---
\multicolumn{2}{l}{\textit{\textbf{Single-Agent Setting}}} \\
\addlinespace
$\calA$ & Set of possible actions. \\
$\calE$ & Set of possible percepts (Observations $\times$ Rewards). \\
$\Turn$ & A single turn (action-percept pair). \\
$\TurnSet$ & The set of all possible turns ($\calA \times \calE$). \\
$\Hist$ & A $t$-turn history. \\
$\HistM$ & A history up to, but not including, time $t$. \\
$\HistA$ & An arbitrary-length finite history. \\
\addlinespace
$\Mge$ & Ground-truth environment. \\
$\Mve$ & Generic environment. \\
$\Mme$ & Mixture of environments; the decoupled agent's belief model. \\
\addlinespace
$\Mgu$ & Ground-truth universe. \\
$\Mvu$ & Generic universe. \\
$\Mmu$ & Mixture of universes; the embedded agent's belief model. \\
\addlinespace
$\Mgp$ & Ground-truth policy. \\
$\Mvp$ & Generic policy. Note: overloading notation with ground-truth policy. \\
$\Mmp$ & Mixture of policies. \\
\addlinespace
$\Mve^{\Mvp}$ & Universe, i.e., measure over histories, resulting from combining $\Mve$ and $\Mvp$ \\
$\Mmu^{\Mvp}$ & Universe resulting from replacing self-model of $\Mmu$ with $\Mvp$ (cf. \eqref{eqn:embedded-br-factorization}). \\
\addlinespace
$V_{\Mve^{\Mvp}}(\Hist)$ & Value of history $\Hist$ w.r.t. policy $\Mvp$ and environment $\Mve$. \\
$V_{\Mvu}(\Hist)$ & Value of history $\Hist$ w.r.t. universe $\Mvu$. \\
$Q_{\Mve^\Mvp}(\HistM, a_{t})$ & $Q$-value of action $a_t$ after history $\HistM$ w.r.t. policy $\Mvp$ and environment $\Mve$. \\
$Q_{\Mvu}(\HistM, a_{t})$ & $Q$-value of action $a_t$ after history $\HistM$ w.r.t. universe $\Mvu$. \\
\addlinespace
$\calMpol$ & Class of policies. \\
$\calMenv$ & Class of environments. \\
$\calMuni$ & Class of universes. \\

% --- Multi-Agent Elements ---
\midrule
\multicolumn{2}{l}{\textit{\textbf{Multi-Agent Setting}}} \\
\addlinespace
$\barcalA$ & Joint action space. \\
$\barcalE$ & Joint percept space. \\
$\MATurn$ & A single multi-agent turn. \\
$\MAHist$ & A $t$-turn multi-agent history. \\
$\MAHistA$ & An arbitrary-length finite multi-agent history. \\
$\HistI$ & The personal history for agent $i$. \\
$-i$ & $[N]/i$: the indices of the other agents. \\
$a^{-i}$ & The joint action of all agents except agent $i$. \\
\addlinespace
$\MultiAgentMge$ & Ground-truth multi-agent environment. \\
$\MultiAgentMve$ & Generic multi-agent environment. \\
$\MultiAgentMvp$ & Ground-truth multi-agent policy. \\
$\MultiAgentMvp$ & Generic multi-agent policy. Note: overloading notation with ground-truth multi-agent policy. \\
$\Mve^i$ & Personal environment resulting from combining $\MultiAgentMve$ with other agent's policies. \\
$(\Mve^i)^{\Mvp^i}$ & Personal distribution over personal histories $\HistI$. \\
\addlinespace

% --- Game-Theoretic Concepts (Section 4) ---
\midrule
\multicolumn{2}{l}{\textit{\textbf{Game-Theoretic Concepts (Section 4)}}} \\
\addlinespace
$\Rgh$ & Repeated game history. \\
$\RghM$ & Repeated game history up to time $t$. \\
$\RghA$ & Arbitrary-length repeated game history. \\
\addlinespace
$\MAgeTailM$ & Tail multi-agent ground-truth environment. \\
$\MgeTailM$ & Tail personal ground-truth environment. \\
$\MgpTailM$ & Tail ground-truth policy. \\
$\MmeTailM$ & Tail mixture environment. \\
$\MmuTailM$ & Tail mixture universe. \\
\addlinespace
NE & Nash Equilibrium. \\
SNE & Subjective Nash Equilibrium. \\
CE & Correlated Equilibrium. \\
SCE & Subjective Correlated Equilibrium. \\
EE & Embedded Equilibrium. \\
SEE & Subjective Embedded Equilibrium. \\
CEE & Correlated Embedded Equilibrium. \\
SCEE & Subjective Correlated Embedded Equilibrium. \\
\addlinespace

% --- Universal AI & MUPI (Section 5) ---
\midrule
\multicolumn{2}{l}{\textit{\textbf{Universal AI \& MUPI (Section 5)}}} \\
\addlinespace
$\calM^{\text{LSCSM}}$ & Class of lower semicomputable semimeasures. \\
$\calMapom$ & Class of Abstract Probabilistic Oracle Machines. \\
$\calMrapom$ & Class of restricted APOMs. \\
$O^\tau$ & A probabilistic oracle. \\
$\tau$ & A probabilistic oracle. \\
RUI & Reflective Universal Inductor. \\
RO & Reflective Oracle. \\
$\Mmu_\tau^w$ & The $(w,\tau)$-universal mixture (the RUI). \\
$\Mmu_\tau$ & The Solomonoff $\tau$-universal mixture. \\
$\calMuni^{w,\tau-\mathrm{RUI}}$ & Hypothesis class of universes with RUI. \\
$\calMuni^{\tau-\mathrm{RO}}$ & Hypothesis class of universes with Reflective Oracle. \\
$\bar{\Mmu}_{\tau\text{-}\mathrm{RO}}^w$ & $\tau$-completed universal mixture (for RO). \\

% --- Functional Similarity ---
\midrule
\multicolumn{2}{l}{\textit{\textbf{Functional Similarity (Sections 3.6 \& 5.4)}}} \\
\addlinespace
$S(\Mvu,w)$ & Degree of functional similarity (pointwise mutual information). \\
$S(w)$ & Average degree of functional similarity. \\
$\calI_w(\Mvp;\Mve)$ & Shannon mutual information between policy and environment. \\
$\wUtau(\Mvu)$ & Solomonoff prior over universes (induced by universal monotone Turing machine $U$ and $\tau$). \\
$\KSubU(\langle M\rangle)$ & Kolmogorov complexity of the binary encoding $\langle M\rangle$ of a machine $M$. \\
$\KSubU(\Mvu)$ & Kolmogorov complexity of a universe $\Mvu$. \\
$\KSubU(\Mgp)$ & Kolmogorov complexity of a policy $\Mgp$. \\
$\KSubU(\Mve)$ & Kolmogorov complexity of an environment $\Mve$. \\
$\checkwUtau(\Mvu)$ & Renormalized Solomonoff prior over fully supported universes. \\
$\checkS(\Mvu, U, \langle \cdot \rangle)$ & Algorithmic degree of functional similarity (\cref{sec:mupi-structural-similarities}). \\
\addlinespace

\bottomrule

\end{longtable}

% Requires \usepackage{longtable} and \usepackage{booktabs}
\begin{longtable}{@{}ll@{}}
\toprule
\textbf{Reference} & \textbf{Definition Name} \\
\midrule
\endhead % This repeats the header on subsequent pages

\cref{def:semi-measures} & (Semimeasures and measures) \\
\cref{def:total-variation-distance} & (Total variation distance) \\
\cref{def:dominance} & (Dominance) \\
\cref{def:computable} & (Computable) \\
\cref{def:lsc} & (Lower semicomputable) \\
\cref{def:limit-computable} & (Limit computable) \\
\cref{def:monotone-turing} & (Monotone Turing machine) \\
\cref{def:grain-of-self-uncertainty} & (A grain of uncertainty) \\
\cref{def:embedded-best-response} & (Embedded best response) \\
\cref{def:1-step-eba} & (One-step planner embedded Bayesian agent) \\
\cref{def:k-step-eba} & ($k$-step planner embedded Bayesian agent) \\
\cref{def:got-property} & (The grain-of-truth property) \\
\cref{def:fully-supported-universe} & (Fully supported universe) \\
\cref{def:repeated-games} & (Repeated Games with Perfect Monitoring) \\
\cref{def:decoupled-agents-repeated-games} & (Decoupled Bayes-optimal agent in repeated games) \\
\cref{def:subj-nash-eq} & (Subjective Nash Equilibrium) \\
\cref{def:nash-eq} & (Nash Equilibrium) \\
\cref{def:tail-game} & (Tail games) \\
\cref{def:corr-eq} & (Correlated Equilibria) \\
\cref{def:subj-corr-eq-kalai} & (Subjective Correlated Equilibria \citep{kalai1995subjective}) \\
\cref{def:embedded-agents-repeated-games} & (Embedded Bayes-optimal agent in repeated games) \\
\cref{def:eps-sde} & ($\epsilon$-Subjective Embedded Equilibrium) \\
\cref{def:conditional-completion} & (Conditional Completion) \\
\cref{def:sde} & (Subjective Embedded Equilibrium) \\
\cref{def:ode} & (Embedded Equilibrium) \\
\cref{def:eps-ode} & ($\epsilon$-Embedded Equilibrium) \\
\cref{def:eps-scde} & ($\epsilon$-Subjective Correlated Embedded Equilibrium) \\
\cref{def:eps-delta-scde} & ($(\epsilon,\delta)$-Subjective Correlated Embedded Equilibrium ($(\epsilon,\delta)$-SCEE)) \\
\cref{def:approximate-eba} & ($(k_t,\epsilon_t)$-embedded Bayesian agent) \\
\cref{def:prefix-free-encoding} & (prefix free encoding) \\
\cref{def:probabilistic_oracle} & (Probabilistic oracle) \\
\cref{def:apom} & (APOM) \\
\cref{def:probabilistic-oracle-machine} & (POM) \\
\cref{def:rapom} & (rAPOM) \\
\cref{def:rpom} & (rPOM) \\
\cref{def:universal-mixture-universes} & (Universal mixture universe) \\
\cref{def:solomonoff-universal-prior} & (Solomonoff universal prior) \\
\cref{def:rui-oracle} & ($w$-reflective universal inductor oracle) \\
\cref{def:reflective-oracle} & (Reflective oracle \citep{fallenstein2015reflective_oracles, fallenstein2015reflective}) \\
\cref{def:tau-completion} & ($\tau$-completion) \\
\cref{def:oracle-estimable} & ($\tau$-estimable \citep{wyeth2025limit}) \\
\cref{def:oracle-lsc} & ($\tau$-lower-semicomputable) \\
\cref{def:tau-sampleable} & ($\tau$-sampleable \citep{wyeth2025limit}) \\
\cref{def:fully-supported-universe-measure} & (Fully supported universe) \\
\cref{def:decoupled-on-polenvcheck} & (Coupledness and decoupledness on $\calMpolenvcheck^\tau$) \\

\bottomrule
\end{longtable}

\normalsize

\section{Preliminaries}\label{app:preliminaries}

This appendix covers the theoretical preliminaries. We begin by reviewing foundational concepts:
measure-theoretic notions for random sequences (\cref{app:preliminaries-measure-theoretic-concepts}),
universal Bayesian prediction and the merging of opinions theorem (\cref{app:preliminaries-universal-bayesian-prediction}),
computability and algorithmic information theory (\cref{app:preliminaries-computability-theory}),
and Solomonoff's theory of induction (\cref{app:preliminaries-solomonoff-induction}).
We then apply these to reinforcement learning, defining the general setup (\cref{app:preliminaries-general-rl})
and the universally intelligent agent AIXI \citep{hutter-aixi} (\cref{app:preliminaries-aixi}).
We conclude by motivating our framework, first by discussing the failure of AIXI in embedded settings (\cref{app:preliminaries-aixi-embedded-failure}),
and second by reviewing JAIXI, a variant introduced by \citet{wyeth2025formalizing} to formalize these embedding failures (\cref{app:preliminaries-jaixi}).

\subsection{Measure-theoretic concepts.}
\label{app:preliminaries-measure-theoretic-concepts}

Let $\mathcal{X}$ be an arbitrary countable set. We generally think of $\mathcal{X}$ as an \textit{alphabet} and refer to its elements as \textit{symbols}.

\begin{definition}
    A semiprobability distribution\footnote{A more general definition that is typically used is to consider mappings $\mathrm{Pow}(\mathcal{X})\to[0,1]$ which are $\sigma$-superadditive; the simpler definition here is sufficient for our purposes.} on $\mathcal{X}$ is a mapping $\sigma:\mathcal{X}\to[0,1]$ such that
    $$\sum_{x\in\mathcal{X}}\sigma(x)\leq 1\,.$$
    If the above holds with equality, we get a probability distribution.

    We denote the set of semiprobability distributions on $\mathcal{X}$ as $\Delta'\mathcal{X}$, and the set of probability distributions as $\Delta\mathcal{X}$.
\end{definition}

A semiprobability distribution $\sigma$ on $\mathcal{X}$ can be turned into a probability distribution on 
\begin{equation}
\label{eq:def-X-tilde}
\tilde{\mathcal{X}}:=\mathcal{X}\cup\{\perp\}  
\end{equation}
by assigning the missing probability mass $1-\sum_{x\in\mathcal{X}}\sigma(x)$ to a symbol `$\perp$' which is assumed to be outside $\mathcal{X}$:

\begin{definition}
    We define the canonical completion of a semiprobability distribution $\sigma\in\Delta'\mathcal{X}$ as the probability distribution $\tilde{\sigma}\in \Delta\tilde{\mathcal{X}}$ on $\tilde{\mathcal{X}}:=\mathcal{X}\cup\{\perp\}$ defined as:
    \begin{align*}
        \tilde{\sigma}(x)&=\sigma(x)\,,\quad\forall x\in\mathcal{X}\,,\\
        \tilde{\sigma}(\perp)&=1-\sum_{x\in\mathcal{X}}\sigma(x)\,.
    \end{align*}
    % \rajai{TODO: Replace the '\perp' symbol with something nicer.}
\end{definition}

\begin{remark}
    The above definition motivates the interpretation of a semiprobability distribution $\sigma\in\Delta'\mathcal{X}$ as formally describing a situation where it is possible to observe a (random) sample from $\mathcal{X}$ but it is also possible not to observe any sample from $\mathcal{X}$.
\end{remark}

We write $\mathcal{X}^{*}$ to denote the set of finite $\mathcal{X}$-sequences, i.e.,
$$\mathcal{X}^*:=\bigcup_{n\geq 0}\mathcal{X}^n\,.$$
We adopt the convention that $\mathcal{X}^0=\{\varepsilon\}$ where $\varepsilon$ is the empty sequence/string.

For $x\in\mathcal{X}^n$ and $y\in\mathcal{X}^m$, we write $xy$ to denote the sequence in $\mathcal{X}^{n+m}$ obtained by concatenating $x$ and $y$.

We write\footnote{Note that $A^B$ denotes the set of mappings from $B$ to $A$. Therefore, $\mathcal{X}^\mathbb{N}$ is the set of infinite sequences in $\mathcal{X}$ with indices in $\mathbb{N}$.} $\mathcal{X}^{\infty}:=\mathcal{X}^\mathbb{N}$ to denote the set of infinite $\mathcal{X}$-sequences, and  $\mathcal{X}^{\#}:=\mathcal{X}^{*}\cup\mathcal{X}^{\infty}$
to denote the set of all (finite and infinite) sequences. For $x,y\in\mathcal{X}^\#$, we write $x\sqsubseteq y$ to denote that $x$ is a prefix of $y$.

For any sequence $x\in \mathcal{X}^{\#}$, we denote the (possibly infinite) length of $x$ as $l(x)$, and for every $1\leq i\leq l(x)$, we denote the $i$-th symbol of $x$ as $x_i$. For $1\leq i\leq j\leq l(x)$ we write $x_{i:j}$ to denote the subsequence $(x_i,\ldots,x_j)$. We use the notation $x_{\leq t}$ and $x_{< t}$ as a shorthand for $x_{1:t}$ and $x_{1:t-1}$, respectively. We also use the notation $x_{>t}$ and $x_{\geq t}$ as a shorthand for $x_{t+1:l(x)}$ and $x_{t:l(x)}$, respectively.

\begin{definition}
    (Cf., e.g., \cite{Hutter:24uaibook2})
    A \emph{semimeasure} on $\mathcal{X}^\infty$ is a mapping $\sigma:\mathcal{X}^*\to\mathbb{R}^+$ satisfying\footnote{It is worth noting that from the second condition, one can show by induction on $l(x)$ that $\sigma(x)\leq\sigma(\varepsilon)$ for all $x\in\mathcal{X}^*$. Combining this with the first property implies that $\sigma(x)\leq 1$ for all $x\in\mathcal{X}^*$, and hence $\sigma$ can be seen as a mapping $\mathcal{X}^*\to[0,1]$.}:
    \begin{enumerate}
        \item $\sigma(\varepsilon)\leq 1$, and
        \item $\displaystyle\sigma(x)\geq \sum_{u\in\mathcal{X}}\sigma(xu)\,,\forall x\in\mathcal{X}^*\,.$
    \end{enumerate}
    If the above two conditions hold with equality, we say that $\sigma$ is a measure.\footnote{Here we use the terminology that is standard in the AIXI literature (e.g., \cite{Hutter:24uaibook2}), which calls a measure only $\sigma$ which satisfies the mentioned equalities. A measure, according to this definition, induces a probability distribution on infinite sequences, as we shall see in a moment.}

    We call a semimeasure $\sigma$ initially-normalized\footnote{The term "initially-normalized" is not standard. We coin this term to distinguish these types of semimeasures because they have a nice intuitive interpretation, as we will see in \cref{rem:initially-normalized-semimeasures}. It is worth noting that our initially-normalized semimeasures corresponds to the definition of semimeasures that is adopted in \cite{HayThesis2007}.} if it satisfies $\sigma(\varepsilon)= 1$.

    A semimeasure for which $\sigma(\varepsilon)=0$ is said to be trivial. All semimeasures we consider in this paper are non-trivial.
\end{definition}

We emphasize that for an initially-normalized semimeasure $\sigma$, the value of $\sigma(x)$ for $x\in\mathcal{X}^*$ is not supposed to represent the probability of observing the substring $x$, but rather the probability of observing a string having $x$ as a prefix. We elaborate on this in the following remark, where we formally describe the intended interpretation of initially-normalized semimeasures.

\begin{remark}
    \label{rem:initially-normalized-semimeasures}
    An initially-normalized semimeasure $\sigma$ (i.e., one that satisfies $\sigma(\varepsilon)=1$) can be viewed as describing a random variable $X_\sigma$ taking values in $\mathcal{X}^{\#}=\mathcal{X}^{*}\cup\mathcal{X}^{\infty}$, which satisfies
    $$\mathbb{P}[X_\sigma=x]=\begin{cases}
    \sigma(x)-\sum_{u\in\mathcal{X}}\sigma(xu)\,,\quad&\text{if }x\in\mathcal{X}^*,\\
    \lim_{t\to\infty}\sigma(x_{\leq t})\,,\quad&\text{if }x\in\mathcal{X}^\infty\,,
    \end{cases}$$
    and
    $$\mathbb{P}[x\sqsubseteq X_\sigma]=\sigma(x)\,.$$
    In other words, $\sigma(x)$ represents the probability that $x$ is a prefix of $X_\sigma$. %\rajai{TODO: Consider elaborating on this}
    
    Under this interpretation of $\sigma$, one can see that for $x\in\mathcal{X}^*$,
    $\sigma(x)>\sum_{u\in\mathcal{X}}\sigma(xu)$ if and only if $\mathbb{P}[X_\sigma=x]>0$. Hence, one can deduce that $\sigma$ is a measure if and only if\footnote{$\mathbb{P}[X_\sigma\in\mathcal{X}^\infty]=1$ is equivalent to $\mathbb{P}[X_\sigma\in\mathcal{X}^*]=0$, i.e., $\mathbb{P}[X_\sigma=x]=0$ for all $x\in\mathcal{X}^*$. By the definition of the probability distribution of the random sequence, this is equivalent to having $\sigma(x)=\sum_{u\in\mathcal{X}}\sigma(xu)$ for all $x\in\mathcal{X}^*$, which would mean that $\sigma$ is a measure as we already know that $\sigma(\varepsilon)=1$.} $\mathbb{P}[X_\sigma\in\mathcal{X}^\infty]=1$, i.e., almost surely, the random sequence $X$ does not stop at any finite length.
\end{remark}

\begin{example}
    Consider a probabilistic monotone Turing machine (i.e., a Turing machine with access to an unlimited number of uniformly random coin flips) with a write-once output tape.  Running this machine gives rise to a (initially-normalized) semimeasure describing the state of the output tape after running until halting or forever if it does not halt. This semimeasure becomes a full measure if and only if the machine almost surely keeps writing symbols on the output tape, i.e., it neither halts nor gets to a situation where it loops forever without writing further symbols on the output tape.
\end{example}

\begin{definition}
\label{def:conditional-semimeasure}
If $\sigma$ is a semimeasure and $x\in\mathcal{X}^*$ is such that $\sigma(x)>0$, then we can define a conditional semimeasure $\sigma(\cdot|x)$ as follows:
$$\sigma(y|x)=\frac{\sigma(xy)}{\sigma(x)}\,,\quad\forall y\in\mathcal{X}^*\,.$$
If $\sigma$ is a measure, then $\sigma(\cdot|x)$ is a measure as well.

The conditional (semi)measure can be intuitively interpreted as the conditional (semi)probability, given that we have observed $x\in\mathcal{X}^*$ so far, that $x$ will be extended with $y\in\mathcal{X}^*$.
\end{definition}

It is possible to turn a semimeasure into a measure through normalization:

\begin{definition}
    For every semimeasure $\sigma$, define the Solomonoff normalization $\overline{\sigma}$ of $\sigma$ recursively as
    \begin{align*}
        \overline{\sigma}(\varepsilon)&=1\,,\\
        \overline{\sigma}(xa)&=\overline{\sigma}(x)\cdot\frac{\sigma(xa)}{\sum_{b\in\mathcal{X}}\sigma(xb)}\,,\quad\forall a\in\mathcal{X}\,,\forall x\in\mathcal{X}^*\,.
    \end{align*}
    If the denominator is zero, then define the fraction arbitrarily in such a way that we get $\sum_{a\in\mathcal{X}}\overline{\sigma}(xa)=\overline{\sigma}(x)$ (e.g., $\overline{\sigma}(xa)=\frac{1}{|\mathcal{X}|}\overline{\sigma}(x)$ if $\mathcal{X}$ is finite).

    It is not hard to see that for every $x\in\mathcal{X}^*$ and every $a\in\mathcal{X}$, we have
    $$\overline{\sigma}(a|x)=\frac{\sigma(xa)}{\sum_{b\in\mathcal{X}}\sigma(xb)}\,,\quad\forall a\in\mathcal{X}\,, \forall x\in\mathcal{X}^*\,.$$

    Furthermore, one can show by induction on $l(x)$ that $\overline{\sigma}(x)\geq\sigma(x)$ for all $x\in\mathcal{X}^*$.
\end{definition}

If $\sigma$ is an initially-normalized semimeasure (i.e., it satisfies $\sigma(\varepsilon)=1$), then we can construct a measure\footnote{Recall that $\tilde{\mathcal{X}}$ was defined as $\mathcal{X}\cup\{\perp\}$ in \eqref{eq:def-X-tilde}.} on $\tilde{\mathcal{X}}^\infty$ inspired by the interpretation in \cref{rem:initially-normalized-semimeasures}. But before describing the construction, it will be useful to introduce the notion of well-formed $\tilde{\mathcal{X}}$-sequences:
\begin{definition}
    We call a (finite or infinite) sequence $\tilde{x}\in\tilde{\mathcal{X}}^\#$ a well-formed sequence if it satisfies $$\forall t\in\mathbb{N}, (\tilde{x}_t=\perp)\Rightarrow (\forall t'\geq t, \tilde{x}_{t'}=\perp)\,.$$
    Equivalently,
    $$\forall t\in\mathbb{N}, (\tilde{x}_t\in\mathcal{X})\Rightarrow (\tilde{x}_{\leq t}\in\mathcal{X}^*)\,.$$
    In other words, a symbol of $\mathcal{X}$ can never appear after the symbol `$\perp$' in a well-formed sequence. We denote the set of well-formed (possibly infinite) sequences as $\tilde{\mathcal{X}}^{\#,wf}$. We write $\tilde{\mathcal{X}}^{*,wf}:=\tilde{\mathcal{X}}^{\#,wf}\cap\tilde{\mathcal{X}}^*$ (resp., $\tilde{\mathcal{X}}^{\infty,wf}:=\tilde{\mathcal{X}}^{\#,wf}\cap\tilde{\mathcal{X}}^\infty$) to denote the set of finite (resp. infinite) well-formed $\tilde{\mathcal{X}}$-sequences.

    There is a canonical bijection $x\mapsto\tilde{x}^{\infty,wf}$ between $\mathcal{X}^\#$ and $\tilde{\mathcal{X}}^{\infty,wf}$ defined as:
    \begin{itemize}
        \item If $x\in\mathcal{X}^\infty$ then $\tilde{x}^{\infty,wf}=x$.
        \item If $x\in\mathcal{X}^*$, then $\tilde{x}^{\infty,wf}_t=x_t$ for $t\leq l(x)$ and $\tilde{x}^{\infty,wf}_t=\,\perp$ for $t>l(x)$.
    \end{itemize}
\end{definition}

Now we are ready to introduce the canonical completion of an initially-normalized semimeasure:

\begin{definition}
    \label{def:canonical-completion}
    Let $\sigma$ be an initially-normalized semimeasure on $\mathcal{X}^\infty$, and let $X$ be the random sequence in $\mathcal{X}^\#$ induced by $\sigma$, as described in \cref{rem:initially-normalized-semimeasures}. We define the canonical completion $\tilde{\sigma}$ of $\sigma$ as the measure on well-formed $\tilde{\mathcal{X}}$-sequences induced by the probability distribution of the infinite sequence $\tilde{\mathcal{X}}^{\infty,wf}$ obtained by applying the canonical bijection $x\mapsto\tilde{x}^{\infty,wf}$ from $\mathcal{X}^\#$ to $\tilde{\mathcal{X}}^{\infty,wf}$ on $X$.
    
    Equivalently, we can define $\tilde{\sigma}$ as follows:
    \begin{itemize}
        \item For $x\in\mathcal{X}^*$, we let $\tilde{\sigma}(x)=\sigma(x)$.
        \item For $\tilde{x}\in\tilde{\mathcal{X}}^*\setminus\tilde{\mathcal{X}}^{*,wf}$, we let $\tilde{\sigma}(\tilde{x})=0$.
        \item For $\tilde{x}\in\tilde{\mathcal{X}}^{*,wf}\setminus \mathcal{X}^*$ (which necessarily means that $\tilde{x}$ is not the empty string and the last symbol of $\tilde{x}$ must be `$\perp$', i.e., $t:=l(\tilde{x})>0$ and $\tilde{x}_t=\perp$), we let
        $$\tilde{\sigma}(\tilde{x})=\sigma(x)-\sum_{x'\in\mathcal{X}}\sigma(xx')\,,$$
        where $x$ is the longest prefix of $\tilde{x}$ that lies in $\mathcal{X}^*$, i.e., $x=\argmax_{l(x)}\{x\in\mathcal{X}^*:x\sqsubseteq\tilde{x}\}$.
    \end{itemize}
\end{definition}

\begin{definition}
    Let $\sigma_1$ and $\sigma_2$ be two measures. For every $k\geq1$, we define the $k$-steps total-variation distance between $\sigma_1$ and $\sigma_2$ as:
    $$D_k(\sigma_1,\sigma_2)=\frac{1}{2}\sum_{x\in\mathcal{X}^k}|\sigma_1(x)-\sigma_2(x)|\,.$$
    It is easy to show that $D_{k+1}(\sigma_1,\sigma_2)\geq D_{k}(\sigma_1,\sigma_2)$. By taking $k\to\infty$, we get the total variation distance:$$D_\infty(\sigma_1,\sigma_2)=\sup_{k\geq 1}\frac{1}{2}\sum_{x\in\mathcal{X}^k}|\sigma_1(x)-\sigma_2(x)|\,.$$

    If $\sigma_1$ and $\sigma_2$ are initially-normalized semimeasures, we define the $D_k$ and $D_\infty$ distances based on their canonical completions as follows:
    $$D_k(\sigma_1,\sigma_2)=D_k(\tilde{\sigma}_1,\tilde{\sigma}_2)\,\text{ and }D_\infty(\sigma_1,\sigma_2)=D_\infty(\tilde{\sigma}_1,\tilde{\sigma}_2)\,.$$

    When we condition measures (or semimeasures) on some $x\in\mathcal{X}^*$, we use the shorthand notation $D_k(\sigma_1,\sigma_2|x)$ and $D_\infty(\sigma_1,\sigma_2|x)$ to denote $D_k(\sigma_1(\cdot|x),\sigma_2(\cdot|x))$ and $D_\infty(\sigma_1(\cdot|x),\sigma_2(\cdot|x))$, respectively.
\end{definition}

\begin{definition}
    \label{def:multiplicative-dominance}
    We say that a semimeasure $\sigma_1$ (multiplicatively) dominates a semimeasure $\sigma_2$, and write $\sigma_1\stackrel{\times}\geq\sigma_2$, if there exists $C>0$ such that $\sigma_1(x)\geq C\cdot\sigma_2(x)$ for all $x\in\mathcal{X}^*$.

    We say that $\mu$ and $\sigma$ are (multiplicatively) equivalent, and write $\mu\stackrel{\times}=\sigma$, if we have $\mu\stackrel{\times}\geq\sigma$ and $\mu\stackrel{\times}\leq\sigma$.
\end{definition}

\subsection{Universal Bayesian prediction theory.}
\label{app:preliminaries-universal-bayesian-prediction}

Assume that we are observing a random sequence $x\in\mathcal{X}^\infty$ whose distribution is described by a measure $\sigma$. We do not know the measure $\sigma$, but we know that it belongs to some class of measures $\calM$ which is countable. If we have observed the first $t$ symbols of $x$ (i.e., $x_{\leq t}$), can we make accurate predictions about the future $x_{>t}$?

We solve this problem with a Bayesian approach. We assume a prior belief distribution $w\in\Delta\calM$ and define the mixture measure
$$\xi^w=\sum_{\sigma\in\calM}w_\sigma \sigma\,.$$

Here, $w_\sigma$ represents the prior belief, before collecting any observation, that $\sigma$ is the true measure describing the distribution of $x\in\mathcal{X}^\infty$. After observing the first $t$ symbols of $x$, our posterior belief about $\sigma$ becomes
$$w(\sigma|x_{\leq t}):=\frac{w_\sigma\sigma(x_{\leq t})}{\sum_{\sigma'\in\calM}w_{\sigma'}\sigma'(x_{\leq t})}=\frac{w_\sigma\sigma(x_{\leq t})}{\xi^w(x_{\leq t})}\,.$$

Our posterior belief, given that we have already observed $x_{\leq t}$, that we will next observe $y\in\mathcal{X}^*$ (i.e., our posterior belief that $x_{t+1:t+l(y)}=y$) is given by
$$\sum_{\sigma\in\calM}w(\sigma|x_{\leq t})\sigma(y|x_{\leq t})=\sum_{\sigma\in\calM}\frac{w_\sigma\sigma(x_{\leq t})}{\xi^w(x_{\leq t})}\frac{\sigma(x_{\leq t}y)}{\sigma(x_{\leq t})}=\sum_{\sigma\in\calM}\frac{w_\sigma \sigma(x_{\leq t}y)}{\xi^w(x_{\leq t})}=\frac{\xi^w(x_{\leq t}y)}{\xi^w(x_{\leq t})}=\xi^w(y|x_{\leq t})\,.$$

Therefore, the conditional measure $\xi^w(\cdot|x_{\leq t})$ of the mixture $\xi^w$ can be conveniently used to describe our updated Bayesian predictions about the future.

Blackwell and Dubins showed that if $w$ satisfies $w_\sigma>0$ for all $\sigma\in\calM$, then $\xi^w(\cdot|x_{\leq t})$ almost surely converges to making correct predictions, as long as the ground-truth measure describing the distribution of $x$ belongs to $\calM$. This motivates the following definition:

\begin{definition}
    We say that $w$ is a universal prior probability (resp., semiprobability) distribution on $\calM$ if $w\in\Delta\calM$ (resp. $w\in\Delta'\calM$) and $w_\sigma>0$ for all $\sigma\in\calM$.
\end{definition}

\begin{theorem}[Merging of opinions \citet{blackwell1962merging}]
\label{thm:merging-opinions}
    For every universal prior $w\in\Delta\calM$ and every $\sigma\in\calM$, we have
    $$\lim_{t\to\infty}D_\infty(\xi^w,\sigma|x_{\leq t})=0\,,\quad\sigma(x)\text{-almost surely}\,.$$
\end{theorem}

It is possible to generalize the above theorem to semimeasures\footnote{Recall that we only consider non-trivial semimeasures in this paper.} by noticing the following facts:
\begin{itemize}
    \item Every semimeasure $\sigma$ on $\mathcal{X}^\infty$ is proportional to an initially-normalized semimeasure $\sigma_{\mathrm{i.n.}}$ (as we can divide $\sigma$ by $\sigma(\varepsilon)>0$).
    \item $\sigma$ and $\sigma_{\mathrm{i.n.}}$ have the same conditional semimeasures, i.e., $\sigma(\cdot|x)=\sigma_{\mathrm{i.n.}}(\cdot|x)$ for all $x\in\mathcal{X}^*$.
    \item A universal mixture of a collection of semimeasures on $\mathcal{X}^\infty$ can be seen as a universal mixture of the corresponding initially-normalized measures on $\mathcal{X}^\infty$, which can in turn be canonically represented as a universal mixture over the collection of measures on $\tilde{\mathcal{X}}^\infty$ through the canonical completion procedure of \cref{def:canonical-completion}.
\end{itemize}
By combining these facts, we can leverage the merging of opinions property for measures on $\tilde{\mathcal{X}}^\infty$ to prove a merging of opinions theorem for semimeasures on $\mathcal{X}^\infty$:

\begin{corollary}{\bf(Merging of opinions for semimeasures)}
    \label{cor:merging-opinions-semimeasures}
    Let $\calM$ be an arbitrary countable class of semimeasures. For every universal semiprobability prior $w\in\Delta'\calM$ and every semimeasure $\sigma\in\calM$, we have\footnote{When $\sigma$ is a semimeasure, then the "$\sigma(x)$-almost surely" notation can be understood in terms of the random sequence $x\in\mathcal{X}^\#$ induced by the corresponding initially-normalized semimeasure $\sigma_{\mathrm{i.n.}}$.}
    $$\lim_{t\to\infty}D_\infty(\xi^w,\sigma|x_{\leq t})=0\,,\quad\sigma(x)\text{-almost surely}\,,$$
    where $\xi^w$ is the universal semimeasure mixture defined as
    $$\xi^w=\sum_{\sigma\in\calM}w_\sigma \sigma\,.$$
\end{corollary}

As we have seen so far, if the real distribution $\sigma$ belongs to a class $\calM$, then Bayesian prediction based on a universal mixture on $\calM$ converges to making accurate predictions. A natural question that arises is: What shall we choose as the class of (semi)measures $\calM$ for our Bayesian mixture, so that the assumption "$\sigma$ belongs to the class $\calM$" is least restrictive? Solomonoff proposed considering the collection of all computable sequences. Solomonoff's approach can be extended to stochastic sequences/processes by considering computable\footnote{Here we use the term "computable" in the Turing-Church sense. This will be described in further details in the next section on computability theory.} (semi)measures.

Before describing Solomonoff induction theory, it will be useful to recall some notions from computability theory and algorithmic information theory.

\subsection{Useful notions from computability theory and algorithmic information theory.}
\label{app:preliminaries-computability-theory}

\begin{definition}
    Let $\mathcal{X}$ and $\mathcal{Y}$ be two countable sets for each of which we presume a fixed canonical encoding for their elements as finite binary strings. We say that a function $f:\mathcal{X}\to\mathcal{Y}$ is computable if there exists a Turing machine that computes $f$ using these encodings.
\end{definition}

For example, the mapping $q\mapsto q^2$ from $\mathbb{Q}$ to itself is computable (presuming standard encoding of rational numbers as binary strings).

\begin{definition}
    Let $\mathcal{X}$ be a countable set for which we presume a fixed canonical encoding for its elements as binary strings.
    A function $f:\mathcal{X}\to\mathbb{R}$ is said be lower semicomputable (l.s.c.) if there exists a computable function $\phi:\mathcal{X}\times\mathbb{N}\to\mathbb{Q}$ such that $\phi(x,n)\leq \phi(x,n+1)$ and $\lim_{n\to\infty}\phi(x,n)=f(x)$ for all $x\in\mathcal{X}$.\footnote{An equivalent definition is to say that $f:\mathcal{X}\to\mathbb{R}$ is lower semicomputable if and only if the set $L_f=\{(x,q)\in\mathcal{X}\times\mathbb{Q}: q<f(x)\}$ is recursively enumerable, i.e., there exists a Turing machine that computes a surjective mapping $\mathbb{N}\to L_f$.}
\end{definition}

In the following, we describe useful notions from algorithmic information theory (e.g., Kolmogorov complexity). For technical reasons, it will be useful to consider the following variants of Turing machines:

\begin{definition}
A monotone\footnote{It is worth noting that in some references (e.g., \cite{Hutter:24uaibook2}), such machines are referred to as prefix/monotone Turing machines, and then the names "prefix Turing machine" and "monotone Turing machine" are used in different contexts to emphasize different aspects of the same machine: In general, the name "prefix Turing machine" is typically used when we are mainly interested in the state of the machine when it halts, whereas the name "monotone Turing machine" is used when we are also interested in the evolution of its state as it runs (and potentially have an infinite computation). In our paper, we choose to use the unified name "monotone Turing machine".} Turing machine is a Turing machine $T$ equipped with:
\begin{enumerate}
    \item A unidirectional read-only input tape.
    \item A unidirectional write-only output tape.
    \item One or more bidirectional read/write working tapes initialized with zeros.
\end{enumerate}
All tapes are binary (i.e., no blank symbols).
\end{definition}

\begin{definition}
    We say that a monotone Turing machine $T$ halts on input $p\in\{0,1\}^*$ with output $x\in\{0,1\}^*$, and write $T(p)=x$, if by putting the string $p$ at the beginning of the input tape, the machine will halt in a state where $p$ is at the left of the input cursor and $x$ will be at the left of the output cursor. I.e., the machine will read $p$ from the input tape (and will not read any further bits from the input tape), and will write $x$ at the beginning of the output tape (and will not write any further bits to the output tape).

    The set $\mathcal{P}_T=\{p:T(p)\text{ halts}\}$ forms a prefix-free\footnote{A prefix-free set is a subset of $\{0,1\}^*$ such that no element is a prefix of any other element.} set. We call the strings in $\mathcal{P}_T$  $T$-self-delimiting programs\footnote{Here we can interpret $p\in\mathcal{P}_T$ as a "program that can run on the monotone Turing machine $T$". We do not assume here that these programs are necessarily able to "simulate all possible computational processes". This would be the case only when the machine $T$ is universal. We describe the concept of universal monotone Turing machines in \cref{def:universal-prefix-turing-machine}.}, or simply self-delimiting programs when the monotone Turing machine $T$ is understood from the context.
\end{definition}

\begin{definition}
    We say that a monotone Turing machine $T$ computes a string starting with $x\in\{0,1\}^*$ on input $p\in\{0,1\}^*$, and write\footnote{Notice that the $*$ symbol in the notation $T(p)=x*$ is not a superscript nor a subscript.} $T(p)=x*$ if by putting the string $p$ at the beginning of the input tape, the machine will write $x$ on the output tape in such a way that when the machine writes the last bit of $x$, the input head will be at position $l(p)$ (i.e., $p$ will be to the left of the input head). It does not matter what exists after $p$ on the input tape: The machine will always write $x$ at the beginning of the output tape as long as $p$ exists at the beginning of the input tape.\footnote{It is worth noting that for $p\in\{0,1\}^*$, there does not necessarily exist a unique $x\in\{0,1\}^*$ for which $T(p)=x*$, as the machine may read $p$ and then write multiple bits before reading any further bits from the input tape. However, for every $x_1,x_2\in\{0,1\}^*$ for which we have both $T(p)=x_1*$ and $T(p)=x_2*$, one of $x_1$ and $x_2$ must be a prefix of the other.}

    We say that a monotone Turing machine $T$ computes an infinite string $\omega\in\{0,1\}^\infty$ on input $p\in\{0,1\}^*$, and write $T(p)=\omega$ if by putting the string $p$ at the beginning of the input tape, the machine will read $p$ (and no further bits) and write $\omega$ on the output tape. Note that $T(p)=\omega$ for $\omega\in \{0,1\}^\infty$ means that the machine never halts.
\end{definition}

Next we turn to the definition of a universal monotone Turing machine. For this we need:
\begin{enumerate}
    \item A fixed canonical way of encoding monotone Turing machines as binary strings such that the set of encodings of all monotone Turing machines is decidable by a Turing machine (which we do not require to be a monotone one). In other words, if we denote the binary encoding of $T$ as $\langle T\rangle$, then there exists a Turing machine which computes a function $\{0,1\}^*\to\{0,1\}$ which outputs 1 if and only if its input is in the set $\{\langle T\rangle : T\text{ is a monotone Turing machine}\}$.
    \begin{itemize}
        \item Note that this induces a fixed canonical enumeration $T_1,\ldots$ of all monotone Turing machines as follows: We let $T_i$ be the $i$-th Turing machine according to the $\prec$ order defined as $T\prec T'$ if and only if $l(\langle T\rangle)< l(\langle T'\rangle)$ or $l(\langle T\rangle)= l(\langle T'\rangle)$ and $\langle T\rangle$ comes before $\langle T'\rangle$ according to the lexicographic order. It is worth noting that the enumeration mapping $n\mapsto\langle T_n\rangle$ is computable.
    \end{itemize}
    \item A computable injective prefix-free encoding $c:\mathbb{N}\to\{0,1\}^*$ of integers as binary strings in such a way that it is "computably invertible", i.e., there exists a computable mapping $c':\{0,1\}^*\to\mathbb{N}\cup\{\perp\}$ such that $c'(c(n))=n$ for all $n\in\mathbb{N}$ and $c'(x)=\perp$ for all $x\notin c(\mathbb{N})$. E.g., we can choose $c(n)$ to be the bitstring having $n$ ones followed by a zero.

    It is worth noting that for every computable function $f:\mathbb{N}\to\mathbb{N}$, the function $f_c:\{0,1\}^*\to\{0,1\}^*$ defined as $$f_c(x)=\begin{cases}c(f(c'(x))\quad&\text{if }x\in c(\mathbb{N})\,, \\\varepsilon\quad&\text{otherwise\,.}\end{cases}$$
    is computable.
\end{enumerate}

Given that we have such fixed canonical choices, we can define universal monotone Turing machines as follows:
\begin{definition}
    \label{def:universal-prefix-turing-machine}
    A universal monotone Turing machine $U$ is a monotone Turing machine which can simulate any other monotone Turing machine in the following sense: There exists a computable injective prefix free-encoding $c:\mathbb{N}\to\{0,1\}^*$, which is computably invertible, and an enumeration $T_1,\ldots,T_n,\ldots$ of monotone Turing machines for which $n\mapsto\langle T_n\rangle$ is computable, such that
    $$U(c(n)p)=T_n(p)\,,\quad\forall n\in\mathbb{N},\forall p\in\{0,1\}^*\,,$$
    and
    $$U(c(n)p)=x* \Leftrightarrow T_n(p)=x*\,,\quad \forall n\in\mathbb{N},\forall p,x\in\{0,1\}^*\,\,.$$
\end{definition}

\begin{theorem}
    Universal monotone Turing machines exist \citep{Hutter:24uaibook2}.
\end{theorem}

Universal monotone Turing machines are not unique. In the remainder of this paper, unless we state otherwise, we will assume that we have a canonical fixed universal monotone Turing machine $U$ that we will use as "the reference universal monotone Turing machine".

Now we are ready to define the (prefix) Kolmogorov complexity of a binary string:

\begin{definition}
    The (prefix) Kolmogorov complexity $K_T(x)$ of a binary possibly-infinite string $x\in\{0,1\}^\#$ relative to a monotone Turing machine $T$ is defined as\footnote{Note that for finite strings $x\in\{0,1\}^*$, the notation $T(p)=x$
    necessarily means that the machine halts on input $p$ after writing $x$. When we consider the possibility of continuing after writing $x$, we write $T(p)=x*$.}
    $$K_T(x):=\min\{l(p): p\in\{0,1\}^*,\,T(p)=x\}\,,$$
    with the convention that $K_T(x)=\infty$ if there is no $p\in\{0,1\}^*$ with $T(p)=x$.

    For the reference universal monotone Turing machine $U$, we drop the subscript and simply write $K(x):=K_U(x)$ and call it the (prefix) Kolmogorov complexity of $x$, or simply the $K$-complexity of $x$.

    We can extend the definition of Kolmogorov complexity to general mathematical objects assuming that we have a fixed canonical representation/encoding of these objects as binary strings, i.e., $K(o)=K(\langle o\rangle)$ where $\langle o\rangle$ is the canonical encoding of $o$ as a binary string.
\end{definition}

From \cref{def:universal-prefix-turing-machine} one can see that if $U$ is a universal monotone Turing machine and $T$ is an arbitrary monotone Turing machine which appears as the $n_T$-th machine in the canonical enumeration (i.e., $T_{n_T}=T$), then for all $x\in\{0,1\}^*$, we have
$$K_U(x)\leq K_T(x)+l(c(n_T))\,.$$

This implies that the prefix Kolmogorov complexity depends on the choice of the universal Turing machine only up to an additive constant:

\begin{theorem}
    The definition of the Kolmogorov complexity depends on the choice of the universal monotone machine only up to additive constants: If $U_1$ and $U_2$ are universal, then there exists $c>0$ such that $K_{U_1}(x)-c\leq K_{U_2}(x)\leq K_{U_1}(x)+c$ for all $x\in\{0,1\}^\#$  \citep{Hutter:24uaibook2}.
\end{theorem}

\subsection{Solomonoff induction.}
\label{app:preliminaries-solomonoff-induction}

\begin{definition}
    Let $\calM_{sol}$ be the class of all lower semicomputable semimeasures, and let $\{\sigma_1,\ldots,\}$ be a fixed canonical enumeration\footnote{A natural choice can be obtained using encodings of Turing machines $T$ which compute functions $\phi_T:\mathcal{X}^*\times\mathbb{N}\to\mathbb{Q}\cap[0,1]$. In this case, if $\langle T\rangle$ is the canonical representation/encoding of the Turing machine $T$ as an integer, then we let $\sigma_{\langle T\rangle}(x)=\lim_{n\to\infty} \max_{1\leq i\leq n}\phi_T(x,i)$. (Note that here we added the $\max_{1\leq i\leq n}$ part to recover the monotonicity in the $n\in\mathbb{N}$ argument, which is part of the definition of lower semicomputable functions.) If $n\in\mathbb{N}$ is not a valid encoding of such a Turing machine, then we let $\sigma_n(x)=0$.} of $\calM_{sol}$ (possibly with repetition), and let $w:\mathbb{N}\to(0,\infty)$ be a lower semicomputable function satisfying $\sum_{i}w_i\leq 1$. The universal mixture over $\calM_{sol}$ induced by the prior $w$ is defined as
    $$\xi^w(x):=\sum_{i\in\mathbb{N}}w_i\sigma_i(x)=\sum_{\sigma\in\calM_{sol}}w_\sigma\sigma(x)\,,$$
    where
    $$w_\sigma:=\sum_{\substack{i\in\mathbb{N}:\\\sigma_i=\sigma}}w_i\,.$$

    The Solomonoff universal prior is the one for which $w_i=2^{-K(i)}$ where $K(i)$ is the Kolmogorov complexity\footnote{It is worth noting that the Kolmogorov complexity is upper semicomputable and hence $i\mapsto 2^{-K(i)}$ is lower semicomputable.} of the binary representation of the integer $i$. In this case we simply write $\xi_U$ to denote the particular universal mixture distribution that arises from the Solomonoff universal prior.\footnote{The Solomonoff universal mixture $\xi_U$ is closely related to the following semimeasure, which is called the Solomonoff distribution $M$, and which was introduced in \cite{solomonoff-complexity-induction-1978}:
    $$M(x)=\sum_{\substack{p\in\{0,1\}^*:\\ U(p)=x*}}2^{-l(p)}\,,\forall x\in\mathcal{X}^*\,.$$
    The semimeasure $M$ can be interpreted as the probability distribution describing the state of the output tape of the universal monotone Turing machine $U$ when fed with uniformly random bits on the input tape (i.e., running a random program on a universal machine). It has been shown that $M$ and $\xi_U$ are multiplicatively equivalent: $M\stackrel{\times}=\xi_U$. Proofs of this fact can be found in \cite{zvonkin1970complexity, hutter2005universal, Hutter_11unipreq, lv19_kolmogorov}.
}
\end{definition}

A direct corollary of \cref{cor:merging-opinions-semimeasures} is that Solomonoff's universal prior can be used to obtain strong predictors:

\begin{theorem}
     % \rajai{Add REF} 
     If $\sigma$ is a computable measure describing the ground-truth probability distribution of an infinite sequence $x\sim\sigma$, then $$\lim_{t\to\infty}D_\infty(\sigma,\xi^U|x_{\leq t})=0\,,\quad \sigma(x)\text{-almost surely.}$$
    The above is also true if $\sigma$ is a lower semicomputable (semi)measure.
\end{theorem}

In other words, by collecting a sufficient number of observations from the true sequence (which we assume to be drawn from a computable measure), we can make accurate predictions about the future observations using the universal mixture $\xi^U$. 

\subsection{General Reinforcement Learning.}
\label{app:preliminaries-general-rl}

Let $\mathcal{A}$ and $\calE$ be two countable sets which we interpret respectively as the action space and the percept space in general reinforcement learning. Let $\TurnSet^* := \TurnSet^*$ be the set of all finite histories of interactions between the agent and the environment. We slightly abuse notation and write $\Hist=a_1e_1\ldots a_te_t$ to describe the first $t$ interactions in a history.
\footnote{Strictly speaking $\Hist$ should be $= (a_1, e_1)(a_2, e_2)\ldots(a_t, e_t)$.}

In \cref{sec:background}, we defined the space of percepts as $\calE=\mathcal{O}\times\mathcal{R}$, where $\mathcal{O}$ is the set of observations and $\mathcal{R}\subset[0,1]$ to be some finite set of rewards.

For a sequence $a\in\mathcal{A}^*$ of actions, and a sequence $e\in\calE^*$ of percepts, if $t\leq \min\{l(a),l(e)\}$, we write $\Hist$ to denote the interleaved sequence $a_1e_1\ldots a_t e_t$. Obviously, $\Hist\in\TurnSet^t\subset\TurnSet^*$.

\begin{definition}
    \label{def:environment}
    An environment-like chronological conditional (semi)measure $\nu$ is characterized by a mapping $f_\nu:\TurnSet^*\times\mathcal{A}\to\Delta'\calE$. 
    For $\Hist\in\TurnSet^*$, $a\in\calA$ and $e\in\calE$, we write $\nu(e|\Hist,a)$ as shorthand notation for $f_\nu(\Hist,a)(e)$ and interpret it as follows: Given a history $\Hist$ of interactions between the agent and the environment, and given that the agent took the action $a$ afterwards, then the conditional (semi)probability that the environment will subsequently produce the percept $e$ is $\nu(e|\Hist,a)$.
    
    We also use the notation
    $$\nu(e_{\leq t}\|a_{\leq t}):=\prod_{i\leq t}\nu(e_i|\HistM[i],a_i)\,,$$
    and interpret it as the conditional (semi)probability that the environment produces the percept sequence $e_{\leq t}$ given that the agent has made the action sequence $a_{\leq t}$. Notice how the structure of $\nu(\cdot\|\cdot)$ respects the chronological order between actions and percepts, and their causal structure.

    The environment-like chronological conditional (semi)measure $\nu$ is said to be lower semicomputable (resp. computable) if the function $\Hist,a,e\mapsto \nu(e|\Hist,a)$ from $\TurnSet^*\times\mathcal{A}\times\calE$ to $[0,1]$ is lower semicomputable (resp. computable).
\end{definition}

\begin{definition}
    \label{def:policy}
    A policy-like chronological conditional (semi)measure $\pi$ is characterized by a mapping $f_\pi:\TurnSet^*\to\Delta'\mathcal{A}$. 
    For $\Hist\in\TurnSet^*$ and $a\in\calA$, we write $\pi(a|\Hist)$ as a shorthand notation for $f_\pi(\Hist)(a)$ and interpret it as follows: Given a history $\Hist$ of interactions between the agent and the environment, then the conditional (semi)probability that the (policy $\pi$ of the) agent will take the action $a$ after $\Hist$ is $\pi(a|\Hist)$.
    
    We also use the notation
    $$\pi(a_{\leq t}\|e_{<t}):=\prod_{i\leq t}\pi(a_i|\HistM[i])\,,$$
    and interpret it as the conditional (semi)probability that the agent makes the action sequence $a_{\leq t}$ given that the environment produces the percept sequence $e_{<t}$. Notice again how the structure of $\pi(\cdot\|\cdot)$ respects the chronological order between actions and percepts, and their causal structure.

    The policy-like chronological conditional (semi)measure $\pi$ is said to be lower semicomputable (resp. computable) if the function $\Hist,a\mapsto \pi(a|\Hist)$ from $\TurnSet^*\times\mathcal{A}$ to $[0,1]$ is lower semicomputable (resp. computable).
\end{definition}

In the remainder of this paper, we write policy (resp. environment) to denote a policy-like (resp. environment-like) chronological conditional semimeasure\footnote{A policy that is a proper conditional semimeasure (i.e., not a conditional measure) can be interpreted as describing a situation where the agent fails to take an action at some point in time after which we assume that the agent ceases to exist (i.e., the agent "dies"). Similarly, an environment that is a proper semimeasure can be interpreted as describing a situation where the "world" is not guaranteed to exist for ever and may end, which would also mean that the agent "dies".}.

Now we can describe the interaction between an environment and a policy:

\begin{definition}
    Given an environment $\nu$ and a policy $\pi$, we can define the semimeasure $\nu^\pi$ on $\TurnSet^*$ describing the interaction between the agent and the environment as follows:
    \begin{equation}
        \label{eq:nupi}
        \nu^\pi(\Hist)=\pi(a_{\leq t}\|e_{<t})\nu(e_{\leq t}\| a_{\leq t})=\prod_{i\leq t}\pi(a_i|\HistM[i])\nu(e_i|\HistM[i],a_i),\quad\forall \Hist\in\TurnSet^*\,.
    \end{equation}

    We can similarly define the conditional semimeasure $\nu^\pi(\cdot|\Hist)$ for all $\Hist\in\TurnSet^*$ as\footnote{Note that in \eqref{eq:conditional-nupi}, we have $\HistM[i]:=\Hist\text{\ae}_{t+1:i-1}$.}
    \begin{equation}
        \label{eq:conditional-nupi}
        \nu^\pi(\text{\ae}_{t+1:t+t'}|\Hist)=\prod_{t<i\leq t+t'}\pi(a_i|\HistM[i])\nu(e_i|\HistM[i],a_i),\quad\forall \text{\ae}_{t+1:t+t'}\in\TurnSet^*\,.
    \end{equation}

    Notice that we have $\nu^\pi(\Hist[t']):=\nu^\pi(\Hist \text{\ae}_{t+1:t+t'})=\nu^\pi(\Hist)\nu^\pi(\text{\ae}_{t+1:t+t'}|\Hist)$ for all $\Hist,\text{\ae}_{t+1:t+t'}\in\TurnSet^*$.
\end{definition}

\begin{remark}
    \label{rem:pol-env-counterfactual-conditional}
    The definition of $\nu^\pi(\cdot|\Hist)$ using \eqref{eq:conditional-nupi} coincides with \cref{def:conditional-semimeasure} when $\nu^\pi(\Hist)>0$, i.e., we have $\nu^\pi(\text{\ae}_{t+1:t+t'}|\Hist)=\frac{\nu^\pi(\Hist \text{\ae}_{t+1:t+t'})}{\nu^\pi(\Hist)}$ whenever $\nu^\pi(\Hist)>0$. Here we are able to extend the definition to cases where $\nu^\pi(\Hist)=0$, and hence we can meaningfully analyze the "$\nu^\pi$-consequences" of counterfactual events that cannot occur under $\nu^\pi$: E.g., under this definition, we can speak in a well-defined way about the conditional expected reward that the agent gets in the next percept under the policy $\pi$ in the environment $\nu$ given that it has observed $\Hist$ even if $\nu^\pi(\Hist)=0$ (we can literally define this as $\sum_{a,e\in\mathcal{A}\times\calE}\nu^\pi(ae|\Hist)r(e)$ where $r(e)$ is the reward associated with the percept $e$.).
\end{remark}

\begin{definition}
    \label{def:value-function}
    The (normalized\footnote{In some references, e.g., \cite{sutton_reinforcement_2018}, the value function is defined without the "normalizing" multiplicative $(1-\gamma)$ term. This would make the value function take values in $[0,1/(1-\gamma)]$ since $\sum_{i\geq 0}\gamma^i=1/(1-\gamma)$. As it is convenient to have values in $[0,1]$, we normalize by the total weight $\sum_{i\geq 0}\gamma^i=1/(1-\gamma)$, which boils down to multiplying by $(1-\gamma)$. Note that the definition of the value function in \cite[Definition 6.6]{Hutter:24uaibook2} is also normalized, though the definition there is general and can handle arbitrary discounting functions that are not necessarily geometric.}) value function of the policy $\pi$ in the environment $\nu$ with discount factor $\gamma\in[0,1)$ is the function $V_{\nu^\pi}:\TurnSet^*\to[0,1]$ defined\footnote{It is worth noting that in some references, e.g., \cite[Definition 6.6]{Hutter:24uaibook2}, the normalized value function is defined as 
    \begin{equation}
        \label{eq:alternative-value-function}
        \lim_{m\to\infty}(1-\gamma)\sum_{\text{\ae}_{t:m}\in\TurnSet^{m-t+1}}\nu^\pi(\text{\ae}_{t:m}|\HistM)\sum_{\substack{i: t\leq i\leq m}}\gamma^{i-t}r(e_i)\,.
    \end{equation}
    For this alternative definition, any future trajectory which "corresponds" to a missing probability mass (due to the semimeasure definition) will not contribute to the value function. In other words, if we take the interpretation that missing probabilities in a semimeasure correspond to the probability of death of an agent, and if we have a trajectory for which the agent dies after two steps, then the rewards obtained in the next two steps will not contribute to the value function as defined in \eqref{eq:alternative-value-function}. In other words, the agent only values trajectories for which it lives forever. The two definitions of the value function in \eqref{eq:value-function-app} and \eqref{eq:alternative-value-function} differ only in the case of semimeasures.} as:
    \begin{equation}
        \label{eq:value-function-app}
        V_{\nu^\pi}(\HistM)=\lim_{m\to\infty}(1-\gamma)\sum_{\substack{i:t\leq i\leq m}}\gamma^{i-t}\sum_{\text{\ae}_{t:i}\in\TurnSet^{i-t+1}}r(e_i)\nu^\pi(\text{\ae}_{t:i}|\HistM)\,,
    \end{equation}
    where $r(e_i)$ is the reward associated with the percept $e_i$.

    The (normalized) action-value function (or $Q$-function) is the function $Q_{\nu^\pi}:\TurnSet^*\times\mathcal{A}\to[0,1]$ defined as:
    \begin{equation}
        \label{eq:q-value-function-app}
        \begin{aligned}
        Q_{\nu^\pi}(\HistM,a_t)&=\sum_{e_t\in\calE}\nu(e_t|\HistM, a_t)\left((1-\gamma)r(e_t)+\gamma V_{\nu^\pi}(\HistM a_te_t)\right)\\
        &=\lim_{m\to\infty}(1-\gamma)\sum_{\substack{i:t\leq i\leq m}}\gamma^{i-t}\sum_{(e_t,\text{\ae}_{t+1:i})\in\calE\times\TurnSet^{i-t}}r(e_i)\nu^\pi(e_t\text{\ae}_{t+1:i}|\HistM a_t)
        \,,
        \end{aligned}
    \end{equation}
    with the convention that $\text{\ae}_{t+1:t}=\varepsilon$ and $\TurnSet^0=\{\varepsilon\}$.

    One can see that for all $\HistM \in \TurnSet^*$, we have
    $$V_{\nu^\pi}(\HistM)=\sum_{a_t\in\mathcal{A}}\pi(a_t|\HistM)Q_{\nu^\pi}(\HistM,a_t)\,.$$
\end{definition}

\begin{definition}
    The optimal value function $V^*_{\nu}:\TurnSet^*\to[0,1]$ of an environment $\nu$ is defined as
    $$V^*_{\nu}(\HistM):=\sup_{\pi}V_{\nu^\pi}(\HistM)\,,\quad\forall \HistM\in\TurnSet^*\,,$$
    where the supremum is taken over all policies (which are not necessarily computable or lower semicomputable).

    We denote as $\pi^*_\nu$ any policy which is optimal for the environment $\nu$, i.e.,
    $$V_{\nu^{\pi^*_{\nu}}}(\HistM)=V^*_{\nu}(\HistM)=\sup_{\pi}V_{\nu^\pi}(\HistM)\,,\quad\forall \HistM\in\TurnSet^*\,,$$
     and we denote the set of optimal policies as $\Pi^*_\nu$.

    %\rajai{TODO: Consider adding a discussion about what is meant by agent and how it relates to the notion of a policy.}

    An agent that acts according to an optimal policy $\pi^*_\nu$ of the environment $\nu$ is denoted as $\mathrm{AI}\nu$.
\end{definition}

So far we have described agents which can act optimally in a known environment $\nu$. What if the agent is uncertain about the true environment? Assume that the agent only knows that the true environment belongs to some class of environments $\calMenv$ which we assume to be countable. Assume further that the agent has a prior Bayesian belief about the true environment: For every $\nu\in\calMenv$, let $w_\nu>0$ be the prior (semi)probability that the true environment is $\nu$. In this case, one can define the universal mixture environment $\xi^w_U$ such that
$$\xi^w_U=\sum_{\nu\in\calMenv}w_\nu \nu\,.$$

A direct corollary that follows from the merging of opinions results (\cref{thm:merging-opinions} and \cref{cor:merging-opinions-semimeasures}) is the following:

\begin{proposition}[on policy convergence]
    For any universal prior $w\in\Delta'\calMenv$, the universal mixture $\xi=\xi_U^w$ satisfies: For any policy $\pi$ and any environment $\nu\in\calMenv$, we have
        $$\lim_{t\to\infty}D_\infty(\xi^\pi,\nu^\pi|\Hist)=0\,,\quad \nu^\pi\text{-almost surely,}$$
        and
        $$\lim_{t\to\infty}|V_{\xi^\pi}(\Hist)-V_{\nu^\pi}(\Hist)|=0\,,\quad \nu^\pi\text{-almost surely.}$$
\end{proposition}

The universal Bayesian agent with respect to the class $\calMenv$ and the prior $w$ is the agent $\mathrm{AI}\xi_U^w$ (also denoted as $\mathrm{AI}\xi$) which acts optimally in the universal mixture environment $\xi^w_U$, i.e., it acts according to $\pi^*_{\xi^w_U}$.

The universal Bayesian agent $\mathrm{AI}\xi$ satisfies a few nice properties such as Pareto optimality and self-optimization \cite{bayes-self-optimization-pareto} %\rajai{TODO: We may want later to mention these properties and describe them explicitly in this section if it turns out to be helpful for the remainder of the paper.}

\subsection{AIXI}
\label{app:preliminaries-aixi}

By combining Solomonoff's induction theory with (universal Bayesian) general reinforcement learning, we get AIXI (introduced by Hutter in \cite{hutter-aixi}).

Hutter considered the class of all lower semicomputable chronological conditional semimeasures $\calM^{\mathrm{semi}}_{\mathrm{lsc}}$, and considered a universal mixture environment
$$\xi_U^w=\sum_{\nu\in \calM^{\mathrm{semi}}_{\mathrm{lsc}}}w_\nu \nu\,,$$
where $w\in \Delta'\calM^{\mathrm{semi}}_{\mathrm{lsc}}$ is some universal mixture which is lower semicomputable. A natural choice of the universal prior would be the one based on Kolmogorov complexity, just as in Solomonoff induction, i.e., we take $w_{\nu}=\sum_{i:M_i\text{ ``lower semi-computes'' }\nu}2^{-K(i)}$, where $M_i$ denotes the $i$-th Turing machine in a canonical enumeration of machines that "lower semicomputes"\footnote{More precisely, the machine $M_i$ computes a function $\phi_i:\TurnSet^*\times\mathcal{A}\times\mathcal{E}\times\mathbb{N}\to[0,1]\cap\mathbb{Q}$ such that $\phi_i(h,a,e,n+1)\geq \phi_i(h,a,e,n)$ for all $n\in\mathbb{N}$ and $\nu_i(e|h,a):=\lim_{n\to\infty}\phi_i(h,a,e,n)$ is an environment-like chronological conditional semimeasure. Furthermore, we assume that the canonical enumeration $M_1,\ldots$ covers all the environment-like chronological conditional semimeasures that can be described in this fashion.} environment-like chronological conditional semimeasures.

AIXI is the universal Bayesian agent $\mathrm{AI}\xi$ in the universal mixture $\xi=\xi_U^w$ w.r.t. the class of all lower semicomputable chronological conditional semimeasures $\calM^{\mathrm{semi}}_{\mathrm{lsc}}$.

AIXI is considered to be a theoretical gold standard for single-agent general reinforcement learning because its hypothesis class consists (in some arguable sense) of the most general environments\footnote{Assuming that the environment is computable is a very unrestrictive assumption.}. By virtue of being a universal Bayesian agent, AIXI satisfies the same nice optimality results that are satisfied by general universal Bayesian agents $\mathrm{AI}\xi$  such as pareto optimality and self-optimization. It was conjectured that AIXI satisfies additional optimality results such as strong asymptotic optimality, which means that for any computable environment, AIXI will almost-surely converge (on policy) to acting optimally in this ground-truth environment.

Unfortunately, it turns out that this is not the case in general as the choice of the universal prior $w$ can affect the optimality properties of AIXI (e.g., \cite{orseau-aixi-non-optimality,orseau-aixi-non-learnability,leike2015bad}): It is possible to choose a bad universal prior $w$ for which AIXI will not converge to acting optimally. A notable example is the "dogmatic prior" \citep{leike2015bad} for which the agent has a strong prior belief that the environment will eternally punish the agent (by giving it 0 rewards forever after) if it deviates from some arbitrary computable policy $\pi$.\footnote{More precisely, the agent gives a very high prior probability $w_\nu$ to the described environment.} In this case, if the value of $\pi$ in the ground-truth environment is not too small, one can show that AIXI will always act according to $\pi$ even if it is not optimal in the ground-truth environment.\footnote{It is worth noting that the Solomonoff prior (which is used to define the AIXI agent) is defined in terms of the Kolmogorov complexity, which in turn is based on the choice of some fixed reference universal monotone Turing machine $U$. One can choose $U$ so that the Solomonoff prior that is induced by $U$ is "dogmatic".}

These issues of asymptotic non-optimality of AIXI can be alleviated by adding good exploration components to it, such as Thompson sampling \citep{leike2016thompson}.

\subsection{Failure of AIXI to model embedded agency.}
\label{app:preliminaries-aixi-embedded-failure}

AIXI is the optimal Bayesian agent for the class of all lower semicomputable environments. AIXI itself is not lower semicomputable, so in some sense, AIXI is more powerful than the environments it considers. This is sufficient for the single-agent reinforcement learning (RL) setup that we have discussed in the previous section as we have considered that the agent and the environment are cleanly separated and communicate via well-defined channels of actions and observations, which is the typical assumption of single-agent RL.

However, in real life, agents are "embedded in their environments", and hence we have to consider the possibility that the environment may contain other agents that may be implementing a policy similar to them. In particular, if we have one agent implementing AIXI, then we have to consider the possibility that the environment (from the perspective of this agent) may also be implementing AIXI, which makes the environment not lower semicomputable, and hence the true environment would not belong to the class of environments that is considered by AIXI. Therefore, any theoretical analysis of Bayesian RL that is based on the assumption that the true environment belongs to the class of environments considered by the Bayesian agent would not apply.

\subsection{JAIXI as a way to formalize embedding failures in universal artificial intelligence.}
\label{app:preliminaries-jaixi}

AIXI is fundamentally a "decoupled Bayesian agent", i.e., it has prior beliefs only about the environment dynamics, and assumes that the choice of its policy is independent from the environment. This fails to capture situations where the environment may contain other agents that may be implementing similar policies (e.g., the psychological twin prisoner's dilemma).

One step forward towards capturing embeddedness is to consider an "embedded Bayesian agent" which considers that the policy of the agent and the environment may be dependent: Instead of having a prior over a hypothesis class of environments (as is the case for Bayesian dualistic agents), an embedded Bayesian agent has a prior over a hypothesis class that jointly describes the interaction of the agent and the environment (cf. \cref{sec:embedded-bayesian-agents}). Since the interaction of an agent and an environment can be modeled as a semimeasure on $\TurnSet^*$, we can say that an embedded Bayesian agent has a prior over a class of semimeasures on $\TurnSet^*$.

In computational terms, one may say that a decoupled Bayesian agent considers that the agent and the environment are described by two different programs whereas an embedded Bayesian agent considers that there is a single joint program that describes both, and it has a prior over the class of such programs that jointly describe the agent and the environment. One may say that the agent and the environment from its perspective are both parts of the same universe, and hence a program that describes the universe would jointly describe the agent and its environment.

In concurrent work, \cite{wyeth2025formalizing} introduced Joint AIXI (JAIXI), which follows the above embedded Bayesian agent approach and considers the class $\calM_{\mathrm{joint}}$ of all lower semicomputable semimeasures on $\TurnSet^*$. Let $w$ be a lower semicomputable universal prior over $\calM_{\mathrm{joint}}$ and let $\xi_{\mathrm{joint}}^{w}$ be the corresponding universal mixture. This induces an environment $\mathrm{env}(\xi_{\mathrm{joint}}^{w}):\TurnSet^*\times\mathcal{A}\to\mathcal{E}$ by considering the conditional semimeasure defined as:\footnote{In \eqref{eq:env_from_univ_embedded_mixture}, we gloss over a formal technicality which is not important for the discussion: The semimeasure $\xi_{\mathrm{joint}}^{w}$ is defined on $\TurnSet^*$, and hence $\xi_{\mathrm{joint}}^{w}(\Hist a)$ is strictly speaking not canonically defined. One may consider that $\mathcal{A}$ and $\mathcal{E}$ are subsets of the same set $\mathcal{S}$ and then consider the class of lower semicomputable semimeasures on $\mathcal{S}^*$ which would then contain both $\TurnSet^* \times \mathcal{A}$ and $\TurnSet^*$. This is the approach followed in \cite{wyeth2025formalizing}. Another possibility would be to simply define $\xi_{\mathrm{joint}}^{w}(\Hist a)=\sum_{e\in\mathcal{E}}\xi_{\mathrm{joint}}^{w}(\Hist ae)$.}
\begin{equation}
    \label{eq:env_from_univ_embedded_mixture}
    \mathrm{env}(\xi_{\mathrm{joint}}^{w})(e|\Hist,a):=\xi_{\mathrm{joint}}^{w}(e|\Hist a):=\frac{\xi_{\mathrm{joint}}^{w}(\Hist ae)}{\xi_{\mathrm{joint}}^{w}(\Hist a)}\,.
\end{equation}
While $\xi_{\mathrm{joint}}^{w}$ is lower semicomputable, the induced environment $\mathrm{env}(\xi_{\mathrm{joint}}^{w})$ is not lower semicomputable\footnote{Note that in general, the ratio of two lower semicomputable functions is not necessarily lower semicomputable.} \citep{leike2016nonparametric,leike2015computability}.

\cite{wyeth2025formalizing} define JAIXI as the optimal agent for the environment $\mathrm{env}(\xi_{\mathrm{joint}}^{w})$, i.e., the agent that implements $\pi^*_{\mathrm{env}(\xi_{\mathrm{joint}}^{w})}$. The authors did not aim to present JAIXI as a good solution for universal embedded intelligence, but rather use it to illustrate the embeddedness failures of this particular approach to universal artificial intelligence, such as failing to learn some adversarially chosen sequences.

The failure of JAIXI to be a good solution for embedded universal intelligence can also be seen from the fact that JAIXI is probably not lower semicomputable,\footnote{The results of \citet{leike2015aixicomputability} can probably be adapted to show that JAIXI is not lower semicomputable.} and if we make it interact with a lower semicomputable environment, we are not guaranteed to get a lower semicomputable semimeasure on $\TurnSet^*$ describing the interaction. In other words, making JAIXI interact with a lower semicomputable environment gives rise to an interaction which may not be in the hypothesis class of JAIXI and hence may not satisfy the grain-of-truth property.

Our work takes a complementary path. We begin by characterizing the behavior and predictions of embedded Bayesian agents under the assumption that they \emph{do} satisfy the grain-of-truth property. We show that this allows them to reason about `functional similarities'—the possibility that the environment contains agents similar to themselves—which in multi-agent scenarios leads to convergence towards a new family of solution concepts: \textit{embedded equilibria}. Subsequently, in Section~\ref{sec:mupi}, we directly solve the grain-of-truth problem by constructing a new model class and a corresponding joint mixture model, using the reflective oracle framework as well as our novel \textit{reflective universal inductor}, which allows us to define a universally intelligent embedded agent.

\section{Proofs}\label{app:proofs}
\subsection{Proof of \cref{prop:behavior-decoupled-beliefs}}\label{app:proof-prop-decoupledness}
We first generalize the proposition towards hypothesis classes $\calMuni$ that can also contain non-fully-supported universes, and then prove the general proposition, which directly implies the correctness of \cref{prop:behavior-decoupled-beliefs} as well. We remind the reader that the notation $\HistA \in \TurnSet^*$ stands for an arbitrary-length history.

When a universe $\Mvu(\HistA)$ is not fully-supported, it does not uniquely factorize into a policy $\pi$ and $\nu$, as the conditionals $\Mvu(a\mid \HistA)$ or $\Mvu(e\mid \HistA a)$ are undefined when $\Mvu(\HistA)=0$ or $\Mvu(\HistA a)=0$, respectively. Hence, different pairs of policy-environment can lead to the same universe distribution. This leads us to the following more general definition of decoupledness of $w(\Mvu)$, also compatible with $\calMuni$ containing non-fully-supported universes. 
\begin{definition}[Decoupled prior]\label{def:generalized-decoupled}
    A probability measure $w(\Mvu)$ over $\Mvu \in \calMuni$ is decoupled iff there exist  spaces $\calMpol$ and $\calMenv$ and probability measure $\tilde{w} \in \Delta (\calMpol \times \calMenv)$ for which the following holds:
    \begin{enumerate}
        \item[(i)] $w(\Mvu) = \sum_{\nu \in \calMenv}\sum_{\pi\in \calMpol} \delta(\nu^\pi = \Mvu) \tilde{w}(\pi,\nu) \quad \forall \Mvu \in \calMuni;$
        \item[(ii)] $\tilde{w}(\pi,\nu) = \tilde{w}(\pi)\tilde{w}(\nu) \quad \forall \pi \in \calMpol, \nu \in \calMenv$ ,
    \end{enumerate}
    with $\delta$ the indicator function, and $\tilde{w}(\Mvp):=\sum_{\Mve\in\calMenv}\tilde{w}(\Mvp,\Mve)$ and $\tilde{w}(\Mve):=\sum_{\Mvp\in\calMpol}\tilde{w}(\Mvp,\Mve)$ are the marginals of $\tilde{w}$ over $\calMpol$ and $\calMenv$, respectively. It is worth noting that the above "factorization" of a decoupled prior $w\in \calMuni$ is not necessarily unique, i.e., there can be more than one such $\tilde{w}$ that yields $w$.
\end{definition}
When $\calMuni$ only contains fully supported universes, this definition simplifies to the notion of decoupledness we used in \cref{sec:emb-bay-agents-structural-similarities}, as each fully supported universe leads to a unique pair $(\pi,\nu)$.
Now we are ready to state and prove the generalized proposition. 
\begin{proposition}\label{prop:behavior-decoupled-beliefs-generalized}
    Consider a hypothesis class $\calMuni$ and corresponding probability measure $w \in \Delta \calMuni$. If the prior beliefs $w$ are decoupled (cf. \cref{def:generalized-decoupled}) with respect to the classes $\calMpol$ and $\calMenv$, with corresponding decoupled probability measure $\tilde{w} \in \Delta (\calMpol \times \calMenv)$, we have that the conditionals of the mixture universe $\Mmu(\HistA):= \sum_{\Mvu \in \calMuni} w(\Mvu)\Mvu(\HistA)$ are equal to 
    \begin{align*}
        \Mmu(a_t \mid \HistM) &= \zeta(a_t \mid \HistM), \quad \Mmu(e_t \mid \HistM a_t) = \xi(e_t \mid \HistM a_t) \\
        \zeta(a_t \mid \HistM)&:= \sum_{\pi \in \calMpol}\tilde{w}_{\textrm{pol}}(\pi \mid \Turn_{<t-1}a_{t-1})\pi(a_t\mid \HistM), \quad \tilde{w}_{\textrm{pol}}(\pi \mid \HistM a_{t}):= \tilde{w}_{\textrm{pol}}(\pi \mid \Turn_{<t-1}a_{t-1})\frac{\pi(a_t\mid \HistM )}{\zeta(a_t \mid \HistM )} \\
        \xi(e_t \mid \HistM a_t)&:= \sum_{\nu \in \calMenv}\tilde{w}_{\textrm{env}}(\nu \mid \HistM )\nu(e_t\mid \HistM a_t), \quad \tilde{w}_{\textrm{env}}(\nu \mid \Hist):= \tilde{w}_{\textrm{env}}(\nu \mid \HistM )\frac{\nu(e_t\mid \HistM a_t)}{\xi(e_t \mid \HistM a_t)}
    \end{align*}
    with $\tilde{w}_{\textrm{pol}}(\pi \mid \varepsilon):=\tilde{w}(\pi)$ and $\tilde{w}_{\textrm{env}}(\nu \mid \varepsilon):=\tilde{w}(\nu)$. Hence, $\Mmu$ uses decoupled posterior beliefs $\tilde{w}_{\textrm{pol}}(\pi \mid \Hist a_{t+1})$ and $\tilde{w}_{\textrm{env}}(\nu \mid \Hist)$. As a result, we have that
    \begin{enumerate}
        \item[(i)] An embedded Bayes-optimal agent using the decoupled beliefs $w(\Mvu)$ to construct its mixture universe model $\Mmu$ (cf. \eqref{eqn:embedded-bayes-mixture}) and implementing an embedded best response w.r.t. $\Mmu$ (cf. \eqref{eqn:embedded-br}) is equivalent to a decoupled Bayesian agent with mixture environment $\xi$ defined above, and implementing a decoupled best response w.r.t. $\xi$ (cf. \eqref{eqn:decoupled-br}).
        \item[(ii)] A $k$-step planner embedded Bayesian agent (cf. \cref{def:k-step-eba}) using the decoupled beliefs $w(\Mvu)$ to construct its mixture universe model $\Mmu$ is equivalent to a $k$-step planner decoupled Bayesian agent with mixture environment $\xi$ and mixture policy $\zeta$ as defined above, and implementing a $k$-step planner policy
        $$a_t \in \argmax_a Q_{\xi^\zeta}^k(\HistM ,a),$$
        with $Q_{\xi^\zeta}^k$ as defined in \eqref{eqn:k-step-qval}.
    \end{enumerate}
\end{proposition}
\begin{proof}
    For any scalar function $F:\calMuni \to \bbR$,
    \begin{align*}
        \sum_{\Mvu \in \calMuni} w(\Mvu)F(\Mvu) &\stackrel{(a)}{=} \sum_{\Mvu\in \calMuni} \sum_{\pi \in \calMpol}\sum_{\nu \in \calMenv} \tilde{w}(\nu)\tilde{w}(\pi) \delta(\nu^\pi = \Mvu) F(\Mvu) \\
        &= \sum_{\pi \in \calMpol}\sum_{\nu \in \calMenv} \tilde{w}(\nu)\tilde{w}(\pi) \sum_{\Mvu \in \calMuni} \delta(\nu^\pi = \Mvu) F(\Mvu) \\
        &=\sum_{\pi \in \calMpol}\sum_{\nu \in \calMenv} \tilde{w}(\nu)\tilde{w}(\pi)  F(\nu^\pi)\,,
    \end{align*}
    where $(a)$ follows directly from \cref{def:generalized-decoupled}. As a result, the mixture universe $\Mmu(\HistA)$ can be rewritten as
    \begin{align*}
        \Mmu(\HistM ) &= \sum_{\pi \in \calMpol}\sum_{\nu \in \calMenv} \tilde{w}(\nu)\tilde{w}(\pi) \nu^\pi(\HistM ) \\
        &=\sum_{\pi \in \calMpol}\tilde{w}(\pi)\pi(a_{<t} \mid \mid e_{<t-1}) \sum_{\nu \in \calMenv} \tilde{w}(\nu) \nu(e_{<t} \mid \mid a_{<t})\,, ~~\text{with} \\
        \pi(a_{<t} \mid \mid e_{<t-1})&:= \prod_{k=1}^{t-1} \pi(a_k \mid \HistM[k]), \quad \nu(e_{<t} \mid \mid a_{<t}):= \prod_{k=1}^{t-1} \nu(e_k \mid \HistM[k]a_k)\,.
    \end{align*}
    Now let us introduce the following notation
    $$\zeta(a_{<t} \mid \mid e_{<t-1}):=\sum_{\pi \in \calMpol}\tilde{w}(\pi)\pi(a_{<t} \mid \mid e_{<t-1}), \quad \xi(e_{<t} \mid \mid a_{<t}):=\sum_{\nu \in \calMenv} \tilde{w}(\nu) \nu(e_{<t} \mid \mid a_{<t})\,.$$
    Then we have that $\Mmu(\HistM ) = \zeta(a_{<t} \mid \mid e_{<t-1})\xi(e_{<t} \mid \mid a_{<t})$, and the conditionals of $\Mmu$ can be rewritten as
    \begin{align*}
        \Mmu(a_t \mid \HistM ) &= \sum_{\Mvu \in \calMuni}w(\Mvu)\frac{\Mvu(\HistM )}{\Mmu(\HistM )}\Mvu(a_t\mid \HistM )\\
        &=\sum_{\pi \in \calMpol}\sum_{\nu \in \calMenv} \tilde{w}(\nu)\tilde{w}(\pi) \frac{\nu^\pi(\HistM )}{\zeta(a_{<t} \mid \mid e_{<t-1})\xi(e_{<t} \mid \mid a_{<t})}\pi(a_t\mid \HistM )\\
        &=\sum_{\pi \in \calMpol}\sum_{\nu \in \calMenv} \tilde{w}(\nu)\tilde{w}(\pi) \frac{\pi(a_{<t} \mid \mid e_{<t-1})\nu(e_{<t} \mid \mid a_{<t})}{\zeta(a_{<t} \mid \mid e_{<t-1})\xi(e_{<t} \mid \mid a_{<t})}\pi(a_t\mid \HistM )\\
        &=\sum_{\nu \in \calMenv} \tilde{w}(\nu) \frac{\nu(e_{<t} \mid \mid a_{<t})}{\xi(e_{<t} \mid \mid a_{<t})}\sum_{\pi \in \calMpol}\tilde{w}(\pi) \frac{\pi(a_{<t} \mid \mid e_{<t-1})}{\zeta(a_{<t} \mid \mid e_{<t-1})}\pi(a_t\mid \HistM )\\
        &= \sum_{\pi \in \calMpol}\tilde{w}(\pi) \frac{\pi(a_{<t} \mid \mid e_{<t-1})}{\zeta(a_{<t} \mid \mid e_{<t-1})}\pi(a_t\mid \HistM ) \\
        &= \zeta(a_t \mid \HistM )\,,
    \end{align*}

    where in the last step we used $\zeta(a_t \mid \HistM )$ as defined in the proposition statement, and the fact that $\tilde{w}_{\textrm{pol}}(\pi \mid \Turn_{<t-1}a_{t-1}) = \tilde{w}(\pi) \frac{\pi(a_{<t} \mid \mid e_{<t-1})}{\zeta(a_{<t} \mid \mid e_{<t-1})}$ which can easily be verified by induction. Using similar reasoning, we arrive at 
    \begin{align*}
        \Mmu(e_t \mid \HistM  a_{t}) = \xi(e_t \mid \HistM a_t),
    \end{align*}
    thereby concluding the proof of the first part of the proposition. Statement (i) follows directly upon observing that $\Mmu^\pi$ from \eqref{eqn:embedded-br-factorization} is equal to $\xi^\pi$, and hence the decoupled best response (cf. \eqref{eqn:decoupled-br}) and embedded best response (cf. \eqref{eqn:embedded-br}) are equivalent. Statement (ii) follows directly from observing that $\Mmu = \xi^\zeta$.

\end{proof}

\subsection{Proof of \cref{thrm:convergence-to-sde}}\label{app:proof-convergence-to-sde}
(i) As $\Mmu^i$ satisfies the grain-of-uncertainty assumption, its conditionals are always well-defined and hence its conditional completion is unique, and as a result there exists an embedded best response w.r.t. $\Mmu^i$ that is equal to $\pi^i$. (ii) From \cref{theorem:convergence-mixture-universe}, it follows from the grain-of-truth property that the tail mixture universes $\Mmu^i_{\RghM}(\bara_{t:t'})$ converge to the tail ground-truth distribution $\MultiAgentMvp_{\RghM}(\bara_{t:t'})$ as $t\to\infty$ $\MultiAgentMvp$-almost-surely. It follows that for each $\epsilon > 0$, there exists a finite time $T(\epsilon)$ such that for all $t \geq T(\epsilon)$, with probability greater than $1-\epsilon$, we have that $D_\infty(\Mmu^i_{\RghM}, \MultiAgentMvp_{\RghM}) \leq \epsilon$, thereby concluding the proof.

\subsection{Proof of \cref{prop:decoupled-beliefs-se-ne}}\label{app:prop-decoupled-beliefs-se-ne}
We start by restating the definition of the subjective Nash equilibrium (\cref{def:subj-nash-eq}, now tailored towards repeated games. Note that \citet{kalai1993subjective} call this repeated-games version of the subjective Nash equilibrium a \textit{subjective equilibrium} in short. %We remind the reader that $h\in \TurnSet^*$ indicates and arbitrary-length history.
\begin{definition}[Subjective equilibrium \citep{kalai1993subjective}]\label{def:subj-eq-rep-games}
    A set of policies $(\pi^i)_{i=1}^N$ and agent models $\{\zeta^i_j\}_{i\in N, j\in N}$ is a subjective equilibrium in the repeated game setting if for each player $i$ it holds that 
    \begin{enumerate}
        \item[(i)] $\zeta^i_i = \pi^i$,
        \item[(ii)] $\pi^i$ is a best response w.r.t. $\Mmp^i_{-i}$, where $\Mmp^i_{-i}(a^{-i} \mid \RghM):=\prod_{j\neq i}\Mmp^i_j(a^{j} \mid \RghM)$, and
        \item[(iii)] $\zeta^i(\bara_{<t}) = \MultiAgentMvp(\bara_{<t})~~\forall \bara_{<t} \in \bar{\calA}^*$ with 
        $$\zeta^i(\bara_{<t}):= \prod_{k=1}^{t-1}\prod_{j\in N} \zeta^i_j(a_k^j \mid \RghM[k] )\,,$$
        $$\MultiAgentMvp(\bara_{<t}):= \prod_{k=1}^{t-1}\prod_{j\in N} \Mgp^j(a_k^j \mid \RghM[k] )\,.$$
        
    \end{enumerate}
\end{definition}
Here, $\Mmp^i_j$ indicates player $i$'s model of player $j$. 

We first show that the set of policies $\{\pi_D^i\}$ that form a subjective embedded equilibrium with decoupled completed conditionals also form a subjective equilibrium according to the above definition. 

As we assume that the completed conditionals are decoupled, i.e., 
$$\Mmu^i(a^{-i} \mid \RghM, a^i) = \prod_{j\neq i}\Mmu^i(a^j\mid \RghM)\,,$$
this leads to a predictive model that can be fully factorized with the marginals: 
$$\Mmu^i(\bara \mid \RghM) = \prod_{j\in N}\Mmu^i(a^j \mid \RghM)\,.$$
Hence, each $\Mmu^i(a^j \mid \RghM)$ can be interpreted as player $i$'s model of player $j$, i.e., $\Mmp^i_j$ in the terminology of the subjective equilibria. To compute an embedded best response following \eqref{eqn:embedded-br}, we need to insert a $\pi^i$ into $\Mmu^i$ following \eqref{eqn:embedded-br-factorization}:
$$\big(\Mmu^{i}\big)^{\pi^i}(\RghM):= \prod_{k=1}^{t-1}\pi^i(a^i_k \mid \RghM[k] )\prod_{j\neq i}\Mmu^i(a^j_k\mid \RghM[k]).$$
Computing an embedded best response w.r.t. the above distribution is equivalent to computing a best response against $\Mmp^i_{-i}$, if $\Mmp^i_{-i}(a^{-i} \mid \RghM)=\prod_{j\neq i}\Mmu^i(a^j\mid \RghM)$ which is true in our setup. Hence, taking $\Mmp^i_j(a^j \mid \RghM) = \Mmu^i(a^j \mid \RghM)$ for $i\neq j$ and $\Mmp^i_i = \pi^i$, combined with best-response policies $\{\pi^i_S\}_{i\in N}$ w.r.t. $\Mmp^i_{-i}$ leads to a subjective equilibrium according to \cref{def:subj-eq-rep-games}, with $\{\pi^i_S\}_{i\in N}$ equal to the embedded best responses $\{\pi^i_D\}_{i\in N}$ from the subjective embedded equilibrium, thereby concluding the first part of the proof. 

Finally, we can use proposition 1 of \citet{kalai1993subjective} stating that for each subjective equilibrium $\{\pi^i_S\}_{i\in N}$, there exists a Nash equilibrium $\{\pi^i_A\}_{i\in N}$ for which $\bar{\pi}_S(\RghM) = \bar{\pi}_A(\RghM)$, thereby concluding the proof.

\subsection{Proof of \cref{cor:convergence-to-eps-nash}}\label{app:proof-convergence-to-eps-nash}
The percepts $e_t^i$ of agent $i$ is equal to the actions $a^{-i}_t$ of the other agents, in our setup of repeated games. Using the proof technique of \cref{prop:behavior-decoupled-beliefs}, it is easy to show that the self-model $\Mmu^i(a^i\mid \RghM)$ and environment model $\Mmu^i(e^i \mid \RghM a^i)$ are equivalent to independent agent models with independent beliefs:
\begin{align*}
    \Mmu^i(a^i \mid \RghM) &= \zeta^i_i(a^i \mid \RghM), \quad \Mmu^i(a^{-i} \mid \RghM a^i) = \prod_{j\neq i} \zeta^i_j(a^j \mid \RghM) \\
    \zeta^i_j(a^j \mid \RghM) &:= \sum_{\pi^j \in \calMpol^j} w^i_j(\pi^j \mid \RghM) \pi^j(a^j \mid \RghM), \\
    w^i_j(\pi^j \mid \Rgh)&:= w^i_j(\pi^j \mid \RghM)\frac{\pi^j(a^j_{t} \mid \RghM)}{\zeta^i_j(a^j_t \mid \RghM)}, ~~ w_j^i(\pi^j \mid \varepsilon):= \tilde{w}^i(\pi^j).
\end{align*}
Using \cref{thrm:convergence-to-sde}, we have that for each $\epsilon>0$, there exists a time $T(\epsilon)$ such that for all $t\geq T(\epsilon)$, with probability at least $1-\eps$, the joint policy $(\pi^i)_{i=1}^N$ and subjective mixture models $\{\Mmu^i\}_{i\in N}$ are an $\epsilon$-subjective embedded equilibrium.
As we have independent agent models, we can use the same proof technique as in \cref{prop:decoupled-beliefs-se-ne} to show that in this case, an $\epsilon$-subjective embedded equilibrium corresponds to an $\epsilon$-subjective Nash equilibrium in repeated games, with an $\epsilon$-subjective Nash equilibrium defined as in \cref{def:subj-eq-rep-games} with the third condition replaced by $\zeta^i(\bara_{<t})$ is $\epsilon$-close to $\MultiAgentMvp(\bara_{<t})$ in total variation distance.
Now we can directly use Theorem 1 of \citet{kalai1993subjective} to conclude our proof: 
\begin{theorem}[Theorem 1 of \citet{kalai1993subjective}]
    In infinitely repeated games, for every $\epsilon>0$, there is $\epsilon' > 0$ such that for all $\epsilon''\leq \epsilon'$, if $\bar{\pi}$ is an $\epsilon''$-subjective Nash equilibrium, then there exists a joint policy $\MultiAgentMvp'$ such that 
    \begin{itemize}
        \item[(i)] $\MultiAgentMvp(\RghM)$ is $\epsilon$-close to $\MultiAgentMvp'(\RghM)$;
        \item[(ii)] $\MultiAgentMvp'$ is an $\epsilon$-Nash equilibrium.
    \end{itemize}
\end{theorem}

\subsection{Proof of \cref{prop:diff-sde-ode}}\label{app:proof-diff-sde-ode}
\begin{proof}
The proof consists of two parts. First, we show that every Embedded Equilibrium (EE) is a Subjective Embedded Equilibrium (SEE). Second, we provide a counterexample to show that the converse is not true.
\subsubsection*{Part 1: Every EE is an SEE}
We do the proof for the general case of MAGRL, which also holds for the repeated games setting.
Let a set of policies $(\pi^i)_{i=1}^N$ and a single dependency distribution $q$ constitute an Embedded Equilibrium (EE), using $q$ conditionally complete the ground-truth universe $\Mgu$:
\begin{align*}
        \Mgu_i(e^{i} \mid \Turn_{<t}^i, a^i) &= \Mgu_i(e^i \mid \Turn_{<t}^i, a^i) &&\text{if}~~ \Mgu_i(\Turn_{<t}^i, a^i)>0\,, \\
        \Mgu_i(e^{i} \mid \Turn_{<t}^i, a^i) &= q(e^i \mid \Turn_{<t}^i, a^i) &&\text{otherwise}\,.
    \end{align*}
We can construct a corresponding set of subjective predictive distributions $\{\Mmu^i\}_{i\in N}$ by setting, for each agent $i \in N$, its subjective model and its conditional completion equal to the conditionally completed $\Mgu$ by $q$.
The best response condition, and the conditionally completed $\Mgu$ of the EE directly imply the two conditions for a Subjective Embedded Equilibrium (SEE) (Subjective Best Response and Uncontradicted Beliefs). It therefore immediately follows from the definitions that every EE is also an SEE.

\subsubsection*{Part 2: An SEE that is not an EE}
We construct a counterexample using a two-player, single-shot normal-form game, as this also serves as a counterexample for the more general MAGRL setting. Let the players be $i \in \{1, 2\}$, with action spaces $\mathcal{A}^1 = \mathcal{A}^2 = \{A, B, C\}$. The payoff matrix $(r^1, r^2)$ is given by
\[
\begin{array}{c|ccc}
 & A & B & C \\
\hline
A & (2,2) & (0,7) & (0,7) \\
B & (7,0) & (6,1) & (1,6) \\
C & (7,0) & (1,6) & (6,1) \\
\end{array}
\]
where the rows correspond to the actions of the first player and the columns correspond to the actions of the second player.

Consider the deterministic joint policy $\bar{\pi}$ where both agents play action $A$, i.e., $\pi^1(A) = 1$ and $\pi^2(A) = 1$. The resulting distribution over histories is a point mass on the outcome $(A,A)$, so $\bar{\pi}(A,A) = 1$.

\textbf{Step A: Show this is a Subjective Embedded Equilibrium (SEE).}
Consider the following subjective beliefs for off-path actions:
\begin{itemize}
    \item For Agent 1: $\Mmu^1(a^2=C \mid a^1=B) = 1$ and $\Mmu^1(a^2=B \mid a^1=C) = 1$.
    \item For Agent 2: $\Mmu^2(a^1=B \mid a^2=B) = 1$ and $\Mmu^2(a^1=C \mid a^2=C) = 1$.
\end{itemize}
With these beliefs, each agent calculates their expected payoff for deviating to B or C as 1, which is less than the equilibrium payoff of 2 from playing A. The beliefs are uncontradicted because the play-path is always $(A,A)$, satisfying $\Mmu^i(A,A) = \bar{\pi}(A,A) = 1$. Thus, this constitutes a valid SEE.

\textbf{Step B: Show this is not an Embedded Equilibrium (EE).}
We show by contradiction that there is no single dependency distribution $q$ which can make this an EE. Assume such an EE exists, supported by a conditional completion of the ground-truth universe $\Mgu$ by dependency distribution $q$. Each conditional completion can be derived from a limit of a sequence $(p_r)_r$ of joint distributions satisfying the grain-of-uncertainty property. Let us take such sequence $(p_r)_r$ to represent the conditional completion of $\Mgu$. The best response condition implies that $Q^i(A) \ge Q^i(B)$ and $Q^i(A) \ge Q^i(C)$. Since
$$2=Q^1(A) \ge Q^1(B)=\expect{q(a^2 \mid a^1=B)}{r^1(B,a^2)}=\lim_{r\to\infty}\expect{p_r(a^2 \mid a^1=B)}{r^1(B,a^2)}\,,$$
there exists $r$ large enough so that
\begin{align*}
    \expect{p_r(a^2 \mid a^1=B)}{r^1(B,a^2)} &= \frac{7p_r(B,A) + 6p_r(B,B) + p_r(B,C)}{p_r(B,A) + p_r(B,B) + p_r(B,C)} \leq Q^i(A)+1=3\,.
\end{align*}
Similarly, we can show that for $r$ large enough, we have
\begin{align*}
    \expect{p_r(a^2 \mid a^1=C)}{r^1(C,a^2)} &= \frac{7p_r(C,A) + p_r(C,B) + 6p_r(C,C)}{p_r(C,A) + p_r(C,B) + p_r(C,C)} \leq Q^i(A)+1=3\,, \\
    \expect{p_r(a^1 \mid a^2=B)}{r^2(a^1,B)} &= \frac{7p_r(A,B) + p_r(B,B) + 6p_r(C,B)}{p_r(A,B) + p_r(B,B) + p_r(C,B)} \leq Q^i(A)+1=3\,, \\
    \expect{p_r(a^1 \mid a^2=C)}{r^2(a^1,C)} &= \frac{7p_r(A,C) + 6p_r(B,C) + p_r(C,C)}{p_r(A,C) + p_r(B,C) + p_r(C,C)} \leq Q^i(A)+1=3\,.
\end{align*}

We can rewrite the above four inequalities as 
\begin{align*}
    4p_r(B,A) + 3p_r(B,B) &\leq  2p_r(B,C)\,,  \\
    4p_r(C,A) + 3p_r(C,C) &\leq  2p_r(C,B)\,,  \\
    4p_r(A,B) + 3p_r(C,B) &\leq  2p_r(B,B)\,,  \\
    4p_r(A,C) + 3p_r(B,C) &\leq  2p_r(C,C)\,,  \\
\end{align*}
which can be simultaneously satisfied only when $$p_r(B,A)=p_r(C,A)=p_r(A,B)=p_r(A,C)=p_r(B,B) = p_r(C,C) = p_r(C,B) = p_r(B,C)=0\,.$$
But this contradicts the grain-of-uncertainty assumption about $p_r$. Hence, there does not exist any $p_r$ that satisfies the grain-of-uncertainty property whose conditionals converge to a completion of $\Mgu$ that satisfies the best-response inequalities. Therefore, no EE exists for this policy profile.
\end{proof}

\subsection{Proof of Theorem \ref{theorem:convergence-to-ode}}\label{app:proof-convergence-to-ode}
We begin by establishing the formal relationship between Subjective Embedded Equilibria (SEE) and Embedded Equilibria (EE) under the assumption of a common mixture model.

\begin{lemma}[SEE with Common Beliefs implies EE]
\label{lem:sde_is_ode}
Let $(\{\Mgp^i\}_{i\in N}, \{\Mmu^i\}_{i\in N})$ be a Subjective Embedded Equilibrium (SEE) for a repeated game with perfect monitoring. If all agents share the same mixture universe and its conditional completion, i.e., $\Mmu^i = \Mmu$ for all $i \in N$, then the set of policies $\{\Mgp^i\}_{i\in N}$ constitutes an Embedded Equilibrium (EE) w.r.t. the dependency distribution $q=\Mmu$.
\end{lemma}

\begin{proof}
By the definition of an SEE (Definition \ref{def:sde}), two conditions hold:
\begin{enumerate}
    \item \textbf{(Best Response)} Each agent's policy $\Mgp^i$ is an embedded best response to the conditional completion of the common mixture universe $\Mmu$.
    \item \textbf{(Uncontradicted Beliefs)} The mixture $\Mmu$ is identical to the ground-truth personal universe $\Mgu^i$ on the play path, i.e., $\Mmu(\Rgh) = \Mgu^i(\Rgh) = (\Mge^i)^{\Mgp^i}(\Rgh)$ for all $\Rgh \in \bar{\calA}^*$.
\end{enumerate}
We show that $\{\Mgp^i\}_{i\in N}$ is an EE. Let the EE policies be $f^i = \Mgp^i$. The corresponding ground-truth universe is $\Mgu_f = (\Mge^i)^{f^i} = (\Mge^i)^{\Mgp^i} = \Mgu^i$. We must specify a dependency distribution $q$ as required by Definition \ref{def:ode}. We define $q$ using the conditional completion of $\Mmu$:
$$ q(a^{-i} \mid \RghM, a^i) := \Mmu(a^{-i} \mid \RghM, a^i) \quad \forall (\RghM, a^i). $$
We now verify that $f^i = \Mgp^i$ is a best response with respect to the $q$-completion of $\Mgu_f$. This $q$-completion is defined by conditionals $\tilde{\Mgu}_f(\cdot \mid \cdot)$ such that:
$$ \tilde{\Mgu}_f(a^{-i} \mid \RghM, a^i) = \begin{cases} \Mgu_f(a^{-i} \mid \RghM, a^i) & \text{if } \Mgu_f(\RghM, a^i) > 0 \\ q(a^{-i} \mid \RghM, a^i) & \text{if } \Mgu_f(\RghM, a^i) = 0 \end{cases} $$
From condition (2), on the play path (where $\Mgu_f(\RghM, a^i) > 0$), we have $\Mgu_f = \Mmu$, and thus $\Mgu_f(\cdot \mid \cdot) = \Mmu(\cdot \mid \cdot)$. Off the play path (where $\Mgu_f(\RghM, a^i) = 0$), the $q$-completion uses $q(\cdot \mid \cdot)$, which we defined as the conditional completion of $\Mmu$.
Therefore, the $q$-completed universe $\tilde{\Mgu}_f$ is identical to the completed mixture universe $\Mmu$. By condition (1), $\Mgp^i$ is a best response to $\Mmu$. It follows that $f^i = \Mgp^i$ is a best response to $\tilde{\Mgu}_f$, satisfying Definition \ref{def:ode}.
\end{proof}

This equivalence allows us to adapt the arguments of \citet{kalai1993subjective} to approximate equilibria. Let us first define a new variant of the SEE that both has an approximate best response and approximate beliefs. 
\begin{definition}[$(\delta, \eta)$-Subjective Embedded Equilibrium]
\label{def:delta_eta_sde}
Let $\delta \ge 0$ and $\eta \ge 0$. A set of policies $\{\Mgp^i\}_{i\in N}$ and subjective mixture universes $\{\Mmu^i\}_{i\in N}$ (each with a specified conditional completion) constitutes a \textbf{$(\delta, \eta)$-Subjective Embedded Equilibrium} if, for each agent $i$:
\begin{enumerate}
    \item \textbf{($\delta$-Subjective Best Response)} The agent's policy $\Mgp^i$ is a $\delta$-best response with respect to its subjective mixture universe $\Mmu^i$:
     $V_{(\Mmu^i)^{\Mgp^i}}(\varepsilon) \ge \max_{\Mvp} V_{(\Mmu^i)^{\Mvp}}(\varepsilon) - \delta $
    \item \textbf{($\eta$-Uncontradicted Beliefs)} The subjective beliefs $\Mmu^i$ are $\eta$-close in total variation distance to the ground-truth personal distribution $\MultiAgentGeGpI$: $ D_\infty(\Mmu^i , \MultiAgentGeGpI \mid \varepsilon) \le \eta $
\end{enumerate}
\end{definition}

\begin{lemma}[Finite Games: $(\delta, \eta)$-SEE plays $\epsilon$-like a $\delta$-EE]
\label{lem:finite_sde_like_ode}
In finitely repeated games, for every $\epsilon > 0$ and $\delta \ge 0$, there exists an $\bar{\eta} > 0$ such that for all $0 < \eta \le \bar{\eta}$, if a set of policies $(\Mgp^i)_i$ and subjective mixtures $(\rho^i)_i$ are a $(\delta, \eta)$-SEE where all agents use an identical subjective mixture $\rho^i=\rho$ with identical conditional completions, then there exists a set of policies $(f^i)_i$ such that:
(i) $(\pi^i)_i$ plays $\epsilon$-like $(f^i)_i$: $D_\infty(\MultiAgentMge^{\MultiAgentMvp}, \MultiAgentMge^{\bar{f}}) < \epsilon$, and
(ii) $(f^i)_i$ is a $\delta$-EE.
\end{lemma}

\begin{proof}
We follow the proof strategy of \citet[Proposition 2]{kalai1993subjective}. Assume for contradiction that $\exists \epsilon > 0, \delta \ge 0$ such that $\forall \bar{\eta} > 0$, $\exists \eta \le \bar{\eta}$ and a $(\delta, \eta)$-SEE $(\Mvp^i,\rho^i)_i$ that does not play $\epsilon$-like any $\delta$-EE. Let us introduce the vector $g = (\Mvp^i,\rho^i)_{i=1}^N$.
The starting assumption allows the construction of a sequence of vectors $\{g(m)\}_{m=1}^\infty$ such that $g(m)$ is a $(\delta, \eta_m)$-SEE with $\eta_m \to 0$ as $m \to \infty$, and no $g(m)$ plays $\epsilon$-like any $\delta$-EE.
The set of behavior strategies in a finitely repeated game is sequentially compact. Therefore, there exists a subsequence $g(m_k)$ that converges to a limit strategy $g$.
By the continuity of the payoff functions, the limit strategy $g$ must be a $(\delta, 0)$-SEE. That is, the policies $(\pi^i)_i$ in $g$ are a $\delta$-best response to the common belief $\Mmu$ in $g$ that is $0$-close (i.e., identical) to the ground-truth $\MultiAgentMge^{\MultiAgentMvp}$.
By the logic of Lemma \ref{lem:sde_is_ode}, this $g$ is a $\delta$-EE.
Since $g(m) \to g$ and the measure-inducing map is continuous, for $m$ sufficiently large (and thus $\eta_{m}$ sufficiently small), $g(m)$ must play $\epsilon$-like $g$.
This contradicts the premise that no $g(m)$ plays $\epsilon$-like any $\delta$-EE, as $g$ itself is a $\delta$-EE. The initial assumption must be false.
\end{proof}

We now bridge the gap between finite and infinite games, proving the core relationship between $\eta$-SEEs and $\epsilon$-EEs.

\begin{lemma}[$\eta$-SEE plays $\epsilon$-like an $\epsilon$-EE]
\label{lem:infinite_eta_sde_like_epsilon_ode}
In infinitely repeated games, for every $\epsilon > 0$, there exists an $\bar{\eta} > 0$ such that for all $0 < \eta \le \bar{\eta}$, if $(\Mgp^i, \Mmu^i)_{i=1}^N$ is an $\eta$-SEE, where all agents use an identical subjective mixture $\Mmu^i=\Mmu$ with identical conditional completions, then there exists a set of policies $(f^i)_{i=1}^N$ such that:
(i) $(\pi^i)_i$ plays $\epsilon$-like $(f^i)_i$: $D_\infty(\MultiAgentMge^{\MultiAgentMvp}, \MultiAgentMge^{\bar{f}}) < \epsilon$, and
(ii) $(f^i)_i$ is an $\epsilon$-EE.
\end{lemma}

\begin{proof}
Let $\epsilon > 0$ be given. Due to the discount factor $\gamma \in [0,1)$, there exists a finite time $l = l(\epsilon) \in \mathbb{N}$ such that the maximum possible discounted payoff attainable from period $l+1$ onwards is less than $\epsilon/2$.
Let $g=(\Mgp^i, \Mmu^i)_{i=1}^N$ be an $\eta$-SEE for the infinite game. By Definition \ref{def:eps-sde}, $\Mgp^i$ is a $0$-best response to its subjective mixture $\Mmu$ which is $\eta$-close to the ground-truth personal distribution $\Mgu^i$.
Consider the $l$-truncation of all strategies and subjective mixtures. The truncated strategy $\Mgp^i_l$ is an $\epsilon/2$-best response to the truncated subjective mixture $\Mmu_l$, as the maximal utility loss from truncation is $\epsilon/2$. The truncated subjective mixture $\Mmu_l$ remains $\eta$-close to the truncated ground-truth $\Mgu_l^i$. Thus, $g_l$ is an $(\epsilon/2, \eta)$-SEE for the $l$-fold finite game.
By Lemma \ref{lem:finite_sde_like_ode} (with $\delta = \epsilon/2$), there exists an $\bar{\eta} > 0$ (which depends on $\epsilon$ via $l$) such that if $\eta \le \bar{\eta}$, $g_l$ plays $\epsilon$-like some $\epsilon/2$-EE, $f_l$.
We construct infinite strategies $(f^i)_i$ by extending $f^i_l$ with $\Mgp^i$ after period $l$:
$$ f^i(\RghA) := \begin{cases} f^i_l(\RghA) & \text{if } l(\RghA) < l \\ \Mgp^i(\RghA) & \text{if } l(\RghA) \ge l \end{cases} $$
We verify the two conditions for $f$:
(i) $(\Mgp^i)_i$ play $\epsilon$-like $(f^i)_i$: As $(\Mgp^i_l)_i$ play $\epsilon$-like $(f_l^i)_i$ in the $l$-fold game, and $f^i$ and $\Mgp^i$ are identical for all histories $\RghA$ with $l(\RghA) \ge l$, the induced distributions $\MultiAgentMge^{\MultiAgentMvp}$ and $\MultiAgentMge^{\bar{f}}$ are $\epsilon$-close.
(ii) $(f^i)_i$ is an $\epsilon$-EE: We check that $f^i$ is an $\epsilon$-best response. Since $(f_l^i)_i$ is an $\epsilon/2$-EE, any deviation in the first $l$ periods yields a payoff gain of at most $\epsilon/2$ during those periods. The maximum possible gain from any action taken after period $l$ is bounded by $\epsilon/2$. Therefore, the total possible gain from any deviation from $f^i$ is at most $\epsilon/2 + \epsilon/2 = \epsilon$. Thus, $(f^i)_i$ is an $\epsilon$-EE, with dependency distribution $q$ having conditionals equal to the conditional completion of the common mixture $\Mmu$.
\end{proof}

We now state and prove the main theorem.

\begin{theorem}[Convergence to $\epsilon$-Embedded Equilibrium]
\label{theorem-app:convergence-to-ode}
Let $\{\Mmu^i\}_{i\in N}$ be Bayesian mixture universes satisfying the grain-of-uncertainty and grain-of-truth conditions, and let $\{\Mgp^i\}_{i\in N}$ be the corresponding embedded best response policies in an infinitely repeated game with perfect monitoring. If the mixtures $\{\Mmu^i\}_{i\in N}$ of all the players are the same mixture, i.e., $\Mmu^i=\Mmu^j$ $\forall i,j\in N$, then, for each $\epsilon > 0$, there exists a finite time $T(\epsilon)$ such that for all $t \geq T(\epsilon)$, with probability greater than $1-\epsilon$, the tail distribution $\bar{\Mge}^{\bar{\Mgp}}_{\bar{a}_{<t}}$ induced by the tail policies $\MgpTailM[\bar{a}_{<t}]^i$ is $\epsilon$-close to the distribution induced by some policies constituting an $\epsilon$-embedded equilibrium in the tail game starting at time $t$.
\end{theorem}

\begin{proof}
Let $\epsilon > 0$ be given.
From Lemma \ref{lem:infinite_eta_sde_like_epsilon_ode}, there exists an $\bar{\eta} > 0$ (which depends on $\epsilon$) such that any $\eta$-SEE with $\eta \le \bar{\eta}$ plays $\epsilon$-like an $\epsilon$-EE.
Let $\eta' = \min(\epsilon, \bar{\eta})$.
As the mixture universes $\Mmu^i$ satisfy grain of truth and grain of uncertainty, and as all agents share the same mixture, it follows by Theorem \ref{thrm:convergence-to-sde} (Convergence to $\epsilon$-SEE) that for $\eta' > 0$, there exists a finite time $T(\eta')$ such that for all $t \ge T(\eta')$, with probability $\ge 1-\eta'$, the agents' tail play $(\MgpTailM[\bar{a}_{<t}]^i, \MmuTailM[\bar{a}_{<t}]^i)$ constitutes an $\eta'$-SEE, where all agents have an equal tail mixture $\MmuTailM[\bar{a}_{<t}]^i=\MmuTailM[\bar{a}_{<t}]$.
Since $\eta' \le \bar{\eta}$, Lemma \ref{lem:infinite_eta_sde_like_epsilon_ode} applies to the tail play $(\MgpTailM[\bar{a}_{<t}]^i, \MmuTailM[\bar{a}_{<t}]^i)$. Therefore, the tail distribution $\bar{\Mge}^{\bar{\Mgp}_{\bara_{<t}}}_{\bar{a}_{<t}}$ is $\epsilon$-close to the distribution induced by some policies $(f^i)_i$ constituting an $\epsilon$-EE for the tail game.
Furthermore, since $\eta' \le \epsilon$, the probability of this event, $1-\eta'$, is $\ge 1-\epsilon$. This completes the proof.
\end{proof}

\subsection{Proof of \cref{thrm:convergence-to-sde-correlated}}\label{app:proof-convergence-to-sde-correlation}
(i) As $\Mmu^i$ satisfies the grain-of-uncertainty assumption, its conditionals are always well-defined and hence its conditional completion is unique, and as a result there exists an embedded best response w.r.t. $\Mmu^i$ that is equal to $\pi^i$. (ii) From \cref{theorem:convergence-mixture-universe}, it follows from the grain-of-truth property that the tail mixture universes $\MmuTailM^i$ converge to the tail ground-truth distribution $\big(\MgeTailM^i\big)^{\MgpTailM^i}$ as $t\to\infty$ $\MultiAgentMge^{\MultiAgentMvp}$-almost-surely. It follows that for each $\epsilon > 0$, there exists a finite time $T(\epsilon)$ such that for all $t \geq T(\epsilon)$, with probability greater than $1-\epsilon$, we have that $D_\infty(\Mmu^i, \big(\Mge^i\big)^{\Mgp^i} \mid \HistMI) \leq \epsilon$. (iii) The tail policies and tail environment are conditioned on the personal histories $\HistMI$, which can differ for each agent due to partial observability; hence these histories serve as a correlation device $((\MATurnSet)^{t-1}, \MultiAgentMge^{\MultiAgentMvp})$ for the subsequent tail policies, leading to an $\epsilon$-subjective correlated equilibrium instead of an $\epsilon$-subjective embedded equilibrium.  

\subsection{Proof of \cref{prop:dogmatic-beliefs}}\label{app:proof-dogmatic-beliefs}
We follow an approach inspired by the \textit{dogmatic beliefs} introduced by \citet{leike2015bad}. We will construct for each agent $i$ a mixture model $\Mmu^i$ with completed conditionals that predicts eternal rewards of 0 after deviating at least once from the deterministic policy $\pi^i$, such that following the policy $\pi^i$ is always an embedded best response according to \eqref{eqn:embedded-br} (in the worst-case, $\pi^i$ also results in an expected future return of 0, but then $\pi^i$ still qualifies as an embedded best response, as all alternatives also lead to an expected return of 0). 

In the following, we assume without loss of generality that the minimum reward in $\calR$ is equal to 0. Let us define the finite set $\calE^i_0$ as the set of percepts encoding a reward of 0:
$$\calE^i_0:= \{e^i \in \calE^i: r^i(e^i)=0\}.$$
Let us define the function $\mathrm{dev}:\big(\TurnSet^i\big)^* \times \calMpol^i \to \bbN^+$ that computes the first timestep $t$ where the action $a^i_t$ in history $\HistA^i$ deviates from the action prescribed by the deterministic policy $\pi^i$. If $\HistA^i$ does not deviate from $\pi^i$, then we let $\mathrm{dev}(\HistA^i,\pi^i):=l(\HistA^i)+1$.
Now let us introduce the following two measures over $\big(\TurnSet^i\big)^* \cup \big(\big(\TurnSet^i\big)^* \times \calA^i\big)$:
\begin{align*}
    \Mvu_{\textrm{dogmatic}}(\HistA^i) &:= \prod_{t=1}^{\mathrm{dev}(\HistA^i,\pi^i)-1} \frac{1}{|\calA^i||\calE^i|} \prod_{t=\mathrm{dev}(\HistA^i,\pi^i)}^{l(\HistA^i)}\frac{\delta(e^i_t \in \calE_0^i)}{|\calA^i||\calE^i_0|}\,,\\  \Mvu_{\textrm{dogmatic}}(\HistA^i a^i)&:=\Mvu_{\textrm{dogmatic}}(\HistA^i)\frac{1}{|\calA^i|}\,,\\
    \Mvu_{\textrm{random}}(\Hist^i)&:= \frac{1}{\left(|\calA^i||\calE^i|\right)^{t}},
    \\\Mvu_{\textrm{random}}(\Hist^i a^i)&:=\Mvu_{\textrm{random}}(\Hist^i)\frac{1}{|\calA^i|}\,,
\end{align*}
with $\delta(e^i_t \in \calE_0^i)$ the indicator function indicating whether $e^i_t \in \calE_0^i$. Hence, for histories $\HistA^i$ where there is at least 1 action deviating from the one prescribed by $\pi^i$, $\Mvu_{\textrm{dogmatic}}$ only assigns non-zero probability to histories that have 0 reward after deviating for the first time from $\pi^i$. 

Take $\big(\Mge^i\big)^{\pi^i}(\HistA^i)$ as the personal trajectory distribution resulting from marginalizing out the $\HistA^{-i}$ in the ground-truth joint-trajectory distribution $\MultiAgentMge^{\MultiAgentMvp}(\MAHistA)$. Now we introduce the following infinite sequence of $\{\Mmu^i_r\}_{r \in \bbN}$ to obtain the conditional completions of $\Mmu^i$
$$\Mmu_r^i:= (1-\epsilon_r - \epsilon_r^2)\big(\Mge^i\big)^{\pi^i} + \epsilon_r \lambda_{\textrm{dogmatic}} + \epsilon_r^2\lambda_{\textrm{random}}\,,$$
with $\lim_{r\to\infty}\epsilon_r = 0$. It is easy to see that $\Mmu_r^i$ satisfies the grain-of-uncertainty condition and hence in the limit defines a valid conditional completion. Furthermore, we have that $\lim_{r\to\infty}\Mmu_r^i = \big(\Mge^i\big)^{\pi^i}$, and hence the resulting $\Mmu^i$ satisfies the \textit{uncontradicted beliefs} condition for the subjective embedded equilibrium (\cref{def:sde}). It remains to be shown that $\pi^i$ is an embedded best response w.r.t. $\Mmu^i$ for all agents $i$. 

Consider a history $\HistMI$ with $\big(\Mge^i\big)^{\pi^i}(\HistMI)>0$, i.e., all the actions follow $\pi^i$. Now consider that at time $t$, the action $a^i_t$ deviates from the deterministic policy $\pi^i(\HistMI)$. In this case, $\big(\Mge^i\big)^{\pi^i}(\HistMI a_t^i)=0$ and hence the predictive distribution $\Mmu_r^i$ is equal to
$$\Mmu_r^i(e_t^i\Turn^i_{>t} \mid \HistMI a_t^i) = \frac{\epsilon_r \lambda_{\textrm{dogmatic}}(\HistMI a_t^ie_t^i\Turn^i_{>t}) + \epsilon_r^2 \lambda_{\textrm{random}}(\HistMI a_t^ie_t^i\Turn^i_{>t})}{\epsilon_r \lambda_{\textrm{dogmatic}}(\HistMI a_t^i) + \epsilon_r^2 \lambda_{\textrm{random}}(\HistMI a_t^i)},$$
and hence
$$\lim_{r\to\infty} \Mmu_r^i(e_t^i\Turn^i_{>t} \mid \HistMI a_t^i) = \lambda_{\textrm{dogmatic}}(e_t^i\Turn^i_{>t} \mid \HistMI a_t^i)\,,$$
for any future trajectory $e_t^i\Turn^i_{>t}$. Notice that $\lambda_{\textrm{dogmatic}}(e_t^i\Turn^i_{>t} \mid \HistMI a_t^i)$ only has non-zero probability for future trajectories $e_t^i\Turn^i_{>t}$ that have a total sum of rewards of 0. Hence, after deviating once from $\pi^i$, the value of any policy is 0, including the optimal policy. Hence, following the deterministic policy $\pi^i$ is always better or equal to following any other policy, as following another policy leads to the minimum value of 0. Hence $\pi^i$ is an embedded best response w.r.t. $\Mmu^i = \lim_{r\to\infty}\Mmu^i_r$, and hence $(\Mgp^i, \Mmu^i)_i$ is an SEE. We can repeat the same proof technique for constructing a SNE $(\Mgp^i, \Mme^i)_i$, taking $\Mme^i(e^i \mid \HistA^i a^i)=\Mmu^i(e^i \mid \HistA^i a^i)$ for the above constructed $\Mmu^i$. This concludes the proof.

\subsection{Proof of \cref{thrm:convergence-to-epsdelta-sde-correlated}}\label{app:proof-convergence-to-epsdelta-sde-correlated}

We need a few lemmas, which we state in the single-agent setting for simplicity.

\begin{lemma}
    \label{lem:k-step-better-than-k-1} For all $k\geq 1$, we have $Q_{\Mmu}^k(\HistM, a_t)\leq Q_{\Mmu}^{k+1}(\HistM, a_t)$.
\end{lemma}
\begin{proof}
    This is an immediate corollary from \eqref{eqn:k-step-qval} that we prove by induction on $k\geq 1$:
    \begin{itemize}
        \item For $k=1$, we have
        \begin{align*}
            Q_{\Mmu}^1(\HistM, a_t)&=Q_{\Mmu}(\HistM, a_t)\\
            &=\sum_{e_t\in\calE}\Mmu(e_t|\HistM a_t)\left[(1-\gamma)r(e_t)+\gamma V_{\Mmu}(\HistM a_te_t)\right]\\
            &=\sum_{e_t\in\calE}\Mmu(e_t|\HistM a_t)\left[(1-\gamma)r(e_t)+\gamma \sum_{a_{t+1}\in\calA}\Mmu(a_{t+1}|\HistM a_te_t) Q_{\Mmu}(\HistM a_te_t, a_{t+1})\right]\\
            &\leq\sum_{e_t\in\calE}\Mmu(e_t|\HistM a_t)\left[(1-\gamma)r(e_t)+\gamma \max_{a\in\calA}Q_{\Mmu}^1(\HistM a_te_t, a)\right]\\
            &=Q_{\Mmu}^2(\HistM, a_t),
        \end{align*}
        hence the lemma is correct for $k=1$.
        \item Let $k>1$, and assume that the lemma is true for $k-1$. We have
        \begin{align*}
            Q_{\Mmu}^k(\HistM, a_t)&=\sum_{e_t\in\calE}\Mmu(e_t|\HistM a_t)\left[(1-\gamma)r(e_t)+\gamma\max_{a\in\calA} Q_{\Mmu}^{k-1}(\HistM a_te_t, a)\right]\\
            &\stackrel{(\ast)}{\leq} \sum_{e_t\in\calE}\Mmu(e_t|\HistM a_t)\left[(1-\gamma)r(e_t)+\gamma\max_{a\in\calA} Q_{\Mmu}^{k}(\HistM a_te_t, a)\right]\\
            &=Q_{\Mmu}^{k+1}(\HistM, a_t)\,,
        \end{align*}
        where $(\ast)$ follows from the induction hypothesis.
    \end{itemize}
    It follows by induction that the lemma holds for all $k\geq 1$.
\end{proof}

\begin{lemma}
    \label{lem:k-step-planning-at-every-step-better-than-once} If $\Mvp$ is a $k$-step planning policy w.r.t. the mixture universe $\Mmu$, then for every history $\HistM$ we have
    $$V_{\Mmu^\Mvp}(\HistM)\geq \max_{a}Q_{\Mmu}^k(\HistM,a)\geq \max_{a}Q_{\Mmu}(\HistM,a)\geq V_\Mmu(\HistM)\,.$$
\end{lemma}
\begin{proof}
    In the light of \cref{lem:k-step-better-than-k-1}, it suffices to show the first inequality.

    From \cref{def:k-step-eba} we have
    $$\Mvp(a_t \mid \HistM)>0\quad \Rightarrow\quad Q_{\Mmu}^k(\HistM, a_t)=\max_{a} Q_{\Mmu}^k(\HistM, a)\,.$$ 
    Therefore,
    \begin{equation}
        \label{eq:q-k-step-max-equals-average-over-maximizer}
        \sum_{a_t\in\calA}\Mvp(a_t|\HistM) Q_{\Mmu}^k(\HistM, a_t)=\max_{a} Q_{\Mmu}^k(\HistM, a)\,.
    \end{equation}

    Now from \cref{lem:k-step-better-than-k-1} and \eqref{eqn:k-step-qval} we have
    \begin{equation}
    \label{eq:q-k-step-inequality-induction-helper}
    \begin{aligned}
        Q_{\Mmu}^k(\HistM, a_t)&\leq Q_{\Mmu}^{k+1}(\HistM, a_t)=\sum_{e_t\in\calE}\Mmu(e_t|\HistM a_t)\left[(1-\gamma)r(e_t)+\gamma\max_{a\in\calA} Q_{\Mmu}^{k}(\HistM a_te_t, a)\right]\\
        &\stackrel{(\ast)}{=}\sum_{e_t\in\calE}\Mmu(e_t|\HistM a_t)\left[(1-\gamma)r(e_t)+\gamma\sum_{a_{t+1}\in\calA}\Mvp(a_{t+1}|\HistM a_t e_t) Q_{\Mmu}^k(\HistM a_t e_t, a_{t+1})\right]\\
        &\stackrel{(\dagger)}{=}(1-\gamma)\sum_{e_t\in\calE}\Mmu^\Mvp(e_t|\HistM a_t)r(e_t)+\gamma\sum_{e_t\in\calE,a_{t+1}\in\calA}\Mmu^\Mvp(e_ta_{t+1}|\HistM a_t ) Q_{\Mmu}^k(\HistM a_t e_t, a_{t+1})
    \end{aligned}
    \end{equation}
    where $(\ast)$ follows from applying \eqref{eq:q-k-step-max-equals-average-over-maximizer} to the history $\HistM a_t e_t$ instead of $\HistM$, and $(\dagger)$ follows from the definition of $\Mmu^\Mvp$.

    We will now show by induction on $\ell\geq 1$ that
    \begin{equation}
    \label{eq:q-k-step-inequality-induction-to-prove}
    \begin{aligned}
    \max_{a} Q_{\Mmu}^k(\HistM, a)&\leq
        (1-\gamma)\sum_{i=0}^{\ell-1}\gamma^{i}\sum_{\text{\ae}_{t:t+i}\in\TurnSet^{i+1}}r(e_{t+i})\Mmu^\Mvp(\text{\ae}_{t:t+i}|\HistM)\\
        &\quad\quad+\gamma^{\ell}\sum_{(\text{\ae}_{t:t+\ell-1},a_{t+\ell})\in\TurnSet^{\ell}\times \calA} \Mmu^\Mvp(\text{\ae}_{t:t+\ell-1}a_{t+\ell}|\HistM) Q_{\Mmu}^k(\HistM\text{\ae}_{t:t+\ell-1},a_{t+\ell})\,.
    \end{aligned}
    \end{equation}

    By combining \eqref{eq:q-k-step-max-equals-average-over-maximizer} and \eqref{eq:q-k-step-inequality-induction-helper} we get
    \begin{align*}
        &\max_{a} Q_{\Mmu}^k(\HistM, a)\\
        &=\sum_{a_t\in\calA}\Mvp(a_t|\HistM) Q_{\Mmu}^k(\HistM, a_t)\\
        &\leq \sum_{a_t\in\calA}\Mvp(a_t|\HistM)\left((1-\gamma)\sum_{e_t\in\calE}\Mmu^\Mvp(e_t|\HistM a_t)r(e_t)+\gamma\sum_{e_t\in\calE,a_{t+1}\in\calA}\Mmu^\Mvp(e_ta_{t+1}|\HistM a_t ) Q_{\Mmu}^k(\HistM a_t e_t, a_{t+1})\right)\\
        &\stackrel{(\ddagger)}{=} (1-\gamma)\sum_{\text{\ae}_{t:t}\in\TurnSet}r(e_{t})\Mmu^\Mvp(\text{\ae}_{t:t}|\HistM)+\gamma\sum_{(\text{\ae}_{t:t},a_{t+1})\in\TurnSet\times \calA} \Mmu^\Mvp(\text{\ae}_{t:t}a_{t+1}|\HistM) Q_{\Mmu}^k(\HistM\text{\ae}_{t:t},a_{t+1})\,,
    \end{align*}
    where $(\ddagger)$ follows from the definition of $\Mmu^\Mvp$ and the convention that $\text{\ae}_{t:t}=a_te_t$. Therefore, \eqref{eq:q-k-step-inequality-induction-to-prove} holds for $\ell=1$.

    Now let $\ell>1$ and assume that \eqref{eq:q-k-step-inequality-induction-to-prove} holds for $\ell-1$. We have

    \begin{align*}
    &\max_{a} Q_{\Mmu}^k(\HistM, a)\\
    &\leq
        (1-\gamma)\sum_{i=0}^{\ell-2}\gamma^{i}\sum_{\text{\ae}_{t:t+i}\in\TurnSet^{i+1}}r(e_{t+i})\Mmu^\Mvp(\text{\ae}_{t:t+i}|\HistM)\\
        &\quad\quad+\gamma^{\ell-1}\sum_{(\text{\ae}_{t:t+\ell-2},a_{t+\ell-1})\in\TurnSet^{\ell-1}\times \calA} \Mmu^\Mvp(\text{\ae}_{t:t+\ell-2}a_{t+\ell-1}|\HistM) Q_{\Mmu}^k(\HistM\text{\ae}_{t:t+\ell-2},a_{t+\ell-1})\\
    &\stackrel{(\diamond)}{\leq}
        (1-\gamma)\sum_{i=0}^{\ell-2}\gamma^{i}\sum_{\text{\ae}_{t:t+i}\in\TurnSet^{i+1}}r(e_{t+i})\Mmu^\Mvp(\text{\ae}_{t:t+i}|\HistM)\\
        &\quad\quad+\gamma^{\ell-1}\sum_{(\text{\ae}_{t:t+\ell-2},a_{t+\ell-1})\in\TurnSet^{\ell-1}\times \calA} \Mmu^\Mvp(\text{\ae}_{t:t+\ell-2}a_{t+\ell-1}|\HistM) \Bigg(\\
        &\quad\quad\quad\quad\quad(1-\gamma)\sum_{e_{t+\ell-1}\in\calE}\Mmu^\Mvp(e_{t+\ell-1}|\HistM\text{\ae}_{t:t+\ell-2} a_{t+\ell-1})r(e_{t+\ell-1})\\
        &\quad\quad\quad\quad\quad+\gamma\sum_{e_{t+\ell-1}\in\calE,a_{t+\ell}\in\calA}\Mmu^\Mvp(e_{t+\ell-1}a_{t+\ell}|\HistM\text{\ae}_{t:t+\ell-2} a_{t+\ell-1} ) Q_{\Mmu}^k(\HistM\text{\ae}_{t:t+\ell-2} a_{t+\ell-1} e_{t+\ell-1}, a_{t+\ell})\Bigg)\\
    &=
        (1-\gamma)\sum_{i=0}^{\ell-1}\gamma^{i}\sum_{\text{\ae}_{t:t+i}\in\TurnSet^{i+1}}r(e_{t+i})\Mmu^\Mvp(\text{\ae}_{t:t+i}|\HistM)\\
        &\quad\quad+\gamma^{\ell}\sum_{(\text{\ae}_{t:t+\ell-1},a_{t+\ell})\in\TurnSet^{\ell}\times \calA} \Mmu^\Mvp(\text{\ae}_{t:t+\ell-1}a_{t+\ell}|\HistM) Q_{\Mmu}^k(\HistM\text{\ae}_{t:t+\ell-1},a_{t+\ell})\,,
    \end{align*}
    where $(\diamond)$ follows from applying \eqref{eq:q-k-step-inequality-induction-helper} to $Q_{\Mmu}^k(\HistM\text{\ae}_{t:t+\ell-2},a_{t+\ell-1})$. We conclude that \eqref{eq:q-k-step-inequality-induction-to-prove} holds for every $\ell\geq 1$. In particular,
    \begin{align*}
        \max_{a} Q_{\Mmu}^k(\HistM, a)&\leq
        \lim_{\ell\to\infty}(1-\gamma)\sum_{i=0}^{\ell-1}\gamma^{i}\sum_{\text{\ae}_{t:t+i}\in\TurnSet^{i+1}}r(e_{t+i})\Mmu^\Mvp(\text{\ae}_{t:t+i}|\HistM)\\
        &\quad\quad+\gamma^{\ell}\sum_{(\text{\ae}_{t:t+\ell-1},a_{t+\ell})\in\TurnSet^{\ell}\times \calA} \Mmu^\Mvp(\text{\ae}_{t:t+\ell-1}a_{t+\ell}|\HistM) Q_{\Mmu}^k(\HistM\text{\ae}_{t:t+\ell-1},a_{t+\ell})\\
        &\stackrel{(*)}{=}\lim_{\ell\to\infty}(1-\gamma)\sum_{i=0}^{\ell-1}\gamma^{i}\sum_{\text{\ae}_{t:t+i}\in\TurnSet^{i+1}}r(e_{t+i})\Mmu^\Mvp(\text{\ae}_{t:t+i}|\HistM)\\
        &\stackrel{(\dagger)}{=}V_{\Mmu^\Mvp}(\HistM)\,,
    \end{align*}
    where in the above equation, $(\ast)$ follows from the fact that $Q_{\Mmu}^k(\HistM\text{\ae}_{t:t+\ell-1},a_{t+\ell})\in[0,1]$ and $\lim_{\ell\to\infty}\gamma^\ell=0$, and $(\dagger)$ follows from the definition of the value function (cf. \cref{def:value-function}).
\end{proof}

We also need the following useful lemma from \citet{leike2016nonparametric}:
    \begin{lemma}[\citet{leike2016nonparametric}, Lemma 4.17]\label{lemma:value-diff-bound}
    For any two environments $\nu_1$ and $\nu_2$, and for any two policies $\pi_1$ and $\pi_2$, we have that
        $$\vert V_{\nu_1^{\pi_1}}(\Hist ) -  V_{\nu_2^{\pi_2}}(\Hist ) \vert \leq D_{\infty}(\nu_1^{\pi_1}, \nu_2^{\pi_2} \mid \Hist )\,.$$
    \end{lemma}

Now we prove an analogous statement of \cref{thrm:convergence-to-epsdelta-sde-correlated} for the single-agent setting.

\begin{lemma}\label{lemma:convergence-to-optimal-planning}
    Let $\pi$ be the policy of an embedded Bayesian agent in environment $\Mge$, implementing a $k$-step-planning policy using Bayesian mixture universe $\Mmu$, satisfying the grain of truth and grain-of-uncertainty conditions, as well as the conditions that $\Mmu$ dominates $\Mmu^\Mgp$ and $\Mmu^\Mgp$ dominates $\Mge^\Mgp$, and the following sensibly off-policy condition: There exists a positive $\alpha<1/\gamma -1$ and a $T$ such that for all $t \geq T$, it holds that
        $$\expect{\Mmu^\pi(\Hist)}{\max_{a} \sum_{e} \Mmu(e \mid \Hist a)[V^*_\Mmu(\Hist ae) - V_\Mmu(\Hist ae)]} \leq (1+\alpha) \expect{\Mmu^\pi(\Hist ae)}{V^*_\Mmu(\Hist ae) - V_\Mmu(\Hist ae)} \,.$$
    Then it holds that 
    $$\expect{\Mge^{\Mgp}(\Hist)}{V^*_\Mmu(\Hist ) - V_{\Mmu^\pi}(\Hist )} \to 0 \quad \text{as} \quad t \to \infty\,,$$
    with $V^*_\Mmu(\Hist)$ the optimal value resulting from an embedded best-response w.r.t. $\Mmu$. 
\end{lemma}

\begin{proof}
    We follow the approach of \citet{catt2023self}, combining their Theorem 16 with Theorem 18, while generalizing the proof to $k$-step planning. We have 
    $$0 \stackrel{(\ast)}\leq V_{\Mmu^\pi}(\Hist) - V_\Mmu(\Hist) \stackrel{(\dagger)}\leq D_\infty(\Mmu^\pi, \Mmu \mid \Hist)\,,$$
    where $(\ast)$ follows from \cref{lem:k-step-planning-at-every-step-better-than-once} as $\pi$ is a $k$-step planner policy, and $(\dagger)$ follows from \cref{lemma:value-diff-bound}. As from the stated assumptions we have that $\Mmu$ dominates $\Mmu^\pi$, we have from the merging of opinions theorem \citep{blackwell1962merging} that $D_\infty(\Mmu, \Mmu^\pi \mid \Hist ) \to 0$ as $t\to \infty$ $\Mmu^\Mgp$-almost surely. Hence, for any $\beta >0$, there exists a $t_1$ such that for all $t\geq t_1$, we have that
    \begin{equation}
    \label{eqn:proof-selfaixi-valuebound}
    \begin{aligned}
        \expect{\Mmu^\pi(\Hist)}{\max_a Q_{\Mmu}(\Hist, a) - V_\Mmu(\Hist )} &\stackrel{(\dagger)}{\leq} \expect{\Mmu^\pi(\Hist)}{V_{\Mmu^\pi}(\Hist ) - V_\Mmu(\Hist )}\\
        &\leq \expect{\Mmu^\pi(\Hist)}{D_\infty(\Mmu^\pi, \Mmu \mid \Hist )} \leq \beta\,, 
    \end{aligned}
    \end{equation}
    where for $(\dagger)$ we used the inequality $V_{\Mmu^\pi}(\Hist ) \geq \max_a Q_{\Mmu}(\Hist, a) \geq V_\Mmu(\Hist )$ from \cref{lem:k-step-planning-at-every-step-better-than-once}. 
    Now let us proceed by showing that for self-models that are good models of the $k$-step planning policy, i.e., that satisfy the above inequality, we have that the resulting value is close to the optimal value corresponding to infinite-horizon optimal planning w.r.t. $\Mmu$, when assuming that the \emph{sensibly off-policy} condition holds. Let us define the optimality gap as
    $$\Delta(\Hist) := V^*_\Mmu(\Hist) - V_{\Mmu}(\Hist)\,,$$
    with $V^*_\Mmu(\Hist)$ the optimal value resulting from infinite-horizon optimal planning w.r.t. $\Mmu$. 
    Take $t \geq \max(T,t_1)$ with $T$ defined in the Lemma statement, and $t_1$ defined for \eqref{eqn:proof-selfaixi-valuebound}. Then we have that
    \begin{align*}
        &\expect{\Mmu^\pi(\Hist)}{\Delta(\Hist )} \\
        &= \bbE[V^*_\Mmu(\Hist ) - V_\Mmu(\Hist )]\\
        &= \expect{\Mmu^\pi(\Hist)}{\max_a Q^*_\Mmu(\Hist, a) - \max_aQ_\Mmu(\Hist, a) + \max_aQ_\Mmu(\Hist, a) - V_\Mmu(\Hist )} \\
        &= \expect{\Mmu^\pi(\Hist)}{\max_a Q^*_\Mmu(\Hist, a) - \max_aQ_\Mmu(\Hist, a)} + \expect{\Mmu^\pi(\Hist)}{\max_aQ_\Mmu(\Hist, a) - V_\Mmu(\Hist )} \\
        &\stackrel{(\dagger)}{\leq} \expect{\Mmu^\pi(\Hist)}{\max_a Q^*_\Mmu(\Hist, a) - \max_aQ_\Mmu(\Hist, a)} + \beta\\
        &\leq \expect{\Mmu^\pi(\Hist)}{\max_a (Q^*_\Mmu(\Hist, a) - Q_\Mmu(\Hist, a))} + \beta \\
        &=\expect{\Mmu^\pi(\Hist)}{\max_a \gamma \expect{\Mmu(e \mid \Hist, a)}{V^*_\Mmu(\Hist ae) - V_\Mmu(\Hist ae)}} + \beta\\
        &\stackrel{(\ddagger)}{\leq} (1+\alpha)\gamma \expect{\Mmu^\pi(\Hist ae)}{V^*_\Mmu(\Hist ae) - V_\Mmu(\Hist ae)} +\beta \\
        &= (1+\alpha)\gamma \expect{\Mmu^\pi(\Hist ae)}{\Delta(\Hist ae)} + \beta\,,
    \end{align*}
    where in $(\dagger)$ we used \eqref{eqn:proof-selfaixi-valuebound} and in $(\ddagger)$ the sensibly off-policy condition. As the above holds for any $t \geq \max(T,t_1)$, we can recursively apply it and obtain
    \begin{align*}
        \expect{\Mmu^\pi(\Hist )}{\Delta(\Hist )} \leq \beta\sum_{l=0}^\infty[(1+\alpha)\gamma]^l = \frac{\beta}{1 - (1+ \alpha)\gamma}\,,
    \end{align*}
    where we used the assumption that $\alpha<1/\gamma -1$. 

    As $\Mmu^\pi$ dominates $\Mge^{\Mgp}$ (cf. the stated assumptions), we have for some finite constant $C$ that 
    $$\expect{\Mge^{\Mgp}(\Hist )}{\Delta(\Hist )} \leq \frac{1}{C}\expect{\Mmu^{\pi}(\Hist )}{\Delta(\Hist )} \leq \frac{\beta}{C(1 - (1+ \alpha)\gamma)}\,.$$
    Finally, as $V^*_\Mmu(\Hist) \geq V_{\Mmu^\pi}(\Hist) \geq V_{\Mmu}(\Hist)$, and as the above holds for any $\beta > 0$, we get that 
    \begin{align*}
        \expect{\Mge^{\Mgp}(\Hist)}{V^*_\Mmu(\Hist ) - V_{\Mmu^\pi}(\Hist )} \to 0 \quad \text{as} \quad t \to \infty\,.
    \end{align*}
    
\end{proof}

We are now ready to prove \cref{thrm:convergence-to-epsdelta-sde-correlated}.
\begin{proof}
    (i) Using \cref{lemma:convergence-to-optimal-planning}, we have that under the stated conditions, the agent's $k$-step planner policies converge to infinite-horizon optimal planning w.r.t. $\Mmu^i$ in $\big(\Mge^i\big)^{\Mgp^i}$-mean. As convergence in mean implies convergence in probability for bounded random variables, we have that for all $\delta > 0$, the probability that the policy $\pi^i$ is a $\delta$-embedded best response w.r.t. subjective environment $\Mmu^i$ for the history $\HistMI $, converges to 1: 
    $$\bbP_{(\Mge^i)^{\Mgp^i}}\left[V^*_{\Mmu^i}(\HistMI ) - V_{{(\Mmu^i)}^{\pi^i}}(\HistMI ) \leq \delta \right] \to 1 ~~\text{as}~~t\to \infty\,,$$
    with $\bbP_{(\Mge^i)^{\Mgp^i}}$ the probability distribution of $\HistMI$ under $\big(\Mge^i\big)^{\Mgp^i}$. (ii) As $\Mmu^i$ dominates $\big(\Mge^i\big)^{\Mgp^i}$, we have from \cref{theorem:convergence-mixture-universe} that the tail mixtures $\MmuTailM^i$ converge to the tail ground-truth distribution $\big(\MgeTailM^i\big)^{\MgpTailM}$ as $t\to\infty$ $\MultiAgentMge^{\MultiAgentMvp}$-almost-surely. Almost-sure convergence implies convergence in probability, hence we have that 
    $$\bbP_{(\Mge^i)^{\Mgp^i}}\left[D_\infty(\Mmu^i, \big(\Mge^i\big)^{\Mgp^i} \mid \HistMI ) \leq \epsilon \right] \to 1 ~~\text{as}~~t\to \infty.$$
    (iii) The policies and environment are conditioned on the history, which can differ for each agent due to partial observability; hence these histories serve as a correlation device $(\MATurnSet^{t-1}, \MultiAgentMge^{\MultiAgentMvp})$ for the subsequent tail policies.
\end{proof}

\subsection{Proof of \cref{prop:convergence-of-approximate-eba-to-sde-correlated}}
\label{app:proof-convergence-of-approximate-eba-to-sde-correlated}
We first prove that the policy of an approximate embedded Bayesian agent with growing planning horizons and improving accuracy has a value that converges to the optimal value with respect to the subjective environment induced by the mixture universe:

\begin{proposition}\label{prop:convergence-of-approximate-eba-to-optimal-value}
Let $\pi$ be the policy of an embedded $(k_t,\epsilon_t)$-approximate Bayes-optimal agent with respect to a mixture universe model $\Mmu$. If $\lim_{t\to\infty}k_t=\infty$ and $\lim_{t\to\infty}\epsilon_t=0$, it holds that
$$\lim_{t\to\infty}V_{\Mmu^\pi}(\HistM)=\lim_{t\to\infty}V^{*}_{\Mmu}(\HistM)\,.$$
\end{proposition}
\begin{proof}
    First notice that for every action $a_t$, we have
    $$Q_{\Mmu}^{k_t}(\HistM,a_t)\geq Q_{\Mmu}^{*}(\HistM,a_t)-\gamma^{k_t}\,.$$
    The reason is that the non-optimality of a planner with a horizon of $k_t$ steps will only come from the contribution of the discounted rewards after $k_t$ steps, which would add up to at most $$(1-\gamma)\sum_{t'\geq t+k_{t}}\gamma^{t'-t}= (1-\gamma)\sum_{t'\geq k_{t}}\gamma^{t'}=\gamma^{k_t}\,.$$
    Therefore, by defining $\tilde{\epsilon}_t=\epsilon_t+\gamma^{k_t}$, we get that
    $$\pi(a_t|\HistM)> 0\quad\Rightarrow\quad Q^{k_t}_{\Mmu}(\HistM,a_t)\geq\max_{a'\in\mathcal{A}}Q_\Mmu^{k_t}(\HistM,a')-\epsilon_t\geq \max_{a'\in\mathcal{A}}Q_\Mmu^{*}(\HistM,a')-\tilde{\epsilon}_t=V_\Mmu^{*}(\HistM)-\tilde{\epsilon}_t\,.$$
    In particular, we also have
    $$\pi(a_t|\HistM)> 0\quad\Rightarrow\quad Q^{*}_{\Mmu}(\HistM,a_t)\geq Q^{k_t}_{\Mmu}(\HistM,a_t)\geq V_\Mmu^{*}(\HistM)-\tilde{\epsilon}_t\,.$$

    Now let us define $\Delta(\HistM)=V_\Mmu^{*}(\HistM)-V_{\Mmu^\pi}(\HistM)$. We have:
    \begin{align*}
        \Delta(\HistM) &=V_\Mmu^{*}(\HistM)-V_{\Mmu^\pi}(\HistM)=\expect{a_t\sim\pi(\cdot|\HistM)} {V_\Mmu^{*}(\HistM)-Q_{\Mmu^\pi}(\HistM,a_t)}\\
        &\leq \expect{a_t\sim\pi(\cdot|\HistM)} {\tilde{\epsilon}_t + Q_\Mmu^{*}(\HistM,a_t)-Q_{\Mmu^\pi}(\HistM,a_t)}\\
        &=\tilde{\epsilon}_t +  \gamma\expect{a_t\sim\pi(\cdot|\HistM)} {\expect{e_t\sim\Mmu(\cdot|\HistM a_t)} {V_\Mmu^{*}(\HistM a_te_t)-V_{\Mmu^\pi}(\HistM a_te_t)}}\\
        &=\tilde{\epsilon}_t + \gamma\expect{a_te_t\sim\Mmu^\pi(\cdot|\HistM )} {\Delta(\HistM a_te_t)}\,.
    \end{align*}

    We can then show by induction on $\ell\geq 0$ that
    \begin{align*}
        \Delta(\HistM ) &\leq \sum_{t'=0}^{\ell-1} \tilde{\epsilon}_{t+t'}\gamma^{t'} +  \gamma^{\ell}\expect{\Turn_{t:t+\ell-1}\sim\Mmu^\pi(\cdot|\HistM )} {\Delta(\HistM \Turn_{t:t+\ell-1})}\,,
    \end{align*}
    and by taking $\ell\to\infty$, we get
    \begin{align*}
        \Delta(\HistM ) &\leq \sum_{t'=0}^{\infty} \tilde{\epsilon}_{t+t'}\gamma^{t'}\leq \left(\max_{t'\geq 0}\tilde{\epsilon}_{t'+t}\right)\sum_{t'=0}^{\infty} \gamma^{t'}=\frac{1}{1-\gamma}\max_{t'\geq 0}\tilde{\epsilon}_{t'+t}\,.
    \end{align*}
    Since $\lim_{t\to\infty}\tilde{\epsilon}_t=0$, we have $\lim_{t\to\infty}\max_{t'\geq 0}\tilde{\epsilon}_{t'+t}=0$, and hence
    \begin{align*}
        \lim_{t\to\infty}\Delta(\HistM ) =0\,.
    \end{align*}
\end{proof}

Now we are ready to prove \cref{prop:convergence-of-approximate-eba-to-sde-correlated}:
(i) As $\Mmu^i$ satisfies the grain-of-uncertainty assumption, its conditionals are always well-defined and hence its conditional completion is unique. The policy $\pi^i$ is a $(k_t,\epsilon_t)$-approximate embedded best response w.r.t. this unique conditional completion. (ii) From \cref{theorem:convergence-mixture-universe}, it follows from the grain-of-truth property that the tail mixtures $\MmuTailM^i$ converges to the tail ground-truth distribution $\big(\MgeTailM^i\big)^{\MgpTailM^i}$ as $t\to\infty$ $\MultiAgentMge^{\MultiAgentMvp}$-almost-surely.
It follows that for each $\epsilon > 0$, there exists a finite time $t_1(\epsilon)$ such that for all $t \geq t_1(\epsilon)$, with probability greater than $1-\epsilon$, we have that $D_\infty(\Mmu^i, \big(\Mge^i\big)^{\Mgp^i} \mid \MAHistI) \leq \epsilon$. (iii) The policies and environment are conditioned on the history, which can differ for each agent due to partial observability; hence these histories serve as a correlation device $(\MATurnSet^{t}, \MultiAgentMge^{\MultiAgentMvp})$ for the subsequent tail policies. (iv) From Proposition \cref{prop:convergence-of-approximate-eba-to-optimal-value} we can see that the value of the policies $\pi^i$ converge to the optimal value with respect to the mixture $\Mmu^i$, which means that for every $\delta>0$, there exists a finite time $t_2(\delta)$ such that for all $t \geq t_2(\delta)$, $\pi^i$ is a $\delta$-best response with respect to $\Mmu^i$. Taking $T=\max(t_1,t_2)$ concludes the proof.

\subsection{Proof of \cref{lem:approximate_universal_inductor}}
\label{app:lem_approximate_universal_inductor_proof}

We start by showing the lemma for the case $l(h')=1$, i.e., $h'=x\in\mathcal{B}$.

For a target precision $\epsilon>0$, we can do the following procedure in order to estimate $\Mmu_\tau^w(x|h)$:
\begin{itemize}
    \item Start with $\mathrm{high}\leftarrow1$ and $\mathrm{low}\leftarrow0$, and then repeat the following as long as $\mathrm{high}-\mathrm{low}>\epsilon$:
    \begin{itemize}
        \item Compute $\mathrm{mid}\leftarrow\frac{\mathrm{low}+\mathrm{high}}{2}$.
        \item Send the request $\langle x,\mathrm{mid},h\rangle$ to the oracle.
        \item If we get 0, then we must have $\Mmu_\tau^w(x|h)\leq\mathrm{mid}$, so we set $\mathrm{high}\leftarrow\mathrm{mid}$.
        \item If we get 1, then we must have $\Mmu_\tau^w(x|h)\geq\mathrm{mid}$, so we set $\mathrm{low}\leftarrow\mathrm{mid}$.
    \end{itemize}
    \item Since the inequality $\mathrm{low}\leq \Mmu_\tau^w(x|h)\leq \mathrm{high}$ is an invariant of the loop, and since $|\mathrm{high}-\mathrm{low}|$ decays exponentially with the number of iterations, we will exit the loop after $\lceil -\log_2\epsilon\rceil$ iterations and the value of $\mathrm{low}$ (resp. $\mathrm{high}$) would be an $\epsilon$-accurate estimate of $\Mmu^w_\tau(x|h)$ from below (resp. from above).
\end{itemize}

Now we turn to the case where $t':=l(h')>1$. We may assume without loss of generality that $\epsilon < 1$ as otherwise we can replace $\epsilon$ by $\min\{\epsilon,1/2\}$.

Let us write $h'=(x_1',\ldots,x_{t'}')$ so that
$$\Mmu^{w}_\tau(h'|h)=\prod_{i=1}^{t'}\Mmu^{w}_\tau(x'_i|hh'_{<i})\,.$$

Let $\epsilon':=\epsilon'(\epsilon,t')>0$ be a number that computably depends only on $\epsilon$ and $t'$ which we will choose later. From the above procedure, we can get for every $1\leq i\leq t'$, an approximation $\tilde{\Mmu}^{w}_\tau(x'_i|hh'_{<i})$ of $\Mmu^{w}_\tau(x'_i|hh'_{<i})$ such that
$$\Mmu^{w}_\tau(x_i'|hh'_{<i})\leq \tilde{\Mmu}^{w}_\tau(x_i'|hh'_{<i})\leq \Mmu^{w}_\tau(x_i'|hh'_{<i})+\epsilon'\,.$$
We can get this using at most $\lceil -\log_2\epsilon'\rceil$ oracle requests for every $1\leq i\leq t'$, and hence we make at most $t'\lceil -\log_2\epsilon'\rceil$ oracle requests in total, which computably depends only on $\epsilon$ and $t'$.

We claim that if $\epsilon'$ is sufficiently small, then
$$\tilde{\Mmu}^{w}_\tau(h'|h):=\prod_{i=1}^{t'}\tilde{\Mmu}^{w}_\tau(x'_i|hh'_{<i})$$
yields the desired $\epsilon$-approximation of $\Mmu^{w}_\tau(h'|h)$. Indeed:
\begin{itemize}
    \item If $\epsilon'\leq\frac{\epsilon}2$ and if there exists some $1\leq i_0\leq t'$ such that $\Mmu^{w}_\tau(x'_{i_0}|hh'_{<i_0})\leq\frac{\epsilon}{2}$, then
    $$0\leq\Mmu^{w}_\tau(h'|h)=\prod_{i=1}^{t'}\Mmu^{w}_\tau(x'_i|hh'_{<i})\leq \Mmu^{w}_\tau(x'_{i_0}|hh'_{<i_0})\leq\frac{\epsilon}{2}\leq\epsilon\,,$$
    and
    $$0\leq\tilde{\Mmu}^{w}_\tau(h'|h)=\prod_{i=1}^{t'}\tilde{\Mmu}^{w}_\tau(x'_i|hh'_{<i})\leq \tilde{\Mmu}^{w}_\tau(x'_{i_0}|hh'_{<i_0})\leq \Mmu^{w}_\tau(x'_{i_0}|hh'_{<i_0})+\epsilon'\leq\frac{\epsilon}2+\frac{\epsilon}{2}=\epsilon\,,$$
    which imply that
    $$|\tilde{\Mmu}^{w}_\tau(h'|h)-\Mmu^{w}_\tau(h'|h)|\leq\epsilon\,.$$
    \item If $\Mmu^{w}_\tau(x'_{i_0}|hh'_{<i_0})>\frac{\epsilon}{2}$ for all $1\leq i\leq t'$, then
    \begin{align*}
        \Mmu^{w}_\tau(x_i'|hh'_{<i})&\leq \tilde{\Mmu}^{w}_\tau(x_i'|hh'_{<i})\leq \Mmu^{w}_\tau(x_i'|hh'_{<i})\left(1+\frac{\epsilon'}{\Mmu^{w}_\tau(x_i'|hh'_{<i})}\right) \leq \Mmu^{w}_\tau(x_i'|hh'_{<i})\left(1+\frac{2\epsilon'}{\epsilon}\right)\\
        &\leq \Mmu^{w}_\tau(x_i'|hh'_{<i})\exp\left(\frac{2\epsilon'}{\epsilon}\right)\,,\quad\forall i\in\{1,\ldots,t'\}\,,
    \end{align*}
    where in the last inequality we used the fact that $1+z\leq \exp(z)$ for all $z\in\mathbb{R}$. Therefore,
    \begin{align*}
    \Mmu^{w}_\tau(h'|h)&=\prod_{i=1}^{t'}\Mmu^{w}_\tau(x'_i|hh'_{<i})\leq \prod_{i=1}^{t'}\tilde{\Mmu}^{w}_\tau(x'_i|hh'_{<i})=\tilde{\Mmu}^{w}_\tau(h'|h)\\
    &=\prod_{i=1}^{t'}\tilde{\Mmu}^{w}_\tau(x_i'|hh'_{<i})\leq \prod_{i=1}^{t'}\Mmu^{w}_\tau(x_i'|hh'_{<i})\exp\left(\frac{2\epsilon'}{\epsilon}\right)\\
    &=\Mmu^{w}_\tau(h'|h)\exp\left(\frac{2t'\epsilon'}{\epsilon}\right)= \Mmu^{w}_\tau(h'|h)\left(1+\exp\left(\frac{2t'\epsilon'}{\epsilon}\right)-1\right)\\
    &\leq \Mmu^{w}_\tau(h'|h)+\left(\exp\left(\frac{2t'\epsilon'}{\epsilon}\right)-1\right)\,,
    \end{align*}
    and hence it suffices to choose $\epsilon'$ so that $\exp\left(\frac{2t'\epsilon'}{\epsilon}\right)-1=\epsilon$.
\end{itemize}
By choosing $\epsilon'=\frac{\epsilon\ln(1+\epsilon)}{2t'}$, we get both $\epsilon'\leq\frac{\epsilon}{2}$ and $\exp\left(\frac{2t'\epsilon'}{\epsilon}\right)-1=\epsilon$, which completes the proof of \cref{lem:approximate_universal_inductor}.

\subsection{Proof of \cref{thm:rui}}
\label{app:thm_rui_proof}

Let us first analyze, abstractly, the $w$-reflective universal inductor oracle ($w$-RUI-oracle) whose existence we want to show. More precisely, let us analyze the relationship between the $w$-RUI-oracle and its corresponding $w$-reflective universal inductor ($w$-RUI): The inductor reacts, i.e., is a consequence of, the oracle that we use. Hence, we need to show the existence of an oracle such that its answers to queries reflect what this inductor -- that depends on the oracle -- would compute. Thus, in some sense we need some kind of fixpoint in a space containing oracles and inductors. This is what we will do. Let us start by defining appropriate spaces. 
Let
$$\Omega_{\textrm{q}}=[0,1]^{\calB\times ([0,1]\cap\mathbb{Q})\times\calB^*}\,,$$
and
$$\Omega_{\textrm{e}}=[0,1]^{\calB\times\calB^*}\,$$
and note the following:
\begin{itemize}
    \item Each element $\mathrm{query}\in\Omega_{\textrm{q}}$ corresponds to a function $\mathrm{query}:\calB\times ([0,1]\cap\mathbb{Q})\times\calB^*\to[0,1]$ which in turn gives rise to a probabilistic oracle $\tau_\mathrm{query}:\{0,1\}^*\to[0,1]$ as follows: Given $x\in\{0,1\}^*$ encoding a triplet $(b,p,h)\in\calB\times(\mathbb{Q}\cap[0,1])\times\calB^*$, i.e., $x=\langle b,p,h\rangle$, we let $\tau_\mathrm{query}(x)=\mathrm{query}(x)$.\footnote{Note that in \cref{def:rui-oracle}, we impose that the encoding $\langle b,p,h \rangle $ is complete, i.e., that all bitstrings $x \in \calB^\ast$ are a valid encoding of a triplet $(b,p,h)$ }
    \item Similarly, each element $\mathrm{eval}\in\Omega_{\textrm{e}}$ corresponds to a function $\mathrm{eval}:\calB\times\calB^*\to[0,1]$ which can be interpreted as a characterization of an inductor/predictor. More precisely, for every history $h\in\calB^*$ and every $b\in\calB$, $\mathrm{eval}(b,h)$ can be seen as the probability of predicting that the next token is $b$ given that the history of observations so far is $h$.
\end{itemize}
We will work with the space $X = \Omega_{\textrm{q}} \times \Omega_{\textrm{e}}$ and define a mapping $\phi$ that maps each element of $X$ to a {\em subset} of $X$. The mapping $\phi$ will be defined in a way that reflects the properties of oracles and inductors. Once we have done this we can apply a classical fixed point theorem to deduce the existence of an $x^* \in X$ that satisfies $x^* \in\phi(x^*)$. This will correspond to a $w$-RUI-oracle.

So, let us now define 
$$\phi:\Omega_{\textrm{q}}\times \Omega_{\textrm{e}}\to \mathrm{Pow}(\Omega_{\textrm{q}}\times \Omega_{\textrm{e}})\,,$$
from $\Omega_{\textrm{q}}\times \Omega_{\textrm{e}}$ to its own power set (i.e., the set of subsets of $\Omega_{\textrm{q}}\times \Omega_{\textrm{e}}$), such that $$(\mathrm{query}',\mathrm{eval}')\in\phi(\mathrm{query}, \mathrm{eval})$$ if and only if
\begin{enumerate}
    \item[(i)] For every $(b,p,h)\in \calB\times ([0,1]\cap\mathbb{Q})\times\calB^*$, we have
    $$\mathrm{eval}(b,h)>p\quad\Rightarrow\quad \mathrm{query}'(b,p,h)=1\,,$$
    and
    $$\mathrm{eval}(b,h)<p\quad\Rightarrow\quad \mathrm{query}'(b,p,h)=0\,.$$
    \item[(ii)] For every $(b,h)\in \calB\times\calB^*$, we have
    $$\eval'(b,h) = \Mmu^w_{\tau_\query}(b|h)= \frac{\sum_{M\in\calMrapom} w_M \lambda_{M}^{\tau_{\mathrm{query}}}(hb)}{\sum_{M\in\calMrapom} w_M\lambda_{M}^{\tau_{\mathrm{query}}}(h)}\,,$$
    where $\calMrapom$ is the space of all restricted abstract probabilistic oracle machines and $w\in\Delta'\calMrapom$ is the lower semicomputable universal prior for which we would like to show that a $w$-RUI-oracle exists.
\end{enumerate}
Let us first check that a fixed point of $\phi$, i.e., a pair $(\mathrm{query}, \mathrm{eval})\in\Omega_{\textrm{q}}\times\Omega_{\textrm{e}}$ satisfying $(\mathrm{query}, \mathrm{eval})\in \phi(\mathrm{query}, \mathrm{eval})$, indeed gives rise to a RUI-oracle $\tau_\query$ and its corresponding reflective universal inductor $\eval=\Mmu^w_{\tau_\query}$. This actually follows directly from the definition of the $w$-RUI-oracle in Definition~\ref{def:rui-oracle} and the above definition of $\phi$.

In order to prove the existence of the desired fixed point we will employ an infinite dimensional version of the classical Kakutani's fixed point theorem:
\begin{theorem}[\citet{fan1952fixed}]\label{theorem:kakutani_infinite}
Let $X=[0,1]^{T}$ be the set of functions from $T$ to $[0,1]$, where $T$ is a countable set, and let $\phi: X \rightarrow \mathrm{Pow}(X)$ be a set-valued function for which the following holds:
\begin{enumerate}
\item[(1)] For all $x \in X$ the set $\phi(x)$ is nonempty and convex.
\item[(2)] The graph of $\phi$ (i.e., the set $\{(x,y)\in X^2:y\in\phi(x)\}$) is closed, i.e., for every sequence $(x_n,y_n)_{n\in\mathbb{N}}$ for which $y_n\in\phi(x_n)$ for every $n\in\mathbb{N}$, if $\lim_{n\to\infty}x_n=x$ and $\lim_{n\to\infty}y_n=y$ (where the convergence is taken to be the pointwise convergence\footnote{Recall that for a sequence $(x_n)_n$ in $X=[0,1]^T$, $(x_n)_n$ converges pointwise to $x\in X$ if $\lim_{n\to\infty}x_n(t)=x(t)$ for every $t\in T$.} in $[0,1]^T$) then $y\in \phi(x)$.
\end{enumerate}
Then there exists a fixed point $x^*$ of $\phi$.
\end{theorem}

We already argued above that in our case $X=\Omega_{\textrm{q}} \times \Omega_{\textrm{e}}$, which can be rewritten as $[0,1]^T$ with $$T=\big(\calB\times(\mathbb{Q}\cap[0,1])\times\calB^*\big)\cup \big(\calB\times\calB^*\big)\,,$$ which satisfies the required countability requirement in \cref{theorem:kakutani_infinite}. We also easily check that our definition of $\phi$ implies that $\phi(\mathrm{query}, \mathrm{eval})$ is always non-empty and convex. So Condition (1)\ is trivially satisfied. 

It remains to show that the graph of $\phi$ is closed. In our notations we can rewrite the closedness condition (as defined in \cref{theorem:kakutani_infinite}) as follows. Suppose we have $$(\query_n,\eval_n,\query_n',\eval_n')\in\Omega_{\mathrm{q}}\times\Omega_{\mathrm{e}}\times \Omega_{\mathrm{q}}\times\Omega_{\mathrm{e}}$$ such that $$(\query_n',\eval_n')\in\phi(\query_n,\eval_n)\,,\quad\forall n\in\mathbb{N}\,,$$ and assume that
$$\lim_{n\to\infty}(\query_n,\eval_n,\query_n',\eval_n')=(\query,\eval,\query',\eval')\,.$$

We need to show that $$(\query',\eval')\in\phi(\query,\eval)\,,$$ i.e., we need to show that these functions satisfy Conditions (i) and (ii) from above. Showing Condition (i) is  straightforward: Let $(b,p,h)\in \calB\times ([0,1]\cap\mathbb{Q})\times\calB^*$ and assume first that $\mathrm{eval}(b,h)>p$. As $\eval_n$ converges to $\eval$ pointwise, we have $\lim_{n\to\infty}\eval_n(b,h)=\eval(b,h)>p$, and hence there must exist $n_0>0$ such that for all $n>n_0$, we have $\eval_n(b,h)>p$. As $(\query'_n,\eval'_n)\in\phi(\query_n,\eval_n)$, it follows that $\query'_n(b,p,h)=1$, for all $n>n_0$, implying that $\query'(b,p,h)=1$ as well, as desired. The case $\mathrm{eval}(b,h)<p$ follows similarly. If $\eval(b,h) = p$, there is no constraint on $\query'(b,p,h)$ that we need to check, so (i) is trivially satisfied in this case as well. 

In order to show that Condition (ii) is satisfied, it suffices to show that for every $h\in\calB^*$ and every $b\in\calB$, we have
$$\query\mapsto \Mmu^w_{\tau_\query}(b|h)= \frac{\sum_{M\in\calMrapom} w_M \lambda_{M}^{\tau_{\mathrm{query}}}(hb)}{ \sum_{M\in\calMrapom} w_M \lambda_{M}^{\tau_{\mathrm{query}}}(h)}$$
is continuous (as a function of $\query$), where continuity is defined using the pointwise convergence\footnote{Equivalently, we endow $\Omega_{\textrm{q}}=[0,1]^{\calB\times ([0,1]\cap\mathbb{Q})\times\calB^*}$ with the product topology and define the continuity of $\query\mapsto \Mmu^w_{\tau_\query}(b|h)$ based on this topology.} as follows: We require that if $\lim_{n\to\infty}\query_n = \query$ pointwise, then $\lim_{n\to\infty}\Mmu^w_{\tau_{\query_n}}(b|h)=\Mmu^w_{\tau_{\query}}(b|h)$.

Define, for all $h\in{\cal B}^*$, 
$$F_h(\query) := \sum_{M\in\calMrapom} w_M \lambda_{M}^{\tau_{\mathrm{query}}}(h) .$$ 
Then $\Mmu^w_{\tau_\query}(b|h) = F_{hb}(\query)/F_h(\query)$. As
both $h$ and $hb$ are members of $\calB^*$, it thus suffices to prove that $\query\mapsto F_h(\query)$ is continuous for every $h\in{\cal B}^*$, as the quotient of two non-zero continuous functions is also continuous. (Here we used that $F_h(\query)$ is always non-zero, which can be seen by considering a restricted abstract probabilistic oracle machine $M_h$ which hard codes $h$ in its definition and writes it with probability one.) 

To show that $\query\mapsto F_h(\query)$ is continuous, we first observe that it actually suffices to show that $\lambda_{M}^{\tau_{\mathrm{query}}}(h)$ is continuous (as a function of $\query$) for arbitrary $h$ and $M$. To see this, recall the following fact from analysis (that we will use repeatedly throughout the remainder of the proof): If $(a_n)$ is a sequence of non-negative real numbers such that $\sum_{n\ge 1} a_n$ is bounded (by one or any other constant), then for every $\eps >0$ there exists $n_\eps\in\mathbb{N}$ such that $\sum_{n > n_\eps} a_n\le \eps$. This fact can be used to show that if $(f_n)_n$ is a sequence of continuous functions from an arbitrary topological space $X$ to $[0,1]$ then $\sum_{n}a_n f_n$ is also continuous.\footnote{We essentially use the following argument: For every $x_0\in X$ and every $\eps>0$ we use the continuity of the functions $(f_n)_{n\leq n_\eps}$ to find a neighborhood of $x_0$ for which $\sum_{n\leq n_\eps}a_n f_n(x)$ is at most $\eps$ away from $\sum_{n\leq n_\eps}a_n f_n(x_0)$ for every $x$ in that neighborhood. This means that $\sum_{n}a_n f_n(x)$ is at most $3\eps$ away from $\sum_{n}a_n f_n(x_0)$.} Now since $\calMrapom$ is countable and $\sum_M w_M\leq 1$ we can now see that if $\query\mapsto \lambda_{M}^{\tau_{\mathrm{query}}}(h)$ is continuous for every $h\in\calB^*$, then $\query\mapsto F_h(\query)$ is also continuous.

Thus all that remains to be seen is that for arbitrary but fixed $h\in{\cal B}^*$ and $M\in{\calMrapom}$ we have that $\lambda_{M}^{\tau_{\mathrm{query}}}(h)$ is continuous (as a function in $\query$).
Towards this end, recall that we may view the execution of $M$ as a, potentially infinite, rooted tree as follows. The root of the tree corresponds to the initial state and each edge corresponds to the transition from one internal state to the next. If $M$  never queries the oracle nor uses the randomness tape, we get a path. If it does, we get a tree. In this tree each node in which we query the oracle (resp. the randomness tape) has two successors, one for getting $0$ as an answer and one for  getting $1$. We assign the probability of getting this answer as a label to the corresponding edge.  For completeness, we label all deterministic edges with $1$. As we are only interested in states with $h$ on the output tape, we may remove all nodes from the tree that do not contain $h$ or a prefix of $h$ on the output tape (recall that the  output tape is written from left to right, thus once it contains a non-prefix of $h$, it will never contain a string starting with $h$). Similarly, we remove all nodes that are (strict) descendants of some node with {\em exactly} $h$ on the output tape. Finally, we remove all nodes for which no node in its subtree contains $h$ on its output tape.

Observe that, by construction, the remaining tree satisfies the following:
\begin{itemize}
\item All leaves contain (exactly) $h$ on the output tape.
\item All nodes that have $h$ on the output tape are leaves.
\item The probability that $M$ reaches a given leaf is given by the product of the labels on the path from the root to this leaf. 
\end{itemize}

Observe also that the oracle $\tau$ only influences the labels of the edges of this tree, but not the structure of the tree. We call the resulting tree $\mathcal{T}_{M,h}$ and we denote its leaves as $\mathcal{L}_{M,h}$. We then have 
$$\lambda_{M}^{\tau_{\mathrm{query}}}(h)=\sum_{v\in\mathcal{L}_{M,h}} w_{M,\query}(v)\,,$$
where $w_{M,\query}(v)$ denotes the product of the labels along the path from the root to $v$ in the tree $\mathcal{T}_{M,h}$ if we use the oracle~$\tau_{\mathrm{\query}}$.

Recall that we need to show that $\lambda_{M}^{\tau_{\mathrm{query}}}$ is continuous (as a function of $\mathrm{query}$). Fix some sequence $\left(\mathrm{query}_n\right)_n$ that converges to some arbitrary $\mathrm{query}\in\Omega_{\mathrm{q}}$.

As a first step, observe that whenever $\mathcal{T}_{M,h}$ is finite the statement is trivially true, as then $\lambda_{M}^{\tau_{\mathrm{query}}}$ depends only on finitely many different values,\footnote{Note that for every $v\in\mathcal{T}_{M,h}$, the mapping $\mathrm{query}\mapsto w_{M,\query}(v)$ is a polynomial function of finitely many values $\{\mathrm{query}(x):\text{ }x\in Q_v\}$, where $Q_v$ is the set of oracle requests $x\in\calB^*$ that were made on the computational path from the root node to $v$. Therefore, the mapping $\mathrm{query}\mapsto w_{M,\query}(v)$ is a continuous function in $\mathrm{query}$, and hence the sum of the weights of finitely many nodes is also continuous in $\mathrm{query}$.}  can trivially be bounded uniformly.
To gain some intuition why the generalization to infinite trees is non-trivial, and in particular why we need the slightly awkward looking bound on the number of oracle calls in \cref{def:rapom}, consider the following machine (which does not restrict the number of oracle calls): It starts by sending a request $x_1\in\calB^*$ to the oracle, if the answer is $1$, it writes $h$ and stops. Otherwise, it questions the oracle again with another request $x_2\in\calB^*\setminus\{x_1\}$ and writes $h$ if the answer is $1$. Otherwise, it questions the oracle again with yet another request $x_3\in\calB^*\setminus\{x_1,x_2\}$, etc.\footnote{We assume that the mapping $n\mapsto x_n$ is computable. For example, we can let $x_n$ be the binary representation of $n$.} Let $p_i:=\tau_{\mathrm{query}}(x_i)$ denote the probability that the $i$th call to the oracle results in a $1$. Then, by construction, we have that the probability that $M$ writes $h$ and stops is given by $\lambda_M := \sum_{i\ge 1} p_i\cdot \prod_{1\le j < i}(1-p_j)$. Now consider different versions on how to set the $p_i$. Fix some $c>0$ arbitrarily and define $p_i^{(n)} = c/n$ if $i\leq n$ and   $p_i^{(n)}=0$ otherwise. Clearly, $p_i^{(n)} \xrightarrow{\raisebox{-0.8ex}{$\scriptscriptstyle n \to \infty$}} 0$ for all $i\in\mathbb{N}$, independently of $c$. However, $\lambda_M^{(n)} = \sum_{i=0}^{n-1} \frac{c}n \cdot (1-c/n)^i = 1-(1-c/n)^n
\xrightarrow{\raisebox{-0.8ex}{$\scriptscriptstyle n \to \infty$}} 1-e^{-c}$.  That is, different values of $c$ give us different halting probabilities, even though we converge locally to the same value on all edges of the tree.

\def\unif{{\ensuremath \text{unif}}}
Next, let us try to understand what the bound in \cref{def:rapom} actually means. For that let us first consider the special case $h=b$ for some $b\in{\cal B}$, i.e., we have only one stage. As our $M$ is randomized, the machine can still use an arbitrarily large number of calls to the oracle, but they will have to come in a somewhat structured way. To gain some intuition let us consider the following example. Assume the machine queries the randomness tape until it sees $1$, and then decides on $4^i$ as a bound if it got a $1$ in the $i$th trial. The important point (and difference to our counterexample from above) is that the machine has to decide on the bound {\em before} its first call to the oracle.\footnote{As a side remark note that in our current example we do not even have a bounded {\em expected} number of calls to the oracle, i.e., we do {\em not} need a bounded expected number of oracle calls per stage.} So let us see now how we can use this fact. As a first step we color all nodes in  $\mathcal{T}_{M,h}$ {\em orange} iff they correspond to the first oracle call (on the path from the root to some leaf). Let ${\cal O}$ be the set of orange nodes in $\mathcal{T}_{M,h}$ and, similarly to the leafs, denote for each $v \in {\cal O}$ by $w_v$ the product of the labels on the path from the root to $v$. Note that neither the set ${\cal O}$ nor the weights $w_v$ of orange nodes depends on the query function.
Note also that regardless of which query function $\query\in \Omega_q$ we use in the remainder of the tree, we always have for all $v \in {\cal O}$ that the sum of the weights of all leaves in the subtree of $v$ can never be larger than $w_v$. We can use this fact to approximate $\lambda_{M}^{\tau_{\mathrm{query}}}(h)$ as follows. Suppose we only sum up the weights of all orange nodes that have depth at most $d$ (here {\em depth} of a vertex $v$ denotes the number of edges of the (unique) path from $v$ to the root of the tree). Clearly, there are only a finite number of such nodes. Also, if we make $d$ larger, we get a better and better approximation of the sum of the weights of all orange nodes. Thus, for every $\eps>0$ there exists a $d_\eps$ such that the sum of the weights of all orange nodes of depth larger than $d_\eps$ is at most $\eps$. As a consequence we get: The sum of the weights of all leaves that are in a subtree of orange nodes of depth larger than $d$ is at most $\eps$. With this we have taken care of all but a finite number of orange nodes. Denote this finite set by ${\cal O}_{d_\eps}$. 

While, by construction, the number of nodes in ${\cal O}_{d_\eps}$ is finite, their subtrees can still contain an infinite number of leaves. To also reduce this number to a finite number, we repeat the trick from above. To do so, we need to handle one additional issue: While the weight of an orange node, by the definition of "orange", is independent of $\tau_{\query}$, this is not the case for the leaves. Consider the oracle $\unif$ that just returns $0/1$ uniformly at random. Observe that for any other function $\query\in\Omega_q$, the weight of any leaf $v'$ in the subtree of some orange vertex $v$ satisfies
$w_{\query}(v') \le 2^{N_v}\cdot w_{\unif}(v')$, where $N_v$ is the bound on the number of oracle calls in the subtree of $v$ (recall that this bound exists according to \cref{def:rapom}, cf. also our discussion above). With this observation at hand we can now proceed as follows.

Fix some orange node $v\in {\cal O}_{d_\eps}$ and consider the tree with respect to $\unif$. Then the sum of the weights of all leaves in the subtree of $v$ that have (total) depth at most $d_\eps + d$ will form, for larger and larger $d$, an ever more precise approximation of the sum over all leaves in the subtree of $v$. Thus, there exists a $d_v$ such that the sum over all leaves of depth larger than $d_v$ is at most $\eps/(2^{N_v}|{\cal O}_{d_\eps}|)$. Note that this implies that if we change the query function from $\unif$ to an arbitrary $\query\in\Omega_q$, we still know that the sum over all leaves of depth larger than $d_v$ is at most $\eps/|{\cal O}_{d_\eps}|$. Now set
$d_T:= \max_{v\in{\cal O}_{d_\eps}} d_v$ and observe that the factor $|{\cal O}_{d_\eps}|$ allows us to conclude that regardless of which oracle  $\query\in\Omega_q$ we consider, the sum of the weights of all leaves of depth larger than $d_T$ is at most $2\eps$ (one $\eps$ for the leaves in subtrees of orange nodes outside of ${\cal O}_{d_\eps}$ and one for those inside). 

As a final step observe that the sum of all leaves of depth at most $d_T$ contains only finitely many different queries to the oracle (as we have a finite tree). Thus, if $\lim_{n\to\infty}\query_n=\query$ pointwise, then for every $\eps>0$ there exists an $n_0\in\mathbb{N}$ such that for all $n\ge n_0$  the sums of the weights of the leaves up to depth $d_T$ with respect to $\query_n$ and $\query$ differ by at most $\eps$. In summary, we have thus shown: For every $\eps >0$ there exists  $n_0\in\mathbb{N}$ such that for all $n\ge n_0$ we have
$$
|\lambda_M^{\tau_{\query_n}}(h) - \lambda_M^{\tau_{\query}}(h)| \le 3\eps\,.
$$
This concludes the proof that $\query\mapsto \lambda_M^{\tau_{\query}}(h)$ is continuous for $h=b\in\calB$. 

To handle the general case $h=b_1\ldots b_\ell$ we will apply the above argument recursively $\ell$ times. Each stage will contribute a term of $2\eps$ for ignoring leaves of stage $i$ that are too deep. There are two important points that allow us to argue like that. Firstly, the fact that the output tape may only be written from left to right, ensures that in order to reach a state with $h$ on the tape, we have to pass first through a node with $b_1$ on it, then with $b_1b_2$ on it, etc. Secondly, we recall that the sum of all weights of leaves in the subtree of some vertex can never be larger than the weight of the root of this subtree. Note that this implies that if, for example, we consider all leaves in $\mathcal{T}_{M,h}$ that are in subtrees of leaves in  $\mathcal{T}_{M,b_1}$ that are contained in one of the two "$\eps$-parts", the total sum of such leaves in $\mathcal{T}_{M,h}$ cannot be larger than $2\eps$. In other words, we may well ignore all such leaves, as they have already been taken care of. Rephrased again, this means that for the second stage we only have to consider leaves of $\mathcal{T}_{M,b_1b_2}$ which are descendants of leaves of $\mathcal{T}_{M,b_1}$  which are at depth at most $d_{T_1}=d_T(\mathcal{T}_{M,b_1})$. And here comes the trick. For any of these leaves, say $v\in\mathcal{L}_{M,b_1}$ (of depth at most $d_{T_1}$), we can design a new machine $M_v$ that, effectively, just starts in the state of $M$ corresponding to $v$. Thus, we can repeat all of our arguments from above (for the case $h=b$) and conclude that there exists a depth $d_v$ such that the sum of the weights of all leaves in  $\mathcal{T}_{M,b_1b_2}$ that are contained in the subtree of $v$ and have depth larger than $d_v$ is bounded by $\eps/K_1$, where $K_1$ is the number of leaves in  $\mathcal{T}_{M,b_1}$ of depth at most $d_{T_1}$. Clearly, this implies that for 
$d_{T_2} := \max d_v$ we have that the sum of the weights of all leaves in $\mathcal{T}_{M,b_1b_2}$ of depth larger than $d_{T_2}$ is bounded by $4\eps$. Proceeding in the same way for all stages we thus get: There exists a $d_{T_\ell}$ such that the sum of weights of all leaves in $\mathcal{T}_{M,b_1\ldots b_\ell}$ of depth larger than $d_{T_\ell}$ is bounded by $2\ell\eps$. Similarly as for the case $h=b$ above this thus allows us to conclude: If $\lim_{n\to\infty}\query_n=\query$ pointwise, then for every $\eps>0$ there exists  $n_0\in\mathbb{N}$ such that for all $n\ge n_0$ we have
$$
|\lambda_M^{\tau_{\query_n}}(h) - \lambda_M^{\tau_{\query}}(h)| \le (2\ell+1)\eps,
$$
which shows that $\query\mapsto \lambda_M^{\tau_{\query}}(h)$ is continuous, as desired.

\subsection{Proof that \eqref{eq:universal-inductor-ro-machine-recursive} implies \eqref{eq:universal-inductor-ro-machine}}\label{app:proof-universal-inductor-ro-machine-recursive}

First, we show by induction on $t\geq 0$ that
\begin{equation}
    \label{eq:direct-non-recursive-weight-ro}
    w(M|h_{1:t})=w(M) \frac{\bar{\Mvu}_M^\tau(h_{1:t})}{\bar{\Mvu}^\tau_{\MwRO}(h_{1:t})}\,.
\end{equation}
\begin{itemize}
    \item For $t=0$, we have $h_{1:t}=\varepsilon$ by convention. Therefore, it follows from the third equality of \eqref{eq:universal-inductor-ro-machine-recursive} that $w(M|h_{1:t})=w(M|\varepsilon)=w(M)$. Now since $\bar{\Mvu}_M^\tau(\varepsilon)=\bar{\Mvu}_{\MwRO}^\tau(\varepsilon)=1$, we can see that \eqref{eq:direct-non-recursive-weight-ro} holds for $t=0$.
    \item For $t>0$, if \eqref{eq:direct-non-recursive-weight-ro} holds for $t-1$, then
    $$w(M|h_{<t})=w(M) \frac{\bar{\Mvu}_M^\tau(h_{<t})}{\bar{\Mvu}^\tau_{\MwRO}(h_{<t})}\,,$$
    and hence from the second equality of \eqref{eq:universal-inductor-ro-machine-recursive} we get
    $$w(M|h_{1: t}) = w(M|h_{<t}) \frac{\bar{\Mvu}_M^\tau(h_t|h_{<t})}{\bar{\Mvu}^\tau_{\MwRO}(h_t|h_{<t})}=w(M) \frac{\bar{\Mvu}_M^\tau(h_{<t})}{\bar{\Mvu}^\tau_{\MwRO}(h_{<t})}\frac{\bar{\Mvu}_M^\tau(h_t|h_{<t})}{\bar{\Mvu}^\tau_{\MwRO}(h_t|h_{<t})}=w(M) \frac{\bar{\Mvu}_M^\tau(h_{1:t})}{\bar{\Mvu}^\tau_{\MwRO}(h_{1:t})}\,.$$
\end{itemize}
We conclude that \eqref{eq:direct-non-recursive-weight-ro} holds for all $t\geq 0$. Now we have

\begin{align*}
    \bar{\Mvu}^\tau_{\MwRO}(h_{1:t})&:= \prod_{k=1}^t  \bar{\Mvu}^\tau_{\MwRO}(h_k \mid h_{<k}) \\
    &= \bar{\Mvu}^\tau_{\MwRO}(h_{<t})\bar{\Mvu}^\tau_{\MwRO}(h_t \mid h_{<t}) \\
    &\geq \bar{\Mvu}^\tau_{\MwRO}(h_{<t})\Mvu^\tau_{\MwRO}(h_t \mid h_{<t}) \\
    &\stackrel{(\ast)}{\geq} \bar{\Mvu}^\tau_{\MwRO}(h_{<t})\sum_{M\in \calMapom}w(M|h_{<t}) \bar{\Mvu}_M^\tau(h_t|h_{<t}) \\
    &\stackrel{(\dagger)}{=}\bar{\Mvu}^\tau_{\MwRO}(h_{<t})\sum_{M\in \calMapom}w(M) \frac{\bar{\Mvu}_M^\tau(h_{<t})}{\bar{\Mvu}^\tau_{\MwRO}(h_{<t})}\bar{\Mvu}_M^\tau(h_t \mid h_{<t}) \\
    &=\sum_{M\in \calMapom}w(M) \bar{\Mvu}_M^\tau(h_{1:t})\,,
\end{align*}
where $(\ast)$ follows from the first inequality of \eqref{eq:universal-inductor-ro-machine-recursive} and $(\dagger)$ follows from \eqref{eq:direct-non-recursive-weight-ro}.

\subsection{Proof of \cref{lemma:tau-lsc-tau-sampled}}\label{app:proof-tau-lsc-tau-sampled}

\begin{proof}
As the conditional semimeasure $\sigma: \calB^* \to \Delta'\calB$ is $\tau$-lower-semicomputable, there exists two $\tau$-POMs $M_0$ and $M_1$, which upon input $h \in \calB^*$ output an infinite sequence of prefix-free encoded rational numbers $\langle q^0_k \rangle$ and $\langle q^1_k \rangle$, respectively, for which $\lim_{k \to \infty} q^b_k = \sigma(h)(b)$ with $b \in \{0,1\}$. Let us use the notation $\phi_0(h)$ and $\phi_1(h)$ to indicate the iterator functions streaming the infinite sequences $(\langle q_k^b\rangle)_k$ for $b=0$ and $b=1$ respectively. We closely follow the proof of Theorem 6 of \citet{wyeth2025limit} (which is similar to the proof technique of  \citet[Lemma 4.3.3]{li2008introduction}) with one minor modification to handle the stochasticity resulting from the probabilistic oracle $\tau$.

The main challenge that a probabilistic oracle introduces for the traditional lower-semicomputable functions $\phi(x,k)$ (cf. \cref{def:lsc}) is that $\tau$-POMs in general have a stochastic output, making it difficult to enforce that the output upon input $(x,k+1)$ is greater or equal to the output upon input $(x,k)$, as those outputs are associated with independent runs of the $\tau$-POM on different inputs and hence can stochastically fluctuate independently of each other. For the proof technique of \citet{wyeth2025limit} to work, it is crucial that the intervals $\phi(x,k+1) - \phi(x,k)$ are all non-negative. 

We solve this challenge by our specialized definition of $\tau$-LSC that instead of making independent calls to a function $f(x,k)$ with increasing values of $k$, we have a single $\tau$-POM that streams an infinite sequence of $(q_k)$. Even though this infinite sequence can be stochastic due to the uncontrollable stochasticity of the probabilistic oracle, the $\tau$-POM can easily enforce that $q_{k+1} \geq q_k$, as the $\tau$-POM has access to $q_k$ when computing $q_{k+1}$, in contrast to the case where $q_k$ and $q_{k+1}$ would be computed by independent calls to some stochastic function $f(x,k)$.

Now we can apply the main proof technique of \citet{wyeth2025limit}, which informally goes as follows. First we partition the interval $[0,1]$ into a list of subintervals of width $\Delta_k^b:= q^b_k-q^b_{k-1}$ labeled by the corresponding bit $b$. Then we sample a random real from $[0,1]$, check which subinterval it is part of and output the label associated with that subinterval. As we have that $\lim_{k\to\infty} q^b_k = \sigma(h)(b)$, and taking $q^b_0:=0$, we have that the sum of subintervals $\sum_{k=1}^\infty \Delta_k^b = \sigma(h)(b)$, and hence random sampling a real from $[0,1]$ and checking which interval it belongs to and outputting the associated label $b$ results in sampling from $\sigma(h)$. Note that since $\sigma(h)\in\Delta'\calB$ is a semiprobability distribution, it might be the case that the sampled real does not belong to any of the subintervals. In this case, the $\tau$-POM $M$ does not halt. We can see that the output distribution $\lambda_M^\tau(b\mid h)$ of $M$ indeed matches the semiprobability $\sigma(h)(b)$.

Algorithm \ref{alg:lsc-tau-sampled} formalizes this construction. The algorithm descriptions follow those of \citet{wyeth2025limit}, with the minimal change of using $\phi_0(h)$ and $\phi_1(h)$--the iterator functions streaming the infinite sequences $(\langle q_k^b\rangle)_k$ for $b=0$ and $b=1$ respectively.

\begin{algorithm}[H]
\caption{append}
\DontPrintSemicolon
\SetKwInput{Input}{Input}
\SetKwInput{Effect}{Effect}
\Input{A list of intervals P, a bit $b \in \mathcal{B}$, and a rational number $\Delta \in \mathbb{Q}$}
\Effect{A subinterval of label $b$ and length $\Delta$ is added to P}
\SetAlgoLined
\uIf{P is empty}{left $\leftarrow$ 0}
\Else{
left $\leftarrow$ P[-1].right
}
P.append(label: $b$, right: left + $\Delta$)\\
\end{algorithm}

\begin{algorithm}[H]
\caption{check}
\DontPrintSemicolon
\SetKwInput{Input}{Input}
\SetKwInput{Effect}{Effect}
\Input{A list of intervals P, and a bitstring $\omega_{1:k}$ of length $k$ interpreted as a number in $\{i2^{-k}:0\leq i\leq 2^{k}-1\}$}
\Effect{If $[\omega_{1:k}, \omega_{1:k}+2^{-k}]$ is a subset of a subinterval within P, return its label $b$}
\SetAlgoLined
left $\leftarrow$ 0\\
\For{($b$, right) $\in$ P}{
    \If{left $\le \omega_{1:k}$ and $\omega_{1:k}+2^{-k} \leq$ right}{
        return $b$\\
    }
    left $\leftarrow$ right\\
}
\end{algorithm}

\begin{algorithm}[H]
\caption{sample}
\label{alg:lsc-tau-sampled}
\DontPrintSemicolon
\SetKwInput{Input}{Input}
\SetKwInput{Require}{Require}
\Input{($\phi_{0}$, $\phi_1$, h)}
\Require{random bitstream $\omega$}
\SetAlgoLined
Output: $b \sim \sigma(h)$\;
Let P be an empty list\;
$\psi_{0} \leftarrow 0$\;
$\psi_{1} \leftarrow 0$\;
\For{$(q^0_k, q^1_k)$ in $\textrm{zip}(\phi_0(h), \phi_1(h))$}{
    \For{ $b \in \{0,1\}$}{
        $\Delta_{b} \leftarrow q_k^b - \psi_{b}$\;
        $\psi_{b} \leftarrow q_k^b$\;
        append(P,$b, \Delta_{b}$)\;
    }
    check(P,$\omega_{1:k}$)\;
}
\end{algorithm}

As noted by \citet{wyeth2025limit}, $\omega_{1:k}$ in this construction is only specified to precision $2^{-k}$. Hence, we need check(P,$\omega_{1:k}$) to only output True when the entire interval $[\omega_{1:k}, \omega_{1:k}+2^{-k}]$ is a subset of some subinterval within P.
\end{proof}

\subsection{Proof of \cref{thrm:ro-ui-sampleable}}\label{app:proof-thrm-ro-ui-sampleable}

\begin{proof}

We will recursively define weight functions $(w_n(M|h))_{n\geq 0}$ and APOMs $(M_n)_{n\geq 0}$ so that they satisfy \eqref{eq:universal-inductor-ro-machine-recursive} on increasingly longer trajectories. More precisely, for every $n\geq 0$ we require that the following desiderata hold:
\begin{enumerate}
    \item $\Mvu_{M_n}^\tau$ is a measure, and hence $\Mvu_{M_n}^\tau=\bar{\Mvu}_{M_n}^\tau$. Furthermore, $\Mvu_{M_n}^\tau(b|h)>0$ for all $h\in\calB^*$ and all $b\in\calB$.
    \item The function $\langle M,h\rangle\mapsto w_n(M|h)$ is $\tau$-LSC and hence there exists a machine $M_{w_n}$ that ``lower-semi-computes'' $w_n$ in the sense of \cref{def:oracle-lsc}.
    \item $w_n(M|h)>0$ for every $M\in\calMapom$ and every $h\in\calB^*$ satisfying $\bar{\Mvu}_M^\tau(h)>0$.
    \item \eqref{eq:universal-inductor-ro-machine-recursive} is satisfied for $w_n$ and $M_n$ for histories up to length $n$:
    \begin{equation}
    \label{eq:universal-inductor-ro-machines-recursive}
\begin{aligned}
    \Mvu^\tau_{M_n}(b|h_{<t}) &\geq \sum_{M\in\calMapom} w_n(M|h_{<t}) \bar{\Mvu}_M^\tau(b|h_{<t})\,, \quad&\forall t\in[n]\\
    w_n(M|h_{1: t}) &= w_n(M|h_{<t}) \frac{\bar{\Mvu}_M^\tau(h_t|h_{<t})}{\bar{\Mvu}^\tau_{M_n}(h_t|h_{<t})}\,,\quad&\forall t\in[n]\\
    w_n(M|\varepsilon)&=w(M)\,,
\end{aligned}
\end{equation}
where $[n]:=\{1,\ldots,n\}$, and we use the convention that $[0]$ is the empty set $\{\}$.
\item $w_n$ and $M_n$ are consistent with their previous versions on shorter histories, i.e., for all $0\leq t\leq n'\leq n$, we have
$$w_n(M|h_{1:t})=w_{n'}(M|h_{1:t})\,,$$
and
$$\Mvu^\tau_{M_n}(b|h_{<t})=\Mvu^\tau_{M_{n'}}(b|h_{<t})\,.$$
\item If $n>0$, we can computationally construct the machines $M_n$ and $M_{w_n}$ from the machines $M_{n-1}$ and $M_{w_{n-1}}$, i.e., there exists a Turing machine that outputs descriptions $\langle M_n\rangle$ and $\langle M_{w_n}\rangle$ when the descriptions $\langle M_{n-1}\rangle$ and $\langle M_{w_{n-1}}\rangle$ of the previous machines are provided as input.
\end{enumerate}

We proceed with the construction recursively as follows:
\begin{itemize}
    \item For $n=0$, we let $w_0(M|h):=w(M)$ for all $h\in\calB^*$ and all $M\in\calMapom$, and we fix $M_0$ to be the machine that always outputs uniform random bits so that $\Mvu_{M_0}^\tau(h)=\bar{\Mvu}_{M_0}^\tau(h)=\frac{1}{2^{l(h)}}>0$. Note that
    \begin{itemize}
        \item By design, $\bar{\Mvu}_{M_0}^\tau$ is a positive measure. Hence, the first desideratum is satisfied.
        \item Since $w$ is LSC, $w_0$ is $\tau$-LSC. Therefore, there exists an APOM $M_{w_0}$ that ``lower-semi-computes'' $w_0$ in the sense of \cref{def:oracle-lsc}. Furthermore, $w_0(M|h)=w(M)>0$ for all $M\in\calMapom$ and all $h\in\calB^*$. Hence, the second and third desiderata are satisfied.
        \item The first and second conditions of
        \eqref{eq:universal-inductor-ro-machines-recursive} trivially hold since $[0]=\{\}$. Now since $w_0(M|h)=w(M)>0$ for all $M\in\calMapom$, we can see that the third condition is also satisfied. Hence, the fourth desideratum is satisfied.
        \item The fifth and sixth desiderata trivially hold for $w_0$ and $M_0$.
    \end{itemize}
    We conclude that all the desiderata hold for $w_0$ and $M_0$.
    \item Now let $n>0$ and assume that we defined $\tau$-LSC weight functions $(w_i)_{i<n}$ and APOMs $(M_i)_{i<n}$ which satisfy the desiderata specified above. Let $(M_{w_i})_{i<n}$ be the machines specified in the second desideratum. We define $w_n$ and two APOMs $M_n'$ and $M_n$ so that:
    \begin{equation}
        \label{eq:universal-inductor-ro-machine-recursive-construction}
    \begin{aligned}
        \Mvu^\tau_{M_n'}(b|h_{<t}) &=\begin{cases} 
        \bar{\Mvu}_{M_{n-1}}^\tau(b|h_{<t})\quad&\text{if } t<n\,,\\
        \\
        \displaystyle\sum_{M\in\calMapom} w_{n-1}(M|h_{<t}) \bar{\Mvu}_M^\tau(b|h_{<t})
        \,, \quad&\text{if } t\geq n\,,\\
        \end{cases}
        \\
        \Mvu^\tau_{M_n}(b|h_{<t})&=\bar{\Mvu}^\tau_{M_n}(b|h_{<t})=\bar{\Mvu}^\tau_{M_n'}(b|h_{<t})\,,\\
        w_n(M|h_{1: t}) &= \begin{cases}
        w_{n-1}(M|h_{1: t})\quad&\text{if }t<n\\
        w_{n-1}(M|h_{<t}) \frac{\bar{\Mvu}_M^\tau(h_t|h_{<t})}{\bar{\Mvu}^\tau_{M_{n}}(h_t|h_{<t})}\quad&\text{if }t\geq n\,,
        \end{cases}
    \end{aligned}
    \end{equation}
    where we use the convention that $h_{1:0}=\varepsilon$. In order to see why we can indeed construct the machines $M_n'$ and $M_n$, note that:
    \begin{itemize}
        \item From \cref{thrm:estimable-completions}, we know that the function $\langle b,h\rangle \mapsto \bar{\Mvu}_{M_{n-1}}^\tau(b|h)$ is $\tau$-LSC.
        \item For all $M\in\calMapom$, the function $\langle b,h\rangle \mapsto \bar{\Mvu}_{M}^\tau(b|h)$ is $\tau$-LSC. Furthermore, since $w_{n-1}$ is $\tau$-LSC, we can see that the function
        $$\langle b,h\rangle \mapsto\sum_{M\in\calMapom} w_{n-1}(M|h) \bar{\Mvu}_M^\tau(b|h) $$
        is $\tau$-LSC.
    \end{itemize}
    Combining the above two observations, we can see that the function
    $$\langle b,h_{<t}\rangle \mapsto\begin{cases} 
    \bar{\Mvu}_{M_{n-1}}^\tau(b|h_{<t})\quad&\text{if } t<n\,,\\
    \\\displaystyle\sum_{M\in\calMapom} w_{n-1}(M|h_{<t}) \bar{\Mvu}_M^\tau(b|h_{<t})
    \,, \quad&\text{if } t\geq n\,,\\
    \end{cases}$$
    is also $\tau$-LSC, and hence from \cref{lemma:tau-lsc-tau-sampled} it is $\tau$-sampleable as well. Therefore, the machine $M_n'$ can be constructed. Furthermore, having access to $\langle M_{n-1}\rangle$ and $\langle M_{w_{n-1}}\rangle$, we can computably\footnote{That is, there exists a Turing machine that can compute $\langle M_n'\rangle$ from $\langle M_{n-1}\rangle$ and $\langle M_{w_{n-1}}\rangle$.} construct the description $\langle M_{n}'\rangle$ of $M_n'$.
    
    Now since $\bar{\Mvu}_{M_n'}^\tau$ is $\tau$-sampleable, we can construct a machine $M_n$ that satisfies $\Mvu_{M_n}^\tau=\bar{\Mvu}_{M_n'}^\tau$, which would mean that $\Mvu_{M_n}^\tau$ is a measure and hence we also have $\Mvu_{M_n}^\tau=\bar{\Mvu}_{M_n}^\tau$. Furthermore, from $\langle M_{n}'\rangle$, we can computably construct $\langle M_{n}\rangle$. In the following, we show that actually all the desiderata are satisfied for $w_n$ and $M_n$:
    \begin{enumerate}
        \item By construction, we made $\bar{\Mvu}_{M_n}^\tau=\Mvu_{M_n}^\tau=\bar{\Mvu}_{M_n'}^\tau$, and hence $\Mvu_{M_n}^\tau$ is a measure. To prove that it is a positive measure, note that for every $h\in\calB^*$, we have $$\Mvu_{M_n}^\tau(h)=\bar{\Mvu}_{M_n'}^\tau(h)\geq \Mvu_{M_n'}^\tau(h)=\prod_{t=1}^{l(h)}\Mvu_{M_n'}^\tau(h_t|h_{<t})\,.$$ Hence, it suffices to show that $\Mvu_{M_n'}^\tau(b|h)>0$ for all $b\in\calB$ and all $h\in\calB^*$. Let $t=l(h)+1$ so that $h=h_{<t}$. We have:
        \begin{itemize}
            \item If $t<n$, we have from \eqref{eq:universal-inductor-ro-machine-recursive-construction} $$\Mvu_{M_n'}^\tau(b|h)=\Mvu_{M_n'}^\tau(b|h_{<t})=\bar{\Mvu}_{M_{n-1}}^\tau(b|h_{<t})\geq \Mvu_{M_{n-1}}^\tau(b|h_{<t})\,,$$ and from the induction hypothesis we know that $\Mvu_{M_{n-1}}^\tau(b|h_{<t})>0$. Hence, $\Mvu_{M_n'}^\tau(b|h)>0$.
            \item If $t\geq n$, we have from \eqref{eq:universal-inductor-ro-machine-recursive-construction}
            $$\Mvu_{M_n'}^\tau(b|h)=\Mvu_{M_n'}^\tau(b|h_{<t})=\sum_{M\in\calMapom} w_{n-1}(M|h_{<t}) \bar{\Mvu}_M^\tau(b|h_{<t})\,.$$
            Now let $M_{hb}$ be an APOM that hard-codes $h=h_{<t}$ and $b$ in its syntax and outputs according to it so that $\Mvu_{M_{hb}}^\tau(h_{<t}b)=1$ and hence $\Mvu_{M_{hb}}^\tau(h_{<t})=1$ and $\Mvu_{M_{hb}}^\tau(b|h_{<t})=1$. We have
            \begin{align*}
                \Mvu_{M_n'}^\tau(b|h)&\geq w_{n-1}({M_{hb}}|h_{<t}) \bar{\Mvu}_{M_{hb}}^\tau(b|h_{<t})\\
                &\geq w_{n-1}({M_{hb}}|h_{<t}) \Mvu_{M_{hb}}^\tau(b|h_{<t}) = w_{n-1}({M_{hb}}|h_{<t})\,.
            \end{align*}
            Now from the induction hypothesis we know that since $\bar{\Mvu}_{M_{hb}}^\tau(h_{<t})\geq\Mvu_{M_{hb}}^\tau(h_{<t})=1>0$, we must have $w_{n-1}({M_{hb}}|h_{<t})>0$. Therefore, $$\Mvu_{M_n'}^\tau(b|h)\geq w_{n-1}({M_{hb}}|h_{<t})>0\,.$$
        \end{itemize}
        We conclude that the first desideratum is satisfied.
        \item From \cref{thrm:estimable-completions}, we know that $\bar{\Mvu}^\tau_{M}$ is both $\tau$-LSC and $\tau$-USC for all $M\in\calMapom$. Furthermore we have just shown that $\bar{\Mvu}^\tau_{M_{n}}(b|h)>0$ for all $(b,h)\in\calB\times\calB^*$. Therefore, the mapping $\langle b,h\rangle \mapsto\frac{1}{\bar{\Mvu}^\tau_{M_{n}}(b|h)}$ is $\tau$-LSC. Now since the mapping $\langle M,h\rangle\mapsto w_{n-1}(M|h)$ is also $\tau$-LSC (from the induction hypothesis), we conclude that the mapping $$(b,h_{1:t})\mapsto w_{n-1}(M|h_{<t}) \frac{\bar{\Mvu}_M^\tau(h_t|h_{<t})}{\bar{\Mvu}^\tau_{M_{n}}(h_t|h_{<t})}$$
        is $\tau$-LSC, and hence $w_n$ is also $\tau$-LSC. Therefore, the second desideratum is satisfied. Furthermore, having access to $\langle M_{w_{n-1}}\rangle$ and $\langle M_{n}\rangle$, we can computably construct the description of $\langle M_{w_n}\rangle$ of a machine $M_{w_n}$ that $\tau$-lower-semi-computes $w_n$ in the sense of \cref{def:oracle-lsc}.
        \item Let $M\in\calMapom$ and $h\in\calB^*$ be such $\bar{\Mvu}_{M}^\tau(h)>0$ and let $t=l(h)$. We have: 
        \begin{itemize}
            \item If $t< n$, then $w_n(M|h)=w_{n-1}(M|h)$ and from the induction hypothesis we know that $w_{n-1}(M|h)>0$. Hence, $w_n(M|h)>0$ .
            \item If $t\geq n$, then
            $$w_n(M|h)=w_n(M|h_{1:t})=w_{n-1}(M|h_{<t})\frac{\bar{\Mvu}_{M}^\tau(h_t|h_{<t})}{\bar{\Mvu}_{M_{n}}^\tau(h_t|h_{<t})}\,.$$
            Now since $\bar{\Mvu}_{M}^\tau(h_{<t})\bar{\Mvu}_{M}^\tau(h_t|h_{<t})=\bar{\Mvu}_{M}^\tau(h_{1:t})=\bar{\Mvu}_{M}^\tau(h)>0$, it follows that we have $\bar{\Mvu}_{M}^\tau(h_{<t})>0$ and $\bar{\Mvu}_{M}^\tau(h_t|h_{<t})>0$. From the induction hypothesis, we know that $\bar{\Mvu}_{M}^\tau(h_{<t})>0$ implies that $w_{n-1}(M|h_{<t})>0$. We also showed (in the proof of the first desideratum) that $\bar{\Mvu}_{M_{n}}^\tau(h_t|h_{<t})>0$. Combining all these facts together, we get $w_n(M|h)>0$.
        \end{itemize}
        We conclude that the third desideratum is satisfied.
        \item In the following, we show that \eqref{eq:universal-inductor-ro-machines-recursive} is satisfied:
        \begin{itemize}
            \item Let $b\in\calB$ and $h\in\calB^*$, and let $t=l(h)+1$ so that $h=h_{<t}$. We have:
            \begin{itemize}
                \item If $t<n$, we have from \eqref{eq:universal-inductor-ro-machine-recursive-construction} that 
                \begin{align*}
                    \Mvu_{M_n}^\tau(b|h_{<t})&=\bar{\Mvu}_{M_n'}^\tau(b|h_{<t})
                    \geq \Mvu_{M_n'}^\tau(b|h_{<t})=\bar{\Mvu}_{M_{n-1}}^\tau(b|h_{<t})\geq \Mvu_{M_{n-1}}^\tau(b|h_{<t})\\
                    &\stackrel{(\ast)}{\geq}\sum_{M\in\calMapom} w_{n-1}(M|h_{<t})\bar{\Mvu}_{M}^\tau(b|h_{<t})=\sum_{M\in\calMapom} w_{n}(M|h_{<t})\bar{\Mvu}_{M}^\tau(b|h_{<t})\,,
                \end{align*}
                where $(\ast)$ follows from the induction hypothesis.
                \item If $t=n$, we have from \eqref{eq:universal-inductor-ro-machine-recursive-construction} that 
                \begin{align*}
                    \Mvu_{M_n}^\tau(b|h_{<t})=\bar{\Mvu}_{M_n'}^\tau(b|h_{<t})
                    \geq \Mvu_{M_n'}^\tau(b|h_{<t})&=\sum_{M\in\calMapom} w_{n-1}(M|h_{<t}) \bar{\Mvu}_M^\tau(b|h_{<t})\\
                    &\stackrel{(\dagger)}{=}\sum_{M\in\calMapom} w_{n}(M|h_{<t})\bar{\Mvu}_{M}^\tau(b|h_{<t})\,,
                \end{align*}
                where $(\dagger)$ follows from the fact that $l(h_{<t})=t-1=n-1<n$.
            \end{itemize}
            Therefore, the first condition of \eqref{eq:universal-inductor-ro-machines-recursive} is satisfied.
            \item Let $h\in\calB^*$ and let $t=l(h)$ so that $h=h_{1:t}$. We have:
            \begin{itemize}
                \item If $t<n$, then it follows from \eqref{eq:universal-inductor-ro-machine-recursive-construction} that
                \begin{align*}
                    w_n(M|h_{1:t})=w_{n-1}(M|h_{1:t})\stackrel{(\ddagger)}= w_{n-1}(M|h_{<t}) \frac{\bar{\Mvu}_M^\tau(h_t|h_{<t})}{\bar{\Mvu}^\tau_{M_{n-1}}(h_t|h_{<t})}=w_n(M|h_{<t}) \frac{\bar{\Mvu}_M^\tau(h_t|h_{<t})}{\bar{\Mvu}^\tau_{M_{n}}(h_t|h_{<t})}\,,
                \end{align*}
                where $(\ddagger)$ follows from the induction hypothesis.
                \item If $t=n$, then from \eqref{eq:universal-inductor-ro-machine-recursive-construction}, we have that
                \begin{align*}
                    w_n(M|h_{1:t})=w_{n-1}(M|h_{<t}) \frac{\bar{\Mvu}_M^\tau(h_t|h_{<t})}{\bar{\Mvu}^\tau_{M_{n}}(h_t|h_{<t})}=w_{n}(M|h_{<t}) \frac{\bar{\Mvu}_M^\tau(h_t|h_{<t})}{\bar{\Mvu}^\tau_{M_{n}}(h_t|h_{<t})}\,,
                \end{align*}
                where the last equality is true because $l(h_{<t})=t-1<n$.
            \end{itemize}
            Therefore, the second condition of \eqref{eq:universal-inductor-ro-machines-recursive} is satisfied.
            \item We have $w_n(M|\varepsilon)=w_{n-1}(M|\varepsilon)$ because $l(\varepsilon)=0<n$, and we have $w_{n-1}(M|\varepsilon)=w(M)$ from the induction hypothesis. Hence, the third condition of \eqref{eq:universal-inductor-ro-machines-recursive} is satisfied.
        \end{itemize}
        We conclude that the fourth desideratum is satisfied.
        \item Let $0\leq t\leq n'\leq n$.
        \begin{itemize}
            \item If $n'=n$, then we obviously have
            $$w_n(M|h_{1:t})=w_{n'}(M|h_{1:t})\quad\text{and}\quad\Mvu^\tau_{M_n}(b|h_{<t})=\Mvu^\tau_{M_{n'}}(b|h_{<t})\,.$$
            \item If $n'<n$, then $n'\leq n-1$. From the induction hypothesis, we know that
            $$w_{n-1}(M|h_{1:t})=w_{n'}(M|h_{1:t})\quad\text{and}\quad\Mvu^\tau_{M_{n-1}}(b|h_{<t})=\Mvu^\tau_{M_{n'}}(b|h_{<t})\,.$$
            On the other hand, since $t\leq n'<n$, we know from \eqref{eq:universal-inductor-ro-machine-recursive-construction} that
            $$w_{n}(M|h_{1:t})=w_{n-1}(M|h_{1:t})\quad\text{and}\quad\Mvu^\tau_{M_{n}}(b|h_{<t})=\bar{\Mvu}^\tau_{M_{n-1}}(b|h_{<t})=\Mvu^\tau_{M_{n-1}}(b|h_{<t})\,.$$
            Therefore,
            $$w_n(M|h_{1:t})=w_{n'}(M|h_{1:t})\quad\text{and}\quad\Mvu^\tau_{M_n}(b|h_{<t})=\Mvu^\tau_{M_{n'}}(b|h_{<t})\,.$$
        \end{itemize}
        We conclude that the fifth desideratum is satisfied.
        \item We have already shown that from $\langle M_{n-1}\rangle$ and $\langle M_{w_{n-1}}\rangle$, we can computably construct $\langle M_{n}'\rangle$, from which we can computably construct $\langle M_{n}\rangle$. We also showed that from $\langle M_{w_{n-1}}\rangle$ and $\langle M_{n}\rangle$, we can computably construct $\langle M_{w_{n}}\rangle$. Therefore, the sixth desideratum is satisfied.
    \end{enumerate}
    We conclude that $(w_n)_{n\geq 0}$ and $(M_n)_{n\geq 0}$ satisfy the six desiderata.
\end{itemize}

We now construct the machine $\MwRO$ based on the recursive definition of the machines $(M_n)_{n\geq 0}$ and $(M_{w_n})_{n\geq 0}$. The machine $\MwRO$ will take an input history $h$, compute the index $n = l(h)$, construct the machine $M_{n+1}$ by recursively applying the process from desideratum 6, and then sample from $M_{n+1}$'s distribution for the history $h$ to produce the next bit $b=h_{n+1}$.

\begin{algorithm}[H]
\caption{The APOM $\MwRO$ satisfying \eqref{eq:universal-inductor-ro-machine-recursive}}
\label{alg:MwRO}
\KwIn{Input history $h \in \calB^*$}
\KwOut{Bit $b$}
\SetAlgoLined
$n \gets l(h)$\;
$\langle M_{\text{current}}\rangle \gets \langle M_0 \rangle$\; \tcp{Initialize with hardcoded description of $M_0$}
$\langle M_{w, \text{current}}\rangle \gets \langle M_{w_0} \rangle$\; \tcp{Initialize with hardcoded description of $M_{w_0}$}
\For{$i \gets 1$ \KwTo $n+1$}{
    $(\langle M_{\text{current}}\rangle, \langle M_{w, \text{current}}\rangle) \gets \text{Construct}(\langle M_{\text{current}}\rangle, \langle M_{w, \text{current}}\rangle)$\; \tcp{Run the computational procedure from desideratum 6}
}
\tcp{At this point, $\langle M_{\text{current}}\rangle = \langle M_{n+1} \rangle$}
$b \gets M_{\text{current}}^\tau(h)\;$ \tcp{Simulate $M_{n+1}^\tau$ on input $h$ and sample one bit $b \sim \Mvu^\tau_{M_{n+1}}(\cdot|h)$}
\KwRet{$b$}\;
\end{algorithm}

We define the APOM $\MwRO$ as the machine described in Algorithm \ref{alg:MwRO}. By its construction, $\MwRO$ is a valid APOM. Its conditional output distribution on input $h$ is, by definition, that of the machine $M_{l(h)+1}$:
$$\Mvu^\tau_{\MwRO}(b|h) := \Mvu^\tau_{M_{l(h)+1}}(b|h)\,.$$
We also define the global weight function $w(M|h)$ based on the length of the history:
$$w(M|h) := w_{l(h)}(M|h)\,.$$
We now show that this machine $\MwRO$ and this weight function $w$ together satisfy the three recursive equations in \eqref{eq:universal-inductor-ro-machine-recursive}:
\begin{enumerate}
    \item We verify the first condition of \eqref{eq:universal-inductor-ro-machine-recursive}. Let $h_{<t}$ be a history. We have
    \begin{align*}
        \Mvu^\tau_{\MwRO}(b|h_{<t}) &\stackrel{(*)}{=} \Mvu^\tau_{M_{l(h_{<t})+1}}(b|h_{<t}) = \Mvu^\tau_{M_t}(b|h_{<t}) \\
        &\stackrel{(\dagger)}{\geq} \sum_{M\in\calMapom} w_{t}(M|h_{<t}) \bar{\Mvu}_M^\tau(b|h_{<t}) \\
        &\stackrel{(\ddagger)}{=} \sum_{M\in\calMapom} w_{t-1}(M|h_{<t}) \bar{\Mvu}_M^\tau(b|h_{<t})\\
        &\stackrel{(*)}{=} \sum_{M\in\calMapom} w(M|h_{<t}) \bar{\Mvu}_M^\tau(b|h_{<t})\,,
    \end{align*}
    where the equalities $(*)$ follow from the definitions of $\MwRO$ and $w(M|h)$, $(\dagger)$ from the fact that $M_t$ satisfies \eqref{eq:universal-inductor-ro-machines-recursive}, and $(\ddagger)$ from the fifth desideratum. Thus, Condition 1 of \eqref{eq:universal-inductor-ro-machine-recursive} holds.
    \item We verify the second condition of \eqref{eq:universal-inductor-ro-machine-recursive}. Let $h_{1:t}$ be a history. We have
    \begin{align*}
    w(M|h_{1:t}) &\stackrel{(*)}{=} w_t(M|h_{1:t}) \\
    &\stackrel{(\dagger)}{=} w_{t-1}(M|h_{<t}) \frac{\bar{\Mvu}_M^\tau(h_t|h_{<t})}{\bar{\Mvu}^\tau_{M_t}(h_t|h_{<t})} \\
    &\stackrel{(*)}{=} w(M|h_{<t}) \frac{\bar{\Mvu}_M^\tau(h_t|h_{<t})}{\Mvu^\tau_{M_{l(h_{<t})+1}}(h_t|h_{<t})} \\
    &\stackrel{(\ast)}= w(M|h_{<t}) \frac{\bar{\Mvu}_M^\tau(h_t|h_{<t})}{\Mvu^\tau_{\MwRO}(h_t|h_{<t})}\,,
\end{align*}
where the equalities $(*)$ follow from the definitions of $w(M|h)$ and $\MwRO$, and $(\dagger)$ from the recursive definition in \eqref{eq:universal-inductor-ro-machine-recursive-construction}. Thus, Condition 2 of  \eqref{eq:universal-inductor-ro-machine-recursive} holds.
    \item Finally, we verify the third condition:
    \begin{align*}
        w(M|\varepsilon) \stackrel{(*)}{=} w_{l(\varepsilon)}(M|\varepsilon) = w_0(M|\varepsilon) \stackrel{(\dagger)}{=} w(M)\,,
    \end{align*}
    where $(*)$ follows from the definition of $w$ and $(\dagger)$ follows from the definition of $w_0$. Thus, Condition 3 of \eqref{eq:universal-inductor-ro-machine-recursive} holds.
\end{enumerate}

Since an APOM $\MwRO$ and a weight function $w$ exist that satisfy all three conditions of \eqref{eq:universal-inductor-ro-machine-recursive}, and we have shown in \cref{app:proof-universal-inductor-ro-machine-recursive} that this implies \eqref{eq:universal-inductor-ro-machine}, the theorem is proven.

\end{proof}

\subsection{Proof of \cref{thm:eaixi-ro}}\label{app:proof-eaixi-ro}
\begin{proof}
    Minor variation upon \citet[Theorem 22]{leike2016formal} and \citet[Theorem 18]{wyeth2025limit}.

    First, let us show that the $k$-step $Q$ values of \eqref{eqn:k-step-qval} are $\tau$-estimable. This can be proved by induction on $k$.

    For one-step $Q$ values ($k=1$), notice from \eqref{eqn:k-step-qval} that for $\Mmu=\Mmu^{RO}$, we have
    \begin{align*}
        Q_{\Mmu}^1(\HistM,a) &= \expect{\Mmu(e_t \mid \HistM,a)}{(1-\gamma)r(e_t) + \gamma V_{\Mmu}(\HistM ae_t)}\\
        &= \sum_{e_t\in\calE} \Mmu(e_t \mid \HistM,a)\left[(1-\gamma)r(e_t) + \gamma V_{\Mmu}(\HistM ae_t)\right]\\
        &=\sum_{e_t\in\calE,\Turn_{t+1}\in\TurnSet} \Mmu(e_t\Turn_{t+1} \mid \HistM,a)\left[(1-\gamma)r(e_t) +(1-\gamma)\gamma r(e_{t+1}) + \gamma^2 V_{\Mmu}(\HistM ae_t\Turn_{t+1})\right]\,.
    \end{align*}
    Similarly, for every $T\geq 1$, we have
    \begin{align*}
        Q_{\Mmu}^1(\HistM,a)&=\sum_{e_t\in\calE,\Turn_{t+1:t+T}\in\TurnSet^T} \Mmu(e_t\Turn_{t+1:t+T} \mid \HistM,a)\\
        &\quad\quad\quad\quad\quad\quad\quad\quad\quad\quad\quad\quad\left[(1-\gamma)\sum_{i=0}^T \gamma^i r(e_{t+i})  + \gamma^{T+1} V_{\Mmu}(\HistM ae_t\Turn_{t+1:t+T})\right]\,.
    \end{align*}
    Now since $0\leq V_{\Mmu}(\HistM ae_ta_{t+1}e_{t+1}\ldots a_{t+T}e_{t+T})\leq 1$, we can see that for every $T\geq 1$,
    $Q_{\Mmu}^1(\HistM,a)$ can be lower-bounded and upper-bounded as
    $$\underline{Q}_{\Mmu,T}(\HistM,a)\leq Q_{\Mmu}^1(\HistM,a)\leq \overline{Q}_{\Mmu,T}(\HistM,a)\,,$$
    where
    \begin{align*}
        \underline{Q}_{\Mmu,T}(\HistM,a)&=(1-\gamma)\sum_{e_t\in\calE,\Turn_{t+1:t+T}\in\TurnSet^T} \Mmu(e_t\Turn_{t+1:t+T} \mid \HistM,a)\sum_{i=0}^T \gamma^i r(e_{t+i})\,,\\
        \overline{Q}_{\Mmu,T}(\HistM,a)&=\sum_{e_t\in\calE,\Turn_{t+1:t+T}\in\TurnSet^T} \Mmu(e_t\Turn_{t+1:t+T} \mid \HistM,a)\left[(1-\gamma)\sum_{i=0}^T \gamma^i r(e_{t+i})+\gamma^{T+1}\right]\,.
    \end{align*}
    Now since $\Mmu$ is $\tau$-estimable, we can see that both $\underline{Q}_{\Mmu,T}(\HistM,a)$ and $\overline{Q}_{\Mmu,T}(\HistM,a)$ are $\tau$-estimable. Furthermore, since
    $$\lim_{T\to\infty}\overline{Q}_{\Mmu,T}(\HistM,a)-\underline{Q}_{\Mmu,T}(\HistM,a)=\lim_{T\to\infty}\gamma^{T+1}=0\,,$$
    we can see that $Q_{\Mmu}^1(\HistM,a)$ is $\tau$-estimable.

    Now for every $k>1$, assume that $Q_{\Mmu}^{k-1}(\HistM,a)$ is $\tau$-estimable, we can see from \eqref{eqn:k-step-qval} that 
    \begin{align*}
         Q_{\Mmu}^k(\HistM, a) &= \expect{\Mmu(e \mid \HistM,a)}{(1-\gamma)r(e) + \gamma\max_{a'}Q^{k-1}_{\Mmu}(\HistM ae,a')}\\
         &=\sum_{e} \Mmu(e \mid \HistM,a)\left[(1-\gamma)r(e) + \gamma\max_{a'}Q^{k-1}_{\Mmu}(\HistM ae,a')\right]\,,
    \end{align*}
    which implies that $Q_{\Mmu}^{k}(\HistM,a)$ is $\tau$-estimable as well. We conclude that for every $k\geq 1$, $Q_{\Mmu}^k(\HistM,a)$ is $\tau$-estimable, and hence also $\tau$-LSC and $\tau$-USC.

    Now since optimal-planning is at least as good as $k$-step planning, we have
    $Q_{\Mmu}^*(\HistM,a)\geq Q_{\Mmu}^k(\HistM,a)$. Furthermore, since optimal planning can bring at most additional expected reward of $\gamma^k$ on top of the returns guaranteed by $k$-steps planning, we can see that $Q_{\Mmu}^*(\HistM,a)\leq Q_{\Mmu}^k(\HistM,a)+\gamma^k$. Therefore, for every $k\geq 1$, we have
    $$Q_{\Mmu}^k(\HistM,a)\leq Q_{\Mmu}^*(\HistM,a)\leq Q_{\Mmu}^k(\HistM,a)+\gamma^k\,,$$
    which shows that $Q_{\Mmu}^*(\HistM,a)$ is also $\tau$-estimable.

    In the remainder of the proof, we show that E-AIXI$^{\mathrm{RO}}$ is implementable on a POM with access to reflective oracle $\tau$. The proof for $k$-E-AIXI$^{\mathrm{RO}}$ is similar and hence we omit it.

    We need to generalize \citet[Theorem 22]{leike2016formal} to incorporate action spaces with $|\calA|>2$. Assume that we have observed history $\HistM$, and we would like to compute $\argmax_{a\in\cal A}Q_{\Mmu}^*(\HistM,a)$. How can we do this using a $\tau$-POM that takes $\HistM$ as input?
    
    Since we have a complete prefix-free encoding of $\calA$, we can construct a binary tree to navigate the prefix-free encoding, with each leaf corresponding to a valid encoding of an action within $\calA$, and the internal nodes representing the common prefix of its descendant leaves. The root node corresponds to the empty string. Choosing an action $a\in\calA$ corresponds to starting from the root node and then descending the tree until reaching a leaf. Each move represents deciding one bit of the code of the action to be taken.
    
    Let us fix a history $\HistM$, and define for each node $n$ in the tree a value $v_n$ as follows:
    \begin{itemize}
        \item Each leaf node has the value $Q_{\Mmu}^*(\HistM,a)$ with $a$ the action encoded by the leaf, which we know to be $\tau$-estimable from the discussion above.
        \item Each parent node has as value the maximum of the values of its two children nodes. Hence its value is the maximum of the values of its descendant leaves, which are all $\tau$-estimable. Therefore, the value of each node is also $\tau$-estimable.
    \end{itemize}
    We can interpret taking the action $\argmax_{a\in\cal A}Q_{\Mmu}^*(\HistM,a)$ as starting on the root node and then descending the tree by recursively going to the child node having the higher value (with ties broken arbitrarily), until reaching a leaf.
    
    Now if we are at a node $n$ which is not a leaf, we will have two children $n_0,n_1$, where $n_0$ (resp. $n_1$) corresponds to appending the bit 0 (resp. 1) to the bit-string of the node $n$. We would like to determine the child node with the maximum value. Notice that $v_{n_1}\geq v_{n_0}$ if and only if $(v_{n_1} - v_{n_0} + 1)/2\geq 1/2$. Now since the values $v_{n_0}$ and $v_{n_1}$ are $\tau$-estimable, the quantity $(v_{n_1} - v_{n_0} + 1)/2\in[0,1]$ is $\tau$-estimable as well, and hence also $\tau$-sampleable, by \cref{lemma:tau-lsc-tau-sampled}. Therefore, there exists a $\tau$-POM $M_n^\tau$ with output distribution satisfying $\Mvu_{M_{n}}^{\tau}(1\mid \HistM) = (v_{n_1} - v_{n_0} + 1)/2$. In summary, we managed to construct for each non-leaf node $n$, a machine $M_n$ such that $\Mvu_{M_{n}}^{\tau}(1\mid \HistM)\geq 1/2$ if and only if $v_{n_1}\geq v_{n_0}$. In particular, we have:
    \begin{itemize}
        \item $O^\tau(\langle M_n,\HistM, \frac{1}{2}\rangle)=1$ implies that $\Mvu_{M_{n}}^{\tau}(1\mid \HistM)\geq 1/2$ and hence $v_{n_1}\geq v_{n_0}$.
        \item $O^\tau(\langle M_n,\HistM, \frac{1}{2}\rangle)=0$ implies that $\Mvu_{M_{n}}^{\tau}(1\mid \HistM)\leq 1/2$ and hence $v_{n_1}\leq v_{n_0}$.
    \end{itemize}
    Therefore, the bit that is returned by $O^\tau(\langle M_n,\HistM, \frac{1}{2}\rangle)$ corresponds to going to the child node with the higher value (with ties broken arbitrarily), which is what we want.
    
    We can now leverage this to construct a single machine $M^*$ implementing an optimal policy as follows: On input $\HistM$, we first set $n$ to the root node, and then as long as $n$ is not a leaf, we repeat the following:
    \begin{enumerate}
        \item $b\leftarrow O^\tau(\langle M_n, \HistM, \frac{1}{2}\rangle)$.
        \item $n\leftarrow n_b$.
    \end{enumerate}
    Once we exit the loop, we get to a leaf node $n$ corresponding to an action $a\in\calA$ with code $\langle a\rangle$, which we would then write on the output tape. This action corresponds to an optimal action maximizing $Q^*(\HistM,a)$.

    Now since ($k$-)E-AIXI$^{\mathrm{RO}}$ agents are implementable on a POM with access to $\tau$, we can see that if such agents are combined with a multi-agent environment~$\MultiAgentMve$ that is implementable on a POM with access to $\tau$, then the resulting POM induces a universe in $\calMuni^{\tau-\mathrm{RO}}$.
\end{proof}

\subsection{Proof of \cref{thm:k_mupi_implementation}}
\label{app:thm_mupi_implementation_proof}
Let $\Mmu=\Mmu^{\mathrm{RUI}}=\Mmu_{\tau-\mathrm{RUI}}^w$. We claim that it is sufficient for our purposes to show that there exists an rPOM with access to a $w$-RUI-oracle $\tau$ which takes as input a binary string $\langle \HistA,a,k,\epsilon\rangle$ encoding $(\HistA,a,k,\epsilon)\in\TurnSet^*\times\mathcal{A}\times\mathbb{N}\times(\mathbb{Q}\cap(0,1))$, it computes an approximation $\tilde{Q}^{k,\epsilon}(\HistA,a)$ which is $\frac{\epsilon}2$-away from $Q^{k}_\Mmu(\HistA,a)$. We can then define the $\Mgp$ policy as the one that returns
$$\argmax_{a'}\tilde{Q}^{k_t,\epsilon_t}(\HistM,a')\,,$$
where ties are broken arbitrarily. Then, we have:
\begin{align*}
    \Mgp(a_t|\HistM)>0
    &\Rightarrow\quad \tilde{Q}^{k_t,\epsilon_t}(\HistM,a_t)=\max_{a'\in\mathcal{A}}\tilde{Q}^{k_t,\epsilon_t}(\HistM,a')\\
    &\Rightarrow\quad Q^{k_t}_\Mmu(\HistM,a_t)+\frac{\epsilon_t}{2}\geq\max_{a'\in\mathcal{A}} Q^{k_t}_\Mmu(\HistM,a')-\frac{\epsilon_t}{2}\\
    &\Rightarrow\quad Q^{k_t}_\Mmu(\HistM,a_t)\geq\max_{a'\in\mathcal{A}} Q_\Mmu^{k_t}(\HistM,a')-\epsilon_t\,,
\end{align*}
as required by \cref{def:approximate-eba}.

Now we turn to showing that we can get an approximation $\tilde{Q}^{k,\epsilon}(\HistM,a_t)$ of $Q^{k}_\Mmu(\HistM, a_t)$ using a rPOM with access to the $w$-RUI-oracle $\tau$, i.e., a $\tau$-POM computation that makes a computably bounded number of requests to the $w$-RUI-oracle $\tau$.

We first show that we can do this for $k=1$. Note that analogously to \eqref{eq:q-value-function-app}, we have
\begin{equation}
    \label{eq:q-value-rho}
    Q_\Mmu^1(\HistM,a_t)=Q_\Mmu(\HistM,a_t)=\lim_{t'\to\infty}(1-\gamma)\sum_{i=t}^{t'}\gamma^{i-t} \sum_{(e_t,\Turn_{t+1:i})\in\mathcal{E}\times\TurnSet^{i-t}}\Mmu(e_t\Turn_{t+1:i}|\HistM a_t)r(e_i)\,.
\end{equation}

Let $t_\eps>0$ be a positive integer that is large enough so that $(1-\gamma)\sum_{i> t_\eps}\gamma^{i}<\frac{\epsilon}{4}$, e.g., $t_\epsilon=\left\lceil \frac{\log(\epsilon/4)}{\log \gamma}\right\rceil$. In this case, we have $(1-\gamma)\sum_{i> t+t_\epsilon}\gamma^{i-t}<\frac{\epsilon}{4}$ for every $t\in\mathbb{N}$. We claim that it is sufficient to obtain for every $a_t\in\mathcal{A}$ and every $\HistM\in \TurnSet^{t-1}$ an approximation $\tilde{Q}^{1,\epsilon}(\HistM, a_t)$ which is at most $\frac{\epsilon}{4}$-away from
\begin{equation}
    \label{eq:truncated-returns}
    (1-\gamma)\sum_{i=t}^{t+t_\epsilon}\gamma^{i-t} \sum_{(e_t,\Turn_{t+1:i})\in\mathcal{E}\times\TurnSet^{i-t}}\Mmu(e_t\Turn_{t+1:i}|\HistM a_t)r(e_i)\,.
\end{equation}
Indeed, if we have this, then we get for all $a_t\in\mathcal{A}$,
\begin{align*}
    |&\tilde{Q}^{1,\epsilon}(\HistM,a_t)-Q_\Mmu^1(\HistM,a_t)|
    \\
    &=\left|\tilde{Q}^{1,\epsilon}(\HistM,a_t)-\lim_{t'\to\infty}(1-\gamma)\sum_{i=t}^{t'}\gamma^{i-t} \sum_{(e_t,\Turn_{t+1:i})\in\mathcal{E}\times\TurnSet^{i-t}}\Mmu(e_t\Turn_{t+1:i}|\HistM a_t)r(e_i)\right|\\
    &\leq \left|\tilde{Q}^{1,\epsilon}(\HistM,a_t)-(1-\gamma)\sum_{i=t}^{t+t_\epsilon}\gamma^{i-t} \sum_{(e_t,\Turn_{t+1:i})\in\mathcal{E}\times\TurnSet^{i-t}}\Mmu(e_t\Turn_{t+1:i}|\HistM a_t)r(e_i)\right|+(1-\gamma)\sum_{i>t+t_\epsilon}\gamma^{i-t}
    \\
    &<\frac{\epsilon}{4}+\frac{\epsilon}{4}\leq \frac{\epsilon}{2}\,.
\end{align*}

Now we show that \eqref{eq:truncated-returns} can indeed be approximated up to an additive error of at most $\frac{\epsilon}{4}$ using a $\tau$-rPOM.

Using \cref{lem:approximate_universal_inductor}, we can get an approximation $\tilde{\Mmu}(e_t \Turn_{t+1:i}|\HistM a_t)$ of $\Mmu(e_t\Turn_{t+1:i}|\HistM a_t)$ for every $(e_t,\Turn_{t+1:i})\in\mathcal{E}\times\TurnSet^{i-t}$ and every $i\in\{t,\ldots,t+t_\epsilon\}$, such that
$$|\tilde{\Mmu}(e_t\Turn_{t+1:i}|\HistM a_t)-{\Mmu}(e_t\Turn_{t+1:i}|\HistM a_t)|\leq \tilde{\epsilon}:=
\frac{\epsilon}{4(t_\epsilon+1)|\mathcal{E}|(|\mathcal{A}|\times |\mathcal{E}|)^{t_\epsilon}}\,,$$
and the number $N_{\epsilon}$ of oracle requests that is needed depends (computably) only on $\epsilon$.\footnote{We can choose $\displaystyle N_\epsilon=\max_{1\leq i\leq t_{\epsilon}+1}\max_{(e_1,\Turn_{2:i})\in\mathcal{E}\times\TurnSet^{i-1}}N_{\epsilon, \langle e_1\Turn_{2:i}\rangle}$, where $N_{\epsilon, h'}$ (for $h':=\langle e_1\Turn_{2:i}\rangle\in\calB^*$) is as in \cref{lem:approximate_universal_inductor}.} By defining
$$\tilde{Q}^{1,\epsilon}(\HistM, a_t):=(1-\gamma)\sum_{i=t}^{t+t_\epsilon}\gamma^{i-t} \sum_{(e_t,\Turn_{t+1:i})\in\mathcal{E}\times\TurnSet^{i-t}}\tilde{\Mmu}(e_t\Turn_{t+1:i}|\HistM a_t)r(e_i)\,,$$
we get our desired approximation. Furthermore, the total number of oracle requests in the above procedure is at most  $\tilde{N}_{1,\epsilon}:=(t_\epsilon+1)|\mathcal{E}|(|\mathcal{A}|\times|\mathcal{E}|)^{t_\epsilon}N_{\epsilon}$, which computably depends only on $|\mathcal{A}|$,  $|\mathcal{E}|$ and $\epsilon$.

Now we turn to showing our claim that $Q^{k}_\Mmu(\HistM, a_t)$ can be approximated using a $\tau$-rPOM. We have just shown how to do the approximation for $k=1$. Now 
since from \eqref{eqn:k-step-qval} we have
\begin{align*}
     Q_{\Mmu}^k(\HistM, a_t) &= \expect{\Mmu(e_t \mid \HistM ,a_t)}{(1-\gamma)r(e_{t}) + \gamma\max_{a'\in\calA}Q^{k-1}_{\Mmu}(\HistM a_te_t,a')}\\
     &=\sum_{e_t\in\calE} \Mmu(e_t \mid \HistM,a_t)\left[(1-\gamma)r(e_{t}) + \gamma\max_{a'\in\calA}Q^{k-1}_{\Mmu}(\HistM a_te_t,a')\right]\,,
\end{align*}
we can use a recursive procedure to approximate $Q_{\Mmu}^k(\HistM, a_t)$ as follows:
\begin{align*}
    \tilde{Q}^{k,\epsilon}(\HistM,a_t)=\sum_{e_t\in\calE} \tilde{\Mmu}'(e_t \mid \HistM,a_t)\left[(1-\gamma)r(e_{t}) + \gamma\max_{a'}\tilde{Q}^{k-1,\epsilon/2}(\HistM a_te_t,a')\right]\,,
\end{align*}
where
$\tilde{\Mmu}'(e_t \mid \HistM,a_t)$ can be obtained using \cref{lem:approximate_universal_inductor} in such a way that $$|\tilde{\Mmu}'(e_t \mid \HistM,a_t)-{\Mmu}(e_t \mid \HistM,a_t)|\leq \tilde{\epsilon}':=
\frac{\epsilon}{4|\mathcal{E}|}\,,$$
and the number $N_{\epsilon}'$ of oracle requests that is needed depends (computably) only on $\epsilon$ and $|\calE|$. The total number of oracle requests $\tilde{N}_{k,\epsilon}$ to compute $\tilde{Q}^{k,\epsilon}(\HistM,a_t)$ is equal to
\begin{align*}
    \tilde{N}_{k,\epsilon}&=|\mathcal{E}|N_{\epsilon}'+|\mathcal{E}||\mathcal{A}|\tilde{N}_{k-1,\epsilon/2}
    = |\mathcal{E}| \sum_{i=0}^{k-2} |\mathcal{E}|^i|\mathcal{A}|^i N_{\epsilon/2^i}' + |\mathcal{E}|^{k-1}|\mathcal{A}|^{k-1} \tilde{N}_{1, \epsilon/2^{k-1}}\,,
\end{align*}
which computably depends only on $|\mathcal{A}|$,  $|\mathcal{E}|$, $\epsilon$ and $k$.

It remains to show that $\tilde{Q}^{k,\epsilon}(\HistM,a_t)$ as defined above indeed yields the desired approximation. Now assuming the inductive hypothesis that $|\tilde{Q}^{k-1,\epsilon}(\HistA,a)-Q^{k-1}_\Mmu(\HistA,a)|\leq\frac{\epsilon}{2}$ for all $(\HistA,a,\epsilon)\in\TurnSet^*\times\calA\times(\bbQ\cap[0,1])$, we have
$$\left|\tilde{Q}^{k-1,\epsilon/2}(\HistM a_te_t,a')-Q^{k-1}_\Mmu(\HistM a_te_t,a')\right|\leq\frac{\epsilon}{4}\,.$$

Therefore,
\begin{align*}
    |\tilde{Q}^{k,\epsilon}(\HistM,a_t)&-Q_\Mmu^{k}(\HistM,a_t)|\\
    &\leq\sum_{e_t\in\calE} |\tilde{\Mmu}'(e_t \mid \HistM,a_t)-\Mmu(e_t \mid \HistM,a_t)|\left[(1-\gamma)r(e_{t}) + \gamma\max_{a'\in\calA}\tilde{Q}^{k-1,\epsilon/2}(\HistM a_te_t,a')\right]\\
    &\quad  +\sum_{e_t\in\calE}\Mmu(e_t \mid \HistM,a_t)\left| \gamma\max_{a'}\tilde{Q}^{k-1,\epsilon/2}(\HistM a_te_t,a')-\gamma\max_{a'\in\calA}Q^{k-1}_\Mmu(\HistM a_te_t,a')\right|\\
    &\leq \sum_{e_t\in\calE}\frac{\epsilon}{4|\mathcal{E}|} + \sum_{e_t\in\calE}\Mmu(e_t \mid \HistM,a_t)\gamma\frac{\epsilon}{4}\leq\frac{\epsilon}{2}\,.
\end{align*}

\subsection{Proof of \cref{thrm:counter-example}: $k$-step planners do not always converge to infinite-horizon optimal planning}\label{app:proof-counterexample}

We start by studying decoupled variants of AIXI that perform $k$-step planning (e.g., Self-AIXI \citep{catt2023self} for which $k=1$). We give a counterexample showing that such agents do not always converge to infinite-horizon planning, hence disproving the sensibly-off policy assumption. We then generalize the counterexample to the coupled case and show that $k$-E-AIXI agents do not always converge to the embedded best response w.r.t. their mixture universe model $\Mmu$.

\subsubsection{Counter-example for $k$-step planning decoupled Bayesian agents such as Self-AIXI}
We demonstrate that Self-AIXI agents and their variants performing $k$-step planning do not always converge to the best response w.r.t. their mixture environment model $\Mme$. This violation occurs because we can adversarially choose the reference universal machine for defining the Kolmogorov complexity to construct some `dogmatic belief' over the agent's own future behavior. As a result, the agent fails to converge to the optimal infinite-horizon policy, becoming trapped by a dogmatic and pessimistic self-model.

We will be interested in $k$-step planning decoupled Bayesian agents w.r.t. the mixture environment $\Mme$ and mixture policy $\Mmp$ which are induced by a Solomonoff prior over the class of environments (resp. policies) which are implementable by APOMs with access to an arbitrary probabilistic oracle $\tau$. 

The overall argument is as follows:
\begin{itemize}
    \item For every $k\geq 1$, we design a specific environment $\Mge_k$ and a specific policy $\Mgp_{\mathrm{bad}}$ such that $k$-step planning w.r.t. $\Mge_k^{\Mgp_{\mathrm{bad}}}$ is not optimal for the environment $\Mge_k$.
    \item Then we argue that we can design $U$ such that $\Mmp_U\approx \Mgp_{\mathrm{bad}}$ and $\Mme_U\approx \Mge_k$, and hence $\Mme_U^{\Mmp_U}\approx \Mge_k^{\Mgp_{\mathrm{bad}}}$. This means that  $k$-step planning with respect to $\Mme_U^{\Mmp_U}$ would not be optimal if the ground-truth environment is $\Mge_k$ (because $k$-step planning w.r.t. $\Mge_k^{\Mgp_{\mathrm{bad}}}$ is not optimal for $\Mge_k$). On the other hand, infinite-horizon planning w.r.t. $\Mme_U\approx \Mge_k$ would be optimal if $\Mge_k$ is the ground-truth environment.
\end{itemize}

Let us define a specific ground-truth environment, $\Mge_{R,k}$, for an integer $k \geq 1$, and a parameter $R\in(0,1)$, as follows:
\begin{itemize}
    \item \textbf{Action Space}: $\calA = \{\text{up}, \text{down}\}$.
    \item \textbf{Observation Space}: A trivial space $\calO = \{o_0\}$.
    \item \textbf{Reward Space}: $\calR:=\{0,R,1\}$.
    \item \textbf{Percept Space}: $\calE = \calO \times \calR$.
    \item \textbf{Reward Function}: The reward at time $t$ depends on the history of actions $a_{1:t}$:
    \begin{equation}
        \label{eq:reward-function-mge-k}
    r_t(a_{1:t}) = \begin{cases} 
    R & \text{if } a_i = \text{up for all } 1 \le i \le t \\ 
    1 & \text{if } t \ge k+1 \text{ and } a_{t-k:t} = (\text{down}, \dots, \text{down}) \\ 
    0 & \text{otherwise} 
    \end{cases}
    \end{equation}
    \item \textbf{Discount Factor}: $\gamma \in (0, 1)$.
\end{itemize}

The agent faces a choice: receive a consistent, smaller reward $R$ by always playing `up', or endure $k$ steps of zero reward to unlock a higher, perpetual reward of 1 by consistently playing `down'. If $R$ is too low then one can show that the optimal policy with respect to $\Mge_{R,k}$ is the policy $\Mvp_{\mathrm{down}}$ which always plays `down'. On the other hand, if $\Mvp_{\mathrm{up}}$ is the policy that always plays `up', then a $k$-step planner w.r.t. $\Mge_{R,k}^{\Mvp_\mathrm{up}}$ will never play the $\mathrm{down}$ action because the self-model, which is used by the $k$-step-planner to reason about its future behavior after $k$ steps, predicts that it will revert back to the $\mathrm{up}$ action after $k$-steps forever after, and all these actions would have a reward of zero.

In fact, we will show a generalization of the aforementioned facts: If $\Mve$ is an environment that is sufficiently close to $\Mge_{R,k}$, then the optimal policy with respect to $\Mve$ is to always play `down'. Similarly, if $\Mvp$ is a policy that is also sufficiently close to $\Mvp_{\mathrm{up}}$, then the $k$-step planner w.r.t. $\Mve^{\Mvp}$ will always play the sub-optimal action `up' when the ground-truth environment is $\Mge_{R,k}$.

In order to establish this, we first develop a few helpful terminology and facts regarding the approximations of $\Mge_{R,k}$ and $\Mvp_\mathrm{up}$, respectively.

\begin{definition}
    \label{def:approx-pol-env} Let $\TurnSet^*_{\mathrm{approx}}\subseteq\TurnSet^*$ and let $\epsilon>0$. We say that a policy $\tilde{\Mvp}:\TurnSet^*\to\Delta'\calA$ $\epsilon$-approximates $\Mvp$ on $\TurnSet^*_{\mathrm{approx}}$ if for every $(\HistA,a)\in\TurnSet^*_{\mathrm{approx}}\times\calA$, we have
    $$|\tilde{\Mvp}(a|\HistA)-\Mvp(a|\HistA)|\leq\epsilon\,.$$
    
    Similarly, we say that an environment $\tilde{\Mve}:\TurnSet^*\times\calA\to\Delta'\calE$ $\epsilon$-approximates $\Mve$ on $\TurnSet^*_{\mathrm{approx}}$ if for every $(\HistA,a,e)\in\TurnSet^*_{\mathrm{approx}}\times\calA\times\calE$, we have
    $$|\tilde{\Mve}(e|\HistA,a)-\Mve(e|\HistA,a)|\leq\epsilon\,.$$
\end{definition}

The following lemma establishes a few useful bounds on the probabilities of policies and environments which $\eps$-approximate other policies and environments.

\begin{lemma}
    \label{lem:approximate-pol-env-bounds}
    If $\tilde{\Mvp}:\TurnSet^*\to\Delta'\calA$ $\epsilon_{\mathrm{p}}$-approximates another policy $\Mvp:\TurnSet^*\to\Delta'\calA$ on $\TurnSet^*_{\mathrm{p},\mathrm{approx}}\subseteq\TurnSet^*$ and if $\tilde{\Mve}:\TurnSet^*\times\calA\to\Delta'\calE$ $\epsilon_{\mathrm{e}}$-approximates another environment $\Mve:\TurnSet^*\times\calA\to\Delta'\calE$ on $\TurnSet^*_{\mathrm{e},\mathrm{approx}}\subseteq\TurnSet^*$, then for every $t>0$, every $i\geq 0$, and every $\Hist[t+i]\in\TurnSet^{t+i}$ for which $\Mve^{\Mvp}(e_t\Turn_{t+1:t+i}|\HistM a_t)=1$ and $\HistM[t+j]\in\TurnSet^*_{p,\mathrm{approx}}$ and $\HistM[t+j]\in\TurnSet^*_{e,\mathrm{approx}}$ for all $0\leq j\leq i$, we have
    $$\tilde{\Mve}^{\tilde{\Mvp}}(e_t\Turn_{t+1:t+i}|\HistM a_t)\geq (1-\epsilon_{\mathrm{p}})^{i}(1-\epsilon_{\mathrm{e}})^{i+1}\Mve^{\Mvp}(e_t\Turn_{t+1:t+i}|\HistM a_t)\,,$$
    and
    $$\tilde{\Mve}^{\tilde{\Mvp}}(e_t\Turn_{t+1:t+i}|\HistM a_t)\leq (1+\epsilon_{\mathrm{p}})^{i}(1+\epsilon_{\mathrm{e}})^{i+1}\Mve^{\Mvp}(e_t\Turn_{t+1:t+i}|\HistM a_t)\,.$$
\end{lemma}
\begin{proof}
We have
\begin{align*}
    \tilde{\Mve}^{\tilde{\Mvp}}(e_t\Turn_{t+1:t+i}|\HistM a_t)&=\prod_{j=1}^i\tilde{\Mvp}(a_{t+j}|\HistM[t+j])\prod_{j=0}^i\tilde{\Mve}(e_{t+j}|\HistM[t+j]a_{t+j})\\
    &\geq\prod_{j=1}^i\Big(\Mvp(a_{t+j}|\HistM[t+j])-\epsilon_{\mathrm{p}}\Big)\prod_{j=0}^i\Big(\Mve(e_{t+j}|\HistM[t+j]a_{t+j})-\epsilon_{\mathrm{e}}\Big)\\
    &=(1-\epsilon_{\mathrm{p}})^{i}(1-\epsilon_{\mathrm{e}})^{i+1}\\
    &= (1-\epsilon_{\mathrm{p}})^{i}(1-\epsilon_{\mathrm{e}})^{i+1}\Mve^{\Mvp}(e_t\Turn_{t+1:t+i}|\HistM a_t)\,,
\end{align*}
where in the last two equalities we used the fact that $\Mve^{\Mvp}(e_t\Turn_{t+1:t+i}|\HistM a_t)=1$. The upper bound is shown similarly.
\end{proof}

We define the set $\TurnSet_{\Mge_{R,k}}^*\subseteq\TurnSet^*$ to be the set of all histories which are compatible with $\Mge_{R,k}$, i.e.,
$$\TurnSet_{\Mge_{R,k}}^*=\{\HistA\in\TurnSet^*:e_i=(o_0,r_i(a_{1:i}))\text{ for all }1\leq i\leq l(\HistA)\}\,.$$

\begin{lemma}
    \label{lem:optimal-q-value-for-env-close-to-mge-k} Let $\epsilon>0$ and let $\Mve$ be an environment that $\epsilon$-approximates $\Mge_{R,k}$ on $\TurnSet_{\Mge_{R,k}}^*$. For every $\HistA\in\TurnSet_{\Mge_{R,k}}^*$, we have
\begin{itemize}
    \item The optimal $Q^*$ value in the environment $\Mve$ after playing `down' can be bounded from below as follows
    $$Q^*_{\Mve}(\HistA,\mathrm{down})\geq \frac{\gamma^k(1-\epsilon)^{k+1}(1-\gamma)}{1 - \gamma + \gamma\epsilon}\,.$$
    \item The optimal $Q^*$ value in the environment $\Mve$ after playing `up' can be bounded from above as follows $$Q^*_{\Mve}(\HistA,\mathrm{up})\leq \frac{\epsilon}{1-\gamma } + \sup_{\ell\geq 1}\left\{(1-\gamma ^\ell)R+\gamma ^{\ell+k}\right\}\,.$$
\end{itemize}

In particular, if $R<\gamma^k$, then there exists a small enough $\epsilon$ such that
$$Q^*_{\Mve}(\HistA,\mathrm{up})<Q^*_{\Mve}(\HistA,\mathrm{down})\,,$$
and hence the optimal planner w.r.t. $\Mve$ would always play the `$\mathrm{down}$' action if the ground-truth environment is $\Mge_{R,k}$, as all produced histories would belong to $\TurnSet_{\Mge_{R,k}}^*$ for which the bounds on $Q^*$ apply.
\end{lemma}
\begin{proof}
    Let $\HistA\in\TurnSet^*_{\Mge_{R,k}}$ and let $t=l(\HistA)+1$ so that $\HistA=\HistM$.
    Let $a_{\geq t}=(\mathrm{down})^{\infty}\in\calA^\infty$ be an infinite sequence of `down' actions, and then recursively define $e_{\geq t}\in\calE^{\infty}$ as $e_{t+i}=(o_0,r_{t+i}(a_{1:t+i}))$ for every $i\geq 0$, where the reward function $r_{t+i}$ is given in \eqref{eq:reward-function-mge-k}. Notice that $\Mge_{R,k}(e_{t+i}|\HistM[t+i]a_{t+i})=1$, and hence
    $\Mve(e_{t+i}|\HistM[t+i]a_{t+i})\geq 1-\eps$ for all $i\geq 0$. Let $\Mvp_{\mathrm{down}}$ be the policy that always takes the `down' action. We have
    
    \begin{align*}
        Q^*_{\Mve}(\HistA,\mathrm{down})&=Q^*_{\Mve}(\HistM,a_t)
        \geq Q_{\Mve^{\pi_{\mathrm{down}}}}(\HistM,a_t)\\
        &\stackrel{(\ast)}=\lim_{m\to\infty}(1-\gamma)\sum_{\substack{i:t\leq i\leq m}}\gamma^{i-t}\sum_{(e_t',\Turn_{t+1:i}')\in\calE\times\TurnSet^{i-t}}r(e_i')\Mve^{\pi_{\mathrm{down}}}(e_t'\Turn_{t+1:i}'|\HistM a_t)\\
        &\geq  (1-\gamma)\sum_{\substack{i\geq k}}\gamma^{i}r(e_{t+i})\Mve^{\pi_{\mathrm{down}}}(e_t\Turn_{t+1:t+i}|\HistM a_t)\\
        &= (1-\gamma)\sum_{\substack{i\geq k}}\gamma^{i}r_{t+i}(a_{1:t+i})\Mve^{\pi_{\mathrm{down}}}(e_t\Turn_{t+1:t+i}|\HistM a_t)\\
        &\stackrel{(\dagger)}= (1-\gamma)\sum_{\substack{i\geq k}}\gamma^{i}\Mve^{\pi_{\mathrm{down}}}(e_t\Turn_{t+1:t+i}|\HistM a_t)\\
        &\stackrel{(\ddagger)}\geq (1-\gamma)\sum_{\substack{i\geq k}}\gamma^{i}(1-\epsilon)^{i+1}=\frac{\gamma^k(1-\epsilon)^{k+1}(1-\gamma)}{1 - \gamma + \gamma\epsilon}\,,\nonumber
    \end{align*}
    where $(\ast)$ follows from \eqref{eq:q-value-function-app}, $(\dagger)$ follows from the fact that $a_{t+i-k:t+i}$ is a substring of $a_{\geq t}$ and hence it is all `down', and $(\ddagger)$ follows from \cref{lem:approximate-pol-env-bounds} and the fact that $\Mvp_{\mathrm{down}}$ 0-approximates itself, and hence $$\Mve^{\pi_{\mathrm{down}}}(e_t\Turn_{t+1:t+i}|\HistM a_t)\geq(1-\epsilon)^{i+1}(1-0)^i \Mge_{R,k}^{\pi_{\mathrm{down}}}(e_t\Turn_{t+1:t+i}|\HistM a_t)=(1-\epsilon)^{i+1}\,.$$

Now we turn to proving the upper bound on $Q^*_{\Mge_{R,k}}(\HistA,\mathrm{up})$ for $\HistA\in\TurnSet_{\Mge_{R,k}}^*$. Again,
let $t=l(\HistA)+1$ so that $\HistA=\HistM$. Let $i\geq 0$ and assume that $\Turn_{t:t+i-1}\in\TurnSet^{i}$ is an extension of $\HistM$ in such a way that $\HistM[t+i]=\HistM \Turn_{t:t+i-1}\in\TurnSet^*_{\Mge_{R,k}}$. Note that for $i=0$,  we use the convention that  $\Turn_{t:t-1}=\varepsilon$ and $\TurnSet^{0}=\{\varepsilon\}$. Let $a_{t+i}\in\calA$ and let $e_{t+i}=(o_0,r_{t+i}(a_{1:t+i}))$ so that $\Hist[t+i]=\HistM[t+i]a_{t+i}e_{t+i}\in\TurnSet^*_{\Mge_{R,k}}$. We have
\begin{align}
    Q^*_{\Mve}(\HistM[t+i],a_{t+i})
    &=\sum_{e_{t+i}'\in\calE}\Mve(e_{t+i}'|\HistM[t+i] a_{t+i})((1-\gamma)r(e_{t+i}')+\gamma\max_{a_{t+i+1}\in\calA}Q^*_{\Mve}(\HistM[t+i] a_{t+i}e_{t+i}',a_{t+i+1}))\nonumber\\
    &=\Mve(e_{t+i}|\HistM[t+i] a_{t+i})((1-\gamma)r(e_{t+i})+\gamma\max_{a_{t+i+1}\in\calA}Q^*_{\Mve}(\HistM[t+i] a_{t+i} e_{t+i},a_{t+i+1}))\nonumber\\
    &\quad\quad+ \sum_{\substack{e_{t+i}'\in\calE:\\
    e_{t+i}'\neq e_{t+i}
    }}\Mve(e_{t+i}'|\HistM[t+i] a_{t+i})((1-\gamma)r(e_{t+i}')+\gamma\max_{a_{t+i+1}\in\calA}Q^*_{\Mve}(\HistM[t+i] a_{t+i} e_{t+i}',a_{t+i+1}))\nonumber\\
    &\stackrel{(\ast)}{\leq} \epsilon + (1-\gamma)r(e_{t+i})+\gamma\max_{a_{t+i+1}\in\calA}Q^*_{\Mve}(\HistM[t+i] a_{t+i} e_{t+i},a_{t+i+1})\nonumber\\
    &= \epsilon + (1-\gamma)r_{t+i}(a_{1:t+i})+\gamma\max_{a_{t+i+1}\in\calA}Q^*_{\Mve}(\HistM[t+i] a_{t+i} e_{t+i},a_{t+i+1})\label{eq:Qstar-upper-bound-for-all-downs}
\end{align}
where $(\ast)$ follows from the following facts:
\begin{itemize}
    \item $\Mve$ $\eps$-approximates $\Mge_{R,k}$ on $\TurnSet^*_{\Mge_{R,k}}$. Now since $\HistM[t+i]\in\TurnSet^*_{\Mge_{R,k}}$, we have $$|\Mge_{R,k}(e_{t+i}|\HistM[t+i],a_{t+i}) - \Mve(e_{t+i}|\HistM[t+i],a_{t+i}) |\leq \epsilon\,.$$
    \item Since $e_{t+i}=(o_0,r_{t+i}(a_{1:t+i}))$ is compatible with the deterministic environment $\Mge_{R,k}$, we have $\Mge_{R,k}(e_{t+i}|\HistM[t+i],a_{t+i})=1$, and hence $|1 - \Mve(e_{t+i}|\HistM[t+i],a_{t+i}) |\leq \epsilon$
    which means that $$1-\epsilon \leq \Mve(e_{t+i}|\HistM[t+i],a_{t+i})\leq 1\,.$$
    \item $(1-\gamma)r(e_t')+\gamma\max_{a_{t+1}\in\calA}Q^*_{\Mve}(\HistM[t+i] a_t e_t',a_{t+1})\leq 1$ and hence,
\begin{align*}
    \sum_{\substack{e_{t+i}'\in\calE:\\
    e_{t+i}'\neq e_{t+i}
    }}&\Mve(e_{t+i}'|\HistM[t+i] a_{t+i})((1-\gamma)r(e_{t+i}')+\gamma\max_{a_{t+i+1}\in\calA}Q^*_{\Mve}(\HistM[t+i] a_{t+i} e_{t+i}',a_{t+i+1}))\\
    &\leq \sum_{\substack{e_{t+i}'\in\calE:\\
    e_{t+i}'\neq e_{t+i}
    }}\Mve(e_{t+i}'|\HistM[t+i] a_{t+i})\leq 1-\Mve(e_{t+i}|\HistM[t+i] a_{t+i})\leq\epsilon\,.
\end{align*}
\end{itemize}

By defining, $\alpha=1-\gamma$, \eqref{eq:Qstar-upper-bound-for-all-downs} simplifies to

\begin{equation}
    \label{eq:Qstar-upper-bound-for-all-downs-simple}
    Q^*_{\Mve}(\HistM[t+i],a_{t+i})
    \leq \epsilon + \alpha r_{t+i}(a_{1:t+i})+\gamma \max_{a_{t+i+1}\in\calA}Q^*_{\Mve}(\HistM[t+i] a_{t+i} e_{t+i},a_{t+i+1})\,.
\end{equation}

We will show by induction on $\ell\geq 1$ that for every $t\geq 0$ and every $\HistM[t]\in\TurnSet^*_{\Mge_{R,k}}$, we have
\begin{equation}
    \label{eq:Qstar-upper-bound-for-all-downs-induction}
    Q^*_{\Mve}(\HistM[t],a_{t})\leq \epsilon\sum_{i=0}^{\ell-1}\gamma ^i +  \max_{a_{t+1:t+\ell}\in\calA^\ell}\left\{\alpha\sum_{i=0}^{\ell-1} \gamma ^i r_{t+i}(a_{1:t+i})+ \gamma ^\ell Q^*_{\Mve}(\HistM[t+\ell], a_{t+\ell}) \right\}\,,
\end{equation}
where for $t\leq i\leq t+\ell$, we define $e_{t+i}=(o_0,r_{t+i}(a_{1:t+i}))$ and we use this in $\HistM[t+\ell]$. For $\ell=1$, the inequality follows immediately from \eqref{eq:Qstar-upper-bound-for-all-downs-simple}.

Now let $\ell>1$ and assume that \eqref{eq:Qstar-upper-bound-for-all-downs-induction} holds for $\ell-1$, i.e., 
for every $t\geq 0$ and every $\HistM[t]\in\TurnSet^*_{\Mge_{R,k}}$, we have
\begin{equation}
    \label{eq:Qstar-upper-bound-for-all-downs-induction-step}
    Q^*_{\Mve}(\HistM[t],a_{t})\leq \epsilon\sum_{i=0}^{\ell-2}\gamma ^i +  \max_{a_{t+1:t+\ell-1}\in\calA^{\ell-1}}\left\{\alpha\sum_{i=0}^{\ell-2} \gamma ^{i} r_{t+i}(a_{1:t+i})+ \gamma ^{\ell-1} Q^*_{\Mve}(\HistM[t+\ell-1], a_{t+\ell-1})  \right\}\,,
\end{equation}
By combining this with \eqref{eq:Qstar-upper-bound-for-all-downs-simple}, we get
\begin{align*}
    Q^*_{\Mve}(&\HistM[t],a_{t})\\
    &\stackrel{(\ast)}{=} \epsilon + \alpha r_{t}(a_{1:t})+\gamma \max_{a_{t+1}\in\calA}Q^*_{\Mve}(\HistM[t] a_{t} e_{t},a_{t+1})\\
    &= \epsilon +  \max_{a_{t+1}\in\calA}\{ \alpha r_{t}(a_{1:t})+\gamma Q^*_{\Mve}(\HistM[t+1],a_{t+1})\}\\
    &\stackrel{(\dagger)}\leq \epsilon +  \max_{a_{t+1}\in\calA}\Bigg\{ \alpha r_{t}(a_{1:t})\\
    &\quad\quad+\gamma \left(\epsilon\sum_{i=0}^{\ell-2}\gamma ^i +  \max_{a_{t+2:t+\ell}\in\calA^{\ell-1}}\left\{\alpha\sum_{i=0}^{\ell-2} \gamma ^{i} r_{t+1+i}(a_{1:t+1+i})+ \gamma ^{\ell-1} Q^*_{\Mve}(\HistM[t+\ell], a_{t+\ell})  \right\}\right)\Bigg\}\\
    &=\epsilon\sum_{i=0}^{\ell-1}\gamma ^i +  \max_{a_{t+1:t+\ell}\in\calA^\ell}\left\{\alpha\sum_{i=0}^{\ell-1} \gamma ^i r_{t+i}(a_{1:t+i})+ \gamma ^\ell Q^*_{\Mve}(\HistM[t+\ell], a_{t+\ell}) \right\}\,,
\end{align*}
where $(\ast)$ follows from \eqref{eq:Qstar-upper-bound-for-all-downs-simple} and $(\dagger)$ follows from  \eqref{eq:Qstar-upper-bound-for-all-downs-induction-step} applied to $Q^*_{\Mve}(\HistM[t+1],a_{t+1})$. We conclude that \eqref{eq:Qstar-upper-bound-for-all-downs-induction} holds for all $\ell\geq 1$. By taking $\ell\to\infty$, we get
\begin{align*}
    Q^*_{\Mve}(\HistM[t],a_{t})&\leq
    \epsilon\sum_{i=0}^{\infty}\gamma ^i +  \sup_{a_{>t}\in\calA^\infty}\left\{\alpha\sum_{i=0}^{\infty} \gamma ^i r_{t+i}(a_{1:t+i})+ \lim_{\ell\to\infty}\gamma ^\ell Q^*_{\Mve}(\HistM[t+\ell], a_{t+\ell}) \right\}\\
    &=
    \frac{\epsilon}{1-\gamma } + \alpha \sup_{a_{>t}\in\calA^\infty}\sum_{i=0}^{\infty} \gamma ^i r_{t+i}(a_{1:t+i})\,.
\end{align*}

Therefore,
\begin{align*}
    Q^*_{\Mve}(\HistM[t],\mathrm{up})&\leq
    \frac{\epsilon}{1-\gamma } + \alpha \sup_{\substack{a_{\geq t}\in\calA^\infty:\\
    a_t=\mathrm{up}}}\sum_{i=0}^{\infty} \gamma ^i r_{t+i}(a_{1:t+i})\,.
\end{align*}

Now for every $\ell>0$, let
$$\mathcal{S}_\ell=\{ a_{>0}\in\calA^\infty: a_{i}=\mathrm{up}\text{ for }i\leq \ell,\text{ and } a_{\ell+1}=\mathrm{down}\}\,,$$
i.e., $\cal{S}_\ell$ is the set of all infinite sequences of actions which start with $\ell$ `up' actions followed by a `down' action. Furthermore let $$\mathcal{S}_\infty=\{(\mathrm{up})^\infty\}\,,$$ be the (singleton) set containing the single infinite sequence of up actions. Clearly,
$$\{a_{1:\infty}\in \calA^\infty:a_1=\mathrm{up}\}=\bigcup_{1\leq \ell\leq \infty}\mathcal{S}_\ell\,,$$
where `$1\leq \ell\leq \infty$' is a shorthand for $\ell\in \mathbb{N}\cup\{\infty\}=\{1,2,\ldots\}\cup\{\infty\}$. Hence, we can rewrite the upper bound on $Q^*_{\Mve}(\HistM,\mathrm{up})$ as follows:
\begin{align*}
    Q^*_{\Mve}(\HistM,\mathrm{up})&\leq \frac{\epsilon}{1-\gamma } + \alpha\sup_{\substack{a_{\geq t}\in\calA^\infty:\\
a_t=\mathrm{up}}}\sum_{i=0}^\infty \gamma ^i r_{t+i}(a_{1:t+i})=\frac{\epsilon}{1-\gamma } + \alpha\sup_{1\leq \ell\leq \infty}\sup_{a_{\geq t}\in\calS_\ell}\sum_{i=0}^\infty \gamma ^i r_{t+i}(a_{1:t+i})\,.
\end{align*}

Now notice that:
\begin{itemize}
    \item If $\ell=\infty$, then
    $$\sup_{a_{\geq t}\in\calS_\ell}\sum_{i=0}^\infty \gamma ^i r_{t+i}(a_{<t}a_{t:t+i})\leq \sum_{i=0}^{\infty}\gamma ^i R=\frac{R}{1-\gamma }\,,$$
    where the upper bound follows because the reward corresponding to an up action can be at most $R$.
    \item If $1\leq \ell<\infty$, then
    \begin{align*}
        \sup_{a_{\geq t}\in\calS_\ell}\sum_{i=0}^\infty \gamma ^i r_{t+i}(a_{<t}a_{t:t+i})&\stackrel{(\ast)}{\leq} \sum_{i=0}^{\ell-1}\gamma ^iR +\sum_{i=\ell+k}^\infty \gamma ^i=\frac{(1-\gamma ^\ell)R}{1-\gamma } +\frac{\gamma ^{\ell+k}}{1-\gamma }\,,
    \end{align*}
    where the upper bound $(\ast)$ follows because for $a_{\geq t}\in\calS_\ell$ we have:
    \begin{itemize}
        \item The actions $(a_{t'})_{t\leq t' <t+\ell}$ are all up and hence they all get a reward of at most $R$.
        \item The action $a_{t+\ell}$ is down. Now since $a_{t+\ell-1}=\mathrm{up}$, all the actions $(a_{t'})_{t+\ell\leq t' <t+\ell+k}$ will get a reward of 0.
        \item The actions $(a_{t'})_{t'\geq t+\ell+k}$ can each get a reward of at most 1.
    \end{itemize}
\end{itemize}
We conclude that
\begin{align*}
    Q^*_{\Mve}(\HistM,\mathrm{up})&\leq \frac{\epsilon}{1-\gamma } + \frac{\alpha}{1-\gamma }\max\left\{R, \sup_{\ell\geq 1}\left\{(1-\gamma ^\ell)R+\gamma ^{\ell+k}\right\}\right\}\\
    &\stackrel{(\dagger)}=\frac{\epsilon}{1-\gamma } + \sup_{\ell\geq 1}\left\{(1-\gamma ^\ell)R+\gamma ^{\ell+k}\right\}\,,
\end{align*}
where $(\dagger)$ follows because $\gamma < 1$, and hence $$\lim_{\ell\to\infty}(1-\gamma ^\ell)R+\gamma ^{\ell+k}=R\,,$$ which implies that the supremum over $\ell \ge 1$ is at least $R$ (rendering the maximum with $R$ redundant).

It remains to show that if $R<\gamma^k$, then there exists $\epsilon>0$ such that  $Q^*_{\Mve}(\HistM,\mathrm{up})\leq Q^*_{\Mve}(\HistM,\mathrm{down})$. For this, notice that if $R<\gamma^k$, then $\gamma^k-R>0$ as well and hence
$$(1-\gamma^\ell)R+\gamma^{\ell+k} = R + \gamma^\ell(\gamma^k - R)\leq R + \gamma(\gamma^k- R)\,,$$
and hence
\begin{align*}
    Q^*_{\Mve}(\HistM,\mathrm{up})&\leq 
    \frac{\epsilon}{1-\gamma } + R + \gamma(\gamma^k- R)\,.
\end{align*}

In particular
\begin{align*}
Q^*_{\Mve}(\HistM,\mathrm{down})-
    Q^*_{\Mve}(\HistM,\mathrm{up})&\geq \frac{\gamma^k(1-\epsilon)^{k+1}(1-\gamma)}{1 - \gamma + \gamma\epsilon} -
    \frac{\epsilon}{1-\gamma } - R - \gamma(\gamma^k- R)\,.
\end{align*}

By taking the limit as $\epsilon\to0$, we get
\begin{align*}
    \lim_{\epsilon\to 0} \frac{\gamma^k(1-\epsilon)^{k+1}(1-\gamma)}{1 - \gamma + \gamma\epsilon} -
    \frac{\epsilon}{1-\gamma } - R - \gamma(\gamma^k- R) = \gamma^k - R - \gamma(\gamma^k -R)=(\gamma^k - R)(1-\gamma)>0\,.
\end{align*}

We conclude that for sufficiently small $\epsilon>0$, we have $Q^*_{\Mve}(\HistM,\mathrm{down})>
    Q^*_{\Mve}(\HistM,\mathrm{up})$ for every $\HistM\in\TurnSet^*_{\Mge_{R,k}}$. In particular, if the ground-truth environment is $\Mge_{R,k}$, then an optimal planner w.r.t. $\Mve$ always takes the `down' action.
\end{proof}

Now we will argue that if $\Mve$ $\epsilon$-approximates $\Mge_{R,k}$ on $\TurnSet^*_{\Mge_{R,k}}$, and if $\Mvp$ $\epsilon$-approximates $\Mvp_{\mathrm{up}}$ on a suitable subset of $\TurnSet^*_{\Mge_{R,k}}$ for sufficiently small $\epsilon>0$, then the $k$-step optimal planner w.r.t. $\Mve^{\Mvp}$ will always play the suboptimal `up' action if the ground-truth environment is $\Mge_{R,k}$.

As a warm-up, let us consider the case where $\Mve=\Mge_{R,k}$ and $\Mvp=\Mvp_{\mathrm{up}}$, and let us analyze the $k$-step planner w.r.t. $\Mge_{R,k}^{\Mgp_{\mathrm{up}}}$. Assume that the ground-truth environment is $\Mge_{R,k}$ and assume that we observed a history $\HistM$ of $t-1$ `up' actions. What will the $k$-step planner do? By definition, it will choose the action $a_t$ which maximizes $Q^k_{\Mge_{R,k}^{\Mgp_{\mathrm{up}}}}(\HistM,a_t)$. Now because $\Mge_{R,k}$ is deterministic, we can write the expression of $Q^k_{\Mge_{R,k}^{\Mgp_{\mathrm{up}}}}(\HistM,a_t)$ as
$$Q^k_{\Mge_{R,k}^{\Mgp_{\mathrm{up}}}}(\HistM,a_t)=\max_{a_{t+1:t+k-1}\in\calA^{k-1}}(1-\gamma)\sum_{i=0}^{k-1}\gamma^i r_{t+i}(a_{1:t+i}) + \gamma^k V_{\Mge_{R,k}^{\Mgp_{\mathrm{up}}}}(\HistM \Turn_{t:t+k-1})\,,$$
where in $\Turn_{t:t+k-1}$, we let $e_{t'}=(o_0,r_{t'}(a_{1:t'}))$ for $t\leq t'<t+k$. Now notice that:
\begin{itemize}
    \item If $a_t=\mathrm{down}$, then $r_{t'}(a_{1:t'})=0$ for all $t\leq t'<t+k$ because a `down' action will make the reward of any subsequent `up' action to be 0, and furthermore, the reward of a `down' action is only worth it when there is a streak of at least $k$ `down' actions. Furthermore, $V_{\Mge_{R,k}^{\Mgp_{\mathrm{up}}}}(\HistM \Turn_{t:t+k-1})$ would also be equal to zero because $\Mge_{R,k}^{\Mgp_{\mathrm{up}}}$ only samples up `actions' which would have zero rewards since they were preceded by $a_t=\mathrm{down}$.\footnote{It is worth noting, that while $\Mge_{R,k}^{\Mgp_{\mathrm{up}}}(\Hist)=0$ if $a_t=\mathrm{down}$, the conditional $\Mge_{R,k}^{\Mgp_{\mathrm{up}}}(\cdot|\HistM a_t)$ is still well defined (cf. \cref{rem:pol-env-counterfactual-conditional}). Therefore, $Q^k_{\Mge_{R,k}^{\Mgp_{\mathrm{up}}}}(\HistM,a_t)$ is well-defined even for $a_t=\mathrm{down}$.}
    \item If $a_t=\mathrm{up}$, then
    $$Q^k_{\Mge_{R,k}^{\Mgp_{\mathrm{up}}}}(\HistM,a_t)\geq Q_{\Mge_{R,k}^{\Mgp_{\mathrm{up}}}}(\HistM,a_t)=R>0\,.$$
\end{itemize}
Therefore the $k$-step planner w.r.t. $\Mge_{R,k}^{\Mgp_{\mathrm{up}}}$ will always play the suboptimal `up' action. The main reason for the failure of the $k$-step planner w.r.t. $\Mge_{R,k}^{\Mgp_{\mathrm{up}}}$ is that its self-model is a bad policy that is incapable of ``continuing on doing the right thing". Because of this, the $k$-step planner is forced to play sub-optimally, as it does not ``trust its future self" on doing the right thing.

Now we turn to showing a similar result for general $\Mve$ and $\Mvp$ that $\eps$-approximate $\Mge_{R,k}$ and $\Mvp_{\mathrm{up}}$, respectively.

Let $\calA^{*\text{-}\mathrm{up}\text{-}k\text{-}\mathrm{down}\text{-}*\text{-}\mathrm{up}}$ be the set of action sequences which contains a streak of `up' actions followed by a streak of exactly $k$ `down' actions followed by a streak of `up' actions, i.e.,
\begin{align*}
\calA^{*\text{-}\mathrm{up}\text{-}k\text{-}\mathrm{down}\text{-}*\text{-}\mathrm{up}} = \bigcup_{t\geq k}\bigcup_{0\leq i\leq t-k}\{\mathrm{up}\}^i\times \{\mathrm{down}\}^k\times \{\mathrm{up}\}^{t-k-i}\,,
\end{align*}
and let
$$\TurnSet^{*\text{-}\mathrm{up}\text{-}k\text{-}\mathrm{down}\text{-}*\text{-}\mathrm{up}}_{\Mge_{R,k}} = \{\HistM:t>0,\, \HistM\in\TurnSet^*_{\Mge_{R,k}}\text{ and }a_{<t}\in\calA^{*\text{-}\mathrm{up}\text{-}k\text{-}\mathrm{down}\text{-}*\text{-}\mathrm{up}}\}\subseteq \TurnSet^*_{\Mge_{R,k}}\,.$$

Furthermore, let $\calA^{*\text{-}\mathrm{up}}$ be the set of action sequences which contain only `up' actions, and let
$$\TurnSet^{*\text{-}\mathrm{up}}_{\Mge_{R,k}} = \{\HistM:t>0,\, \HistM\in\TurnSet^*_{\Mge_{R,k}}\text{ and }a_{<t}\in\calA^{*\text{-}\mathrm{up}}\}\subseteq \TurnSet^*_{\Mge_{R,k}}\,.$$

\begin{lemma}
    \label{lem:k-step-approx-Mvpup-suboptimal} Let $\epsilon>0$, let $\Mve$ be an environment that $\epsilon$-approximates $\Mge_{R,k}$ on $\TurnSet^*_{\Mge_{R,k}}$, and let $\Mvp$  be a policy that  $\epsilon$-approximates $\Mvp_{\mathrm{up}}$ on $\TurnSet^{*\text{-}\mathrm{up}\text{-}k\text{-}\mathrm{down}\text{-}*\text{-}\mathrm{up}}_{\Mge_{R,k}} \cup \TurnSet^{*\text{-}\mathrm{up}}_{\Mge_{R,k}}$. For every $\HistM\in \TurnSet^{*\text{-}\mathrm{up}}_{\Mge_{R,k}}$ we have
    \begin{align*}
        Q^k_{{\Mve^{\Mvp}}}(\HistM,\mathrm{up})\geq \frac{(1-\gamma)(1-\epsilon)R}{1-\gamma(1-\epsilon)^2}\,,
    \end{align*}
    and
    \begin{align*}
        Q^k_{{\Mve^{\Mvp}}}(\HistM,\mathrm{down})\leq k\epsilon+\gamma^k(\gamma+2\epsilon)\,.
    \end{align*}
    
    In particular, for sufficiently small $\epsilon>0$ and $R>\gamma^{k+1}$, we have
    $$Q^k_{\Mve^\Mvp}(\HistM,\mathrm{down})<Q^k_{\Mve^\Mvp}(\HistM,\mathrm{up})\,,$$
    and hence the $k$-step planner w.r.t. $\Mve^\Mvp$ will always play the `up' action if the ground-truth environment is $\Mge_{R,k}$.
\end{lemma}
\begin{proof}

Let us first prove the lower bound on $Q^k_{\Mve^\Mvp}(\HistM,\mathrm{up})$. Let $a_{\geq t}=(\mathrm{up})^{\infty}\in\calA^\infty$ be an infinite sequence of `up' actions, and then recursively define $e_{\geq t}\in\calE^{\infty}$ as $e_{t+i}=(o_0,r_{t+i}(a_{1:t+i}))$ for every $i\geq 0$, where the reward function $r_{t+i}$ is given in \eqref{eq:reward-function-mge-k}. Clearly $\HistM[t+i]\in \TurnSet^{*\text{-}\mathrm{up}}_{\Mge_{R,k}}$ for all $i\geq 0$. Furthermore, since $a_{1:t+i}$ consists only of up actions, we have $r(e_{t+i})=r_{t+i}(a_{1:t+i})=R$ for all $i\geq 0$. Notice that
\begin{align*}
  Q^k_{\Mve^\Mvp}(\HistM,\mathrm{up}) &\stackrel{(\ast)}{\geq} Q^1_{\Mve^\Mvp}(\HistM,\mathrm{up})=Q_{\Mve^\Mvp}(\HistM,\mathrm{up})\\
  &\stackrel{(\dagger)}=\lim_{m\to\infty}(1-\gamma)\sum_{\substack{i:t\leq i\leq m}}\gamma^{i-t}\sum_{(e_t',\Turn_{t+1:i}')\in\calE\times\TurnSet^{i-t}}r(e_i')\Mve^\Mvp(e_t'\Turn_{t+1:i}'|\HistM a_t)\\
  &\geq \lim_{m\to\infty}(1-\gamma)\sum_{\substack{i:t\leq i\leq m}}\gamma^{i-t}r(e_i)\Mve^\Mvp(e_t\Turn_{t+1:i}|\HistM a_t)\\
  &= (1-\gamma)\sum_{i\geq 0}\gamma^{i}R\Mve^\Mvp(e_t\Turn_{t+1:t+i}|\HistM a_t)\\
  &\stackrel{(\ddagger)}\geq (1-\gamma)\sum_{i\geq 0}\gamma^{i}R(1-\epsilon)^{2i+1}\Mge_{R,k}^{\Mvp_{\mathrm{up}}}(e_t\Turn_{t+1:t+i}|\HistM a_t)\\
  &= (1-\gamma)\sum_{i\geq 0}\gamma^{i}R(1-\epsilon)^{2i+1}=\frac{(1-\gamma)(1-\epsilon)R}{1-\gamma(1-\epsilon)^2}
  \,,
\end{align*}
where $(\ast)$ follows from \cref{lem:k-step-better-than-k-1}, $(\dagger)$ follows from \eqref{eq:q-value-function-app}, $(\ddagger)$ follows from \cref{lem:approximate-pol-env-bounds}.

Now we turn to proving an upper bound on $Q^k_{\Mve^\Mvp}(\HistM,\mathrm{down})$. For this, let us define $a_t=\mathrm{down}$. We will show by induction on $k\geq 1$ that

\begin{equation}
        \label{eq:Q-k-step-upper-bound-approximate}
    Q^k_{{\Mve^{\Mvp}}}(\HistM,a_t)\leq k\epsilon +\max_{a_{t+1:t+k-1}\in\calA^{k-1}}\left\{(1-\gamma)\sum_{i=0}^{k-1} \gamma^i r(e_{t+i}) + \gamma^k V_{{\Mve^{\Mvp}}}(\HistM[t+k])\right\}\,,
\end{equation}
    where we recursively define $e_{t+i}=(o_0,r_{t+i}(a_{1:t+i}))$ for $0\leq i< k$ and we use those in $\HistM[t+k]$. We again use the convention that for $k=1$, we have $\calA^{k-1}=\calA^{0}=\{\varepsilon\}$ and $a_{t+1:t+k-1}=a_{t+1:t}=\varepsilon$.

We prove \eqref{eq:Q-k-step-upper-bound-approximate} by induction on $k$. For $k=1$, we have
\begin{align*}
    Q_{{\Mve^{\Mvp}}}^1(\HistM,a_t)&=Q_{{\Mve^{\Mvp}}}(\HistM,a_t)\\
    &=\sum_{e_t'\in\calE}{\Mve^{\Mvp}}(e_t'|\HistM a_t)\left((1-\gamma)r(e_t')+\gamma V_{\Mve^{\Mvp}}(\HistM a_t e_t')\right)\\
    &= \sum_{\substack{e_t'\in\calE:\\e_t'\neq e_t}}{\Mve^{\Mvp}}(e_t'|\HistM a_t)\left((1-\gamma)r(e_t')+\gamma V_{\Mve^{\Mvp}}(\HistM a_t e_t')\right)\\
    &\quad\quad\quad\quad+ {\Mve^{\Mvp}}(e_t|\HistM a_t)\left[(1-\gamma)r(e_t)+\gamma V_{\Mve^{\Mvp}}(\HistM a_t e_t)\right]\\
    &\stackrel{(\ast)}\leq  \left(\sum_{\substack{e_t'\in\calE:\\e_t'\neq e_t}}{\Mve^{\Mvp}}(e_t'|\HistM a_t)\right) + \left[(1-\gamma)r(e_t)+\gamma V_{\Mve^{\Mvp}}(\HistM a_t e_t)\right]\\
    &\stackrel{(\dagger)}\leq \epsilon + \left[(1-\gamma)r(e_t)+\gamma V_{\Mve^{\Mvp}}(\HistM a_t e_t)\right]
        \,,
\end{align*}
where in $(\ast)$ we used the fact that $(1-\gamma)r(e_t')+\gamma V_{\Mve^{\Mvp}}(\HistM a_t e_t')\leq 1$ for all $e_t'\in\cal E$, and that ${\Mve^{\Mvp}}(e_t'|\HistM a_t)\leq 1$, and in $(\dagger)$ we used the fact that ${\Mve^{\Mvp}}$ is a semimeasure, and hence
$$\sum_{\substack{e_t'\in\calE:\\e_t'\neq e_t}}{\Mve^{\Mvp}}(e_t'|\HistM a_t)\leq 1 - {\Mve^{\Mvp}}(e_t|\HistM a_t)=1-\Mve(e_t|\HistM a_t)=\Mge_{R,k}(e_t|\HistM a_t)-\Mve(e_t|\HistM a_t)\leq \epsilon\,.$$
Therefore, \eqref{eq:Q-k-step-upper-bound-approximate} holds for $k=1$.

Now let $k>1$ and assume that \eqref{eq:Q-k-step-upper-bound-approximate} holds for $k-1$, i.e., for every $t>0$ and every $(\HistM,a_t)\in \TurnSet^{t-1}\times\calA$, we have
\begin{align*}
    Q^{k-1}_{{\Mve^{\Mvp}}}(\HistM,a_t)\leq (k-1)\epsilon +\max_{a_{t+1:t+k-2}\in\calA^{k-2}}\left\{(1-\gamma)\sum_{i=0}^{k-2} \gamma^i r(e_{t+i}) + \gamma^{k-1} V_{{\Mve^{\Mvp}}}(\HistM[t+k-1])\right\}\,,
\end{align*}
where $e_{t+i}$ for $0\leq i<k-1$ are recursively defined as $e_{t+i}=(o_0,r_{t+i}(a_{1:t+i}))$ for $0\leq i< k-1$ and we use those in $\HistM[t+k-1]$.

Then, it follows from \eqref{eqn:k-step-qval} that
\begin{align*}
    Q_{\Mve^{\Mvp}}^k(\HistM, a_t)&=\sum_{e_t'\in\calE}{\Mve^{\Mvp}}(e_t'|\HistM,a_t)\left[(1-\gamma) r(e_t')+\gamma \max_{a_{t+1}\in\calA}Q_{\Mve^{\Mvp}}^{k-1}(\HistM a_t e_t',a_{t+1})\right]\\
    &=\sum_{\substack{e_t'\in\calE:\\e_t'\neq e_t}}{\Mve^{\Mvp}}(e_t'|\HistM,a_t)\left[(1-\gamma) r(e_t')+\gamma \max_{a_{t+1}\in\calA}Q_{\Mve^{\Mvp}}^{k-1}(\HistM a_t e_t',a_{t+1})\right]\\
    &\quad\quad\quad\quad + {\Mve^{\Mvp}}(e_t|\HistM a_t)\left[(1-\gamma) r(e_t)+\gamma \max_{a_{t+1}\in\calA}Q_{\Mve^{\Mvp}}^{k-1}(\HistM a_t e_t,a_{t+1})\right]\\
    &\leq \epsilon + (1-\gamma) r(e_t)+\gamma \max_{a_{t+1}\in\calA}Q_{\Mve^{\Mvp}}^{k-1}(\HistM a_t e_t,a_{t+1})\,,
\end{align*}
where in the last inequality we used the fact that $1-\epsilon\leq {\Mve^{\Mvp}}(e_t|\HistM a_t)\leq 1$. Therefore,
\begin{align*}
    Q_{\Mve^{\Mvp}}^k(\HistM, a_t)\leq \epsilon + (1-\gamma) r(e_t)+\gamma \max_{a_{t+1}\in\calA}Q_{\Mve^{\Mvp}}^{k-1}(\HistM a_t e_t,a_{t+1})\,.
\end{align*}
By using the induction hypothesis on the history $\HistM[t+1]=\HistM[t]a_t e_t$, we get
\begin{align*}
    Q_{\Mve^{\Mvp}}^k(\HistM, a_t)&\leq \epsilon + (1-\gamma) r(e_t)\\
    &\quad\quad+\gamma \max_{a_{t+1}\in\calA}\left\{
    (k-1)\epsilon +\max_{a_{t+2:t+k-1}\in\calA^{k-2}}\left\{(1-\gamma)\sum_{i=0}^{k-2} \gamma^i r(e_{t+1+i}) + \gamma^{k-1} V_{{\Mve^{\Mvp}}}(\HistM[t+k])\right\}
    \right\}\\
    &\leq k\epsilon + \max_{a_{t+1:t+k-1}\in\calA^{k-1}}\left\{(1-\gamma)\sum_{i=0}^{k-1} \gamma^i r(e_{t+i}) + \gamma^{k} V_{{\Mve^{\Mvp}}}(\HistM[t+k])\right\}
    \,,
\end{align*}
where $e_{t+i}$ for $0\leq i<k$ are again recursively defined as $e_{t+i}=(o_0,r_{t+i}(a_{1:t+i}))$ and we use those in $\HistM[t+k]$. It follows by induction that \eqref{eq:Q-k-step-upper-bound-approximate} holds for every $k\geq 1$. Now for every $a_{t+1:t+k-1}\in\calA^{k-1}$ and every $1\leq i\leq k-1$, we have
\begin{itemize}
    \item If $a_{t+i}=\mathrm{up}$, then $r(e_{t+i})=r_{t+i}(a_{1:t+i})=0$ because $a_{t+i}$ is preceded by the $\mathrm{down}$ action.
    \item If $a_{t+i}=\mathrm{down}$, then $r(e_{t+i})=r_{t+i}(a_{1:t+i})=0$ because $a_{<t}$ is all-up, and hence $a_{t+i-k:t+i}$ is not all down.
\end{itemize}
We conclude that
\begin{align*}
    Q^k_{{\Mve^{\Mvp}}}(\HistM,a_t)\leq k\epsilon +\gamma^k\max_{a_{t+1:t+k-1}\in\calA^{k-1}} V_{{\Mve^{\Mvp}}}(\HistM[t+k])\,.
\end{align*}

Now we turn to proving an upper bound on $V_{{\Mve^{\Mvp}}}(\HistM[t+k])$, which can be written as 
\begin{align*}
    V_{{\Mve^{\Mvp}}}(\HistM[t+k])&=\sum_{a_{t+k}\in\calA}\Mvp(a_{t+k}|\HistM[t+k])Q_{\Mve^\Mvp}(\HistM[t+k],a_{t+k})\,,
\end{align*}
Let $a_{t+k}\in\calA$ and let $e_{t+k}=(o_0,r_{t+k}(a_{1:t}))$. We have
\begin{align*}
Q_{\Mve^\Mvp}(\HistM[t+k],a_{t+k})&=\sum_{e_{t+k}'\in\calE}\Mve(e_{t+k}'|\HistM[t+k]a_{t+k})((1-\gamma)r(e_{t+k}')+\gamma V_{\Mve^\Mvp}(\HistM[t+k]a_{t+k}e_{t+k}'))    \\
&\leq\gamma +(1-\gamma)\sum_{e_{t+k}'\in\calE}\Mve(e_{t+k}'|\HistM[t+k]a_{t+k})r(e_{t+k}')\\
&=\gamma +(1-\gamma)\Mve(e_{t+k}|\HistM[t+k]a_{t+k})r(e_{t+k}) + (1-\gamma)\sum_{\substack{e_{t+k}'\in\calE:\\e_{t+k}'\neq e_{t+k}}}\Mve(e_{t+k}'|\HistM[t+k]a_{t+k})r(e_{t+k}')\\
&\leq\gamma + r(e_{t+k}) + \sum_{\substack{e_{t+k}'\in\calE:\\e_{t+k}'\neq e_{t+k}}}\Mve(e_{t+k}'|\HistM[t+k]a_{t+k})\leq \gamma + r(e_{t+k}) + 1-\Mve(e_{t+k}|\HistM[t+k]a_{t+k})\\
&\stackrel{(\ast)}{\leq} \gamma + r(e_{t+k}) + \epsilon\,,
\end{align*}
where $(\ast)$ follows from the fact that $\Mve$ $\eps$-approximates $\Mge_{R,k}$ on $\TurnSet^*_{\Mge_{R,k}}$. Now notice that

\begin{itemize}
    \item If $a_{t+1:t+k-1}$ contains an `up' action, then $r(e_{t+k})=r_{t+k}(a_{1:t+k})=0$ regardless of whether $a_{t+k}$ is up or down. Therefore,
    \begin{align*}
        V_{{\Mve^{\Mvp}}}(\HistM[t+k])&=\sum_{a_{t+k}\in\calA}\Mvp(a_{t+k}|\HistM[t+k])Q_{\Mve^\Mvp}(\HistM[t+k],a_{t+k})\\
        &\leq\sum_{a_{t+k}\in\calA}\Mvp(a_{t+k}|\HistM[t+k])(\gamma + r(e_{t+k}) + \epsilon)\\
        &=\sum_{a_{t+k}\in\calA}\Mvp(a_{t+k}|\HistM[t+k])(\gamma + \epsilon)=\gamma+\epsilon\,.
    \end{align*}
    \item If $a_{t+1:t+k-1}$ is all-down, then $a_{t:t+k-1}$ is all-down, and hence $a_{<t+k}\in\calA^{*\text{-}\mathrm{up}\text{-}k\text{-}\mathrm{down}\text{-}*\text{-}\mathrm{up}}$. Now since $\Mvp$ $\epsilon$-approximates $\Mvp_{\mathrm{up}}$ on $\TurnSet^{*\text{-}\mathrm{up}\text{-}k\text{-}\mathrm{down}\text{-}*\text{-}\mathrm{up}}$, we get
    \begin{align*}
        V_{{\Mve^{\Mvp}}}(\HistM[t+k])&=\sum_{a_{t+k}\in\calA}\Mvp(a_{t+k}|\HistM[t+k])Q_{\Mve^\Mvp}(\HistM[t+k],a_{t+k})\\
        &=\Mvp(\mathrm{up}|\HistM[t+k])Q_{\Mve^\Mvp}(\HistM[t+k],\mathrm{up})+\sum_{\substack{a_{t+k}\in\calA:\\a_{t+k}\neq\mathrm{up}}}\Mvp(a_{t+k}|\HistM[t+k])Q_{\Mve^\Mvp}(\HistM[t+k],a_{t+k})\\
        &\leq\Mvp(\mathrm{up}|\HistM[t+k])Q_{\Mve^\Mvp}(\HistM[t+k],\mathrm{up})+\sum_{\substack{a_{t+k}\in\calA:\\a_{t+k}\neq\mathrm{up}}}\Mvp(a_{t+k}|\HistM[t+k])\\
        &\leq Q_{\Mve^\Mvp}(\HistM[t+k],\mathrm{up})+1-\Mvp(\mathrm{up}|\HistM[t+k])\\
        &= Q_{\Mve^\Mvp}(\HistM[t+k],\mathrm{up})+\Mvp_{\mathrm{up}}(\mathrm{up}|\HistM[t+k])-\Mvp(\mathrm{up}|\HistM[t+k])\\
        &\leq Q_{\Mve^\Mvp}(\HistM[t+k],\mathrm{up})+\epsilon\,.
    \end{align*}
    Now fix $a_{t+k}=\mathrm{up}$. We get
    $$V_{{\Mve^{\Mvp}}}(\HistM[t+k])\leq Q_{\Mve^\Mvp}(\HistM[t+k] a_{t+k})+\epsilon\leq  \gamma + r(e_{t+k}) + 2\epsilon\,.$$
    Since $a_t=\mathrm{down}$, we have $r(e_{t+k})=0$, and hence
    $$V_{{\Mve^{\Mvp}}}(\HistM[t+k])\leq \gamma + 2\epsilon\,.$$
\end{itemize}
We conclude that
\begin{align*}
    Q^k_{{\Mve^{\Mvp}}}(\HistM,\mathrm{down})\leq k\epsilon +\gamma^k\max_{a_{t+1:t+k-1}\in\calA^{k-1}} V_{{\Mve^{\Mvp}}}(\HistM[t+k])\leq k\epsilon+\gamma^k(\gamma+2\epsilon)\,.
\end{align*}

Now since
\begin{align*}
    Q^k_{{\Mve^{\Mvp}}}(\HistM,\mathrm{up})-Q^k_{{\Mve^{\Mvp}}}(\HistM,\mathrm{down})\geq \frac{(1-\gamma)(1-\epsilon)R}{1-\gamma(1-\epsilon)^2} -k\epsilon - \gamma^k(\gamma+2\epsilon)\,,
\end{align*}
and since
$$\lim_{\epsilon\to 0}\frac{(1-\gamma)(1-\epsilon)R}{1-\gamma(1-\epsilon)^2} -k\epsilon - \gamma^k(\gamma+2\epsilon) = R-\gamma^{k+1}\,,$$
we conclude that if $R>\gamma^{k+1}$, then there exists $\epsilon>0$ such that $Q^k_{{\Mve^{\Mvp}}}(\HistM,\mathrm{up})> Q^k_{{\Mve^{\Mvp}}}(\HistM,\mathrm{down})$ for every $\HistM\in\TurnSet^*_{\Mge_{R,k}}$ consisting of only up actions. For such $\epsilon$, the $k$-step-planner w.r.t. $\Mve^{\Mvp}$ will always take the `up' action if the ground-truth environment is $\Mge_{R,k}$.
\end{proof}

\begin{corollary}
    \label{cor:optimal-and-k-step-approx}
    Let $\epsilon>0$, let $\Mve$ be an environment that $\epsilon$-approximates $\Mge_{R,k}$ on $\TurnSet^*_{\Mge_{R,k}}$, and let $\Mvp$ be a policy that  $\epsilon$-approximates $\Mvp_{\mathrm{up}}$ on $\TurnSet^{*\text{-}\mathrm{up}\text{-}k\text{-}\mathrm{down}\text{-}*\text{-}\mathrm{up}}_{\Mge_{R,k}} \cup \TurnSet^{*\text{-}\mathrm{up}}_{\Mge_{R,k}}$. Define the following two policies:
    \begin{itemize}
        \item Let $\pi^*_{\Mve^\Mvp}$ be the embedded best response w.r.t. $\Mve^\Mvp$, which is the same as the optimal policy $\pi^*_{\Mve}$ w.r.t. $\Mve$.\footnote{Note that the conditionals of $\Mve^\Mvp$ are always well defined (cf. \cref{rem:pol-env-counterfactual-conditional}) and we can define the embedded best response w.r.t. those conditionals. This boils down to optimal planning w.r.t. $\Mve$.}
        \item Let $\pi^k_{\Mve^\Mvp}$ be the $k$-step optimal planner w.r.t. $\Mve^\Mvp$.
    \end{itemize}
    If $\gamma^{k+1}<R<\gamma^k$, and if $\epsilon>0$ is sufficiently small, then if $\Mge_{R,k}$ is the ground-truth environment, the $\pi^*_{\Mve^\Mvp}=\pi^*_{\Mve}$ always plays the `down' action while the  $\pi^k_{\Mve^\Mvp}$ policy always plays the `up' action. Furthermore, there exists a gap between the returns of $\pi^*_{\Mve^\Mvp}$ and $\pi^k_{\Mve^\Mvp}$:
    \begin{align*}
        \lim_{t\to\infty}V_{\Mge_{R,k}^{\pi^k_{\Mve^\Mvp}}}(\Hist)=R\,,\quad \Mge_{R,k}^{\pi^k_{\Mve^\Mvp}}\text{-almost surely},
    \end{align*}
    and
    \begin{align*}
        \lim_{t\to\infty}V_{\Mge_{R,k}^{\pi^*_{\Mve^\Mvp}}}(\Hist)=1\,,\quad \Mge_{R,k}^{\pi^*_{\Mve^\Mvp}}\text{-almost surely}.
    \end{align*}
\end{corollary}
\begin{proof}
    This is an immediate corollary of \cref{lem:optimal-q-value-for-env-close-to-mge-k} and \cref{lem:k-step-approx-Mvpup-suboptimal}.
\end{proof}

Now we turn to applying \cref{cor:optimal-and-k-step-approx} to the mixture policy and the mixture environment that are induced by the Solomonoff prior.

Let us first establish some useful notation, and a few helpful facts. For every APOM $M$, let $\mathbb{P}[M^\tau(x)=y]$ be the probability that on input $x\in\calB^*$, the $\tau$-POM $M$ halts with $y\in\calB^*$ written on the output tape. Then, we can define the policy $\Mvp_M^\tau:\TurnSet^*\to\Delta'\calA$ and environment $\Mve_M^\tau:\TurnSet^*\times\calA\to\Delta'\calE$ as follows:
$$\Mvp_M^\tau(a|\HistA)=
\mathbb{P}[M^\tau(\langle \HistA\rangle)=\langle a\rangle]\,,\quad\text{and}\quad\Mve_M^\tau(e|\HistA,a)=
\mathbb{P}[M^\tau(\langle \HistA,a\rangle)=\langle e\rangle]\,.$$
We say that a policy $\Mvp$ is $\tau$-POM-implementable if there exists $M\in\calMapom$ such that $\Mvp_M^\tau=\Mvp$. Similarly, we say that an environment $\Mve$ is $\tau$-POM-implementable if there exists $M\in\calMapom$ such that $\Mve_M^\tau=\Mve$. We denote the class of all $\tau$-POM-implementable policies as $\calMpol^\tau$ and the class of all $\tau$-POM-implementable environments as $\calMenv^\tau$.

Now for a universal monotone Turing machine $U$, define the mixtures $\Mmp_{U,\tau}$ and $\Mme_{U,\tau}$ as follows:
$$\Mmp_{U,\tau}=\sum_{M\in\calMapom}2^{-K_U(\langle M\rangle)}\Mvp_M^\tau\,,$$
and
$$\Mme_{U,\tau}=\sum_{M\in\calMapom}2^{-K_U(\langle M\rangle)}\Mve_M^\tau\,.$$

By defining
$$w_{\mathrm{pol}}^{U,\tau}(\Mvp)=\sum_{\substack{M\in\calMapom:\\\Mvp_M^\tau=\Mvp}}2^{-K_U(\langle M\rangle)}\,,\quad\forall\Mvp\in\calMpol^\tau\,,$$
and
$$w_{\mathrm{env}}^{U,\tau}(\Mve)=\sum_{\substack{M\in\calMapom:\\\Mve_M^\tau=\Mve}}2^{-K_U(\langle M\rangle)}\,,\quad\forall\Mve\in\calMenv^\tau\,,$$
we get
$$\Mmp_{U,\tau}=\sum_{\Mvp\in\calMpol^\tau}w_{\mathrm{pol}}^{U,\tau}(\Mvp)\Mvp\quad\text{and}\quad\Mme_{U,\tau}=\sum_{\Mve\in\calMenv^\tau}w_{\mathrm{env}}^{U,\tau}(\Mve)\Mve\,.$$

\begin{lemma}
    \label{lem:common-pol-env-machine}
    Let $\tau$ be an arbitrary probabilistic oracle. For every $\Mvp\in\calMpol^\tau$ and every $\Mve\in\calMenv^\tau$, there exists an APOM $M_{\Mvp,\Mve}$ such that the $\tau$-POM $M_{\Mvp,\Mve}^\tau$ implements both $\Mvp$ and $\Mve$, i.e., $\Mvp_{M_{\Mvp,\Mve}}^\tau=\Mvp$ and $\Mve_{M_{\Mvp,\Mve}}^\tau=\Mve$.
\end{lemma}
\begin{proof}
    Let $M_p$ and $M_e$ be two APOMs such that $\Mvp_{M_p}=\Mvp$ and $\Mve_{M_e}=\Mve$. Then we can design an APOM $M_{\Mvp,\Mve}$ such that on input $x\in\calB^*$ it does the following:
    \begin{itemize}
        \item If $x=\langle\HistA\rangle$ for some $\HistA\in\TurnSet^*$, then it simulates $M_p$ on input $x$ and writes the output of the simulated $M_p$ on its output tape.
        \item If $x=\langle\HistA,a\rangle$ for some $(\HistA,a)\in\TurnSet^*\times\calA$, then it simulates $M_e$ on input $x$ and writes the output of the simulated $M_e$ on its output tape.
        \item Otherwise, it loops forever.
    \end{itemize}
    Obviously, we have $\Mvp_{M_{\Mvp,\Mve}}^\tau=\Mvp_{M_p}^\tau=\Mvp$ and $\Mve_{M_{\Mvp,\Mve}}^\tau=\Mve_{M_e}^\tau=\Mve$.
\end{proof}

The following lemma shows that for every sufficiently big finite set $\mathcal{S}_0\subset\calB^*$ of binary strings, and every injective computable function $F:\calB^*\to\calB^*$ that keeps $\calS_0$ invariant, i.e., $F(x)=x$ for all $x\in\calS$, we can always design a universal monotone Turing machine $U$ such that (i) $\mathcal{S}_0$ is very likely under the Solomonoff prior, and (ii) for every (potentially infinite) subset $\calS\subseteq\calB^*$ of binary strings which is closed under $F$, i.e., $F(\calS)=\{F(x):x\in\calS\}\subseteq\calS$, the set $\calS\cap F(\calB^*)=\calS\setminus(\calS\setminus F(\calB^*))$ is almost as likely as $\calS$ according to the induced Solomonoff distribution.

\begin{lemma}
    \label{lem:choose-UTM-to-make-Fx-likely}
    Let $\eps>0$, and let $\calS_0$ be an arbitrary finite set of binary strings containing at least $\lceil\log(2/\epsilon)\rceil$ elements. Let $F:\calB^*\to\calB^*$ be a an arbitrary computable function for which $F(x)=x$ for all $x\in\calS_0$. There exists a universal monotone Turing machine $U$ such that
    \begin{enumerate}
        \item The set $\calS_0$ has at least $1-\epsilon$ probability according to the Solomonoff distribution:
        $$\sum_{x\in\calS_0}2^{-K_U(x)}\geq 1-\epsilon\,.$$
        \item For every subset $\calS\subseteq\calB^*$ that is closed under $F$, i.e.,
    $$F(\calS)=\{F(x):x\in\calS\}\subseteq\calS\,,$$
    and every weight function $w:\calB^*\to[0,1]$ that is non-decreasing under $F$, i.e., $w(F(x))\geq w(x)$ for all $x\in\calB^*$, we have
    $$\sum_{x\in \calS\cap F(\calB^*)}w(x)2^{-K_U(x)}\geq (1-\epsilon)\sum_{x\in \calS}w(x)2^{-K_U(x)}.$$

    In particular,
    $$\sum_{x\in \calS\cap F(\calB^*)}2^{-K_U(x)}\geq (1-\epsilon)\sum_{x\in \calS}2^{-K_U(x)}.$$
    \end{enumerate}
\end{lemma}
\begin{proof}
Let $\ell=\lceil\log(2/\epsilon)\rceil$, and let $x_1,\ldots,x_\ell$ be the elements of $\calS_0$. For $1\leq i\leq\ell$, let us fix some arbitrary monotone Turing machine $T_{x_i}$ which writes $x_i$ on its output tape (without reading any bit from its input tape) and then halts.

For every monotone Turing machine $T$, define the monotone Turing machine $S_F(T)$ which simulates $T$ until the simulated machine $T$ halts:
\begin{itemize}
    \item Prior to the halting of the simulated machine $T$, the real machine $S_F(T)$ does not write any bit on its output tape. We only store the value of the (simulated) output tape of the simulated $T$ on the working tape of $S_F(T)$.
    \item When the simulated $T$ attempts to read a bit from its input tape, we read the bit from the actual input tape of $S_F(T)$.
    \item If the simulated $T$ does not halt, then $S_F(T)$ loops forever without writing any bit on its output tape (even if the simulated $T$ writes some bits on its simulated output tape).
    \item If the simulated $T$ halts, the machine $S_F(T)$ computes $F(x)$, where $x$ is the output of the simulated $T$, and then writes $F(x)$ on the output tape.
\end{itemize}

Clearly, $S_F(T)(p)=F(T(p))$ for every $p\in\calB^*$. Furthermore, we can computably get a description $\langle S_F(T)\rangle$ of $S_F(T)$ from a description $\langle T\rangle$ of $T$.

Now let $T_1',\ldots,T_n',\ldots$ be a computable enumeration of all monotone Turing machines, i.e., the mapping $n\mapsto \langle T_n'\rangle$ is computable.

We define another enumeration $T_1,\ldots,T_n,\ldots$ as follows:
\begin{align*}
    T_{n}=\begin{cases}
        T_{x_n}\quad&\text{if }1\leq n\leq \ell\,,\\
        T'_{\frac{n-\ell}\ell}\quad&\text{if }n>\ell\text{ and }\ell\text{ divides }n\,,\\
        S_F( T'_{\lceil \frac{n-\ell}{\ell}\rceil})\quad&\text{otherwise}\,.
    \end{cases}
\end{align*}
Clearly, the mapping $n\mapsto \langle T_n\rangle$ is also computable.

Let $c:\mathbb{N}\to\calB^*$ be the complete prefix-free encoding of positive natural numbers where $c(n)$ is the bitstring having $n-1$ ones followed by a zero.

We can then define the machine $U$ so that $U(c(n)p)=T_n(p)$ as follows: 
\begin{itemize}
    \item The machine $U$ starts reading the input tape bit by bit until it reads the first zero.
    \item Let $n$ be the number of bits that were read. The machine then computes the representation $\langle T_n\rangle$ of the $n$-th machine $T_n$ in the enumeration and then start simulating $T_n$.
\end{itemize}

This machine $U$ is a universal monotone Turing machine by construction (as it can simulate any machine in the enumeration $(T_n)_{n\geq 1}$ which covers all monotone Turing machine).

For every $1\leq i\leq\ell$, we have $U(c(i))=T_i(\varepsilon)=T_{x_i}(\varepsilon)=x_i$, and hence $K_U(x_i)\leq l(c(i))=i$. Therefore,
$$\sum_{x\in\calS_0}2^{-K_U(x)}=\sum_{i=1}^\ell 2^{-K_U(x_i)}\geq\sum_{i=1}^\ell 2^{-i}=1-2^{-\ell}\geq 1-\epsilon\,.$$

Now let $\calS$ be some arbitrary set which is closed under $F$, i.e., $F(\calS)\subseteq\calS$. Let $x\in\calS\setminus F(\calB^*)$. Since $F(x')=x'$ for all $x'\in\calS_0$, we must have $x\notin\calS_0$ and hence for all $1\leq i\leq \ell$, we must have $x\neq x_i=T_i(\varepsilon)=U(c(i))$. Let $p_x$ be an arbitrary bit-string for which $U(p_x)=x$. From the structure of $U$, we know that there must exist $n>0$ such that $p_x=c(n)p'_x$ for some $p'_x\in\calB^*$. Hence, $x=U(p_x)=U(c(n)p'_x)=T_n(p'_{x})$. Since $x\neq x_i=T_i(\varepsilon)=U(c(i))$ for all $1\leq i\leq\ell$, we must have $n>\ell$.

If $\ell$ does not divide $n$, then $T_{n}=S_F( T'_{\lceil (n-\ell)/\ell\rceil})$ and hence $$x=T_n(p'_x)=S_F( T'_{\lceil (n-\ell)/\ell\rceil})(p'_x)=F(T'_{\lceil (n-\ell)/\ell\rceil}(p'_x))\,,$$ which is a contradiction because $x\in\calS\setminus F(\calB^*)$. Therefore, $\ell$ must divide $n$. In particular,
$$x=T_n(p'_{x})=T'_{(n-\ell)/\ell}(p'_x)\,.$$
Now if we define $\tilde{p}_x=c(n-\ell+1)p'_x$, we get
\begin{align*}
    U(\tilde{p}_x)&=U(c(n-\ell+1)p'_x)=T_{n-\ell+1}(p'_{x})=S_F(T'_{\lceil (n-2\ell+1)/\ell\rceil})(p'_x)\\
    &=S_F(T'_{(n-\ell)/\ell})(p'_x)=F(T'_{(n-\ell)/\ell}(p'_x))=F(x)\,.
\end{align*}
Therefore, for every bit-string $p_x\in\calB^*$ for which $U(p_x)=x$, there exists a shorter bit-string $\tilde{p}_x\in\calB^*$ such that $U(\tilde{p}_x)=F(x)$ and $l(\tilde{p}_x)=l(p_x)-\ell+1$, hence,
$$K_U(F(x))\leq K_U(x)-\ell+1\,.$$

Thus, for every weight function $w:\calB^*\to[0,1]$ that is non-decreasing under $F$, we have
\begin{align*}
    \sum_{x\in \calS\setminus F(\calB^*)} w(x) 2^{-K_U(x)}&\leq \sum_{x\in \calS\setminus F(\calB^*)} w(x) 2^{-K_U(F(x))-\ell+1}\stackrel{(\ast)}\leq \epsilon \sum_{x\in \calS\setminus F(\calB^*)} w(x) 2^{-K_U(F(x))}\\
    &\stackrel{(\dagger)}\leq\epsilon \sum_{x\in \calS\setminus F(\calB^*)} w(F(x)) 2^{-K_U(F(x))}\stackrel{(\ddagger)}= \epsilon \sum_{y\in F(\calS\setminus F(\calB^*))}w(y)2^{-K_U(y)}\\
    &\stackrel{(\diamond)}\leq \epsilon \sum_{y\in \calS}w(y)2^{-K_U(y)}\,,
\end{align*}
where $(\ast)$ follows from our choice of $\ell$, $(\dagger)$ follows from the fact that $w$ is non-decreasing under $F$, $(\ddagger)$ follows from the injectivity of $F$, and $(\diamond)$ follows from the fact that $\calS$ is closed under $F$, and hence $F(\calS\setminus F(\calB^*))\subseteq F(\calS)\subseteq\calS$.

We conclude that
\begin{align*}
    \sum_{x\in \calS\cap F(\calB^*)}w(x)2^{-K_U(x)}&=\sum_{x\in \calS\setminus(\calS\setminus F(\calB^*))}w(x)2^{-K_U(x)}\\
    &=\sum_{x\in \calS}w(x)2^{-K_U(x)}-\sum_{x\in \calS\setminus F(\calB^*)}w(x)2^{-K_U(x)} \\
    &\geq (1-\epsilon)\sum_{x\in \calS}w(x)2^{-K_U(x)}.
\end{align*}
\end{proof}

The following lemma leverages \cref{lem:choose-UTM-to-make-Fx-likely} and shows how we can design a universal monotone Turing machine $U$ for which $\Mmp_{U,\tau}$ $\eps$-approximates $\Mvp_{\mathrm{up}}$ and $\Mme_{U,\tau}$ $\eps$-approximates $\Mge_{R,k}$.

\begin{lemma}
    \label{lem:decoupled-solomonoff-mixture-eps-approximates-MgeRk-andMvpup}
    For every $\epsilon>0$, there exists a monotone Turing machine $U$ such that $\Mme_{U,\tau}$ $\epsilon$-approximates $\Mge_{R,k}$ on $\TurnSet^*_{\Mge_{R,k}}$, and $\Mmp_{U,\tau}$ $\epsilon$-approximates $\Mvp_{\mathrm{up}}$ on $\TurnSet^{*\text{-}\mathrm{up}\text{-}k\text{-}\mathrm{down}\text{-}*\text{-}\mathrm{up}}_{\Mge_{R,k}} \cup \TurnSet^{*\text{-}\mathrm{up}}_{\Mge_{R,k}}$.
\end{lemma}
\begin{proof}
    Let $\ell=\lceil\log(4/\epsilon)\rceil$, and let $M_{\Mvp_{\mathrm{up}},\Mge_{R,k}}$ be an APOM that implements both $\Mvp_{\mathrm{up}}$ and $\Mge_{R,k}$, as in \cref{lem:common-pol-env-machine}. Let $M_1,\ldots,M_\ell$ be $\ell$ APOMs which are syntactically different, but which are computationally equivalent to $M_{\Mvp_{\mathrm{up}},\Mge_{R,k}}$. For example, for every $1\leq i\leq \ell$, the machine $M_i$ starts with executing a loop of $i$ iterations in which it does some useless computation and afterwards it starts simulating $M_{\Mgp_{\mathrm{up}},\Mge_{R,k}}$. What matters is that $\Mvp_{M_i}=\Mvp_{\mathrm{up}}$ and $\Mve_{M_i}=\Mge_{R,k}$ for all $1\leq i\leq \ell$. Let $$\mathcal{S}_0=\left\{\langle M_i\rangle:1\leq i\leq \ell\right\}\,.$$

    Now for every APOM $M\in\calMapom$, define the machine $S_k(M)$ as follows:

    \begin{algorithm}[H]
    \KwIn{$x=\langle \HistM[t]\rangle$ or $x=\langle \HistM[t],a_t\rangle$}
    \KwOut{Action $a_{t}$ or Percept $e_t$}
    Parse the input $x$ using the complete prefix-free encodings of $\calA$ and $\calE$\;
    \If{$x=\langle\HistM\rangle$}{
    Let $a_{<t}$ be the action sequence from $\HistM[t]$\;
    $\mathrm{streak}\text{\_}\mathrm{found} \leftarrow \text{false}$\;
    \For{$t' \leftarrow 1$ \KwTo $t - k$}{
        \If{$a_{t':t'+k-1} = (\mathrm{down})^k$}{
            $\mathrm{streak}\text{\_}\mathrm{found} \leftarrow \text{true}$\;
            \textbf{break}\;
        }
    }
    \If{$\mathrm{streak}\text{\_}\mathrm{found}$}{
        \KwRet{`$\mathrm{up}$'}
    }
    \Else{
        $\langle a_{t}\rangle \leftarrow M(\langle \HistM[t]\rangle)$ \tcp*{Simulate $M$ on input $\langle\HistM\rangle$}
        \KwRet{$\langle a_{t}\rangle$}
    }
    }
    \Else{
        \tcp{I.e., $x=\langle \HistM[t],a_t\rangle$}
        Let $a_{<t}$ be the action sequence from $\HistM[t]$\;
        \tcp{We return a percept according to $\Mge_{R,k}$}
        \KwRet{$\langle o_{0},r_t(a_{1:t})\rangle$}
    }
    \caption{The "reverting" APOM $S_k(M)$}
    \label{alg:reverting-apom}
\end{algorithm}
Notice that $\Mve_{S_k(M)}^\tau=\Mge_{R,k}$ for every $M\in\calMapom$. We assume that the ``boiler-plate" structure of $S_k$ is done in a way that ensures that $S_k(M)\neq M_i$ for all $1\leq i\leq \ell$, e.g., we can just add some useless loop that does not exist in any of the machines $(M_i)_{1\leq i\leq \ell}$.

Now define the function $F:\calB^*\to\calB^*$ as
\begin{equation}
    \label{eq:definition-F-in-terms-of-Sk}
    F(x)=\begin{cases}
        \langle S_k(M)\rangle \quad&\text{if } x=\langle M\rangle\text{ for some }M\in\calMapom\setminus\{M_1,\ldots,M_\ell\}\,,\\
        x\quad&\text{otherwise},
    \end{cases}
\end{equation}
Clearly, $F$ is computable and injective. Let $U$ be the universal monotone Turing machine of \cref{lem:choose-UTM-to-make-Fx-likely} applied to $\calS_0=\{\langle M_1\rangle,\ldots,\langle M_\ell\rangle\}$,  $F$ and $\frac{\epsilon}{2}$. We have
\begin{equation}
    \label{eq:lower-bound-for-M1-to-Mell}
    \sum_{i=1}^\ell 2^{-K_U(\langle M_i\rangle)}=\sum_{x\in\calS_0} 2^{-K_U(x)}\geq 1-\frac\epsilon2
\end{equation}

Now let $$\mathcal{S}:=\left\{\langle M\rangle:M\in\calMapom\setminus\{M_1,\ldots,M_\ell\}\right\}=\left\{\langle M\rangle:M\in\calM_\calS\right\}\,,$$
where
$$\calM_\calS:=\calMapom\setminus\{M_1,\ldots,M_\ell\}\,.$$
Clearly, $\mathcal{S}$ is closed \footnote{Recall that we designed the wrapper $S_k$ in a way that ensures that $S_k(M)\neq M_i$ for every $M\in\calMapom$ and every $1\leq i\leq\ell$.} under $F$, i.e., $F(\mathcal{S})\subseteq \mathcal{S}$. Furthermore, if $x\notin\calS$ then $F(x)=x\notin\calS$, hence
$$\calS\cap F(\calB^*)=F(\calS)\,,$$
and from \cref{lem:choose-UTM-to-make-Fx-likely} we have
$$\sum_{x\in F(\calS)}w(x)2^{-K_U(x)}\geq\left(1-\frac{\epsilon}2\right)\sum_{x\in \calS}w(x)2^{-K_U(x)}\geq (1-\epsilon)\sum_{x\in \calS}w(x)2^{-K_U(x)},$$
for any weight function $w:\calB^*\to[0,1]$ that is non-decreasing under $F$. We can reformulate this by saying that for any weight function $w:\calM_\calS\to[0,1]$ that is non-decreasing under $S_k$, i.e., it satisfies $w(S_k(M))\geq w(M)$ for every $M\in\calM_\calS$, we have
\begin{equation}
    \label{eq:weighted-lower-bound-on-machines}
    \sum_{M\in\calM_\calS}w(S_k(M))2^{-K_U(\langle S_k(M)\rangle)}\geq (1-\epsilon)\sum_{M\in\calM_\calS}w(M)2^{-K_U(\langle M\rangle)}\,,
\end{equation}
Given these properties of the chosen universal monotone Turing machine $U$, we proceed to prove the lemma.

We notice first that since $\Mvp_{M_i}=\Mvp_{\mathrm{up}}$ and $\Mve_{M_i}=\Mge_{R,k}$ for all $1\leq i\leq \ell$, we have
\begin{align*}
    w_{\mathrm{env}}^{U,\tau}(\Mge_{R,k})&=\sum_{\substack{M\in\calMapom:\\ \Mve_M^\tau=\Mge_{R,k}}} 2^{-K_U(\langle M\rangle)}\geq \sum_{i=1}^\ell 2^{-K_U(\langle M_i\rangle)}\stackrel{(\ast)}\geq 1-\frac\epsilon2\,,
\end{align*}
where $(\ast)$ follows from \eqref{eq:lower-bound-for-M1-to-Mell}
and similarly we can show that
\begin{align*}
    w_{\mathrm{pol}}^{U,\tau}(\Mvp_{\mathrm{up}})\geq 1-\frac\epsilon2\,.
\end{align*}

We will show by induction on $t\geq 0$ that for every $\Hist\in\TurnSet^{t}\cap\TurnSet_{\Mge_{R,k}}^*$, we have
$$w(\Mge_{R,k}|\Hist)\geq 1-\frac\epsilon2\,.$$
For $t=0$, we have $w(\Mge_{R,k}|\Hist)=w(\Mge_{R,k}|\varepsilon)=w(\Mge_{R,k})\geq 1-\frac\epsilon2$. Now let $t>0$, and assume that the claim is true for $t-1$. We have from \cref{prop:behavior-decoupled-beliefs} that
$$w(\Mge_{R,k}|\Hist)=w(\Mge_{R,k}|\HistM)\frac{\Mge_{R,k}(e_t|\HistM a_t)}{\Mme_{U,\tau}(e_t|\HistM a_t)}\stackrel{(\ast)}\geq w(\Mge_{R,k}|\HistM)\stackrel{(\dagger)}{\geq}1-\frac\epsilon2\,,$$
where $(\ast)$ follows from the fact that $\Mge_{R,k}(e_t|\HistM a_t)=1$ for $\Hist\in\TurnSet_{\Mge_{R,k}}^*$ and $(\dagger)$ follows from the induction hypothesis\footnote{Note that $\Hist\in\TurnSet_{\Mge_{R,k}}^*$ implies that $\HistM\in\TurnSet_{\Mge_{R,k}}^*$ as well.}. Hence the claim is true for all $t\geq 0$.

Now fix $t>0$ and let $\HistM\in\TurnSet^{t-1}\cap \TurnSet_{\Mge_{R,k}}^*$. For every $a_t\in\calA$ and every $e_t\in\calE$, we have from \cref{prop:behavior-decoupled-beliefs} that
$$\Mme_{U,\tau}(e_t|\HistM a_t)=\sum_{\Mve\in\calMenv^\tau} w(\Mve|\HistM)\Mve(e_t|\HistM a_t)\,,$$
hence
\begin{align*}
    |\Mme_{U,\tau}(e_t|\HistM a_t)-\Mge_{R,k}(e_t|\HistM a_t)|&\leq |w(\Mge_{R,k}|\HistM)-1|\Mge_{R,k}(e_t|\HistM a_t)+\sum_{\Mve\in\calMenv^\tau\setminus\{\Mge_{R,k}\}} w(\Mve|\HistM)\Mve(e_t|\HistM a_t)\\
    &\leq |w(\Mge_{R,k}|\HistM)-1|+\sum_{\Mve\in\calMenv^\tau\setminus\{\Mge_{R,k}\}} w(\Mve|\HistM)\leq 2(1-w(\Mge_{R,k}|\HistM))\\
    &\leq 2\frac{\epsilon}{2}=\epsilon\,.
\end{align*}
We conclude that $\Mme_{U,\tau}$ $\epsilon$-approximates $\Mge_{R,k}$ on $\TurnSet^*_{\Mge_{R,k}}$.

Similarly, one can show by induction that for histories $\HistM \in \TurnSet^{*\text{-}\mathrm{up}}_{\Mge_{R,k}}$ containing only up actions, that $w(\Mvp_{\mathrm{up}}\mid \HistM) \geq  w_{\mathrm{pol}}^{U,\tau}(\Mvp_{\mathrm{up}})\geq 1-\frac\epsilon2$, and hence that for any $\HistM \in \TurnSet^{*\text{-}\mathrm{up}}_{\Mge_{R,k}}$ and $a_t \in \calA$ we have that $\Mmp_{U,\tau}$ $\epsilon$-approximates $\Mvp_{\mathrm{up}}$ on $\TurnSet^{*\text{-}\mathrm{up}}_{\Mge_{R,k}}$:
$$|\Mmp_{U,\tau}(a_t \mid \HistM) - \Mvp_{\mathrm{up}}(a_t \mid \HistM)| \leq \epsilon\,.$$

Now we turn to showing that $\Mmp_{U,\tau}$ $\epsilon$-approximates $\Mvp_{\mathrm{up}}$ on $\TurnSet^{*\text{-}\mathrm{up}\text{-}k\text{-}\mathrm{down}\text{-}*\text{-}\mathrm{up}}_{\Mge_{R,k}}$. Note that the shortest history in $\TurnSet^{*\text{-}\mathrm{up}\text{-}k\text{-}\mathrm{down}\text{-}*\text{-}\mathrm{up}}_{\Mge_{R,k}}$ has length $k$. Let $t>k$ and let $\HistM\in\TurnSet^{t-1}\cap \TurnSet^{*\text{-}\mathrm{up}\text{-}k\text{-}\mathrm{down}\text{-}*\text{-}\mathrm{up}}_{\Mge_{R,k}}$. Since the shortest history in $\TurnSet^{*\text{-}\mathrm{up}\text{-}k\text{-}\mathrm{down}\text{-}*\text{-}\mathrm{up}}_{\Mge_{R,k}}$ has length $k$, we must have $t\geq k>0$. Let $a_t\in\calA$ and for every $0< T\leq t$, let

\begin{align*}
    \Mmp_{U,\tau}(a_{1:T} \mid\mid e_{<T}):=\prod_{t'=1}^{T} \Mmp_{U,\tau}(a_{t'} \mid \HistM[t'])\,,
\end{align*}
so that
\begin{equation}
    \label{eq:Mmutau-as-a-chronological-fraction}
    \Mmp_{U,\tau}(a_{t} \mid \HistM[t])=\frac{\Mmp_{U,\tau}(a_{1:t} \mid\mid e_{<t})}{\Mmp_{U,\tau}(a_{<t} \mid\mid e_{<t-1})}\,.
\end{equation}

From \cref{prop:behavior-decoupled-beliefs}, we have
    $$\Mmp_{U,\tau}(a_T \mid \HistM[T])=\sum_{\Mvp\in\calMpol^\tau}w_{\mathrm{pol}}^{U,\tau}(\Mvp \mid \HistM[T])\Mvp(a_T \mid \HistM[T])\,,$$
    where the posterior $w_{\mathrm{pol}}^{U,\tau}(\Mvp \mid \HistM[T])$ is updated recursively as 
    \[
    w_{\mathrm{pol}}^{U,\tau}(\Mvp \mid \HistM[T]) = \frac{w_{\mathrm{pol}}^{U,\tau}(\Mvp \mid \HistM[T-1]) \Mvp(a_{T-1} \mid \HistM[T-1])}{\Mmp_{U,\tau}(a_{T-1} \mid \HistM[T-1])}\,.
    \]
    By recursive application (and starting from $w_{\mathrm{pol}}^{U,\tau}(\Mvp \mid \varepsilon) = w_{\mathrm{pol}}^{U,\tau}(\Mvp)$), this gives:
    \[
    w_{\mathrm{pol}}^{U,\tau}(\Mvp \mid \HistM[T]) = \frac{w_{\mathrm{pol}}^{U,\tau}(\Mvp) \prod_{t'=1}^{T-1} \Mvp(a_{t'} \mid \HistM[t'])}{\prod_{t'=1}^{T-1} \Mmp_{U,\tau}(a_{t'} \mid \HistM[t'])} = \frac{w_{\mathrm{pol}}^{U,\tau}(\Mvp) \Mvp(a_{1:T-1} \mid\mid e_{<T-1})}{\Mmp_{U,\tau}(a_{1:T-1} \mid\mid e_{<T-1})}\,.
    \]
    Therefore,
\begin{align*}
    \Mmp_{U,\tau}(a_{1:T} \mid\mid e_{<T})&=\Mmp_{U,\tau}(a_{1:T-1} \mid\mid e_{<T-1}) \cdot \Mmp_{U,\tau}(a_T \mid \HistM[T])\\
    &=\Mmp_{U,\tau}(a_{1:T-1} \mid\mid e_{<T-1})\sum_{\Mvp\in\calMpol^\tau}w_{\mathrm{pol}}^{U,\tau}(\Mvp \mid \HistM[T])\Mvp(a_T \mid \HistM[T])\\
    &=\Mmp_{U,\tau}(a_{1:T-1} \mid\mid e_{<T-1})\sum_{\Mvp\in\calMpol^\tau}\frac{w_{\mathrm{pol}}^{U,\tau}(\Mvp) \Mvp(a_{1:T-1} \mid\mid e_{<T-1})}{\Mmp_{U,\tau}(a_{1:T-1} \mid\mid e_{<T-1})}\Mvp(a_T \mid \HistM[T])\\
    &=\sum_{\Mvp\in\calMpol^\tau}w_{\mathrm{pol}}^{U,\tau}(\Mvp) \Mvp(a_{1:T} \mid\mid e_{<T})=\sum_{\Mvp\in\calMpol^\tau}\sum_{\substack{M\in\calMapom:\\\Mvp_M^\tau=\Mvp}}2^{-K_U(\langle M\rangle)} \Mvp(a_{1:T} \mid\mid e_{<T})\\
    &=\sum_{M\in\calMapom}2^{-K_U(\langle M\rangle)}\Mvp_M^\tau(a_{1:T} \mid\mid e_{<T})\,.
\end{align*}

Combining this with \eqref{eq:Mmutau-as-a-chronological-fraction}, we get
\begin{align}
    \Mmp_{U,\tau}(a_{t} \mid \HistM[t])&=\frac{\Mmp_{U,\tau}(a_{1:t} \mid\mid e_{<t})}{\Mmp_{U,\tau}(a_{<t} \mid\mid e_{<t-1})} = \frac{\sum_{M \in \calMapom} 2^{-K_U(\langle M \rangle)} \Mvp_M^\tau(a_{1:t} \mid\mid e_{<t})}{\sum_{M \in \calMapom} 2^{-K_U(\langle M \rangle)} \Mvp_M^\tau(a_{<t} \mid\mid e_{<t-1})}\nonumber\\
    &\stackrel{(\ast)}=  \frac{\sum_{M \in \calM_\calS} 2^{-K_U(\langle M \rangle)} \Mvp_M^\tau(a_{1:t} \mid\mid e_{<t})}{\sum_{M \in \calM_\calS} 2^{-K_U(\langle M \rangle)} \Mvp_M^\tau(a_{<t} \mid\mid e_{<t-1})}\label{eq:Mmutau-as-a-chronological-fraction-explicit}\,,
\end{align}
where in $(\ast)$ we used the fact that for $M\in\calMapom\setminus\calM_\calS$, we have $M=M_i$ for some $1\leq i\leq\ell$ and hence $\Mvp_M^\tau=\Mvp_{\mathrm{up}}$, which means that $\Mvp_M^\tau(a_{1:t} \mid\mid e_{<t})=\Mvp_M^\tau(a_{<t} \mid\mid e_{<t-1})=0$ because $a_{<t}$ contains a substring of $k$ down actions.

Now assume that $a_{t} = \text{up}$, and define
$$\calM_{\calS,S_k}=\left\{S_k(M):M\in\calM_\calS\right\}\subseteq\calM_\calS\,,$$
and notice that for every $T> t$. We have
\begin{equation*}
\begin{aligned}
     \sum_{M \in \calM_{\calS}} 2^{-K_U(\langle M \rangle)} \Mvp_M^\tau(a_{1:t} \mid\mid e_{<t})     &\geq \sum_{M' \in \calM_{\calS,S_k}} 2^{-K_U(\langle M' \rangle)} \Mvp_{M'}^\tau(a_{1:t} \mid\mid e_{<t}) \\ 
     &\stackrel{(a)}{=}\sum_{M \in \calM_{\calS}}2^{-K_U(\langle S_k(M) \rangle)} \Mvp_{S_k(M)}^\tau(a_{1:t} \mid\mid e_{<t})\\
     &=\sum_{M \in \calM_{\calS}} 2^{-K_U(\langle S_k(M) \rangle)} \Mvp_{S_k(M)}^\tau(a_{<t} \mid\mid e_{<t-1})\Mvp_{S_k(M)}^\tau(a_{t} \mid \HistM[t])\\
    &\stackrel{(b)}{=} \sum_{M \in \calM_{\calS}} 2^{-K_U(\langle S_k(M) \rangle)} \Mvp_{S_k(M)}^\tau(a_{<t} \mid\mid e_{<t-1})\,.  \\
    %&= (1-\epsilon)\sum_{M \in \calM_{\calS}} 2^{-K_U(\langle M \rangle)} \Mvp_{M}^\tau(a_{<t} \mid\mid e_{<t-1})\,,
\end{aligned}
\end{equation*}
where in $(a)$ we used the definition of $\calM_{\calS,S_k}$, in $(b)$ we used the fact that $\Mvp_{S_k(M)}^\tau(a_{t}\mid\HistM[t]) = 1$ because $a_{t}=\mathrm{up}$ and $\HistM\in\TurnSet^{*\text{-}\mathrm{up}\text{-}k\text{-}\mathrm{down}\text{-}*\text{-}\mathrm{up}}_{\Mge_{R,k}}$ and hence contains the sequence $(\mathrm{down})^k$ as a substring (which means that $S_k(M)$ would deterministically return `up' on input $\HistM[t']$).

By defining the weight function $w:\calM_\calS\to[0,1]$ as $w(M)=\Mvp_{M}^\tau(a_{<t} \mid\mid e_{<t-1})$ for all $M\in\calM_\calS$, we get
\begin{equation}
\label{eq:reverting-lower-bound-in-terms-of-w}
\begin{aligned}
     \sum_{M \in \calM_{\calS}} 2^{-K_U(\langle M \rangle)}\Mvp_M^\tau(a_{1:t} \mid\mid e_{<t}) \geq \sum_{M \in \calM_{\calS}} 2^{-K_U(\langle S_k(M) \rangle)} w(S_k(M))\,.
\end{aligned}
\end{equation}

We will show that $w$ is non-decreasing under $S_k$, i.e, $w(S_k(M))\geq w(M)$ for all $M\in\calM_\calS$. Let $1\leq t'\leq t-k$ be such that $a_{t':t'+k-1}=(\mathrm{down})^k$, which exists because $\HistM\in\TurnSet^{*\text{-}\mathrm{up}\text{-}k\text{-}\mathrm{down}\text{-}*\text{-}\mathrm{up}}_{\Mge_{R,k}}$. We have
\begin{align*}
    w(M)&=\Mvp_{M}^\tau(a_{<t} \mid\mid e_{<t-1})=\prod_{i=1}^{t-1}\Mvp_{M}^\tau(a_{i}|\HistM[i])\stackrel{(\ast)}\leq \prod_{i=1}^{t-1}\Mvp_{S_k(M)}^\tau(a_{i}|\HistM[i])=\Mvp_{S_k(M)}^\tau(a_{<t} \mid\mid e_{<t-1})=w(S_k(M))\,,
\end{align*}
where $(\ast)$ follows from the following two facts:
\begin{itemize}
    \item If $i<t'+k$, then $S_k(M)$ produces the output of $M$ on input $\HistM[i]$ because $a_{<i}$ does not contain a streak of $k$ `down' actions. Therefore, $\Mvp_{S_k(M)}^\tau(a_i|\HistM[i])=\Mvp_{M}^\tau(a_{i}|\HistM[i])$.
    \item If $i\geq t'+k$, then $a_{<i}$ contains a streak of $k$ downs and $a_i$ must be the up action,\footnote{This follows from the structure of histories in $\TurnSet^{*\text{-}\mathrm{up}\text{-}k\text{-}\mathrm{down}\text{-}*\text{-}\mathrm{up}}_{\Mge_{R,k}}$.} hence $\Mvp_{S_k(M)}^\tau(a_{i}|\HistM[i])=1\geq \Mvp_{M}^\tau(a_{i}|\HistM[i])$.
\end{itemize}
We conclude that $w$ is non-decreasing under $w$, and hence it follows from \eqref{eq:reverting-lower-bound-in-terms-of-w} and \eqref{eq:weighted-lower-bound-on-machines} that
\begin{equation*}
\label{eq:reverting-lower-bound-after-streak}
\begin{aligned}
     \sum_{M \in \calM_{\calS}} 2^{-K_U(\langle M \rangle)}\Mvp_M^\tau(a_{1:t} \mid\mid e_{<t})    &\geq \sum_{M \in \calM_{\calS}} 2^{-K_U(\langle S_k(M) \rangle)} w(S_k(M))  \\
     &\geq (1-\epsilon) \sum_{M \in \calM_{\calS}} 2^{-K_U(\langle M \rangle)} w(M)\\
     &= (1-\epsilon) \sum_{M \in \calM_{\calS}} 2^{-K_U(\langle M \rangle)} \Mvp_M^\tau(a_{<t} \mid\mid e_{<t-1})\,.
\end{aligned}
\end{equation*}

Substituting this back into the expression of $\Mmp_{U,\tau}(a_{t} \mid \HistM[t])$ in \eqref{eq:Mmutau-as-a-chronological-fraction-explicit}, we get
\[
\Mmp_{U,\tau}(\mathrm{up} \mid \HistM[t])=\Mmp_{U,\tau}(a_{t} \mid \HistM[t])\geq 1-\epsilon= \Mvp_{\mathrm{up}}(\mathrm{up}|\HistM)-\epsilon\,.
\]
Now for $a\neq \mathrm{up}$, we have
$$\Mmp_{U,\tau}(a \mid \HistM[t])\leq 1- \Mmp_{U,\tau}(\mathrm{up} \mid \HistM[t])\leq \epsilon = \Mvp_{\mathrm{up}}(a|\HistM)+\epsilon\,.$$
In summary, for every $a\in\calA$, we have
$$|\Mmp_{U,\tau}(a \mid \HistM[t])- \Mvp_{\mathrm{up}}(a \mid \HistM[t])|\leq\epsilon\,.$$

Therefore, $\Mmp_{U,\tau}$ $\epsilon$-approximates $\Mvp_{\mathrm{up}}$ on $\TurnSet^{*\text{-}\mathrm{up}\text{-}k\text{-}\mathrm{down}\text{-}*\text{-}\mathrm{up}}_{\Mge_{R,k}}$.
\end{proof}

Putting everything together, we get the following theorem:

\begin{theorem}\label{thrm:counter-example-decoupled}
Consider a $k$-AIXI$^{\tau}$ agent, i.e., a decoupled Bayesian agent employing $k$-step planning for a fixed integer $k \ge 1$, with a decoupled mixture universe model $\Mmu=\Mme_{U,\tau}^{\Mmp_{U,\tau}}$ constructed from a mixture policy $\Mmp_{U,\tau}$ and a mixture environment $\Mme_{U,\tau}$ which are themselves constructed from a Solomonoff prior over the $\tau$-POM-computable policies and $\tau$-POM-computable environments, respectively. Then, there exists a computable ground-truth environment $\Mge$ and a reference universal monotone Turing machine $U$ for defining the Kolmogorov complexity $K_U$ in the Solomonoff prior, such that the agent interacting with $\Mge$ does not converge to the infinite-horizon optimal planner w.r.t. $\Mmu=\Mme_{U,\tau}^{\Mmp_{U,\tau}}$, despite the fact that this setup satisfies the grain-of-truth property (in case, e.g., of a reflective oracle). In particular, Self-AIXI does not always converge to AIXI.
\end{theorem}
\begin{proof}
    Fix $R$ so that $\gamma^{k+1}<R<\gamma^k$. Let $\epsilon>0$ be sufficiently small as indicated in \cref{cor:optimal-and-k-step-approx} and then construct $U$ as in 
    \cref{lem:decoupled-solomonoff-mixture-eps-approximates-MgeRk-andMvpup}.
\end{proof}

\subsubsection{Counterexample for embedded Bayesian agents}

For a universal monotone Turing machine $U$, define the mixture universe $\Mmu_{U,\tau}$ as follows:
$$\Mmu_{U,\tau}=\sum_{M\in\calMapom}2^{-K_U(\langle M\rangle)}\Mvu_M^\tau\,.$$

By defining
$$\calMuni^\tau=\{\Mvu_M^\tau:M\in \calMapom\}\,,$$
and
$$w_{\mathrm{uni}}^{U,\tau}(\Mvu)=\sum_{\substack{M\in\calMapom:\\\Mvu_M^\tau=\Mvu}}2^{-K_U(\langle M\rangle)}\,,\quad\forall\Mvu\in\calMuni^\tau\,,$$
we get
$$\Mmu_{U,\tau}=\sum_{\Mvu\in\calMuni^\tau}w_{\mathrm{uni}}^{U,\tau}(\Mvu)\Mvu\,.$$

We start by showing a stronger version of \cref{lem:choose-UTM-to-make-Fx-likely}.

\begin{lemma}
    \label{lem:choose-UTM-to-make-Fx-and-Fprimex-likely}
    Let $\eps>0$, and let $\calS_0$ be an arbitrary finite set of binary strings containing at least $\lceil\log(2/\epsilon)\rceil$ elements. Let $F_1:\calB^*\to\calB^*$ and $F_2:\calB^*\to\calB^*$ be two arbitrary computable functions for which $F_1(x)=F_2(x)=x$ for all $x\in\calS_0$, and $F_1(F_2(x))=F_2(F_1(x))$ for all $x\in\calB^*$. There exists a universal monotone Turing machine $U$ such that
    \begin{enumerate}
        \item The set $\calS_0$ has at least $1-\epsilon$ probability according to the Solomonoff distribution:
        $$\sum_{x\in\calS_0}2^{-K_U(x)}\geq 1-\epsilon\,.$$
        \item For every $m\in\{1,2\}$ and every subset $\calS_m\subseteq\calB^*$ that is closed under $F_m$, i.e.,
    $$F_m(\calS_m)=\{F_m(x):x\in\calS_m\}\subseteq\calS_m\,,$$
    and every weight function $w_m:\calB^*\to[0,1]$ that is non-decreasing under $F_m$, i.e., $w_m(F_m(x))\geq w_m(x)$ for all $x\in\calB^*$, we have
    $$\sum_{x\in \calS_m\cap F_m(\calB^*)}w_m(x)2^{-K_U(x)}\geq (1-\epsilon)\sum_{x\in \calS_m}w_m(x)2^{-K_U(x)}.$$

    In particular,
    $$\sum_{x\in \calS_m\cap F_m(\calB^*)}2^{-K_U(x)}\geq (1-\epsilon)\sum_{x\in \calS_m}2^{-K_U(x)}.$$
    \end{enumerate}
\end{lemma}
\begin{proof}

Let $\ell=\lceil\log(2/\epsilon)\rceil$, and let $x_1,\ldots,x_\ell$ be the elements of $\calS_0$. For $1\leq i\leq\ell$, let us fix some arbitrary monotone Turing machine $T_{x_i}$ which writes $x_i$ on its output tape (without reading any bit from its input tape) and then halts. Furthermore, define the function $F_0:\calB^*\to\calB^*$ as
$$F_0(x)=F_1(F_2(x))\stackrel{(\ast)}=F_2(F_1(x))\,,\quad\forall x\in\calB^*\,,$$
where $(\ast)$ follows from the assumption in the lemma.

For every monotone Turing machine $T$, and every $m'\in\{0,1,2\}$ define the monotone Turing machine $S_{F_{m'}}(T)$ which simulates $T$ until the simulated machine $T$ halts:
\begin{itemize}
    \item Prior to the halting of the simulated machine $T$, the real machine $S_{F_{m'}}(T)$ does not write any bit on its output tape. We only store the value of the (simulated) output tape of the simulated $T$ on the working tape of $S_{F_{m'}}(T)$.
    \item When the simulated $T$ attempts to read a bit from its input tape, we read the bit from the actual input tape of $S_{F_{m'}}(T)$.
    \item If the simulated $T$ does not halt, then $S_{F_{m'}}(T)$ loops forever without writing any bit on its output tape (even if the simulated $T$ writes some bits on its simulated output tape).
    \item If the simulated $T$ halts, the machine $S_{F_{m'}}(T)$ computes ${F_{m'}}(x)$, where $x$ is the output of the simulated $T$, and then writes ${F_{m'}}(x)$ on the output tape.
\end{itemize}

Clearly, $S_{F_{m'}}(T)(p)={F_{m'}}(T(p))$ for every $p\in\calB^*$. Furthermore, we can computably get a description $\langle S_{F_{m'}}(T)\rangle$ of $S_{F_{m'}}(T)$ from a description $\langle T\rangle$ of $T$.

Now let $T_1',\ldots,T_n',\ldots$ be a computable enumeration of all monotone Turing machines, i.e., the mapping $n\mapsto \langle T_n'\rangle$ is computable.

We define another enumeration $T_1,\ldots,T_n,\ldots$ as follows:
\begin{align*}
    T_{n}=\begin{cases}
        T_{x_n}\quad&\text{if }1\leq n\leq \ell\,,\\
        T'_{\frac{n-\ell}{3\ell}}\quad&\text{if }n>\ell\text{ and }3\ell\text{ divides }n-\ell\,,\\
        S_{F_{\left\lfloor \frac{n-\ell}{\ell}\right\rfloor \bmod 3}}( T'_{\lceil \frac{n-\ell}{3\ell}\rceil})\quad&\text{otherwise}\,.
    \end{cases}
\end{align*}
Clearly, the mapping $n\mapsto \langle T_n\rangle$ is also computable. The following figure illustrates how the machines $(T_n)_{n\geq 1}$ are related to $(T_{x_i})_{1\leq i\leq \ell}$ and $(T_n')_{n\geq 1}$:

\begin{center}
\begin{tikzpicture}[
    % Define styles for the blocks
    block/.style={
        draw, 
        rectangle, 
        minimum width=4cm, 
        text centered, 
        font=\small,
        inner sep=0pt
    },
    % Base style for braces (will be modified for direction)
    brace/.style={
        decorate,
        decoration={brace, amplitude=5pt},
        thick
    },
    % Style for labels on the right side
    label text right/.style={
        font=\footnotesize, 
        align=left,
        xshift=0.5em,
        anchor=west
    },
    % Style for labels on the left side (rotated)
    label text left/.style={
        font=\bfseries, 
        align=right,
        xshift=-1.5em, % Shift further left away from brace
        anchor=south,  % Anchor at the "bottom" (which becomes right side after rotation)
        rotate=90
    }
]

% --- 1. The Initial Set S_0 (Bottom) ---
% Split into distinct machines to show granularity
\node[block, fill=gray!20, minimum height=0.6cm] (tx1) at (0,0) {$T_{x_1}$};
\node[above=0.05cm of tx1] (dots_init) {$\vdots$};
\node[block, fill=gray!20, minimum height=0.6cm, above=0.05cm of dots_init] (txl) {$T_{x_\ell}$};

% Brace on the LEFT side for the group S_0
\draw[thick, decorate, decoration={brace, amplitude=5pt, mirror}] 
    (txl.north west) -- (tx1.south west) 
    node[midway, label text left] {\scriptsize $\ell$ machines (for $\mathcal{S}_0$)};

% Individual braces on the RIGHT side
% T_x1 is T_1
\draw[brace] (tx1.north east) -- (tx1.south east) node[midway, label text right] {$= T_1$};
% T_xl is T_ell
\draw[brace] (txl.north east) -- (txl.south east) node[midway, label text right] {$= T_\ell$};

% --- 2. The First Block (Based on T'_1) ---
% S_F0 -- CHANGED: ell-1 machines, height 1.25cm
% Positioned DIRECTLY above txl (0pt spacing)
\node[block, fill=blue!10, minimum height=1.25cm, above=0pt of txl] (b1_f0) {$S_{F_0}(T'_1)$};
\draw[brace] (b1_f0.north east) -- (b1_f0.south east) node[midway, label text right] {$\ell-1$ machines:\\ $T_{\ell+1},\ldots,T_{2\ell-1}$};

% S_F1 -- Unchanged: ell machines, height 1.5cm
\node[block, fill=red!10, minimum height=1.5cm, above=0pt of b1_f0] (b1_f1) {$S_{F_1}(T'_1)$};
\draw[brace] (b1_f1.north east) -- (b1_f1.south east) node[midway, label text right] {$\ell$ machines:\\
$T_{2\ell},\ldots,T_{3\ell-1}$
};

% S_F2 -- CHANGED: ell machines, height 1.5cm
\node[block, fill=green!10, minimum height=1.5cm, above=0pt of b1_f1] (b1_f2) {$S_{F_2}(T'_1)$};
\draw[brace] (b1_f2.north east) -- (b1_f2.south east) node[midway, label text right] {$\ell$ machines:\\
$T_{3\ell},\ldots,T_{4\ell-1}$};

% T'_1
\node[block, fill=orange!20, minimum height=0.6cm, above=0pt of b1_f2] (b1_t) {$T'_1$};
\draw[brace] (b1_t.north east) -- (b1_t.south east) node[midway, label text right] {1 machine: $T_{4\ell}$};

% Group Label for T'_1 (Left Side)
\draw[thick, decorate, decoration={brace, amplitude=10pt, mirror}] 
    (b1_t.north west) -- (b1_f0.south west) 
    node[midway, label text left] {\scriptsize Block for $T'_1$};

% --- 3. The Second Block (Based on T'_2) ---
% S_F0 -- CHANGED: ell-1 machines, height 1.25cm
% Positioned DIRECTLY above b1_t (0pt spacing)
\node[block, fill=blue!10, minimum height=1.25cm, above=0pt of b1_t] (b2_f0) {$S_{F_0}(T'_2)$};
\draw[brace] (b2_f0.north east) -- (b2_f0.south east) node[midway, label text right] {$\ell-1$ machines:\\ $T_{4\ell+1},\ldots,T_{5\ell-1}$};

% S_F1 -- Unchanged
\node[block, fill=red!10, minimum height=1.5cm, above=0pt of b2_f0] (b2_f1) {$S_{F_1}(T'_2)$};
\draw[brace] (b2_f1.north east) -- (b2_f1.south east) node[midway, label text right] {$\ell$ machines:\\
$T_{5\ell},\ldots,T_{6\ell-1}$};

% S_F2 -- CHANGED: ell machines, height 1.5cm
\node[block, fill=green!10, minimum height=1.5cm, above=0pt of b2_f1] (b2_f2) {$S_{F_2}(T'_2)$};
\draw[brace] (b2_f2.north east) -- (b2_f2.south east) node[midway, label text right] {$\ell$ machines:\\
$T_{6\ell},\ldots,T_{7\ell-1}$};

% T'_2
\node[block, fill=orange!20, minimum height=0.6cm, above=0pt of b2_f2] (b2_t) {$T'_2$};
\draw[brace] (b2_t.north east) -- (b2_t.south east) node[midway, label text right] {1 machine: $T_{7\ell}$};

% Group Label for T'_2 (Left Side - Outward)
\draw[thick, decorate, decoration={brace, amplitude=10pt, mirror}] 
    (b2_t.north west) -- (b2_f0.south west) 
    node[midway, label text left] {\scriptsize Block for $T'_2$};

% --- Continuation ---
\node[above=0.2cm of b2_t, font=\large] {$\vdots$};

\end{tikzpicture}
\end{center}

Let $c:\mathbb{N}\to\calB^*$ be the complete prefix-free encoding of positive natural numbers where $c(n)$ is the bitstring having $n-1$ ones followed by a zero.

We can then define the machine $U$ so that $U(c(n)p)=T_n(p)$ as follows: 
\begin{itemize}
    \item The machine $U$ starts reading the input tape bit by bit until it reads the first zero.
    \item Let $n$ be the number of bits that were read. The machine then computes the representation $\langle T_n\rangle$ of the $n$-th machine $T_n$ in the enumeration and then start simulating $T_n$.
\end{itemize}

This machine $U$ is a universal monotone Turing machine by construction (as it can simulate any machine in the enumeration $(T_n)_{n\geq 1}$ which covers all monotone Turing machine).

For every $1\leq i\leq\ell$, we have $U(c(i))=T_i(\varepsilon)=T_{x_i}(\varepsilon)=x_i$, and hence $K_U(x_i)\leq l(c(i))=i$. Therefore,
$$\sum_{x\in\calS_0}2^{-K_U(x)}=\sum_{i=1}^\ell 2^{-K_U(x_i)}\geq\sum_{i=1}^\ell 2^{-i}=1-2^{-\ell}\geq 1-\epsilon\,.$$

Now fix $m\in\{1,2\}$ and let $\calS_m$ be some arbitrary set which is closed under $F_m$, i.e., $F_m(\calS_m)\subseteq\calS_m$. Let $x\in\calS_m\setminus F_m(\calB^*)$. Since $F_m(x')=x'$ for all $x'\in\calS_0$, we must have $x\notin\calS_0$ and hence for all $1\leq i\leq \ell$, we must have $x\neq x_i=T_i(\varepsilon)=U(c(i))$. Let $p_x$ be an arbitrary bit-string for which $U(p_x)=x$. From the structure of $U$, we know that there must exist $n>0$ such that $p_x=c(n)p'_x$ for some $p'_x\in\calB^*$. Hence, $x=U(p_x)=U(c(n)p'_x)=T_n(p'_{x})$. Since $x\neq x_i=T_i(\varepsilon)=U(c(i))$ for all $1\leq i\leq\ell$, we must have $n>\ell$.

We consider two cases for the index $n$:
\begin{enumerate}
    \item $3\ell$ does not divide $n-\ell$. In this case, $T_n = S_{F_{m'}}(T'_{n'})$ where
    $n'=\lceil \frac{n-\ell}{3\ell}\rceil$
    and $m' =\lfloor(n-\ell)/\ell\rfloor\bmod 3 \in \{0,1,2\}$.
    This implies $x = T_n(p'_x) = F_{m'}(T'_{n'}(p'_x)) = F_{m'}(y)$ for $y:=T'_{n'}(p'_x)$.
    
    \begin{itemize}
        \item If $m'=m$, then $x=F_m(y)$, which contradicts $x \in \calS_m \setminus F_m(\calB^*)$.
        \item If $m'=0$, then since $F_0 = F_m \circ F_{3-m}$, we have $x = F_m(F_{3-m}(y))$, which also contradicts $x \in \calS_m \setminus F_m(\calB^*)$.
        \item Thus, $m'$ must be equal to $3-m$, and hence $$F_m(x)=F_m(F_{m'}(y))=F_m(F_{3-m}(y))=F_0(y)=F_0(T'_{n'}(p'_x))=S_{F_0}(T'_{n'})(p'_x)\,.$$
        Now notice that for $\tilde{n}=(3n'-2)\ell+1$, which is greater than $\ell$ and not divisible by $\ell$, we have
        $$\left\lceil\frac{\tilde{n}-\ell}{3\ell}\right\rceil=\left\lceil\frac{(3n'-2)\ell+1-\ell}{3\ell}\right\rceil=\left\lceil\frac{3(n'-1)\ell+1}{3\ell}\right\rceil=n'\,,$$
        and\footnote{We may assume without loss of generality that $\epsilon< 1/10$ so that $\ell=\lceil \log(2/\epsilon)\rceil\geq 2$ and hence $\lfloor 1/\ell\rfloor=0$.}
        \begin{align*}
            \left\lfloor\frac{\tilde{n}-\ell}{\ell}\right\rfloor\bmod 3=\left\lfloor\frac{(3n'-3)\ell+1}{\ell}\right\rfloor\bmod 3=(3n'-3)\bmod 3=0\,,
        \end{align*}
        and hence if we let $\tilde{p}_x=c(\tilde{n})p'_x$, we get 
        \begin{align*}
            U(\tilde{p}_x)=U(c(\tilde{n})p'_x)=T_{\tilde{n}}(p'_x)=S_{F_{\left\lfloor \frac{\tilde{n}-\ell}{\ell}\right\rfloor \bmod 3}}( T'_{\lceil \frac{\tilde{n}-\ell}{3\ell}\rceil})(p'_x)=S_{F_0}(T'_{n'})(p'_x)=F_m(x)\,.
        \end{align*}
        Furthermore,
        \begin{equation}
            \label{eq:shorter-tildepx-first-case}
            \begin{aligned}
            l(\tilde{p}_x)-l(p_x)&=l(c(\tilde{n})p'_x)-l(c(n)p_x')=\tilde{n}-n=(3n'-2)\ell+1-n\\
            &\stackrel{(\ast)}\leq (3n'-2)\ell+1-(3n'-1)\ell\leq -\ell+1\,,
            \end{aligned}
        \end{equation}
        where $(\ast)$ follows from the following fact:
        \begin{itemize}
            \item As $n'=\lceil \frac{n-\ell}{3\ell}\rceil$, we have
    $n'-1 <  \frac{n-\ell}{3\ell}\leq n'$ and hence
    $$3n'-3<\frac{n-\ell}{\ell}\leq 3n'\,.$$
            Combining this with $m'=\lfloor(n-\ell)/\ell\rfloor\bmod 3=3-m>0$, we get that we must have $(n-\ell)/\ell\geq (3n'-3)+1$ and hence $n\geq (3n'-1)\ell$.
        \end{itemize}
    \end{itemize}
    
    \item $3\ell$ divides $n-\ell$. In this case, $T_n$ is equal to the ``raw'' machine $T'_{n'}$ with $n'=\frac{n-\ell}{3\ell}$.
    Notice that for $$\tilde{n} = (3n' - 2 + m)\ell\,,$$
    we have
    \begin{align*}
        \left\lceil\frac{\tilde{n}-\ell}{3\ell}\right\rceil &= \left\lceil\frac{(3n'-2+m)\ell-\ell}{3\ell}\right\rceil = \left\lceil\frac{(3n'-3+m)\ell}{3\ell}\right\rceil \\
        &= \left\lceil n' - 1 + \frac{m}{3}\right\rceil = n'\,,
    \end{align*}
    where the last equality is true because $m\in\{1,2\}$ and hence $0< \frac{m}{3}<1$. On the other hand, we have
    \begin{align*}
        \left\lfloor\frac{\tilde{n}-\ell}{\ell}\right\rfloor \bmod 3 &= \left\lfloor \frac{(3n'-3+m)\ell}{\ell} \right\rfloor \bmod 3 \\
        &= (3n' - 3 + m) \bmod 3 = m \bmod 3 = m\,.
    \end{align*}
    Thus $T_{\tilde{n}} = S_{F_m}(T'_{n'})$.
    Defining $\tilde{p}_x = c(\tilde{n})p'_x$, we get $$U(\tilde{p}_x)=U(c(\tilde{n})p'_x) = T_{\tilde{n}}(p'_x) = S_{F_m}(T'_{n'})(p'_x)= F_m(T'_{n'}(p'_x)) = F_m(x)\,.$$
    Furthermore,
    \begin{equation}
        \label{eq:shorter-tildepx-second-case}
        \begin{aligned}
        l(\tilde{p}_x) - l(p_x) &= \tilde{n} - n = (3n' - 2 + m)\ell - (\ell + 3n'\ell) \\
        &= (m-3)\ell \stackrel{(\ast)}{\leq} -\ell\leq-\ell + 1\,,
        \end{aligned}
    \end{equation}
    where $(\ast)$ follows from the fact that $m \in \{1, 2\}$ and hence $m\geq 1$.
\end{enumerate}
We conclude that for every $p_x\in\calB^*$ for which $U(p_x)=x$, there exists $\tilde{p}_x\in\calB^*$ satisfying $U(\tilde{p}_x)=F_m(x)$ and $l(\tilde{p}_x)\leq l(p_x)-\ell+1$. Therefore,
$$K_U(F_m(x))\leq K_U(x)-\ell+1\,,$$
and hence
\begin{align*}
    \sum_{x\in \calS_m\setminus F_m(\calB^*)} w_m(x) 2^{-K_U(x)}&\leq \sum_{x\in \calS_m\setminus F_m(\calB^*)} w_m(x) 2^{-K_U(F_m(x))-\ell+1}\\
    &\stackrel{(\ast)}\leq \epsilon\sum_{x\in \calS_m\setminus F_m(\calB^*)} w_m(x) 2^{-K_U(F_m(x))}\\
    &\stackrel{(\dagger)}\leq \epsilon \sum_{x\in \calS_m\setminus F_m(\calB^*)} w_m(F_m(x)) 2^{-K_U(F_m(x))}\\
    &\stackrel{(\ddagger)}= \epsilon \sum_{y\in F_m(\calS_m\setminus F_m(\calB^*))} w_m(y) 2^{-K_U(y)}\\
    &\stackrel{(\diamond)}\leq \epsilon \sum_{y\in \calS_m}w_m(y)2^{-K_U(y)}\,,
\end{align*}
where $(\ast)$ follows from our choice of $\ell$, $(\dagger)$ from the fact that $w_m$ is non-decreasing under $F_m$, $(\ddagger)$ follows because $F_m$ is injective, and $(\diamond)$ follows because $\calS_m$ is closed under $F_m$ and hence $F_m(\calS_m\setminus F_m(\calB^*))\subseteq F_m(\calS_m)\subseteq\calS_m$.

We conclude that
\begin{align*}
    \sum_{x\in \calS_m\cap F_m(\calB^*)}w_m(x)2^{-K_U(x)}&=\sum_{x\in \calS_m}w_m(x)2^{-K_U(x)}-\sum_{x\in \calS_m\setminus F_m(\calB^*)}w_m(x)2^{-K_U(x)} \\
    &\geq (1-\epsilon)\sum_{x\in \calS_m}w_m(x)2^{-K_U(x)}.
\end{align*}
\end{proof}

The following lemma is a version of \cref{lem:decoupled-solomonoff-mixture-eps-approximates-MgeRk-andMvpup} for the Solomonoff mixture $\Mmu_{U,\tau}$ in the (coupled) embedded case. We write $\Mvp_{\Mmu_{U,\tau}}:\TurnSet^*\to\Delta'\calA$ and $\Mve_{\Mmu_{U,\tau}}:\TurnSet^*\times\calA\to\Delta'\calE$ to denote the ``policy-part" and the ``environment-part" of the mixture universe $\Mmu_{U,\tau}$, respectively, i.e.,
$$\Mvp_{\Mmu_{U,\tau}}(a|\HistA)=\Mmu_{U,\tau}(a|\HistA)\quad\text{and}\quad\Mve_{\Mmu_{U,\tau}}(e|\HistA,a)=\Mmu_{U,\tau}(e|\HistA,a)\,,\quad\forall(\HistA,a,e)\in\TurnSet^*\times\calA\times\calE\,.$$
Clearly, $\Mmu_{U,\tau}=\Mve_{\Mmu_{U,\tau}}^{\Mvp_{\Mmu_{U,\tau}}}$.

\begin{lemma}
    \label{lem:coupled-solomonoff-mixture-eps-approximates-MgeRk-andMvpup}
    For every $\epsilon>0$, there exists a monotone Turing machine $U$ such that the environment part $\Mve_{\Mmu_{U,\tau}}$ of $\Mmu_{U,\tau}$ $\epsilon$-approximates $\Mge_{R,k}$ on $\TurnSet^*_{\Mge_{R,k}}$, and the policy part $\Mvp_{\Mmu_{U,\tau}}$ of $\Mmu_{U,\tau}$ $\epsilon$-approximates $\Mvp_{\mathrm{up}}$ on $\TurnSet^{*\text{-}\mathrm{up}\text{-}k\text{-}\mathrm{down}\text{-}*\text{-}\mathrm{up}}_{\Mge_{R,k}} \cup \TurnSet^{*\text{-}\mathrm{up}}_{\Mge_{R,k}}$.
\end{lemma}
\begin{proof}
    Let $\ell=\lceil\log(2/\epsilon)\rceil$, and let $M_1,\ldots,M_\ell$ be $\ell$ APOMs which are syntactically different, but which all implement the same universe $\Mge_{R,k}^{\Mvp_{\mathrm{up}}}$, i.e., $\Mvu_{M_i}^\tau=\Mge_{R,k}^{\Mvp_{\mathrm{up}}}$ for all $1\leq i\leq \ell$. For example, we can pick any machine $M_1$ for which $\Mvu_{M_1}^\tau=\Mge_{R,k}^{\Mvp_{\mathrm{up}}}$ and then for every $1< i\leq \ell$, the machine $M_i$ starts with executing a loop of $i$ iterations in which it does some useless computation and afterwards it starts simulating $M_1$. Let $$\mathcal{S}_0=\left\{\langle M_i\rangle:1\leq i\leq \ell\right\}\,.$$

    Now for every APOM $M\in\calMapom$, we define an APOM $S_{\Mge_{R,k}}(M)$ such that the ``environment part" of the universe $\Mvu_{S_{\Mge_{R,k}}(M)}^\tau$ is equivalent to (the deterministic environment) $\Mge_{R,k}$, but at the same time it has the same distribution as $\Mvu_{M}^\tau$ on histories $\HistA a\in\TurnSet^*\times\calA$ that end with an action and which are compatible with $\Mge_{R,k}$. More precisely, we would like to have the following guarantees:
    \begin{align*}
        \Mvu_{S_{\Mge_{R,k}}(M)}^\tau(e_t|\HistM,a_t)=\Mge_{R,k}(e_t|\HistM,a_t)\,,\quad&\text{if }\Mvu_{S_{\Mge_{R,k}}(M)}^\tau(\HistM a_t)>0\,,
    \end{align*}
    and
    \begin{equation}
        \Mvu_{S_{\Mge_{R,k}}(M)}^\tau(\HistM a_t)=\begin{cases}
            \Mvu_M^\tau(\HistM a_t)\,\quad&\text{if }\Mge_{R,k}(e_{t'}|\HistM[t'],a_{t'})=1\text{ for all }1\leq t'<t\,,\\
            0\quad&\text{otherwise.}
        \end{cases}
        \label{eq:MvuSMgeRkM-compatible-with-MvuM}
    \end{equation}

    Since $\Mge_{R,k}$ is deterministic, we can ensure the above by designing the ``policy part" of $S_{\Mge_{R,k}}$ so that it satisfies
    
    \begin{equation}
        \label{eq:conditional-SMgeRkM}
        \Mvu_{S_{\Mge_{R,k}}(M)}^\tau(a_t|\HistM)=\begin{cases}
            \Mvu_{M}^\tau(a_t|\HistM)\quad&\text{if }t=1\,,\\
            \Mvu_{M}^\tau(a_t|\HistM)\Mvu_{M}^\tau(e_{t-1}|\HistM[t-1]a_{t-1})\quad&\text{if }t>1\text{ and }\Mge_{R,k}(e_{t-1}|\HistM[t-1],a_{t-1})=1\,,\\
           0\quad&\text{otherwise.}\\
        \end{cases}
    \end{equation}

    In order to properly implement this using an APOM, we need to be able to run a simulation of $M$ and we would need the ability to restart the simulation of $M$ from a previous state. Let $\mathcal{S}_M$ be state space describing the internal state of the APOM $M$, i.e., an element $s$ of $\mathcal{S}_M$ has a complete description of the current content of the working tapes, the output tape, and the oracle tape.\footnote{Recall that when we run $M$ to produce a universe, the input tape is ignored, so we do not need a representation of the input tape.} One computational step of $M$ can be simulated as $s\leftarrow \mathrm{Simulate}(\langle M\rangle,s)$ where $\mathrm{Simulate}$ is a function that applies the simulation. With these conventions, we can implement $S_{\Mge_{R,k}}(M)$ as follows:

    \begin{algorithm}[H]
    % \SetKwProg{Fn}{Function}{:}{}

        \BlankLine
    
        $t\leftarrow1$\;
        $\HistA\leftarrow\varepsilon$\;
        \tcp{We will keep $t=l(\HistA)+1$ so that we generally have $\HistA=\HistM$.}
    
        \BlankLine
    
        \tcp{Initialize the internal state $s$ to be used in a simulation of $M$}
        $s\leftarrow \mathrm{InitializeState}()$\;
    
        \BlankLine
    
        \While{True}{
            \tcp{Continue the simulation of $M$ until it produces an action. This part contributes the term $\Mvu_{M}^\tau(a_t|\HistM)$ in the expression of $\Mvu_{S_{\Mge_{R,k}}(M)}^\tau(a_t|\HistM)$ (which would only be a factor when $t>1$) (cf. \eqref{eq:conditional-SMgeRkM}).}
            \While{True}{
                $s\leftarrow \mathrm{Simulate}(\langle M\rangle, s)$\;
                \BlankLine
                \If{$\mathrm{IsHalt}(s)$}{
                    Halt the real machine $S_{\Mge_{R,k}}(M)$\;
                }
                \BlankLine
                %\tcp{Check if the simulated output tape ends with a valid representation of an action $a\in\calA$}
                \If{$\mathrm{OutputTape}(s)=\langle \HistA a\rangle$ for some $a\in\calA$}{
                    Extract $a$ from $\mathrm{OutputTape}(s)$\;
                    $a_t\leftarrow a$\;
                    \textbf{Break}\;
                }
            }
            Append $\langle a_t\rangle$ to the real output tape of $S_{\Mge_{R,k}}(M)$\;
            
            \BlankLine
    
            \tcp{Compute a percept $e_t$ according to $\Mge_{R,k}$. This ensures that $\Mvu_{S_{\Mge_{R,k}}(M)}^\tau(e_t|\HistM,a_t)=\Mge_{R,k}(e_t|\HistM,a_t)$.}
            $e_t\leftarrow\langle o_{0},r_t(a_{1:t})\rangle$\;
            Append $\langle e_t\rangle$ to the real output tape of $S_{\Mge_{R,k}}(M)$\;
    
            \BlankLine
    
            \tcp{Evolve the simulation of $M$ until it produces a percept $e\in\calE$. If the simulation halts before producing a percept or if it produces a percept $e\neq e_t$, we halt the real machine. This part contributes a multiplicative factor $\Mvu_{M}^\tau(e_{t}|\HistM[t]a_{t})$ in the expression of $\Mvu_{S_{\Mge_{R,k}}(M)}^\tau(a_{t+1}|\Hist)$ (cf. \eqref{eq:conditional-SMgeRkM}).}
            \While{True}{
                $s\leftarrow \mathrm{Simulate}(\langle M\rangle, s)$\;
                \BlankLine
                \If{$\mathrm{IsHalt}(s)$}{
                    Halt the real machine $S_{\Mge_{R,k}}(M)$\;
                }
                \BlankLine
                %\tcp{Check if the simulated output tape ends with a valid representation of an action $e\in\calA$}
                \If{$\mathrm{OutputTape}(s)=\langle \HistM a_t e\rangle$ for some $e\in\calE$}{
                    \textbf{Break}\;
                }
            }
            \If{$e=e_t$}{
                \textbf{Break};
            }
            \Else{
                Halt the real machine $S_{\Mge_{R,k}}(M)$\;
            }
    
            \BlankLine
            
            $\HistA\leftarrow\HistA a_t e_t$\;
            $t\leftarrow t+1$\;
            
        }
        \caption{The "reverting" APOM $S_{\Mge_{R,k}}(M)$}
        \label{alg:reverting-apom-SMgeRkM}
    \end{algorithm}

We assume that the ``boiler-plate" structure of $S_{\Mge_{R,k}}(M)$ is done in a way that ensures that $S_{\Mge_{R,k}}(M)\neq M_i$ for all $1\leq i\leq \ell$, e.g., we can just add some useless loop that does not exist in any of the machines $(M_i)_{1\leq i\leq \ell}$.

Now we also design a machine $S_{*\text{-}\mathrm{up}\text{-}\mathrm{after}\text{-}k\text{-}\mathrm{down}}(M)$ which would follow the simulation of $M$ until it produces $k$ consecutive `down' actions, after which it continues the simulation according to $\Mge_{R,k}^{\Mvp_{\mathrm{up}}}$.

    \begin{algorithm}[H]
    % \SetKwProg{Fn}{Function}{:}{}

        \BlankLine
    
        $t\leftarrow1$\;
        $\HistA\leftarrow\varepsilon$\;
        \tcp{We will keep $t=l(\HistA)+1$ so that we generally have $\HistA=\HistM$.}

        \BlankLine
        
        $\mathrm{streak}\text{\_}\mathrm{found}\leftarrow\mathrm{false}$\;
    
        \BlankLine
    
        \tcp{Initialize the internal state $s$ to be used in a simulation of $M$}
        $s\leftarrow \mathrm{InitializeState}()$\;
    
        \BlankLine
    
        \While{True}{
            \If{$\mathrm{streak}\text{\_}\mathrm{found}$}{
                \tcp{If $k$ down actions have been found, produce an `up' action and a percept that is consistent with $\Mge_{R,k}$.}
                $a_t\leftarrow\mathrm{up}$\;
                $e_t\leftarrow\langle o_{0},r_t(a_{1:t})\rangle$\;
                $\Turn_t\leftarrow a_t e_t$\;
            }
            \Else{
                \tcp{Continue the simulation of $M$ until it produces an action-percept pair $\Turn\in\TurnSet$.}
                \While{True}{
                    $s\leftarrow \mathrm{Simulate}(\langle M\rangle, s)$\;
                    \BlankLine
                    \If{$\mathrm{IsHalt}(s)$}{
                        Halt the real machine $S_{*\text{-}\mathrm{up}\text{-}\mathrm{after}\text{-}k\text{-}\mathrm{down}}(M)$\;
                    }
                    \BlankLine
                    \If{$\mathrm{OutputTape}(s)=\langle \HistA \Turn\rangle$ for some $\Turn\in\TurnSet$}{
                        Extract $\Turn$ from $\mathrm{OutputTape}(s)$\;
                        $\Turn_t\leftarrow \Turn$\;
                        Break\;
                    }
                }
            }
            Append $\langle \Turn_t\rangle$ to the real output tape of $S_{*\text{-}\mathrm{up}\text{-}\mathrm{after}\text{-}k\text{-}\mathrm{down}}(M)$\;
            
            \BlankLine
    
            \If{$t\geq k$ and $a_{t-k+1:t}=(\mathrm{down})^k$}{
                $\mathrm{streak}\text{\_}\mathrm{found}\leftarrow \mathrm{true}$
            }

            $\HistA\leftarrow\HistA \Turn_t$\;
            $t\leftarrow t+1$\;
        }
        \caption{The "reverting" APOM $S_{*\text{-}\mathrm{up}\text{-}\mathrm{after}\text{-}k\text{-}\mathrm{down}}(M)$}
        \label{alg:reverting-apom-S-upafterkdwon}
    \end{algorithm}

We further assume that the ``boiler-plate" structure of $S_{*\text{-}\mathrm{up}\text{-}\mathrm{after}\text{-}k\text{-}\mathrm{down}}(M)$ ensures it is distinct from the machines in $\calS_0$ and from the outputs of $S_{\Mge_{R,k}}$. Note that by construction, $S_{\Mge_{R,k}}$ enforces the environment conditions of $\Mge_{R,k}$, while $S_{*\text{-}\mathrm{up}\text{-}\mathrm{after}\text{-}k\text{-}\mathrm{down}}$ enforces the policy conditions of $\Mvp_{\mathrm{up}}$ (as well as the environment conditions of $\Mge_{R,k}$) after a streak of $k$ downs.

Now how does $S_{*\text{-}\mathrm{up}\text{-}\mathrm{after}\text{-}k\text{-}\mathrm{down}}(S_{\Mge_{R,k}}(M))$ compare to $S_{\Mge_{R,k}}(S_{*\text{-}\mathrm{up}\text{-}\mathrm{after}\text{-}k\text{-}\mathrm{down}}(M))$? Syntactically they are different, but they induce the same distribution, i.e., $\Mvu_{S_{*\text{-}\mathrm{up}\text{-}\mathrm{after}\text{-}k\text{-}\mathrm{down}}(S_{\Mge_{R,k}}(M))}^\tau=\Mvu_{S_{\Mge_{R,k}}(S_{*\text{-}\mathrm{up}\text{-}\mathrm{after}\text{-}k\text{-}\mathrm{down}}(M))}^\tau$.

Let $$\mathcal{S} := \{\langle M \rangle : M \in \calMapom \setminus \calS_0\}=\left\{\langle M\rangle:M\in\calM_\calS\right\}\,,$$
    where
    $$\calM_\calS:=\calMapom\setminus\{M_1,\ldots,M_\ell\}\,,$$
We would like to define two functions $F_1$ and $F_2$ based on $S_{\Mge_{R,k}}$ and $S_{*\text{-}\mathrm{up}\text{-}\mathrm{after}\text{-}k\text{-}\mathrm{down}}$ (similarly to the definition of $F$ in \eqref{eq:definition-F-in-terms-of-Sk} in the proof of \cref{lem:decoupled-solomonoff-mixture-eps-approximates-MgeRk-andMvpup}) so that we can apply \cref{lem:choose-UTM-to-make-Fx-and-Fprimex-likely}. However, due to the constraint that we need $F_1$ and $F_2$ to commute, we cannot use $S_{\Mge_{R,k}}$ and $S_{*\text{-}\mathrm{up}\text{-}\mathrm{after}\text{-}k\text{-}\mathrm{down}}$ directly because they do not commute. Because of this, we will define $S_1:\calM_\calS\to\calM_\calS$ to exactly match $S_{\Mge_{R,k}}$, i.e.,
$$S_1(M) = S_{\Mge_{R,k}}(M)\,,\quad\forall M\in\calM_\calS\,,$$
but we will define $S_2:\calM_\calS\to\calM_\calS$ in a way that does not exactly match $S_{*\text{-}\mathrm{up}\text{-}\mathrm{after}\text{-}k\text{-}\mathrm{down}}$, but nevertheless is ``distributionally equivalent to it''.

For this purpose, let us define for every $M\in\calMapom$ the number $n_{S_{1}}(M)$ as the maximal $n\geq 0$ for which $S_{1}^n(M')=M$ for some $M'\in\calMapom$, where $S_{1}^n(M')$ denotes the machine obtained by applying the wrapper $S_{\Mge_{R,k}}$ $n$ times on $M'$, and we adopt the convention that $S_{1}^0(M)=M$.\footnote{Hence, if $n_{S_{1}}(M)=0$ then $M\neq S_{\Mge_{R,k}}(M')$ for all $M'\in \calMapom$.} Such a maximal $n$ is guaranteed to exist because the wrapper always adds extra overhead (of at least one bit) in the representation of the machine, and hence $n_{S_{1}}(M)\leq l(\langle M\rangle)$. Now for every $M\in\calMapom$, let $S_{1}^{-n_{S_{1}}(M)}(M)$ be the machine $M'$ for which $M=S_{1}^{n_{S_{1}}(M)}(M')$. We can now define $S_2:\calM_\calS\to\calM_\calS$ as follows

\begin{align*}
    S_2(M) = S_{1}^{n_{S_{1}}(M)}\left(S_{*\text{-}\mathrm{up}\text{-}\mathrm{after}\text{-}k\text{-}\mathrm{down}}\left( S_{1}^{-n_{S_{1}}(M)} (M)\right)\right)\,.
\end{align*}

A few comments are in order:
\begin{itemize}
    \item $n_{S_1}(S_1(M))=n_{S_1}(M)+1$ and $S_1^{-n_{S_1}(S_1(M))}(S_1(M))=S_1^{-n_{S_1}(M)}(M)$. Therefore,
    \begin{align*}
        S_2(S_1(M))&=S_{1}^{n_{S_{1}}(S_1(M))}\left(S_{*\text{-}\mathrm{up}\text{-}\mathrm{after}\text{-}k\text{-}\mathrm{down}}\left( S_{1}^{-n_{S_{1}}(S_1(M))} (S_1(M))\right)\right)\\
        &=S_{1}^{n_{S_{1}}(M)+1}\left(S_{*\text{-}\mathrm{up}\text{-}\mathrm{after}\text{-}k\text{-}\mathrm{down}}\left( S_1^{-n_{S_1}(M)}(M)\right)\right)\\
        &=S_1\left(S_{1}^{n_{S_{1}}(M)}\left(S_{*\text{-}\mathrm{up}\text{-}\mathrm{after}\text{-}k\text{-}\mathrm{down}}\left( S_1^{-n_{S_1}(M)}(M)\right)\right)\right)\\
        &= S_1(S_2(M))\,,
    \end{align*}
    which means that $S_1$ and $S_2$ commute.
    \item Now since $S_{*\text{-}\mathrm{up}\text{-}\mathrm{after}\text{-}k\text{-}\mathrm{down}}$  and $S_{\Mge_{R,k}}$ ``distributionally commute'' in the sense that $$\Mvu_{S_{*\text{-}\mathrm{up}\text{-}\mathrm{after}\text{-}k\text{-}\mathrm{down}}(S_{\Mge_{R,k}}(M))}^\tau=\Mvu_{S_{\Mge_{R,k}}(S_{*\text{-}\mathrm{up}\text{-}\mathrm{after}\text{-}k\text{-}\mathrm{down}}(M))}^\tau\,,\quad\forall M\in\calM_\calS\,,$$ we can see that
    \begin{align*}
        \Mvu_{S_2(M)}^\tau &= \Mvu_{S_{1}^{n_{S_{1}}(M)}\left(S_{*\text{-}\mathrm{up}\text{-}\mathrm{after}\text{-}k\text{-}\mathrm{down}}\left( S_{1}^{-n_{S_{1}}(M)} (M)\right)\right)}^\tau\\
        &=\Mvu_{S_{*\text{-}\mathrm{up}\text{-}\mathrm{after}\text{-}k\text{-}\mathrm{down}}\left(S_{1}^{n_{S_{1}}(M)}\left( S_{1}^{-n_{S_{1}}(M)} (M)\right)\right)}^\tau\\
        &=\Mvu_{S_{*\text{-}\mathrm{up}\text{-}\mathrm{after}\text{-}k\text{-}\mathrm{down}}(M)}^\tau\,,
    \end{align*}
    and hence $S_2(M)$ and $S_{*\text{-}\mathrm{up}\text{-}\mathrm{after}\text{-}k\text{-}\mathrm{down}}(M)$ are ``distributionally equivalent''. This will be sufficient for our purposes.
\end{itemize}

Now define two functions $F_1, F_2:\calB^*\to\calB^*$ as follows
\begin{align*}
    F_1(x)&=\begin{cases}
    \langle S_1(M)\rangle \quad&\text{if } x=\langle M\rangle\text{ for some }M\in\calM_\calS\,,\\
    x\quad&\text{otherwise},
\end{cases}\\
    F_2(x) &= \begin{cases}
        \langle S_2(M)\rangle \quad&\text{if } x=\langle M\rangle\text{ for some }M\in\calM_\calS\,,\\
        x\quad&\text{otherwise.}
    \end{cases}
\end{align*}
Both $F_1$ and $F_2$ are computable and injective, and map $\calS_0$ to itself (as the identity). Since $S_2$ and $S_1$ commute, $F_1$ and $F_2$ commute as well, i.e., $F_1(F_2(x))=F_2(F_1(x))$ for all $x\in\calB^*$.

Let $U$ be the universal monotone Turing machine from \cref{lem:choose-UTM-to-make-Fx-and-Fprimex-likely} applied to $\calS_0$, $F_1$, $F_2$, and $\epsilon$. This ensures two key properties:
\begin{enumerate}
    \item The set $\calS_0$ has high probability:
    \begin{equation}
        \label{eq:S0-bound-lower-bound}
        \sum_{M \in \calS_0} 2^{-K_U(\langle M \rangle)} \geq 1 - \epsilon.
    \end{equation}
    \item For $$\mathcal{S} := \{\langle M \rangle : M \in \calMapom \setminus \calS_0\}=\left\{\langle M\rangle:M\in\calM_\calS\right\}\,,$$
    with
    $$\calM_\calS:=\calMapom\setminus\{M_1,\ldots,M_\ell\}\,,$$ we get that $\calS$ is closed under $F_1$ and $F_2$. By following a similar reasoning to how we proved \eqref{eq:weighted-lower-bound-on-machines}, we can show that for any $m\in\{1,2\}$ and any non-decreasing weight function $w:
    \calM_\calS\to[0,1]$ under $S_m$, i.e., it satisfies $w(S_m(M))\geq w(M)$ for every $M\in\calM_\calS$, we have
    \begin{equation}
        \label{eq:Sm-lower-bound}
        \sum_{M \in \calM_\calS} w(S_m(M)) 2^{-K_U(\langle S_m(M)\rangle)} \geq \left(1 - \epsilon\right) \sum_{M \in \calM_\calS} w(M) 2^{-K_U(\langle M \rangle)}\,, \quad \text{for } m \in \{1, 2\}\,.
    \end{equation}
\end{enumerate}

We now prove the two approximation claims.

We start by showing the environment approximation, i.e., that $|\Mve_{\Mmu_{U,\tau}}(e|\HistM a) - \Mge_{R,k}(e|\HistM a)| \leq \epsilon$ for any $\HistM a \in \TurnSet^*_{\Mge_{R,k}} \times \calA$.
The environment part of the mixture univere $\Mmu_{U,\tau}$ is given by:
\[
\Mve_{\Mmu_{U,\tau}}(e|\HistM a) = \frac{\Mmu_{U,\tau}(\HistM a e)}{\Mmu_{U,\tau}(\HistM a)} = \frac{\sum_{M \in \calMapom} 2^{-K_U(\langle M \rangle)} \Mvu_M^\tau(\HistM a e)}{\sum_{M \in \calMapom} 2^{-K_U(\langle M \rangle)} \Mvu_M^\tau(\HistM a)}\,.
\]
Since $\HistM \in \TurnSet^*_{\Mge_{R,k}}$, any machine $M \in \calS_0$ perfectly matches $\Mge_{R,k}$, meaning $$\Mvu_M^\tau(\HistM a e) = \Mge_{R,k}(e|\HistM a)\Mvu_M^\tau(\HistM a)\,.$$

For machines $M \in \calM_\calS$, consider the mapping $S_1$. By construction of Algorithm \ref{alg:reverting-apom-SMgeRkM}, for any $M\in\calMapom$, the machine $S_1(M)=S_{\Mge_{R,k}}(M)$ enforces the environment $\Mge_{R,k}$ as long as $\Mvu_{S_{\Mge_{R,k}}(M)}^\tau(\HistM,a_t)>0$. Thus, for any $M \in \calM_\calS$, we have

$$\Mvu_{S_{1}(M)}^\tau(\HistM a e) =\Mvu_{S_{\Mge_{R,k}}(M)}^\tau(\HistM a e) = \Mge_{R,k}(e|\HistM a)\Mvu_{S_{\Mge_{R,k}}(M)}^\tau(\HistM a)=\Mge_{R,k}(e|\HistM a)\Mvu_{S_{1}(M)}^\tau(\HistM a)\,.$$

Let $e^* = (o_0, r_t(a_{<t}a))$ so that $\Mge_{R,k}(e^*|\HistM a) = 1$. We have:
\begin{equation}
    \label{eq:Mmu-tau-expression-for-S1}
\begin{aligned}
    \Mmu_{U,\tau}(\HistM a e^*) &= \sum_{M \in \calS_0} 2^{-K_U(\langle M \rangle)} \Mvu_M^\tau(\HistM ae^*) + \sum_{M \in \calM_\calS} 2^{-K_U(\langle M \rangle)} \Mvu_M^\tau(\HistM a e^*)\\
    &\geq \sum_{M \in \calS_0} 2^{-K_U(\langle M \rangle)} \Mvu_M^\tau(\HistM ae^*) + \sum_{M \in \calM_\calS} 2^{-K_U(\langle S_{1}(M) \rangle)} \Mvu_{S_{1}(M)}^\tau(\HistM a e^*)\\
    &= \sum_{M \in \calS_0} 2^{-K_U(\langle M \rangle)} \Mvu_M^\tau(\HistM a) + \sum_{M \in \calM_\calS} 2^{-K_U(\langle S_{1}(M) \rangle)} \Mvu_{S_{1}(M)}^\tau(\HistM a)\,,
\end{aligned}
\end{equation}

By defining $w(M)=\Mvu_{M}^\tau(\HistM a)$, it follows from \eqref{eq:MvuSMgeRkM-compatible-with-MvuM} and the fact that $\HistM\in\TurnSet_{\Mge_{R,k}}^*$ that $$w(S_1(M))=\Mvu_{S_{1}(M)}^\tau(\HistM a)=\Mvu_{S_{\Mge_{R,k}}(M)}^\tau(\HistM a)=\Mvu_{M}^\tau(\HistM a)=w(M)\,,$$
and hence $w$ is non-decreasing under $S_1$. It now follows from \eqref{eq:Sm-lower-bound} that
\begin{align*}
    \sum_{M \in \calM_\calS} 2^{-K_U(\langle S_{1}(M) \rangle)} \Mvu_{S_{1}(M)}^\tau(\HistM a)&=\sum_{M \in \calM_\calS} 2^{-K_U(\langle S_{1}(M) \rangle)} w(S_1(M))\\
    &\geq(1-\epsilon) \sum_{M \in \calM_\calS} 2^{-K_U(\langle M\rangle)} w(M)\\
    &=(1-\epsilon) \sum_{M \in \calM_\calS} 2^{-K_U(\langle M\rangle)} \Mvu_{M}^\tau(\HistM a)\,.
\end{align*}

Combining this with \eqref{eq:Mmu-tau-expression-for-S1}, we get
\begin{align*}
    \Mmu_{U,\tau}(\HistM a e^*) &\geq \sum_{M \in \calS_0} 2^{-K_U(\langle M \rangle)} \Mvu_M^\tau(\HistM a) + (1-\epsilon) \sum_{M \in \calM_\calS} 2^{-K_U(\langle M \rangle)} \Mvu_{M}^\tau(\HistM a)\\
    &\geq (1-\epsilon) \Mmu_{U,\tau}(\HistM a)\,.
\end{align*}
Thus
$$\Mve_{\Mmu_{U,\tau}}(e^*|\HistM a)=\frac{ \Mmu_{U,\tau}(\HistM a e^*) }{ \Mmu_{U,\tau}(\HistM a) } \geq 1 - \epsilon=\Mge_{R,k}(e^*|\HistM a)-\epsilon\,.$$
For $e\in\calE\setminus \{e^*\}$, we have $\Mge_{R,k}(e|\HistM a)=0$ and $\Mve_{\Mmu_{U,\tau}}(e|\HistM a)\leq 1-\Mve_{\Mmu_{U,\tau}}(e^*|\HistM a)\leq \epsilon$, hence
$$\Mve_{\Mmu_{U,\tau}}(e|\HistM a)\leq\epsilon+\Mge_{R,k}(e|\HistM a)\,.$$
Therefore, $\Mve_{\Mmu_{U,\tau}}$ $\epsilon$-approximates $\Mge_{R,k}$ on $\TurnSet^*_{\Mge_{R,k}}$.

Now we turn to proving the policy approximation, i.e., that $|\Mvp_{\Mmu_{U,\tau}}(a|\HistM) - \Mvp_{\mathrm{up}}(a|\HistM)| \leq \epsilon$ for every $\HistM \in\TurnSet^{*\text{-}\mathrm{up}\text{-}k\text{-}\mathrm{down}\text{-}*\text{-}\mathrm{up}}_{\Mge_{R,k}} \cup \TurnSet^{*\text{-}\mathrm{up}}_{\Mge_{R,k}}$ and every $a\in\calA$. We split this into two cases based on the history $\HistM$.

\textbf{Case 1:} $\HistM \in \TurnSet^{*\text{-}\mathrm{up}}_{\Mge_{R,k}}$ (Histories containing only `up' actions, and also compatible with $\TurnSet^{*\text{-}\mathrm{up}}_{\Mge_{R,k}}$).
In this case, $\Mvp_{\mathrm{up}}(\text{up}|\HistM) = 1$.
Since $\calS_0$ contains machines implementing $\Mge_{R,k}^{\Mvp_{\text{up}}}$, and since $\HistM \in \TurnSet^{*\text{-}\mathrm{up}}_{\Mge_{R,k}}$, we get that $\Mvu_M^\tau(\HistM \text{up}) = \Mvu_M^\tau(\HistM)=1$. Therefore,
\begin{align*}
    \Mmu_{U,\tau}(\HistM \text{up}) &\geq \sum_{M \in \calS_0} 2^{-K_U(\langle M \rangle)} \Mvu_M^\tau(\HistM \text{up}) =\sum_{M \in \calS_0} 2^{-K_U(\langle M \rangle)} \geq 1-\epsilon\,,
\end{align*}
where the last inequality follows from \eqref{eq:S0-bound-lower-bound}. Therefore,
\[
\Mvp_{\Mmu_{U,\tau}}(\text{up} \mid \HistM )=\Mmu_{U,\tau}(\text{up}|\HistM )= \frac{\Mmu_{U,\tau}(\HistM \text{up})}{\Mmu_{U,\tau}(\HistM)} \geq \Mmu_{U,\tau}(\HistM \text{up})\geq 1-\epsilon\,.
\]
\textbf{Case 2:} $\HistM \in \TurnSet^{*\text{-}\mathrm{up}\text{-}k\text{-}\mathrm{down}\text{-}*\text{-}\mathrm{up}}_{\Mge_{R,k}}$ (Histories containing a streak of $k$ downs, and also compatible with $\TurnSet^{*}_{\Mge_{R,k}}$).
Since the history contains $(\text{down})^k$, and since machines in $\calS_0$ implement $\Mvp_{\text{up}}$, we have $\Mvu_{M}^\tau(\HistM) = 0$ for $M \in \calS_0$. The probability mass is entirely supported by $\calM_\calS$.
Consider the transformation $S_2(M)$ and recall that the distributions of $\Mvu_{S_{*\text{-}\mathrm{up}\text{-}\mathrm{after}\text{-}k\text{-}\mathrm{down}}(M)}^\tau$ and $\Mvu_{S_2(M)}^\tau$ are equal. By definition (Algorithm \ref{alg:reverting-apom-S-upafterkdwon}), if the history contains $k$ downs, $S_{*\text{-}\mathrm{up}\text{-}\mathrm{after}\text{-}k\text{-}\mathrm{down}}(M)$ forces the action `up'.
Thus, for any $M'$, $\Mvu_{S_{*\text{-}\mathrm{up}\text{-}\mathrm{after}\text{-}k\text{-}\mathrm{down}}(M')}^\tau(\text{up}|\HistM) = 1$.

Now define $w(M) = \Mvu_M^\tau(\HistM)$ and let $t'<t$ be such that $a_{t'-k+1:t'}=(\mathrm{down})^k$. We have
\begin{align*}
    w(S_2(M))&=\Mvu_{S_2(M)}^\tau(\HistM)=\Mvu_{S_{*\text{-}\mathrm{up}\text{-}\mathrm{after}\text{-}k\text{-}\mathrm{down}}(M)}^\tau(\HistM)\\
    &=\Mvu_{S_{*\text{-}\mathrm{up}\text{-}\mathrm{after}\text{-}k\text{-}\mathrm{down}}(M)}^\tau(\Hist[t'])\Mvu_{S_{*\text{-}\mathrm{up}\text{-}\mathrm{after}\text{-}k\text{-}\mathrm{down}}(M)}^\tau(\Turn_{t'+1:t-1}\mid \Hist[t'])\\
    &\stackrel{(\ast)}=\Mvu_{M}^\tau(\Hist[t'])\geq \Mvu_{M}^\tau(\HistM)\,,
\end{align*}
where $(\ast)$ follows from the fact that the wrapper $S_{*\text{-}\mathrm{up}\text{-}\mathrm{after}\text{-}k\text{-}\mathrm{down}}$ mimics $M$ until the streak is complete after which it follows $\Mge_{R,k}$ and $\Mvp_{\mathrm{up}}$
(cf. Algorithm \ref{alg:reverting-apom-S-upafterkdwon}) which means that $\Mvu_{S_{*\text{-}\mathrm{up}\text{-}\mathrm{after}\text{-}k\text{-}\mathrm{down}}(M)}^\tau(\Turn_{t'+1:t-1}\mid \Hist[t'])=1$ because $\HistM \in \TurnSet^{*\text{-}\mathrm{up}\text{-}k\text{-}\mathrm{down}\text{-}*\text{-}\mathrm{up}}_{\Mge_{R,k}}$. We conclude that $w(M)= \Mvu_{M}^\tau(\HistM)$ is non-decreasing under $S_2$, and hence it follows from \eqref{eq:Sm-lower-bound} that
$$\sum_{M \in \calM_\calS} 2^{-K_U(\langle S_2(M) \rangle)} \Mvu_{S_2(M)}^\tau(\HistM)\\
    \geq (1-\epsilon) \sum_{M \in \calM_\calS} 2^{-K_U(\langle M \rangle)} \Mvu_{M}^\tau(\HistM)\,.$$

Therefore,

\begin{align*}
    \Mmu_{U,\tau}(\HistM \text{up}) &= \sum_{M \in \calM_\calS} 2^{-K_U(\langle M \rangle)} \Mvu_M^\tau(\HistM \text{up}) \\
    &\geq \sum_{M \in \calM_\calS} 2^{-K_U(\langle S_2(M) \rangle)} \Mvu_{S_2(M)}^\tau(\HistM \text{up}) \\
    &= \sum_{M \in \calM_\calS} 2^{-K_U(\langle S_2(M) \rangle)} \Mvu_{S_2(M)}^\tau(\HistM) \cdot 1\quad \quad \text{(since } F_2 \text{ forces up)}\\
    &\geq (1-\epsilon) \sum_{M \in \calM_\calS} 2^{-K_U(\langle M \rangle)} \Mvu_{M}^\tau(\HistM)= (1-\epsilon) \Mmu_{U,\tau}(\HistM)\,.
\end{align*}
Thus, $\Mvp_{\Mmu_{U,\tau}}(\text{up}|\HistM) \geq 1 - \epsilon$. Since $\Mvp_{\mathrm{up}}(\text{up}|\HistM)=1$, the approximation holds.
\end{proof}

Putting everything together, we get the following theorem:

\begin{theorem}\label{thrm:counter-example-coupled}
Consider a $k$-E-AIXI$^{\tau}$ agent, i.e., an embedded agent employing $k$-step planning for a fixed integer $k \ge 1$, with a mixture universe model $\Mmu=\Mmu_{U,\tau}$ constructed from a Solomonoff prior over the $\tau$-POM-computable universes. Then, there exists a computable ground-truth environment $\Mge$ and a reference universal monotone Turing machine $U$ for defining the Kolmogorov complexity $K_U$ in the Solomonoff prior, such that the agent interacting with $\Mge$ does not converge to the infinite-horizon optimal planner w.r.t. $\Mmu=\Mmu_{U,\tau}$, despite the fact that this setup satisfies the grain-of-truth property (in case, e.g., of a reflective oracle).
\end{theorem}
\begin{proof}
    Fix $R$ so that $\gamma^{k+1}<R<\gamma^k$. Let $\epsilon>0$ be sufficiently small as indicated in \cref{cor:optimal-and-k-step-approx} and then construct $U$ as in 
    \cref{lem:coupled-solomonoff-mixture-eps-approximates-MgeRk-andMvpup}. Since $\Mmu_{U,\tau}=\Mve_{\Mmu_{U,\tau}}^{\Mvp_{\Mmu_{U,\tau}}}$, the result follows.
\end{proof}

\section{Embedded Bayesian agents and related solution concepts: additional information}\label{app:eba-solution-concpets}
\subsection{Recursive belief updates of mixture universes}\label{app:recursive-belief-updates}
The embedded agents we consider use as their predictive model a Bayesian mixture universe $\Mmu$: Given a countable class of universes $\calMuni$, we start with a prior belief distribution $w(\Mvu):=w(\Mvu \mid \varepsilon)$ for all $\Mvu \in \calMuni$, which induces a mixture $\Mmu$ as follows
$$\Mmu:=\sum_{\Mvu\in\calMuni}w(\Mvu)\Mvu\,.$$

Now if we observe a history $\HistM$, then Bayes' rule implies that the posterior beliefs about the universe $\Mvu$ should be updated to
$$w(\Mvu|\HistM):=\frac{w(\Mvu)\Mvu(\HistM)}{\sum_{\Mvu'\in\calMuni}w(\Mvu')\Mvu'(\HistM)}=\frac{w(\Mvu)\Mvu(\HistM)}{\Mmu(\HistM)}\,.$$

On the other hand, if we observe a history $\HistM a_t$, then Bayes' rule implies that the posterior beliefs about the universe $\Mvu$ should be updated to
$$w(\Mvu|\HistM a_t):=\frac{w(\Mvu)\Mvu(\HistM a_t)}{\Mmu(\HistM a_t)}=\frac{w(\Mvu)\Mvu(\HistM)\Mvu(a_t|\HistM)}{\Mmu(\HistM)\Mmu(a_t|\HistM)}= w(\Mvu\mid \HistM) \frac{\Mvu(a_t \mid \HistM)}{\Mmu(a_t \mid \HistM)}\,.$$

Similarly, one can show that
$$w(\Mvu\mid \Hist) = w(\Mvu\mid \HistM a_t) \frac{\Mvu(e_t \mid \HistM a_t)}{\Mmu(e_t \mid \HistM a_t)}.$$

Now according to the mixture $\Mmu$, the probability of observing $a_t$ after having observed $\HistM$ is given by
\begin{align*}
    \Mmu(a_t|\HistM)&=\frac{\Mmu(\HistM a_t)}{\rho(\HistM)}=\frac{\sum_{\Mvu\in\calMuni}w(\Mvu)\Mvu(\HistM a_t)}{\rho(\HistM)}\\
    &=\sum_{\Mvu\in\calMuni}\frac{w(\Mvu)\Mvu(\HistM)\Mvu(a_t|\HistM)}{\rho(\HistM)}=\sum_{\Mvu\in\calMuni}w(\Mvu|\HistM)\Mvu(a_t|\HistM)\,.
\end{align*}
Similarly, one can show that
\begin{align*}
    \Mmu(e_t|\HistM a_t)&=\sum_{\Mvu\in\calMuni}w(\Mvu|\HistM a_t)\Mvu(e_t|\HistM a_t)\,.
\end{align*}

We can summarize the above discussion with the following recursive equations, which could be taken as recursive definitions of the involved quantities: 
\begin{equation}\label{eqn-app:embedded-bayes-mixture}
\begin{split}
    \Mmu(a_t \mid \HistM) = \sum_{\Mvu\in \calMuni} w(\Mvu\mid \HistM)\Mvu(a_t \mid \HistM)\,, \quad \quad w(\Mvu\mid \Hist) = w(\Mvu\mid \HistM a_t) \frac{\Mvu(e_t \mid \HistM a_t)}{\Mmu(e_t \mid \HistM a_t)}\,, \\
    \Mmu(e_t \mid \HistM a_t) = \sum_{\Mvu\in \calMuni} w(\Mvu\mid \HistM a_t)\Mvu(e_t \mid \HistM a_t)\,, \quad \quad w(\Mvu\mid \HistM a_{t}) = w(\Mvu\mid \HistM) \frac{\Mvu(a_t \mid \HistM)}{\Mmu(a_t \mid \HistM)}\,.
\end{split}
\end{equation}

\subsection{Additional information on the differences between the decoupled and embedded formalisms}\label{app:decoupled-vs-embedded}
\begin{remark}[On the differences between the embedded and decoupled formalisms]
    \label{rem-app:coupled-vs-decoupled}
    The reader might be wondering whether the formalism of embedded Bayesian agents is really different from the formalism of decoupled Bayesian agents: Assuming that we have a class of universes $\calMuni$ with a prior $w_{\mathrm{uni}}(\Mvu)$ over universes $\Mvu\in\calMuni$, which induces a mixture universe $\Mmu=\sum_\Mvu w_{\mathrm{uni}}(\Mvu)\Mvu$, is it possible to define a class of environments $\calMenv$ with a prior $w_{\mathrm{env}}(\Mve)$ over environments $\Mve\in\calMenv$ inducing a mixture environment $\Mme=\sum_\Mve w_{\mathrm{env}}(\Mve)\Mve$ in such a way that $\Mme(e_t|\HistM a_t)=\Mmu(e_t|\HistM a_t)$ (which would mean that the decoupled best response policy w.r.t. $\Mme$ is the same as the embedded best response w.r.t. $\Mmu$)?

    A seemingly reasonable approach that one might try is to define for every universe $\Mvu\in\calMuni$ its induced environment $\Mve_\Mvu$ given by\footnote{We assume for the sake of this discussion that $\Mvu(\HistM a_t)>0$ for all $\Mvu\in\calMuni$ so that $\Mvu(e_t|\HistM a_t)$ is well defined.} $\Mve_\Mvu(e_t|\HistM a_t)=\Mvu(e_t|\HistM a_t)$, and then define $\calMenv=\{\Mve_\Mvu:\Mvu\in\calMuni\}$ with the prior\footnote{Note that this prior is effectively the probability distribution of $\Mve_{\Mvu}$ when $\Mvu$ is chosen according to the prior $w_{\mathrm{uni}}$.}
    $$w_{\mathrm{env}}(\Mve)=\sum_{\substack{\Mvu\in\calMuni:\\\Mve_\Mvu=\Mve}}w_{\mathrm{uni}}(\Mvu)\,,$$
    i.e., for every $\Mve\in\calMenv$ we add up the probabilities (according to the $w_{\mathrm{uni}}$ prior) of all universes whose induced environments are all equal to $\Mve$.\footnote{It is possible for two different universes in $\calMuni$ to have the same induced environment, but this would of course require that they have a different induced policy $\Mvu(a_t|\HistM)$.} The resulting mixture environment is
    $$\Mme=\sum_{\Mve\in\calMenv}w_{\mathrm{env}}(\Mve)\Mve=\sum_{\Mve\in\calMenv}\sum_{\substack{\Mvu\in\calMuni:\\\Mve_\Mvu=\Mve}}w_{\mathrm{uni}}(\Mvu)\Mve=\sum_{\Mvu\in\calMuni}w_{\mathrm{uni}}(\Mvu)\Mve_\Mvu\,.$$
    However, the mixture environment $\Mme=\sum_{\Mvu\in\calMuni}w_{\mathrm{uni}}(\Mvu)\Mve_\Mvu$ in this approach would not necessarily match the environment $\Mmu(e_t|\HistM a_t)$ induced by the mixture universe $\Mmu$: Consider the case where $\calA=\{a_0,a_1\}$ and $\calE=\{e_0,e_1\}$, and assume that we have two universes $\Mvu_0$ and $\Mvu_1$ satisfying\footnote{For the purposes of the example, we only care about the marginal distribution of $\Mvu_0$ and $\Mvu_1$ over the first turn.}
    \begin{equation*}
        \begin{aligned}
        \Mvu_0(a_0e_0)&=(1-\epsilon)^2\,,\\
        \Mvu_0(a_0e_1)&=\epsilon(1-\epsilon)\,,\\
        \Mvu_0(a_1e_0)&=\epsilon(1-\epsilon)\,,\\
        \Mvu_0(a_1e_1)&=\epsilon^2\,,\\
        \end{aligned}
        \quad\quad\text{and}\quad\quad
        \begin{aligned}
        \Mvu_1(a_0e_0)&=\epsilon^2\,,\\
        \Mvu_1(a_0e_1)&=\epsilon(1-\epsilon)\,,\\
        \Mvu_1(a_1e_0)&=\epsilon(1-\epsilon)\,,\\
        \Mvu_1(a_1e_1)&=(1-\epsilon)^2\,,\\
        \end{aligned}
    \end{equation*}
    where $\epsilon\in(0,1/2)$ is some fixed small number. Define $\calMuni=\{\Mvu_0,\Mvu_1\}$ and assume the uniform prior $w_{\mathrm{uni}}(\Mvu_0)=w_{\mathrm{uni}}(\Mvu_1)=\frac{1}{2}$. A simple calculation reveals that:
    \begin{itemize}
        \item $\Mmu(a_0e_0)=\frac{1}{2}(1-\epsilon)^2+\frac{1}{2}\epsilon^2$ and $\Mmu(a_0e_1)=\epsilon(1-\epsilon)$, and hence $\Mmu(e_0|a_0)=\frac{\Mmu(a_0e_0)}{\Mmu(a_0e_0)+\Mmu(a_0e_1)}=1-2\epsilon(1-\epsilon)$.
        \item $\Mve_{\Mvu_0}(e_0|a_0)=\Mvu_0(e_0|a_0)=\frac{\Mvu_0(a_0e_0)}{\Mvu_0(a_0e_0)+\Mvu_0(a_0e_1)}=1-\epsilon$, and similarly $\Mve_{\Mvu_1}(e_0|a_0)=\Mvu_1(e_0|a_0)=\epsilon$. This implies that
        $$\Mme(e_0|a_0)=w_{\mathrm{uni}}(\Mvu_0)\Mve_{\Mvu_0}(e_0|a_0)+w_{\mathrm{uni}}(\Mvu_1)\Mve_{\Mvu_1}(e_0|a_0)=\frac{1}{2}(1-\epsilon)+\frac{1}{2}\epsilon=\frac{1}{2}\,,$$
        which is different from $\Mmu(e_0|a_0)=1-2\epsilon(1-\epsilon)$ because $\epsilon<\frac{1}{2}$.
    \end{itemize}
    Therefore, the mixture environment $\Mme=\sum_{\Mvu\in\calMuni}w_{\mathrm{uni}}(\Mvu)\Mve_\Mvu$ is different from the environment induced by the mixture universe $\Mmu=\sum_{\Mvu\in\calMuni}w_{\mathrm{uni}}(\Mvu)\Mvu$.
    
    The above discussion does not preclude the possibility of defining some other class $\calMenv$ with an associated prior $w_{\mathrm{env}}$ inducing a mixture environment $\Mme$ matching  the environment induced by $\Mmu$: A trivial way to achieve this is to just define a single environment $\Mve_\Mmu$ as $\Mve_\Mmu(e_t|\HistM a_t)=\Mmu(e_t|\HistM a_t)$, and then let $\calMenv=\{\Mve_\Mmu\}$ with the prior $w_{\mathrm{env}}(\Mve_\Mmu)=1$, which would make $\Mme=\Mve_\Mmu$ and hence $\Mme(e_t|\HistM a_t)=\Mmu(e_t|\HistM a_t)$, as desired.

    As far as the Bayes-optimal policy is concerned, the embedded setting with respect to $w_{\mathrm{uni}}$ is mathematically equivalent to the decoupled setting with respect to the prior $w_{\mathrm{env}}$ satisfying $w_{\mathrm{env}}(\Mve_\Mmu)=1$, because $\Mme(e_t|\HistM a_t)=\Mmu(e_t|\HistM a_t)$. Nevertheless, the two settings differ in the Bayesian belief structure (assuming $\calMuni$ contains at least two different universes): The decoupled Bayesian agent optimizing w.r.t. $\Mme$ defined above really has a single environment $\Mve_\Mmu$ in its hypothesis class $\calMenv$, with $w_{\mathrm{env}}(\Mve_\Mmu)=1$, and hence it is \emph{certain about the environment it is interacting with}, while the embedded Bayesian agent optimizing w.r.t. (the environment induced by) $\Mmu$ is \emph{uncertain about the universe it is living in} since there is no single $\Mvu\in\calMuni$ satisfying $w_{\mathrm{uni}}(\Mvu)=1$. If we are only interested in Bayes-optimal agents, then this difference is mainly philosophical, because the Bayes-optimal policy is the same in both cases. However, if one considers variants of the Bayes-optimal agents which also do some stronger form of exploration\footnote{E.g., something similar to the Thompson sampling variant of AIXI \citep{leike2016thompson}, or the strongly optimal agent Inq introduced in \citet{cohen2019inq}.} to reduce the uncertainty about the ground-truth universe (resp. ground-truth environment) before applying the best response policy, then the difference between $w_{\mathrm{uni}}$ and $w_{\mathrm{env}}$ becomes important because the exploration strategy would necessarily require the use of the underlying Bayesian beliefs in order to reduce the uncertainty over the universe (resp. environment) in a principled manner. If the prior over environments satisfies $w_{\mathrm{env}}(\Mve_\Mmu)=1$, then there is no need to do any exploration for a "decoupled exploring agent" because there is no uncertainty on the environment. On the other hand, an "embedded exploring agent" would need to explore to reduce its uncertainty about the universe it is living in. We do not consider embedded exploring agents in this paper, and leave the investigation of such agents for future research.

    In \cref{sec:emb-bay-agents-structural-similarities,sec:subj-emb-eq}, we analyze the differences between the coupled and decoupled settings in further depth.
\end{remark}

\subsection{Spohn's dependency equilibria and its relation to embedded equilibria}
\label{app:equivalence-code-de}

In \cref{sec:eq-behavior-repeated-games}, we introduced the subjective embedded equilibrium (SEE) and its common knowledge counterpart, the embedded equilibrium (EE), which were inspired by the \textit{dependency equilibrium} (DE) of \citet{spohn2007dependency}. The goal of this section is to formalize this relationship. We first define Spohn's original dependency equilibrium and then investigate its precise connection to our framework. As Spohn's dependency equilibria were originally defined for normal-form (single-shot) games, we restrict our analysis in this section to that setting. We will show that the dependency equilibrium is a special case of a \textit{correlated} version of our embedded equilibrium.

\begin{definition}[Dependency Equilibrium - DE]
For a normal-form game, a dependency distribution $p(\bar{a})$ with a corresponding conditional completion constitutes a \textit{dependency equilibrium} if: For each agent $i$, any action $a^i$ with marginal probability $p(a^i)>0$ maximizes its expected utility with respect to the completed conditionals $p(a^{-i}|a'^i)$:
    $$a^i \in \arg\max_{a'^i \in \mathcal{A}^i} \mathbb{E}_{p(a^{-i}|a'^i)}[r^i(a'^i, a^{-i})]\,.$$
\end{definition}

The primary difference between our embedded equilibrium (EE, \cref{def:ode}) and Spohn's dependency equilibrium (DE) lies in what the best response is computed against. For an EE, the best response is computed w.r.t. the conditional completion of the \textit{ground-truth universe} $\Mgu^i = (\Mge^i)^{\Mgp^i}$, which is induced by the agents' actual policies. This grounds the equilibrium in the observable reality of the agents' interactions. In contrast, a DE computes its best response w.r.t. a given dependency distribution $p(\bar{a})$ which does not necessarily have to correspond to the ground-truth universe on the play-path. However, as we will show, when we introduce a correlation device \citep{aumann1974subjectivity} to define a \textit{correlated embedded equilibrium} (CEE), this concept encapsulates Spohn's original dependency equilibrium.

\begin{definition}[Correlated Embedded Equilibrium - CEE]\label{def:correlated-ode}
Let $(M, p)$ be a correlation device, let $(\Mgp^i(a^i|m^i))_{i \in N}$ be a set of policies, and let $q(a^{-i} \mid a^i, \bar{m})$ be conditional completions of a correlated dependency distribution $q(\overline{a}|\overline{m})$. Let the ground-truth distribution over actions given the joint message $\bar{m}$ be given by $$\Mgu(\bar{a} \mid \bar{m}) = \prod_{j \in N} \Mgp^j(a^j \mid m^j)\,.$$

We say that the tuple $(\{\Mgp^i\}_{i\in[N]},(M,p),q)$ forms a \textit{correlated embedded equilibrium} (CEE) for a normal-form game if for each agent $i$ and each message $m^i$ with $p(m^i) > 0$:
The agent's policy $\Mgp^i(\cdot \mid m^i)$ is an embedded best response w.r.t. the conditional completion of $\Mgu$ by $q$. That is, the expected reward under the policy $\Mgp^i$ is maximal, i.e.,
    \begin{equation}
    \label{eq:cee-br}
    \begin{aligned}
    &\mathbb{E}_{\Mgp^i(a^i \mid m^i)}\left[\mathbb{E}_{p(m^{-i}\mid m^i)}\left[\mathbb{E}_{\Mgu(a^{-i} \mid a^i, \bar{m})}[r^i(a^i, a^{-i})]\right]\right] = \max_{a'^i \in \mathcal{A}^i} \mathbb{E}_{p(m^{-i}\mid m^i)}\left[\mathbb{E}_{\Mgu(a^{-i} \mid a'^i, \bar{m})}[r^i(a'^i, a^{-i})]\right]\,,
    \end{aligned}
    \end{equation}
    with
    $$\Mgu(a^{-i} \mid a'^i, \bar{m}) = \begin{cases}\Mgu(a^{-i} \mid a'^i, \bar{m}) \quad &\text{if}~~ \Mgu(a'^i\mid \bar{m})=\Mgp^i(a'^i \mid m^i)>0\,, \\
     q(a^{-i} \mid a'^i, \bar{m}) \quad &\text{otherwise},\end{cases}$$
where the term $\Mgu(a^{-i} \mid  a'^i,\bar{m})$ on the left-hand side is the completed conditional, whereas the term $\Mgu(a^{-i} \mid  a'^i,\bar{m})$ on the right-hand side (for the case $ \Mgu(a'^i|\bar{m})>0$) is the conditional from the original ground-truth measure $\Mgu$. As can be seen, the conditional $q(a^{-i} \mid  a'^i,\bar{m})$ helps extend the definition of $\Mgu(a^{-i} \mid  a'^i,\bar{m})$ to the cases where $\Mgu( a'^i|\bar{m})=0$.

\end{definition}

\begin{proposition}\label{prop:equivalence-de-cee}
Each dependency equilibrium can be formulated as a correlated embedded equilibrium in the following sense: For every dependency distribution $p(\bar{a})$ with conditional completions $(p(a^{-i}|a'^i))_{i\in[N]}$ which form a dependency equilibrium, there exists a correlated embedded equilibrium $(\{\Mgp^i\}_{i\in[N]},(M,p'),q)$ such that
$$p(\overline{a})=\expect{p'(\overline{m})}{\prod_{i\in[N]}\Mgp^i(a^i|m^i)}\,.$$
\end{proposition}

\begin{proof}
Let $p(\bar{a})$ be a dependency equilibrium (DE) with completed conditionals $p(a^{-i}|a'^i)$. We will construct an equivalent correlated embedded equilibrium (CEE), denoted $(\{\Mgp^i\}, (M, p'), q)$.

Let the message space for each agent be its action space, $M^i = \mathcal{A}^i$. Define the probability distribution $p'$ over the joint message space $M = \prod_i \mathcal{A}^i$ to be the dependency distribution $p$ from the DE. So, $p'(\bar{m}) := p(\bar{m})$. The message $\bar{m}$ can be interpreted as a "recommended" joint action.

Now we define the agents' policies to be the ones that obediently follow the recommendation, i.e., for each agent $i$, we define its policy $\Mgp^i$ as:
    $$
    \Mgp^i(a^i|m^i) = \delta(a^i=m^i)= \begin{cases} 1 & \text{if } a^i = m^i\,, \\ 0\,, & \text{otherwise.} \end{cases}
    $$

Notice that for every $\overline{a}=(a^i)_{i\in[N]}$, we have
\begin{align*}
    \expect{p'(\overline{m})}{\prod_{i\in[N]}\Mgp^i(a^i|m^i)} =\expect{p'(\overline{m})}{\prod_{i\in[N]}\delta(a^i=m^i)} =\expect{p'(\overline{m})}{\delta(\overline{a}=\overline{m})}=p'(\overline{a}) =p(\overline{a})\,.
\end{align*}

Now define the correlated dependency distribution $q(\overline{a}|\overline{m})$ and its conditional completions $q(a^{-i} \mid a'^i, \bar{m})$ using the DE's dependency distribution $p(\overline{a})$ and its conditional completions $p(a^{-i} \mid a'^i)$ as follows:
\begin{align*}
    q(\overline{a} \mid  \bar{m}) := p(\overline{a})\,,\quad\text{and}\quad q(a^{-i} \mid a'^i, \bar{m}) := p(a^{-i} \mid a'^i)\,.
\end{align*}

Let us now show that $(\{\Mgp^i\}_{i\in[N]},(M,p'),q)$ indeed forms a CEE by checking the conditions of \cref{def:correlated-ode}. The ground-truth universe is given by $$\Mgu(\bar{a} \mid \bar{m}) = \prod_j \Mgp^j(a^j \mid m^j) = \delta(\bar{a} = \bar{m})\,,$$ where $\delta(\cdot)$ is the indicator function.
        
We must show that the obedient policy $\Mgp^i$ is a best response for each $i$ and $m^i$ with $p'(m^i) > 0$. The CEE best-response condition in \eqref{eq:cee-br} requires the expected reward from the policy (LHS) to equal the maximum possible expected reward (RHS).
        
The LHS in \eqref{eq:cee-br} evaluates the reward for the obedient action $a^i = m^i$. This action is on the play-path ($\Mgp^i(a^i|m^i)=\Mgp^i(m^i|m^i)=1 > 0$), so the inner expectation uses the ground-truth conditional $\Mgu(a^{-i} \mid a^i, \bar{m})$:
$$ \Mgu(a^{-i} \mid a^i, \bar{m}) = \frac{\Mgu(\bar{a} \mid \bar{m})}{\Mgu(a^i \mid \bar{m})} = \frac{\delta(\bar{a}=\bar{m})}{\delta(a^i=m^i)} = \delta(a^{-i}=m^{-i})\,, $$
where in the last equality we used the fact that $\delta(a^i=m^i)=1$, as $a^i=m^i$.
The expected reward for the obedient action $a^i=m^i$ is:
\begin{align*}
\text{LHS} &= \mathbb{E}_{p'(m^{-i}\mid m^i)}\left[\mathbb{E}_{\delta(a^{-i}=m^{-i})}[r^i(m^i, a^{-i})]\right] \\
&= \mathbb{E}_{p'(m^{-i}\mid m^i)}\left[r^i(m^i, m^{-i})\right] = \sum_{m^{-i}} \frac{p'(\bar{m})}{p'(m^i)} r^i(m^i, m^{-i})\,.
\end{align*}
Since $p'(\bar{m}) = p(\bar{m})$, this is exactly $\mathbb{E}_{p(a^{-i}|a^i)}[r^i(a^i, a^{-i})]$ (with $a^i=m^i$).

The RHS in \eqref{eq:cee-br} considers an arbitrary action $a'^i$. If $a'^i \neq m^i$, this action is off the play-path ($\Mgp^i(a'^i|m^i)=0$). Therefore, the inner expectation uses the dependency distribution $q$:
$$ \Mgu(a^{-i} \mid a'^i, \bar{m}) = q(a^{-i} \mid a'^i, \bar{m}) = p(a^{-i} \mid a'^i) \,.$$
The expected reward for deviating to $a'^i$ is:
$$ \mathbb{E}_{p'(m^{-i}\mid m^i)}\left[\mathbb{E}_{p(a^{-i} \mid a'^i)}[r^i(a'^i, a^{-i})]\right] = \mathbb{E}_{p(a^{-i} \mid a'^i)}[r^i(a'^i, a^{-i})]\,. $$
The DE best-response condition states that for any $a^i$ with $p(a^i)>0$ (which is equivalent to $p'(m^i)>0$ for $m^i=a^i$), $a^i$ maximizes the expected utility:
$$ \mathbb{E}_{p(a^{-i}|a^i)}[r^i(a^i, a^{-i})] \geq \mathbb{E}_{p(a^{-i}|a'^i)}[r^i(a'^i, a^{-i})] \quad \forall a'^i\,. $$
This is exactly $\text{LHS} \geq \text{RHS}$ for any $a'^i$. Thus, the obedient action $a^i=m^i$ achieves the maximum, satisfying the CEE best-response condition.

Thus, any DE can be formulated as an equivalent CEE.
\end{proof}

\subsection{Additional information on \cref{example:bayesian-agents-cooperation}}\label{app:example-bayesian-agents-cooperation}

Take $Q^i_{\Mmu^i}(\RghM, a^i) := \expect{\Mmu^i(a^{-i} \mid \RghM,a^i)}{r^i(a^i, a^{-i})}$.
We will show that for $\alpha>\frac{m^{*}_\infty}{1 + m^*_\infty}$ (with $m^{*}_\infty \in (0,1]$ and will be defined later), we have that there exists some time $T$ such that for every $t>T$ and every trajectory $\RghM$ satisfying $a^i_{t'} = a^{-i}_{t'}$ for each $t'<t$, we have $Q^i_{\Mmu^i}(\RghM, a^i=C) > Q^i_{\Mmu^i}(\RghM, a^i=D)$. This means that there exists some time $T$ for which both agents choose to cooperate for all $t>T$, thereby proving the main claim of \cref{example:bayesian-agents-cooperation}. 

% Let us take $h$ a trajectory of arbitrary length where for each $t$ we have that $a^i_t = a^{-i}_t$. Hence, in the following, we have that $\RghM=h^{-i}$.
Let us define the following helper variables:
\begin{align*}
    L(\pi^i,\RghM)&:= \prod_{k=1}^{t-1}\pi^i(a^i_k \mid \RghM[k] )\,, \quad  L(\pi^i,\RghM a^i):= L(\pi^i,\RghM)\pi^i(a^i\mid \RghM)\,,\\
    m(\RghM) &:= \sum_{\pi^i \in \calMpol}\tilde{w}(\pi^i) L(\pi^i,\RghM)\,, \quad m(\RghM a^i):=  \sum_{\pi^i \in \calMpol}\tilde{w}(\pi^i)L(\pi^i,\RghM a^i)\,.
\end{align*}
Since all the policies of $\calMpol$ are deterministic, we have $L(\pi^i,\RghM) \in \{0,1\}$. Using this notation, we have that
\begin{align*}
    \Mmu^i(\RghM)&:= \sum_{\Mvu \in \calMuni}w^i(\Mvu)\Mvu(\RghM) = \sum_{\pi^i \in \calMpol}\sum_{\pi^{-i} \in \calMpol}\hat{w}^i(\pi^i,\pi^{-i})\MultiAgentMvp(\RghM) \\
    &=\sum_{\pi^i \in \calMpol}\sum_{\pi^{-i} \in \calMpol}\left[\alpha \tilde{w}(\pi^i)\delta(\pi^i = \pi^{-i}) + (1-\alpha) \tilde{w}(\pi^i)\tilde{w}(\pi^{-i})\right]L(\pi^i,\RghM)L(\pi^{-i},\RghM) \\
    &\stackrel{(\ast)}=\alpha\sum_{\pi^i \in \calMpol}\sum_{\pi^{-i} \in \calMpol}\tilde{w}(\pi^i)L(\pi^i,\RghM)^2 + (1-\alpha)\sum_{\pi^i \in \calMpol}\sum_{\pi^{-i} \in \calMpol}\tilde{w}(\pi^i)\tilde{w}(\pi^{-i})L(\pi^i,\RghM)L(\pi^{-i},\RghM) \\
    &\stackrel{(\dagger)}=\alpha\sum_{\pi^i \in \calMpol}\sum_{\pi^{-i} \in \calMpol}\tilde{w}(\pi^i)L(\pi^i,\RghM) + (1-\alpha) m(\RghM)^2,\\
    &=\alpha m(\RghM) + (1-\alpha) m(\RghM)^2\,,
\end{align*}
where in $(\ast)$ we used the fact that $\RghM$ is symmetric, i.e., $a_t^i=a_t^{-i}$ for all timesteps, and hence $L(\pi^i,\RghM) = L(\pi^{-i},\RghM)$ in case $\pi^i = \pi^{-i}$. In $(\dagger)$ we used the fact that $L(\pi^i,\RghM) \in \{0,1\}$ and hence $L(\pi^i,\RghM)^2 = L(\pi^i,\RghM)$.
Using a similar derivation, we have that 
\begin{align*}
    \Mmu^i(\RghM a^i) &= \sum_{\pi^i \in \calMpol}\sum_{\pi^{-i} \in \calMpol}\left[\alpha \tilde{w}(\pi^i)\delta(\pi^i = \pi^{-i}) + (1-\alpha) \tilde{w}(\pi^i)\tilde{w}(\pi^{-i})\right]L(\pi^i,\RghM a^i)L(\pi^{-i},\RghM) \\
    &= \alpha m(\RghM a^i) + (1-\alpha)m(\RghM)m(\RghM a^i)\,,
\end{align*}
where we used the fact that $L(\pi^i,\RghM a^i)=1 \implies L(\pi^i,\RghM)=1$. Finally, we have that 
\begin{align*}
    \Mmu^i(\RghM a^ia^{-i}) &= \sum_{\pi^i \in \calMpol}\sum_{\pi^{-i} \in \calMpol}\left[\alpha \tilde{w}(\pi^i)\delta(\pi^i = \pi^{-i}) + (1-\alpha) \tilde{w}(\pi^i)\tilde{w}(\pi^{-i})\right]L(\pi^i,\RghM a^i)L(\pi^{-i},\RghM a^{-i}) \\
    &= \alpha m(\RghM a^i)\delta(a^i=a^{-i}) + (1-\alpha)m(\RghM a^i)m(\RghM a^{-i})\,, \\
    \Mmu^i(a^{-i} \mid \RghM a^i) &= \frac{\Mmu^i(\RghM a^ia^{-i})}{\Mmu^i(\RghM a^i)} = \frac{\alpha m(\RghM a^i)\delta(a^i=a^{-i}) + (1-\alpha)m(\RghM a^i)m(\RghM a^{-i})}{\alpha m(\RghM a^i) + (1-\alpha)m(\RghM)m(\RghM a^i)} \\
    &= \frac{\alpha \delta(a^i=a^{-i}) + (1-\alpha)m(\RghM a^{-i})}{\alpha + (1-\alpha)m(\RghM)}\,,
\end{align*}
where we used the fact that $m(\RghM a^i) > 0$ as the grain-of-uncertainty property is satisfied. Using the rewards $r(C,C)=2, r(D,D)=1, r(C,D)=0, r(D,C)=3$, we have that
\begin{align*}
    Q^i_{\Mmu^i}(\RghM, a^i=C) &- Q^i_{\Mmu^i}(\RghM, a^i=D) \\
    &= \expect{\Mmu^i(a^{-i} \mid \RghM,a^i=C)}{r^i(a^i, a^{-i})} - \expect{\Mmu^i(a^{-i} \mid \RghM,a^i=D)}{r^i(a^i, a^{-i})} \\
    &=\frac{\alpha + (1-\alpha)m(\RghM C)}{\alpha + (1-\alpha)m(\RghM)} r(C,C)+\frac{(1-\alpha)m(\RghM D)}{\alpha + (1-\alpha)m(\RghM)} r(C,D)\\
    &\quad- \frac{(1-\alpha)m(\RghM C)}{\alpha + (1-\alpha)m(\RghM)} r(D,C)-\frac{\alpha + (1-\alpha)m(\RghM D)}{\alpha + (1-\alpha)m(\RghM)} r(D,D) \\
    &=\frac{\alpha + (1-\alpha)m(\RghM C)}{\alpha + (1-\alpha)m(\RghM)} \times 2 - \frac{(1-\alpha)m(\RghM C)}{\alpha + (1-\alpha)m(\RghM)} \times 3 -\frac{\alpha + (1-\alpha)m(\RghM D)}{\alpha + (1-\alpha)m(\RghM)} \times 1 \\
    &=\frac{1}{\alpha + (1-\alpha)m(\RghM)} \left[ \alpha - (1-\alpha)(m(\RghM C) + m(\RghM D)) \right] \\
    &= \frac{\alpha - (1-\alpha)m(\RghM)}{\alpha + (1-\alpha)m(\RghM)}\,. \\
    % &\geq \frac{\alpha - (1-\alpha)}{\alpha + (1-\alpha)m(\RghM)} 
\end{align*}
Hence, $Q^i_{\Mmu^i}(\RghM, a^i=C)> Q^i_{\Mmu^i}(\RghM, a^i=D)$ iff
\begin{align*}
    \alpha > \frac{m(\RghM)}{1 + m(\RghM)}\,.
\end{align*}
As $m(\RghM) \in [0,1]$, $\frac{m(\RghM)}{1 + m(\RghM) }$ is decreasing with decreasing $m(\RghM) $. Furthermore, as $\RghM $ is a prefix of $\Rgh$, we have that $m(\Rgh) \leq m(\RghM)$, i.e., $m(\RghM )$ is a non-increasing sequence with time, which converges to some value $m_\infty$. Now let us define the maximum possible $m_\infty$ as 
\begin{align*}
    m^*_{\infty} := \lim_{t\to \infty} \max_{\Rgh \in \TurnSet^t} m(\Rgh)\,.
\end{align*}
Now if we take $\alpha > \frac{m^*_\infty}{1 + m^*_\infty}$, we are guaranteed that there exists some time $T$ such that for all $t>T$, we have that $\alpha > \frac{m(\RghM )}{1 + m(\RghM )}$ and hence both agents mutually cooperate for all $t > T$. Such policies that forever cooperate after some time $T$ are contained within the hypothesis class, and hence the grain-of-truth property is satisfied.

In fact, we can obtain a complete characterization of the behavior of the EBR policy in case of self-play: Let $$m_{k}^{\textrm{defect}}=m((D,D)^k)\,,$$
where we adopt the convention that $m_{0}^{\textrm{defect}}=m((D,D)^0)=m(\varepsilon)=1$, and
$$m_{\infty}^{\textrm{defect}}=\lim_{k\to\infty} m((D,D)^k)\,.$$

Let $T$ be the smallest $k$ for which $\alpha> \frac{m_k^{\textrm{defect}}}{1+m_k^{\textrm{defect}}}$. Then, we have $\alpha\leq  \frac{m((D,D)^k)}{1+m((D,D)^k)}$ for all $k< T$ which means that the EBR policy will defect\footnote{Here we assume that the default action in case of a tie is defection. If the default action in case of a tie is cooperation, then we let $T$ be the smallest $k$ for which $\alpha\geq \frac{m_k^{\textrm{defect}}}{1+m_k^{\textrm{defect}}}$.} up until the $T$-th round. Now since $\alpha> \frac{m_T^{\textrm{defect}}}{1+m_T^{\textrm{defect}}}= \frac{ m((D,D)^T)}{1+ m((D,D)^T)}$, we can see that we get cooperation on the $(T+1)$-th round. Furthermore, since\footnote{This is because $m((D,D)^T(C,C)^\ell)\leq m((D,D)^T)$.} $\alpha> \frac{m((D,D)^T)}{1+m((D,D)^T)}\geq  \frac{m((D,D)^T(C,C)^\ell)}{1+m((D,D)^T(C,C)^\ell)} $ for every $\ell\geq 0$, we can see that we will get cooperation in all rounds after the $T$-th round.

If $\alpha\leq\frac{m_{\infty}^{\textrm{defect}}}{1+m_{\infty}^{\textrm{defect}}}$, then we have that at each timestep $Q^i_{\Mmu^i}(\RghM, a^i=C)\leq Q^i_{\Mmu^i}(\RghM, a^i=D)$ and hence both agents will always defect, starting from timestep 1. 

\subsection{Additional information on \cref{example:bayesian-agents-cooperation-sne}}\label{app:example-bayesian-agents-cooperation-sne}

We analyze the decision-making of a decoupled Bayesian agent $i$ at the start of round $t$, given a personal history $\HistMI$. Since the discount factor is $\gamma=0$, the agent's optimal action $a^i_t$ is the one that maximizes the expected immediate reward, given its beliefs. This is captured by the $Q$-value function:
$$Q_{\Mme^i}(\HistMI, a^i) := \mathbb{E}_{\Mme^i(e^i \mid \HistMI, a^i)}[r^i(e^i)]\,.$$
Given that the percept is $e^i = (a^{-i}, r^i)$ and the reward function $r(a^i, a^{-i})$ is known, the $Q$-value simplifies to an expectation over the opponent's action $a^{-i}$:
$$Q_{\Mme^i}(\HistMI, a^i) = \sum_{a^{-i} \in \{C,D\}} \Mme^i(a^{-i} \mid \HistMI, a^i) r(a^i, a^{-i})\,.$$
where $\Mme^i(a^{-i} \mid \HistMI, a^i)$ is the agent's predictive marginal probability for the opponent's action, derived from its mixture environment $\Mme^i$.

To proceed, we must assume a symmetric history, which arises because both agents are identical, start with the same priors, and follow the same deterministic (tie-breaking) best response policy. Thus, $a^i_k = a^{-i}_k = a_k$ for all $k < t$, and the personal history is
\begin{equation}
\label{eq:symmetric-history}
\HistMI = (a_1, (a_1, r(a_1, a_1)), \dots, a_{t-1}, (a_{t-1}, r(a_{t-1}, a_{t-1})))\,.
\end{equation}

Let's first define the likelihood of this history $\HistMI$ under each hypothesis $\nu \in \calMenv$.
\begin{itemize}
    \item For an environment $\nu_\Mvp \in \calMpol$ (corresponding to an opponent policy $\Mvp$),
    the likelihood is $$\Mve_\Mvp(e_{<t}^i \mid \mid a_{<t}^i) = \prod_{k=1}^{t-1} \nu_\Mvp(e^i_k \mid \HistMI[k], a^i_k) = \prod_{k=1}^{t-1} \left[ \Mvp(a^{-i}_k \mid \HistMI[k]) \delta(r^i_k = r(a^i_k, a^{-i}_k)) \right]\,.$$
    On our symmetric history, $a^i_k = a^{-i}_k = a_k$ and $r^i_k = r(a_k, a_k)$, this simplifies to:
    $$\Mve_\Mvp(e_{<t}^i \mid \mid a_{<t}^i) = \prod_{k=1}^{t-1} \left[ \Mvp(a_k \mid \HistMI[k]) \delta(r(a_k, a_k) = r(a_k, a_k)) \right] = \prod_{k=1}^{t-1} \Mvp(a_k \mid \HistMI[k])\,.$$
    We define this likelihood as $L(\Mvp, \HistMI) := \prod_{k=1}^{t-1} \Mvp(a_k \mid \HistMI[k])$. (Note that the opponent's policy $\Mvp$ interprets $\HistMI[k]$ as its own history, which is identical to the ego-agent's history in this symmetric case).
    
    \item For the $\nu_{\textrm{copy}}$ environment:
    $$\nu_{\textrm{copy}}(e_{<t}^i \mid \mid a_{<t}^i) = \prod_{k=1}^{t-1} \nu_{\textrm{copy}}(e^i_k \mid \HistMI[k], a^i_k) = \prod_{k=1}^{t-1} \left[ \delta(a^{-i}_k = a^i_k) \delta(r^i_k = r(a^i_k, a^{-i}_k)) \right]\,.$$
    On our symmetric history, this is $\prod_{k=1}^{t-1} \left[ \delta(a_k = a_k) \delta(r(a_k, a_k) = r(a_k, a_k)) \right] = 1$.
\end{itemize}

From \cref{prop:behavior-decoupled-beliefs} we know that the agent's posterior belief $w(\nu \mid \HistMI)$ is updated as follows:
$$w(\nu \mid \HistI)=w(\nu \mid \HistMI)\frac{\nu(e_t^i|\HistMI a_t^i)}{\xi^i(e_t^i|\HistMI a_t^i)}\,,$$
with
$$\xi^i(e_t^i|\HistMI a_t^i):=\sum_{\nu\in\calMenv}w(\nu \mid \HistMI)\nu(e_t^i|\HistMI a_t^i)\,.$$

A simple (reverse) induction argument on $1\leq \ell\leq t$ reveals that
$$\prod_{k=\ell}^t \xi^i(e_k^i|\HistMI[k]a_k^i) = \sum_{\nu\in\calMenv} w(\nu|\HistMI[\ell]) \prod_{k=\ell}^t \nu(e_k^i|\HistMI[k]a_k^i)\,.$$

In particular for $\ell=1$, we get

$$\xi^i(e_{1:t}^i||a_{1:t}^i)=\sum_{\nu\in\calMenv}w(\nu|\varepsilon)\nu(e_{1:t}^i||a_{1:t}^i)=\sum_{\nu\in\calMenv}w(\nu)\nu(e_{1:t}^i||a_{1:t}^i)\,,$$
where
$$\xi^i(e_{1:t}^i||a_{1:t}^i)=\prod_{k=1}^t \xi^i(e_k^i|\HistMI[k]a_k^i)\quad\text{and}\quad \nu(e_{1:t}^i||a_{1:t}^i)=\prod_{k=1}^t \nu(e_k^i|\HistMI[k]a_k^i)\,.$$

Let $\HistMI$ be as in \eqref{eq:symmetric-history}, and let $a_t^i\in\calA$ and $e_t^i=(a^{-i}_t,r_t^i)\in\calE=\calA\times\calR$ be such that $r_t^i=r(a^i,a^{-i})$. Since $\xi^i(e_t^i|\HistMI[k]a_t^i)=\frac{\xi^i(e_{1:t}^i||a_{1:t}^i)}{\xi^i(e_{<t}^i||a_{<t}^i)}$, we can calculate $\xi^i(e_t^i|\HistMI[k]a_t^i)$ by analyzing $\xi^i(e_{<t}^i||a_{<t}^i)$ and $\xi^i(e_{1:t}^i||a_{1:t}^i)$:

\begin{align*}
    \Mme^i(e_{<t}^i \mid \mid a_{<t}^i) &= \sum_{\nu \in \calMenv} w(\nu) \nu(\HistMI) = w(\nu_{\textrm{copy}}) \nu_{\textrm{copy}}(e_{<t}^i \mid \mid a_{<t}^i) + \sum_{\Mvp \in \calMpol} w(\nu_\Mvp) \Mve(e_{<t}^i \mid \mid a_{<t}^i) \\
    &\stackrel{(\ast)}= \alpha \cdot 1 + \sum_{\Mvp \in \calMpol} (1-\alpha) \tilde{w}(\Mvp) L(\Mvp, \HistMI) \\
    &\stackrel{(\dagger)}= \alpha + (1-\alpha) m(\HistMI)\,,
\end{align*}
where in $(\ast)$ we use the fact that $\HistMI$ is a symmetric history and hence $\nu_{\textrm{copy}}(e_{<t}^i \mid \mid a_{<t}^i)=1$, and in $(\dagger)$ we used the definition of $m(\cdot)$ from \cref{example:bayesian-agents-cooperation-sne}:
$$m(\HistMI) := \sum_{\Mvp \in \calMpol} \tilde{w}(\Mvp) L(\Mvp, \HistMI)\,.$$

Similarly,
\begin{align*}
    \Mme^i(e_{1:t}^i \mid \mid a_{1:t}^i) &= \sum_{\nu \in \calMenv} w(\nu) \nu(\HistMI) = w(\nu_{\textrm{copy}}) \nu_{\textrm{copy}}(e_{1:t}^i \mid \mid a_{1:t}^i) + \sum_{\Mvp \in \calMpol} w(\nu_\Mvp) \Mve(e_{1:t}^i \mid \mid a_{1:t}^i) \\
    &= \alpha \cdot \delta(a^{-i}_t = a^i_t)\delta(r^{i}_t = r(a^i,a^{-i})) + \sum_{\Mvp \in \calMpol} (1-\alpha) \tilde{w}(\Mvp) L(\Mvp, \HistI) \\
    &\stackrel{(\ddagger)}= \alpha\delta(a^{-i}_t = a_t^i) + (1-\alpha) m(\HistMI a^{-i}_t)\,,
\end{align*}
where in $(\ddagger)$ we used the definition $$m(\HistMI a) := \sum_{\Mvp \in \calMpol} \tilde{w}(\Mvp) L(\Mvp, \HistMI) \Mvp(a \mid \HistMI)\,.$$

We conclude that
$$\Mme^i(a^{-i}_t \mid \HistMI, a^i_t):= \Mme^i(e^{i}_t \mid \HistMI, a^i_t) =\frac{\xi^i(e_{1:t}^i||a_{1:t}^i)}{\xi^i(e_{<t}^i||a_{<t}^i)}= \frac{\alpha \delta(a^{-i}_t = a^i_t) + (1-\alpha) m(\HistMI a^{-i}_t)}{\alpha + (1-\alpha) m(\HistMI)}\,.$$
This is mathematically identical to the conditional probability $\Mmu^i(a^{-i} \mid \RghM a^i)$ in \cref{app:example-bayesian-agents-cooperation}.

Finally, we compute the $Q$-value difference, using $r(C,C)=2$, $r(C,D)=0$, $r(D,C)=3$, $r(D,D)=1$. Using $Z:=\alpha + (1-\alpha) m(\HistMI)$, we get
\begin{align*}
    Q_{\Mme^i}(\HistMI, a^i=C) &= \Mme^i(C \mid \HistMI, C) r(C,C) + \Mme^i(D \mid \HistMI, C) r(C,D) \\
    &= \frac{\alpha \cdot 1 + (1-\alpha) m(\HistMI C)}{Z} \times 2 + \frac{\alpha \cdot 0 + (1-\alpha) m(\HistMI D)}{Z} \times 0 \\
    &= \frac{2(\alpha + (1-\alpha) m(\HistMI C))}{Z}\,,
\end{align*}
\begin{align*}
    Q_{\Mme^i}(\HistMI, a^i=D) &= \Mme^i(C \mid \HistMI, D) r(D,C) + \Mme^i(D \mid \HistMI, D) r(D,D) \\
    &= \frac{\alpha \cdot 0 + (1-\alpha) m(\HistMI C)}{Z} \times 3 + \frac{\alpha \cdot 1 + (1-\alpha) m(\HistMI D)}{Z} \times 1 \\
    &= \frac{3(1-\alpha) m(\HistMI C) + \alpha + (1-\alpha) m(\HistMI D)}{Z}\,.
\end{align*}
The difference is:
\begin{align*}
    Q_{\Mme^i}(\HistMI, C) - Q_{\Mme^i}(\HistMI, D) &= \frac{1}{Z} \left[ (2\alpha + 2(1-\alpha) m(\HistMI C)) - (3(1-\alpha) m(\HistMI C) + \alpha + (1-\alpha) m(\HistMI D)) \right] \\
    &= \frac{1}{Z} \left[ \alpha - (1-\alpha) m(\HistMI C) - (1-\alpha) m(\HistMI D) \right] \\
    &= \frac{1}{Z} \left[ \alpha - (1-\alpha) (m(\HistMI C) + m(\HistMI D)) \right] \\
    &= \frac{\alpha - (1-\alpha) m(\HistMI)}{\alpha + (1-\alpha) m(\HistMI)}\,.
\end{align*}
Therefore, cooperation is the better action (i.e., $Q_{\xi^i}(\cdot, C) > Q_{\xi^i}(\cdot, D)$) if and only if $\alpha - (1-\alpha) m(\HistMI) > 0$, which rearranges to:
$$\alpha > \frac{m(\HistMI)}{1 + m(\HistMI)}\,.$$
This is the exact same condition as in \cref{app:example-bayesian-agents-cooperation}. Let $\Turn^i_{1:k}$ be the personal history corresponding to $k$ rounds of mutual defection. We use the example's notation $m_k^{\textrm{defect}} := m(\Turn^i_{1:k})$. At time $t=k+1$, the agent's history satisfies $m(\HistMI) = m_k^{\textrm{defect}}$.

The condition for cooperation at time $k+1$ is $\alpha > \frac{m_k^{\textrm{defect}}}{1 + m_k^{\textrm{defect}}}$. As $k$ increases, $m_k^{\textrm{defect}}$ decreases and converges to $m_\infty^{\textrm{defect}}$.
If $\alpha > \frac{m_\infty^{\textrm{defect}}}{1 + m_\infty^{\textrm{defect}}}$, there will be a finite time $T$ (the smallest $k$ such that $\alpha > \frac{m_k^{\textrm{defect}}}{1 + m_k^{\textrm{defect}}}$) after which the agent's BR is to cooperate. Since both agents are identical, they both switch to cooperation at $T+1$. Furthermore, since $\alpha> \frac{m((D,D)^T)}{1+m((D,D)^T)}\geq  \frac{m((D,D)^T(C,C)^\ell)}{1+m((D,D)^T(C,C)^\ell)} $ for every $\ell\geq 0$, we can see that we will get cooperation in all rounds after the $T$-th round. 
If $\alpha \leq \frac{m_\infty^{\textrm{defect}}}{1 + m_\infty^{\textrm{defect}}}$, the condition to cooperate is never met (assuming defection is the tie-breaking action), and both agents defect forever. This confirms the claims in the example.

\section{Functional similarities through the lens of algorithmic information: additional information}\label{app:structural-similarities}
In this section, we generalize the concept of coupledness for priors over fully supported universes $\calMpolenvcheck^\tau$ towards possibly non-fully-supported universes in $\calMpolenv$, and provide additional results and proofs supporting \cref{sec:mupi-structural-similarities}. Throughout this section, we assume a fixed choice of $\calA$ and $\calE$ and a complete prefix-free encoding thereof. 

Let us first define how we can combine two priors $w^{\mathrm{p}}\in\Delta\calM_{\mathrm{pol}}^\tau$ and $w^{\mathrm{e}}\in\Delta\calM_{\mathrm{env}}^\tau$ into a prior over the set of universes $\calMpolenv$.

\begin{definition}
    \label{def:ind-priors-on-env-and-pol}
    Let $w^{\mathrm{p}}\in\Delta\calM_{\mathrm{pol}}^\tau$ and $w^{\mathrm{e}}\in\Delta\calM_{\mathrm{env}}^\tau$ be two probability priors on the spaces of policies and environments, respectively. We define the probability prior $w^{\mathrm{p}}\otimes w^{\mathrm{e}}\in\Delta\calM^\tau_{\mathrm{pol}\textrm{-}\mathrm{env}}$ as follows:
    \begin{equation}
    \label{eq:ind-priors-on-env-and-pol}
        (w^{\mathrm{p}}\otimes w^{\mathrm{e}})_{\lambda}=\sum_{\pi \in \calM^{\tau}_{\mathrm{pol}}}\sum_{\substack{\nu \in \calM^{\tau}_{\mathrm{env}}:\\\nu^\pi=\lambda}} w^{\mathrm{p}}_\pi w^{\mathrm{e}}_\nu\,,\quad \forall \lambda\in \calM^\tau_{\mathrm{pol}\textrm{-}\mathrm{env}}\,.
    \end{equation}
    In other words, $w^{\mathrm{p}}\otimes w^{\mathrm{e}}$ is the probability distribution on $\calM^\tau_{\mathrm{pol}\textrm{-}\mathrm{env}}$ which is induced by the interaction of a policy and an environment that are independently drawn from $\calM^{\tau}_{\mathrm{pol}}$ and $\calM^{\tau}_{\mathrm{env}}$ according to $w^{\mathrm{p}}$ and $w^{\mathrm{e}}$, respectively.
\end{definition}

It is worth noting that the main reason why we have the complicated \eqref{eq:ind-priors-on-env-and-pol} rather than simply defining $(w^{\mathrm{p}}\otimes w^{\mathrm{e}})_{\nu^\pi}= w^{\mathrm{p}}_\pi w^{\mathrm{e}}_\nu$ is because it is possible for two different pairs $(\pi,\nu)\neq(\tilde\pi,\tilde\nu)$ to give rise to the same universe $\nu^\pi={\tilde\nu}^{\tilde\pi}$: This can happen, e.g., if $\pi(a|\HistA )=\tilde{\pi}(a|\HistA )=0$ for some $(\HistA, a)\in(\TurnSet^*)\times\mathcal{A}$ which would then allow for $\nu$ and $\tilde\nu$ to differ on $(\HistA, a)$ (i.e., $\nu(e \mid \HistA, a)\neq \tilde\nu(e \mid \HistA, a)$ for some $e \in \calE$) while still have $\nu^\pi={\tilde\nu}^{\tilde\pi}$. The following remark summarizes this:

\begin{remark}
\label{rem:diff-pol-env-same-univ}
If $\lambda\in\calM_{\mathrm{pol}\textrm{-}\mathrm{env}}^{\tau}$, then the universe $\lambda$ does not necessarily "factorize" into a unique pair of a policy and an environment, i.e., there can be more than one pair $(\pi,\nu)\in\calM_{\mathrm{pol}}^{\tau}\times\calM_{\mathrm{env}}^{\tau}$ such that $\lambda=\nu^\pi$.
\end{remark}

\begin{definition}
    \label{def:decoupled-w-tau}
    We say that the prior $w^\tau\in\Delta'\calM_{\mathrm{univ}}^\tau$ is decoupled on $\calM^\tau_{\mathrm{pol}\textrm{-}\mathrm{env}}$, if the restriction of
    $w^\tau$ to $\calM^\tau_{\mathrm{pol}\textrm{-}\mathrm{env}}$ can be seen as the result of interaction between a policy and an environment which are chosen independently, i.e., there exist some $w^{\mathrm{p}}\in\Delta\calM_{\mathrm{pol}}^\tau$ and $w^{\mathrm{e}}\in\Delta\calM_{\mathrm{env}}^\tau$ such that for all $\lambda\in \calM^\tau_{\mathrm{pol}\textrm{-}\mathrm{env}}$, we have
        \begin{equation}
            \label{eq:decoupled-w-tau}
            w_{\lambda}^\tau=C\cdot(w^{\mathrm{p}}\otimes w^{\mathrm{e}})_{\lambda}=C\sum_{\pi \in \calM^{\tau}_{\mathrm{pol}}}\sum_{\substack{\nu \in \calM^{\tau}_{\mathrm{env}}:\\\nu^\pi=\lambda}} w^{\mathrm{p}}_\pi w^{\mathrm{e}}_\nu\,,
        \end{equation}
        where
        $$C:=\sum_{\lambda\in \calM^\tau_{\mathrm{pol}\textrm{-}\mathrm{env}}} w_{\lambda}^\tau\,.$$
\end{definition}

We define the (normalized) probability prior $w^{\tau,\mathrm{pol}\textrm{-}\mathrm{env}}\in\Delta\calM^\tau_{\mathrm{pol}\textrm{-}\mathrm{env}}$ as follows:
$$w^{\tau,\mathrm{pol}\textrm{-}\mathrm{env}}_\lambda:=\frac{w^{\tau}_\lambda}{\sum_{\lambda'\in \calM_{\mathrm{pol}\textrm{-}\mathrm{env}}^{\tau}} w^{\tau}_{\lambda'}}\,,\quad\forall\lambda\in\calM^\tau_{\mathrm{pol}\textrm{-}\mathrm{env}}\,.$$

\begin{remark}
    \label{rem:decoupled-normalized-prior}
    It follows immediately from \cref{def:decoupled-w-tau} that $w^\tau$ is decoupled on $\calM^\tau_{\mathrm{pol}\textrm{-}\mathrm{env}}$ if and only if there exist some $w^{\mathrm{p}}\in\Delta\calM_{\mathrm{pol}}^\tau$ and $w^{\mathrm{e}}\in\Delta\calM_{\mathrm{env}}^\tau$ such that $w^{\tau,\mathrm{pol}\textrm{-}\mathrm{env}}=w^{\mathrm{p}}\otimes w^{\mathrm{e}}$.

    However, due to \cref{rem:diff-pol-env-same-univ}, if $w^\tau\in\Delta'\calM_{\mathrm{univ}}^\tau$ is decoupled on $\calM^\tau_{\mathrm{pol}\textrm{-}\mathrm{env}}$, then there can be multiple pairs of priors $(w^{\mathrm{p}}, w^{\mathrm{e}})\in\Delta\calM_{\mathrm{pol}}^\tau\times \Delta\calM_{\mathrm{env}}^\tau$ such that $w^{\tau,\mathrm{pol}\textrm{-}\mathrm{env}}_\lambda=w^{\mathrm{p}}\otimes w^{\mathrm{e}}$. In other words, a decoupled prior does not necessarily "factorize" into a unique pair of priors on policies and environments, respectively.
\end{remark}

Motivated by the above remark, we define a notion of a degree of functional similarity relative to priors $w^\tau\in\Delta'\calM_{\mathrm{univ}}^\tau$, $w^{\mathrm{p}}\in\Delta\calM_{\mathrm{pol}}^\tau$ and $w^{\mathrm{e}}\in\Delta\calM_{\mathrm{env}}^\tau$:

\begin{definition}
    \label{def:degree-of-functional-similarity}

    Let $w^\tau\in\Delta'\calM_{\mathrm{univ}}^\tau$ be a semiprobability prior on $\calM_{\mathrm{univ}}^\tau$, and let $w^{\mathrm{p}}\in\Delta\calM_{\mathrm{pol}}^\tau$ and $w^{\mathrm{e}}\in\Delta\calM_{\mathrm{env}}^\tau$ be two probability priors on the spaces of policies and environments, respectively.

    We define the degree of functional similarity in a universe $\lambda\in\calM_{\mathrm{pol}\textrm{-}\mathrm{env}}^\tau$ with respect to the priors $w^\tau$, $w^{\mathrm{p}}$ and $w^{\mathrm{e}}$ as
    $$S\left(\lambda,w^\tau,w^{\mathrm{p}},w^{\mathrm{e}}\right):=\log \frac{w_{\lambda}^{\tau,\mathrm{pol}\textrm{-}\mathrm{env}}}{\left(w^{\mathrm{p}}\otimes w^{\mathrm{e}}\right)_{\lambda}}\,.$$
    
    For a policy $\pi\in \calM_{\mathrm{pol}}^{\tau}$ and an environment $\nu\in \calM_{\mathrm{env}}^{\tau}$, we define the degree of functional similarity between $\pi$ and $\nu$ with respect to the priors $w^\tau$, $w^{\mathrm{p}}$ and $w^{\mathrm{e}}$ as
    $$S\left(\pi,\nu,w^\tau,w^{\mathrm{p}},w^{\mathrm{e}}\right):=S\left(\nu^\pi,w^\tau,w^{\mathrm{p}},w^{\mathrm{e}}\right)\,.$$

    The average degree of functional similarity in $\calM^\tau_{\mathrm{pol}\textrm{-}\mathrm{env}}$ with respect to the priors $w^\tau$, $w^{\mathrm{p}}$ and $w^{\mathrm{e}}$ is defined as
    $$S\left(\calM^\tau_{\mathrm{pol}\textrm{-}\mathrm{env}},w^\tau,w^{\mathrm{p}},w^{\mathrm{e}}\right):=\frac{\sum_{\lambda\in \calM^\tau_{\mathrm{pol}\textrm{-}\mathrm{env}}}w^\tau_\lambda S(\lambda,w^\tau,w^{\mathrm{p}},w^{\mathrm{e}})}{\sum_{\lambda\in \calM^\tau_{\mathrm{pol}\textrm{-}\mathrm{env}}}w^\tau_\lambda}=\sum_{\lambda\in \calM^\tau_{\mathrm{pol}\textrm{-}\mathrm{env}}}w^{\tau,\mathrm{pol}\textrm{-}\mathrm{env}}_\lambda S(\lambda,w^\tau,w^{\mathrm{p}},w^{\mathrm{e}})\,.$$
\end{definition}

One thing that is not very satisfying about the above definition is that it does not define the degree of functional similarity only relative to the prior $w^\tau\in\Delta'\calM_{\mathrm{univ}}^\tau$ but it also requires the choice of priors $w^{\mathrm{p}}\in\Delta\calM_{\mathrm{pol}}^\tau$ and $w^{\mathrm{e}}\in\Delta\calM_{\mathrm{env}}^\tau$. We will come back to this point later, but for now let us study the properties of the notion the degree of functional similarity of \cref{def:degree-of-functional-similarity} to provide further intuition about it:

\begin{proposition}
    \label{prop:average-functional-similarity-is-KL}
        The average degree of functional similarity in $\calM^\tau_{\mathrm{pol}\textrm{-}\mathrm{env}}$ with respect to the priors $w^\tau$, $w^{\mathrm{p}}$ and $w^{\mathrm{e}}$ is equal to the KL divergence between $w^{\tau,\mathrm{pol}\textrm{-}\mathrm{env}}$ and $w^{\mathrm{p}}\otimes w^{\mathrm{e}}$:
    $$S\left(\calM^\tau_{\mathrm{pol}\textrm{-}\mathrm{env}},w^\tau,w^{\mathrm{p}},w^{\mathrm{e}}\right) = KL\left(w^{\tau,\mathrm{pol}\textrm{-}\mathrm{env}},w^{\mathrm{p}}\otimes w^{\mathrm{e}}\right)\,.$$
\end{proposition}
\begin{proof}
    The proof is a straightforward implication of the definitions:
    \begin{align*}
    S\left(\calM^\tau_{\mathrm{pol}\textrm{-}\mathrm{env}},w^\tau,w^{\mathrm{p}},w^{\mathrm{e}}\right)&=\sum_{\lambda\in \calM^\tau_{\mathrm{pol}\textrm{-}\mathrm{env}}}w^{\tau,\mathrm{pol}\textrm{-}\mathrm{env}}_\lambda S(\lambda,w^\tau,w^{\mathrm{p}},w^{\mathrm{e}}) \\
    &= \sum_{\lambda\in \calM^\tau_{\mathrm{pol}\textrm{-}\mathrm{env}}}w^{\tau,\mathrm{pol}\textrm{-}\mathrm{env}}_\lambda  \log \frac{w_{\lambda}^{\tau,\mathrm{pol}\textrm{-}\mathrm{env}}}{\left(w^{\mathrm{p}}\otimes w^{\mathrm{e}}\right)_\lambda}=KL\left(w^{\tau,\mathrm{pol}\textrm{-}\mathrm{env}},w^{\mathrm{p}}\otimes w^{\mathrm{e}}\right)\,.
    \end{align*}
\end{proof}

\begin{corollary}
    \label{cor:zero-functional-similarity}
        For every  $w^{\mathrm{p}}\in\Delta\calM_{\mathrm{pol}}^\tau$ and $w^{\mathrm{e}}\in\Delta\calM_{\mathrm{env}}^\tau$, the following are equivalent:
    \begin{enumerate}
        \item[(a)] For every $\pi\in\calM^{\tau}_{\mathrm{pol}}$ and every $\nu\in\calM^{\tau}_{\mathrm{env}}$, we have $S(\nu,\pi,w^\tau,w^{\mathrm{p}},w^{\mathrm{e}})=0$.
        \item[(b)] The  average degree of functional similarity in $\calM^\tau_{\mathrm{pol}\textrm{-}\mathrm{env}}$ with respect to $w^{\mathrm{p}}$ and $w^{\mathrm{e}}$ is equal to zero:
        $$S\left(\calM^\tau_{\mathrm{pol}\textrm{-}\mathrm{env}},w^\tau,w^{\mathrm{p}},w^{\mathrm{e}}\right)=0\,.$$
    \end{enumerate}
\end{corollary}
\begin{proof}
    In the light of \cref{prop:average-functional-similarity-is-KL}, we have
    \begin{align*}
    S\left(\calM^\tau_{\mathrm{pol}\textrm{-}\mathrm{env}},w^\tau,w^{\mathrm{p}},w^{\mathrm{e}}\right)=0\quad
    &\Leftrightarrow \quad KL\left(w^{\tau,\mathrm{pol}\textrm{-}\mathrm{env}},w^{\mathrm{p}}\otimes w^{\mathrm{e}}\right)=0\\
    &\Leftrightarrow \quad w^{\tau,\mathrm{pol}\textrm{-}\mathrm{env}} = w^{\mathrm{p}}\otimes w^{\mathrm{e}}\\
    &\Leftrightarrow \quad w^{\tau,\mathrm{pol}\textrm{-}\mathrm{env}}_\lambda=\left(w^{\mathrm{p}}\otimes w^{\mathrm{e}}\right)_\lambda\,,\quad\forall\lambda\in \calM^\tau_{\mathrm{pol}\textrm{-}\mathrm{env}}\\
    &\Leftrightarrow \quad S\left(\pi,\nu,w^\tau,w^{\mathrm{p}},w^{\mathrm{e}}\right)=0 \,,\quad\forall \pi\in\calM^{\tau}_{\mathrm{pol}}\,,\forall \nu\in\calM^{\tau}_{\mathrm{env}}\,.
    \end{align*}
\end{proof}

\begin{corollary}
    \label{cor:zero-functional-similarity-decoupled}
        A prior $w^\tau\in\Delta'\calM_{\mathrm{univ}}^\tau$ is decoupled on $\calM^\tau_{\mathrm{pol}\textrm{-}\mathrm{env}}$ if and only if there exist some $w^{\mathrm{p}}\in\Delta\calM_{\mathrm{pol}}^\tau$ and $w^{\mathrm{e}}\in\Delta\calM_{\mathrm{env}}^\tau$ such that
        $$S\left(\calM^\tau_{\mathrm{pol}\textrm{-}\mathrm{env}},w^\tau,w^{\mathrm{p}},w^{\mathrm{e}}\right)=0\,.$$
\end{corollary}
\begin{proof}
    This follows immediately from \cref{cor:zero-functional-similarity} and \cref{def:decoupled-w-tau}.
\end{proof}

Our ultimate goal in this section is to show the "coupledness" of the Solomonoff prior. In order to simplify the analysis and overcome the technicalities arising from the non-unique "factorization" of a decoupled prior (as described in \cref{rem:diff-pol-env-same-univ,rem:decoupled-normalized-prior}), we will focus on studying the "coupledness and functional similarity aspects" on the sets of fully-supported policies and environments, which we define as follows (see also \cref{def:fully-supported-universe-measure}):

\begin{definition}
    \label{def:fully-supported-env-pol}
    We say that a policy $\pi\in \calM_{\mathrm{pol}}^{\tau}$ is fully supported if $\pi(a|\HistA )>0$ for all $(\HistA, a)\in(\TurnSet^*)\times\mathcal{A}$, and $\sum_{a\in \mathcal{A}}\pi(a|\HistA )=1$ for all $\HistA \in \TurnSet^*$. We denote the set of fully supported policies as $\check{\calM}_{\mathrm{pol}}^{\tau}$.

    We say that an environment $\nu\in \calM_{\mathrm{env}}^{\tau}$ is fully supported if $\nu(e \mid \HistA, a)>0$ for all $(\HistA, a, e)\in(\TurnSet^*)\times\mathcal{A}\times\calE$, and $\sum_{e \in \calE}\nu(e \mid \HistA, a)=1$ for all $(\HistA, a)\in(\TurnSet^*)\times\mathcal{A}$. We denote the set of fully supported environments as $\check{\calM}_{\mathrm{env}}^{\tau}$.

    We say that a universe $\lambda\in\calM_{\mathrm{univ}}^\tau$ is fully supported if $\lambda(\HistA )>0$ for all $\HistA\in\TurnSet^*$, and $\lambda$ is a measure.\footnote{I.e., $\sum_{\Hist\in \TurnSet^t}\lambda(\Hist )=1$ for all $t>0$.} We denote the set of fully supported universes as $\check{\calM}_{\mathrm{univ}}^\tau$.

    We let
    $$\check{\calM}^{\tau}_{\mathrm{pol}\textrm{-}\mathrm{env}}:=\left\{\nu^\pi:\nu\in \check{\calM}_{\mathrm{env}}^{\tau}, \pi \in \check{\calM}_{\mathrm{pol}}^{\tau}\right\}\subseteq \check{\calM}^{\tau}_{\mathrm{univ}}$$
    be the set of universes which are obtained by letting fully-supported policies and fully-supported environments interact.\footnote{It is worth noting that one can show that $\check{\calM}^{\tau}_{\mathrm{pol}\textrm{-}\mathrm{env}}$ and $\check{\calM}^{\tau}_{\mathrm{univ}}$ are in fact equal. In order to see why we also have $\check{\calM}^{\tau}_{\mathrm{univ}}\subseteq \check{\calM}^{\tau}_{\mathrm{pol}\textrm{-}\mathrm{env}}$, let $\Mvu\in \check{\calM}^{\tau}_{\mathrm{univ}}$. Since $\Mvu$ is a fully-spported measure, there exist a unique policy $\Mvp:\TurnSet^*\to\Delta\calA$ and a unique environment $\Mve:\TurnSet^*\times\calA\to\Delta\calE$ such that $\Mvu=\Mve^\Mvp$, so we need to show that $\Mvp\in\check{\calM}_{\mathrm{pol}}^{\tau}$ and $\Mve\in\check{\calM}_{\mathrm{env}}^{\tau}$. Since $\Mvu\in \check{\calM}^{\tau}_{\mathrm{univ}}$, there exists $M\in\calMapom$ such that $\Mvu=\Mvu_M^\tau$. We can use $M$ to construct an implementation of $\Mvp$ (resp. $\Mve$) as a $\tau$-POM through rejection sampling (refer to the proof of \cref{thm:mupi-structural-similarities} below for further details). Obviously  $\Mvp$ and $\Mve$ are fully supported and hence $\Mvp\in\check{\calM}_{\mathrm{pol}}^{\tau}$ and $\Mve\in\check{\calM}_{\mathrm{env}}^{\tau}$, which implies that $\Mvu=\Mve^\Mvp\in\check{\calM}^{\tau}_{\mathrm{pol}\textrm{-}\mathrm{env}}$.}
    
\end{definition}

The following two (almost trivial) propositions describe the structure of an interaction between a fully-supported policy with a fully-supported environment:

\begin{proposition}
    \label{prop:fully-supp-pol-env}
    For every $\pi\in{\calM}_{\mathrm{pol}}^{\tau}$ and every $\nu\in{\calM}_{\mathrm{env}}^{\tau}$, we have $\nu^\pi\in \check{\calM}_{\mathrm{pol}\textrm{-}\mathrm{env}}^{\tau}$ if and only if $\pi\in\check{\calM}_{\mathrm{pol}}^{\tau}$ and $\nu\in\check{\calM}_{\mathrm{env}}^{\tau}$.
\end{proposition}
\begin{proof}
    The proof follows immediately from the definitions and by observing that for $\HistA \in \TurnSet^*$ 
    we have
    $\nu^\pi(\HistA )>0$ if and only if $\pi(a_i|\HistM[i])>0$ and $\nu(e_i|\HistM[i],a_i)>0$ for all $1\leq i\leq l(\HistA)$.
\end{proof}

\begin{proposition}
    \label{prop:fully-supp-pol-env-give-uniq-univ}
    For every $\pi,\tilde{\pi}\in\check{\calM}_{\mathrm{pol}}^{\tau}$ and $\nu,\tilde{\nu}\in\check{\calM}_{\mathrm{env}}^{\tau}$, we have $\nu^\pi=\tilde{\nu}^{\tilde{\pi}}$ if and only if $\nu=\tilde{\nu}$ and $\pi=\tilde{\pi}$.
\end{proposition}
\begin{proof}
    The proof is an immediate consequence of the fact that for every $(\HistA, a)\in(\TurnSet^*)\times\mathcal{A}$, we have
    $$\pi(a|\HistA )=\frac{\sum_{e \in \calE}\nu^\pi(\HistA a e)}{\nu^\pi(\HistA )}\,,$$
    and for every $(\HistA, a, e)\in(\TurnSet^*)\times\mathcal{A}\times\calE$, we have
    $$\nu(e \mid \HistA, a)=\frac{\nu^\pi(\HistA a e)}{\sum_{e'\in\calE}\nu^\pi(\HistA a e')}\,.$$
\end{proof}

An immediate corollary of the above two propositions is a simplification of \eqref{eq:ind-priors-on-env-and-pol} in the case of fully supported policies and environments:

\begin{corollary}
    \label{cor:fully-supp-pol-env-priors-uniq-factorization}
    For every priors $w^{\mathrm{p}}\in\Delta{\calM}_{\mathrm{pol}}^{\tau}$ and $w^{\mathrm{e}}\in\Delta{\calM}_{\mathrm{env}}^{\tau}$, if $\pi\in\check{\calM}_{\mathrm{pol}}^{\tau}$ and $\nu\in\check{\calM}_{\mathrm{env}}^{\tau}$ then
    $$\left(w^{\mathrm{p}}\otimes w^{\mathrm{e}}\right)_{\nu^\pi}=w^{\mathrm{p}}_\pi w^{\mathrm{e}}_\nu\,.$$

    In particular, for every $w^{\mathrm{p}},\tilde{w}^{\mathrm{p}}\in\Delta\check{\calM}_{\mathrm{pol}}^{\tau}$ and $w^{\mathrm{e}},\tilde{w}^{\mathrm{e}}\in\Delta\check{\calM}_{\mathrm{env}}^{\tau}$, we have $w^{\mathrm{p}}\otimes w^{\mathrm{e}}=\tilde{w}^{\mathrm{p}}\otimes \tilde{w}^{\mathrm{e}}$ if and only if $w^{\mathrm{p}}=\tilde{w}^{\mathrm{p}}$ and $w^{\mathrm{e}}=\tilde{w}^{\mathrm{e}}$.
\end{corollary}
\begin{proof}
    The proof follows immediately from \cref{def:ind-priors-on-env-and-pol} and \cref{prop:fully-supp-pol-env,prop:fully-supp-pol-env-give-uniq-univ}.
\end{proof}

Motivated by the above results, we introduce the notion of "decoupledness on $\check{\calM}^\tau_{\mathrm{pol}\textrm{-}\mathrm{env}}$" (which is the analog of \cref{def:decoupled-w-tau}) as follows:

\begin{definition}
    \label{def:decoupled-w-tau-fully-supp}
    We say that the prior $w^\tau\in\Delta'\calM_{\mathrm{univ}}^\tau$ is decoupled on $\check{\calM}^\tau_{\mathrm{pol}\textrm{-}\mathrm{env}}$, if the restriction of
    $w^\tau$ to $\check{\calM}^\tau_{\mathrm{pol}\textrm{-}\mathrm{env}}$ can be seen as the result of interaction between a policy and an environment which are chosen independently, i.e., there exist some $w^{\mathrm{p}}\in\Delta\check{\calM}_{\mathrm{pol}}^\tau$ and $w^{\mathrm{e}}\in\Delta\check{\calM}_{\mathrm{env}}^\tau$ such that for all $\pi\in \check{\calM}_{\mathrm{pol}}^\tau$ and all $\nu\in \check{\calM}_{\mathrm{env}}^\tau$, we have
        \begin{equation}
            \label{eq:decoupled-w-tau-fully-supp}
            w_{\nu^\pi}^\tau= C\left(w^{\mathrm{p}}\otimes w^{\mathrm{e}}\right)_{\nu^\pi}=C w^{\mathrm{p}}_\pi w^{\mathrm{e}}_\nu\,,
        \end{equation}
        where
        $$C:=\sum_{\lambda\in \check{\calM}^\tau_{\mathrm{pol}\textrm{-}\mathrm{env}}} w_{\lambda}^\tau\,.$$
\end{definition}

We define the (normalized) probability prior $\check{w}^{\tau,\mathrm{pol}\textrm{-}\mathrm{env}}\in\Delta\check{\calM}^\tau_{\mathrm{pol}\textrm{-}\mathrm{env}}$ as follows:
\begin{equation}
\label{eq:prior-on-fully-supp}
    \check{w}^{\tau,\mathrm{pol}\textrm{-}\mathrm{env}}_\lambda:=\frac{w^{\tau}_\lambda}{\sum_{\lambda'\in \check{\calM}_{\mathrm{pol}\textrm{-}\mathrm{env}}^{\tau}} w^{\tau}_{\lambda'}}\,,\quad\forall\lambda\in\check{\calM}^\tau_{\mathrm{pol}\textrm{-}\mathrm{env}}\,.
\end{equation}

\begin{remark}
    \label{rem:decoupled-normalized-prior-fully-supp}
    It follows immediately from \cref{def:decoupled-w-tau-fully-supp} that $w^\tau$ is decoupled on $\check{\calM}^\tau_{\mathrm{pol}\textrm{-}\mathrm{env}}$ if and only if there exist some $w^{\mathrm{p}}\in\Delta\check{\calM}_{\mathrm{pol}}^\tau$ and $w^{\mathrm{e}}\in\Delta\check{\calM}_{\mathrm{env}}^\tau$ such that $\check{w}^{\tau,\mathrm{pol}\textrm{-}\mathrm{env}}_\lambda=w^{\mathrm{p}}\otimes w^{\mathrm{e}}$.
\end{remark}

An immediate corollary of the definitions is that coupledness on $\check{\calM}^\tau_{\mathrm{pol}\textrm{-}\mathrm{env}}$ implies coupledness on ${\calM}^\tau_{\mathrm{pol}\textrm{-}\mathrm{env}}$:

\begin{proposition}
    \label{prop:coupledness-on-fully-supp-suffices}
    If $w^\tau$ is decoupled on ${\calM}^\tau_{\mathrm{pol}\textrm{-}\mathrm{env}}$, then it is decoupled on $\check{\calM}^\tau_{\mathrm{pol}\textrm{-}\mathrm{env}}$ as well.

    In particular, if our goal is to show that the Solomonoff prior is always coupled ${\calM}^\tau_{\mathrm{pol}\textrm{-}\mathrm{env}}$, it suffices to show that it is coupled on $\check{\calM}^\tau_{\mathrm{pol}\textrm{-}\mathrm{env}}$.
\end{proposition}

Now since decoupled priors on $\calMpolenvcheck^\tau$ "factorize uniquely" (as implied by \cref{cor:fully-supp-pol-env-priors-uniq-factorization}), we can explicitly characterize the factors:

\begin{definition}
For every policy $\pi\in \check{\calM}_{\mathrm{pol}}^{\tau}$ and every environment $\nu\in \check{\calM}_{\mathrm{env}}^{\tau}$, define

\begin{equation}
    \label{eq:w-pol-and-w-env-fully-supp}
    \check{w}^{\tau,\mathrm{env}}_\nu :=\sum_{\pi\in \check{\calM}^{\tau}_{\mathrm{pol}}}\check{w}_{\nu^\pi}^{\tau,\mathrm{pol}\textrm{-}\mathrm{env}}
\quad\text{and}\quad
\check{w}^{\tau,\mathrm{pol}}_\pi :=\sum_{\nu\in \check{\calM}^{\tau}_{\mathrm{env}}}\check{w}_{\nu^\pi}^{\tau,\mathrm{pol}\textrm{-}\mathrm{env}}\,,
\end{equation}
\end{definition}

\begin{proposition}
    \label{prop:factors-of-decoupled-fully-supp}
    $w^\tau$ is decoupled on $\check{\calM}^\tau_{\mathrm{pol}\textrm{-}\mathrm{env}}$ if and only if $\check{w}^{\tau,\mathrm{pol}\textrm{-}\mathrm{env}}=\check{w}^{\tau,\mathrm{pol}}\otimes \check{w}^{\tau,\mathrm{env}}$.
\end{proposition}
\begin{proof}
    If $\check{w}^{\tau,\mathrm{pol}\textrm{-}\mathrm{env}}=\check{w}^{\tau,\mathrm{pol}}\otimes \check{w}^{\tau,\mathrm{env}}$ then $w^\tau$ is trivially decoupled on $\check{\calM}^\tau_{\mathrm{pol}\textrm{-}\mathrm{env}}$ due to \cref{rem:decoupled-normalized-prior-fully-supp}.

    Conversely, if $w^\tau$ is decoupled on $\check{\calM}^\tau_{\mathrm{pol}\textrm{-}\mathrm{env}}$, then \cref{rem:decoupled-normalized-prior-fully-supp} implies that there exist some $w^{\mathrm{p}}\in\Delta\check{\calM}_{\mathrm{pol}}^\tau$ and $w^{\mathrm{e}}\in\Delta\check{\calM}_{\mathrm{env}}^\tau$ such that $\check{w}^{\tau,\mathrm{pol}\textrm{-}\mathrm{env}}=w^{\mathrm{p}}\otimes w^{\mathrm{e}}$. Then, for all $\nu\in \check{\calM}_{\mathrm{env}}^\tau$, we have

    \begin{align*}
        \check{w}^{\tau,\mathrm{env}}_\nu &=\sum_{\pi\in \check{\calM}^{\tau}_{\mathrm{pol}}}\check{w}_{\nu^\pi}^{\tau,\mathrm{pol}\textrm{-}\mathrm{env}} =\sum_{\pi\in \check{\calM}^{\tau}_{\mathrm{pol}}}\left(w^{\mathrm{p}}\otimes w^{\mathrm{e}}\right)_{\nu^\pi}\stackrel{(\ast)}{=} \sum_{\pi\in \check{\calM}^{\tau}_{\mathrm{pol}}}w^{\mathrm{p}}_\pi w^{\mathrm{e}}_\nu=w^{\mathrm{e}}_\nu\,,
    \end{align*}
    where $(\ast)$ follows from \cref{cor:fully-supp-pol-env-priors-uniq-factorization}. We deduce that $\check{w}^{\tau,\mathrm{env}}=w^{\mathrm{e}}$. A similar argument shows that  $\check{w}^{\tau,\mathrm{pol}}=w^{\mathrm{p}}$. Therefore, $\check{w}^{\tau,\mathrm{pol}\textrm{-}\mathrm{env}}_\lambda=\check{w}^{\tau,\mathrm{pol}}\otimes \check{w}^{\tau,\mathrm{env}}$.
\end{proof}

Motivated by the above proposition, we define a notion of a degree of functional similarity for fully supported universes relative to a prior $w^\tau\in\Delta'\calM_{\mathrm{univ}}^\tau$:

\begin{definition}
    \label{def:degree-of-functional-similarity-fully-supp}
    Let $w^\tau\in\Delta'\calM_{\mathrm{univ}}^\tau$ be a semiprobability prior on $\calM_{\mathrm{univ}}^\tau$. We define the degree of functional similarity in a fully-supported universe $\lambda\in\check{\calM}_{\mathrm{pol}\textrm{-}\mathrm{env}}^\tau$ with respect to the prior\footnote{Notice here that we only specify a prior $w^\tau\in\Delta'\calM_{\mathrm{univ}}^\tau$ relative to which we define the degree of functional similarity. This is in contrast with \cref{def:degree-of-functional-similarity} where we also specify priors $w^{\mathrm{p}}$ and $w^{\mathrm{e}}$.} $w^\tau$ as
    $$\check{S}\left(\lambda,w^\tau\right):=S\left(\lambda,\check{w}^{\tau,\mathrm{pol}\textrm{-}\mathrm{env}},\check{w}^{\tau,\mathrm{pol}},\check{w}^{\tau,\mathrm{env}}\right)=\log \frac{\check{w}_{\lambda}^{\tau,\mathrm{pol}\textrm{-}\mathrm{env}}}{\left(\check{w}^{\tau,\mathrm{pol}}\otimes \check{w}^{\tau,\mathrm{env}}\right)_{\lambda}}\,.$$
    
    For a policy $\pi\in \check{\calM}_{\mathrm{pol}}^{\tau}$ and an environment $\nu\in \check{\calM}_{\mathrm{env}}^{\tau}$, we define the degree of functional similarity between the fully-supported $\pi$ and $\nu$ with respect to the prior $w^\tau$ as
    $$\check{S}\left(\pi,\nu,w^\tau\right):=\check{S}\left(\nu^\pi,w^\tau\right)=\log \frac{\check{w}_{\nu^\pi}^{\tau,\mathrm{pol}\textrm{-}\mathrm{env}}}{\check{w}^{\tau,\mathrm{pol}}_\pi \check{w}^{\tau,\mathrm{env}}_{\nu}}\,.$$

    The average degree of functional similarity in $\check{\calM}^\tau_{\mathrm{pol}\textrm{-}\mathrm{env}}$ with respect to the prior $w^\tau$ is defined as

    \begin{align*}
    \check{S}\left(\check{\calM}^\tau_{\mathrm{pol}\textrm{-}\mathrm{env}},w^\tau\right)&:=\frac{\sum_{\lambda\in \check{\calM}^\tau_{\mathrm{pol}\textrm{-}\mathrm{env}}}w^\tau_\lambda \check{S}(\lambda,w^\tau)}{\sum_{\lambda\in \check{\calM}^\tau_{\mathrm{pol}\textrm{-}\mathrm{env}}}w^\tau_\lambda}=\sum_{\lambda\in \check{\calM}^\tau_{\mathrm{pol}\textrm{-}\mathrm{env}}}\check{w}^{\tau,\mathrm{pol}\textrm{-}\mathrm{env}}_\lambda \check{S}(\lambda,w^\tau)\\
    &\;\;=\sum_{\lambda\in \check{\calM}^\tau_{\mathrm{pol}\textrm{-}\mathrm{env}}}\check{w}^{\tau,\mathrm{pol}\textrm{-}\mathrm{env}}_\lambda \log \frac{\check{w}_{\lambda}^{\tau,\mathrm{pol}\textrm{-}\mathrm{env}}}{\left(\check{w}^{\tau,\mathrm{pol}}\otimes \check{w}^{\tau,\mathrm{env}}\right)_{\lambda}}=KL\left(\check{w}^{\tau,\mathrm{pol}\textrm{-}\mathrm{env}},\check{w}^{\tau,\mathrm{pol}}\otimes \check{w}^{\tau,\mathrm{env}}\right)\,.
    \end{align*}

\end{definition}

\begin{proposition}
    \label{prop:decoupled-fully-supp-degree-functional-similarity}
    Let $w^\tau\in\Delta'\calM_{\mathrm{univ}}^\tau$ be a semiprobability prior on $\calM_{\mathrm{univ}}^\tau$. The following statements are equivalent:
    \begin{itemize}
        \item[(a)] The prior $w^\tau\in\Delta'\calM_{\mathrm{univ}}^\tau$ is decoupled on $\check{\calM}^\tau_{\mathrm{pol}\textrm{-}\mathrm{env}}$.
        \item[(b)] For every $\pi\in\check{\calM}^{\tau}_{\mathrm{pol}}$ and every $\nu\in\check{\calM}^{\tau}_{\mathrm{env}}$, we have $\check{S}(\nu,\pi,w^\tau)=0$.
        \item[(c)] The average degree of functional similarity in $\check{\calM}^\tau_{\mathrm{pol}\textrm{-}\mathrm{env}}$ with respect to $w^{\tau}$ is equal to zero:        $$\check{S}\left(\check{\calM}^\tau_{\mathrm{pol}\textrm{-}\mathrm{env}},w^\tau\right)=0\,.$$
    \end{itemize}
\end{proposition}
\begin{proof}
    The equivalence between (a) and (b) follows from \cref{prop:factors-of-decoupled-fully-supp} and the definition of $\check{S}(\nu,\pi,w^\tau)$.
    
    The same proof for \cref{cor:zero-functional-similarity-decoupled} can be used to show the equivalence between (b) and (c).
\end{proof}

We now introduce the concept of bounded coupledness and unbounded coupledness on $\calMpolenvcheck^\tau$.

\begin{definition}
\label{def:unbounded-coupledness} 
Let $w^\tau\in\Delta'\calM_{\mathrm{univ}}^\tau$ be a semiprobability prior on $\calM_{\mathrm{univ}}^\tau$. We call $w^{\tau}$ \textit{boundedly coupled on $\calMpolenvcheck^\tau$} iff there exists absolute positive constants $C_1<C_2$ such that for all $\Mvu=\Mve^\Mvp \in \calMpolenvcheck^\tau$ we have
$$C_1\check{w}^{\tau,\mathrm{pol}}(\pi)\check{w}^{\tau,\mathrm{env}}(\Mve)\leq \check{w}^{\tau,\mathrm{pol}\textrm{-}\mathrm{env}}(\Mve^\Mvp) \leq C_2 \check{w}^{\tau,\mathrm{pol}}(\pi)\check{w}^{\tau,\mathrm{env}}(\Mve)\,.$$
From \cref{prop:factors-of-decoupled-fully-supp} we can see that a decoupled prior on $\calMpolenvcheck^\tau$ is also boundedly coupled on $\calMpolenvcheck^\tau$.

We call $w^{\tau}$ unboundedly coupled on $\calMpolenvcheck^\tau$ iff it is not boundedly coupled on $\calMpolenvcheck^\tau$.
\end{definition}

\begin{lemma}
    \label{lem:bounded-coupledness} $w^{\tau}$ is boundedly coupled on $\calMpolenvcheck^\tau$ iff there exist two absolute real numbers $s_1<s_2$ such that for every $\pi\in\check{\calM}^{\tau}_{\mathrm{pol}}$ and every $\nu\in\check{\calM}^{\tau}_{\mathrm{env}}$, we have $s_1<\check{S}(\nu,\pi,w^\tau)<s_2$.
\end{lemma}
\begin{proof}
    This follows immediately from \cref{def:degree-of-functional-similarity-fully-supp,def:unbounded-coupledness}.
\end{proof}

Now we are ready to study the functional similarity and the coupledness of the Solomonoff prior.

Let $\calMapom$ be the set of abstract probabilistic oracle machines and for every $M\in\calMapom$, let $\langle M\rangle$ be the binary encoding of $M$ according to some canonical encoding.

Let $U$ be a universal monotone Turing machine and let $w_M^U=2^{-K_U(\langle M\rangle)}$ be the Solomonoff prior over abstract probabilistic oracle machines with respect to $U$. Let $w^{U,\tau}\in\Delta'\calM^\tau_{\mathrm{univ}}$ be the induced prior on the space $\calM^\tau_{\mathrm{univ}}$ of universes that are implementable as POMs with access to the probabilistic oracle $\tau$, i.e.,
$$w_\lambda^{U,\tau}:=\sum_{\substack{M\in\calMapom:\\\lambda_M^\tau=\lambda}}w^U_M=\sum_{\substack{M\in\calMapom:\\\lambda_M^\tau=\lambda}}2^{-K_U(\langle M\rangle)}\,.$$

The next theorem shows that the restriction of the Solomonoff prior on $\calM^{\tau}_{\mathrm{univ}}$ is always (unboundedly) coupled:

\begin{theorem}
    \label{thm:mupi-structural-similarities}
    For every fixed choices of:
    \begin{itemize}
        \item a canonical encoding of $\langle M\rangle$ of APOMs as binary strings,
        \item a universal monotone Turing machine $U$,
        \item a probabilistic oracle $\tau$ (which may or may not be a RUI-oracle),
        \item finite spaces $\calA$ and $\calE$, with $|\mathcal{A}|\geq 2$ and $|\calE|\geq 2$, and complete prefix encodings thereof,
    \end{itemize}
    we have:
    \begin{enumerate}
        \item[(a)] The Solomonoff prior $w^{U,\tau}$ is always coupled on ${\calM}^\tau_{\mathrm{pol}\textrm{-}\mathrm{env}}$, i.e., for every $w^{\mathrm{p}}\in\Delta\calM^{\tau}_{\mathrm{pol}}$ and every $w^{\mathrm{e}}\in\Delta\calM^{\tau}_{\mathrm{env}}$, there are always a policy $\pi\in \calM^{\tau}_{\mathrm{pol}}$ and an environment $\nu\in \calM^{\tau}_{\mathrm{env}}$ such that
    $$w^{U,\tau}_{\nu^\pi}\neq C\cdot\left( w^{\mathrm{e}} \otimes w^{\mathrm{p}}\right)_{\nu^\pi}\,,$$
    where
    $$C:=\sum_{\lambda\in \calM^\tau_{\mathrm{pol}\textrm{-}\mathrm{env}}} w_{\lambda}^{\tau ,U}\,.$$
        \item[(b)] The Solomonoff prior $w^{U,\tau}$ is always unboundedly coupled on $\calMpolenvcheck^\tau$.
        \item[(c)] There are fully-supported policies and environments sharing arbitrarily large degrees of functional similarity with respect to $w^\tau$, i.e., for every $s>0$, there is at least some $\pi\in \check{\calM}^{\tau}_{\mathrm{pol}}$ and some $\nu\in \check{\calM}^{\tau}_{\mathrm{env}}$ such that
        $$\check{S}\left(\pi,\nu,w^{\tau ,U}\right)>s\,.$$
    \end{enumerate}
\end{theorem}
\begin{proof}

From \cref{lem:bounded-coupledness} we can see that $(c)$ implies $(b)$.

Now assume that $(b)$ is correct, i.e., $w^{U,\tau}$ is unboundedly coupled on $\calMpolenvcheck^\tau$. It follows from \cref{def:unbounded-coupledness} and \cref{prop:factors-of-decoupled-fully-supp} that $w^{U,\tau}$ is coupled on $\calMpolenvcheck^\tau$ and hence from
\cref{prop:coupledness-on-fully-supp-suffices}, we can see that $w^{U,\tau}$ is also coupled on ${\calM}^\tau_{\mathrm{pol}\textrm{-}\mathrm{env}}$. This means that $(b)$ implies $(a)$, and hence $(c)$ implies both $(a)$ and $(b)$. Therefore, it suffices to prove $(c)$.

For simplicity, we will write $w^{\tau}$ to denote the Solomonoff prior $w^{U,\tau}$ that is induced by $U$.

Let us fix some subsets $\tilde{\mathcal{A}}\subseteq\mathcal{A}$ and $\tilde{\calE}\subseteq\calE$ such that $|\tilde{\mathcal{A}}|=|\tilde{\calE}|=2$ and let $\tilde{a}_0,\tilde{a}_1$ (resp. $\tilde{e}_0,\tilde{e}_1$) be the elements of $\tilde{\mathcal{A}}$ (resp. $\tilde{\calE}$).

Let us further fix some computable probability distribution $p_{\tilde{\mathcal{A}}^c}$ on $\mathcal{A}\setminus\tilde{\mathcal{A}}$, and some computable probability distribution $p_{\tilde{\calE}^c}$ on $\calE\setminus\tilde{\calE}$.

Let us focus our attention on the family $\calM_{\mathrm{sim}}^\tau$ of fully-supported universes $\lambda=\nu^\pi\in \check{\calM}^\tau_{\mathrm{pol}\textrm{-}\mathrm{env}}$ which share the following structure:
\begin{itemize}
    \item Percepts and actions at time $t$ are conditionally independent given history $\HistM $, i.e., for every $(\HistM ,a_t,e_t)\in(\TurnSet^*)\times\mathcal{A}\times\calE$, the conditional probability $\lambda(a_te_t|\HistM )$ factorizes as
    $$\lambda(a_te_t|\HistM )=\pi(a_t|\HistM )\nu(e_t|\HistM a_t)\,,$$
    with $\nu(e_t|\HistM a_t)$ depending only on $\HistM $. We will slightly abuse notation and write $\nu(e_t|\HistM )$ instead of $\nu(e_t|\HistM a_t)$ to emphasize that it depends only on $\HistM $. With this notation, we can write
    $$\lambda(a_te_t|\HistM )=\pi(a_t|\HistM )\nu(e_t|\HistM )\,.$$
    \item For every $(\HistM ,a_t)\in(\TurnSet^*)\times\mathcal{A}$, we have
    \begin{itemize}
        \item $\pi(a_t|\HistM )\in\{1/4,1/2\}$ if $a_t\in\tilde{\mathcal{A}}$.
        \item $\pi(a_t|\HistM )=\frac{1}{4}p_{\tilde{\mathcal{A}}^c}(a_t)$ if $a_t\notin\tilde{\mathcal{A}}$.
    \end{itemize}
    \item For every $(\HistM ,e_t)\in\TurnSet^*\times\calE$, we have
    \begin{itemize}
        \item $\nu(e_t|\HistM )\in\{1/4,1/2\}$ if $e_t\in\tilde{\calE}$.
        \item $\nu(e_t|\HistM )=\frac{1}{4}p_{\tilde{\calE}^c}(e_t)$ if $e_t\notin\tilde{\calE}$.
    \end{itemize}
    \item For every $\HistA \in \TurnSet^*$, we have $\pi(\tilde{a}_0|\HistA )=\nu(\tilde{e}_0|\HistA )$ and $\pi(\tilde{a}_1|\HistA )=\nu(\tilde{e}_1|\HistA )$.
\end{itemize}
It is worth noting that $\calM_{\mathrm{sim}}^\tau$ is countably infinite: For example, there are infinitely many computable functions $F:\TurnSet^*\to\{0,1\}$, and each $F$ yields a different $\lambda_F=\nu_F^{\pi_F}\in\calM_{\mathrm{sim}}^\tau$ by letting
\begin{align*}
    \pi_F(\tilde{a}_0|\HistM )&=\frac{1}{4}+\frac{1}{4}F(\HistM )\,,\\
\pi_F(\tilde{a}_1|\HistM )&=\frac{3}{4}-\pi_F(\tilde{a}_0|\HistM )\,,\\
\nu_F(\tilde{e}_0|\HistM )&=\frac{1}{4}+\frac{1}{4}F(\HistM )\,,\\
\nu_F(\tilde{e}_1|\HistM )&=\frac{3}{4}-\nu_F(\tilde{e}_0|\HistM )\,.
\end{align*}

We claim that it is sufficient for our purposes (i.e., proving the claim $(c)$ of the theorem) to show that there are absolute constants $\alpha_{\mathrm{p}},\alpha_{\mathrm{e}}>0$ such that for every $\lambda=\nu^\pi\in\calM_{\mathrm{sim}}^\tau$, we have
\begin{equation}
    \label{eq:sim-univ-pol-sandwich}
    \alpha_{\mathrm{p}} \check{w}^{\tau,\mathrm{pol}}_{\pi} \leq \check{w}^{\tau,\mathrm{pol}\textrm{-}\mathrm{env}}_{\nu^\pi}\,,
\end{equation}
and
\begin{equation}
    \label{eq:sim-univ-env-sandwich}
    \alpha_{\mathrm{e}} \check{w}^{\tau,\mathrm{env}}_{\nu} \leq \check{w}^{\tau,\mathrm{pol}\textrm{-}\mathrm{env}}_{\nu^\pi}\,.
\end{equation}

Indeed, if the above two equations hold, then
$$\check{S}\left(\pi,\nu,w^{\tau ,U}\right)=\log \frac{\check{w}^{\tau,\mathrm{pol}\textrm{-}\mathrm{env}}_{\nu^\pi}}{\check{w}^{\tau,\mathrm{pol}}_{\pi}\check{w}^{\tau,\mathrm{env}}_{\nu}}\geq \log \frac{\alpha_{\mathrm{p}}\alpha_{\mathrm{e}}\check{w}^{\tau,\mathrm{pol}\textrm{-}\mathrm{env}}_{\nu^\pi}}{\left(\check{w}^{\tau,\mathrm{pol}\textrm{-}\mathrm{env}}_{\nu^\pi}\right)^2}=\log\alpha_{\mathrm{p}}+\log\alpha_{\mathrm{e}}-\log\check{w}^{\tau,\mathrm{pol}\textrm{-}\mathrm{env}}\,.$$
Now since the set $\calM_{\mathrm{sim}}^\tau$ is countably infinite and since $\check{w}^{\tau,\mathrm{pol}\textrm{-}\mathrm{env}}$ is a universal prior on $\check{\calM}^\tau_{\mathrm{pol}\textrm{-}\mathrm{env}}$ which contains the countably infinite set $\calM_{\mathrm{sim}}^\tau$, we conclude that for every $s>0$, there exists some $\lambda=\nu^\pi\in\calM_{\mathrm{sim}}^\tau$ such that $0<\check{w}^{\tau,\mathrm{pol}\textrm{-}\mathrm{env}}_{\nu^\pi}< \alpha_{\mathrm{p}}\alpha_{\mathrm{e}}e^{-s}$, hence
$$\check{S}\left(\pi,\nu,w^{\tau ,U}\right)\geq\log\alpha_{\mathrm{p}}+\log\alpha_{\mathrm{e}}-\log\check{w}^{\tau,\mathrm{pol}\textrm{-}\mathrm{env}}> s\,.$$
In other words, there are fully-supported policies and environments sharing arbitrarily large degrees of functional similarity with respect to $w^\tau$, as desired.

It remains to prove \eqref{eq:sim-univ-pol-sandwich} and \eqref{eq:sim-univ-env-sandwich}. In the following we will only prove \eqref{eq:sim-univ-pol-sandwich}, as the proof of the other equation is similar.

From \eqref{eq:prior-on-fully-supp}, we see that if 
\begin{equation}
    \label{eq:C-of-fully-supported}
    C:=\sum_{\lambda'\in \check{\calM}_{\mathrm{pol}\textrm{-}\mathrm{env}}^{\tau}} w^{\tau}_{\lambda'}\,,
\end{equation}
then

\begin{align*}
    \check{w}^{\tau,\mathrm{pol}\textrm{-}\mathrm{env}}_{\nu^\pi} &=\frac{1}{C}w_{\nu^\pi}^\tau=\frac{1}{C}\sum_{\substack{M\in\calM:\\\lambda_M^\tau=\nu^\pi}}2^{-K_U(\langle M\rangle)}\,.
\end{align*}

Similarly, from \eqref{eq:prior-on-fully-supp}, \eqref{eq:w-pol-and-w-env-fully-supp} and \eqref{eq:C-of-fully-supported}, we have

\begin{align*}
    \check{w}^{\tau,\mathrm{pol}}_\pi &=\sum_{\tilde{\nu}\in \check{\calM}^{\tau}_{\mathrm{env}}}\check{w}_{\tilde{\nu}^\pi}^\tau=\frac{1}{C}\sum_{\tilde{\nu}\in \check{\calM}^{\tau}_{\mathrm{env}}}w_{\tilde{\nu}^\pi}^\tau=\frac{1}{C}\sum_{\tilde{\nu}\in \check{\calM}^{\tau}_{\mathrm{env}}}\sum_{\substack{M\in\calM:\\\lambda_M^\tau=\tilde{\nu}^\pi}}2^{-K_U(\langle M\rangle)}\,.
\end{align*}

Therefore, if we define
$$\calM(\nu^\pi)=\{M\in\calMapom:\lambda_M^\tau=\nu^\pi\}\,,$$
and
$$\calM(\pi)=\big\{M\in\calMapom:\exists\tilde{\nu}\in \check{\calM}^{\tau}_{\mathrm{env}}\,, \lambda_M^\tau=\tilde{\nu}^\pi\}\,,$$

then
\begin{align*}
    \check{w}^{\tau,\mathrm{pol}\textrm{-}\mathrm{env}}_{\nu^\pi} =\frac{1}{C}\sum_{M\in\calM(\nu^\pi)}2^{-K_U(\langle M\rangle)}\,.
\end{align*}

and

\begin{align*}
    \check{w}^{\tau,\mathrm{pol}}_\pi &=\frac{1}{C}\sum_{M\in\calM(\pi)}2^{-K_U(\langle M\rangle)}\,.
\end{align*}

We claim that it is sufficient for the proof of \eqref{eq:sim-univ-pol-sandwich} to show that for every $\lambda=\nu^\pi\in\calM_{\mathrm{sim}}^\tau$ there exists an injective function
$$f:\calM(\pi)\to\calM(\nu^\pi)$$
such that there exists an absolute constant $C_K>0$ (which does not depend on $\lambda$) such that
\begin{equation}
    \label{eq:K-ineq-for-inj-f-from-pol-to-univ}
    K_U(\langle f( M)\rangle)\leq K_U(\langle M\rangle)+C_K\,,\forall M\in\calM(\pi)\,.
\end{equation}
Indeed, if this is the case then
\begin{align*}
\check{w}^{\tau,\mathrm{pol}\textrm{-}\mathrm{env}}_{\nu^\pi} = \frac{1}{C}\sum_{M\in\calM(\nu^\pi)}2^{-K_U(\langle M\rangle)}\stackrel{(\ast)}{\geq} \frac{1}{C}\sum_{M\in\calM(\pi)}2^{-K_U(\langle f(M)\rangle)}\geq \frac{1}{C}\sum_{M\in\calM(\pi)}2^{-K_U(\langle M\rangle)-C_K}=2^{-C_K}\check{w}^{\tau,\mathrm{pol}}_\pi\,,
\end{align*}
where $(\ast)$ follows from the fact that $f$ is injective. By choosing $\alpha_{\mathrm{p}}=2^{-C_K}$ we get our \eqref{eq:sim-univ-pol-sandwich}.

It remains to show the existence of an injective $f:\calM(\pi)\to\calM(\nu^\pi)$ satisfying \eqref{eq:K-ineq-for-inj-f-from-pol-to-univ}. For this, let us fix $\lambda=\nu^\pi\in\calM_{\mathrm{sim}}^\tau$ and an arbitrary machine $M\in\calM(\pi)$. In the following, we will describe how we can construct the machine $f(M)\in\calM(\nu^\pi)$.

Notice that since $M\in\calM(\pi)$, then there exists some $\tilde{\nu}\in \check{\calM}^{\tau}_{\mathrm{env}}$ such that $\lambda_M^\tau= \tilde{\nu}^\pi$. In particular, for every $(\HistA, a)\in(\TurnSet^*)\times\mathcal{A}$ we have\footnote{Note that $\pi$ and $\tilde{\nu}$ are fully supported, which makes $\lambda_M^\tau=\tilde{\nu}^\pi$ fully supported as well.}
$$\lambda_M^\tau(a\mid \HistA)=\pi(a\mid \HistA)\,.$$

So $M$ already has the potential of "simulating the policy part" of $\lambda=\nu^\pi$. We will formalize this notion by constructing an oracle machine $M_{\mathrm{p}}$ which takes $\HistA \in \TurnSet^*$ as input and produces $a\in\mathcal{A}$ such that by equipping $M_{\mathrm{p}}$ with the oracle $\tau$, we get that $M_{\mathrm{p}}^\tau(a\mid \HistA)=\pi(a\mid \HistA)$ for all $(\HistA, a)\in(\TurnSet^*)\times\mathcal{A}$:

\begin{algorithm}[H]
\caption{Oracle machine $M_{\mathrm{p}}$ such that when equipped with $\tau$, it samples from $\pi(a\mid \HistA)$}
\label{alg:sampling}

\KwIn{An encoded history $\langle \HistA \rangle$ with $\HistA \in \TurnSet^*$.}
\KwOut{$\langle a\rangle$ with $a\in\mathcal{A}$}
\BlankLine

$h' \leftarrow \varepsilon$\;

\While{$h' \neq \langle \HistA \rangle$}{
    Simulate running the machine $M$ to produce $l(\langle \HistA \rangle)$ bits\;
    Let $h'$ be the first $l(\langle \HistA \rangle)$ bits on the output tape of the simulated machine $M$\;
    % Let $a$ be the last symbol on the output tape of the simulated machine $M$\;
}
$a \leftarrow \varepsilon$\;
\While{$a$ is not a valid coding word of an action in $\calA$}{
    Continue simulating the machine $M$ to produce one additional output bit\;
    Append the last output bit to $a$\;
}

\KwRet{$a$}\;
\end{algorithm}
Note that since $M$ has an output distribution $\lambda_M^\tau$ which is a fully supported universe we have:
\begin{itemize}
    \item $\lambda_M^\tau$ a measure (cf. \cref{def:fully-supported-env-pol}), hence the simulation of $M$ does not get stuck into infinite loops (otherwise $\lambda_M$ would be a proper semimeasure, not a measure). Therefore, every simulation of $M$ is guaranteed to produce at least $l(\langle \HistA\rangle)$ bits.
    \item $\lambda_M^\tau(\HistA)>0$ and hence the while loop in the above algorithm is guaranteed to eventually produce the history $\HistA$.
\end{itemize}
We conclude that the above algorithm correctly samples $\pi(a\mid \HistA)$ because it follows a rejection sampling procedure.

Now since $\nu$ and $\pi$ share a common structure as described in the definition of the family $\calM_{\mathrm{sim}}^\tau$, we can use the ability of $M_{\mathrm{p}}$ to simulate $\pi$ to also simulate $\nu$. We formalize this by constructing a machine $M_{\mathrm{e}}$ as follows:

\begin{algorithm}[H]
\caption{Oracle machine $M_{\mathrm{e}}$ such that when equipped with $\tau$, it samples from $\nu(e\mid \HistA )$}
\label{alg:sampling2}

\KwIn{An encoded history $h=\langle \HistA \rangle$ with $\HistA \in \TurnSet^*$.}
\KwOut{$\langle e\rangle$ with $e\in\mathcal{E}$}
\BlankLine

$a \leftarrow M_{\mathrm{p}}(h)$\;

\tcp{Convert $a\in\mathcal{A}$ to $e \in \calE$ so that $M_{\mathrm{e}}^\tau(e\mid \HistA )=\nu(e\mid \HistA )$}
\uIf{$a = \langle \tilde{a}_0 \rangle$}{
    $e \leftarrow \langle \tilde{e}_0 \rangle$\;
}
\uElseIf{$a = \langle \tilde{a}_1\rangle $}{
    $e \leftarrow \langle \tilde{e}_1 \rangle$\;
}
\Else(\tcp*[f]{I.e., $a \in \mathcal{A} \setminus \tilde{\mathcal{A}}$}){
    Sample $e$ from $\calE \setminus \tilde{\calE}$ according to the distribution $p_{\tilde{\calE}^c}$\;
}
\KwRet{$e$}\;
\end{algorithm}

Now we are ready to describe the computational procedure of the machine $f(M)$:

\begin{algorithm}[H]
\caption{Oracle machine $f(M)$ such that $\lambda_{f(M)}^\tau=\nu^\pi$}
\label{alg:sampling3}

$h\leftarrow \varepsilon$\;
$t\leftarrow 1$\;
\While{True}{
    $a_t\leftarrow M_{\mathrm{p}}(h)$\;
    Append $a_t$ to the output tape\;
    $e_t\leftarrow M_{\mathrm{e}}(h)$\;
    Append $e_t$ to the output tape\;
    $h\leftarrow ha_te_t$\;
}
\end{algorithm}

It is not hard to see that the $K_U(f(M))\leq K_U(M)+C_K$ where $C_K$ is some absolute constant related to the program code overhead needed to apply the above procedure on top of using the code of $M$ (recall that we can define the procedure of $M$ once and then refer to it in the simulation of $M_{\mathrm{p}}$ and $M_{\mathrm{e}}$). Furthermore, we can see that changing the machine $M$ yields a syntactically different machine $f(M)$. Therefore, $M\mapsto f(M)$ is indeed injective.
\end{proof}

\begin{remark}
    \label{rem:mupi-structural-similarities-general-Solomonoff}
    \cref{thm:mupi-structural-similarities} is not too sensitive to the particular form $w_M^U=2^{-K_U(\langle M\rangle)}$ of the Solomonoff prior: If
    \begin{itemize}
        \item $g:\mathbb{N}\to[0,1]$ is a decreasing function whose decay is not faster than exponential, i.e., $\limsup_{t\to\infty}\frac{-\log g(t)}{t}<\infty$, and
        \item $\sum_{M\in\calMapom}g(K_U(\langle M\rangle))\leq 1$,
    \end{itemize}
    then we can adapt the proof of \cref{thm:mupi-structural-similarities} to show that it still applies to $w_M=g(K_U(\langle M\rangle))$.
\end{remark}

\subsection{Universes sharing a common algorithmic structure}
\label{app:structural-similarities-algorithmic}

Recall that in the proof of \cref{thm:mupi-structural-similarities}, in order to show that there are (fully-supported) policies and environments sharing arbitrary large degrees of functional similarity according to the Solomonoff prior, we constructed a family $\calM_{\mathrm{sim}}^\tau$ of (fully-supported) universes in which policies and environments share a common structure, and then we showed that inside $\calM_{\mathrm{sim}}^\tau$ we can always find $\lambda=\nu^\pi$ for which the degree of functional similarity between $\nu$ and $\pi$ according to the Solomonoff prior is arbitrarily large.

This fact does not depend on the particular form of the structure which is shared among all universes in $\calM_{\mathrm{sim}}^\tau$. In this section, we formalize a general notion of "algorithmic structure" and then show that as long as this structure satisfies a few conditions, then we can always find policies and environments respecting this structure and which share arbitrarily large degrees of functional similarity with respect to the Solomonoff prior.

Intuitively, an algorithmic structure can be understood as a "boilerplate template" structure that can be respected by multiple computer programs. Let us consider the following two programs

\begin{algorithm}[H]
\caption{Pseudocode using the subroutine function $f(x) = x+1$}
\label{alg:program_with_f_x_to_x_plus_1}

\SetKwProg{Fn}{Function}{:}{}
\SetKwFunction{F}{f}
\SetKwFunction{Main}{main\_func}

\Fn{\F{$x$}}{
    \KwRet{$x+1$}\;
}
\BlankLine

\Fn{\Main{$x$}}{
    $y \leftarrow 0$\;
    \For{$i \leftarrow 1$ \KwTo $x$}{
        $y \leftarrow \F(y + \F(x))$\;
    }
    \KwRet{$y$}\;
}
\end{algorithm}

\begin{algorithm}[H]
\caption{Pseudocode using the subroutine function $f(x) = x^2$}
\label{alg:program_with_f_x_to_x_squared}

\SetKwProg{Fn}{Function}{:}{}
\SetKwFunction{F}{f}
\SetKwFunction{Main}{main\_func}

\Fn{\F{$x$}}{
    \KwRet{$x^2$}\;
}
\BlankLine

\Fn{\Main{$x$}}{
    $y \leftarrow 0$\;
    \For{$i \leftarrow 1$ \KwTo $x$}{
        $y \leftarrow \F(y + \F(x))$\;
    }
    \KwRet{$y$}\;
}
\end{algorithm}

As one can see, the main function \texttt{main\_func} has a similar algorithmic structure in both programs, as it uses the subroutine function \texttt{f} in the same way, and they differ only by the specification of $f$.

We will use this idea to define a notion of an algorithmic structure that can be shared by a family of universes. We will basically define a fixed "boilerplate template" structure which refers to some subroutine that is not specified, and we will define the family of all universes that can be implemented by an algorithm that respects the "boilerplate template" structure. Different specifications of the subroutine yield different universes in the family.

The reader might have noticed that the notion of subroutine is very similar to the notion of an oracle that can be called by an oracle machine, and this is indeed how we will formalize this notion.

For this, we need to introduce machines that can use more than one oracle.

\begin{definition}[Multi-Oracle Machines]
    An \textbf{$n$-abstract probabilistic oracle machine ($n$-APOM)} is a monotone Turing machine equipped with $n$ distinct oracle tapes, denoted $T_1, \dots, T_n$, and a corresponding set of $n$ distinct query instructions, $Q_1, \dots, Q_n$. When instruction $Q_i$ is executed, the machine interacts with oracle tape $T_i$.
    
    An \textbf{$n$-probabilistic oracle machine ($n$-POM)} is a pair $(M, (\tau_1, \dots, \tau_n))$, where $M$ is an $n$-APOM and each $\tau_i: \{0,1\}^* \to [0,1]$ is a probabilistic oracle\footnote{Recall that if $x$ is the query to the oracle $\tau_i$, the oracle returns 1 with probability $\tau_i(x)$, and returns 0 with probabilty $1-\tau_i(x)$.} . When instruction $Q_i$ is called, the content of the tape $T_i$ is interpreted as a query which is answered by oracle $\tau_i$ and the content of the tape $T_i$ is replaced with the answer of the oracle. The standard APOM and POM are equivalent to a $1$-APOM and $1$-POM, respectively.
\end{definition}

Each of the $n$ probabilistic oracles can be understood as a distinct, black-box \textbf{subroutine} that the main program (the $n$-APOM) can call. The $n$-APOM defines the high-level logic or "template", while the specific oracles provide the implementation details for its sub-components.

\begin{remark}
    While the above definition only uses probabilistic oracles that return single random bits, the definition is general enough to model probabilistic oracles that can return random bit strings. Basically, if $O:\{0,1\}^*\to\Delta\{0,1\}^*$ is a probabilistic oracle that returns random bit strings, we can effectively represent it using two probabilistic oracles, $\tau_e:\{0,1\}^*\to[0,1]$ and $\tau_b:\{0,1\}^*\to[0,1]$, where:
    \begin{itemize}
        \item $\tau_e(\langle x,y\rangle)$ represents the probability that $O(x)=y$ conditioned on $y$ being a prefix of $O(x)$.
        \item $\tau_b(\langle x,y\rangle)$ represents the probability that $O(x)=y1$ conditioned on $y$ being a strict prefix of $O(x)$.
    \end{itemize}
    The following pseudocode illustrates how we can use $\tau_e$ and $\tau_b$ to simulate sampling from $O$:
    
    \begin{algorithm}[H]
        \caption{Sampling from $O$ using $\tau_e$ and $\tau_b$}
        \label{alg:simulate_sample_multibit_oracle}
        \KwIn{A bit string $x\in\{0,1\}^*$.}
        \KwOut{A random sample from $O(x)$}
        \BlankLine

        $y\leftarrow\varepsilon$

        \While{True}{
            \If{$O^{\tau_e}(\langle x,y\rangle)=1$}{
                \KwRet{$y$};
            }
            \uElseIf{$O^{\tau_b}(\langle x,y\rangle)=1$}{
                $y\leftarrow y1$\;
            }
            \Else{
                $y\leftarrow y0$\;
            }
        }
    \end{algorithm}
\end{remark}

The framework with multiple oracles allows us to compose machines. An $n$-APOM can be reduced to a machine with fewer free subroutines by specifying what some of its oracles are. For instance, if we fix a base probabilistic oracle $\tau$, we can construct a complex $1$-POM by using an $n$-APOM where the first oracle remains $\tau$, but the other $n-1$ oracles are themselves implemented by other $1$-POMs that also rely on $\tau$. The resulting composite machine is a single, more complex $1$-POM that depends only on the base oracle $\tau$.

More formally, we can think of this composition syntactically. Given an $n$-APOM, $M$, and $n-1$ standard $1$-APOMs, $M_2, \dots, M_n$, we can construct a new $1$-APOM, denoted $M[M_2, \dots, M_n]$. This new machine simulates $M$ as follows:
\begin{itemize}
    \item When $M$ calls its first oracle, $Q_1$, the new machine passes this query to its own single oracle.
    \item When $M$ calls any other oracle $Q_i$ (for $i \in \{2, \dots, n\}$), it instead simulates the execution of the corresponding machine $M_i$, feeding the query string as input to $M_i$ and using the output of $M_i$ as the oracle's answer. While executing the machine $M_i$, any call to the (single) oracle of $M_i$ is directed to the single oracle of the new machine.
\end{itemize}

This allows us to formally define the notion of a shared algorithmic structure.

\begin{definition}[Algorithmic Structure]
    An $n$-APOM, $M$, defines an \textbf{algorithmic structure}. A $1$-APOM, $M'$, is said to \textbf{possess the structure} defined by $M$ if there exist $n-1$ other $1$-APOMs, $M_2, \dots, M_n$, such that $M'$ is computationally equivalent to the composite machine $M[M_2, \dots, M_n]$.
\end{definition}

Now we can define specific types of algorithmic structures for policies, environments and universes.

\begin{definition}[Algorithmic Structure for policies]
     An \textbf{algorithmic structure for policies} is an $n$-APOM, $M$, that is syntactically constrained to take as inputs valid encodings $\langle \HistA \rangle$ from $\TurnSet^*$ and produce outputs $\langle a \rangle$  with $a\in\mathcal{A}$. We say that a policy $\pi:\TurnSet^*\to\Delta'\mathcal{A}$ possesses the structure $M$ with respect to a probabilistic oracle $\tau:\{0,1\}^*\to[0,1]$ if there exists a 1-APOM $M_\pi$ such that
     \begin{itemize}
        \item $\pi(a \mid \HistA)=\mathbb{P}[M_\pi^\tau(\langle \HistA\rangle)=\langle a\rangle]$, with $\mathbb{P}[M_\pi^\tau(\langle \HistA\rangle)=\langle a\rangle]$ denoting the probability that $M_\pi^\tau$ outputs $\langle a \rangle$ upon input $\langle \HistA \rangle$, and
        \item $M_\pi$ possesses the structure defined by $M$.
     \end{itemize}
     We denote the set of policies that possess the structure $M$ with respect to $\tau$ as $\calM_{\mathrm{pol}}^\tau(M)$.
\end{definition}

\begin{definition}[Algorithmic Structure for environments]
     An \textbf{algorithmic structure for environments} is an $n$-APOM, $M$, that is syntactically constrained to take as inputs $\langle \HistA,a\rangle$ with $(\HistA, a) \in \TurnSet^*\times\mathcal{A}$ and produce outputs $\langle e \rangle$ with $e \in \calE$. We say that an environment $\nu:\TurnSet^*\times \calA\to\Delta'\mathcal{E}$ possesses the structure $M$ with respect to a probabilistic oracle $\tau:\{0,1\}^*\to[0,1]$ if there exists a 1-APOM $M_\nu$ such that
     \begin{itemize}
        \item $\nu(e \mid \HistA a)=\mathbb{P}[M_\nu^\tau(\langle \HistA, a\rangle)=\langle e\rangle]$, with $\mathbb{P}[M_\nu^\tau(\langle \HistA, a\rangle)=\langle e\rangle]$  denoting the probability that $M_\nu^\tau$ outputs $\langle e \rangle$ upon input $\langle \HistA ,a\rangle$, and
        \item $M_\nu$ possesses the structure defined by $M$.
     \end{itemize}
     We denote the set of environments that possess the structure $M$ with respect to $\tau$ as $\calM_{\mathrm{env}}^\tau(M)$.
\end{definition}

\begin{definition}[Algorithmic Structure for universes]
     An \textbf{algorithmic structure for universes} is an $n$-APOM, $M$, that does not have any inputs, and writes sequences from $\mathcal{B}^\#$ on its output tape. We say that a semimeasure $\sigma:\mathcal{B}^*\to [0,1]$ possesses the structure $M$ with respect to a probabilistic oracle $\tau$ if there exists a 1-APOM $M_\sigma$ such that
     \begin{itemize}
        \item $\sigma=\lambda_{M_\sigma}^\tau$, and
        \item $M_\sigma$ possesses the structure defined by $M$.
     \end{itemize}
     We denote the set of universes that possess the structure $M$ with respect to $\tau$ as $\calM_{\mathrm{univ}}^\tau(M)$.
\end{definition}

One way to get an algorithmic structure for universes is by specifying a pair of algorithmic structures, one for policies and one for environments:

\begin{definition}
     Let $M_p$ be an algorithmic structure for policies and $M_e$ be an algorithmic structure for environments with the same number $n$ of oracles. We can syntactically define an $n$-APOM that alternates between applying $M_p$ and $M_e$. This induces an algorithmic structure for universes. We refer to the set of universes which possess this induced structure with respect to a probabilistic oracle $\tau$ as  $\calM_{\mathrm{univ}}^\tau(M_p,M_e)$.
\end{definition}

It is worth noting that if $\pi\in \calM_{\mathrm{pol}}^\tau(M_p)$ and $\nu\in \calM_{\mathrm{env}}^\tau(M_e)$, then we do not necessarily have $\nu^\pi\in\calM_{\mathrm{univ}}^\tau(M_p,M_e)$. The reason is that $\nu$ and $\pi$ might be using different sets of 1-APOMs $M_2,\ldots,M_n$ for the specification of their subroutines. More precisely, if $M_{2,\pi},\ldots,M_{n,\pi}$ are 1-APOMs such that $\pi=M_p[M_{2,\pi},\ldots,M_{n,\pi}]^\tau$, and if $M_{2,\nu},\ldots,M_{n,\nu}$ are 1-APOMs such that $\nu=M_e[M_{2,\nu},\ldots,M_{n,\nu}]^\tau$, then $\nu^\pi$ is not guaranteed to be in $\calM_{\mathrm{univ}}^\tau(M_p,M_e)$ if $(M_{2,\pi},\ldots,M_{n,\pi})\neq(M_{2,\nu},\ldots,M_{n,\nu})$. However, if $(M_{2,\pi},\ldots,M_{n,\pi})=(M_{2,\nu},\ldots,M_{n,\nu})$, then we would indeed have $\nu^\pi\in\calM_{\mathrm{univ}}^\tau(M_p,M_e)$.

We will show that if the structures $M_p$ and $M_e$ satisfy a few properties, then we can always find $\pi\in \calM_{\mathrm{pol}}^\tau(M_p)$ and $\nu\in \calM_{\mathrm{env}}^\tau(M_e)$ such that $\nu^\pi\in\calM_{\mathrm{univ}}^\tau(M_p,M_e)$ and $\pi$ and $\nu$ share arbitrarily large degrees of functional similarity with respect to the Solomonoff prior. A key property that is needed is the "identifiability" of subroutines. In the following, we will formally define this property, but first, we need to introduce proper subroutines:

\begin{definition}[Proper subroutines]
    A POM $M^\tau$ is said to be a proper subroutine if for every input $x\in\calB^*$, the POM $M^\tau$ almost surely halts with some $y\in\calB$ written on the output tape. In other words, $M^\tau$ induces a mapping $M^\tau:\calB^*\to\Delta\calB$.
\end{definition}

\begin{definition}[Distributional equivalence of POMs]
    Let $M_1$ and $M_2$ be two 1-APOMs. We say that $M_1^\tau$ and $M_2^\tau$ are distributionally equivalent if for every $x,y\in\calB^*$, we have
    $$\mathbb{P}[M_1^\tau(x)=y]=\mathbb{P}[M_2^\tau(x)=y]\,.$$
    We write $M_1^\tau\equiv M_2^\tau$ to denote the distributional equivalence of $M_1^\tau$ and $M_2^\tau$.
\end{definition}

\begin{definition}[Identifiable subroutines]
    Let $M$ be an $n$-APOM. We say that the algorithmic structure $M$ has identifiable subroutines with respect to the probabilistic oracle $\tau$ if there are $n-1$ 2-APOMs $I_2,\ldots,I_n$ such that for every $n-1$ 1-APOMs $M_2,\ldots,M_n$ for which $M_2^\tau,\ldots,M_n^\tau$ are proper subroutines, and every $2\leq i\leq n$, the POMs $I_i[M[M_2,\ldots,M_n]]^\tau\equiv M_i^\tau$, i.e., for every $x\in\mathcal{B}^*$ and every $y\in\calB$,\footnote{Note that by definition, a proper subroutine can only output single bits, so we only need to check $y\in\calB$.} we have
    $$\mathbb{P}[I_i[M[M_2,\ldots,M_n]]^\tau(x)=y]=\mathbb{P}[M_i^\tau(x)=y]\,.$$

    In other words, if we had access to an oracle that on any query $x$ returns samples from $M[M_2,\ldots,M_n]^\tau(x)$, then we can write a program to simulate samples from $M_i^\tau(x)$ for all $x$ as well.
\end{definition}

Let us construct a simple example to illustrate the concept of subroutine identifiability:

\begin{example}[A simple example of an Algorithmic Structure with Identifiable Subroutines]
\label{example:identifiable-subroutine-simple}
    Consider the algorithmic structure defined by the following 2-APOM $M_{struct}$:
    
    \begin{algorithm}[H]
        \caption{Algorithmic Structure $M_{struct}$}
        \KwIn{An input string $x \in \{0,1\}^*$.}
        \KwOut{A canonical encoding of a pair of strings $\langle y, z \rangle$.}
        \BlankLine
        $y \leftarrow Q_2(x)$\;
        $z \leftarrow Q_1(y)$\;
        \KwRet{$\langle y, z \rangle$}\;
    \end{algorithm}
    
    This structure has "identifiable subroutines" because we can construct an "identifier" machine, $I_2$, that can recover the behavior of any proper subroutine $M_2$ that may be plugged into the second oracle slot of $M_{struct}$.
    
    \begin{algorithm}[H]
        \caption{Identifier machine $I_2$}
        \KwIn{An input string $x \in \{0,1\}^*$.}
        \KwOut{A string $y$.}
        \BlankLine
        $s \leftarrow Q_2(x)$\;
        Decode $s$ into a pair $\langle y, z \rangle$\;
        \KwRet{$y$}\;
    \end{algorithm}
    
    Now, let us verify that $I_2$ can indeed identify any subroutine specification for the second oracle $Q_2$ of $M_{struct}$. Let $M_2$ be an arbitrary 1-APOM. We need to show that the 1-POM $I_2[M_{struct}[M_2]]^\tau$ is distributionally equivalent to $M_2^\tau$. Let's trace the execution of $I_2[M_{struct}[M_2]]^\tau$ on an input $x$:
    \begin{enumerate}
        \item The machine begins by simulating $I_2$. The first step of $I_2$ is to call its second oracle, $Q_2$, with input $x$.
        \item According to the composition rule, this call to $I_2$'s second oracle is replaced by a simulation of the machine plugged into that slot, which is $M_{struct}[M_2]$. This simulation is run with the base oracle $\tau$.
        \item So, we execute $M_{struct}[M_2]^\tau(x)$. This machine, in turn, first simulates $M_2$ with the oracle $\tau$. Let the probabilistic output of $M_2^\tau(x)$ be the string $y_{out}$.
        \item Next, $M_{struct}[M_2]^\tau$ calls its first oracle (which is passed through to the base oracle $\tau$) with $y_{out}$. Let the result be $z_{out}$. It then outputs the pair $\langle y_{out}, z_{out} \rangle$.
        \item This pair is the result returned to $I_2$ from its oracle call in step 1.
        \item $I_2$ then proceeds to its second step: It decodes the pair $\langle y_{out}, z_{out} \rangle$ and outputs the first element, which is $y_{out}$.
    \end{enumerate}
    The final output is $y_{out}$, which is precisely the probabilistic output of $M_2^\tau(x)$. Therefore, the machines are distributionally equivalent, and we have successfully "identified" or extracted the behavior of the subroutine $M_2$.
\end{example}

The structure $M_{struct}$ in the above example has identifiable subroutines because it explicitly includes the output of its subroutines in its own final output. This "pass-through" design is a straightforward way to ensure that the behavior of sub-components is not obscured by subsequent computations. Nevertheless, there are more complex algorithmic structures that can still have identifiable subroutines, as illustrated in the following example:

\begin{example}[A Slightly More Complex Example of an Algorithmic Structure with Identifiable Subroutines]
\label{example:complex-algorithmic-structure-identifiable-subroutines}
    Consider the algorithmic structure defined by the following 3-APOM $M_{struct}$:
    
    \begin{algorithm}[H]
        \caption{A Slightly More Complex Algorithmic Structure $M_{struct}$}
        \KwIn{An input string $x \in \{0,1\}^*$.}
        \KwOut{A canonical encoding of a pair $\langle r,y\rangle$ or $\langle x',y\rangle$.}
        \BlankLine
        \uIf{$x = \varepsilon$}{
            Flip a random coin $r \in \{0,1\}$\;
            \uIf{$r=0$}{
                $y \leftarrow Q_2(\varepsilon)$\;
            }
            \Else{
                $y \leftarrow Q_3(\varepsilon)$\;
            }
            \KwRet{$\langle r, y \rangle$}\;
        }
        \Else{
            Draw a random string $x'\in\{0,1\}^{l(x)}$ \;
            $x'' \leftarrow x \oplus x'$ \tcp*[l]{We compute the bit-wise XOR between $x$ and $x'$.}
            \uIf{the number of ones in $x'$ is even}{
                $y \leftarrow Q_2(x'')$\;
            }
            \Else{
                $y \leftarrow Q_3(x'')$\;
            }
            \KwRet{$\langle x', y \rangle$}\;
        }
    \end{algorithm}
    The structure $M_{struct}$ also has identifiable subroutines, although extracting the subroutines requires a more sophisticated process than simply decoding the output. To identify $M_2^\tau$, we must repeatedly sample from the composite machine $M_{struct}[M_2, M_3]^\tau$ and filter for the cases where the random string $x'$ has an even number of ones.

    The following pseudocode illustrates the algorithm of $I_2$.

    \begin{algorithm}[H]
    \caption{Identifier machine $I_2$}
    \label{alg:identifier_I2}
    \KwIn{An input string $x_{in} \in \{0,1\}^*$.}
    \KwOut{A random sample from $M_2^\tau(x_{in})$}
    \BlankLine
    \If{$x_{in} = \varepsilon$}{
        \While{True}{
            Call the second oracle $Q_2$ with input $x_{in} = \varepsilon$. Let the result be $s$\;
            Decode $s$ as a pair $\langle r, y \rangle$\;
            \If{$r=0$}{
                \KwRet{$y$}\;
            }
        }
    }
    \Else{
        \While{True}{
            Draw a random string $x \in \{0,1\}^{l(x_{in})}$ using the internal randomness tape\;
            Call the second oracle $Q_2$ with input $x$. Let the result be $s$\;
            Decode $s$ as a pair $\langle x', y \rangle$\;
            \If{$x \oplus x' = x_{in}$ \textbf{and} the number of ones in $x'$ is even}{
                \KwRet{$y$}\;
            }
        }
    }
    \end{algorithm}
    An analogous machine $I_3$ can be designed by checking for $r=1$ when $x_{in}=\varepsilon$, and checking for an odd number of ones in $x'$ in case $x_{in}\neq\varepsilon $.
    
    To prove that the structure is identifiable, we must show that $I_2[M_{struct}[M_2, M_3]]^\tau$ is distributionally equivalent to $M_2^\tau$. We analyze the two cases based on the input $x_{in}$ to $I_2$.
    
    \begin{itemize}
        \item \textbf{Case 1: $x_{in} \neq \varepsilon$.} The machine $I_2$ enters a loop where it draws a random string $x$ of the same length as $x_{in}$ and queries its second oracle (the composite machine $M_{struct}[M_2, M_3]^\tau$) with this $x$. The composite machine, in turn, draws another random mask $x'$ and calls either $M_2$ or $M_3$ with the input $x'' = x \oplus x'$. The identifier $I_2$ accepts the result $y$ only if the number of ones in $x'$ is even (indicating $M_2$ was called) and if $x \oplus x' = x_{in}$. Since both $x$ and $x'$ are drawn uniformly at random, the resulting input $x''$ to the subroutine is also uniformly random over strings of that length. Therefore, for any target $x_{in}$, the condition $x'' = x_{in}$ will eventually be met. When it is, the returned $y$ is a correct sample from $M_2^\tau(x_{in})$.
        \item \textbf{Case 2: $x_{in} = \varepsilon$.} The machine $I_2$ enters a simpler loop. It repeatedly queries its second oracle with the empty string $\varepsilon$. In response, the composite machine $M_{struct}[M_2, M_3]^\tau$ flips a coin $r$. If $r=0$, it calls $M_2(\varepsilon)$; otherwise, it calls $M_3(\varepsilon)$. It returns the pair $\langle r, y \rangle$. The identifier $I_2$ waits until it receives a pair where $r=0$. Since the coin flip is fair, this is guaranteed to happen eventually. When it does, the corresponding $y$ is a correct sample from $M_2^\tau(\varepsilon)$.
    \end{itemize}
    
    In both cases, the identifier machine $I_2$ successfully simulates a sample from $M_2^\tau(x_{in})$ using its oracle access to the composite machine. This process of filtering executions is a form of rejection sampling. Thus, the structure has identifiable subroutines.
\end{example}

With this general framework for algorithmic structures, we can now state the main result of this section.

\begin{theorem}
    \label{thm:mupi-structural-similarities-through-algorithmic-structures}
    For every fixed choice of:
    \begin{itemize}
        \item a canonical encoding $\langle M\rangle$ of APOMs as binary strings,
        \item a universal monotone Turing machine $U$,
        \item a probabilistic oracle $\tau$ (which may or may not be a RUI-oracle),
        \item sets $\mathcal{A}$, and $\mathcal{E}$, with $|\mathcal{A}|\geq 2$ and $|\calE|\geq 2$, and corresponding complete prefix-free codes,
    \end{itemize}
    if $M_p$ and $M_e$ are algorithmic structures for policies and environments, respectively, such that:
    \begin{itemize}
        \item $M_p$ and $M_e$ have the same number of oracles $n\geq 2$,
        \item $M_p$ and $M_e$ have identifiable subroutines with respect to $\tau$, and
        \item the set of fully-supported universes in $\calM_{\mathrm{univ}}^\tau(M_p,M_e)$ is infinite,
    \end{itemize}
    then there are fully-supported policies $\pi\in\calM_{\mathrm{pol}}^\tau(M_p)$ and fully-supported environments $\nu\in\calM_{\mathrm{env}}^\tau(M_e)$ sharing arbitrarily large degrees of functional similarity with respect to the Solomonoff prior $w^{U,\tau}$, i.e., for every $s>0$, there exists at least one such pair $(\pi, \nu)$ for which
    $$\check{S}\left(\pi,\nu,w^{U,\tau}\right) > s\,.$$
    Furthermore, we can choose this pair $(\pi, \nu)$ such that their subroutines are identical, which implies that their interaction $\nu^\pi$ is an element of $\calM_{\mathrm{univ}}^\tau(M_p,M_e)$.
\end{theorem}
\begin{proof}
The proof strategy mirrors that of \cref{thm:mupi-structural-similarities}. We first establish the existence of an infinite family of suitable pairs $(\pi, \nu)$, then show that complexity savings arguments lead to arbitrarily large degrees of functional similarity for some members of this family.

The theorem assumes that the set of fully-supported universes in $\calM_{\mathrm{univ}}^\tau(M_p,M_e)$ is infinite. By definition, any universe $\lambda$ in this set is formed by the interaction of a policy $\pi \in \calM_{\mathrm{pol}}^\tau(M_p)$ and an environment $\nu \in \calM_{\mathrm{env}}^\tau(M_e)$ that share the \textit{same} underlying subroutines $\{M_i\}_{i=2}^n$. Since each such universe is fully-supported, the pair $(\pi, \nu)$ that generates it is unique (\cref{prop:fully-supp-pol-env-give-uniq-univ}) and both $\pi$ and $\nu$ are also fully-supported (\cref{prop:fully-supp-pol-env}). Therefore, we have an infinite family of pairs $(\pi, \nu)$ of fully-supported policies  $\pi \in \calM_{\mathrm{pol}}^\tau(M_p)$ and fully-supported environments  $\nu \in \calM_{\mathrm{env}}^\tau(M_e)$ such that $\nu^\pi\in \calM_{\mathrm{univ}}^\tau(M_p,M_e)$.

Our goal is to show that for any $s>0$, there is a pair $(\pi, \nu)$ in this family with $\check{S}(\pi, \nu, w^{U,\tau}) > s$. Recall that $\check{S}\left(\pi,\nu,w^{U,\tau}\right)=\log \frac{\check{w}_{\nu^\pi}^{\tau,\mathrm{pol}\textrm{-}\mathrm{env}}}{\check{w}^{\tau,\mathrm{pol}}_\pi \check{w}^{\tau,\mathrm{env}}_{\nu}}$. As in the proof of \cref{thm:mupi-structural-similarities}, it is sufficient to show that there are absolute constants $\alpha_p, \alpha_e > 0$ such that for every pair $(\pi, \nu)$ in our infinite family, the following inequalities hold:
\begin{equation}
    \label{eq:gen-sim-univ-pol-sandwich}
    \alpha_{\mathrm{p}} \check{w}^{\tau,\mathrm{pol}}_{\pi} \leq \check{w}^{\tau,\mathrm{pol}\textrm{-}\mathrm{env}}_{\nu^\pi}\quad \text{and} \quad \alpha_{\mathrm{e}} \check{w}^{\tau,\mathrm{env}}_{\nu} \leq \check{w}^{\tau,\mathrm{pol}\textrm{-}\mathrm{env}}_{\nu^\pi}\,.
\end{equation}
If this condition holds, then $\check{S}(\pi,\nu,w^{U,\tau}) \geq \log(\alpha_p\alpha_e) - \log(\check{w}_{\nu^\pi}^{\tau,\mathrm{pol}\textrm{-}\mathrm{env}})$. Since our family of universes is infinite and the Solomonoff prior $w^{U,\tau}$ must be non-zero for all of them, then for any $\delta > 0$, there must exist a universe $\nu^\pi$ in the family with prior weight $\check{w}_{\nu^\pi}^{\tau,\mathrm{pol}\textrm{-}\mathrm{env}} < \delta$. By choosing $\delta$ small enough, we can make $-\log(\check{w}_{\nu^\pi}^{\tau,\mathrm{pol}\textrm{-}\mathrm{env}})$ arbitrarily large, thus making $\check{S}$ exceed any given $s$.

Let us now fix a pair $(\pi, \nu)$ from our infinite family, such that $\pi = M_p[M_2, \dots, M_n]^\tau$ and $\nu = M_e[M_2, \dots, M_n]^\tau$ for some set of 1-APOMs $\{M_i\}_{i=2}^n$, and their interaction is $\nu^\pi \in \calM_{\mathrm{univ}}^\tau(M_p,M_e)$. We will prove the first inequality in \eqref{eq:gen-sim-univ-pol-sandwich}; the proof for the second is symmetric.

Let $\calM(\nu^\pi)$ be the set of 1-APOMs that compute the universe $\nu^\pi$, i.e., $$\calM(\nu^\pi) = \{ M \in \calMapom : M^\tau = \nu^\pi \}\,.$$
Let $\calM(\pi)$ be the set of 1-APOMs that compute a universe whose policy part is $\pi$, i.e., $$\calM(\pi)=\big\{M\in\calMapom:\exists\tilde{\nu}\in \check{\calM}^{\tau}_{\mathrm{env}}\,, \lambda_M^\tau=\tilde{\nu}^\pi\}\,.$$

We construct an injective map $f: \calM(\pi) \to \calM(\nu^\pi)$ such that for any $M \in \calM(\pi)$, we have $K_U(f(M)) \le K_U(M) + C_p$ for some absolute constant $C_p$ independent of $M$ and $\pi$.

Let $M \in \calM(\pi)$. By definition, this machine computes some universe $\tilde{\nu}^\pi$ whose policy component is our target policy $\pi$.
First, we construct a policy simulator machine, $M_\pi$, from $M$. This machine, given oracle access to $\tau$, can sample from $\pi(\cdot\mid \HistA )$ for any history $\HistA$. It does so by rerunning a simulation of $M^\tau$ and generating bitstrings of length $l(\langle \HistA \rangle)$ until it generates $\langle \HistA \rangle$ and then it outputs the encoding of the next action. As described in the proof of \cref{thm:mupi-structural-similarities}, we can see that the output distribution of $M_\pi^\tau$ is equal to $\Mvp$.

Since the policy structure $M_p$ has identifiable subroutines with respect to $\tau$, there are identifier machines $\{I_{p,i}\}_{i=2}^n$ such that for all $2\leq i\leq n$, we have
$$I_{p,i}[M_p[M_2, \dots, M_n]]^\tau \equiv M_i^\tau\,,$$
where $\equiv$ denotes the distributional equivalence, i.e., for every $x\in\calB^*$ and every $y\in\calB$, we  have $\mathbb{P}[I_{p,i}[M_p[M_2, \dots, M_n]]^\tau(x)=y]=\mathbb{P}[M_i^\tau(x)=y]$.

We can now construct a set of 1-APOMs $\{M_i'\}_{i=2}^n$, where each $M_i'$ is distributionally equivalent to the corresponding subroutine $M_i$ when equipped with $\tau$: Let $M_i' = I_{p,i}[M_\pi]$, then
\begin{align*}
    M_i'^\tau \equiv I_{p,i}[M_\pi]^\tau \stackrel{(\ast)}{\equiv} I_{p,i}[M_p[M_2, \dots, M_n]]^\tau \equiv M_i^\tau\,,
\end{align*}
where $(\ast)$ follows from the fact that $M_\pi^\tau\equiv M_p[M_2, \dots, M_n]^\tau$ (as they both simulate the policy $\pi$).

Now that we can simulate the subroutines $M_i^\tau$, we can construct a 1-APOM $M_\nu$ that simulates the target environment $\nu$. We define $M_\nu = M_e[M_2', \dots, M_n']$. When equipped with $\tau$, we get:
$$M_\nu^\tau \equiv M_e[M_2', \dots, M_n']^\tau \stackrel{(\dagger)}{\equiv} M_e[M_2, \dots, M_n]^\tau \equiv \nu\,,$$
where $(\dagger)$ follows from the fact that $M_i'^\tau\equiv M_i^\tau$ for all $2\leq i\leq n$.

The machine $f(M)$ can now be defined as a 1-APOM that alternates between running the policy simulator $M_\pi$ and the environment simulator $M_\nu$. The complete machine $f(M)$ can be described by a fixed universal program that takes the description of $M$ as input and implements the logic described above. This universal program needs the code for the identifiers $\{I_{p,i}\}$ and the structures $M_p, M_e$, all of which are fixed. Therefore, its complexity is bounded by $K_U(f(M)) \le K_U(M) + C_p$ for some constant $C_p$ that depends only on the fixed structures. The map $M \mapsto f(M)$ is injective because the description of $M$ is part of the description of $f(M)$.

With this injective map and complexity bound, the argument is identical to the final part of the proof of \cref{thm:mupi-structural-similarities}. We have:
\begin{align*}
\check{w}^{\tau,\mathrm{pol}\textrm{-}\mathrm{env}}_{\nu^\pi} &= \frac{1}{C}\sum_{M' \in \calM(\nu^\pi)} 2^{-K_U(\langle M' \rangle)} \\
&\ge \frac{1}{C}\sum_{M \in \calM(\pi)} 2^{-K_U(\langle f(M)\rangle)} \quad (\text{since } f \text{ is injective}) \\
&\ge \frac{1}{C}\sum_{M \in \calM(\pi)} 2^{-K_U(\langle M\rangle) - C_p} \\
&= 2^{-C_p} \left( \frac{1}{C}\sum_{M \in \calM(\pi)} 2^{-K_U(\langle M\rangle)} \right) = 2^{-C_p} \check{w}^{\tau,\mathrm{pol}}_\pi\,.
\end{align*}
By choosing $\alpha_p = 2^{-C_p}$, we get the first inequality. A symmetric argument using the identifiability of the subroutines of $M_e$ yields the second inequality for a constant $\alpha_e = 2^{-C_e}$. This completes the proof.
\end{proof}

\section{Multi-agent environments and their relation to universes}\label{app:multi-agent-env}

Here, we provide some additional details on multi-agent environments and how to `embed' agent policies by interfacing them with this multi-agent environment.

\begin{definition}
     A $k$-multi-agent environment $\bar{\nu}$ is a tuple $$\left(k,\mathcal{A}^{(1)},\ldots,\mathcal{A}^{(k)},\mathcal{O}^{(1)},\ldots,\mathcal{O}^{(k)},\mathcal{R}^{(1)},\ldots,\mathcal{R}^{(k)},f_\nu\right)\,,$$ which satisfies:
    \begin{itemize}
        \item $\mathcal{A}^{(1)},\ldots,\mathcal{A}^{(k)}$ are $k$ finite sets which we interpret as the action spaces of the $k$ agents, respectively.
        \item $\mathcal{O}^{(1)},\ldots,\mathcal{O}^{(k)}$ are $k$ finite sets which we interpret as the partial observation spaces of the $k$ agents, respectively.
        \item $\mathcal{R}^{(1)},\ldots,\mathcal{R}^{(k)}$ are $k$ finite subsets of $[0,1]$ which we interpret as the reward spaces of the $k$ agents, respectively. We write $$\mathcal{E}^{(i)}:=\mathcal{O}^{(i)}\times \mathcal{R}^{(i)}$$ to denote the percept space of the $i$-th agent. We also use the notation
        $\TurnSet^{(i)*}=(\mathcal{A}^{(i)}\times \mathcal{E}^{(i)})^*$ for the set of histories from the perspective of the $i$-th agent.
        \item $f_{\bar{\nu}}$ is a
        mapping $\TurnSet^{(1:k)*}\times\mathcal{A}^{(1:k)}\to\Delta'\mathcal{E}^{(1:k)}$, where
        $$\mathcal{A}^{(1:k)}:=\prod_{i=1}^k\mathcal{A}^{(i)}\,,$$
        $$\mathcal{E}^{(1:k)}:=\prod_{i=1}^k\mathcal{E}^{(i)}\,,$$
        and
        $$\TurnSet^{(1:k)*}:=\left(\mathcal{A}^{(1:k)}\times\mathcal{E}^{(1:k)}\right)^*\,.$$
    \end{itemize}
    For every $\Turn^{(1:k)}_{<t}\in\left(\mathcal{A}^{(1:k)}\times \mathcal{E}^{(1:k)}\right)^{t-1}$, every $e^{(1:k)}_t=\left(e^{(1)}_t,\ldots, e^{(k)}_t\right)\in\mathcal{E}^{(1:k)}$, and every $a^{(1:k)}_t=\left(a^{(1)}_t,\ldots, a^{(k)}_t\right)\in\mathcal{A}^{(1:k)}$, we introduce the notation
    \begin{align*}
        \bar{\nu}\left(e^{(1:k)}_t\middle|\Turn^{(1:k)}_{<t},a^{(1:k)}_t\right)&:=\bar{\nu}\left(e^{(1)}_t,\dots,e^{(k)}_t\middle|\Turn^{(1:k)}_{<t},a^{(1)}_t,\dots,a^{(k)}_t\right)\\&:=f_{\bar{\nu}}\left(\Turn^{(1:k)}_{<t},\left(a^{(1)}_t,\dots,a^{(k)}_t\right)\right)\left(e^{(1)}_t,\dots,e^{(k)}_t\right)\,.
    \end{align*}

    A policy for the $i$-th agent of ${\bar{\nu}}$ is a policy-like chronological conditional semimeasure $\pi^{(i)}:\TurnSet^{(i)*}\to\Delta'\mathcal{A}^{(i)}$.

    If $\Turn^{(1:k)}_{<t}\in\left(\mathcal{A}^{(1:k)}\times \mathcal{E}^{(1:k)}\right)^{t-1}$ describes the complete interaction of the $k$ agents with the $k$-multi-agent environment for $t-1$ steps, then we write $\Turn^{(i)}_{<t}\in\left(\mathcal{A}^{(i)}\times \mathcal{E}^{(i)}\right)^{t-1}$ to denote the corresponding (subjective) history from the perspective of the $i$-th agent.
\end{definition}

The interactions between a $k$-multi-agent-environment and policies of the $k$ agents can be defined in a natural way which is analogous to how we defined it in the single-agent setting:

\begin{definition}
      Let ${\bar{\nu}}$ be a $k$-multi-agent environment with action spaces $\mathcal{A}^{(1)},\ldots,\mathcal{A}^{(k)}$, observation spaces $\mathcal{O}^{(1)},\ldots,\mathcal{O}^{(k)}$ and reward spaces $\mathcal{R}^{(1)},\ldots,\mathcal{R}^{(k)}$, and let $\pi^{(1)},\ldots,\pi^{(k)}$ be policies for the $k$ agents of ${\bar{\nu}}$, respectively. We define the semimeasure ${\bar{\nu}}^{\pi^{(1)},\ldots,\pi^{(k)}}$ on $\TurnSet^{(1:k)*}$ recursively as follows:
      $${\bar{\nu}}^{\pi^{(1)},\ldots,\pi^{(k)}}(\varepsilon)=1\,,$$
      and
      \begin{align*}
            {\bar{\nu}}^{\pi^{(1)},\ldots,\pi^{(k)}}&\left(\Turn_{<t}^{(1:k)}a_t^{(1:k)}e_t^{(1:k)}\right)\\
            &={\bar{\nu}}^{\pi^{(1)},\ldots,\pi^{(k)}}\left(\Turn_{<t}^{(1:k)}\right)\times\left(\prod_{1\leq i\leq k}\pi^{(i)}\left(a_t^{(i)}\middle|\Turn_{<t}^{(i)}\right)\right)\times{\bar{\nu}}\left(e^{(1:k)}_t\middle|\Turn^{(1:k)}_{<t},a^{(1:k)}_t\right)\\
            &=\prod_{1\leq t'\leq t}\left(\prod_{1\leq i\leq k}\pi^{(i)}\left(a^{(i)}_{t'}\middle|\Turn_{<t'}^{(i)}\right)\right)\times{\bar{\nu}}\left(e^{(1:k)}_{t'}\middle|\Turn^{(1:k)}_{<t'},a^{(1:k)}_{t'}\right)\,,
      \end{align*}
      for all $\Turn^{(1:k)}_{<t}\in\left(\mathcal{A}^{(1:k)}\times \mathcal{E}^{(1:k)}\right)^{t-1}$, all $e^{(1:k)}_t=\left(e^{(1)}_t,\ldots, e^{(k)}_t\right)\in\mathcal{E}^{(1:k)}$, and all $a^{(1:k)}_t=\left(a^{(1)}_t,\ldots, a^{(k)}_t\right)\in\mathcal{A}^{(1:k)}$.
\end{definition}

If we fix the policies of a subset of agents in a multi-agent environment, we get a multi-agent environment with a smaller number of agents. This can be seen as one possible way of formalizing the concept of "embedding an agent in an environment":

\begin{definition}
    \label{def:embedding-some-agents-in-multi-agents-environemnt}
     Let ${\bar{\nu}}$ be a $k$-multi-agent environment with action spaces $\mathcal{A}^{(1)},\ldots,\mathcal{A}^{(k)}$, observation spaces $\mathcal{O}^{(1)},\ldots,\mathcal{O}^{(k)}$ and reward spaces $\mathcal{R}^{(1)},\ldots,\mathcal{R}^{(k)}$. Let $1\leq\ell<k$ and let $\pi^{(i_1)},\pi^{(i_2)},\ldots,\pi^{(i_\ell)}$ be policies for the $i_1$-th, the $i_2$-th, \ldots, and the $i_\ell$-th agent of ${\bar{\nu}}$, respectively. We define the $(k-\ell)$-multi-agent environment ${\bar{\nu}}^{\pi^{(i_1)},\ldots,\pi^{(i_\ell)}}_{(i_1,\ldots,i_\ell)}$ as follows:
     \begin{itemize}
         \item The action spaces of ${\bar{\nu}}^{\pi^{(i_1)},\ldots,\pi^{(i_\ell)}}_{(i_1,\ldots,i_\ell)}$ are $\tilde{\mathcal{A}}^{(1)}=\mathcal{A}^{(\tilde{i}_1)},\ldots,\tilde{\mathcal{A}}^{(k-\ell)}=\mathcal{A}^{(\tilde{i}_{k-\ell})}$, where $\tilde{i}_j$ is the $j$-th element of $\{1,\ldots,k\}\setminus\{i_1,\ldots,i_\ell\}$. The observation spaces $\tilde{\mathcal{O}}^{(1)},\ldots,\tilde{\mathcal{O}}^{(k-\ell)}$ and the reward spaces $\tilde{\mathcal{R}}^{(1)},\ldots,\tilde{\mathcal{R}}^{(k-\ell)}$ are defined similarly.
         \item For all $\tilde{\Turn}^{(1:k-\ell)}_{<t}\in\left(\tilde{\mathcal{A}}^{(1:k-\ell)}\times \tilde{\mathcal{E}}^{(1:k-\ell)}\right)^{t-1}$, all $\tilde{a}^{(1:k-\ell)}_t\in\tilde{\mathcal{A}}^{(1:k-\ell)}$, and all $\tilde{e}^{(1:k-\ell)}_t\in\tilde{\mathcal{E}}^{(1:k-\ell)}$, we have
      \begin{align*}
            {\bar{\nu}}^{\pi^{(i_1)},\ldots,\pi^{(i_\ell)}}_{(i_1,\ldots,i_\ell)}&\left(\tilde{e}_t^{(1:k-\ell)}\middle| \tilde{\Turn}_{<t}^{(1:k-\ell)}\tilde{a}_t^{(1:k-\ell)}\right)\\
            &=\frac{\displaystyle\sum_{\substack{a_{1:t}^{(i_1)}\in\left(\mathcal{A}^{(i_1)}\right)^t,\;e_{1:t}^{(i_1)}\in\left(\mathcal{E}^{(i_1)}\right)^t\,,\\
            \vdots\\
            a_{1:t}^{(i_\ell)}\in\left(\mathcal{A}^{(i_\ell)}\right)^t,\;e_{1:t}^{(i_\ell)}\in\left(\mathcal{E}^{(i_\ell)}\right)^t
            }} \prod_{1\leq t'\leq t}\left(\prod_{1\leq j\leq \ell}\pi^{(i_j)}\left(a^{(i_j)}_{t'}\middle|\Turn_{<t'}^{(i_j)}\right)\right)\times{\bar{\nu}}\left(e^{(1:k)}_{t'}\middle|\Turn^{(1:k)}_{<t'},a^{(1:k)}_{t'}\right)}{\displaystyle\sum_{\substack{a_{1:t}^{(i_1)}\in\left(\mathcal{A}^{(i_1)}\right)^t,\;e_{1:t}^{(i_1)}\in\left(\mathcal{E}^{(i_1)}\right)^t\,,\\
            \vdots\\
            a_{1:t}^{(i_\ell)}\in\left(\mathcal{A}^{(i_\ell)}\right)^t,\;e_{1:t}^{(i_\ell)}\in\left(\mathcal{E}^{(i_\ell)}\right)^t,\\
            e_t^{(\tilde{i}_1)}\in\calE^{(\tilde{i}_1)},\ldots,e_t^{(\tilde{i}_{k-\ell})}\in\calE^{(\tilde{i}_{k-\ell})}
            }} \prod_{1\leq t'\leq t}\left(\prod_{1\leq j\leq \ell}\pi^{(i_j)}\left(a^{(i_j)}_{t'}\middle|\Turn_{<t'}^{(i_j)}\right)\right)\times{\bar{\nu}}\left(e^{(1:k)}_{t'}\middle|\Turn^{(1:k)}_{<t'},a^{(1:k)}_{t'}\right)}\,.
      \end{align*}
      where in the above we wrote $e^{(1:k)}_{t'}$ to denote $e^{(1:k)}_t=(e^{(1)}_{t'},\ldots,e^{(k)}_{t'})\in\mathcal{E}^{(1:k)}$ obtained by extending the elements $e^{(i_1)}_{t'},\ldots,e^{(i_\ell)}_{t'}$ determined by the summation sign as follows: If $\tilde{i}_j$ is the $j$-th element of $\{1,\ldots,k\}\setminus\{i_1,\ldots,i_\ell\}$, then we let $e^{(\tilde{i}_j)}_{t'}=\tilde{e}^{(j)}_{t'}$, which is the $j$-th entry in $\tilde{e}^{(1:k-\ell)}_{t'}=\left(\tilde{e}^{(1)}_{t'},\ldots,\tilde{e}^{(k-\ell)}_{t'}\right)\in\tilde{\mathcal{E}}^{(1:k-\ell)}$. The definition of $a^{(1:k)}_{t'}$ and $\Turn_{<t'}^{(1:k)}$ is done similarly. The aforementioned description of $e^{(1:k)}_{t'}$ has one exception when $t'=t$ in the denominator as we already have all the values of $e^{(1)}_{t},\ldots,e^{(k)}_{t}$ given in the summation, hence we do not use the values of $\tilde{e}^{(1)}_{t},\ldots,\tilde{e}^{(k-\ell)}_{t}$ to get $(e^{(\tilde{i}_j)})_{1\leq j\leq k-\ell}$ but rather use the values given from the summation directly.
     \end{itemize}
\end{definition}

\section{A few desiderata for a Bayesian theory of embedded universal intelligence}
\label{app:desiderata-embedded-uai-theory}
This section formulates a set of desiderata which, we believe, a good Bayesian theory of embedded universal intelligence should satisfy. The desiderata that we specify here capture the issues of AIXI and JAIXI that we informally described in Appendix \ref{app:preliminaries-jaixi}, and which made them fail in embedded settings.

It is worth noting that the desiderata that we specify in this section are not meant to capture every aspect of embedded agency. There are several aspects of embedded agency which were mentioned in \cite{demski2019embedded} and which we do not capture in the desiderata that we specify here (e.g., we do not consider here the possibility that the policy of an agent may be self-modified by the agent, or modified by the environment). Our aim here is mainly to mathematically and rigorously formalize a set of goals which capture some of the aspects described in \cite{demski2019embedded} and for which AIXI and JAIXI fail. This helps motivate our theory of Embedded Universal Predictive Intelligence (MUPI), as well as provide intuition and clarify the motivation behind the mathematical concepts in our MUPI theory.

We start by formulating one set of desiderata that captures a few embeddedness aspects from a purely single-agent perspective:\footnote{We later formulate a more general set of desiderata which capture some embeddedness aspects from a multi-agent perspective. The more general set of desiderata will be able to capture the notion that a single-agent environment, as seen from the perspective of one agent, may contain/embed other agents which may be implementing similar policies to the policy of the considered agent.}

\begin{problem}
    \label{prob:embedded_hypothesis_classes_informal} (Informal) We aim to find three classes $\calM_{\mathrm{pol}},\calM_{\mathrm{env}},\calMuni$ and a universal prior $w\in\Delta'\calMuni$ such that the following desiderata are satisfied:
    \begin{enumerate}
        \item {\bf Generality of the considered policies:} $\calM_{\mathrm{pol}}$ is a countable class of policies (i.e., policy-like chronological conditional semimeasures) which contains all computable policies.
        \item {\bf Generality of the considered environments:} $\calM_{\mathrm{env}}$ is a countable class of environments (i.e., environment-like chronological conditional semimeasures) which contains all computable environments.
        \item {\bf Generality of the considered universes:} $\calMuni$ is a countable class of semimeasures on $\TurnSet^*$ which contains all computable semimeasures on $\TurnSet^*$. The class $\calMuni$ can be thought of as a class of (distributions of histories of) interactions between an agent and an environment. We will use the term "universe" to refer to a semimeasure on $\TurnSet^*$ which can be seen as jointly describing an agent and an environment.
        \item {\bf The interaction between a considered policy and a considered environment yields a universe that is considered in $\calMuni$:} For every $\pi\in \calM_{\mathrm{pol}}$ and every $\nu\in \calM_{\mathrm{env}}$, we have $\nu^\pi\in\calMuni$.
        \item {\bf Embedded Bayesian agents with respect to the $\calMuni$ hypothesis class are part of the considered policies:}  There exists a policy $\pi^*\in\calM_{\mathrm{pol}}$ which performs "a form of universal embedded Bayesian optimization" as detailed in \cref{sec:embedded-br,sec:k-step-planning,sec:eq-behavior-k-step}
        %\footnote{This is essentially the informal part of Problem \ref{prob:embedded_hypothesis_classes_informal}.} 
        with respect to the mixture universe $$\Mmu:=\sum_{\lambda\in\calMuni}w(\lambda) \lambda\,.$$ %We will come back later to what exactly "universal embedded Bayesian optimization" means.
    \end{enumerate}
\end{problem}

Notice how the desiderata in the above problem captures the issue that we mentioned about JAIXI in \ref{app:preliminaries-jaixi}: JAIXI is not lower semicomputable, and hence if we let it interact with a computable environment, we may get a universe which is not lower semicomputable, i.e., we get something which is outside the hypothesis class of JAIXI. Putting this in the language of the above desiderata: If $\calM_{\mathrm{pol}}$ contains JAIXI and $\calM_{\mathrm{env}}$ contains all lower semicomputable semimeasures, then $\calMuni$ must contain universes which are not lower semicomputable, and hence the hypothesis class of JAIXI would not cover all the universes in $\calMuni$, i.e., JAIXI would not be a valid solution of the last point of Problem \ref{prob:embedded_hypothesis_classes_informal}.

Now given a solution to Problem \ref{prob:embedded_hypothesis_classes_informal}, if we make the policy $\pi^*$ interact with an environment in $\calM_{\mathrm{env}}$, then the fourth point implies that we get a universe which is in the hypothesis class $\calMuni$ of the "embedded Bayesian" policy $\pi^*$. This is in contrast with JAIXI which does not necessarily give rise to a lower semicomputable interaction if we make it interact with a computable environment.

Now we are ready to state our desiderata for a good theory of embedded universal intelligence, from a multi-agent perspective:

\begin{problem}
    \label{prob:embedded_hypothesis_classes_multi_agent_informal} (Informal) We aim to find:
    \begin{itemize}
        \item for every choice of finite sets $\mathcal{A}, \mathcal{O}$ and $\mathcal{R}$ with complete prefix-free encodings thereof, a countable class $\calM_{\mathrm{pol}}^{\mathcal{A},\mathcal{O},\mathcal{R}}$ of policies with action space $\mathcal{A}$, observation space $\mathcal{O}$ and reward space $\mathcal{R}$, and
        \item for every $k\geq 1$ and every choice of finite sets $\mathcal{A}^{(1)},\ldots,\mathcal{A}^{(k)}, \mathcal{O}^{(1)},\ldots,\mathcal{O}^{(k)}$, and $\mathcal{R}^{(1)},\ldots,\mathcal{R}^{(k)}$ with complete prefix-free encodings thereof, a countable class $\calM_{k,\mathrm{env}}^{\mathcal{A}^{(1:k)},\mathcal{O}^{(1:k)},\mathcal{R}^{(1:k)}}$ of $k$-multi-agent environments with action spaces $\mathcal{A}^{(1)},\ldots,\mathcal{A}^{(k)}$, observation spaces $ \mathcal{O}^{(1)},\ldots,\mathcal{O}^{(k)}$, and reward spaces $\mathcal{R}^{(1)},\ldots,\mathcal{R}^{(k)}$, and
        \item a countable class $\calMuni$ of semimeasures on $\mathcal{B}^*$ (which we will refer to as universes), together with a universal prior $w\in\Delta'\calMuni$, which can be interpreted as semimeasures over $\TurnSet^*$ and $\TurnSet^{(1:k)*}$ for suitably chosen complete prefix-free encodings of $\mathcal{A}, \mathcal{O}$, $\mathcal{R}$, $\mathcal{A}^{(1)},\ldots,\mathcal{A}^{(k)}, \mathcal{O}^{(1)},\ldots,\mathcal{O}^{(k)}$, and $\mathcal{R}^{(1)},\ldots,\mathcal{R}^{(k)}$
    \end{itemize}
    such that the following desiderata are satisfied:
    \begin{enumerate}
        \item {\bf Generality of considered policies:} For all finite sets $\mathcal{A}, \mathcal{O}$ and $\mathcal{R}$, the class
        $\calM_{\mathrm{pol}}^{\mathcal{A},\mathcal{O},\mathcal{R}}$ contains all computable policies of action space $\mathcal{A}$, observation space $\mathcal{O}$ and reward space $\mathcal{R}$.
        \item {\bf Generality of considered multi-agent environments:} For all $k>1$, and all finite sets $\mathcal{A}^{(1)},\ldots,\mathcal{A}^{(k)}, \mathcal{O}^{(1)},\ldots,\mathcal{O}^{(k)}$, and $\mathcal{R}^{(1)},\ldots,\mathcal{R}^{(k)}$, the class $\calM_{k,\mathrm{env}}^{\mathcal{A}^{(1:k)},\mathcal{O}^{(1:k)},\mathcal{R}^{(1:k)}}$ contains all computable $k$-multi-agent environments with actions spaces $\mathcal{A}^{(1)},\ldots,\mathcal{A}^{(k)}$, observations spaces $ \mathcal{O}^{(1)},\ldots,\mathcal{O}^{(k)}$, and reward spaces $\mathcal{R}^{(1)},\ldots,\mathcal{R}^{(k)}$.
        \item {\bf Generality of considered universes:} $\calMuni$ contains all computable semimeasures on $\mathcal{B}^*$.
        \item {\bf Interactions between considered policies and considered single-agent environments yield universes that are considered in $\calMuni$:} For all finite sets $\mathcal{A},\mathcal{O}$ and $\mathcal{R}$ , every $\pi\in \calM_{\mathrm{pol}}^{\mathcal{A},\mathcal{O},\mathcal{R}}$ and every single-agent environment $\nu\in \calM_{1,\mathrm{env}}^{\mathcal{A},\mathcal{O},\mathcal{R}}$, we have $\nu^\pi\in\calMuni$ when interpreting the bitstrings as histories $\HistA \in \TurnSet$ through complete prefix-free encodings of $\mathcal{A},\mathcal{O}$ and $\mathcal{R}$.
        \item {\bf Embedding a set of considered policies inside a considered multi-agent environment yields a multi-agent environment (with less agents) which is also considered:} For every $k>1$, and every finite sets $\mathcal{A}^{(1)},\ldots,\mathcal{A}^{(k)}, \mathcal{O}^{(1)},\ldots,\mathcal{O}^{(k)}$, and $\mathcal{R}^{(1)},\ldots,\mathcal{R}^{(k)}$, if
        \begin{itemize}
            \item ${\bar{\nu}}\in\calM_{k,\mathrm{env}}^{\mathcal{A}^{(1:k)},\mathcal{O}^{(1:k)},\mathcal{R}^{(1:k)}}$,
            \item $\{i_1,\ldots,i_\ell\}\subset\{1,\ldots,k\}$ with $\ell<k$, and 
            \item $\pi^{(i_j)}\in\calM_{\mathrm{pol}}^{\mathcal{A}^{(i_j)},\mathcal{O}^{(i_j)},\mathcal{R}^{(i_j)}}$ for all $1\leq j\leq \ell$,
        \end{itemize} 
        then ${\bar{\nu}}^{\pi^{(i_1)},\ldots,\pi^{(i_\ell)}}_{(i_1,\ldots,i_\ell)}\in\calM_{k-\ell,\mathrm{env}}^{\tilde{\mathcal{A}}^{(1:k-\ell)},\tilde{\mathcal{O}}^{(1:k-\ell)},\tilde{\mathcal{R}}^{(1:k-\ell)}}$  (cf. \cref{def:embedding-some-agents-in-multi-agents-environemnt}), where $\tilde{\mathcal{A}}^{(j)}:=\mathcal{A}^{(\tilde{i}_j)},\tilde{\mathcal{O}}^{(j)}:=\mathcal{O}^{(\tilde{i}_j)},\tilde{\mathcal{R}}^{(j)}:=\mathcal{R}^{(\tilde{i}_j)}$ with $\tilde{i}_j$ being the $j$-th element of $\{1,\ldots,k\}\setminus\{i_1,\ldots,i_\ell\}$.
        \item {\bf Embedded Bayesian agents with respect to the $\calMuni$ hypothesis class are part of the considered policies:} For every finite sets $\mathcal{A},\mathcal{O},\mathcal{R}$, there exists a policy $\pi^*_{\mathcal{A},\mathcal{O},\mathcal{R}}\in\calM_{\mathrm{pol}}$ which performs "a form of universal embedded Bayesian optimization" as detailed in \cref{sec:embedded-br,sec:k-step-planning,sec:eq-behavior-k-step} with respect to a universal mixture model $$\Mmu:=\sum_{\Mvu \in \calMuni}w(\Mvu)\Mvu,$$ using a lower-semicomputable semiprior $w$, and with respect to the spaces $\mathcal{A},\mathcal{O}$ and $\mathcal{R}$ and a complete prefix-free encoding thereof. 
    \end{enumerate}
\end{problem}

A few comments are in order:
\begin{itemize}
    \item Given a solution to the above problem, if we make the policy $\pi^*_{\mathcal{A},\mathcal{O},\mathcal{R}}$ interact with an environment in $\calM_{\mathrm{1,env}}^{\mathcal{A},\mathcal{O},\mathcal{R}}$, then we get a universe which is in the hypothesis class $\calMuni$ of the "embedded Bayesian" policy $\pi^*_{\mathcal{A},\mathcal{O},\mathcal{R}}$.
    \item It is possible for a single-agent environment $\nu\in \calM_{\mathrm{1,env}}^{\mathcal{A},\mathcal{O},\mathcal{R}}$ to be embedding other agents in the sense that there may be a $k$-multi-agent environment ${\bar{\nu}}$ in such a way that $\nu$ can be obtained from ${\bar{\nu}}$ by providing policies to $k-1$ agents of ${\bar{\nu}}$.
    \item The formalism in Problem \ref{prob:embedded_hypothesis_classes_multi_agent_informal} allows to capture notions of similarity between agents. For example, the psychological twin prisoner's dilemma may be captured as follows: Take ${\bar{\nu}}$ as the 2-multi-agent environment given by the prisoner's dilemma\footnote{We can assume, e.g., that the game stops after one round.}, and for every policy $\pi$ let ${\bar{\nu}}_{(2)}^{\pi}$ be the single-agent environment from the perspective of the first agent of ${\bar{\nu}}$ which is induced by fixing the policy of the second agent to $\pi$. By choosing a universal prior which has almost all its weight on the universes of the form $\left({\bar{\nu}}_{(2)}^{\pi}\right)^\pi$, we get a prior that gives a very high credence to the hypothesis that both agents implement the same policy. This captures the essence of the psychological twin variant of the prisoner's dilemma.
\end{itemize}

\subsection{The MUPI theory satisfies our desiderata}

\label{app:mupi-theory-satisfies-desiderata}

By inspecting Problems \ref{prob:embedded_hypothesis_classes_informal} and \ref{prob:embedded_hypothesis_classes_multi_agent_informal}, we can see that we essentially have all the ingredients:
\begin{itemize}
    \item We can take $\calMuni$ to be the class of all universes which are equipped with a $w$-RUI-oracle $\tau$, where $w\in
    \Delta'\calMrapom$ is some lower semicomputable prior on the class of restricted abstract probabilistic oracle machines $\calMrapom$. Note that $\calMuni$ is countable and contains all lower semicomputable semimeasures on $\mathcal{B}^*$.
    \item For finite sets $\mathcal{A}, \mathcal{O}$ and $\mathcal{R}$ with prefix-free encodings thereof, we can take the class of policies $\calM_{\mathrm{pol}}^{\mathcal{A},\mathcal{O},\mathcal{R}}$ as those that can be implemented by a $\tau$-rPOM $P^\tau$ which takes as input the prefix-free encoding $\langle\Hist\rangle$ of a history $\Hist\in\TurnSet^*$ and produces the prefix-free encoding $\langle a\rangle$ of an action $a\in\calA$.
    \item For $k\geq 1$, and for finite sets $\mathcal{A}^{(1)},\ldots,\mathcal{A}^{(k)}, \mathcal{O}^{(1)},\ldots,\mathcal{O}^{(k)}$, and $\mathcal{R}^{(1)},\ldots,\mathcal{R}^{(k)}$ with prefix-free encodings thereof, we can take the class of all $k$-multi-agent environments $\calM_{k,\mathrm{env}}^{\mathcal{A}^{(1:k)},\mathcal{O}^{(1:k)},\mathcal{R}^{(1:k)}}$ to be those that can be implemented by a $\tau$-rPOM $E^\tau$ which takes as input the prefix-free encoding $\langle\HistM^{(1)},\ldots,\HistM^{(k)},a^{(1)}_t,\ldots,a^{(k)}_t\rangle$ of histories $(\HistM^{(1)},\ldots,\HistM^{(k)})\in\TurnSet^{(1:k)*}$ and actions $(a^{(1)}_t,\ldots,a_t^{(k)})\in\calA^{(1:k)}$ and produces the prefix-free encoding $\langle e^{(1)}_t,\ldots,e_t^{(k)}\rangle$ of percepts $(e^{(1)}_t,\ldots,e_t^{(k)})\in\calE^{(1:k)}$.
    \item As shown by \cref{thm:k_mupi_implementation}, the set of considered policies contains approximate embedded Bayesian agents: All $(k_t,\epsilon_t)_t$-E-AIXI$^{\mathrm{RUI}}$ agents are members of the policy classes defined above. Furthermore, the approximations of these agents can be made arbitrarily good.
    \item The interaction between a considered policy and a considered single agent environment (fourth point in Problem \ref{prob:embedded_hypothesis_classes_multi_agent_informal}) does indeed yield a universe in $\calMuni$: One just needs to take the machines $P$ and $E$ and build a machine $M$ which makes $P$ and $E$ interact by alternately exchanging inputs and outputs appropriately. We omit the technical details here.
    \begin{itemize}
        \item The fifth point of Problem \ref{prob:embedded_hypothesis_classes_multi_agent_informal} regarding the reduction of a $k$-multi-agent environment into a $(k-\ell)$-multi-agent environment by fixing $\ell$ policies for $\ell$ agents, can be shown similarly.
    \end{itemize}
\end{itemize}

We can also solve \ref{prob:embedded_hypothesis_classes_informal} and \ref{prob:embedded_hypothesis_classes_multi_agent_informal} using the reflective oracle framework by considering the classes of all policies (resp., environment, $k$-multi-agent environments, universes) which are implementable using POMs with access to a reflective oracle $\tau$. \cref{thm:eaixi-ro} shows that the universal embedded Bayesian agents E-AIXI$^{\mathrm{RO}}$ and $k$-E-AIXI$^{\mathrm{RO}}$ for $k>0$ are all implementable as $\tau$-POMs.

\end{document}